\documentclass[12pt]{article}
\usepackage{xr}
\usepackage{microtype}
\usepackage{graphicx}
\usepackage{subfigure}
\usepackage{rotating}
\usepackage{setspace}
\usepackage{booktabs} 
\usepackage{algorithm}
\usepackage{authblk}
\usepackage{setspace} 
\usepackage{algpseudocode}

\usepackage{enumitem}

\usepackage{hyperref}

\usepackage{amsmath}
\usepackage{amssymb}
\usepackage{mathtools}
\usepackage{amsthm}
\usepackage[capitalize,noabbrev]{cleveref}
\usepackage[english]{babel}
\makeatletter
\theoremstyle{definition}
    \ifx\thechapter\undefined
      \newtheorem{defn}{\protect\definitionname}
    \else
      \newtheorem{defn}{\protect\definitionname}[chapter]
    \fi
\theoremstyle{plain}
    \ifx\thechapter\undefined
      \newtheorem{lem}{\protect\lemmaname}
    \else
      \newtheorem{lem}{\protect\lemmaname}[chapter]
    \fi
\theoremstyle{plain}
    \ifx\thechapter\undefined
      \newtheorem{thm}{\protect\theoremname}
    \else
      \newtheorem{thm}{\protect\theoremname}[chapter]
    \fi
\theoremstyle{remark}
    \ifx\thechapter\undefined
      \newtheorem{rem}{\protect\remarkname}
    \else
      \newtheorem{rem}{\protect\remarkname}[chapter]
    \fi
\theoremstyle{plain}
    \ifx\thechapter\undefined
  \newtheorem{cor}{\protect\corollaryname}
\else
      \newtheorem{cor}{\protect\corollaryname}[chapter]
    \fi

\makeatother

\usepackage{babel}
\usepackage{natbib}

\providecommand{\definitionname}{Definition}
\providecommand{\lemmaname}{Lemma}
\providecommand{\theoremname}{Theorem}
\providecommand{\corollaryname}{Corollary}
\providecommand{\remarkname}{Remark}
\usepackage{pifont}

\algnewcommand{\Input}[1]{%
  \State \noindent \textbf{Input:}
  \Statex \hspace*{\algorithmicindent}\parbox[t]{.8\linewidth}{\raggedright #1}
}

\algnewcommand{\Output}[1]{%
  \State \noindent \textbf{Output:}
  \Statex \hspace*{\algorithmicindent}\parbox[t]{.8\linewidth}{\raggedright #1}
}

\DeclareMathOperator*{\argmax}{arg\,max}

\theoremstyle{plain}

\theoremstyle{definition}

\theoremstyle{remark}

\usepackage[textsize=tiny]{todonotes}

\begin{document}
\def\spacingset#1{\renewcommand{\baselinestretch}%
{#1}\small\normalsize} \spacingset{1}

%%%%%%%%%%%%%%%%%%%%%%%%%%%%%%%%%%%%%%%%%%%%%%%%%%%%%%%%%%%%%%%%%%%%%%%%%%%%%%

%%%%%%%%%%%%%%%%%%%%%%%%%%%%%%%%%%%%%%%%%%%%%%%%%%%%%%%%%%%%%%%%%%%%%%%%%%%%%%
  \title{\bf Deep Clustering Evaluation: How to Validate Internal Clustering Validation Measures}
  \author[1]{Zeya Wang \thanks{Correspondence: zeya.wang@uky.edu}}
  \author[1]{Chenglong Ye \thanks{Correspondence: chenglong.ye@uky.edu}}
 \date{\vspace{-5ex}}

  \affil[1]{Dr. Bing Zhang Department of Statistics, University of Kentucky}

  \maketitle

\bigskip
\begin{abstract}
Deep clustering partitions complex high-dimensional data using deep neural networks for clustering. It involves projecting data into lower-dimensional embeddings before partitioning, which embarks unique evaluation challenges. Traditional clustering validation measures, designed for low-dimensional spaces, are problematic for deep clustering for two reasons: 1) the curse of dimensionality when applied to the high-dimensional input data, and 2) unreliable comparison of clustering results when applied to embedded data from different embedding spaces, owing to variations in training procedures and model parameter settings. This paper addresses these unresolved and often overlooked challenges in evaluating clustering within deep learning. We propose a systematic evaluation framework for internal clustering validation measures that: (1) theoretically establishes why traditional measures are ineffective when applied to input data or across disparate embedding spaces paired with partitioning outcomes; (2) identifies embedding spaces that endorse reliable evaluations by detecting groups with high agreement in ranking partitioning outcomes; and (3) develops a stable and robust scoring scheme by weighting index values computed across these identified embedding spaces. Experiments show that this new framework aligns better with external measures, effectively reducing the misguidance from the improper use of internal validation measures in deep clustering evaluation.
\end{abstract}

\noindent%

\vfill

%%%%%%%%%%%%%%%%%%%%%%%%%%%%%%%%%%%%%%%%%%%%%%%%%%%%%%%%%%%%

\section{Introduction}\label{sec:intro}
%%%%%%%%%%%%%%%%%%%%%%%%%%%%%%%%%%%%%%%%%%%%%%%%%%%%%%%%%%%%
\vspace{-4mm}
Clustering, a core task in unsupervised learning, is essential in various applications, from image analysis to data segmentation. With advances in artificial intelligence (AI), deep networks have excelled in label prediction and feature extraction from unlabeled data. This progress has spawned deep clustering methods \citep{yang2016joint,ghasedi2017deep,caron2018deep}, which enhance the scalability of traditional clustering techniques to high-dimensional data by projecting data into a lower-dimensional feature space (or named the {\it embedding space}) and performing partitioning in this more manageable space. 

The primary learning objective of most deep clustering methods involves minimizing a clustering loss through the generated embedded data. Two main categories of deep clustering methods are: autoencoder-based and clustering deep neural network (CDNN)-based approaches \citep{min2018survey}. An autoencoder is a neural network featuring an encoder that compresses input data into an embedding space and a decoder that reconstructs it. Autoencoder-based methods conduct clustering on the embedded data from the encoder component~\citep{min2018survey}. Within this category, some approaches either train or fine-tune data embeddings from autoencoders and and apply conventional clustering techniques like $k$-means for cluster estimation~\citep{yang2017towards}. Another category of deep clustering methods jointly learns clusters and embeddings without incorporating an autoencoder~\citep{yang2016joint,caron2018deep, wang2021dnb}. CDNN-based methods introduce an end-to-end clustering pipeline, enhancing model scalability by directly minimizing a clustering loss atop a network~\citep{yang2016joint,caron2018deep,wang2021dnb}. These methods necessitate a clustering loss and involve an iterative learning process to jointly update the network and cluster labels. Without a decoder, this efficiency enables their wider applicability to large-scale datasets, as exemplified by \emph{DeepCluster}~\citep{caron2018deep}, which achieves promising performance on ImageNet~\citep{deng2009imagenet}.

Deep clustering methods demonstrate significant success across various fields. For example, in medical imaging, by reducing the cost of labor-intensive dataset labeling, deep clustering methods enable the efficient organization of large-scale medical images \citep{kart2021deepmcat} and help cancer patients stratification for targeted treatments \citep{zhao2020joint}. For clinical and multi-omics data, such as somatic mutations and RNA-seq data, deep clustering is instrumental in identifying cell types and detecting patient subtypes \citep{rohani2020classifying, li2023attention}. Other applications include but are not limited to financial analysis \citep{zhou2022comprehensive, hwang2023identifying}, trajectory analysis \citep{yue2019detect, wang2022deep}, and Internet of Things (IoT) applications \citep{abedin2020towards}.

Evaluating clustering results in machine learning is essential to ensure algorithmic quality and optimal partitioning. This evaluation typically involves two types \citep{liu2010understanding}: \emph{external measures} and \emph{internal measures} (also known as validity index). External measures evaluate the alignment of estimated and true partition labels. Two widely used external measures are normalized mutual information (NMI) and clustering accuracy (ACC). NMI quantifies the similarity between two distinct cluster assignments, while ACC measures the proportion of correctly matched pairs resulting from aligning true class labels with predicted cluster labels (see Section S1 of the Supplementary Material (SM) for definitions). Due to the lack of true labels, external measures have limited applicability in many evaluation settings. In contrast, internal measures are developed to evaluate clustering results based on the intrinsic characteristics of the data without relying on external labels. Examples include the Silhouette score \citep{rousseeuw1987silhouettes}, Calinski-Harabasz index \citep{calinski1974dendrite}, Davies-Bouldin index \citep{davies1979cluster}, Cubic clustering criterion (CCC) \citep{sarle1983sas}, Dunn index \citep{dunn1974well}, Cindex \citep{hubert1976general}, SDbw index \citep{halkidi2001clustering}, and CDbw index \citep{halkidi2008density}. More detailed descriptions of each index are deferred to SM Section S2. The development of various internal measures has emerged as a crucial research focus in clustering evaluation.

\begin{figure}[h]\label{fig1}
\centering
\includegraphics[scale=0.51]{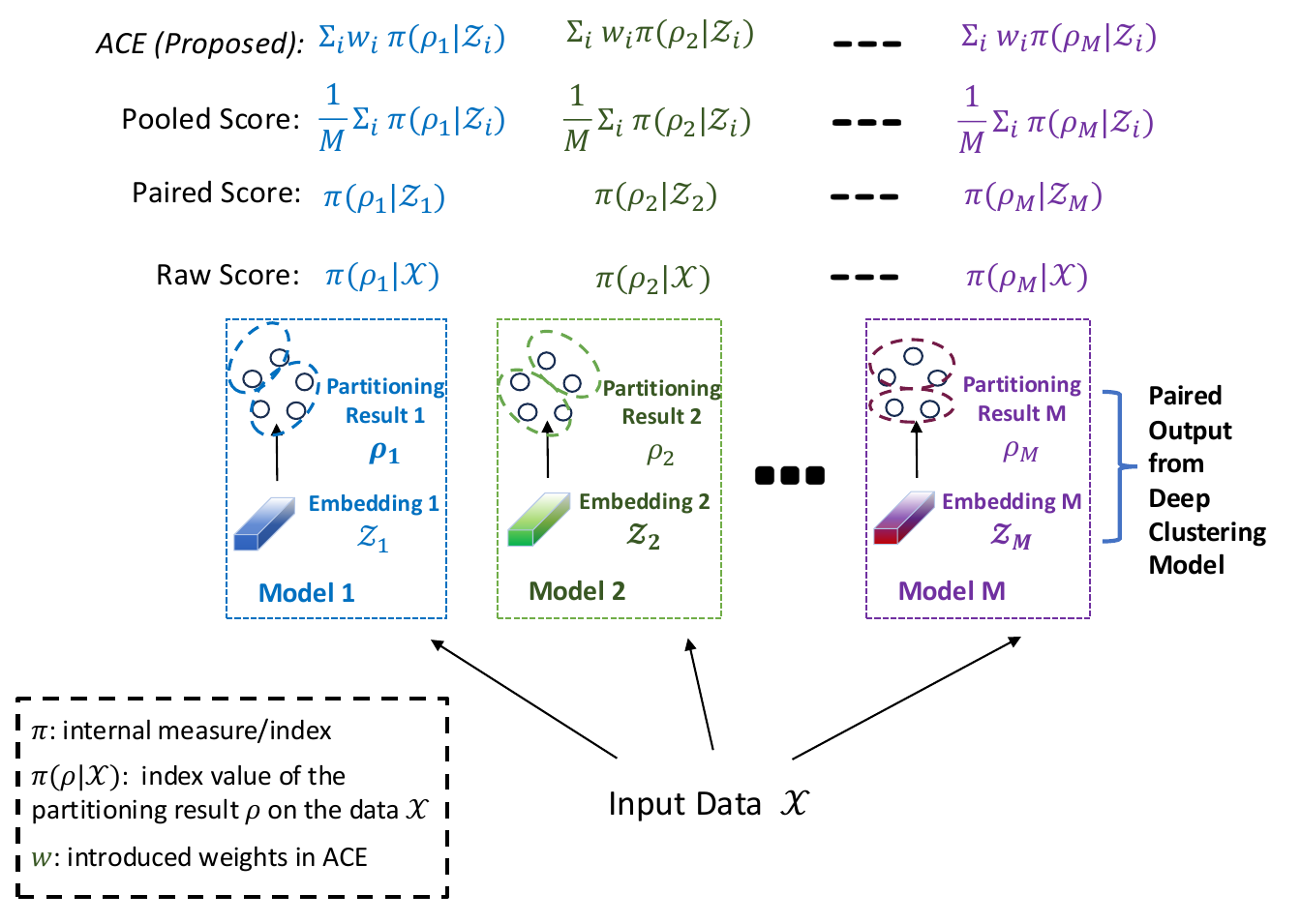}\caption{\footnotesize Different deep clustering evaluation approaches. \emph{Raw score} computes the internal measure based on the input data space; \emph{paired score} calculates the internal measure based on the paired embedding space; \emph{pooled score} average internal measures across embedding spaces, while \emph{ACE} is the proposed Adaptive Clustering Evaluation strategy.}
\end{figure}

Before diving into more details of internal measures, we first denote the input data as $\mathbf{X}$. Deep clustering algorithms generate embedded data $\mathbf{Z}:=f(\mathbf{X})$ for some function $f$ (typically an encoder network) and output a partition, denoted as $\rho$. We use the notation $\pi(\rho|\mathbf{X})$ to represent an internal measure for evaluating the quality of partition  $\rho$ on the raw data $\mathbf{X}$, referred to as the \emph{raw score} in this paper. We refer to $\pi(\rho|\mathbf{Z})$ as the \emph{paired score} to highlight that both $\rho$ and $\mathbf{Z}$ originate from the same execution of the algorithm. Calculating the raw score $\pi(\rho|\mathbf{X})$ of the internal measure often falters due to the curse of dimensionality, where the distance between two points in a high-dimensional space becomes meaningless. Due to the significantly reduced dimensionality of the embedded data, many works in the literature \citep{wang2018deep, wang2021dnb, huang2021dice, ronen2022deepdpm, hadipour2022deep, li2023attention} calculate the paired scores of the deep clustering algorithms as a validation criterion. Figure \ref{fig1} illustrates the difference between the two evaluation approaches using raw and paired scores. Despite the ability of embedded data to mitigate the curse of dimensionality, the application of the paired score to compare different partitioning results is problematic. The embedding space, where the embedded data resides, even differs between various runs (e.g., different hyperparameters, random initializations, data shuffling) of the same deep clustering algorithms. Internal measures are typically designed assuming the evaluated data come from the same feature space. Consequently, variations in embedding spaces compromise the reliability of comparisons using paired scores. For instance, one model creates more separated clusters in the embedding space but with slight boundary errors, while another forms tighter clusters with perfect classification. Despite its less precise partitioning, the first model may score higher on an internal measure like the Silhouette score, which leverages inter-cluster and intra-cluster distances. The questionable reliance on the paired score in much of the existing literature, as noted earlier, highlights the need to properly validate internal measures for deep clustering evaluation. This paper provides a statistical understanding that such comparisons across different embedding spaces may fail due to the embedding space discrepancy. Ideally, we want to compare clustering results based on a single embedding space. However, in real practice, we lack knowledge about which space is ideally separable. To address this, we propose a simple yet effective logic and strategy to guide the usage of internal measures in deep clustering evaluation.

In summary, our contribution is two-fold.  \emph{(1) Theoretical justifications:} We provide formal theoretical proofs demonstrating that employing either high-dimensional raw data or separate embedded data paired with individual partitioning results does not ensure the convergence of the comparative relationship between clustering results and the truth. We introduce the concept of an ``admissible space" (Definition \ref{def2}) to characterize spaces for reliable evaluations. In particular, it is an embedding space in which the internal measure remains consistent with the external measure for most pairwise comparisons of clustering results. Furthermore, we establish theoretical properties for identifying admissible spaces among all embedding spaces obtained with clustering results. These properties provide the foundational framework for developing a strategy to select optimal spaces for evaluation. To our knowledge, we are the first to provide theoretical justifications and explore the significance of feature spaces for deep clustering evaluation. \emph{(2) Evaluation strategy:} 
Building on our theoretical analysis, we present a strategy to identify and select admissible embedding spaces during evaluation. By integrating the calculated internal measure values from these chosen spaces, we enhance the robustness of the evaluation results. Through extensive experiments, we demonstrate the effectiveness and importance of the proposed framework for evaluating deep clustering methods.

The rest of the paper is organized as follows. Section \ref{sec:2} theoretically analyzes why comparing deep clustering results using raw and paired scores is inadequate and offers new insights into identifying optimal embedding spaces. Section \ref{sec:3} introduces our weighted score-based evaluation framework. Sections \ref{sec:sim} and \ref{sec:exp} demonstrate its superior performance. Finally, Section \ref{sec:5} concludes the paper and discusses future research directions.
\vspace{-4mm}
%%%%%%%%%%%%%%%%%%%%%%%%%%%%%%%%%%%%%%%%%%%%%%%%%%%%%%%%%%%%
%%%%%%%%%%%%%%%%%%%%%%%%%%%%%%%%%%%%%%%%%%%%%%%%%%%%%%%%%%%%%%%
\section{Theoretical Analysis for Deep Clustering Evaluation}\label{sec:2}
\vspace{-4mm}
\subsection{Preliminaries and Notations}\label{sec:dcl}
\vspace{-4mm}
 Let $\mathbf{X} = \{\mathbf{x}_1, \cdots \mathbf{x}_n\}$ denote a collection of unlabeled $n$ observations, where $\mathbf{x}_i$ is i.i.d. generated from some unknown distribution $P_X$. We denote the unknown labels of the observations as $Y = \{y_1, \cdots, y_n\}$, where each $y_i \in \{1, \cdots, K\}$ and $K$ represents the true number of groups.  Clustering methods map (up to permutations) the data $\mathbf{X}$ to $\{1,...,K\}$, which we represent as $\phi: \mathbf{X} \rightarrow \{1,...,K\}$. The outcomes of $\phi$ form a partition, denoted as $\rho_\phi:=\{C_1,\cdots,C_K\}$, of the index set $\{1,\cdots,n\}$, where $\hat{y}_i:=\phi(\mathbf{x}_i) = k$ if and only if $i \in C_k$ for any $k=1,...,K$ and $i=1,...,n$. Deep clustering approaches transform the high-dimensional space $\mathcal{X}$ to a significantly lower-dimensional space $\mathcal{Z}$ through an encoder network, denoted $f$, that maps $\mathbf{x}_i\in {\mathcal{X}}$ to $\mathbf{z}_i\in {\mathcal{Z}}$, denoted as $\mathbf{Z}:=f(\mathbf{X})$. The low-dimensional space $\mathcal{Z}$ is referred to in the literature as the embedding space. Subsequently, clustering is performed on the lower-dimensional data $\mathbf{Z} := \{\mathbf{z}_1, \cdots \mathbf{z}_n\}$ to generate labels $Y$. In this context, we employ $g(\cdot): \mathbf{Z} \rightarrow Y$ to represent the mapping from $\mathbf{Z}$ to $Y$. Then, the clustering algorithm $\phi$ can be expressed as a composite function $\phi(\cdot) = g(f(\cdot))$. We denote $\pi(\rho_\phi|\mathbf{X})$ as an internal measure of partition $\rho_\phi$ generated by $\phi$ based on the observed data $\mathbf{X}$. Similarly, $\pi(\rho_\phi|\mathbf{Z})$ denotes the measure based on the embedded data $\mathbf{Z}$. 

 In our subsequent discussion, to underscore the variations arising from the embedding space, we use $\mathcal{Z}$ to denote both the embedding space and the embedded data. Additionally, for notational convenience, we omit $\rho$ in our theoretical analysis and write
$\pi(\phi|\mathcal{Z})$ instead of $\pi(\rho_\phi|\mathbf{Z})$. Unlike $\mathcal{X}$, $\mathcal{Z}$ is affected by $f$ and therefore also by $\phi$. When we refer to $\pi(\phi|\mathcal{Z})$, $\mathcal{Z}$ may or may not be related to $\phi$. For example, we have two deep clustering algorithms $\phi_1(\cdot) = g_1(f_1(\cdot))$ and $\phi_2(\cdot) = g_2(f_2(\cdot))$, and we denote the embedding spaces related to $f_1$ and $f_2$ as $\mathcal{Z}_1$ and $\mathcal{Z}_2$, respectively. When using their respective embedding spaces to evaluate the results, we obtain $\pi(\phi_1|\mathcal{Z}_1)$ and $\pi(\phi_2|\mathcal{Z}_2)$, which is the aforementioned paired score. However, we can also evaluate the result of $\phi_1$ in the embedding space generated by $\phi_2$, which we denote as $\pi(\phi_1|\mathcal{Z}_2)$.
%%%%%%%%%%%%%%%%%%%%%%%%%%%%%%%%%%%%%%%%%%%%%%%%%%%%%%%%%%%%%%%
\vspace{-4mm}
\subsection{Main Results}
\vspace{-4mm}
 We first show that internal measures calculated based on distance functions in high-dimensional spaces yield unreliable evaluations, leading to the common practice of calculating these indices in low-dimensional embedding spaces. Theorem \ref{thm2}  provides a theoretical perspective on why this practice can lead to problematic rankings of partitioning results.

\begin{lem}\label{lemma1}[Distance Meaningless in High Dimensions (Theorem 1 in \citet{10.1007/3-540-49257-7_15})]
Denote $n$ random points $\{X_{1},...,X_{n}\}$ drawn from a common distribution where each
point $X_{i}$ is a $p$-dimensional vector. Let $X_{0}$ be a random
query point that is chosen independently from $\{X_{1},...,X_{n}\}$. For any distance function $d$, define $d_{\max}=\max_{i\in\{1,...,n\}}d(X_{i},X_{0})$
and $d_{\min}=\min_{i\in\{1,...,n\}}d(X_{i},X_{0})$. Given a fixed
$n$, if $\lim_{p\rightarrow\infty} \mathrm{Var}(\frac{d(X_{i},X_{0})}{\mathbf{E}[d(X_{i},X_{0})]}) = 0$, then for any $\epsilon>0$, we have  $$\lim_{p\rightarrow\infty}\mathbb{P}(\text{\ensuremath{\frac{d_{\max}}{d_{\min}}\le1+\epsilon)=1}}.$$
\end{lem}

\begin{rem}
\label{rem:high}
As shown in Lemma \ref{lemma1}, under proper conditions, as the dimensionality increases, the distance between data points converges, making the distance in the input space $\mathcal{X}$ meaningless. Moreover, high-dimensional data, such as images, often have an intrinsic structure that lies in a lower-dimensional latent space. In such cases, distances computed in the input space may poorly approximate similarities defined by the underlying structure in the latent space. Consequently, internal measures (e.g., Silhouette score, Calinski-Harabasz index, and Davies-Bouldin index), which rely on distance or similarity computations, may provide unreliable clustering evaluations in high-dimensional settings.

\end{rem}

\begin{rem}
\label{rem:high:con}
When $\frac{d_{\max}}{d_{\min}}\xrightarrow{p} 1$, many distance-based internal measures will converge to a constant. For example, we have the Silhouette score as $\pi= \frac{1}{K} \sum_{k=1}^K \frac1{N_k} \sum_{i \in C_k} s\left(i\right)$, where $s(i)=\frac{b(i) - a(i)}{\max\{a(i),b(i)\}}=\frac{b(i)/ a(i)-1}{\max\{1,b(i)/a(i)\} }.$ Since $\frac{d_{\min}}{d_{\max}} \le \frac{b(i)}{ a(i)} \le \frac{d_{\max}}{d_{\min}}$ and $\frac{d_{\max}}{d_{\min}}\xrightarrow{p} 1$, we have $\frac{b(i)}{ a(i)}\xrightarrow{p} 1$ and thus $\pi \xrightarrow{p} 0$ as $n$ stays fixed and $p\rightarrow \infty$, given Lemma \ref{lemma1}. Similar reasoning can be extended to other distance-based internal measures. Therefore, the convergence of internal measures hold when $n$ is fixed. To see the case where $n$ also grows, we refer to Theorem 1 in \cite{peng2023interpreting}. According to this theorem, for Euclidean distance $d$ and an i.i.d. uniform distribution $Unif(0,1)$ in each dimension (the results can be extended to more general distributions with unbounded support), we have $\frac{2}{\sqrt{5\pi p}}\le \frac{d_{\max}}{d_{\min}} -1 \le \frac{n(n-1)}{\sqrt{5\pi p}} $ with high probability when $p \rightarrow \infty$. Thus, $\frac{d_{\max}}{d_{\min}}$ converges either when $n$ is fixed or when $\frac{n(n-1)}{\sqrt{p}}\rightarrow 0$. The condition $\frac{n(n-1)}{\sqrt{p}}\rightarrow 0$ is satisfied in the ultra-high-dimensional case where $p$ is at least of order $\exp(n)$, ensuring the convergence of internal measures in this scenario.
\end{rem}

Next, we demonstrate in Theorem \ref{thm2} that comparing different partitioning results based on their paired embedding spaces will fail, even when all the embedding spaces are ideal. Before presenting the theorem, we introduce some definitions.

\begin{defn}[Comparative Quality of Partitions]
Let $\rho^{*}$ denote the unknown true partition. For two partitions, \textit{$\rho_{i}$ is better than $\rho_{j}$}
if $V(\rho^{*},\rho_{i})>V(\rho^{*},\rho_{j})$, where we denote $V$ as an external measure.
\end{defn}
\begin{rem}
Since our goal is to evaluate internal measures, throughout this paper we consider the external measure $V$ as the oracle measure. 
\end{rem}

\begin{defn}[Consistency of Internal Measure Values]\label{def2}
Let $\varrho(\mathbf{X})$
denote the collection of all possible partitions on the given data $\mathbf{X}$. Define \sloppy
$
A:=\{(\phi(\mathbf{X}),\phi^{'}(\mathbf{X}))|\phi(\mathbf{X}),\phi^{'}(\mathbf{X})\in\varrho(\mathbf{X}),(\pi(\phi|{\cal Z})-\pi(\phi^{'}|{\cal Z}))\cdot(V(\rho^{*},\phi)-V(\rho^{*},\phi^{'}))\ge0\}
$
as the set of pairs of partitions whose ranking in terms of internal measure values is consistent with the truth. An internal measure $\pi$ is \textit{$\epsilon_{{\cal Z}}$-consistent
}in space ${\cal Z}$ if $\lim_{n\rightarrow\infty}\mathbb{P}(\mathbf{X}\in A)=\epsilon_{{\cal Z}}$ for some constant $\epsilon_{{\cal Z}}>0$.  In particular, $\pi$ is\textit{ inadmissible} if $\epsilon_{{\cal Z}}<0.5$
and $\pi$ is\textit{ admissible} if $\epsilon_{{\cal Z}}\ge0.5$.
In addition, $\pi$ is \textit{consistent} if $\epsilon_{{\cal Z}}=1$
and $\pi$ is \textit{inconsistent} if $\epsilon_{{\cal Z}}=0$. 
\end{defn}

\begin{rem}
    Note that the constant $\epsilon_{{\cal Z}}$ depends on the space ${\cal Z}$. In turn, we call a space $\mathcal{Z}$ \textit{admissible} for the internal measure $\pi$ if $\epsilon_{{\cal Z}}\ge0.5$
and $\mathcal{Z}$ is \textit{inadmissible} if $\epsilon_{{\cal Z}}<0.5$. In particular, we call $\cal{Z}$ a {\it perfect space} if $\epsilon_{{\cal Z}}=1$.
\end{rem}
\begin{rem}
   The score $\epsilon_{{\cal Z}}$ characterizes the asymptotic agreement between the internal measure evaluated on the space $\mathcal{Z}$ and the external measure. 
\end{rem}

\begin{defn}[{Comparative Quality of Embedding Spaces}]\label{def3} 
A space ${\cal Z}$ is {\it as good as} another space ${\cal Z}^{'}$ if
$\mathbb{P}(\pi(\phi(\mathbf{X})|{\cal Z})-\pi(\phi(\mathbf{X})|{\cal Z}^{'})\ge0)\rightarrow1$ as $n\rightarrow \infty$
for any clustering method $\phi$, which we denote as ${\cal Z}\succeq{\cal Z}^{'}$. 
\end{defn}

The above definition is intended for theoretical analysis, as it is challenging in practice to consistently find cases where the scale of an internal measure based on $\mathcal{Z}$ is consistently larger than that on $\mathcal{Z}'$.

\begin{rem}
It follows from the above definition that ${\cal {\cal Z}}$ is not
as good as ${\cal Z}^{'}$ if $\mathbb{P}(\pi(\phi(\mathbf{X})|{\cal Z})-\pi(\phi(\mathbf{X})|{\cal Z}^{'})\ge0)$
does not converge to 1, which we denote as ${\cal Z}\prec{\cal Z}^{'}$.
Note that ${\cal Z}\prec{\cal Z}^{'}$ and ${\cal Z}^{'}\prec{\cal Z}$
can happen simultaneously. For the purpose of theoretical analysis,
for a pair of spaces $({\cal Z},{\cal Z}^{'})$, we only consider
three cases: ${\cal Z}\succeq{\cal Z}^{'}$, ${\cal Z}\prec{\cal Z}^{'}$
, or the two spaces are the same (denoted as ${\cal Z}={\cal Z}^{'}$). 
\end{rem}
\begin{defn}[Distinguishable Spaces]\label{def4}
Two spaces ${\cal Z},{\cal Z}^{'}$ are {\it distinguishable} if the set
\sloppy $B_{\phi}:=\left\{\max_{\phi^{'}}\left[\pi(\phi^{'}(\mathbf{X})|{\cal Z})-\pi(\phi(\mathbf{X})|{\cal Z})\right]< \pi(\phi(\mathbf{X})|{\cal Z}^{'})-\pi(\phi(\mathbf{X})|{\cal Z})\right\} $
satisfies that $\lim_{n\rightarrow\infty}\mathbb{P}(B_{\phi})=c_{\phi}$
for any given $\phi$, where $0<c_{\phi}\le1$. 
\end{defn}

\begin{rem}
Definition \ref{def4} requires the existence of non-zero probability that the difference between the internal measure values on two distinguishable spaces is larger than the maximum difference between partitions generated by two algorithms on the same space. 
\end{rem}

\begin{thm}[Inconsistency of the Paired Score]\label{thm2}
Consider two distinguishable spaces ${\cal Z}_{1},{\cal Z}_{2}$ and
an internal measure $\pi$ that is consistent in both ${\cal Z}_{1}$
and ${\cal Z}_{2}$. Assume the partition $\phi_{1}(\mathbf{X})$ is
as good as $\phi_{2}(\mathbf{X})$. Then $\mathbb{P}(\pi(\phi_{1}(\mathbf{X})|{\cal Z}_{1})\ge\pi(\phi_{2}(\mathbf{X})|{\cal Z}_{2}))$
does not converge to 1. 
\end{thm}
\begin{rem}
Given an oracle measure, when the outcome generated by method $\phi_1$ is no worse than that of $\phi_2$, Theorem \ref{thm2} suggests that even with a perfect internal measure $\pi$, the commonly used scoring method, the paired score, will not consistently conclude that $\phi_1$ is as good as $\phi_2$ with a probability approaching one.
\end{rem}
By Theorem \ref{thm2}, comparing the goodness between the partitions $\phi:=g(f(\cdot))$ and $\phi^{'}:=g^{'}(f^{'}(\cdot))$ is not equivalent to comparing $\pi(\phi(\cdot) | f(\cdot))$ and $\pi(\phi^{'}(\cdot) | f^{'}(\cdot))$. The optimal practice is to perform comparisons based on a perfect embedding space. However, defining and deriving such a space poses challenges methodologically and practically. Therefore, our focus shifts towards identifying a collection of admissible spaces from the available embedding spaces, specifically those generated with all the clustering results under consideration.

Next, Theorem \ref{thm3} shows that internal measure values on admissible spaces can exhibit high rank correlation, guiding the space grouping step (Step 2) in our proposed evaluation procedure outlined in Algorithm \ref{Algorithm1}. Additionally, Remark \ref{rem:link} informs the ensemble analysis step (Step 3). Together, they formalize the intuitions underlying our evaluation framework.

\begin{thm}\label{thm3} \emph{[Rank Correlation of Admissible Spaces]}
Consider two spaces ${\cal Z}_{1},{\cal Z}_{2}$ and $\pi$ is admissible
in both ${\cal Z}_{1}$ and ${\cal Z}_{2}$. For any pair of partitions
$\phi_{1}(\mathbf{X})$ and $\phi_{2}(\mathbf{X})$, \sloppy 
$\lim_{n\rightarrow\infty}\mathbb{P}\left((\pi(\phi_{1}(\mathbf{X})|{\cal Z}_{1})-\pi(\phi_{2}(\mathbf{X})|{\cal Z}_{1}))\cdot(\pi(\phi_{1}(\mathbf{X})|{\cal Z}_{2})-\pi(\phi_{2}(\mathbf{X})|{\cal Z}_{2}))\ge0\right)\ge0.5.$
\end{thm}
\begin{rem}\label{rem5}
If two spaces satisfy that $\epsilon_{{\cal Z}_{1}}=\epsilon_{{\cal Z}_{2}}$, then Theorem \ref{thm3} still holds. 
\end{rem}
Theorem \ref{thm3} implies that if admissible spaces exist among all
available embedding spaces, one may identify these spaces by finding clusters of
spaces where the internal measure consistently produces highly correlated rankings.
Therefore, Theorem \ref{thm3} provides a strategy for finding admissible spaces by looking
for groups of embedding spaces that show agreement in their rankings of clustering results. However, as pointed out in Remark \ref{rem5}, it is also possible for two
inadmissible spaces to show a high rank correlation in their rankings. To address this, we refer to Definition \ref{def3} and focus on selecting spaces with higher average internal measure values since inadmissible spaces are generally associated with
lower measure values. This inspires us to cluster the embedding spaces based
on rank correlation and then choose the group with the highest average measure
values as the set of admissible spaces. This group serves as our best estimate of the
``perfect space", and we proceed by conducting an ensemble analysis to aggregate
the internal measure values from these selected spaces into a final, more robust aggregated score. More details are in Section \ref{sec:3}.

\begin{cor}\label{cor1} \emph{[Rank Agreement of Admissible Spaces]}
Assume we have $M$ partitioning results to compare: $\phi_{1}(\mathbf{X}),...,\phi_{M}(\mathbf{X})$, with independent pairwise comparisons. Assume $\pi$ is admissible in both ${\cal Z}_{1}$ and ${\cal Z}_{2}$.
Then the scores $\mathbf{a}:=(\pi(\phi_{1}|{\cal Z}_{1}),...,\pi(\phi_{M}|{\cal Z}_{1}))$
and $\mathbf{b}:=(\pi(\phi_{1}|{\cal Z}_{2}),...,\pi(\phi_{M}|{\cal Z}_{2}))$
satisfy \sloppy
$ \lim_{n\rightarrow\infty}\mathbb{P}\left(\textrm{the rankings in } \mathbf{a}\textrm{ and }\mathbf{b}\textrm{ agree}\right)=\left(1-(\epsilon_{{\cal Z}_{1}}+\epsilon_{{\cal Z}_{2}}-2\epsilon_{{\cal Z}_{1}}\epsilon_{{\cal Z}_{2}})\right)^{{O(M\log (M))}}.$
\end{cor}
\begin{rem}\label{rem4}
As we can see, the probability is affected by $M$. When $M$ increases,
the probability $P\left(\textrm{the rankings in } \mathbf{a}\textrm{ and }\mathbf{b}\textrm{ agree}\right)$
will converge to a small quantity. In fact, when $M\rightarrow\infty,$
we have $\lim_{M\rightarrow\infty}\lim_{n\rightarrow\infty}P\left(\textrm{ rank correlation of }\mathbf{a}\textrm{ and }\mathbf{b}\textrm{ is }1\right)=0$
if $\epsilon_{{\cal Z}_{1}}+\epsilon_{{\cal Z}_{2}}<2$. The only
case $\lim_{M\rightarrow\infty}\lim_{n\rightarrow\infty}P\left(\textrm{ rank correlation of }\mathbf{a}\textrm{ and }\mathbf{b}\textrm{ is }1\right)=1$
is when $\pi$ is consistent in both ${\cal Z}_{1}$ and ${\cal Z}_{2}$,
i.e., $\epsilon_{{\cal Z}_{1}}=\epsilon_{{\cal Z}_{2}}=1$. It suggests that the choice of internal measure $\pi$ is important for comparing multiple deep clustering results. If the internal measure is not consistent, a large $M$ will naturally make the task of comparing different clustering algorithms challenging, even infeasible. 
\end{rem}

\begin{rem}\label{rem:link}

We consider a weighted scheme based on two key factors. First, if we obtain a collection of estimates ($\{\pi(\cdot|{\cal Z}_{1}),\pi(\cdot|{\cal Z}_{2}),...\}$) for the unknown perfect score \(\pi(\cdot|{\cal Z}^{*})\) from an unknown perfect space \({\cal Z}^{*}\), such that for all \(j\), \(E[\pi(\phi|{\cal Z}_j) - \pi(\phi|{\cal Z}^{*})] \xrightarrow{p} 0\) and \(Var[\pi(\phi|{\cal Z}_j) - \pi(\phi|{\cal Z}^{*})] \xrightarrow{p} \sigma_j^2\) for some constants \(\sigma_j\), then using a weighted score \(\sum_{j}w_j\pi(\phi|{\cal Z}_j)\) with appropriately chosen weights \(w_j\) helps reduce variance. Second, consider the special case of \( M\) clustering results \( \{({\cal Z}_m, \rho_m)\}_{m=1}^M \) where $\rho_1=...=\rho_M$ (up to permutations) and ${\cal Z}_{m_1}\neq {\cal Z}_{m_2}$ for at least a pair of $m_1\neq m_2\in \{1,...,M\}$, an evaluation criterion that prioritizes clustering accuracy over embedding space quality should assign equal scores to all results. This property is satisfied by the weighted score but not by the paired score. 

A traditional approach in model averaging assigns weights \( w_j \) to the internal measure calculated on  \( {\cal Z}_j \) based on an assumed likelihood. However, due to the unsupervised and nonparametric nature of clustering evaluation, we instead use link analysis to determine the weights $w_j$. Link analysis is a valuable technique for assessing node relationships and assigning node importance. By treating each candidate space ${\cal Z}_j$ as a node, we define the edge strength between two spaces based on the pairwise rank correlation of their internal measures. Then, the weight of a node (i.e., embedding space in our setting) is determined by the extent to which one embedding space is connected with the other spaces, such as the total connectivity of a node or the stationary probability associated with a stochastic matrix. More details of the link analysis are illustrated in SM Section S4.3. 
\end{rem}
\vspace{-4mm}
%%%%%%%%%%%%%%%%%%%%%%%%%%%%%%%%%%%%%%%%%%%%%%%%%%%%%%%%%%%%%%%
\section{The Proposed Adaptive Clustering Evaluation (ACE)}\label{sec:3}
%%%%%%%%%%%%%%%%%%%%%%%%%%%%%%%%%%%%%%%%%%%%%%%%%%%%%%%%%%%%%%%
\vspace{-4mm}

Consider $M$ deep clustering outputs,  each comprising embedded data and its corresponding partitioning outcome for comparison, denoted as $(\mathcal{Z}_{m}, \rho_{m})$, where $m=1,...,M$. These outputs may arise from different algorithms or configurations but are evaluated on the same task. Our objective is to detect a group of admissible spaces $\mathcal{Z}_1, \ldots, \mathcal{Z}_L$, $L \le M$ for the selected internal measure $\pi$, aiming for a rank measurement more likely to align with the external measure than not. As noted in Remark \ref{rem:link}, we employ an ensemble-style scoring scheme to estimate a final score across selected spaces. A straightforward version of this ensemble-style score is simply averaging the scores over all available embedding spaces $\pi(\rho|\mathcal{Z}_1), \cdots, \pi(\rho|\mathcal{Z}_M)$, referred as the \emph{pooled score} (Figure \ref{fig1}), which we include as a comparative approach. Based on these ideas, we introduce an \underline{A}daptive \underline{C}lustering \underline{E}valuation (\emph{ACE}) strategy for deep clustering assessment, which includes multimodality testing, space screening and grouping, and ensemble analysis, as outlined in Algorithm \ref{Algorithm1}.

%%%%%%%%%%%%%%%%%%%%%%%%%%%
\begin{algorithm*}\small
\caption{\small \small Adaptive clustering evaluation (ACE) for deep clustering models}
\label{Algorithm1} 
 {\bf Input: }Clustering outputs $ \{(\mathcal{Z}_{m}, \rho_{m})\}_{m=1}^M$; internal measure $\pi$
\begin{algorithmic}[1]
\State Multimodality test:  for each $\mathcal{Z}_{m}$, perform Dip test and get the \emph{p}-value, and apply a multiple testing procedure to select retained spaces.  If no spaces are retained, skip this step. To ease the notation, we still denote $\mathcal{Z}_1,...,\mathcal{Z}_{M}$ as the retained spaces.
\State Space screening and grouping:
\begin{enumerate}[noitemsep]
    \item For each retained embedding space $m \in \{1,...,M\}$, calculate $\boldsymbol{\pi}_m=(\pi(\rho_{1} | \mathcal{Z}_m),\pi(\rho_{2} | \mathcal{Z}_m),...,\pi(\rho_{M} | \mathcal{Z}_m))$.
    \item Calculate rank correlation $r_{mm^{'}}:=RankCorr(\boldsymbol{\pi}_m, \boldsymbol{\pi}_{m^{'}})$ for each pair $(m, m^{'})$.
    \item Based on the rank correlation matrix $\{r_{mm^{'}}\}_{m,m^{'}=1}^{M}$, perform density-based stage-wise grouping (SM Section S4.2) to divide the $M$ embedding spaces into $S$ mutually exclusive subgroups $\{G_s\}_{s=1}^S$. 
\end{enumerate}
\State Ensemble analysis:
\begin{enumerate}[noitemsep]
    \item For each subgroup $G_s$, build an undirected graph $\mathcal{G}_s=(V_s,E_s)$ where $V_s=G_s$ and $E_s=\{e_{mm^{'}}\}_{m,m^{'} \in G_s}$ with $e_{mm^{'}} = r_{mm^{'}}$ for significantly positive-correlated spaces $\mathcal{Z}_{m}$ and $\mathcal{Z}_{m^{'}}$, else $e_{mm^{'}} = 0$.
    \item For each $\mathcal{Z}_{m}$ in the $s$-th group $G_s$, run a link analysis to get the rating $w^{(s)}_{m}$. Then calculate $\pi(\rho_{m^{'}} | G_s) = \sum_{m \in G_s} w^{(s)}_{m} \pi(\rho_{m^{'}} | \mathcal{Z}_{m})$ for each $m^{'}=1,\cdots,M$.
    \item Select $G_s* = \argmax_{G_s} \sum_{m^{'}=1}^M\pi(\rho_{m^{'}} | G_s)/M$
\end{enumerate}
\end{algorithmic}
{\bf Output: }$\pi(\rho_{1} | G_{s^*}),\cdots, \pi(\rho_{M} | G_{s^*})$
\end{algorithm*}
%%%%%%%%%%%%%%%%
\noindent \textbf{Multimodality test.} According to  \citet{liu2008statistical}, one of the key challenges in statistical clustering is
determining whether the clusters observed are genuine clustering within the data or
simply random artifacts resulting from sampling variations. Genuine clusters typically
manifest as distinct groups or peaks in the distribution of embedded data. Therefore, it is reasonable to expect that admissible spaces--where our internal measure $\pi$ is
trustworthy--should display multimodality, indicating multiple peaks or distinct regions in
the data distribution. We employ the widely applied multimodality testing method, \emph{Dip test} \citep{hartigan1985dip}, which assesses the presence of multiple modes in the data distribution without assuming a specific underlying distribution. We retain the embedding spaces with significant multimodality. The Dip test quantifies clusterability by estimating the discrepancy between the empirical cumulative distribution function (CDF) and the nearest unimodal CDF. \emph{p}-values, obtained via Monte Carlo simulations with a uniform null model, undergo multiple testing to control family-wise error rates (FWERs), with a default threshold of \(\alpha_{dip} = 0.05\). Only embedding spaces that reject the null hypothesis are retained, indicating a non-unimodal distribution (see SM Section S4.1 for details). If no spaces are retained, we skip this step and proceed with all spaces in the subsequent analysis. While the Dip test does not inherently extend to multivariate data, unidimensional reduction techniques like Principal Component Analysis are commonly used to adapt it for high-dimensional settings (SM Section S5.2). We choose the Dip test due to its common usage and simplicity. However, other statistical tests for multimodality could also be applied in this context \citep{liu2008statistical, huang2015statistical, tibshirani2001estimating}.

\noindent \textbf{Space screening and grouping.}
For each retained embedding space $\mathcal{Z}_{m}$, based on the chosen internal measure, we calculate the measure values across all partitioning results, denoted as $(\pi(\rho_{1} | \mathcal{Z}_m), \pi(\rho_{2} | \mathcal{Z}_m), ..., \pi(\rho_{M} | \mathcal{Z}_m))$. 
Following our discussions in Section \ref{sec:2}, we divide the retained spaces into groups based on their rank correlation of measure values and select the group with the highest value of the internal measure (see more details in the step of ensemble analysis). Considering the absence of prior knowledge about the number of groups, we adopt density-based clustering approaches like HDBSCAN \citep{mcinnes2017hdbscan} as suitable methods. These approaches are particularly well-suited because they eliminate the need to specify the number of groups and can identify outlier spaces during grouping. Alternative statistical approaches, such as Nonparametric Bayesian clustering or clustering with internal validation, can be considered when the number of groups is unknown. However, they may struggle to detect outliers, require prior specifications, or lack a direct way to incorporate pairwise correlations. After forming the initial groups, we create subgroups within each group to refine our selection further. These subgroups are based on the scale of the measure values, excluding spaces that, while correlated, may differ significantly in scale and could potentially introduce noise into the analysis. This results in a stage-wise grouping scheme guided by a density-based approach: we first group embedding spaces based on their rank correlations; subsequently, we form subgroups, denoted as \( \{G_s\}_{s=1}^S \), within the initial groups based on the index values. Ultimately, we select the subgroup that yields the highest aggregated index values as the final evaluation result. The method for computing these aggregated scores for a subgroup will be discussed in the subsequent section. Please refer to SM Section S4.2 for more details on implementing the stage-wise algorithm. 

\begin{figure}[h]
\centering
\includegraphics[scale=0.34]{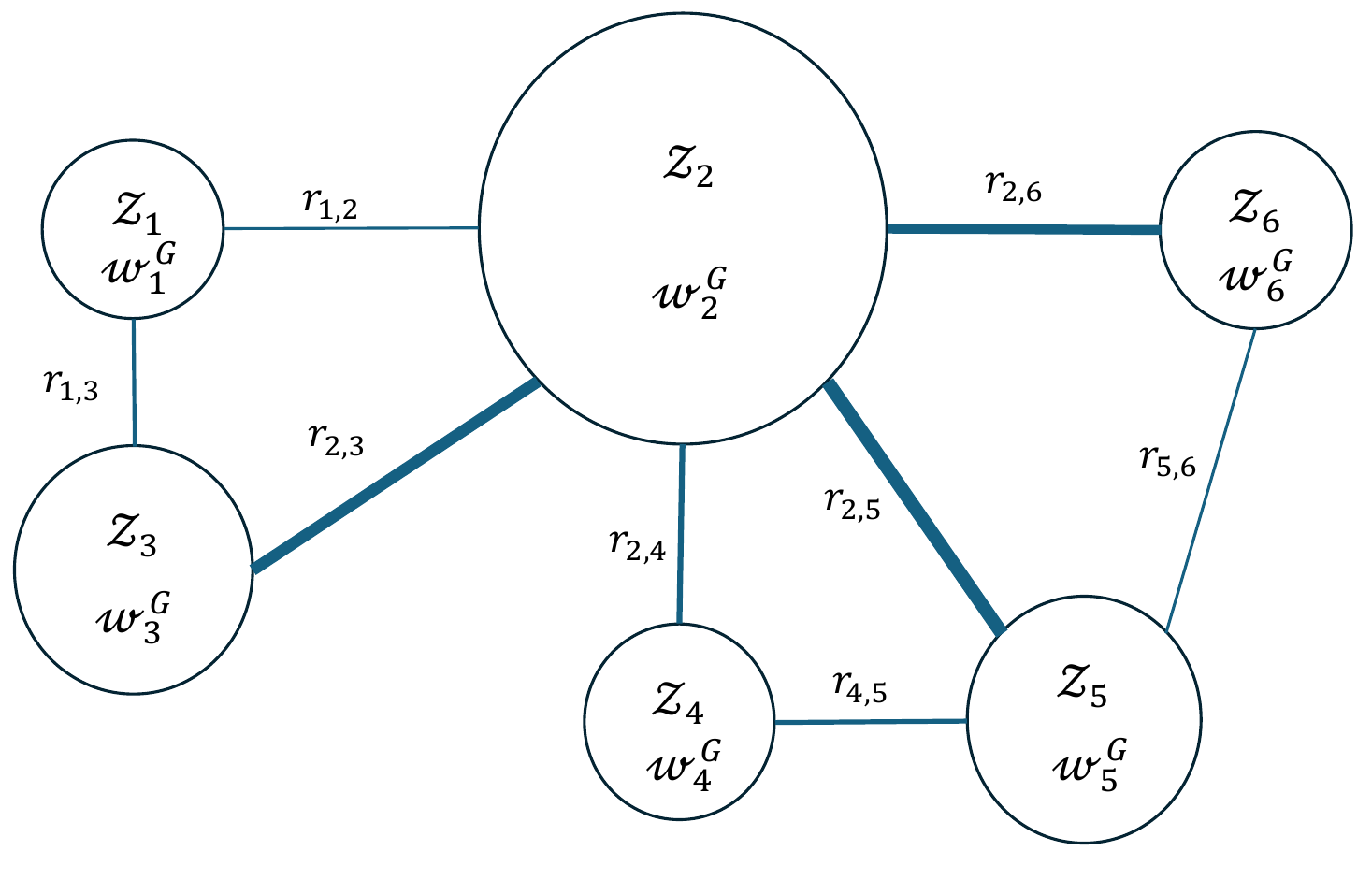}\caption{\footnotesize A graphical representation of the undirected graph $\mathcal{G}$ for a subgroup containing six embedding spaces. Each node represents a space retained in the subgroup, and an edge is drawn between two nodes if their corresponding spaces exhibit significant rank correlation. The rank correlation coefficient, denoted as $r_{mm^{\prime}}$, measures the strength of the correlation between two nodes. A link analysis algorithm calculates $w_{m}$ to evaluate each node's importance by considering the strength of both its direct and indirect connections within the graph. In this illustration, bolder edges correspond to higher values of $r_{mm^{\prime}}$, while larger nodes indicate higher $w_{m}$ values, reflecting the greater significance of the corresponding node in the graph. } \label{fig:graph}
\end{figure}

\noindent \textbf{Ensemble analysis.} For each subgroup with more than one space, we propose an ensemble analysis to obtain an aggregated score. Consider a subgroup $G$ with $m_G$ embedding spaces denoted as $\{\mathcal{Z}_{m}\}_{m \in G}$. Within this subgroup, we treat each space as a vertex and represent the pairwise rank correlations with an undirected graph, $\mathcal{G}= (V, E)$, where $V$ denotes embedding spaces, and $E$ consists of edges weighted by the rank correlation $r_{mm^{'}}$ as defined in Algorithm \ref{Algorithm1}. To construct the edge set, we connect vertices only if their corresponding spaces exhibit statistically significant rank correlation, determined through a multiple testing procedure that controls FWERs, with a default threshold of \(\alpha_{edges} = 0.1\). In this procedure, each individual test operates under the null hypothesis of no Spearman's correlation between the two spaces, with the alternative hypothesis of a positive correlation. Figure \ref{fig:graph} provides a graphical illustration of this graph for a subgroup containing six embedding spaces. After obtaining the graph, we can run a link analysis to rate each space based on the magnitude of its link to other spaces. The basic idea is that a top-rated space in a subgroup should be a hub, demonstrating high rank correlation with many other spaces in the same subgroup. We consider link analysis algorithms, such as \emph{PageRank} \citep{ding2002pagerank} (our default choice) or \emph{HITS} \citep{kleinberg1999authoritative}, with brief introductions to these algorithms provided in SM Section S4.3 and implementation details in SM Section S5.2. With this procedure, we can obtain a normalized rating $w^G_{m}$ for each space, which serves as a weight in the aggregation. Using these ratings, we generate a score by aggregating the internal measure values of all the embedding spaces within the subgroup, represented as $\pi(\cdot | G) = \sum_{m \in G} w^G_{m} \pi(\cdot | \mathcal{Z}_{m})$. If a subgroup contains only a single space, its score is directly taken as the aggregated score. After computing the score vector \( (\pi(\rho_{1} | G_s), \dots, \pi(\rho_{M} | G_s)) \) for each subgroup \( G_s \), we ultimately select the subgroup \( G_{s^*} \) whose score vector has the highest average value among all subgroups. This ensures the selection of embedding spaces that exhibit both strong rank correlation and high values.

\vspace{-8mm}
%%%%%%%%%%%%%%%%%%%%%%%%%%%%%%%%%%%%%%%%%%
\section{Simulations} \label{sec:sim}
\vspace{-4mm}
We present two simulation examples demonstrating the robust performance of ACE.%We start with Section \ref{sec:datagenerate} by introducing the data generating process. 
\vspace{-5mm}
\subsection{Simulation Setup\label{sec:datagenerate}}
\vspace{-4mm}
\noindent\textbf{Raw data.} The raw data $\mathbf{X}\in \mathbb{R}^{n\times p}$ consists of \( K = 20 \) clusters, where $n=1000$ and $p=50000$. For the $k$-th cluster, $k=1,...,K$, we generate $n_k=\frac{n}{K}=50$ i.i.d. observations from $N(\boldsymbol{\mu}_k,\Sigma)$, where $\boldsymbol{\mu}_{k}=(\boldsymbol{\alpha}_{k},\boldsymbol{\beta}_{k},0,...,0)^{T}$ for some $\boldsymbol{\alpha}_k \in \mathbb{R}^{p_1}$,$\boldsymbol{\beta}_k \in \mathbb{R}^{p_2}$, and $\Sigma=\left(\begin{array}{cc}
\sigma^{2}I_{p_{1}+p_{2}} & 0\\
0 & I_{p-p_{1}-p_{2}}
\end{array}\right)$ with $\sigma \in \{1,2\}$.
%More details of $p_1$ and $p_2$ are deferred to Section \ref{sec:ex1}
We can see that the first $p_1+p_2$ features are informative, and the rest are non-informative. Details of $p_1$ and $p_2$ will be provided in later examples. In each replicated experiment, we randomly generate $\boldsymbol{\alpha}_k$ and $10\cdot\boldsymbol{\beta}_k$ from the standard Gaussian distribution.

\noindent\textbf{Embedding space.} Let $[n]:=\{1,...,n\}$ for any integer $n$. Given the data matrix $\mathbf{X}:=\{{x_{ij}\}_{n\times p}}$, for any index set $S\subseteq [p]$, we denote $\mathbf{X}_S:=\{X_{ij}\}_{i\in[n],j\in S}$ as the submatrix by concatenating the columns of $\mathbf{X}$ with index in $S$. Given a raw dataset, we simulate the embedded data by selecting 100 features, including both informative and non-informative ones, and applying a transformation. Specifically, the embedded data is \( {\cal Z} := f_{\text{transform}}(\mathbf{X}_{S_1 \cup S_2}) \), where \( S_1 \subseteq [p_1] \), \( S_2 \subseteq [p] \setminus [p_1] \), and the total cardinality \( |S_1| + |S_2| = 100 \). For the transformation function $f_{transform}$, we consider both the identity function $f(x)=x$ and a set of nonlinear transformations. Specifically, for nonlinear transformations, we apply UMAP transformation \citep{mcinnes2018umap}, denoted as $f_{\textrm{UMAP}(a,b)}(\cdot):\mathbb{R}^{100}\rightarrow \mathbb{R}^{100}$, where $a\in\{5, 20, 50\}$ is the local neighborhood size and $b\in\{0.1,0.5,0.99\}$ is the minimum allowable distance between points. The index set of $S_1$ is randomly generated such that $|S_1|=10,15,20, 25$ for the low-noise case $\sigma=1$ and $|S_1|=10,25,40, 50$ for the high-noise case $\sigma=2$ (in each replicated experiment, $S_1$ and $S_2$ remain fixed across different transformations). In summary, by varying the transformation function and $|S_1|$, we generate 40 embedding spaces that span a range of performance levels.
%\noindent\textbf{}

\begin{figure}[htbp]
\centering
\subfigure{\includegraphics[width = 0.45\linewidth]{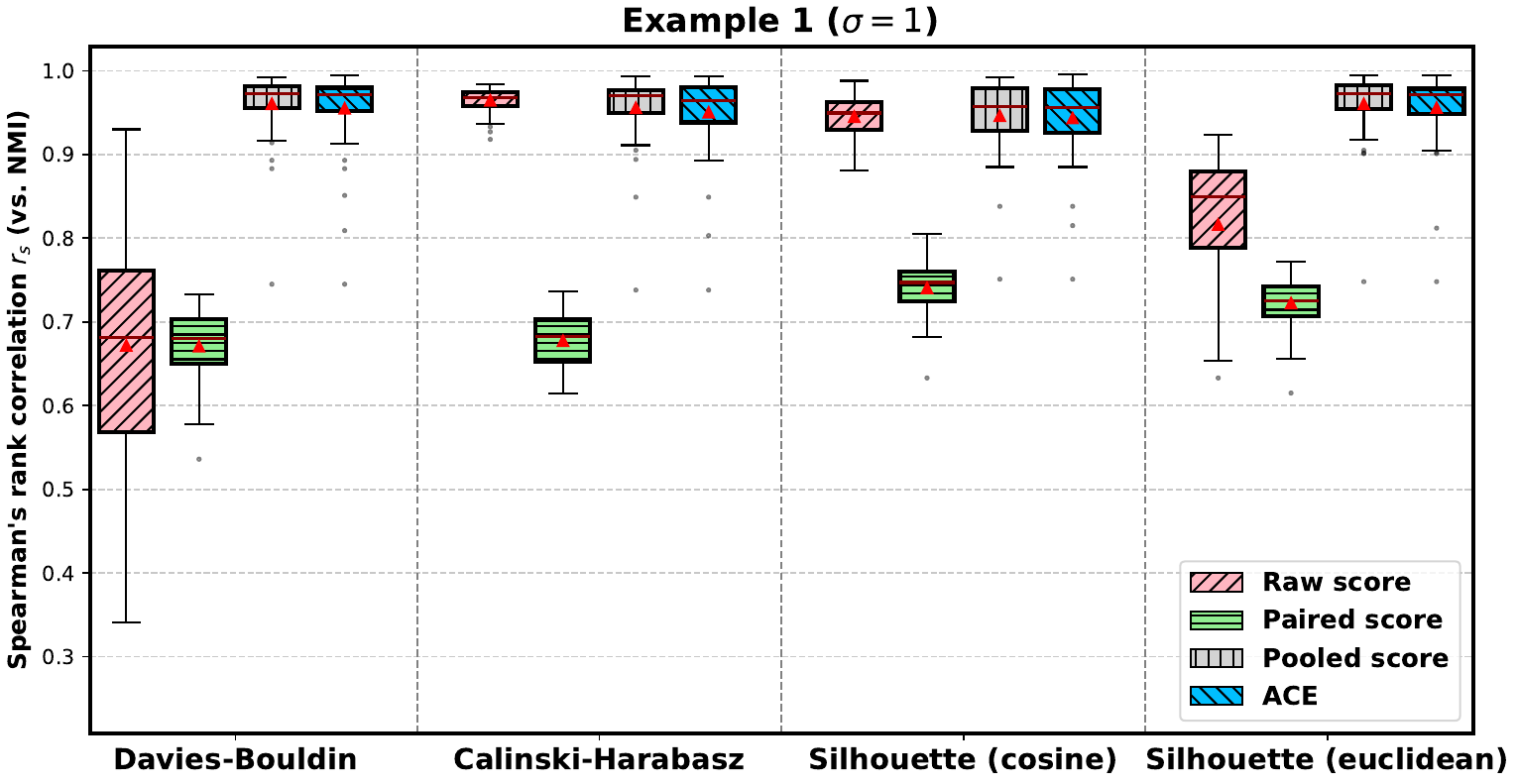}}
\subfigure{\includegraphics[width = 0.45\linewidth]{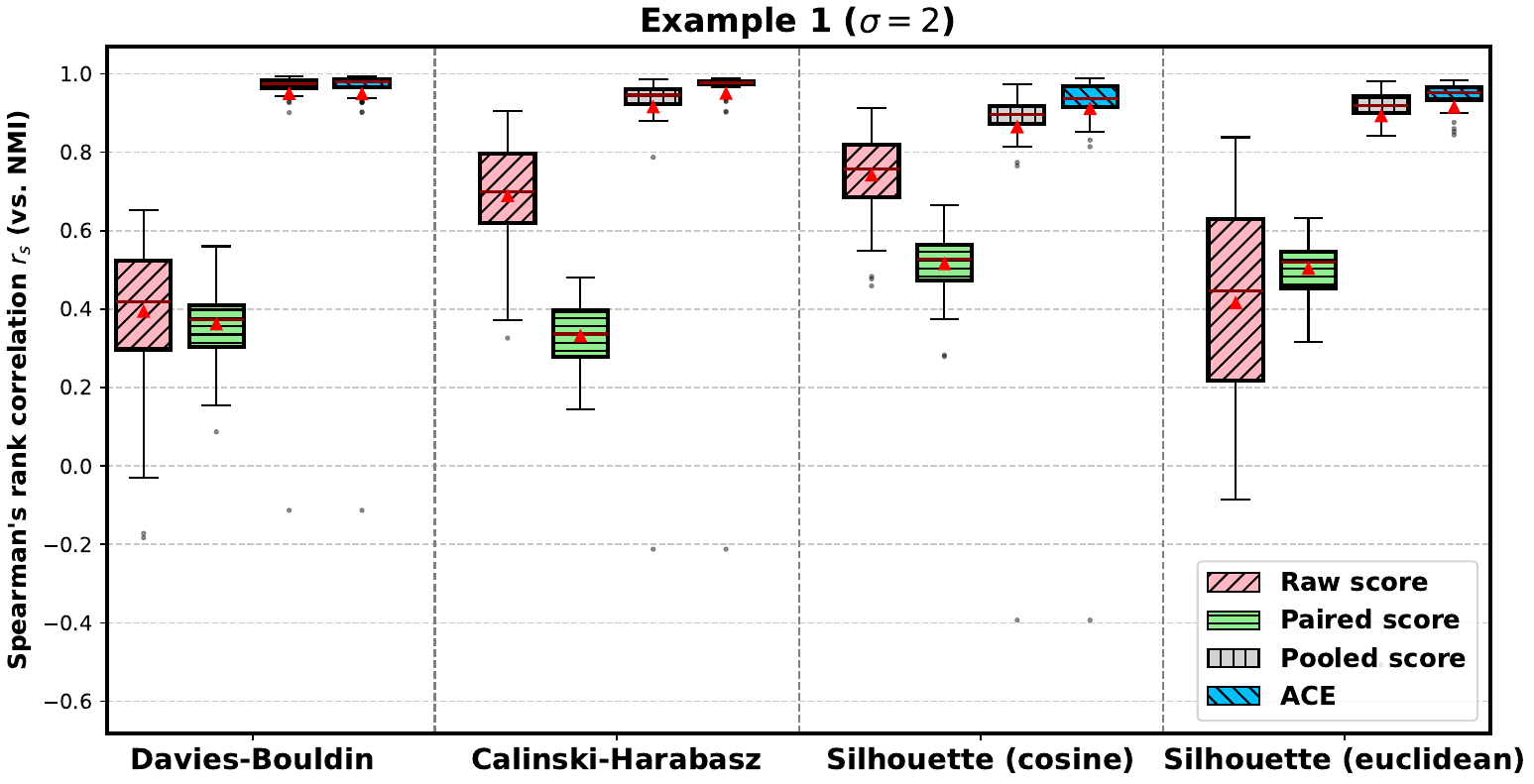}}\\
\subfigure{\includegraphics[width = 0.45\linewidth]{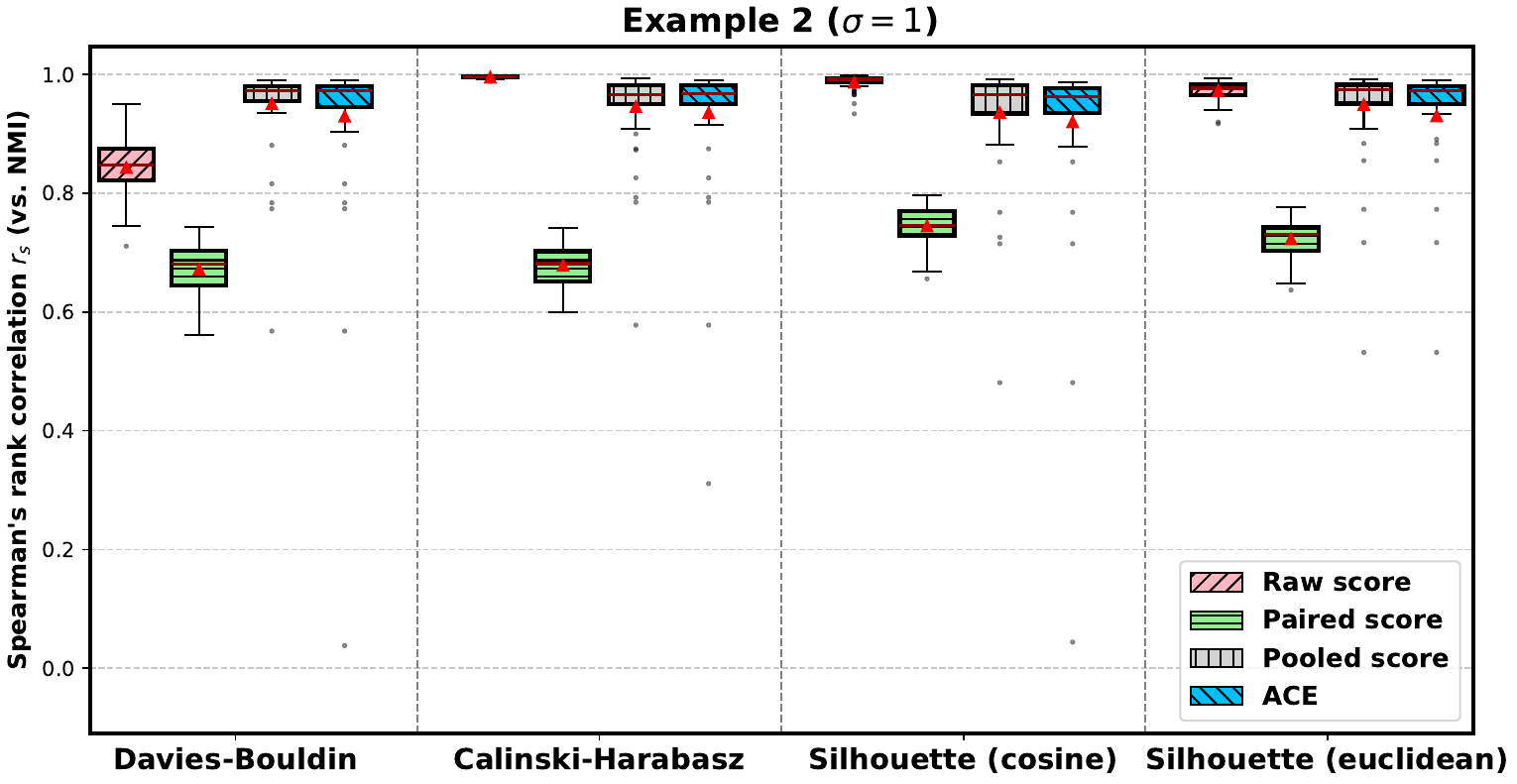}}
\subfigure{\includegraphics[width = 0.45\linewidth]{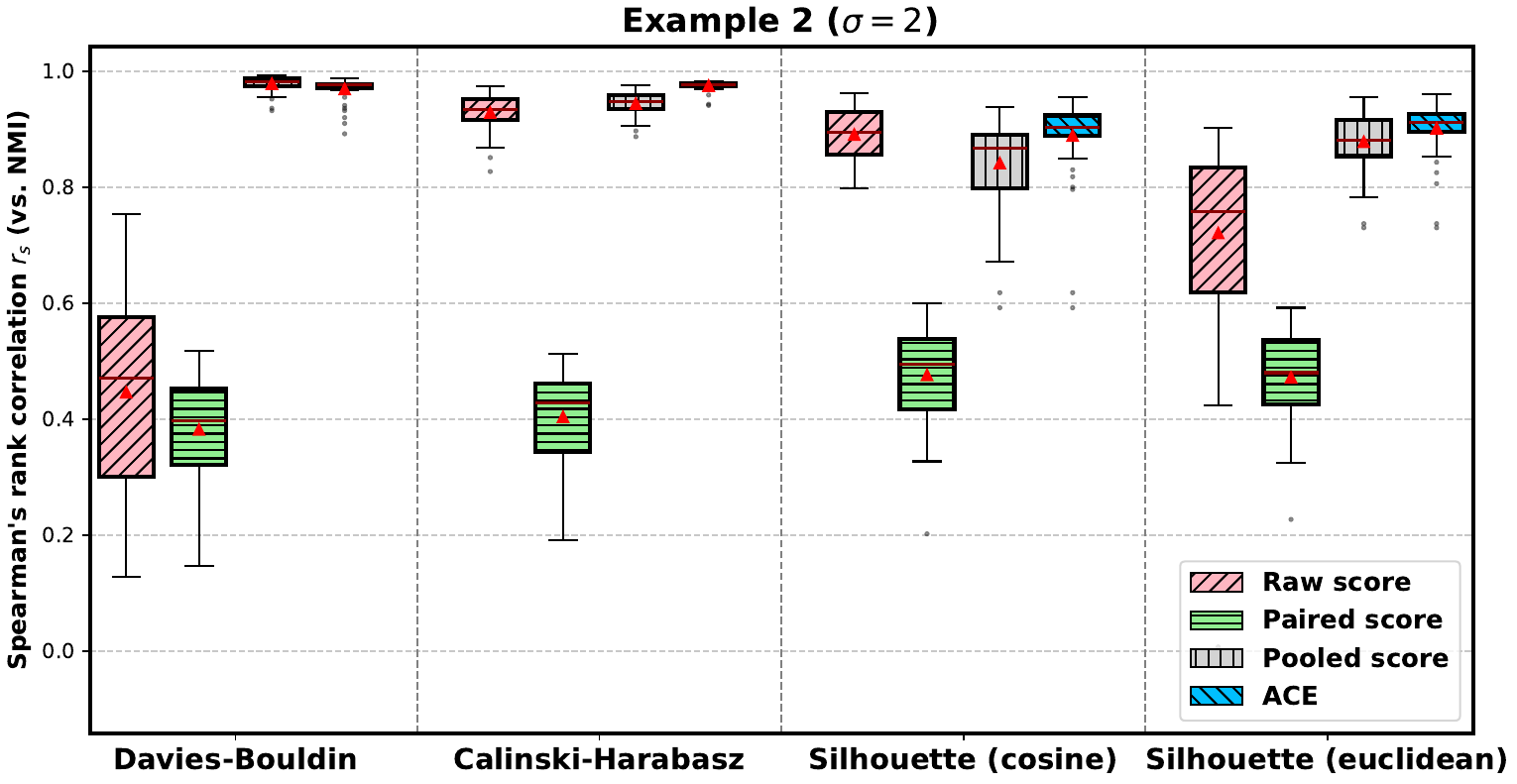}}
\caption{\small Boxplots illustrate rank consistency based on Spearman rank correlation coefficients ($r_s$) with NMI. The two figures in the first row correspond to Simulation Example 1, and the two figures in the second row correspond to Simulation Example 2. The mean value for each evaluation approach across simulation studies is highlighted with a red triangle in each plot.}
\label{fig:main:sim}
\end{figure}

\noindent\textbf{Clustering method \& evaluation criteria.} Clustering results are obtained by applying K-Means to each embedded data. We denote $\rho_m:=\textrm{KMeans}({\cal Z}_m)$ for $m=1,...,40$. We compare the performance of ACE with three evaluation approaches (raw score, paired score, and pooled score (after Dip test)) as in Figure \ref{fig1}. We calculate the rank correlation between the internal measure value obtained under these approaches and the external measure. For internal measures, we consider three widely used metrics: the Silhouette score (with Euclidean or Cosine distance), the Calinski-Harabasz index, and the Davies-Bouldin index. As external measures, we use normalized mutual information (NMI) and clustering accuracy (ACC). To quantify rank consistency, we reported Spearman's rank correlation coefficient ($r_s$) and Kendall's rank coefficient ($\tau_B$), as defined in SM Section S5.1. Additional implementation details can be found in SM Section S5.2. Due to space limitations, Figure \ref{fig:main:sim} shows boxplots of the Spearman rank correlations between each approach and NMI over 50 replicated experiments for each setting. Boxplots for other rank correlation types and external measures from the same experiments are provided in Figures S1-S3.
\vspace{-4mm}
\subsection{Two Examples}\label{sec:ex1}
\vspace{-2mm}
\textbf{Example 1: sparse mean difference.} We set $p_1=50$ and $p_2=0$ so that 50 features in the raw dataset are informative for differentiating clusters.

In Example 1, ACE outperforms the other three approaches across different internal measures, variance levels (\(\sigma^2\)), and external measures. It is worth noting that the other three approaches perform comparably to ACE in some cases. For instance, under the low-variance setting (\(\sigma = 1\)), the pooled score performs similarly, and the raw score shows comparable performance on certain indices. As variance increases from \(\sigma = 1\) to \(\sigma = 2\), ACE remains the best-performing approach in terms of Spearman correlation, while the performance of the raw and paired scores deteriorates significantly. The pooled score also declines, with its Spearman correlation remaining lower than that of ACE across most internal measures. This degradation stems from increased data spread across clusters, leading to greater overlap between clusters. The intra-cluster distance then becomes comparable to the inter-cluster distance, elevating the noise level. Consequently, internal measures become less reliable as proxies for actual clustering performance. This inconsistency is particularly pronounced for the raw score due to the curse of dimensionality, which makes clustering less distinguishable. As the signal-to-noise ratio decreases, discrepancies between the candidate embedding spaces \(\mathcal{Z}_m\) grow, making simple averaging of performance scores (pooled score) less effective.

\noindent\textbf{Example 2: dense mean difference.} We set \( p_1 = 50 \) and \( p_2 = 10000 \), resulting in $10050$ informative features for differentiating clusters, with $10000$ of them having weaker signals.

In Example 2, similar conclusions to Example 1 can be drawn: ACE consistently achieves the best performance in most cases. Therefore, we focus on comparing Example 1 and Example 2. Notably, when cluster centers have dense differences, i.e., the number of features differentiating two clusters is not sparse, the performance of the raw score improves significantly. A greater number of distinguishing features enhances distance discernibility, mitigating the effects of the curse of dimensionality.  However, the inclusion of dense features has minimal impact on scores derived from the embedding space (ACE, pooled score, and paired score). Similar trends are observed in Spearman’s rank correlation with ACC and Kendall’s rank correlation with both NMI and ACC (Figures S1-S3). The stability of ACE across both dense-difference and sparse-difference settings underscores its effectiveness in comparing clustering results from various embedded representations of the same high-dimensional data, a key characteristic of deep clustering.

\vspace{-3mm}
%%%%%%%%%%%%%%%%%%%%%%%%%%%%%%%%%%%%%%%%%%
\section{Real Data Analysis} \label{sec:exp}
\vspace{-2mm}
In this section, we evaluate how ACE performs in evaluating deep clustering methods on real datasets. Specifically, we focus on evaluating two well-known deep clustering methods: \emph{DEPICT} \citep{ghasedi2017deep} and \emph{JULE} \citep{yang2016joint}  as representatives of autoencoder-based and CDNN-based approaches, respectively, on clustering tasks. We ran the source code of DEPICT and JULE on datasets specified in their original papers for image clustering tasks. These datasets consist of COIL20 and COIL100 (multi-view object image datasets)~\citep{nene1996columbia}, USPS and MNIST$_{\text{test}}$ (handwritten digits datasets)~\citep{hull2002database,lecun1998gradient}, and UMist, FRGC-v2.0, CMU-PIE, and Youtube-Face (YTF) (face image datasets)~\citep{graham1998characterising,frgc2024,sim2002cmu,wolf2011face}. USPS, MNIST$_{\text{test}}$, YTF, FRGC, and CMU-PIE are employed in both JULE and DEPICT papers, while COIL-20, COIL-100, and YTF are used exclusively in JULE. 

Table \ref{app:tab:0} provides details on sample size, image size, and the number of classes for all datasets. Using these two deep clustering methods and datasets, we conducted experiments to address key applications in model validation and result evaluation for deep clustering, including hyperparameter tuning and evaluating the number of clusters, with the results presented in the main text. The evaluation criteria are consistent with those in Section \ref{sec:sim}. Additionally, we assess internal measures such as the CCC index, Dunn index, C-index, SDbw index, and CDbw index, with the results provided in SM Sections S8.1 and S8.3. We also investigate checkpoint selection, a key aspect of deep clustering model validation. Using \emph{DeepCluster} \citep{caron2018deep}, known for its success on large-scale datasets like ImageNet~\citep{deng2009imagenet}, we ran experiments on ImageNet's validation set. Due to space constraints, checkpoint selection results are deferred to SM Section S8.6. A related literature review on these deep clustering methods is also provided in SM Section S6.

\begin{table}[h]
\small
\caption{Dataset information}
\centering
    \resizebox{0.9\textwidth}{!}{
\begin{tabular}{ccccccccc}
\hline
 & COIL20 & COIL100 & CMU-PIE & YTF & USPS & MNIST$_{\text{test}}$ & UMist & FRGC \\
\hline
\#Samples & 1440 & 7200 & 2856 & 10000 & 11000 & 10000 & 575 & 2462 \\
Size & 128$\times$128 & 128$\times$128 & 32$\times$32 & 55$\times$55 & 16$\times$16 & 28$\times$28 & 112$\times$92 & 32$\times$32 \\
\#Classes & 20 & 100 & 68 & 41 & 10 & 10 & 20 & 20 \\
\hline
\end{tabular}
}
\label{app:tab:0}
\end{table}
\vspace{-10mm}
\subsection{Application 1: Hyperparameter Tuning}
\vspace{-2mm}
We consider the hyperparameter tuning process in deep clustering. Different clustering outputs are generated by applying the same deep clustering model (JULE or DEPICT) with varying hyperparameter configurations. Given \(M\) hyperparameter pairs to explore, for each dataset, we train the deep clustering model using the \(m\)-th pair of hyperparameters and obtain the corresponding clustering output. Specifically, for JULE, we explore \(42\) pairs of values for two key hyperparameters: learning rate and unfolding rate. For DEPICT, we investigate \(18\) pairs for the learning rate and balancing parameter. Additional details regarding the hyperparameter configurations and implementation are provided in SM Section S5.2. We summarize the key findings from Tables \ref{tab:nmi:hyper} and S1 as follows.

\begin{table}[h]
\caption{\small Rank consistency with NMI for different evaluation approaches in Application 1. Here, $r_{s}$ and $\tau_{B}$ denote the Spearman and Kendall rank correlation coefficients. Empty cells indicate cases where the result is either missing or impractical to obtain. The top results for each dataset are highlighted in bold. For rank consistency with ACC, refer to Table S1.}
\centering \resizebox{\textwidth}{!}{ 
\begin{tabular}{lllllllllllllllllll}
\toprule
 & \multicolumn{2}{l}{USPS} & \multicolumn{2}{l}{YTF} & \multicolumn{2}{l}{FRGC} & \multicolumn{2}{l}{MNIST$_{\text{test}}$} & \multicolumn{2}{l}{CMU-PIE} & \multicolumn{2}{l}{UMist} & \multicolumn{2}{l}{COIL-20} & \multicolumn{2}{l}{COIL-100} & \multicolumn{2}{l}{Average} \\
& $r_s$ & $\tau_B$ & $r_s$ & $\tau_B$ & $r_s$ & $\tau_B$ & $r_s$ & $\tau_B$ & $r_s$ & $\tau_B$ & $r_s$ & $\tau_B$ & $r_s$ & $\tau_B$ & $r_s$ & $\tau_B$ & $r_s$ & $\tau_B$ \\
\midrule
\hline 
 \multicolumn{19}{c}{\emph{JULE}: Calinski-Harabasz index} \\
 \hline 
Raw score & 0.58 & 0.47 & 0.79 & 0.62 & -0.44 & -0.28 & 0.81 & 0.62 & -0.99 & -0.93 & -0.57 & -0.40 & -0.31 & -0.18 & 0.32 & 0.21 & 0.02 & 0.01 \\
Paired score & 0.17 & 0.13 & 0.52 & 0.40 & -0.13 & -0.10 & 0.49 & 0.34 & -0.13 & -0.08 & 0.70 & 0.50 & 0.53 & 0.38 & 0.20 & 0.19 & 0.29 & 0.22 \\
Pooled score & \noindent \textbf{0.84} & \noindent \textbf{0.68} & \noindent \textbf{0.91} & \noindent \textbf{0.79} & 0.29 & 0.22 & 0.82 & 0.67 & 0.94 & 0.82 & \noindent \textbf{0.81} & 0.60 & \noindent \textbf{0.62} & \noindent \textbf{0.47} & 0.89 & 0.73 & 0.77 & 0.62 \\
\noindent \textbf{ACE} & 0.80 & 0.63 & 0.90 & 0.73 & \noindent \textbf{0.39} & \noindent \textbf{0.26} & \noindent \textbf{0.87} & \noindent \textbf{0.71} & \noindent \textbf{0.98} & \noindent \textbf{0.90} & \noindent \textbf{0.81} & \noindent \textbf{0.61} & 0.60 & 0.45 & \noindent \textbf{0.95} & \noindent \textbf{0.82} & \noindent \textbf{0.79} & \noindent \textbf{0.64} \\
\hline 
 \multicolumn{19}{c}{\emph{JULE}: Davies-Bouldin index} \\
 \hline 
Raw score & -0.48 & -0.30 & -0.47 & -0.32 & -0.43 & -0.30 & -0.83 & -0.67 & -0.97 & -0.88 & -0.70 & -0.50 & -0.58 & -0.40 & -0.79 & -0.61 & -0.66 & -0.50 \\
Paired score & -0.10 & -0.03 & -0.32 & \noindent \textbf{-0.21} & -0.08 & -0.05 & -0.13 & -0.06 & 0.26 & 0.20 & \noindent \textbf{0.62} & \noindent \textbf{0.44} & \noindent \textbf{0.61} & \noindent \textbf{0.42} & 0.43 & \noindent \textbf{0.35} & 0.16 & 0.13 \\
Pooled score & -0.26 & -0.12 & -0.46 & -0.34 & 0.11 & 0.07 & -0.16 & -0.07 & \noindent \textbf{0.92} & \noindent \textbf{0.78} & 0.30 & 0.20 & -0.25 & -0.17 & -0.46 & -0.35 & -0.03 & 0.00 \\
\noindent \textbf{ACE} & \noindent \textbf{-0.08} & \noindent \textbf{-0.02} & \noindent \textbf{-0.30} & \noindent \textbf{-0.21} & \noindent \textbf{0.22} & \noindent \textbf{0.16} & \noindent \textbf{0.73} & \noindent \textbf{0.55} & 0.10 & 0.06 & 0.38 & 0.27 & 0.23 & 0.22 & \noindent \textbf{0.48} & 0.33 & \noindent \textbf{0.22} & \noindent \textbf{0.17} \\
\hline 
 \multicolumn{19}{c}{\emph{JULE}: Silhouette score (cosine distance)} \\
 \hline 
Raw score & 0.68 & 0.51 & 0.84 & 0.69 & 0.03 & 0.01 & 0.64 & 0.49 & 0.66 & 0.50 & -0.46 & -0.34 & -0.14 & -0.11 & 0.12 & 0.08 & 0.30 & 0.23 \\
Paired score & 0.28 & 0.22 & 0.73 & 0.56 & 0.09 & 0.06 & 0.63 & 0.47 & 0.50 & 0.36 & 0.71 & 0.50 & \noindent \textbf{0.68} & \noindent \textbf{0.50} & 0.74 & 0.54 & 0.54 & 0.40 \\
Pooled score & 0.70 & 0.56 & \noindent \textbf{0.93} & 0.81 & 0.40 & 0.27 & 0.79 & 0.64 & 0.95 & 0.85 & 0.77 & 0.56 & 0.27 & 0.16 & 0.68 & 0.52 & 0.69 & 0.55 \\
\noindent \textbf{ACE} & \noindent \textbf{0.89} & \noindent \textbf{0.73} & \noindent \textbf{0.93} & \noindent \textbf{0.83} & \noindent \textbf{0.52} & \noindent \textbf{0.35} & \noindent \textbf{0.81} & \noindent \textbf{0.66} & \noindent \textbf{0.99} & \noindent \textbf{0.93} & \noindent \textbf{0.79} & \noindent \textbf{0.59} & 0.44 & 0.38 & \noindent \textbf{0.92} & \noindent \textbf{0.78} & \noindent \textbf{0.79} & \noindent \textbf{0.66} \\
\hline 
 \multicolumn{19}{c}{\emph{JULE}: Silhouette score (Euclidean distance)} \\
 \hline 
Raw score & 0.81 & 0.62 & 0.85 & 0.70 & 0.07 & 0.04 & 0.71 & 0.53 & 0.32 & 0.29 & -0.45 & -0.32 & -0.13 & -0.05 & 0.23 & 0.15 & 0.30 & 0.24 \\
Paired score & 0.27 & 0.20 & 0.72 & 0.55 & 0.04 & 0.03 & 0.56 & 0.41 & 0.42 & 0.30 & 0.70 & 0.50 & \noindent \textbf{0.64} & \noindent \textbf{0.46} & 0.55 & 0.41 & 0.49 & 0.36 \\
Pooled score & 0.71 & 0.58 & \noindent \textbf{0.90} & \noindent \textbf{0.77} & 0.41 & \noindent \textbf{0.28} & 0.78 & 0.63 & 0.96 & 0.85 & 0.79 & 0.57 & 0.26 & 0.16 & 0.70 & 0.54 & 0.69 & 0.55 \\
\noindent \textbf{ACE} & \noindent \textbf{0.88} & \noindent \textbf{0.72} & 0.89 & 0.75 & \noindent \textbf{0.42} & \noindent \textbf{0.28} & \noindent \textbf{0.81} & \noindent \textbf{0.65} & \noindent \textbf{0.98} & \noindent \textbf{0.90} & \noindent \textbf{0.88} & \noindent \textbf{0.70} & 0.41 & 0.36 & \noindent \textbf{0.92} & \noindent \textbf{0.78} & \noindent \textbf{0.77} & \noindent \textbf{0.64} \\
\hline 
 \multicolumn{19}{c}{\emph{DEPICT}: Calinski-Harabasz index} \\
 \hline 
Raw score & -0.05 & -0.10 & \noindent \textbf{0.73} & \noindent \textbf{0.62} & 0.43 & 0.25 & 0.43 & 0.35 & -0.95 & -0.83 &  &  &  &  &  &  & 0.12 & 0.06 \\
Paired score & 0.76 & 0.57 & 0.44 & 0.26 & 0.76 & 0.57 & 0.89 & 0.72 & 0.49 & 0.44 &  &  &  &  &  &  & 0.67 & 0.51 \\
Pooled score & \noindent \textbf{0.96} & \noindent \textbf{0.83} & 0.53 & 0.41 & 0.90 & 0.77 & \noindent \textbf{0.96} & \noindent \textbf{0.87} & 0.61 & 0.56 &  &  &  &  &  &  & 0.79 & 0.69 \\
\noindent \textbf{ACE} & 0.91 & 0.77 & 0.56 & 0.44 & \noindent \textbf{0.94} & \noindent \textbf{0.82} & \noindent \textbf{0.96} & \noindent \textbf{0.87} & \noindent \textbf{0.96} & \noindent \textbf{0.87} &  &  &  &  &  &  & \noindent \textbf{0.87} & \noindent \textbf{0.75} \\
\hline 
 \multicolumn{19}{c}{\emph{DEPICT}: Davies-Bouldin index} \\
 \hline 
Raw score & 0.05 & -0.10 & 0.63 & 0.48 & 0.48 & 0.32 & -0.01 & -0.03 & -0.14 & -0.18 &  &  &  &  &  &  & 0.20 & 0.10 \\
Paired score & 0.81 & 0.59 & 0.45 & 0.31 & 0.90 & 0.74 & 0.89 & 0.72 & 0.63 & 0.59 &  &  &  &  &  &  & 0.73 & 0.59 \\
Pooled score & \noindent \textbf{0.96} & \noindent \textbf{0.88} & 0.49 & 0.35 & 0.64 & 0.48 & 0.43 & 0.32 & -0.77 & -0.61 &  &  &  &  &  &  & 0.35 & 0.28 \\
\noindent \textbf{ACE} & 0.91 & 0.82 & \noindent \textbf{0.76} & \noindent \textbf{0.58} & \noindent \textbf{0.91} & \noindent \textbf{0.79} & \noindent \textbf{0.96} & \noindent \textbf{0.87} & \noindent \textbf{0.98} & \noindent \textbf{0.92} &  &  &  &  &  &  & \noindent \textbf{0.90} & \noindent \textbf{0.80} \\
\hline 
 \multicolumn{19}{c}{\emph{DEPICT}: Silhouette score (Cosine distance)} \\
 \hline 
Raw score & 0.37 & 0.29 & 0.68 & 0.53 & 0.68 & 0.54 & 0.80 & 0.60 & 0.46 & 0.32 &  &  &  &  &  &  & 0.60 & 0.46 \\
Paired score & 0.81 & 0.62 & 0.45 & 0.33 & 0.90 & 0.75 & 0.89 & 0.72 & 0.77 & 0.58 &  &  &  &  &  &  & 0.76 & 0.60 \\
Pooled score & 0.96 & 0.86 & 0.68 & \noindent \textbf{0.56} & \noindent \textbf{0.94} & \noindent \textbf{0.82} & \noindent \textbf{0.97} & \noindent \textbf{0.90} & 0.93 & 0.79 &  &  &  &  &  &  & 0.90 & 0.78 \\
\noindent \textbf{ACE} & \noindent \textbf{0.97} & \noindent \textbf{0.90} & \noindent \textbf{0.71} & \noindent \textbf{0.56} & \noindent \textbf{0.94} & \noindent \textbf{0.82} & \noindent \textbf{0.97} & \noindent \textbf{0.90} & \noindent \textbf{0.94} & \noindent \textbf{0.83} &  &  &  &  &  &  & \noindent \textbf{0.91} & \noindent \textbf{0.80} \\
\hline 
 \multicolumn{19}{c}{\emph{DEPICT}: Silhouette score (Euclidean distance)} \\
 \hline 
Raw score & 0.50 & 0.36 & \textbf{0.76} & \textbf{0.61} & 0.57 & 0.41 & 0.74 & 0.59 & -0.21 & -0.12 &  &  &  &  &  &  & 0.47 & 0.37 \\
Paired score & 0.73 & 0.50 & 0.47 & 0.36 & 0.79 & 0.65 & 0.86 & 0.69 & 0.59 & 0.52 &  &  &  &  &  &  & 0.69 & 0.54 \\
Pooled score & 0.96 & 0.86 & 0.65 & 0.53 & 0.94 & 0.82 & 0.97 & \noindent \textbf{0.90} & 0.92 & 0.75 &  &  &  &  &  &  & 0.89 & 0.77 \\
\noindent \textbf{ACE} & \noindent \textbf{0.97} & \noindent \textbf{0.88} & 0.65 & 0.50 & \noindent \textbf{0.95} & \noindent \textbf{0.83} & \noindent \textbf{0.98} & \noindent \textbf{0.90} & \noindent \textbf{0.94} & \noindent \textbf{0.82} &  &  &  &  &  &  & \noindent \textbf{0.90} & \noindent \textbf{0.79} \\
\bottomrule
\end{tabular}
}
\label{tab:nmi:hyper} 
\end{table}

\begin{figure}[htbp]
\centering
\subfigure[\scriptsize  Selected \emph{JULE}/Silhouette score (cosine distance)]{\includegraphics[width = 0.35\linewidth]{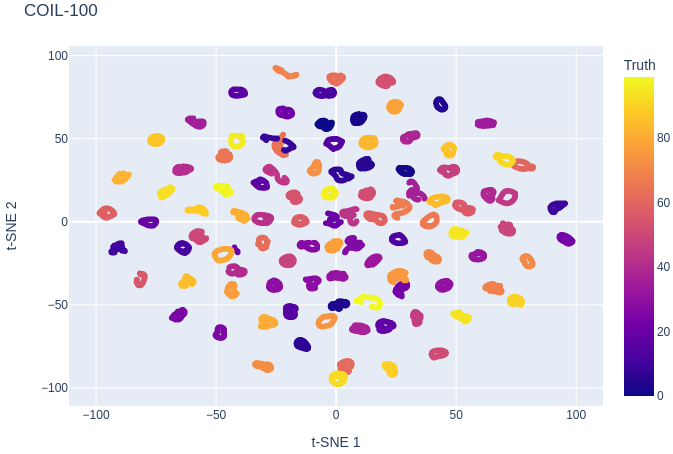}}
\subfigure[\scriptsize  Excluded \emph{JULE}/Silhouette score (cosine distance)]{\includegraphics[width = 0.35\linewidth]{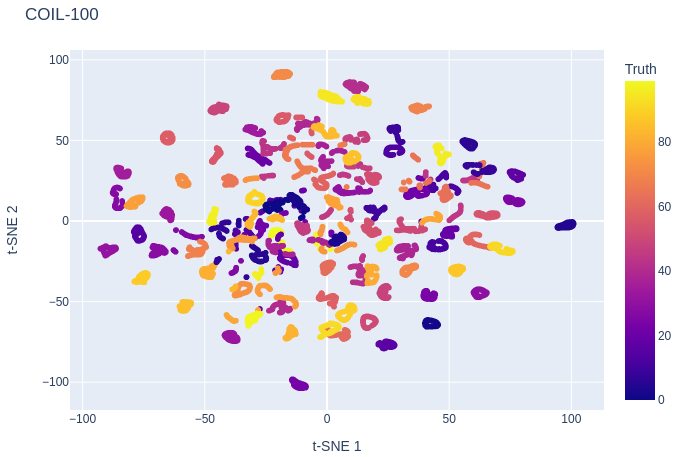}}\\
\subfigure[\scriptsize  Selected \emph{DEPICT}/Calinski-Harabasz index]{\includegraphics[width = 0.35\linewidth]{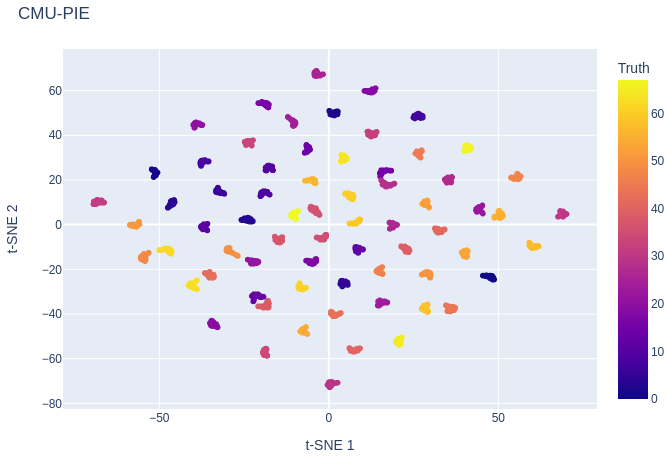}}
\subfigure[\scriptsize  Excluded \emph{DEPICT}/Calinski-Harabasz index]{\includegraphics[width = 0.35\linewidth]{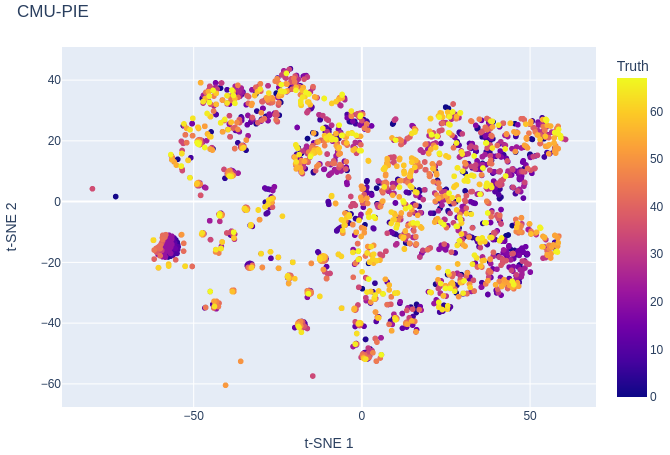}}
\caption{\footnotesize t-SNE visualizes the embedded data from ACE-selected embedding spaces (Left) compared to those excluded (Right) in Application 1, with each color indicating a true cluster label.}
\label{fig:main}
\end{figure}

%%%%%%%%%%%%%%%%%%%%%%%%%%%%%%%%%%%%%%%%%%%%%%%%%%%%%%%%%%%%%%%
{\it (1) Superior performance of ACE.} The proposed ACE consistently outperforms the other three evaluation approaches across different internal measures. Additionally, we utilize t-SNE \citep{van2008visualizing} to visualize the embedded data from ACE-selected embedding spaces in two dimensions. As shown in Figure \ref{fig:main}, the selected spaces display more compact and well-separated clusters of data points aligned with their true cluster labels. In contrast, many excluded spaces have no clear boundaries for different clusters, and points are less compact in a cluster. More examples and discussions of the t-SNE visualizations are available in SM Section S8.2.

{\it (2) Curse of dimensionality.} The raw score performs worse than the other three scores that are computed in low-dimensional embedding spaces. As suggested by Remark \ref{rem:high}, internal measures become unreliable in high-dimensional spaces.

{\it (3) Inconsistency of the paired score.}  Furthermore, consistent with Theorem \ref{thm2}, we observe the lower performance of the paired score compared to the pooled and ACE scores.
 
{\it (4) Choice of internal measure.} Internal measures affect the relative performance of the raw score, the paired score, and the pooled score. For example, the pooled score outperforms the paired score for the Calinski-Harabasz index, but on the contrary, for the Davies-Bouldin index. Additionally, compared to the other two measures, the improvement from the raw and paired scores to the pooled score and ACE is more pronounced for the Calinski-Harabasz index. To further investigate the differences among internal measures, we visualize the rank correlations of selected spaces after ACE's multimodality Test (Step 1) using an undirected graph (Figures S4-S18), highlighting notable structural variations. These differing behaviors are likely due to the distinct focuses of each measure: the Silhouette score emphasizes individual data points and their relationships, the Davies-Bouldin index considers overall compactness and separation, and the Calinski-Harabasz index measures variance ratios. As mentioned in Remark \ref{rem4}, a bad choice of the internal measure $\pi$ may lead to the infeasibility of comparing different deep clustering results.

{\it (5) Choice of distance metrics.} Similar performances are observed between the Silhouette score with Cosine distance and that with Euclidean distance, suggesting that the choice of distance metric in the Silhouette score may not significantly impact the ranking results. We emphasize the importance of selecting a good internal measure over the distance metric selection.

\vspace{-6mm}
\subsection{Application 2: Evaluating the Number of Clusters}
\vspace{-2mm}

We consider the problem of evaluating different values of \(K\) (i.e., the number of clusters) and selecting the optimal \(K\) in deep clustering. The clustering outputs are obtained by applying the same deep clustering model (JULE or DEPICT) across a range of \(K\) values. Specifically, we consider \(10\) evenly distributed values of \(K\), ensuring that the true \(K\) (or a value close to it) is included within the evaluation range. The \(K\) values for each dataset, along with further implementation details, are provided in SM Section S5.2.

As shown in Tables \ref{tab:nmi:det} and S4, ACE consistently outperforms the other methods in most cases. The conclusions drawn from the three common measures align with conclusions (1)–(5) from Application 1. While broadly generalizable, exceptions arose in both applications with less common internal measures (SM Sections S8.1 and S8.3), particularly when low or negative rank correlations indicated limitations of the internal measure itself. Notably, our applications evaluates high-dimensional image data, a key deep clustering application, where raw scores perform poorly. However, this may not extend to other high-dimensional data like genomic data. Our simulations capturing the high-dimensionality but lacking the spatial structure of image data show different raw score performance.

t-SNE visualizations of embedded data are shown in Figures S23-S37 and discussed in SM Section S8.5. Similar to Application 1, Figures S22-S36 depict the rank correlations of selected spaces after multimodality test. Unlike Application 1, where variations in grouping behavior of spaces are evident, there is a stronger tendency for spaces to form a single group. This is because we only have a candidate set of \(M=10\) embedding spaces. Hence, to have a practical evaluation, we suggest \(M \ge 10\) when applying ACE. 

\begin{figure}[htbp!]
\centering
\subfigure[\scriptsize  JULE:CMU-PIE]{\includegraphics[width = 0.35\linewidth]{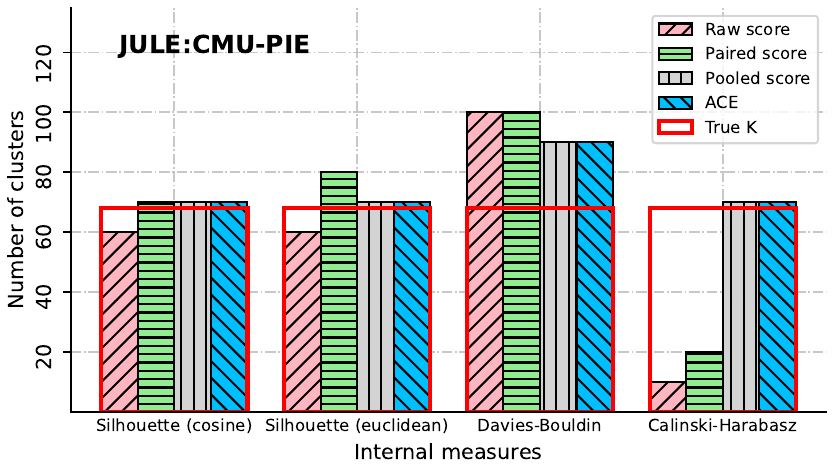}}
\subfigure[\scriptsize  JULE:YTF]{\includegraphics[width = 0.35\linewidth]{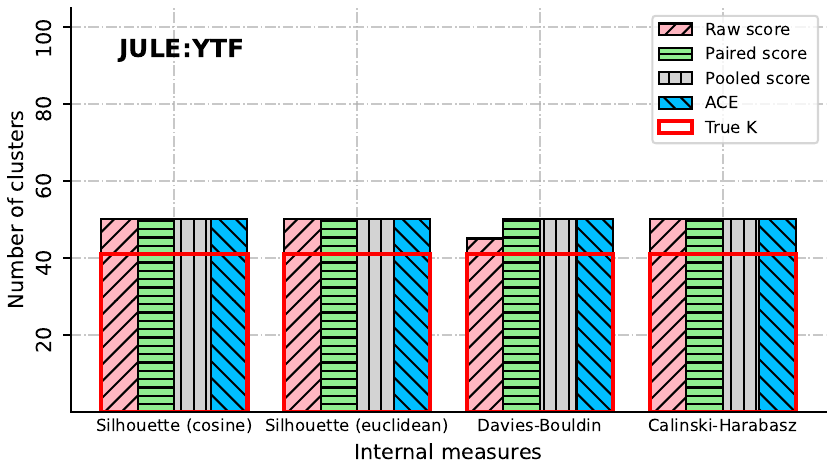}} \\
\subfigure[\scriptsize  DEPICT:CMU-PIE]{\includegraphics[width = 0.35\linewidth]{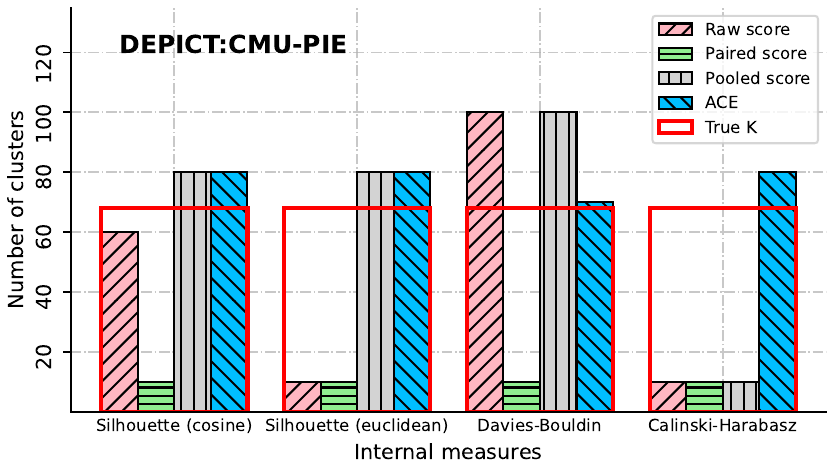}} 
\subfigure[\scriptsize  DEPICT:YTF]{\includegraphics[width = 0.35\linewidth]{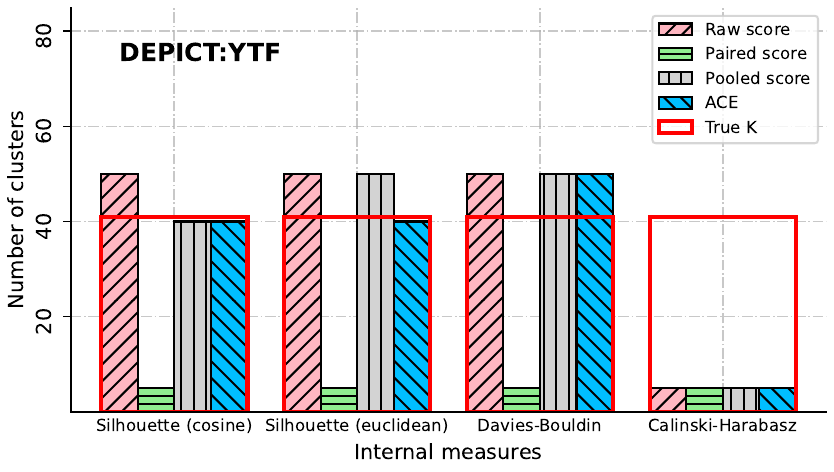}}
\caption{\footnotesize Bar plots of the optimal \( K \) detected by each evaluation approach. The true \( K \) is marked by a red, outlined, hollow box, while the estimated \( \hat{K} \) values are represented as solid boxes with hatching.}
\label{app:fig:ncl}
\end{figure}

We present the optimal number of clusters (\( K \)) identified by each evaluation approach for JULE and DEPICT across the CMU-PIE and YTF datasets in Figure~\ref{app:fig:ncl}. In DEPICT, ACE suggests $\hat{K}=40$ and $\hat{K}=50$ for several indices for YTF, which are closer to the true $K=41$, while the paired score indicates $\hat{K}=5$. Similarly, for CMU-PIE with true $K=68$, ACE selects $\hat{K}=70$ or $\hat{K}=80$, closer to the truth, while the paired score suggests very low values like $\hat{K}=10$. In JULE, when using the Calinski-Harabasz index, both ACE and pooled scores select $\hat{K}=70$. In contrast, the raw score and the paired score respectively suggest $\hat{K}=10$ and $\hat{K}=20$, significantly lower than the true $K$ for CMU-PIE. Additional results for the optimal \( K \) identified by each evaluation approach across all tested datasets are provided in Figures S20 and S21 (SM Section S8.4). It is worth noting that the primary goal of ACE is to provide a comprehensive ranking of clustering performance for candidate results rather than identifying the most accurate \(\hat{K}\). However, developing an evaluation approach dedicated to selecting \(K\) presents an interesting avenue for future research.

\begin{table}[h]
 \caption{\small 
 Rank consistency with NMI for different evaluation approaches in Application 2. For rank consistency with ACC, refer to Table S4.}
\centering 
\resizebox{1\textwidth}{!}{ %
\begin{tabular}{lllllllllllllllllll}
\toprule
 & \multicolumn{2}{l}{USPS} & \multicolumn{2}{l}{YTF} & \multicolumn{2}{l}{FRGC} & \multicolumn{2}{l}{MNIST$_{\text{test}}$} & \multicolumn{2}{l}{CMU-PIE} & \multicolumn{2}{l}{UMist} & \multicolumn{2}{l}{COIL-20} & \multicolumn{2}{l}{COIL-100} & \multicolumn{2}{l}{Average} \\
 & $r_s$ & $\tau_B$ & $r_s$ & $\tau_B$ & $r_s$ & $\tau_B$ & $r_s$ & $\tau_B$ & $r_s$ & $\tau_B$ & $r_s$ & $\tau_B$ & $r_s$ & $\tau_B$ & $r_s$ & $\tau_B$ & $r_s$ & $\tau_B$ \\
\midrule
\hline 
 \multicolumn{19}{c}{\emph{JULE}: Calinski-Harabasz index} \\
 \hline 
Raw score & 0.44 & 0.56 & \noindent \textbf{0.95} & \noindent \textbf{0.89} & -0.93 & -0.83 & 0.43 & 0.51 & -0.37 & -0.24 & -0.33 & -0.24 & 0.74 & 0.64 & 0.53 & 0.47 & 0.18 & 0.22 \\
Paired score & \noindent \textbf{0.65} & \noindent \textbf{0.64} & 0.1 & 0.06 & -0.93 & -0.83 & \noindent \textbf{0.64} & \noindent \textbf{0.6} & -0.03 & -0.02 & \noindent \textbf{-0.13} & \noindent \textbf{-0.07} & \noindent \textbf{0.76} & \noindent \textbf{0.71} & 0.74 & 0.56 & 0.22 & 0.21 \\
Pooled score & \noindent \textbf{0.65} & \noindent \textbf{0.64} & 0.9 & 0.78 & -0.87 & -0.72 & \noindent \textbf{0.64} & \noindent \textbf{0.6} & \noindent \textbf{0.9} & \noindent \textbf{0.73} & -0.14 & -0.11 & 0.74 & 0.64 & 0.72 & 0.64 & 0.44 & 0.40 \\
\noindent \textbf{ACE} & \noindent \textbf{0.65} & \noindent \textbf{0.64} & 0.93 & 0.83 & \noindent \textbf{-0.72} & \noindent \textbf{-0.67} & \noindent \textbf{0.64} & \noindent \textbf{0.6} & 0.88 & \noindent \textbf{0.73} & -0.14 & -0.11 & 0.74 & 0.64 & \noindent \textbf{0.79} & \noindent \textbf{0.69} & \noindent \textbf{0.47} & \noindent \textbf{0.42} \\
\hline 
 \multicolumn{19}{c}{\emph{JULE}: Davies-Bouldin index} \\
 \hline 
Raw score & -0.27 & -0.29 & \noindent \textbf{0.92} & \noindent \textbf{0.78} & \noindent \textbf{0.87} & \noindent \textbf{0.72} & -0.46 & -0.42 & 0.72 & 0.47 & \noindent \textbf{0.19} & \noindent \textbf{0.16} & -0.88 & -0.79 & -0.92 & -0.82 & 0.02 & -0.02 \\
Paired score & 0.54 & 0.38 & 0.15 & 0.17 & 0.85 & 0.67 & 0.43 & 0.29 & 0.78 & 0.56 & -0.08 & 0.02 & \noindent \textbf{-0.26} & \noindent \textbf{-0.14} & \noindent \textbf{-0.9} & \noindent \textbf{-0.78} & 0.19 & 0.15 \\
Pooled score & \noindent \textbf{0.98} & \noindent \textbf{0.91} & 0.83 & 0.67 & 0.82 & 0.61 & \noindent \textbf{0.79} & \noindent \textbf{0.6} & 0.82 & 0.64 & -0.21 & -0.02 & -0.76 & -0.57 & -0.92 & -0.82 & 0.29 & 0.25 \\
\noindent \textbf{ACE} & \noindent \textbf{0.98} & \noindent \textbf{0.91} & 0.83 & 0.67 & 0.87 & 0.72 & \noindent \textbf{0.79} & \noindent \textbf{0.6} & \noindent \textbf{0.85} & \noindent \textbf{0.69} & -0.21 & -0.02 & -0.69 & -0.57 & -0.94 & -0.82 & \noindent \textbf{0.31} & \noindent \textbf{0.27} \\
\hline 
 \multicolumn{19}{c}{\emph{JULE}: Silhouette score (cosine distance)} \\
 \hline 
Raw score & 0.69 & 0.51 & \noindent \textbf{1.0} & \noindent \textbf{1.0} & 0.67 & 0.5 & 0.07 & 0.02 & -0.28 & -0.11 & \noindent \textbf{0.13} & \noindent \textbf{0.07} & -0.52 & -0.43 & 0.42 & 0.24 & 0.27 & 0.23 \\
Paired score & \noindent \textbf{0.99} & \noindent \textbf{0.96} & 0.3 & 0.22 & \noindent \textbf{0.72} & \noindent \textbf{0.61} & 0.87 & 0.69 & \noindent \textbf{0.98} & \noindent \textbf{0.91} & -0.07 & 0.07 & 0.52 & 0.36 & 0.39 & 0.2 & 0.59 & 0.50 \\
Pooled score & 0.95 & 0.87 & 0.98 & 0.94 & 0.68 & 0.56 & \noindent \textbf{0.96} & \noindent \textbf{0.87} & \noindent \textbf{0.98} & \noindent \textbf{0.91} & -0.07 & -0.02 & 0.71 & \noindent \textbf{0.57} & 0.41 & 0.24 & 0.70 & 0.62 \\
\noindent \textbf{ACE} & 0.95 & 0.87 & 0.98 & 0.94 & 0.7 & 0.61 & \noindent \textbf{0.96} & \noindent \textbf{0.87} & \noindent \textbf{0.98} & \noindent \textbf{0.91} & -0.07 & -0.02 & \noindent \textbf{0.74} & 0.5 & \noindent \textbf{0.46} & \noindent \textbf{0.33} & \noindent \textbf{0.71} & \noindent \textbf{0.63} \\
\hline 
 \multicolumn{19}{c}{\emph{JULE}: Silhouette score (Euclidean distance)} \\
 \hline 
Raw score & 0.56 & 0.47 & \noindent \textbf{1.0} & \noindent \textbf{1.0} & -0.18 & -0.17 & 0.61 & 0.47 & 0.55 & 0.38 & \noindent \textbf{0.19} & \noindent \textbf{0.16} & -0.41 & -0.36 & 0.39 & 0.2 & 0.34 & 0.27 \\
Paired score & 0.85 & 0.73 & 0.33 & 0.28 & 0.72 & 0.61 & 0.88 & 0.69 & 0.96 & 0.87 & 0.07 & 0.16 & 0.55 & 0.43 & 0.44 & 0.29 & 0.60 & 0.51 \\
Pooled score & \noindent \textbf{0.95} & \noindent \textbf{0.87} & 0.97 & 0.89 & 0.68 & 0.56 & \noindent \textbf{0.95} & \noindent \textbf{0.82} & \noindent \textbf{0.98} & \noindent \textbf{0.91} & 0.14 & 0.11 & \noindent \textbf{0.76} & \noindent \textbf{0.57} & \noindent \textbf{0.47} & \noindent \textbf{0.33} & \noindent \textbf{0.74} & 0.63 \\
\noindent \textbf{ACE} & \noindent \textbf{0.95} & \noindent \textbf{0.87} & 0.98 & 0.94 & \noindent \textbf{0.78} & \noindent \textbf{0.67} & \noindent \textbf{0.95} & \noindent \textbf{0.82} & \noindent \textbf{0.98} & \noindent \textbf{0.91} & 0.14 & 0.11 & 0.71 & 0.43 & \noindent \textbf{0.47} & \noindent \textbf{0.33} & \noindent \textbf{0.74} & \noindent \textbf{0.64} \\
\hline 
 \multicolumn{19}{c}{\emph{DEPICT}: Calinski-Harabasz index} \\
 \hline 
Raw score & \noindent \textbf{0.46} & \noindent \textbf{0.6} & -0.69 & -0.56 & -0.88 & -0.78 & \noindent \textbf{0.46} & \noindent \textbf{0.6} & -0.92 & -0.82 &  &  &  &  &  &  & -0.31 & -0.19 \\
Paired score & \noindent \textbf{0.46} & \noindent \textbf{0.6} & -0.99 & -0.96 & -0.85 & -0.72 & 0.44 & 0.56 & -0.92 & -0.82 &  &  &  &  &  &  & -0.37 & -0.27 \\
Pooled score & \noindent \textbf{0.46} & \noindent \textbf{0.6} & -0.98 & -0.91 & -0.85 & -0.72 & \noindent \textbf{0.46} & \noindent \textbf{0.6} & 0.44 & 0.56 &  &  &  &  &  &  & -0.09 & 0.03 \\
\noindent \textbf{ACE} & \noindent \textbf{0.46} & \noindent \textbf{0.6} & \noindent \textbf{-0.66} & \noindent \textbf{-0.51} & \noindent \textbf{0.77} & \noindent \textbf{0.61} & \noindent \textbf{0.46} & \noindent \textbf{0.6} & \noindent \textbf{0.92} & \noindent \textbf{0.82} &  &  &  &  &  &  & \noindent \textbf{0.39} & \noindent \textbf{0.42} \\
\hline 
 \multicolumn{19}{c}{\emph{DEPICT}: Davies-Bouldin index} \\
 \hline 
Raw score & -0.39 & -0.42 & \noindent \textbf{0.99} & \noindent \textbf{0.96} & 0.68 & 0.39 & -0.22 & -0.16 & 0.92 & 0.82 &  &  &  &  &  &  & 0.40 & 0.32 \\
Paired score & 0.46 & \noindent \textbf{0.6} & -0.78 & -0.64 & -0.85 & -0.72 & 0.44 & 0.56 & -0.1 & 0.02 &  &  &  &  &  &  & -0.17 & -0.04 \\
Pooled score & 0.6 & 0.51 & 0.88 & 0.73 & -0.13 & -0.17 & 0.74 & 0.64 & 0.92 & 0.82 &  &  &  &  &  &  & 0.60 & 0.51 \\
\noindent \textbf{ACE} & \noindent \textbf{0.62} & 0.6 & 0.95 & 0.87 & \noindent \textbf{0.77} & \noindent \textbf{0.67} & \noindent \textbf{0.78} & \noindent \textbf{0.69} & \noindent \textbf{0.96} & \noindent \textbf{0.91} &  &  &  &  &  &  & \noindent \textbf{0.82} & \noindent \textbf{0.75} \\
\hline 
 \multicolumn{19}{c}{\emph{DEPICT}: Silhouette score (cosine distance)} \\
 \hline 
Raw score & -0.13 & -0.11 & \noindent \textbf{1.0} & \noindent \textbf{1.0} & \noindent \textbf{0.97} & \noindent \textbf{0.89} & 0.71 & 0.56 & -0.43 & -0.33 &  &  &  &  &  &  & 0.42 & 0.40 \\
Paired score & 0.44 & 0.56 & -0.7 & -0.6 & -0.85 & -0.72 & 0.44 & 0.56 & 0.07 & 0.11 &  &  &  &  &  &  & -0.12 & -0.02 \\
Pooled score & 0.6 & 0.51 & 0.61 & 0.47 & 0.07 & 0.06 & 0.71 & 0.64 & 0.98 & 0.91 &  &  &  &  &  &  & 0.59 & 0.52 \\
\noindent \textbf{ACE} & \noindent \textbf{0.65} & \noindent \textbf{0.64} & 0.87 & 0.78 & 0.93 & 0.83 & \noindent \textbf{0.85} & \noindent \textbf{0.78} & \noindent \textbf{0.99} & \noindent \textbf{0.96} &  &  &  &  &  &  & \noindent \textbf{0.86} & \noindent \textbf{0.80} \\
\hline 
 \multicolumn{19}{c}{\emph{DEPICT}: Silhouette score (Euclidean distance)} \\
 \hline 
Raw score & -0.34 & -0.29 & \noindent \textbf{1.0} & \noindent \textbf{1.0} & \noindent \textbf{0.3} & \noindent \textbf{0.11} & 0.39 & 0.33 & -0.43 & -0.33 &  &  &  &  &  &  & 0.18 & 0.16 \\
Paired score & 0.44 & 0.56 & -0.61 & -0.47 & -0.85 & -0.72 & 0.44 & 0.56 & -0.12 & -0.02 &  &  &  &  &  &  & -0.14 & -0.02 \\
Pooled score & \noindent \textbf{0.6} & 0.51 & 0.98 & 0.91 & 0.07 & 0.06 & 0.73 & 0.69 & \noindent \textbf{0.99} & \noindent \textbf{0.96} &  &  &  &  &  &  & \noindent \textbf{0.67} & 0.63 \\
\noindent \textbf{ACE} & 0.46 & \noindent \textbf{0.6} & 0.94 & 0.87 & 0.02 & 0.06 & \noindent \textbf{0.85} & \noindent \textbf{0.78} & 0.98 & 0.91 &  &  &  &  &  &  & 0.65 & \noindent \textbf{0.64} \\
\bottomrule
\end{tabular}
}
\label{tab:nmi:det} 
\end{table}
%%%%%%%%%%%%%%%%%%%%%%%%%%%%%%%%%%%%%%%%%%%%%%%%%%%%%%
\vspace{-6mm}
\subsection{Perturbation Analysis}
\vspace{-4mm}
We perform a perturbation analysis to Applications 1 and 2 to demonstrate how sensitive ACE is to the algorithmic change (e.g., removing Step 1, adjusting parameter values, or using an alternative link analysis algorithm). Below, we summarize the key findings, with a detailed analysis available in SM Section S9.

\noindent\textbf{(1) Multimodality test.} Including the multimodality test in ACE enhances performance in certain tasks, although its effect on the pooled score is minimal. For further details, refer to Tables S8-S11 in SM Section S9.1.

\noindent\textbf{(2) Significance level, $\alpha_{edges}$.}  ACE is robust across varying levels of $\alpha_{edges}$. We examined $\alpha_{edges} = 0.05$ and $\alpha_{edges} = 0.1$ used for constructing the graph’s edge set and observed no significant performance differences. While performance is generally similar when including all positive edges, it declined in certain scenarios, highlighting the necessity of the testing procedure. Details are available in SM Section S9.2.

\noindent\textbf{(3) Density-based clustering methods.} We compared HDBSCAN, used for space screening and grouping in our algorithm, with DBSCAN \citep{ester1996density}. Both methods yielded comparable evaluation performance (see Tables S16-S19). HDBSCAN was chosen for its simplicity in implementation.  Additional details are in SM Section S9.3.

\noindent\textbf{(4) Link analysis algorithms.}  We compared HITS with PageRank for ensemble analysis. As shown in Tables S20-S23, both algorithms exhibit very similar performance, with PageRank performing slightly better. This result highlights the robustness of ACE to the choice of link analysis algorithm. Additional details are in SM Section S9.4.

%likely due to its simultaneous consideration of both incoming and outgoing links.
These findings provide deeper insights into the impact of different components on ACE's performance. Overall, ACE is robust to parameter choices and algorithmic variations.

%%%%%%%%%%%%%%%%%%%%%%%%%%%%%%%%%%%%%%%%%%%%%%%%%%%%%%
\vspace{-8mm}
\section{Discussion and Future Work}\label{sec:5}
\vspace{-4mm}
This paper reveals crucial yet overlooked challenges in deep clustering evaluation and addresses them with a theoretical framework that revisits the limitations of traditional approaches. With theoretical intuitions and justifications, we propose an adaptive evaluation strategy, named ACE, by identifying admissible embedding spaces and adaptively aggregating calculated internal measures from these spaces. Extensive experiments demonstrate the effectiveness of ACE across different scenarios. ACE is computationally lightweight. In our experiments, it was implemented on a machine equipped with 12 Intel Core i7-6800K CPUs and 128 GB of RAM (see SM Section S5.2 for additional experimental computational details). ACE is straightforward and user-friendly, requiring minimal parameter tuning (only the significance levels for the multiple testing procedures need to be predefined), which can be applied to large-scale AI models and datasets. Once internal measures across different embedding spaces are computed, ACE can evaluate the dataset within seconds. Moreover, these computations of internal measures are independent, enabling seamless parallel execution. Together, these features ensure ACE's efficiency and scalability to meet the demands of modern large-scale AI systems. The source code and implementation of ACE are publicly available at \url{https://github.com/ZWStatLab/ACE}.

ACE enhances clustering evaluation and has broad applicability in deep clustering studies. For instance, many studies \citep{zhao2020joint, li2023attention, rohani2020classifying, cen2022financial, hwang2023identifying} that utilize deep clustering techniques depend on internal validation for comparing clustering approaches, optimizing model configurations, and evaluating the number of clusters. By improving evaluation reliability, ACE enables more precise cluster selection and model tuning, potentially leading to better subgrouping in applications like patient stratification and cell type detection. This improved precision could contribute to enhanced treatment strategies and a reduced risk of adverse reactions. 

It is worth mentioning that ACE can be generalized to broader evaluation scenarios beyond deep clustering. (1) In many unsupervised problems, traditional machine learning techniques struggle with high-dimensional data, leading to the use of dimensionality reduction methods or deep networks for extracting lower-dimensional representations. A common practice is to leverage pre-trained encoders (e.g., ResNet, CLIP embeddings, or autoencoders) to obtain low-dimensional embeddings before applying clustering techniques (e.g., k-means or graph clustering) for efficiency and scalability. Unlike deep clustering, this two-step approach requires selecting a pre-trained model. Similar evaluation challenges may arise as clustering results from embeddings of different pre-trained models may not be directly comparable. In this context, ACE can help identify the pre-trained model (whether different models or the same model with varying hyperparameters) that yields the most reliable clustering results. (2) ACE also has potential applications in classification tasks under a transfer learning setup, particularly in unsupervised domain adaptation, where substantial distribution shifts occur, and the target domain lacks labeled data. Despite extensive development of methods in this area, model validation and selection remain difficult due to the absence of target domain labels \citep{zhuang2020comprehensive,zhou2022domain}. Given that these tasks also produce estimated labels and embedded data, extending ACE to this context could provide a valuable solution to these persistent evaluation issues. (3) Deep anomaly detection is another challenging area, where many deep learning-based methods now outperform traditional detection approaches in real-world applications \citep{pang2021deep}. Many of these methods jointly learn anomaly scores and feature embeddings \citep{pang2021deep}, creating evaluation challenges similar to those in deep clustering. Fully unsupervised evaluation approaches for traditional anomaly detection rely on internal evaluations of the solution based solely on the model output and input data \citep{idan2024towards,marques2020internal,goix2016evaluate,marques2015internal}. However, these conventional evaluation methods may not be adequate for complex deep anomaly detection models. Our proposed method, therefore, could potentially be adapted to tackle these specific evaluation challenges as well.

One limitation of ACE is its reliance on the existence of admissible spaces. When few or no admissible spaces exist (see discussion in SM Sections S8.6 and S9.5), ACE may fail. In such cases, the pooled score serves as a practical alternative, particularly when $M$ is small, or when no spaces are retained after the multimodality test. Practitioners are also encouraged to leverage empirical knowledge and exploratory data visualization techniques (e.g., t-SNE) to decide which spaces to include. Our t-SNE analysis (SM Sections S8.2 and S8.5) suggests that effective spaces typically feature compact and well-separated clusters. Another limitation of ACE is that it does not handle the selection of internal measures, which remains a potential direction for future work.\\

\def\spacingset#1{\renewcommand{\baselinestretch}%
{#1}\small\normalsize} \spacingset{1}

\renewcommand{\thefigure}{S\arabic{figure}} 
\setcounter{figure}{0} 
\renewcommand{\thetable}{S\arabic{table}} 
\setcounter{table}{0} 
\renewcommand{\thesection}{S\arabic{section}}
\setcounter{section}{0}  
\renewcommand{\thealgorithm}{S\arabic{algorithm}}
\setcounter{algorithm}{0}

%%%%%%%%%%%%%%%%%%%%%%%%%%%%%%%%%%%%%%%%%%%%%%%%%%%%%%%%%%%%%%%%%%%%%%%%%%%%%%

  \bigskip
  \bigskip
  \bigskip
  \begin{center}
    {\LARGE\bf  Deep Clustering Evaluation: How to Validate Internal Clustering Validation Measures}\\
    \bigskip
    \hspace{0.5cm}
        {\LARGE\bf Supplementary Material}
\end{center}
  \medskip

In the Supplementary Material, we provide additional details regarding the metrics, technical proofs, supplementary algorithmic and implementation information, an overview of the deep clustering algorithms used in our evaluation, further experimental results, and extended discussion, which were omitted from the main paper due to space constraints.

\tableofcontents

%%%%%%%%%%%%%%%%%%%%%%%%%%%%%%%%%%%%%%%%%%%%%%%%%%%%
\section{External Validation Measures} \label{app:evm}
 \subsection{Normalized mutual information \citep{vinh2009information}} Normalized Mutual Information (NMI) is a widely adopted metric for gauging the similarity between two cluster assignments, denoted by sets $A$ and $B$. It can be computed using the formula: 
\begin{equation}
    NMI(A;B) = \frac{2 \times I(A; B)}{H(A) + H(B)}
\end{equation}
where \( I(A, B) \) denotes the mutual information between \( A \) and \( B \), and \( H(A) \) and \( H(B) \) stand for their entropies. The NMI ranges between 0 (indicating no mutual information) and 1 (reflecting perfect correlation). In the context of clustering performance evaluation, when provided with true partition labels denoted as $Y$ and estimated partition labels denoted as $\hat{Y}$, we can leverage $NMI(Y; \hat{Y})$ as a reliable metric. 
 
\subsection{Clustering accuracy \citep{xie2016unsupervised}} Clustering accuracy (ACC) is defined as the proportion of correctly matched pairs resulting from the optimal alignment between true labels and predicted cluster labels. The ACC of $\hat{Y}$ with respect to $Y$ is expressed as:

\begin{equation}
    ACC(Y; \hat{Y}) = \max_{\text{map}\in P}\frac{\sum_{i=1}^{N}I\{y_i=\text{map}(\hat{y}_i)\}}{N}
\end{equation}
where $P$ denotes the set of mappings corresponding to all permutations of partition indices. Like accuracy in classification, clustering ACC computes the ratio of correctly assigned labels to total samples. However, it differs from classification accuracy by accounting for the best one-to-one mapping between predicted partition indices and the ground-truth class set.
%%%%%%%%%%%%%%%%%%%%%%%%%%%%%%%%%%%%%%%%%%%%%%%%%%%%

\section{Internal Validation Measures} \label{app:index}
This section summarizes and provides additional details on the clustering indices referenced in the main paper. These include the Silhouette score\citep{rousseeuw1987silhouettes}, Dunn index \citep{dunn1974well,desgraupes2013clustering},cubic clustering criterion (CCC) \citep{sarle1983sas}, Cindex (CIND) \citep{hubert1976general,desgraupes2013clustering}, Calinski-Harabasz index \citep{calinski1974dendrite,desgraupes2013clustering}, Davies-Bouldin index (DB) \citep{davies1979cluster,desgraupes2013clustering}, SDBW index (SDBW) \citep{halkidi2001clustering,desgraupes2013clustering}, and CDbw index (CDbw) \citep{halkidi2008density}. Readers can refer to the original publications for complete definitions and mathematical formulations.

The data in $\mathbb{R}^{p}$ used for clustering and evaluation purposes is denoted as $x_1, \cdots, x_N$. Let us consider \( \mathbf{C} \) as a partition of the data into $K$ clusters. The index set for the \( k \)-th cluster ($1 \le k \le K$) is denoted as \( C_k \), with its size given by \( n_k \). When referring to two different clusters, we denote them as \( C_k \) and \( C_{k'} \), respectively. Throughout our discussion of internal measures, we follow the notation conventions and definitions established in \cite{desgraupes2013clustering} for most of the introduced indices. Let $\mu^{\{k\}}$ represent the centroid of the observations in cluster $C_{k}$, and let $\mu$ denote the centroid of all observations.
\begin{equation}
\begin{aligned}
\mu^{\{k\}} & =\frac{1}{n_{k}} \sum_{i \in C_{k}} x_{i} \\
\mu & =\frac{1}{N} \sum_{i=1}^{N} x_{i}
\end{aligned}
\end{equation}

\subsection{Silhouette score \citep{rousseeuw1987silhouettes}}
For a chosen distance function \( d(i, j) \) that measures the distance between observations \( i \) and \( j \) (i.e., \( x_i \) and \( x_j \)), let \( a(i) \) denote the mean distance between the \( i \)-th observation and all other observations within the same cluster \( C_k \) to which $i$ belongs.
\begin{equation}
    a(i) = \frac{1}{n_k - 1} \sum_{j \in C_k, j \neq i} d(i, j) 
\end{equation}

For the \( i \)-th observation, let \( b(i) \) denote the minimum mean distance to all observations in any other cluster.

\begin{equation}
    b(i) = \min_{k' \neq k} \frac{1}{n_{k'}} \sum_{j \in C_{k'}} d(i, j)
\end{equation}

Then, the silhouette value for observation \( i \) is defined as: 
\begin{equation}
    s(i) = \frac{b(i) - a(i)}{\max\{a(i),b(i)\}}
\end{equation}

The Silhouette score is defined as the average silhouette value across clusters \citep{desgraupes2013clustering} via their mean values or across observations \citep{malika2014nbclust}: \\
\begin{equation}
    \pi_{Silhouette} = \frac{1}{K} \sum_{k=1}^K \frac1{N_k} \sum_{i \in C_k} s\left(i\right) \quad \text{or} \quad \frac{1}{N} \sum_{i=1}^{N} s(i)
\end{equation}

\subsection{Dunn index \citep{dunn1974well}}

Let $d_{\min}$ represent the minimal distance between observations from different clusters, which is defined as: 
\begin{equation}
d_{\min }=\min _{k \neq k^{\prime}} d_{k k^{\prime}}
\end{equation}

where $d_{k k^{\prime}}=\min_{\substack{i \in C_{k} \\ j \in C_{k^{\prime}}}}\left\|x_{i}-x_{j}\right\|$ denotes the distance between the closest observations of clusters $C_{k}$ and $C_{k^{\prime}}$ (Euclidean distance is shown, but it need not be Euclidean distance).

Let $d_{\max}$ denote the largest within-cluster distance, which is defined as :

\begin{equation}
d_{\max }=\max _{1 \leq k \leq K} d_{k}
\end{equation}

where $d_{k}=\max _{\substack{i, j \in C_{k} \\ i \neq j}}\left\|x_{i}-x_{j}\right\|$ is the largest distance between two distinct observations within the same cluster $C_{k}$.

The Dunn index is defined as the ratio of $d_{\min }$ to $d_{\max }$ :

\begin{equation}
    \pi_{Dunn}=\frac{d_{\min }}{d_{\max }}
\end{equation}

\subsection{Davies-Bouldin index \citep{davies1979cluster}}
Let $\delta_{k}$ denote the average distance of the observations in cluster $C_{k}$ to their centroid $\mu^{\{k\}}$:

\begin{equation}
\delta_{k}=\frac{1}{n_{k}} \sum_{i \in C_{k}}\left\|x_{i}-\mu^{\{k\}}\right\|
\end{equation}

Let \( \Delta_{k k^{\prime}} \) denote the distance between \( \mu^{\{k\}} \), the centroid of cluster \( C_k \), and \( \mu^{\{k^{\prime}\}} \), the centroid of cluster \( C_{k^{\prime}} \).

\begin{equation}
\Delta_{k k^{\prime}}=d\left(\mu^{\{k\}}, \mu^{\left\{k^{\prime}\right\}}\right)=\left\|\mu^{\left\{k^{\prime}\right\}}-\mu^{\{k\}}\right\|
\end{equation}

For each cluster $k$, $M_{k}$ is defined as: 

\begin{equation}
M_{k} = \max _{k^{\prime} \neq k}\left(\frac{\delta_{k}+\delta_{k^{\prime}}}{\Delta_{k k^{\prime}}}\right)
\end{equation}

The Davies-Bouldin index is the average of $M_{k}$ across all the clusters:

\begin{equation}
\pi_{Davies-Bouldin}=\frac{1}{K} \sum_{k=1}^{K} M_{k}
\end{equation}

Unlike the Silhouette score, a lower Davies-Bouldin index indicates better clustering. To align it with external metrics like NMI and clustering accuracy, where higher is better, we negate the output index value.

\subsection{Calinski-Harabasz index \citep{calinski1974dendrite}}

The within-group (cluster) sum of squares (or dispersion) $WGSS^{k}$ is defined as the sum of squared distances between each observation in the cluster $C_k$ and its centroid $\mu^{{k}}$:
\begin{equation}
W G S S^{k}  =\sum_{i \in C_{k}}||x_{i}-\mu^{(k)}||^2 =\frac{1}{n_{k}} \sum_{i<j \in C_{k}}||x_{i}-x_{j}||^2
\end{equation}

The pooled within-group sum of squares (WGSS) is calculated as the total within-cluster sum of squares across all clusters.

\begin{equation}
W G S S=\sum_{k=1}^{K} W G S S^{k}
\end{equation}

Define the between-group sum of squares (BGSS) as the separation of the cluster centers $\mu^{\{k\}}$ relative to the overall dataset center $\mu$.

\begin{equation}
B G S S = \sum_{k=1}^{K} n_{k}\left\|\mu^{\{k\}}-\mu\right\|^{2} 
\end{equation}

The Calinski-Harabasz index is defined as:

\begin{equation}
\pi_{Calinski-Harabasz}=\frac{B G S S /(K-1)}{W G S S /(N-K)}
\end{equation}

\subsection{Cindex \citep{hubert1976general}}
Let \( N_W = \sum_{k=1}^K \frac{n_k(n_k-1)}{2} \) be the total number of observation pairs within the same clusters, and let \( N_T = \frac{N(N-1)}{2} \) represent the total number of observation pairs in the entire dataset. Define \( S_W \) as the sum of the \( N_W \) distances (where \( N_W \) is the number of distances being summed) between all observation pairs within each cluster. Additionally, let \( S_{\min} \) and \( S_{\max} \) denote the sums of the \( N_W \) smallest and \( N_W \) largest distances, respectively, among all \( N_T \) pairs in the dataset.

The $\mathrm{C}$ index is defined as:

\begin{equation}
\pi_{Cindex}=\frac{S_{W}-S_{\min }}{S_{\max }-S_{\min }}
\end{equation}

Since lower C-index values indicate better clustering quality, we negate the output value during evaluation to align with metrics where higher is better.

\subsection{SDBW index \citep{halkidi2001clustering}}
Consider the vector of variances for each variable in the data set $X = (x_1^T, \cdots, x_N^T)^T$:

\begin{equation}
\mathcal{V} = diag(Cov(X))
\end{equation}
For cluster \( C_k \), let the corresponding data be represented as \( X_k \). Then, we have the vector:

\begin{equation}
\mathcal{V}^{(k)} = diag(Cov(X_k))
\end{equation}

Let $\mathcal{S}$ be the average of the norms of $\mathcal{V}^{(k)}$'s, divided by the norm of $\mathcal{V}$:

\begin{equation}
\mathcal{S} = \frac{\frac{1}{K}\sum_{k=1}^K ||\mathcal{V}^{(k)}||}{||\mathcal{V}||}
\end{equation}

Let \( \sigma \) be the square root of the total norms of \( \mathcal{V}^{\{k\}} \)'s, divided by $K$.

\begin{equation}
\sigma=\frac{1}{K} \sqrt{\sum_{k=1}^{K}\left\|\mathcal{V}^{\{k\}}\right\|}
\end{equation}

The density \( \gamma_{k k^{\prime}} \) at a given observation, with respect to clusters \( C_k \) and \( C_{k^{\prime}} \), is defined as the number of observations from these clusters within a distance \( \sigma \) of the observation. This is determined by counting the observations in \( C_k \cup C_{k^{\prime}} \) that lie inside a ball of radius \( \sigma \) centered at the given observation. For each pair \( (k, k') \), the ratio \( R_{k k^{\prime}} \) is defined as the density at the midpoint \( H_{k k^{\prime}} \) of the line segment connecting the centroids of \( C_k \) and \( C_{k^{\prime}} \), divided by the greater of the two densities at the centroids \( \mu^{\{k\}} \) and \( \mu^{\{k^{\prime}\}} \).
\begin{equation}
R_{k k^{\prime}}=\frac{\gamma_{k k^{\prime}}\left(H_{k k^{\prime}}\right)}{\max \left(\gamma_{k k^{\prime}}\left(\mu^{\{k\}}\right), \gamma_{k k^{\prime}}\left(\mu^{\left\{k^{\prime}\right\}}\right)\right)}
\end{equation}

Then the between-cluster density $\mathcal{G}$ is obtained from averaging the ratios $R_{k k^{\prime}}$:

\begin{equation}
\mathcal{G}=\frac{2}{K(K-1)} \sum_{k<k^{\prime}} R_{k k^{\prime}}
\end{equation}

The SDbw index is defined as :
\begin{equation}
\pi_{SDbw}=\mathcal{S}+\mathcal{G}
\end{equation}

Since lower SDbw values indicate better clustering quality, we negate the index during evaluation.

\subsection{Cubic clustering criterion \citep{sarle1983sas}}
Let $A_{N\times K}$ represent a one-hot encoding matrix for the clustering membership of the observations in the data set. Now assuming $X$ is the centered data, the matrix of cluster means $\overline{X}$ can be computed as:
\begin{equation}
\overline{X} = (A^TA)^{-1}A^TX
\end{equation}
Define the total-sample sum-of-square and crossproducts (SSCP) matrix $T$ as:
\begin{equation}
T = X^TX
\end{equation}
Define the between-cluster SSCP matrix $B$ as:
\begin{equation}
B = \overline{X}^TA^TA\overline{X}
\end{equation}
Then the with-cluster SSCP matrix $W$ is defined as:
\begin{equation}
W = T-B
\end{equation}
Then the observed $\hat{R}^2$ for the clustering result can be expressed as:
\begin{equation}
\hat{R}^2 = 1 - \frac{trace(W)}{trace(T)}
\end{equation}

Consider approximating \( R^2 \) for a population uniformly distributed within a hyperbox composed of equal-sized hypercubes, assuming edges aligned along the coordinate axes. To remain consistent with the notation convention of the original paper, we use $n$ instead of $N$ to denote the number of observations. Let \( s_j \) denote the hyperbox's edge length in the \( j \)-th dimension, computed as the square root of the \( j \)-th eigenvalue of \( T / (n-1) \), with \( s_j \) values ordered in decreasing magnitude. Denote the hyperbox volume along the first $p^*$ dimensions as \( v^* \). If the hyperbox can be partitioned into \( q \) (i.e., \( K \)) hypercubes of edge length \( c \), its total volume equals the sum of the hypercube volumes. Let \( u_j \) denote the number of hypercubes along the \( j \)-th dimension, and define \( p^* \) as the largest integer less than \( q \) for which \( u_{p^*} \) is at least one. Hence, we have

\begin{equation}
\begin{aligned}
& v^{*}=\prod_{j=1}^{p^*} s_j, \\
& c=\left(\frac{v^{*}}{q}\right)^{\frac{1}{p^*}}, \\
& u_{j}=\frac{s_{j}}{c},
\end{aligned}
\end{equation}

Thus, we obtain the following small-sample approximation for the expectation of \( R^2 \):

\begin{equation}
E\left(R^{2}\right)=1-\left[\frac{\sum_{j=1}^{p^{*}} \frac{1}{n+u_{j}}+\sum_{j=p^{*}+1}^{p} \frac{u_{j}^{2}}{n+u_{j}}}{\sum_{j=1}^{p} u_{j}^{2}}\right]\left[\frac{(n-q)^{2}}{n}\right]\left[1+\frac{4}{n}\right] .
\end{equation}

The CCC is computed as
\begin{equation}
\pi_{CCC}=\ln \left[\frac{1-E\left(R^{2}\right)}{1-\hat{R}^{2}}\right] \frac{\sqrt{\frac{n p^{*}}{2}}}{\left(0.001+E\left(R^{2}\right)\right)^{1.2}}
\end{equation}

\subsection{CDbw index \citep{halkidi2008density}}
Consider each observation as a piont. Let \( V_k \) be the set of $r$ representative points for cluster \( C_k \), capturing its geometry (see the original paper \citep{halkidi2008density} for a detailed definition of representative points). A representative point \( v_{k,i} \) in cluster \( C_k \) (where \( i \in C_k \)) is defined as the closest representative to a representative point \( v_{k',j} \) in cluster \( C_{k'} \) (where \( j \in C_{k'} \)), if \( v_{k,i} \) has the minimum distance to \( v_{k',j} \) among all representatives in \( C_k \).  The Respective Closest Representative points (\( RCR_{kk'} \)) between clusters \( C_k \) and \( C_{k'} \) are defined as the set of mutually closest representatives of the two clusters. Let \( \text{clos\_rep}_{kk'}^l = (v_{k,i}, v_{k',j}) \) denote the \( l \)-th pair of respective closest representative points between clusters \( C_k \) and \( C_{k'} \).
The density between two clusters $C_k$ and $C_{k'}$ is defined as:

\begin{equation}
\operatorname{Dens}(C_k, C_{k'}) = \frac{1}{\left| RCR_{kk'} \right|} \sum_l \left( \frac{d\left(\operatorname{clos\_rep}_{kk'}^{l} \right)}{2 \cdot \operatorname{stdev}} \cdot \left| u_{kk'}^{l} \right| \right)
\end{equation}

where \( d\left(\text{clos\_rep}_{k k'}^{l}\right) \) represents the Euclidean distance between the pionts in the pair \( \text{clos\_rep}_{k k'}^{l} \in RCR_{k k'} \). The term \( \left| RCR_{k k'} \right| \) denotes the cardinality of the set \( RCR_{k k'} \), while \( \text{stdev} \) refers to the average standard deviation of the clusters being considered. Additionally, \( \left| u_{k k'}^{l} \right| \) represents the proportion of points in clusters \( C_k \) and \( C_{k'} \) that belong to the neighborhood of \( u_{k k'}^{l} \), the midpoint of the line segment defined by $\text{clos\_rep}_{k k'}^{l}$.

The inter-cluster density of a partition \( \mathbf{C} \) averages the highest density between each cluster \( C_k \) and any other cluster in \( \mathbf{C} \):

\begin{equation}
\operatorname { Inter\_dens }(\mathbf{C})=\frac{1}{K} \sum \max _{ k \neq k'}\left\{\operatorname{Dens}\left(C_{k}, C_{k'}\right)\right\}
\end{equation}

Cluster separation measures the separation between clusters based on inter-cluster density and distance:

\begin{equation}
\operatorname{Sep}(\mathbf{C})=\frac{\frac{1}{K} \sum \min _{\substack{{k} \neq {k'}}}\left\{\operatorname{Dist}\left({C}_{{k}}, {C}_{{k'}}\right)\right\}}{1+\operatorname { Inter\_dens}(\mathbf{C})}, \quad {K}>1
\end{equation}

where $\operatorname{Dist}(C_k, C_{k'}) = \frac{1}{\left| RCR_{kk'} \right|} \sum d\left(\operatorname{clos\_rep}_{kk'}^{l}\right)$.

Then the relative intra-cluster density given a shrinkage factor $s$ is defined as follows:
\begin{equation}
\operatorname{Intra\_dens}(\mathbf{C}, \mathrm{s})=\frac{ { \operatorname{Dens\_cl} }(\mathbf{C}, \mathrm{s})}{\mathrm{K} \cdot \operatorname{stdev}}, \mathrm{K}>1 
\end{equation}

where $\operatorname{Dens\_cl}(\mathbf{C}, \mathrm{s})
=\frac{1}{\mathrm{r}} \sum_{C_k \in \mathbf{C}} \sum_{i \in V_k} \left|v_{k,i}\right|$. The cardinality of a point \( v_{k,i} \) (a representative point shrunk by a factor $s$), denoted as \( \left| v_{k,i} \right| \), measures the proportion of points in cluster \( C_k \) that fall within the neighborhood of \( v_{k,i} \), where the neighborhood is defined as a hypersphere centered at \( v_{k,i} \) with radius equal to \(\operatorname{stdev}\). For a detailed definition, explanation, and discussion of the shrunken representative, the shrinking factor $s$ and its selection, please refer to the original paper.

The compactness of a partition $\mathbf{C}$ with respect to density is defined as:

\begin{equation}
    \operatorname{Compactness}(\mathbf{C})=\sum_{s} \operatorname{Intra\_dens}(\mathbf{C}, s) / n_{s}
\end{equation}
where $n_{s}$ represents the number of distinct values taken by $s$, determining the density at different regions within clusters.

Then, Intra-density change is defined as:

\begin{equation}
    \operatorname{Intra\_change}(\mathbf{C})=\frac{\sum \mid \operatorname{Intra\_dens}\left(\mathbf{C}, \mathrm{s}_{\mathrm{i}}\right)- \operatorname{Intra\_dens}\left(\mathbf{C}, \mathrm{s}_{\mathrm{i}-1}\right) \mid}{\left(\mathrm{n}_{\mathrm{s}}-1\right)}
\end{equation}

Cohesion is defined as:

\begin{equation}
    \operatorname{ Cohesion }(\mathbf{C})=\frac{ \operatorname{ Compactness }(\mathbf{C})}{1+ \operatorname{ Intra\_change }(\mathbf{C})}
\end{equation}

SC (Separation with respect to Compactness) is defined as:  %measures cluster separation (inter-cluster density), in relation to compactness (intra-cluster density):

\begin{equation}
\operatorname{SC}(\mathbf{C})=\operatorname{Sep}(\mathbf{C}) \cdot \text { Compactness }(\mathbf{C})
\end{equation}

Finally, the CDbw index is defined as:

\begin{equation}
\pi_{CDbw} =\text { Cohesion }(\mathbf{C}) \cdot \mathrm{SC}(\mathbf{C})
\end{equation}

%%%%%%%%%%%%%%%%%%%%%%%%%%%%%%%%%%%%%%%%%%%%%%%%%%%%
\section{Technical Proofs} \label{app:proof}

\subsection{Proof of Theorem \ref{thm2}} \label{app:proof:1}
\begin{proof}
Since $\pi$ is a consistent measure in ${\cal Z}_{1}$ and ${\cal Z}_{2}$, we have $\pi(\phi_{1}(X)|{\cal Z}_{2})\ge\pi(\phi_{2}(X)|{\cal Z}_{2})$ with probability tending to $1$.

(1) If ${\cal Z}_{1}\succeq{\cal Z}_{2},$ by definition we have $\mathbb{P}(\pi(\phi_{1}(X)|{\cal Z}_{1})-\pi(\phi_{1}(X)|{\cal Z}_{2})\ge0)\rightarrow1$.
Thus 
\begin{align*}
 & \mathbb{P}(\pi(\phi_{1}(X)|{\cal Z}_{1})\ge\pi(\phi_{2}(X)|{\cal Z}_{2}))\\
\ge & \mathbb{P}(\pi(\phi_{1}(X)|{\cal Z}_{1})\ge\pi(\phi_{1}(X)|{\cal Z}_{2})\textrm{ and }\pi(\phi_{1}(X)|{\cal Z}_{2})\ge\pi(\phi_{2}(X)|{\cal Z}_{2}))\\
\ge & \mathbb{P}(\pi(\phi_{1}(X)|{\cal Z}_{1})\ge\pi(\phi_{1}(X)|{\cal Z}_{2}))+\mathbb{P}(\pi(\phi_{1}(X)|{\cal Z}_{2})\ge\pi(\phi_{2}(X)|{\cal Z}_{2}))-1\\
\rightarrow & 1+1-1=1
\end{align*}

as $n\rightarrow\infty$. 

(2) If ${\cal Z}_{1}\prec{\cal Z}_{2}$, 

i) Consider the case where $\phi_{1}(X)=\phi_{2}(X)$, i.e., $\phi_{1}(X)$
and $\phi_{2}(X)$ are the same. 
\begin{align*}
 & \mathbb{P}(\pi(\phi_{1}(X)|{\cal Z}_{1})-\pi(\phi_{2}(X)|{\cal Z}_{2})\ge0)\\
= & \mathbb{P}(\pi(\phi_{1}(X)|{\cal Z}_{1})-\pi(\phi_{1}(X)|{\cal Z}_{2})\ge0)
\end{align*}

It follows by the definition of ${\cal Z}_{1}\prec{\cal Z}_{2}$ that $\mathbb{P}(\pi(\phi_{1}(X)|{\cal Z}_{1})-\pi(\phi_{2}(X)|{\cal Z}_{2})\ge0)$
does not converge to 1.

ii) Consider the case where $\phi_{1}(X)\neq\phi_{2}(X)$, without
loss of generality we assume $\phi_{1}(X)$ is better than $\phi_{2}(X)$. Then we have
the following decomposition:
\[
\pi(\phi_{1}(X)|{\cal Z}_{1})-\pi(\phi_{2}(X)|{\cal Z}_{2})=\left[\pi(\phi_{1}(X)|{\cal Z}_{1})-\pi(\phi_{2}(X)|{\cal Z}_{1})\right]-\left[\pi(\phi_{2}(X)|{\cal Z}_{2})-\pi(\phi_{2}(X)|{\cal Z}_{1})\right].
\]
The first quantity $\left[\pi(\phi_{1}(X)|{\cal Z}_{1})-\pi(\phi_{2}(X)|{\cal Z}_{1})\right]$
represents the clustering difference on space ${\cal Z}_{1}$, and
the second quantity $\left[\pi(\phi_{2}(X)|{\cal Z}_{2})-\pi(\phi_{2}(X)|{\cal Z}_{1})\right]$
represents the space difference. If the clustering difference is greater
than or equal to the space difference, we then have $\pi(\phi_{1}(X)|{\cal Z}_{1}) \ge \pi(\phi_{2}(X)|{\cal Z}_{2})$.
Since ${\cal Z}_{1}$ and ${\cal Z}_{2}$ are distinguishable, by
definition we have $\mathbb{P}(\max_{\phi_{1}}\left[\pi(\phi_{1}(X)|{\cal Z}_{1})-\pi(\phi_{2}(X)|{\cal Z}_{1})\right]<\left[\pi(\phi_{2}(X)|{\cal Z}_{2})-\pi(\phi_{2}(X)|{\cal Z}_{1})\right])\rightarrow c$
for some $0<c\le 1$. So 
\begin{align*}
 & \mathbb{P}(\pi(\phi_{1}(X)|{\cal Z}_{1})-\pi(\phi_{2}(X)|{\cal Z}_{2})\ge 0)\\
= & 1-\mathbb{P}(\pi(\phi_{1}(X)|{\cal Z}_{1})-\pi(\phi_{2}(X)|{\cal Z}_{1})<\pi(\phi_{2}(X)|{\cal Z}_{2})-\pi(\phi_{2}(X)|{\cal Z}_{1}))\\
\le & 1-\mathbb{P}(\max_{\phi_{1}}\left[\pi(\phi_{1}(X)|{\cal Z}_{1})-\pi(\phi_{2}(X)|{\cal Z}_{1})\right]<\pi(\phi_{2}(X)|{\cal Z}_{2})-\pi(\phi_{2}(X)|{\cal Z}_{1}))\\
\rightarrow & 1-c<1.
\end{align*}
In summary, $\mathbb{P}(\pi(\phi_{1}(X)|{\cal Z}_{1})\ge\pi(\phi_{2}(X)|{\cal Z}_{2}))\rightarrow1$
happens only when ${\cal Z}_{1}={\cal Z}_{2}$ (trivial case) and ${\cal Z}_{1}\succeq{\cal Z}_{2}$. 
\end{proof}

\subsection{Proof of Theorem \ref{thm3}} \label{app:proof:2}
\begin{proof}
By definition we have 
\[
\lim_{n\rightarrow\infty}\mathbb{P}((\pi(\phi_{1}(X)|{\cal Z}_{1})-\pi(\phi_{2}(X)|{\cal Z}_{1}))\cdot(V(\rho^{*},\phi_{1}(X))-V(\rho^{*},\phi_{2}(X)))\ge0)=\epsilon_{{\cal Z}_{1}}
\]
and
\[
\lim_{n\rightarrow\infty}\mathbb{P}((\pi(\phi_{1}(X)|{\cal Z}_{2})-\pi(\phi_{2}(X)|{\cal Z}_{2}))\cdot(V(\rho^{*},\phi_{1}(X))-V(\rho^{*},\phi_{2}(X)))\ge0)=\epsilon_{{\cal Z}_{2}}.
\]
Denote the above events as $A:=\{(\pi(\phi_{1}(X)|{\cal Z}_{1})-\pi(\phi_{2}(X)|{\cal Z}_{1}))\cdot(V(\rho^{*},\phi_{1}(X))-V(\rho^{*},\phi_{2}(X))) \allowbreak \ge0\} $ and $B:=\{(\pi(\phi_{1}(X)|{\cal Z}_{2})-\pi(\phi_{2}(X)|{\cal Z}_{2}))\cdot(V(\rho^{*},\phi_{1}(X))-V(\rho^{*},\phi_{2}(X)))\ge0\}$. Although $\phi_1(X)$ and $\phi_2(X)$ are dependent, it is reasonable to assume that the two spaces ${\cal Z}_1$ and ${\cal Z}_2$ are independent in comparing any two cluster assignments, and thus $\pi(\phi(X)|{\cal Z}_1)$ and $\pi(\phi(X)|{\cal Z}_2)$ can be treated as independent.  It is also reasonable to assume the each space ${\cal Z}_i, i=1,2$, is independent from the external measures $V(\rho^{*},\phi_{1}(X))$ and $V(\rho^{*},\phi_{2}(X))$ when comparing any two cluster assignments. We assume the two external measure values produce no ties. Thus, it follows that the events $A$ and $B$ are independent. Then, 
\begin{align*}
&\lim_{n\rightarrow\infty}\mathbb{P}((\pi(\phi_{1}(X)|{\cal Z}_{1})-\pi(\phi_{2}(X)|{\cal Z}_{1}))\cdot(\pi(\phi_{1}(X)|{\cal Z}_{2})-\pi(\phi_{2}(X)|{\cal Z}_{2}))\ge0)\\ &=1-\lim_{n\rightarrow\infty}\mathbb{P}(A\backslash B)-\lim_{n\rightarrow\infty}\mathbb{P}(B\backslash A) \\ &= 1-\lim_{n\rightarrow\infty}[\mathbb{P}(A\cup B)-P(A\cap B)]\\ & =1-(\epsilon_{{\cal Z}_{1}}+\epsilon_{{\cal Z}_{2}}-2\epsilon_{{\cal Z}_{1}}\epsilon_{{\cal Z}_{2}})\\
 & \ge0.5
\end{align*}
since $\epsilon_{{\cal Z}_{1}}\ge0.5$ and $\epsilon_{{\cal Z}_{2}}\ge0.5$. 

For the special case where $\pi$ is consistent in both ${\cal Z}_{1}$
and ${\cal Z}_{2}$, we have $\pi(\phi_{1}(X)|{\cal Z}_{1})\ge\pi(\phi_{2}(X)|{\cal Z}_{1})$
 if and only if $\pi(\phi_{1}(X)|{\cal Z}_{2}) \ge \pi(\phi_{2}(X)|{\cal Z}_{2})$ with probability tending to $1$, i.e.,
\[\lim_{n\rightarrow\infty}\mathbb{P}((\pi(\phi_{1}(X)|{\cal Z}_{1})-\pi(\phi_{2}(X)|{\cal Z}_{1}))\cdot(\pi(\phi_{1}(X)|{\cal Z}_{2})-\pi(\phi_{2}(X)|{\cal Z}_{2}))\ge0)=1.\]
\end{proof}

\subsection{Proof of Corollary \ref{cor1}} \label{app:proof:3}
\begin{proof}
To set up the rank among the $m$ clusterings, assuming each pairwise comparison is independent, we need to do $O(m \log m)$ (under transitivity) or ${m \choose 2}$ (under no transitivity) times of pairwise comparison. 

For any $i\neq j\in\{1,...,m\}$, by definition we have 
\[
\lim_{n\rightarrow\infty}\mathbb{P}((\pi(\phi_{i}|{\cal Z}_{1})-\pi(\phi_{j}|{\cal Z}_{1}))\cdot(V(\rho^{*},\phi_{i})-V(\rho^{*},\phi_{j}))\ge0)=\epsilon_{{\cal Z}_{1}}
\]
and $\lim_{n\rightarrow\infty}\mathbb{P}((\pi(\phi_{i}|{\cal Z}_{2})-\pi(\phi_{j}|{\cal Z}_{2}))\cdot(V(\rho^{*},\phi_{i})-V(\rho^{*},\phi_{j}))\ge0)=\epsilon_{{\cal Z}_{2}}$.
So, for any fixed pair of distinct indices $(i,j)$, by Proof \ref{app:proof:2}, we have
\[
\lim_{n\rightarrow\infty}P\left((\pi(\phi_{i}|{\cal Z}_{1})-\pi(\phi_{j}|{\cal Z}_{1}))\cdot(\pi(\phi_{i}|{\cal Z}_{2})-\pi(\phi_{j}|{\cal Z}_{2}))\ge 0\right)=1-(\epsilon_{{\cal Z}_{1}}+\epsilon_{{\cal Z}_{2}}-2\epsilon_{{\cal Z}_{1}}\epsilon_{{\cal Z}_{2}})
\]
 and thus 
\begin{align*}
 & \lim_{n\rightarrow\infty}P\left(\textrm{the rankings in } \mathbf{a}\textrm{ and }\mathbf{b}\textrm{ agree}\right)\\
= & \lim_{n\rightarrow\infty}P\left((\pi(\phi_{i}|{\cal Z}_{1})-\pi(\phi_{j}|{\cal Z}_{1}))\cdot(\pi(\phi_{i}|{\cal Z}_{2})-\pi(\phi_{j}|{\cal Z}_{2}))\ge 0 \textrm{ for any distinct pair } (i, j) \textrm{ to compare}\right)\\
= &
 \left(1-(\epsilon_{{\cal Z}_{1}}+\epsilon_{{\cal Z}_{2}}-2\epsilon_{{\cal Z}_{1}}\epsilon_{{\cal Z}_{2}})\right)^{{O(m \log m)}}
\quad \text{(w/ transitivity)} \\
& \text{or} \quad
 \left(1-(\epsilon_{{\cal Z}_{1}}+\epsilon_{{\cal Z}_{2}}-2\epsilon_{{\cal Z}_{1}}\epsilon_{{\cal Z}_{2}})\right)^{{m \choose 2}} \quad \text{(w/o transitivity)}.
\end{align*}
\end{proof}

\section{Additional Algorithms Details} \label{app:alg}

\subsection{Dip statistics \citep{hartigan1985dip}} \label{app:dip}
Given embedded data from each space $\mathcal{Z}_{m}$, we use the widely applied Dip test for testing clusterability. The Dip test is designed to measure the discrepancy between the cumulative distribution function (CDF) of the data and the nearest unimodal function. For a given CDF $F(z)$, the Dip $D(F)$ is defined as $\inf_{G\in \mathcal{A}} \sup_z |F(z) - G(z)|$, where $\mathcal{A}$ represents the class of unimodal CDFs. Considering the empirical CDF $F_n(z)$ of the data $z_1, z_2, \cdots, z_n$ (e.g., dimension-reduced embeddings), the Dip of $F_n(z)$ asymptotically converges to the Dip of $F$ (i.e., $D(F_n) \rightarrow D(F)$ a.s.). In the Dip test, a uniform distribution $Unif(0,1)$ is chosen as a``null" model. \citet{hartigan1985dip} suggested that $Unif(0,1)$ is the ``asymptotically least favorable" unimodal distribution, essentially, the most challenging to distinguish from multimodal distributions as $n$ increases. Following that, \emph{p}-values for the Dip test under the null hypothesis that $F$ is a unimodal distribution can be derived via precomputed table interpolation or through Monte Carlo simulations with $Unif(0,1)$ \citep{maechler2013package}. Based on the computed \emph{p}-values from different $\mathcal{Z}_{m}$, we apply the Holm–Bonferroni method that controls the family-wise error rate (FWER) at 0.05 \citep{holm1979simple} to identify embedding spaces where the null hypothesis is rejected, indicating that \( F \) is not unimodal. If no spaces are selected, we bypass the Dip test and proceed with all spaces to the next step.

The computation of $D(F)$ can be conceptualized as analogous to stretching a taut string. The stretched string idea with relevant concepts (e.g., greatest convex minorant and least concave majorant) is detailed in the original paper \citep{hartigan1985dip}, and its implementation can be summarized as follows. Consider $z_1, \cdots, z_n$, the atoms of $F$. 

 \begin{itemize}
    \item Initialize: $z_L =z_1$, $z_U=z_n$, $D=0$.
    \item Repeat:
     \begin{enumerate}
    \item Calculate the greatest convex minorant $G$ and least concave majorant $L$ for $F$ in $[z_L,z_U]$. Denote the points where \( G \) and \( L \) make contact with \( F \) be \( g_1, g_2, \dots, g_i, \dots \) and \( l_1, l_2, \dots, l_j, \dots \), respectively.
    \item If \( \sup |G(g_i)-L(g_i)| > \sup |G(l_i)-L(l_i)| \) with the supremum occurring at \( l_j \leq g_i \leq l_{j+l} \), then set \( z^0_L = g_i \), \( z^0_U = l_{j+l} \), and \( d = \sup |G(g_i)-L(g_i)| \); otherwise if \( \sup |G(l_i)-L(l_i)| \geq \sup |G(g_i)-L(g_i)| \) with the supremum occurring at \( g_i \leq l_j \leq g_{i+l} \), then set \( z^0_L = g_i \), \( z^0_U = l_j \), and \( d = \sup |G(l_i)-L(l_i)| \).
    \item If \( d \leq D \), terminate and set \( D(F) = D \); otherwise, update 
    $$D=\sup\{D,\sup_{z_L \le z \le z_L^0} |G(z) -F(z)|, \sup_{z^0_U \le z \le z_U} |L(z)- F(z)|\}$$
    and set $z_U = z^0_U$, $z_L = z_L^0$ 
 \end{enumerate}
 \end{itemize}

\subsection{Stage-wise grouping} \label{app:stagecls}
The details of the stage-wise grouping algorithm can be seen in Algorithm \ref{Algorithm2}. 
\begin{algorithm}[H]
\caption{\small Algorithm for stage-wise grouping}
\label{Algorithm2} 
\begin{algorithmic}[1]
\Input{ $\{RankCorr(\pi_i,\pi_j)\}_{i,j\in M}$ and $\{\mathbf{\pi}_{i}\}_{i\in M}$, where $M$ is the set of retained spaces after multimodality test}

\State Phase 1: clustering based on rank correlations:
\begin{enumerate}
\itemsep0em 
    \item For each $i,j \in M$, define the distance $d_{ij} = 1 - RankCorr(\pi_i,\pi_j)$
    \item Run a density-based clustering method based on $\{d_{i,j}\}_{i,j\in M}$
    \item Obtain $S^{(\text{phase1})}$ groups of spaces $\{G^{(\text{phase1})}_s\}_{s=1}^{S^{(\text{phase1})}}$ (each $G^{(\text{phase1})}_s \subseteq M$) with outlier spaces excluded 
\end{enumerate}
\State Phase 2: clustering based on score values
\begin{enumerate}
    \item For each group $G_s^{(\text{phase1})}$, apply density-based clustering to $\{\mathbf{\pi}_{i}\}_{i\in G_s^{(\text{phase1})}}$
    \item Within each group $G_s^{(\text{phase1})}$, collect the subgroups $SG^{(s)}_1, \cdots SG^{(s)}_{N_s}$ together with outlier spaces $\{O^{(s)}_l\}_{l=1}^{L_{s}}$ produced by clustering 
    \item Treat each outlier space $O_l^{(s)}$ as a singleton subgroup. Combine these singleton subgroups with $\{SG_n^{(s)}\}_{n=1}^{N_s}$over all $s$ to yield the final set of $S$ mutually exclusive subgroups $\{G_s\}_{s=1}^S$ (where the notation is simplified from $SG$ to $G$)
\end{enumerate}
\Output {$\{G_s\}_{s=1}^S$, where each $G_s \subseteq M$}
\end{algorithmic}
\end{algorithm}
Note that when calculating rank correlation across different embedding spaces, any embedding space with constant scores should be removed. In Algorithm \ref{Algorithm2}, we omit outlier spaces from density-based clustering in the first phase, treating them as approximately rank uncorrelated spaces (i.e., spaces exhibiting low or negative rank correlation with other spaces). In the second phase, we handle and incorporate outlier spaces as singleton subgroups. The distinction lies in the fact that the second phase is solely intended for grouping spaces with similar internal measure scales, while the first phase is employed to identify rank-correlated spaces.  Further discussion on the decision to include or exclude outlier spaces in Phase 1 can be found in Supplementary Section \ref{app:outlier}. Additionally, to ensure meaningful clustering, we proceed with Phase 2 clustering for a group \( G_s^{\text{phase1}} \) only if its sample size is not too small (e.g., $> 5$) (density-based clustering may flag all points as outliers in when the sample size is small). In the rare case where all spaces to be clustered are flagged as outliers (a possible outcome in density-based methods), we can apply a refinement step that leverages outlier scores (e.g., GLOSH scores \citep{campello2015hierarchical}) and filtering techniques such as the Interquartile Range (IQR) filter \citep{tukey1977exploratory} to identify only the spaces with high scores as outliers.

\subsection{Link analysis} \label{app:link}
Given a graph or network, link analysis is a valuable technique for assessing relationships between nodes and assigning an importance score $w$ to each node. Two prominent algorithms commonly employed for link analysis are:
\paragraph{Hyperlink-Induced Topic Search (HITS) algorithm~\citep{kleinberg1999authoritative}} The HITS algorithm is based on an intuition that a good \emph{authority} node is linked by numerous high quality hub nodes, and a good \emph{hub} node points to numerous trusted authorities. For each node $v_i$, HITS computes an $auth(v_i)$ value based on incoming links and a $hub(v_i)$ value based on outgoing links. This mutually reinforcing relationship is mathematically expressed through the following operations \citep{ding2002pagerank, langville2005survey}:

\begin{equation}
    auth(v_i) = \sum_{j: e_{j i} \in E} hub(v_j), \quad hub(v_i) = \sum_{j: e_{i j} \in E} auth(v_j)
\end{equation}

The final authority and hub scores for each node are obtained through an iterative updating process. Since HITS produces two distinct scores that typically do not converge to the same values, it is less suitable for our setting compared to the PageRank algorithm described below. Additional details can be found in \citet{ding2002pagerank} and \citet{langville2005survey}.

\paragraph{PageRank (PR) algorithm~\citep{page1998pagerank}} The PageRank (PR) algorithm shares a similar idea with HITS, suggesting that a good node should be linked to or linked by other good nodes. PR employs a web surfing model rooted in a Markov process, introducing a different approach for determining the scores compared to the mutual reinforcement concept in HITS \citep{ding2002pagerank,hamzehei2017topic}. Let $P = \left(P_{ij}\right)$ be a stochastic matrix, obtained by rescaling the adjacency matrix so that each row sums to one. Here, $P_{ij}$ represents the probability of transitioning from node $v_i$ to $v_j$. Incorporating the idea of link-interrupting jumps, the matrix $P$ is adjusted by adding a matrix $\Lambda$ consisting of all ones, resulting in $P' =\alpha P + (1-\alpha)\frac{1}{M}\Lambda$ ($M$ is the number of nodes), where $0 < \alpha < 1$. Then the resulting score in PR, indicating each node's importance, is determined by the equilibrium distribution $\zeta$, satisfying $\sum_{k}\zeta_k = 1$ through the equation:

\begin{equation}
    P'^T \mathbf{\zeta} = \mathbf{\zeta}
\end{equation}

This solution can be obtained iteratively. Further details are available in \citet{ding2002pagerank} and \citet{langville2005survey}.

In our ensemble analysis, we require normalized weights, so when using these scores as ratings, we divide them by their sum. In practice, the implementations of these algorithms we use usually produce scores that already sum to one.

\section{Additional Experimental Details} 
\subsection{Evaluation metrics} \label{app:exp:evm}
\paragraph{Spearman's rank correlation coefficient \citep{spearman1961proof,kendall1979advanced,press2007numerical}}
Spearman's rank correlation coefficient, denoted as $r_s$, is a nonparametric measure of agreement between rankings. The coefficient assesses both the direction and magnitude of a monotonic association between two random variables. It is calculated by determining the Pearson correlation, denoted as $r_p$, between the ranks of the variables and has a range between $-1$ and $1$.

Given two measures, \( \pi \) (e.g., a scoring approach based on an internal measure) and \( S \) (e.g., an external measure), each producing \( n \) scores, the scores are first converted into their respective ranks, denoted as \( \operatorname{R}(\pi) \) and \( \operatorname{R}(S) \). The rank correlation coefficient \( r_s \) is then calculated using these ranks as follows:

\begin{equation}
     r_s =
 r_p({\operatorname{R}(\pi),\operatorname{R}(S)}) =
 \frac{\operatorname{cov}(\operatorname{R}(\pi), \operatorname{R}(S))}
      {\sigma_{\operatorname{R}(\pi)} \sigma_{\operatorname{R}(S)}},
\end{equation}

where $\operatorname{cov}(\operatorname{R}(\pi), \operatorname{R}(S))$ is the covariance of $\operatorname{R}(\pi)$ and $\operatorname{R}(S)$. $\sigma_{\operatorname{R}(\pi)}$ and $\sigma_{\operatorname{R}(S)}$ are their respective standard deviations.

The significance of Spearman's rank correlation can be tested under the null hypothesis ($H_0$): $r_s = 0$, which indicates no monotonic relationship in the population. The alternative hypothesis ($H_1$) can be two-sided: $r_s \neq 0$ (nonzero correlation), left-sided:  $r_s < 0$ (negative correlation), or right-sided: $r_s > 0$ (positive correlation). The test statistic is expressed as:
\begin{equation}
    t_s = r_s \sqrt{\frac{n - 2}{1 - r_s^2}}
\end{equation}
which approximately follows a Student's t-distribution $t_{n-2}$ under the null hypothesis \citep{kendall1979advanced,press2007numerical}.

\paragraph{Kendall rank correlation coefficient \citep{kendall1938new, kendall1945treatment, agresti2010analysis}}
The Kendall rank correlation coefficient ($\tau$) serves as a statistical metric quantifying the ordinal association between two variables. As a measure of rank correlation, it ranges from $-1$ (perfect inversion) to $1$ (perfect agreement), with a value of zero signifying an absence of association. A higher $\tau$ between two variables suggests that their observations share similar rank ordering, while a lower correlation indicates disparity in their rank orderings.

Consider a set of observations \((\pi_1, s_1), \dots, (\pi_n, s_n)\) for the joint random variables \(\pi\) and \(S\). A pair of observations \((\pi_i, s_i)\) and \((\pi_j, s_j)\), where \(i < j\), is concordant if the order of \((\pi_i, \pi_j)\) matches that of \((s_i, s_j)\), meaning either both \(\pi_i > \pi_j\) and \(s_i > s_j\) or both \(\pi_i < \pi_j\) and \(s_i < s_j\). If either \(\pi_i = \pi_j\) or \(s_i = s_j\), the pair is tied; otherwise, it is discordant.

The Kendall coefficient $\tau_B$ is defined as:

\begin{equation}
   \tau_B = \frac{n_c-n_d}{\sqrt{(n_0-n_1)(n_0-n_2)}}
\end{equation}

where
$n_0  = n(n-1)/2$. $n_c$ represents the count of concordant pairs, while $n_d$ indicates the count of discordant pairs. $n_1$ is the number of pairs tied on the first variable (e.g., $\pi$ for the pair $\{\pi,S\}$), and $n_2$ is the number of pairs tied on the second variable (e.g., $S$ for the pair $\{\pi,S\}$). The count of discordant pairs can be obtained from the inversion number, representing the count of rearrangements needed to permute the $S$-sequence with the order of the $\pi$-sequence.\\

In our evaluation, we assume that higher scores indicate better clustering quality, aligning with the external measure, so higher rank correlations reflect stronger agreement. For internal measures where lower scores are better, we negate the values before applying the scoring strategies.
%%%%%%%%%%%%%%%%%%%%%%%%%%%%%%%%%%%%%%%%%%%%%%%
%%%%%%%%%%%%%%%%%%%%%%%%%%%%%%%%%%%%%%%%%%%%%%%
\subsection{Additional implementation details} \label{app:exp:ad}
In this section, we provide additional details regarding our experiments. 

In our real data analysis, we implement the deep clustering models, JULE{\footnote{\url{https://github.com/jwyang/JULE.torch}}}, DEPICT{\footnote{\url{https://github.com/herandy/DEPICT}}}, and DeepCluster{\footnote{\url{https://github.com/facebookresearch/deepcluster}}}, along with their data processing steps using the original source code provided in their respective papers. In our simulation studies, we perform K-Means clustering using the \emph{sklearn} \citep{scikit-learn} library and apply UMAP using the \emph{umap} \citep{mcinnes2018umap} package, and normalize both the raw and embedded data for clustering and evaluation to have zero mean and unit variance. For computing internal measures such as Silhouette score \citep{rousseeuw1987silhouettes}, Calinski-Harabasz index \citep{calinski1974dendrite}, and Davies-Bouldin index \citep{davies1979cluster}, we utilize functions from the \emph{sklearn} library in \emph{Python}. The computation of Cubic clustering criterion (CCC) \citep{sarle1983sas}, Dunn index \citep{dunn1974well}, Cindex \citep{hubert1976general}, SDbw index \citep{halkidi2001clustering} involves using R and reimplementations based on the core functions provided by \emph{NbClust} \citep{malika2014nbclust}. For CDbw index \citep{halkidi2008density}, we calculate scores using the function provided by the \emph{R} package \emph{fpc} \citep{fpc}. For these index calculations in R, we impose a one-hour time limit for computations on the input data space to prevent excessive runtime, and a two-minute limit on the embedding space for the indices that are computationally intensive. When the calculated scores yield \texttt{NA} or invalid values, we impute them using the minimum non-\texttt{NA} value within each corresponding evaluation for the input-space and embedding-space scores, respectively. The Dip test is implemented in R using the function from the R package \emph{clusterability} \citep{dip}. By default, the package conducts a PCA dimension reduction on the tested data before performing the test. The link analysis algorithms, PageRank and HITS, are implemented using the Python library \emph{networkx} \citep{hagberg2008exploring}. PageRank is configured with a damping parameter of \(0.9\), while all other parameters are set to their default values. If any negative scores are generated by these algorithms, their outputs are disregarded, and equal weights are assigned to all spaces in the ensemble analysis. The multiple tests are conducted using the Python library \emph{statsmodels} \citep{seabold2010statsmodels}, with multiple comparisons adjusted using the `holm' method (Holm–Bonferroni \citep{holm1979simple}). When constructing the edge set for the ensemble analysis, all negative edges were excluded following the multiple testing procedure. The \emph{HDBSCAN} clustering and outlier detection are implemented using the Python library \emph{hdbscan} \citep{mcinnes2017hdbscan}, and \emph{DBSCAN} is implemented using \emph{sklearn}. Additional analyses (e.g., t-SNE), and metric calculations, (e.g., rank correlation), are performed using the \emph{sklearn} and \emph{scipy} \citep{2020SciPy-NMeth} packages in Python. The implementation of the ACE algorithm is carried out in Python.

JULE and DEPICT were executed on machines equipped with NVIDIA TITAN X and NVIDIA GeForce GTX 1080 Ti GPUs, with each training run utilizing a single GPU. DeepCluster was run on a machine with eight NVIDIA RTX A5500 GPUs. The internal measure calculations and Dip test implementation were executed on Slurm nodes equipped with Intel Icelake or Xeon Gold CPUs and 128 GB of memory, with each task allocated two CPUs. ACE was executed on a machine equipped with 12 Intel Core i7-6800K CPUs and 128 GB of RAM.

\paragraph{Hyperparameter tuning} \label{app:exp:ad1}
For the JULE algorithm, we construct the search space by selecting the learning rate from the list $[0.0005, 0.001, 0.005, 0.01, 0.05, 0.1]$ and the unfolding rate ($\eta$) from the list $[0.2, 0.3, 0.4, 0.5, 0.7, 0.8, 0.9]$, resulting in $42$ hyperparameter combinations. For the DEPICT algorithm, we define the search space by choosing the learning rate from the list $[0.00005, 0.0001, 0.0005, 0.001, 0.005, 0.01]$ and the balancing parameter of the reconstruction loss function from the list $[0.1, 1.0, 10.0]$, yielding $18$ hyperparameter combinations. For each combination, we execute the deep clustering algorithm, and if a training trial fails, we consider the clustering results as missing and exclude that specific combination from the final evaluation.

\paragraph{Evaluating the number of clusters} \label{app:exp:ad2}
In this experimental setup, when running JULE and DEPICT across datasets, we search for $K$ among $10$ different values that are evenly distributed, covering the true $K$. To create the search space for $K$, we specify the following intervals: for the datasets FRGC, MNIST-test, USPS, UMist, YTF, and COIL-20, we use $linspace(5, 50, num=10)$; for CMU-PIE, we choose $linspace(10, 100, num=10)$; and for COIL-100, we apply $linspace(20, 200, num=10)$. Here, $linspace(start, end, num)$ denotes the generation of evenly spaced numbers over the specified interval $[start, end]$ with a total of $num$ values. For each \( K \), we run the clustering algorithm. If a training trial fails, we retry with a nearby \( K \). If it fails again, we consider the clustering result as missing and exclude that \( K \) from the final evaluation.

\paragraph{Selection of checkpoints} \label{app:exp:ad3}
In consideration of training time and computational resources, we choose to download and use the validation set from ImageNet \citep{deng2009imagenet}{\footnote{\url{https://www.kaggle.com/datasets/titericz/imagenet1k-val}}
, which consists of approximately $50,000$ images distributed across $1000$ classes. We train a VGG16 network~\citep{simonyan2014very} following the source repository of DeepCluster, modifying the maximum number of clusters, which is set to 1000. The deep clustering process runs for $100$ epochs, with checkpoints saved every five epochs, resulting in a total of $20$ checkpoints. At each checkpoint, we save the generated $256$-dimensional features that are used for clustering and the corresponding estimated cluster assignments for evaluation.

%%%%%%%%%%%%%%%%%%%%%%%%%%%%%%%%%%%%%%%%%%%%%%%%%%%%

\section{Deep Clustering Algorithms} \label{app:dcl}
In the following sections, we provide a brief overview of the deep clustering algorithms evaluated in this paper: JULE \citep{yang2016joint}, DEPICT \citep{ghasedi2017deep} and DeepCluster \citep{caron2018deep}.

\subsection{JULE~\citep{yang2016joint} }
JULE is notable as a joint unsupervised learning approach that employs agglomerative clustering, a type of hierarchical clustering technique, to train its feature extractor,  without relying on the use of autoencoders. JULE performs joint learning within a recurrent framework. In this framework, each step treats the merging operations of agglomerative clustering as a forward pass for generating cluster labels, while the representation learning of deep neural networks serves as the backward pass. JULE introduces a unified weighted triplet loss, which is optimized to enhance the estimation of cluster labels and embeddings. The loss serves two purposes: reducing inner-cluster distances and increasing intra-cluster distances. Consequently, in each step, JULE systematically merges clusters, and computes the loss for the backward pass. 

\subsection{DEPICT~\citep{ghasedi2017deep}}
DEPICT follows an autoencoder-based framework. The approach applies a multinomial logistic regression function to a convolutional autoencoder. The method introduces a clustering scheme that projects data into an embedding space and predicts cluster assignments from this space. It employs a joint learning framework that simultaneously minimizes both the reconstruction loss and the clustering loss. The clustering loss is formulated via KL divergence minimization, with an additional regularization term that incorporates a prior on the frequency of cluster assignments.

\subsection{DeepCluster~\citep{caron2018deep}}
DeepCluster is an end-to-end approach that simultaneously updates the network and image clusters for large-scale deep clustering tasks. This method employs $k$-means on features extracted from large deep convolutional neural networks, such as VGG-16 ~\citep{simonyan2014very}, to predict cluster assignments. Subsequently, it makes use of these cluster assignments as ``pseudo-labels'' to guide the learning and update the parameters of the convolutional neural networks. Successfully applied to extensive datasets like ImageNet~\citep{deng2009imagenet}, this method has exhibited promising performance in learning visual features across different tasks.

\section{Additional Results for Simulation Studies} \label{app:exp:sim}
In this section, we present additional results from the simulation studies, including boxplots illustrating Spearman's rank correlation coefficients with ACC, as well as Kendall's rank coefficients ($\tau_B$) with NMI and ACC.

\begin{figure}[htbp!]
\centering
\subfigure{\includegraphics[width = 0.475\linewidth]{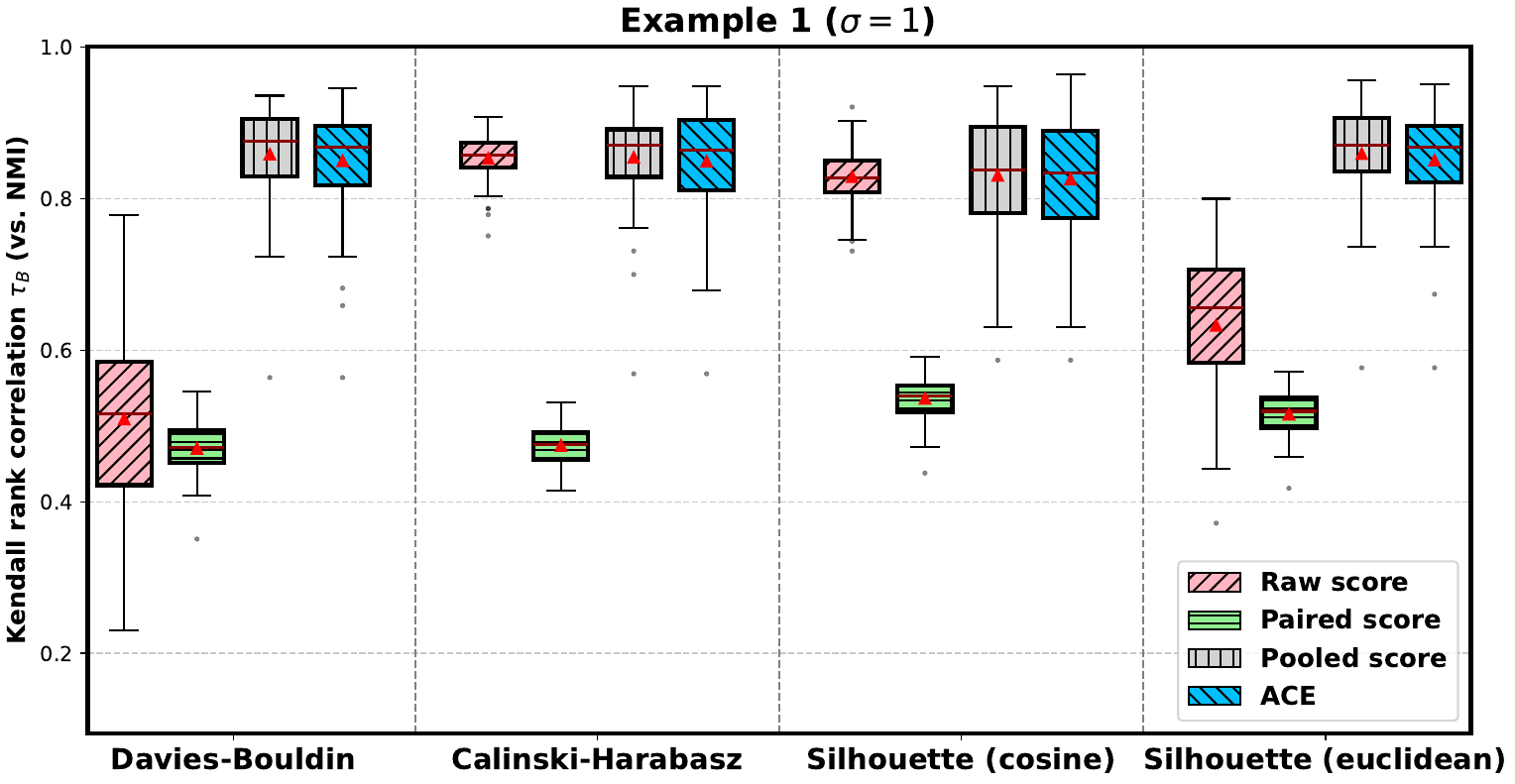}}
\subfigure{\includegraphics[width = 0.475\linewidth]{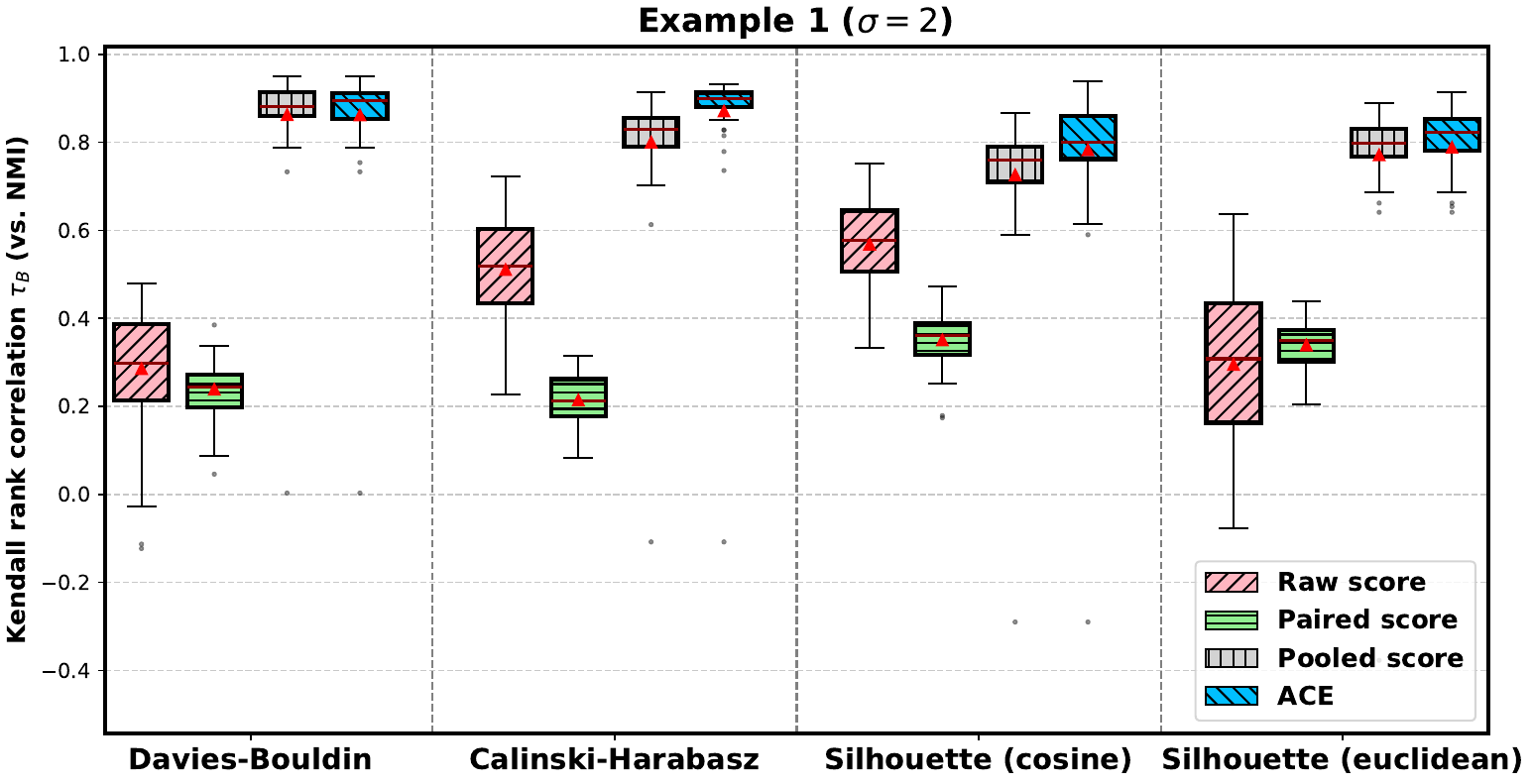}}\\
\subfigure{\includegraphics[width = 0.475\linewidth]{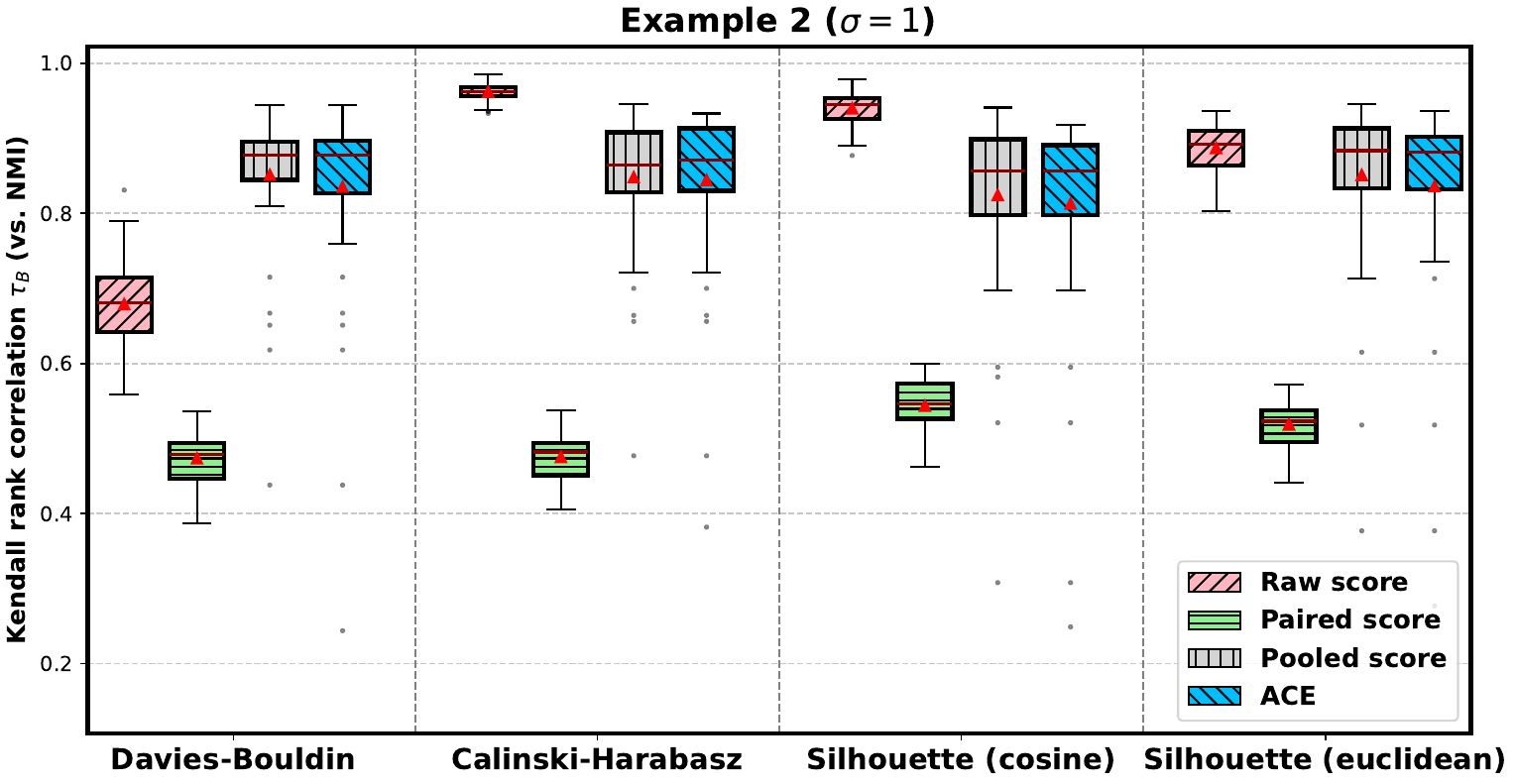}}
\subfigure{\includegraphics[width = 0.475\linewidth]{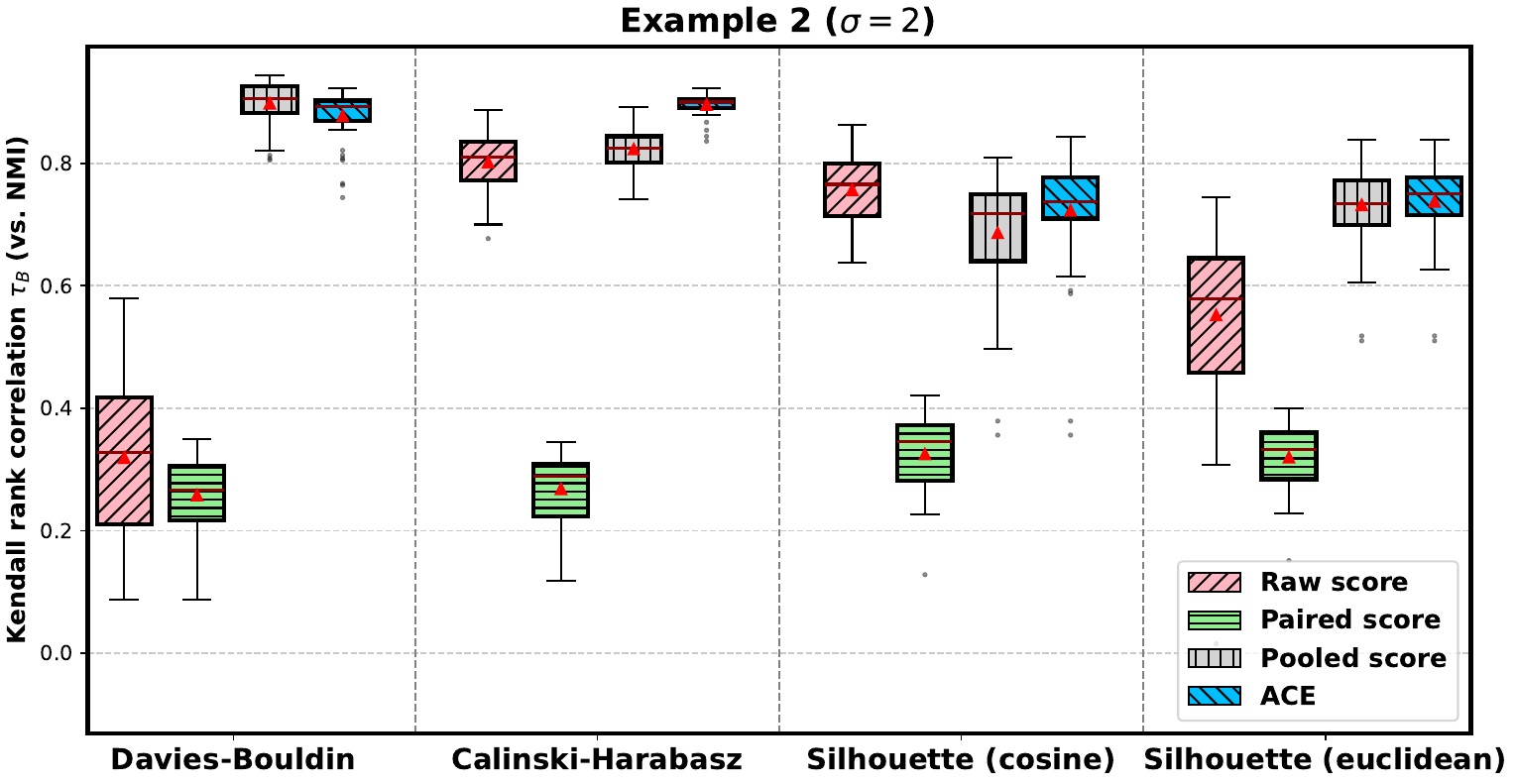}}
\caption{\small Boxplots illustrate rank consistency based on Kendall's rank coefficients ($\tau_s$) with NMI across simulation studies for various scoring approaches under different simulation settings. The two figures in the first row correspond to Example 1, with $\sigma = 1$ and $\sigma = 2$, respectively, while the two figures in the second row correspond to Example 2, also with $\sigma = 1$ and $\sigma = 2$. Different colors represent distinct scoring approaches. The mean value for each scoring approach across simulation studies is highlighted with a red triangle in each plot.}
\label{fig:app:sim1}
\end{figure}
\begin{figure}[htbp!]
\centering
\subfigure{\includegraphics[width = 0.475\linewidth]{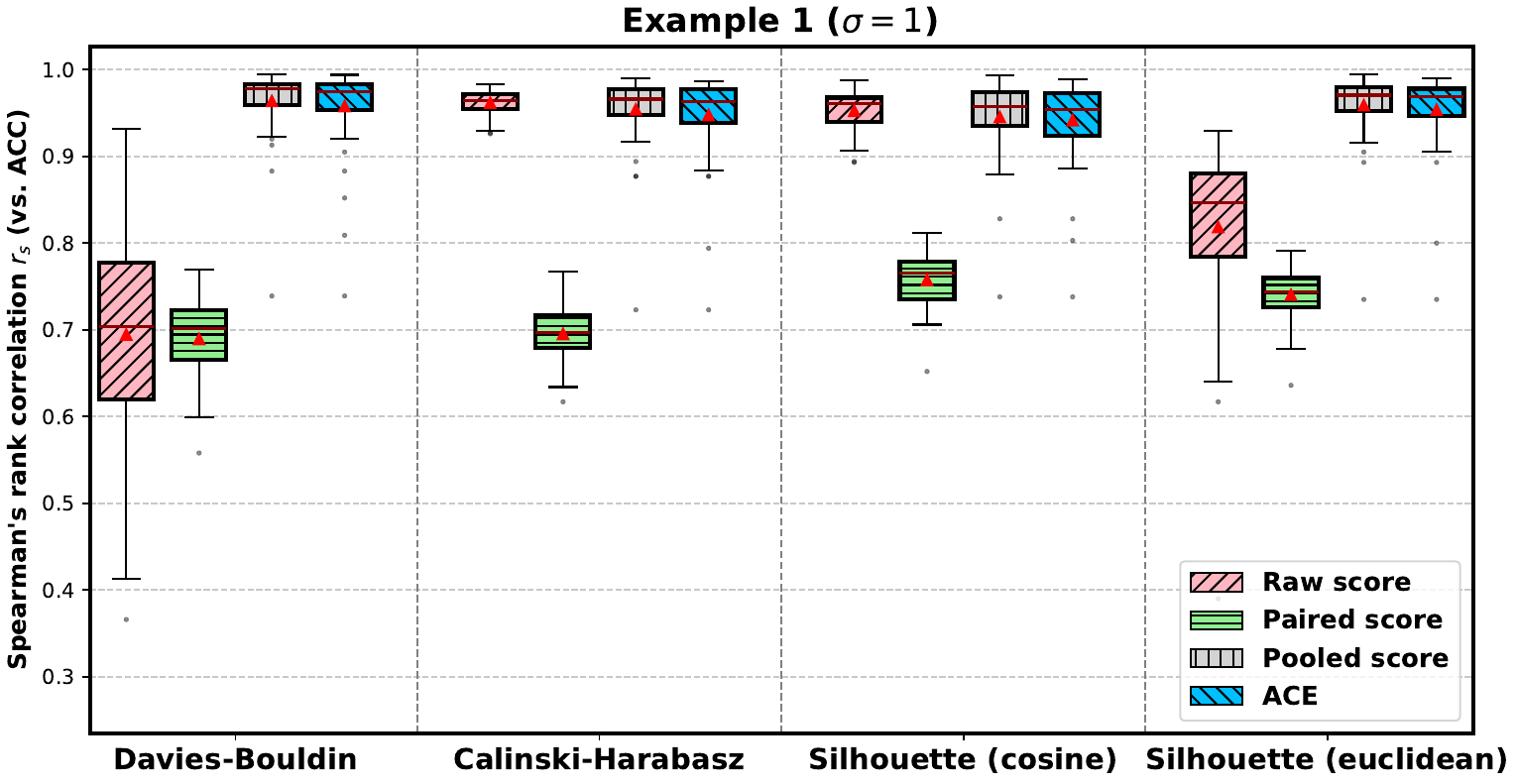}}
\subfigure{\includegraphics[width = 0.475\linewidth]{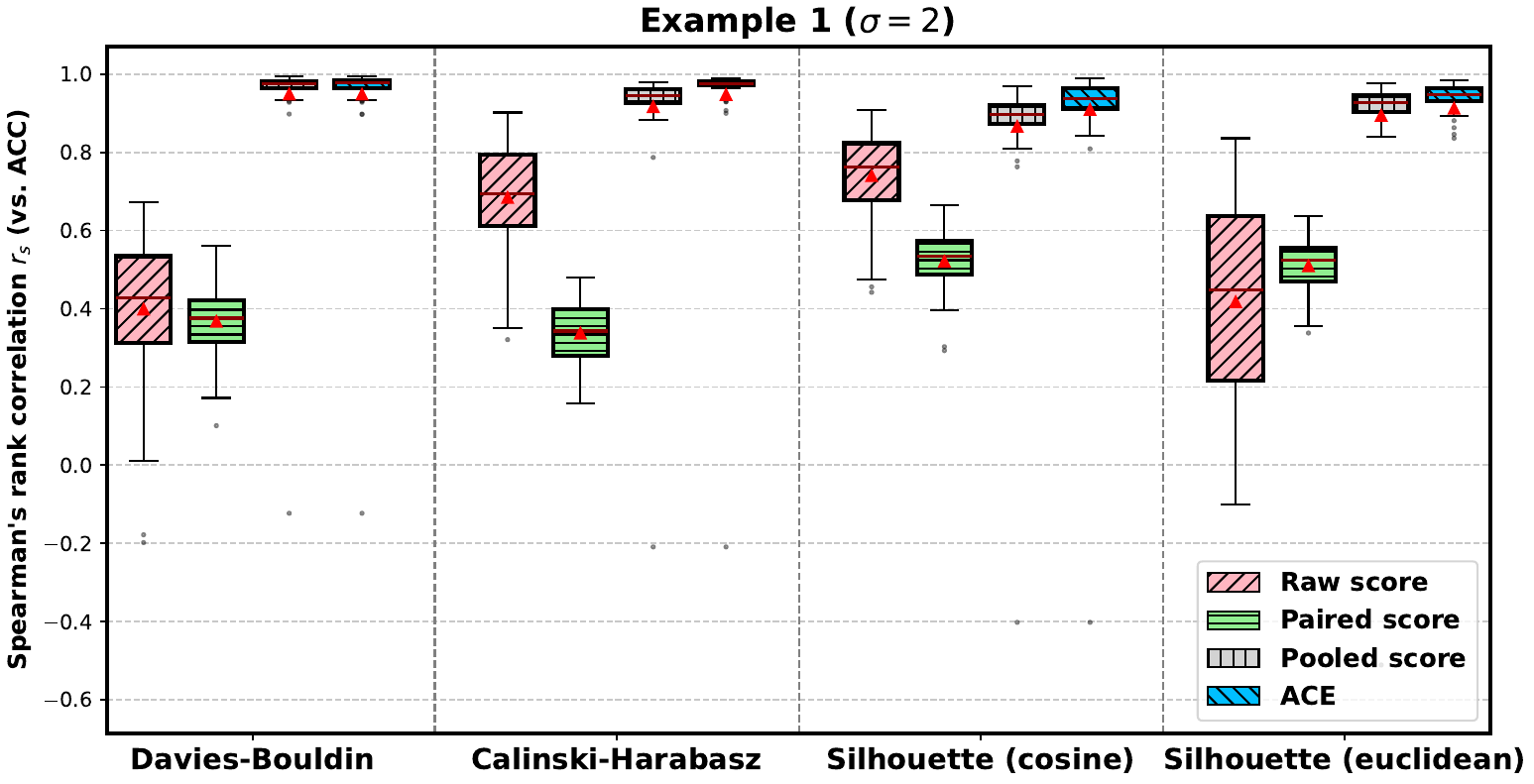}}\\
\subfigure{\includegraphics[width = 0.475\linewidth]{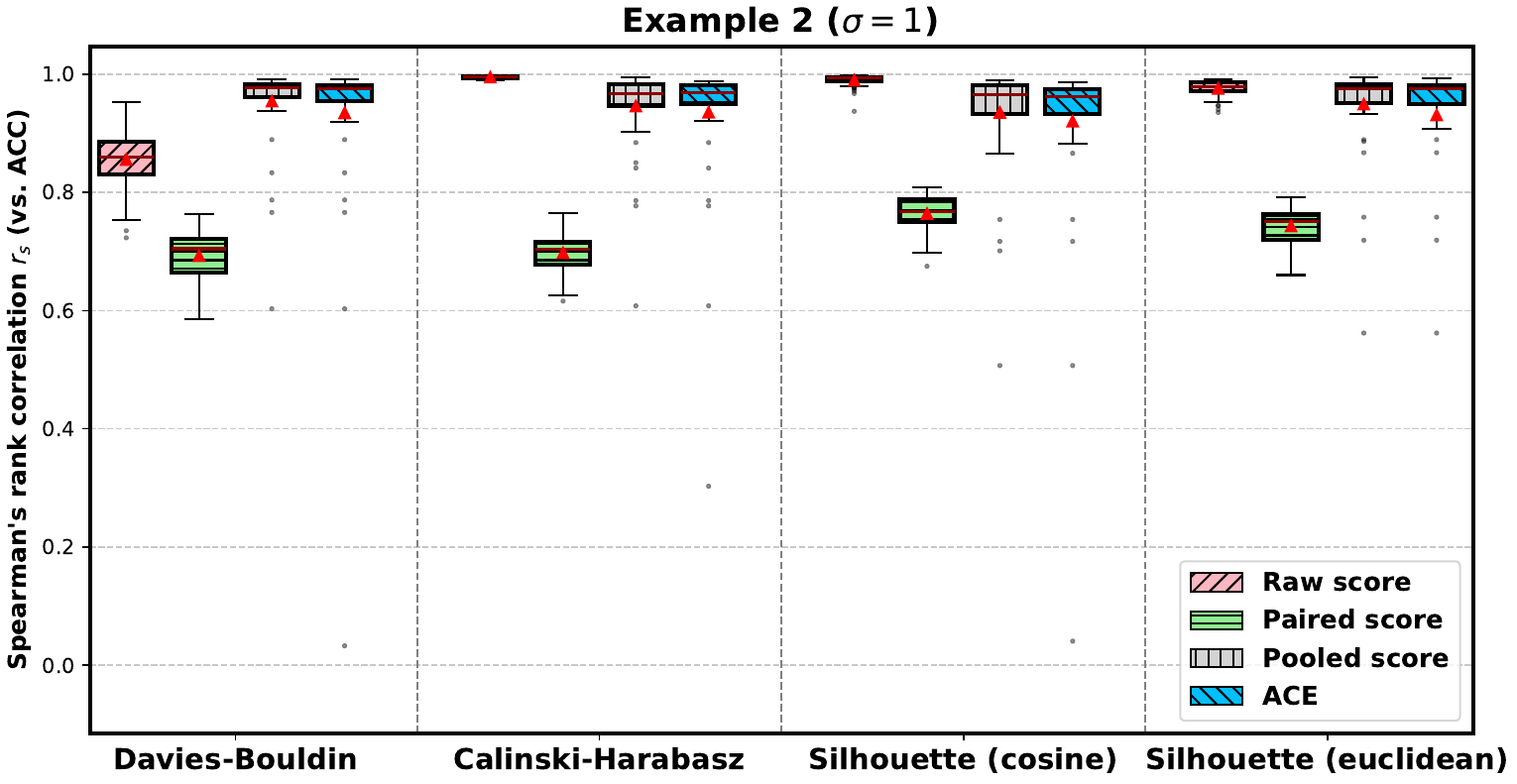}}
\subfigure{\includegraphics[width = 0.475\linewidth]{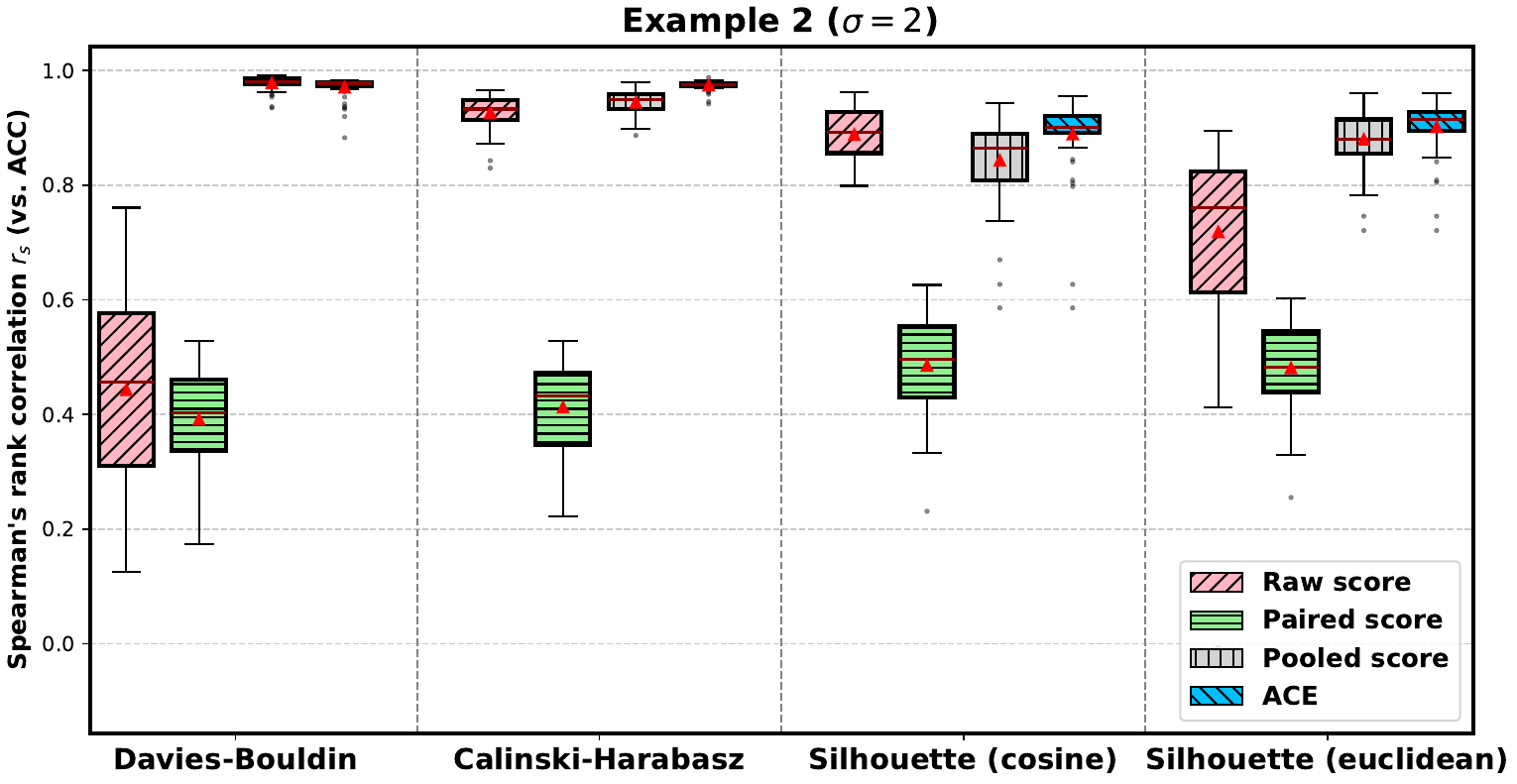}}
\caption{\small Boxplots illustrate rank consistency based on Spearman rank correlation coefficients ($r_s$) with ACC across simulation studies for various scoring approaches under different simulation settings.}
\label{fig:app:sim2}
\end{figure}
\begin{figure}[htbp!]
\centering
\subfigure{\includegraphics[width = 0.475\linewidth]{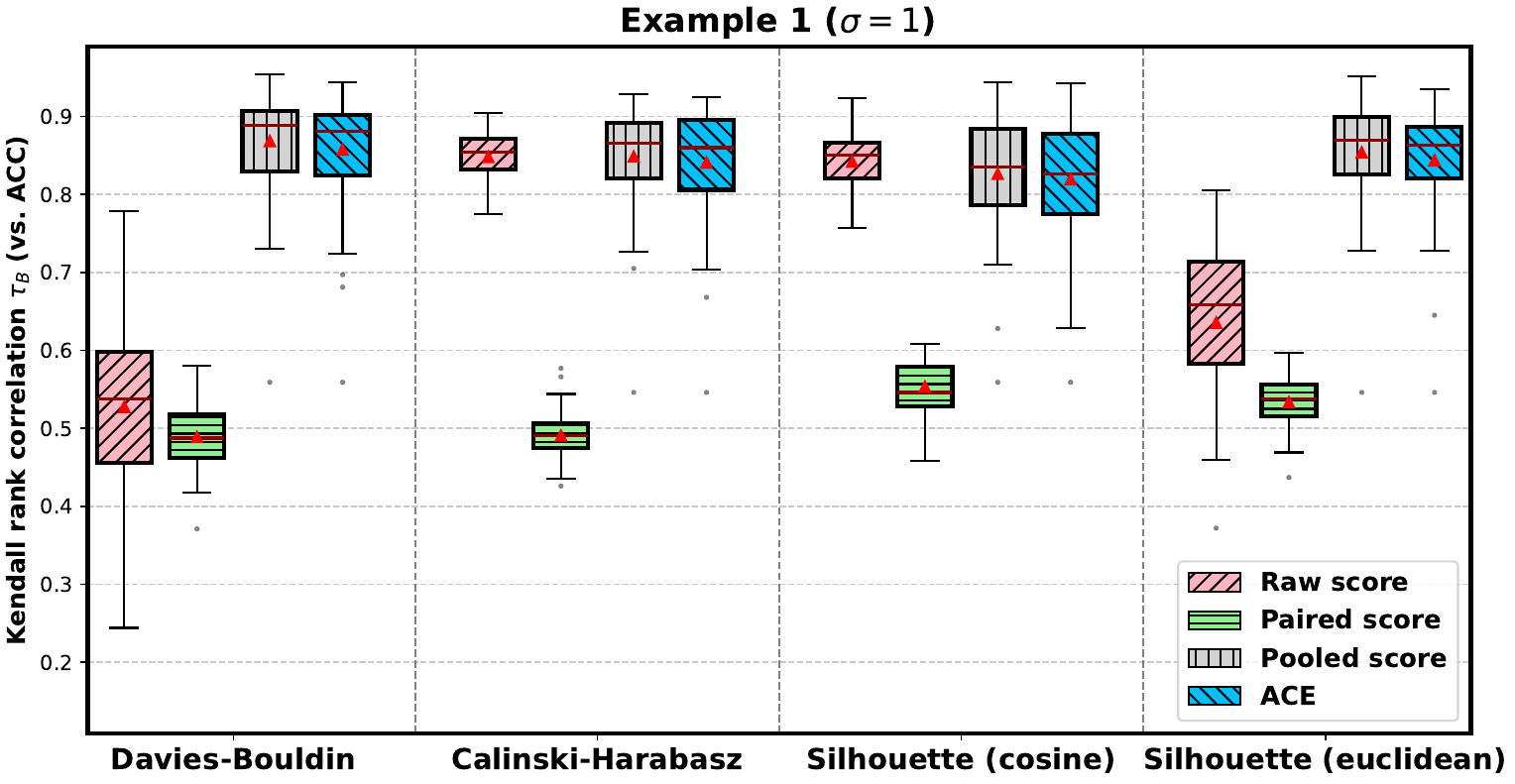}}
\subfigure{\includegraphics[width = 0.475\linewidth]{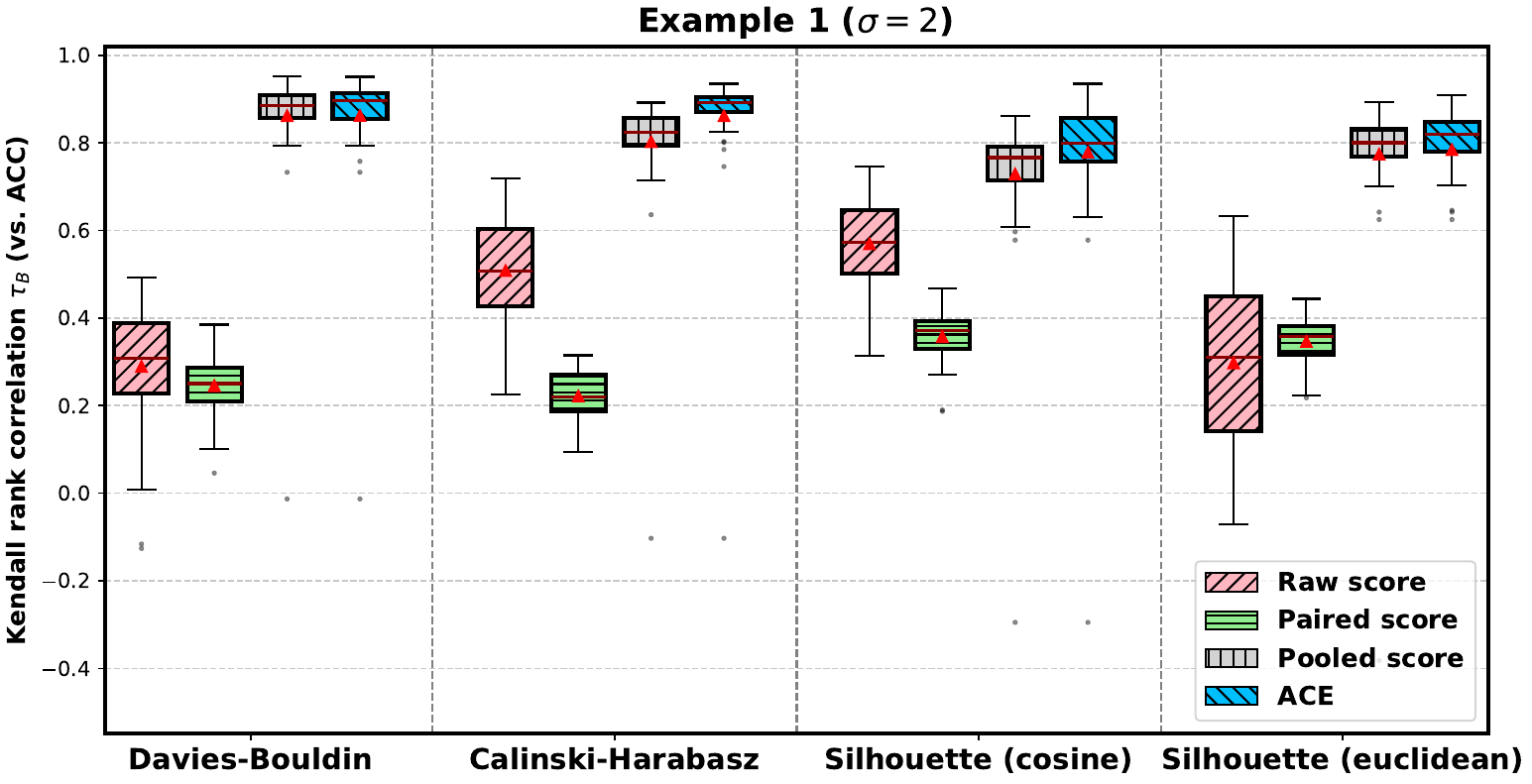}}\\
\subfigure{\includegraphics[width = 0.475\linewidth]{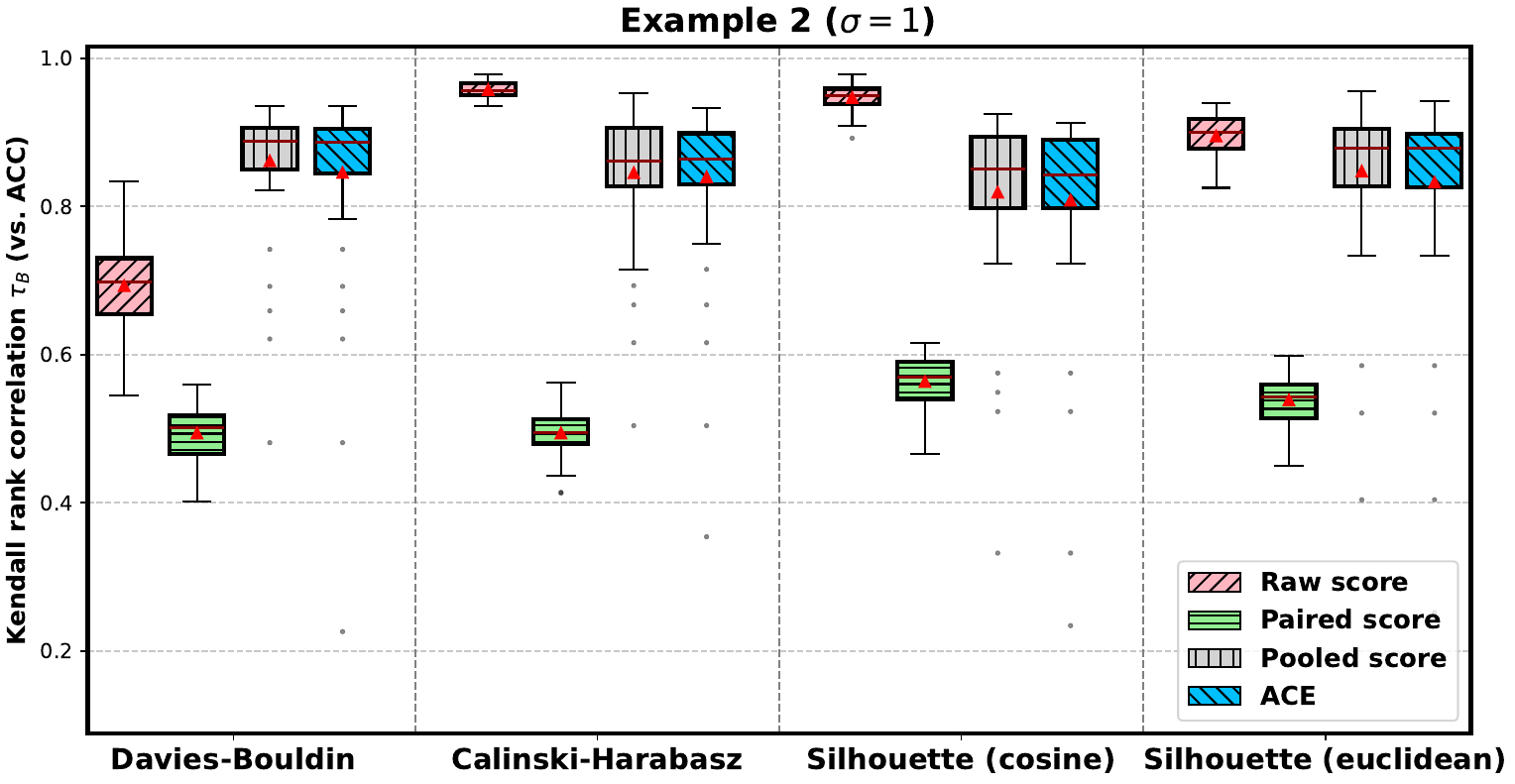}}
\subfigure{\includegraphics[width = 0.475\linewidth]{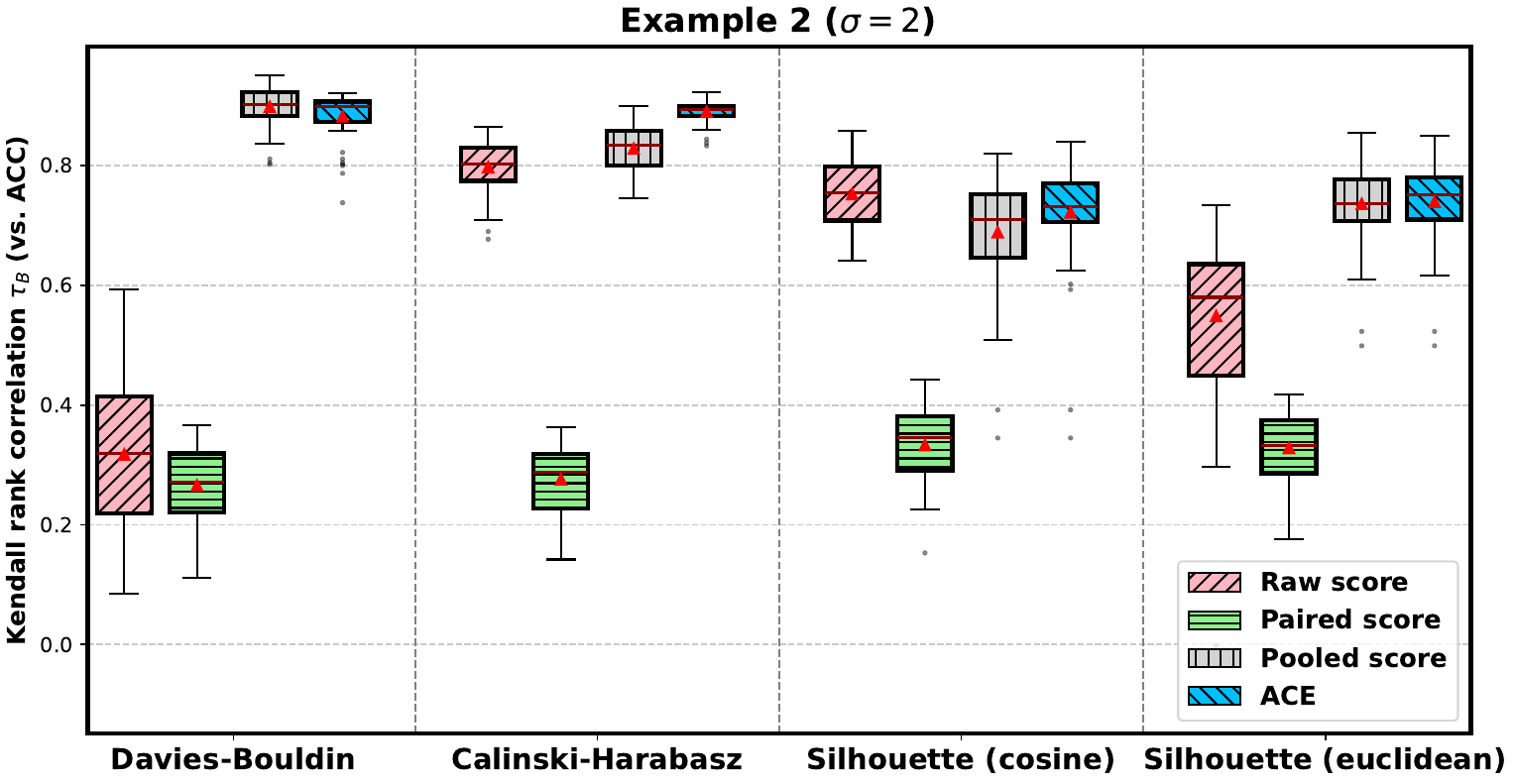}}
\caption{\small Boxplots illustrate rank consistency based on Kendall's rank coefficients ($\tau_s$) with ACC across simulation studies for various scoring approaches under different simulation settings.}
\label{fig:app:sim3}
\end{figure}

%%%%%%%%%%%%%%%%%%%%%%%%%%%%%%%%%%%%%%%%%%%%%%%%%%%%%%%%%
\newpage

\section{Additional Results for Experiments on Real-World Applications}\label{app:exp:ar}

\subsection{Application 1 - supplementary results for additional tables} \label{app:exp:ar:1}
In this section, we present additional tables for Application 1. First, we provide Table \ref{tab:acc:hyper}, which shows rank consistency with ACC based on the three internal measures discussed in the main text across different evaluation approaches. The findings in Table \ref{tab:acc:hyper} are generally consistent with those in Table \ref{tab:nmi:hyper} from the main text, which evaluates performance using NMI, further reinforcing our conclusions.

%%%%%%%%%%%%%%%%%%%%%%%%%%%%%%%%%%%%%%%%%%%%%%%%%%%%%%%%%%%%%%%
\begin{table*}[htbp!]
  \centering
  \caption{Rank consistency with ACC based on the three internal measures discussed in the main text for different evaluation approaches in Application 1. The top results for each dataset are highlighted in bold.}
  \resizebox{\textwidth}{!}{
\begin{tabular}{lllllllllllllllllll}
\toprule
 & \multicolumn{2}{l}{USPS} & \multicolumn{2}{l}{YTF} & \multicolumn{2}{l}{FRGC} & \multicolumn{2}{l}{MNIST-test} & \multicolumn{2}{l}{CMU-PIE} & \multicolumn{2}{l}{UMist} & \multicolumn{2}{l}{COIL-20} & \multicolumn{2}{l}{COIL-100} & \multicolumn{2}{l}{Average} \\
 & $r_s$ & $\tau_B$ & $r_s$ & $\tau_B$ & $r_s$ & $\tau_B$ & $r_s$ & $\tau_B$ & $r_s$ & $\tau_B$ & $r_s$ & $\tau_B$ & $r_s$ & $\tau_B$ & $r_s$ & $\tau_B$ & $r_s$ & $\tau_B$ \\
\midrule
\hline 
 \multicolumn{19}{c}{JULE: Calinski-Harabasz index} \\
 \hline 
Raw score & 0.70 & 0.59 & 0.54 & 0.39 & -0.52 & -0.35 & 0.91 & 0.76 & -0.98 & -0.91 & -0.50 & -0.35 & -0.29 & -0.17 & 0.36 & 0.23 & 0.03 & 0.02 \\
Paired score & 0.04 & 0.05 & 0.39 & 0.27 & -0.26 & -0.18 & 0.31 & 0.21 & -0.20 & -0.12 & 0.64 & 0.45 & 0.57 & 0.40 & 0.09 & 0.08 & 0.20 & 0.14 \\
Pooled score & \textbf{0.91} & \textbf{0.78} & \textbf{0.78} & \textbf{0.61} & 0.30 & 0.21 & 0.91 & 0.77 & 0.95 & 0.83 & \textbf{0.81} & 0.60 & \textbf{0.58} & \textbf{0.43} & 0.90 & 0.75 & 0.77 & 0.62 \\
\textbf{ACE} & 0.90 & 0.77 & 0.73 & 0.54 & \textbf{0.49} & \textbf{0.36} & \textbf{0.95} & \textbf{0.82} & \textbf{0.97} & \textbf{0.87} & \textbf{0.81} & \textbf{0.61} & 0.57 & 0.40 & \textbf{0.93} & \textbf{0.81} & \textbf{0.79} & \textbf{0.65} \\
\hline 
 \multicolumn{19}{c}{JULE: Davies-Bouldin index} \\
 \hline 
Raw score & -0.67 & -0.43 & -0.45 & -0.30 & -0.04 & -0.01 & -0.94 & -0.80 & -0.96 & -0.86 & -0.77 & -0.60 & -0.56 & -0.38 & -0.83 & -0.64 & -0.65 & -0.50 \\
Paired score & \textbf{-0.27} & -0.15 & -0.14 & -0.09 & -0.23 & -0.14 & -0.35 & -0.19 & 0.20 & 0.16 & \textbf{0.53} & \textbf{0.36} & \textbf{0.63} & \textbf{0.44} & 0.33 & 0.26 & 0.09 & 0.08 \\
Pooled score & -0.49 & -0.20 & -0.35 & -0.23 & 0.48 & 0.36 & -0.35 & -0.21 & \textbf{0.89} & \textbf{0.75} & 0.17 & 0.11 & -0.29 & -0.22 & -0.48 & -0.34 & -0.05 & 0.00 \\
\textbf{ACE} & -0.30 & \textbf{-0.09} & \textbf{-0.07} & \textbf{-0.07} & \textbf{0.53} & \textbf{0.38} & \textbf{0.79} & \textbf{0.64} & 0.07 & 0.03 & 0.27 & 0.20 & 0.21 & 0.18 & \textbf{0.44} & \textbf{0.28} & \textbf{0.24} & \textbf{0.19} \\
\hline 
 \multicolumn{19}{c}{JULE: Silhouette score (Cosine distance)} \\
 \hline 
Raw score & 0.77 & 0.59 & 0.64 & 0.47 & 0.31 & 0.21 & 0.79 & 0.61 & 0.69 & 0.54 & -0.37 & -0.27 & -0.16 & -0.13 & 0.06 & 0.02 & 0.34 & 0.26 \\
Paired score & 0.17 & 0.14 & 0.59 & 0.41 & 0.07 & 0.06 & 0.47 & 0.33 & 0.45 & 0.33 & 0.64 & 0.46 & \textbf{0.70} & \textbf{0.51} & 0.64 & 0.45 & 0.47 & 0.34 \\
Pooled score & 0.74 & 0.68 & 0.73 & \textbf{0.55} & 0.71 & 0.53 & 0.90 & 0.73 & 0.96 & 0.88 & 0.75 & 0.55 & 0.20 & 0.11 & 0.61 & 0.44 & 0.70 & 0.56 \\
\textbf{ACE} & \textbf{0.96} & \textbf{0.85} & \textbf{0.74} & \textbf{0.55} & \textbf{0.82} & \textbf{0.65} & \textbf{0.92} & \textbf{0.78} & \textbf{0.98} & \textbf{0.92} & \textbf{0.78} & \textbf{0.58} & 0.41 & 0.32 & \textbf{0.84} & \textbf{0.68} & \textbf{0.81} & \textbf{0.67} \\
\hline 
 \multicolumn{19}{c}{JULE: Silhouette score (Euclidean distance)} \\
 \hline 
Raw score & 0.92 & 0.77 & 0.59 & 0.43 & 0.27 & 0.19 & 0.83 & 0.66 & 0.35 & 0.32 & -0.35 & -0.24 & -0.14 & -0.05 & 0.14 & 0.08 & 0.33 & 0.27 \\
Paired score & 0.14 & 0.12 & 0.54 & 0.39 & -0.08 & -0.02 & 0.41 & 0.27 & 0.36 & 0.27 & 0.64 & 0.46 & \textbf{0.67} & \textbf{0.48} & 0.44 & 0.31 & 0.39 & 0.28 \\
Pooled score & 0.73 & 0.67 & \textbf{0.66} & \textbf{0.49} & 0.70 & \textbf{0.53} & 0.89 & 0.72 & 0.97 & 0.88 & 0.77 & 0.57 & 0.20 & 0.11 & 0.62 & 0.45 & 0.69 & 0.55 \\
\textbf{ACE} & \textbf{0.93} & \textbf{0.78} & 0.63 & 0.48 & \textbf{0.71} & \textbf{0.53} & \textbf{0.92} & \textbf{0.78} & \textbf{0.98} & \textbf{0.91} & \textbf{0.86} & \textbf{0.68} & 0.39 & 0.30 & \textbf{0.84} & \textbf{0.68} & \textbf{0.78} & \textbf{0.64} \\
\hline 
 \multicolumn{19}{c}{DEPICT: Calinski-Harabasz index} \\
 \hline 
Raw score & -0.10 & -0.19 & \textbf{0.65} & \textbf{0.50} & 0.54 & 0.38 & 0.59 & 0.47 & -0.95 & -0.83 &  &  &  &  &  &  & 0.14 & 0.07 \\
Paired score & 0.56 & 0.40 & 0.54 & 0.35 & 0.76 & 0.57 & 0.88 & 0.69 & 0.48 & 0.43 &  &  &  &  &  &  & 0.64 & 0.49 \\
Pooled score & \textbf{0.94} & \textbf{0.82} & 0.54 & 0.45 & \textbf{0.92} & 0.79 & 0.95 & 0.86 & 0.62 & 0.55 &  &  &  &  &  &  & 0.79 & 0.69 \\
\textbf{ACE} & 0.82 & 0.72 & 0.61 & 0.45 & 0.91 & \textbf{0.82} & \textbf{0.97} & \textbf{0.91} & \textbf{0.96} & \textbf{0.87} &  &  &  &  &  &  & \textbf{0.86} & \textbf{0.75} \\
\hline 
 \multicolumn{19}{c}{DEPICT: Davies-Bouldin index} \\
 \hline 
Raw score & 0.06 & -0.09 & 0.48 & 0.33 & 0.53 & 0.39 & 0.13 & 0.07 & -0.14 & -0.20 &  &  &  &  &  &  & 0.21 & 0.10 \\
Paired score & 0.61 & 0.42 & 0.48 & 0.32 & \textbf{0.92} & \textbf{0.74} & 0.88 & 0.69 & 0.62 & 0.56 &  &  &  &  &  &  & 0.70 & 0.55 \\
Pooled score & 0.95 & 0.84 & 0.40 & 0.28 & 0.64 & 0.48 & 0.38 & 0.28 & -0.76 & -0.60 &  &  &  &  &  &  & 0.32 & 0.26 \\
\textbf{ACE} & \textbf{0.99} & \textbf{0.96} & \textbf{0.65} & \textbf{0.46} & 0.90 & \textbf{0.74} & \textbf{0.99} & \textbf{0.96} & \textbf{0.96} & \textbf{0.87} &  &  &  &  &  &  & \textbf{0.90} & \textbf{0.80} \\
\hline 
 \multicolumn{19}{c}{DEPICT: Silhouette score (Cosine distance)} \\
 \hline 
Raw score & 0.43 & 0.33 & 0.69 & 0.52 & 0.77 & 0.62 & 0.83 & 0.64 & 0.43 & 0.26 &  &  &  &  &  &  & 0.63 & 0.47 \\
Paired score & 0.62 & 0.45 & 0.53 & 0.42 & 0.91 & 0.75 & 0.88 & 0.69 & 0.77 & 0.58 &  &  &  &  &  &  & 0.74 & 0.58 \\
Pooled score & \textbf{0.96} & 0.87 & \textbf{0.75} & \textbf{0.59} & \textbf{0.94} & \textbf{0.82} & \textbf{0.96} & \textbf{0.88} & 0.93 & 0.76 &  &  &  &  &  &  & \textbf{0.91} & \textbf{0.78} \\
\textbf{ACE} & 0.95 & \textbf{0.88} & 0.70 & 0.54 & 0.91 & 0.77 & \textbf{0.96} & \textbf{0.88} & \textbf{0.94} & \textbf{0.83} &  &  &  &  &  &  & 0.89 & \textbf{0.78} \\
\hline 
 \multicolumn{19}{c}{DEPICT: Silhouette score (Euclidean distance)} \\
 \hline 
Raw score & 0.45 & 0.27 & \textbf{0.75} & \textbf{0.59} & 0.69 & 0.51 & 0.79 & 0.63 & -0.23 & -0.13 &  &  &  &  &  &  & 0.49 & 0.37 \\
Paired score & 0.52 & 0.33 & 0.57 & 0.45 & 0.80 & 0.62 & 0.85 & 0.65 & 0.59 & 0.48 &  &  &  &  &  &  & 0.67 & 0.51 \\
Pooled score & 0.94 & 0.84 & 0.72 & 0.57 & \textbf{0.94} & \textbf{0.82} & 0.96 & 0.88 & 0.92 & 0.75 &  &  &  &  &  &  & \textbf{0.90} & 0.77 \\
\textbf{ACE} & \textbf{0.95} & \textbf{0.87} & 0.63 & 0.49 & 0.91 & 0.78 & \textbf{0.97} & \textbf{0.91} & \textbf{0.95} & \textbf{0.84} &  &  &  &  &  &  & 0.88 & \textbf{0.78} \\
\bottomrule
\end{tabular}
}
\label{tab:acc:hyper}
\end{table*}

%%%%%%%%%%%%%%%%%%%%%%%%%%%%%%%%%%%%%%%%%%%%%%%%%%%%%%%%%%%%%%%%%%%%%%%%%%%%%%%%

The five additional internal measures, Cubic Clustering Criterion (CCC), Dunn index, Cindex, SDbw index, and CDbw index, are presented in Table \ref{tab:app:nmi:hyper}. Both ACE and pooled scores demonstrate superior rank correlation with NMI in comparison to paired scores and raw scores, showcasing their effectiveness. It's important to highlight the practical challenges of obtaining raw scores due to the high dimensional input data, as indicated by the dash mark ($-$). This mark denotes results that could not be obtained because of excessive runtime, memory exhaustion, or computational failure, even though we allowed substantially longer runtime for computing internal metrics on the raw data compared with the embedding data. Furthermore, in certain cases, all four evaluation approaches exhibit negative rank correlation with NMI, suggesting a possible lack of admissible spaces for this metric in the dataset. Additionally, for JULE, the density-based internal measure CDbw shows a noteworthy negative correlation of NMI with paired scores, pooled scores, and ACE scores across several datasets. However, it achieves high correlation on datasets UMist, COIL-20, and COIL-100, which displays non-convex shaped clusters in the output embedding spaces (suggested by Figures \ref{fig:tsne:dav:0} and \ref{fig:tsne:ch:0})). This observation suggests that density-based internal measures may offer more accurate evaluations for non-convex shape clustering results. Furthermore, Table \ref{tab:app:acc:hyper1} presents the rank correlation between different scores and ACC across the internal measures, revealing similar patterns.

\begin{table*}[htbp!]
  \centering
  \caption{Rank consistency with NMI for different evaluation approaches based on five additonal internal measures in Application 1.}
  \resizebox{\textwidth}{!}{
\begin{tabular}{lllllllllllllllllll}
\toprule
{} & \multicolumn{2}{l}{USPS} & \multicolumn{2}{l}{YTF} & \multicolumn{2}{l}{FRGC} & \multicolumn{2}{l}{MNIST-test} & \multicolumn{2}{l}{CMU-PIE} & \multicolumn{2}{l}{UMist} & \multicolumn{2}{l}{COIL-20} & \multicolumn{2}{l}{COIL-100} & \multicolumn{2}{l}{Average} \\
 & $r_s$ & $\tau_B$ & $r_s$ & $\tau_B$ & $r_s$ & $\tau_B$ & $r_s$ & $\tau_B$ & $r_s$ & $\tau_B$ & $r_s$ & $\tau_B$ & $r_s$ & $\tau_B$ & $r_s$ & $\tau_B$ & $r_s$ & $\tau_B$ \\
\midrule
\hline 
 \multicolumn{19}{c}{\emph{JULE}: Cubic clustering criterion} \\
 \hline 
Raw score & - & - & - & - & - & - & - & - & - & - & - & - & - & - & - & - & - & - \\
Paired score & 0.17 & 0.13 & 0.62 & 0.49 & 0.61 & 0.45 & 0.46 & 0.33 & 0.82 & 0.66 & 0.71 & 0.51 & 0.74 & 0.57 & 0.76 & 0.58 & 0.61 & 0.46 \\
Pooled score & 0.84 & 0.68 & 0.92 & 0.80 & 0.30 & 0.22 & 0.82 & 0.67 & 0.94 & 0.81 & 0.80 & 0.59 & 0.61 & 0.46 & 0.89 & 0.73 & 0.77 & 0.62 \\
\textbf{ACE} & 0.87 & 0.72 & 0.93 & 0.83 & 0.23 & 0.15 & 0.82 & 0.65 & 0.98 & 0.91 & 0.84 & 0.64 & 0.93 & 0.78 & 0.93 & 0.80 & 0.82 & 0.68 \\
\hline 
 \multicolumn{19}{c}{\emph{JULE}: Dunn index} \\
 \hline 
Raw score & 0.12 & 0.10 & 0.56 & 0.43 & -0.07 & -0.04 & -0.17 & -0.13 & -0.37 & -0.21 & 0.34 & 0.23 & 0.54 & 0.44 & 0.71 & 0.52 & 0.21 & 0.17 \\
Paired score & -0.23 & -0.16 & 0.59 & 0.42 & 0.42 & 0.29 & -0.23 & -0.14 & 0.90 & 0.74 & 0.37 & 0.28 & 0.58 & 0.43 & 0.37 & 0.25 & 0.35 & 0.26 \\
Pooled score & -0.12 & -0.04 & 0.71 & 0.53 & 0.50 & 0.34 & -0.42 & -0.30 & 0.89 & 0.76 & 0.63 & 0.48 & 0.73 & 0.54 & 0.85 & 0.67 & 0.47 & 0.37 \\
\textbf{ACE} & -0.57 & -0.39 & 0.63 & 0.47 & 0.27 & 0.19 & -0.13 & -0.09 & 0.93 & 0.82 & 0.61 & 0.47 & 0.74 & 0.54 & 0.80 & 0.59 & 0.41 & 0.33 \\
\hline 
 \multicolumn{19}{c}{\emph{JULE}: Cindex} \\
 \hline 
Raw score & 0.49 & 0.37 & 0.27 & 0.20 & -0.46 & -0.31 & 0.17 & 0.14 & -0.81 & -0.68 & 0.10 & 0.07 & 0.50 & 0.36 & 0.80 & 0.62 & 0.13 & 0.10 \\
Paired score & 0.27 & 0.19 & 0.09 & 0.06 & -0.28 & -0.19 & 0.47 & 0.33 & -0.49 & -0.35 & 0.53 & 0.37 & 0.06 & 0.04 & -0.17 & -0.09 & 0.06 & 0.05 \\
Pooled score & 0.65 & 0.45 & 0.67 & 0.52 & 0.02 & 0.02 & 0.73 & 0.57 & -0.11 & -0.08 & 0.58 & 0.42 & 0.51 & 0.37 & 0.76 & 0.57 & 0.48 & 0.36 \\
\textbf{ACE} & 0.78 & 0.62 & 0.20 & 0.13 & -0.16 & -0.11 & 0.83 & 0.67 & -0.55 & -0.35 & 0.58 & 0.42 & 0.77 & 0.58 & 0.69 & 0.52 & 0.39 & 0.31 \\
\hline 
 \multicolumn{19}{c}{\emph{JULE}: SDbw index} \\
 \hline 
Raw score & -0.44 & -0.26 & -0.59 & -0.43 & -0.18 & -0.11 & -0.76 & -0.58 & -0.99 & -0.92 & -0.70 & -0.53 & -0.17 & -0.07 & - & - & -0.55 & -0.42 \\
Paired score & -0.16 & -0.08 & -0.54 & -0.38 & -0.12 & -0.08 & -0.44 & -0.30 & -0.25 & -0.16 & 0.69 & 0.48 & 0.52 & 0.37 & 0.24 & 0.22 & -0.01 & 0.01 \\
Pooled score & -0.38 & -0.24 & -0.62 & -0.45 & 0.18 & 0.12 & -0.56 & -0.40 & -0.76 & -0.65 & 0.24 & 0.17 & 0.10 & 0.13 & 0.61 & 0.41 & -0.15 & -0.11 \\
\textbf{ACE} & -0.35 & -0.18 & -0.64 & -0.45 & 0.47 & 0.36 & -0.18 & -0.11 & -0.52 & -0.52 & 0.64 & 0.46 & 0.61 & 0.45 & 0.74 & 0.54 & 0.10 & 0.07 \\
\hline 
 \multicolumn{19}{c}{\emph{JULE}: CDbw index} \\
 \hline 
Raw score & -0.26 & -0.21 & - & - & - & - & - & - & -0.31 & -0.26 & - & - & - & - & - & - & -0.28 & -0.23 \\
Paired score & -0.24 & -0.16 & -0.23 & -0.17 & -0.38 & -0.27 & -0.60 & -0.43 & -0.07 & -0.05 & 0.07 & 0.06 & 0.33 & 0.21 & 0.50 & 0.35 & -0.08 & -0.06 \\
Pooled score & -0.38 & -0.25 & -0.55 & -0.40 & 0.26 & 0.17 & -0.73 & -0.54 & 0.71 & 0.63 & 0.73 & 0.52 & 0.85 & 0.68 & 0.90 & 0.72 & 0.22 & 0.19 \\
\textbf{ACE} & -0.31 & -0.20 & -0.58 & -0.41 & 0.31 & 0.21 & -0.70 & -0.52 & 0.62 & 0.52 & 0.75 & 0.55 & 0.80 & 0.61 & 0.97 & 0.85 & 0.23 & 0.20 \\
\hline 
 \multicolumn{19}{c}{\emph{DEPICT}: Cubic clustering criterion} \\
 \hline 
Raw score & - & - & - & - & - & - & - & - & - & - &  &  &  &  &  &  & - & - \\
Paired score & 0.74 & 0.52 & 0.50 & 0.35 & 0.92 & 0.79 & 0.89 & 0.72 & 0.89 & 0.70 &  &  &  &  &  &  & 0.79 & 0.62 \\
Pooled score & 0.96 & 0.83 & 0.61 & 0.48 & 0.92 & 0.82 & 0.97 & 0.88 & 0.95 & 0.84 &  &  &  &  &  &  & 0.88 & 0.77 \\
\textbf{ACE} & 0.95 & 0.83 & 0.76 & 0.62 & 0.95 & 0.83 & 0.96 & 0.87 & 0.95 & 0.84 &  &  &  &  &  &  & 0.91 & 0.80 \\
\hline 
 \multicolumn{19}{c}{\emph{DEPICT}: Dunn index} \\
 \hline 
Raw score & 0.56 & 0.41 & 0.42 & 0.26 & 0.59 & 0.47 & 0.88 & 0.73 & 0.15 & 0.05 &  &  &  &  &  &  & 0.52 & 0.39 \\
Paired score & 0.85 & 0.66 & 0.55 & 0.41 & 0.81 & 0.62 & 0.91 & 0.78 & 0.39 & 0.29 &  &  &  &  &  &  & 0.70 & 0.55 \\
Pooled score & 0.85 & 0.67 & 0.75 & 0.59 & 0.82 & 0.62 & 0.91 & 0.74 & 0.81 & 0.65 &  &  &  &  &  &  & 0.83 & 0.66 \\
\textbf{ACE} & 0.92 & 0.78 & 0.68 & 0.53 & 0.65 & 0.50 & 0.84 & 0.67 & 0.94 & 0.80 &  &  &  &  &  &  & 0.80 & 0.66 \\
\hline 
 \multicolumn{19}{c}{\emph{DEPICT}: Cindex} \\
 \hline 
Raw score & -0.27 & -0.19 & -0.35 & -0.27 & 0.52 & 0.41 & 0.09 & 0.06 & -0.23 & -0.28 &  &  &  &  &  &  & -0.05 & -0.05 \\
Paired score & 0.53 & 0.36 & -0.03 & -0.02 & 0.24 & 0.19 & 0.44 & 0.35 & -0.18 & 0.01 &  &  &  &  &  &  & 0.20 & 0.18 \\
Pooled score & 0.70 & 0.54 & 0.53 & 0.40 & 0.88 & 0.74 & 0.92 & 0.80 & 0.53 & 0.37 &  &  &  &  &  &  & 0.71 & 0.57 \\
\textbf{ACE} & 0.90 & 0.75 & 0.61 & 0.45 & 0.91 & 0.77 & -0.39 & -0.27 & 0.73 & 0.57 &  &  &  &  &  &  & 0.55 & 0.45 \\
\hline 
 \multicolumn{19}{c}{\emph{DEPICT}: SDbw index} \\
 \hline 
Raw score & 0.18 & 0.09 & 0.64 & 0.48 & 0.57 & 0.43 & 0.14 & 0.09 & -0.94 & -0.80 &  &  &  &  &  &  & 0.12 & 0.06 \\
Paired score & 0.84 & 0.67 & 0.55 & 0.36 & 0.91 & 0.77 & 0.89 & 0.74 & 0.57 & 0.46 &  &  &  &  &  &  & 0.75 & 0.60 \\
Pooled score & 0.93 & 0.79 & 0.62 & 0.48 & 0.75 & 0.61 & 0.96 & 0.87 & 0.67 & 0.49 &  &  &  &  &  &  & 0.79 & 0.65 \\
\textbf{ACE} & 0.93 & 0.79 & 0.64 & 0.48 & 0.89 & 0.75 & 0.97 & 0.90 & 0.95 & 0.84 &  &  &  &  &  &  & 0.87 & 0.75 \\
\hline 
 \multicolumn{19}{c}{\emph{DEPICT}: CDbw index} \\
 \hline 
Raw score & - & - & - & - & - & - & - & - & 0.38 & 0.29 &  &  &  &  &  &  & 0.38 & 0.29 \\
Paired score & 0.48 & 0.35 & 0.61 & 0.40 & 0.83 & 0.66 & 0.63 & 0.46 & 0.88 & 0.70 &  &  &  &  &  &  & 0.69 & 0.51 \\
Pooled score & 0.95 & 0.86 & 0.64 & 0.48 & 0.78 & 0.63 & 0.50 & 0.32 & 0.92 & 0.79 &  &  &  &  &  &  & 0.76 & 0.62 \\
\textbf{ACE} & 0.69 & 0.53 & 0.64 & 0.48 & 0.81 & 0.66 & 0.50 & 0.32 & 0.94 & 0.83 &  &  &  &  &  &  & 0.72 & 0.56 \\
\bottomrule
\end{tabular}
}

\label{tab:app:nmi:hyper}
\end{table*}

\newpage

\begin{table*}[htbp!]
  \centering
  \caption{Rank consistency with ACC for different evaluation approaches based on five additonal internal measures in Application 1.}
  \resizebox{\textwidth}{!}{
\begin{tabular}{lllllllllllllllllll}
\toprule
{} & \multicolumn{2}{l}{USPS} & \multicolumn{2}{l}{YTF} & \multicolumn{2}{l}{FRGC} & \multicolumn{2}{l}{MNIST-test} & \multicolumn{2}{l}{CMU-PIE} & \multicolumn{2}{l}{UMist} & \multicolumn{2}{l}{COIL-20} & \multicolumn{2}{l}{COIL-100} & \multicolumn{2}{l}{Average} \\
{} & $r_s$ & $\tau_B$ & $r_s$ & $\tau_B$ & $r_s$ & $\tau_B$ &      $r_s$ & $\tau_B$ &   $r_s$ & $\tau_B$ & $r_s$ & $\tau_B$ &   $r_s$ & $\tau_B$ &    $r_s$ & $\tau_B$ &   $r_s$ & $\tau_B$ \\
\midrule
\hline 
 \multicolumn{19}{c}{\emph{JULE}: Cubic clustering criterion} \\
 \hline 
Raw score & - & - & - & - & - & - & - & - & - & - & - & - & - & - & - & - & - & - \\
Paired score & 0.04 & 0.05 & 0.43 & 0.30 & 0.52 & 0.35 & 0.30 & 0.20 & 0.84 & 0.67 & 0.65 & 0.48 & 0.76 & 0.58 & 0.73 & 0.54 & 0.53 & 0.40 \\
Pooled score & 0.91 & 0.78 & 0.76 & 0.59 & 0.33 & 0.23 & 0.91 & 0.77 & 0.95 & 0.83 & 0.80 & 0.59 & 0.57 & 0.42 & 0.90 & 0.76 & 0.77 & 0.62 \\
\textbf{ACE} & 0.94 & 0.82 & 0.76 & 0.59 & 0.21 & 0.14 & 0.88 & 0.72 & 0.99 & 0.92 & 0.84 & 0.65 & 0.90 & 0.74 & 0.93 & 0.79 & 0.81 & 0.67 \\
\hline 
 \multicolumn{19}{c}{\emph{JULE}: Dunn index} \\
 \hline 
Raw score & 0.02 & 0.02 & 0.21 & 0.17 & -0.18 & -0.13 & -0.09 & -0.08 & -0.33 & -0.20 & 0.39 & 0.25 & 0.50 & 0.40 & 0.62 & 0.44 & 0.14 & 0.11 \\
Paired score & -0.36 & -0.24 & 0.29 & 0.19 & 0.55 & 0.39 & -0.20 & -0.14 & 0.89 & 0.73 & 0.31 & 0.24 & 0.56 & 0.42 & 0.40 & 0.28 & 0.31 & 0.23 \\
Pooled score & -0.35 & -0.16 & 0.35 & 0.23 & 0.62 & 0.44 & -0.52 & -0.37 & 0.87 & 0.72 & 0.59 & 0.45 & 0.72 & 0.53 & 0.73 & 0.54 & 0.38 & 0.30 \\
\textbf{ACE} & -0.77 & -0.56 & 0.38 & 0.25 & 0.46 & 0.33 & -0.10 & -0.09 & 0.92 & 0.79 & 0.58 & 0.44 & 0.73 & 0.54 & 0.67 & 0.49 & 0.36 & 0.27 \\
\hline 
 \multicolumn{19}{c}{\emph{JULE}: Cindex} \\
 \hline 
Raw score & 0.56 & 0.45 & 0.35 & 0.24 & -0.52 & -0.37 & 0.24 & 0.23 & -0.80 & -0.67 & 0.14 & 0.10 & 0.56 & 0.40 & 0.78 & 0.61 & 0.17 & 0.13 \\
Paired score & 0.13 & 0.10 & -0.09 & -0.06 & -0.47 & -0.32 & 0.33 & 0.21 & -0.56 & -0.40 & 0.54 & 0.38 & 0.09 & 0.06 & -0.21 & -0.14 & -0.03 & -0.02 \\
Pooled score & 0.82 & 0.63 & 0.62 & 0.45 & -0.24 & -0.16 & 0.87 & 0.67 & -0.06 & -0.05 & 0.67 & 0.51 & 0.60 & 0.45 & 0.77 & 0.59 & 0.51 & 0.38 \\
\textbf{ACE} & 0.85 & 0.70 & 0.17 & 0.13 & -0.39 & -0.28 & 0.93 & 0.76 & -0.52 & -0.32 & 0.67 & 0.50 & 0.81 & 0.63 & 0.68 & 0.52 & 0.40 & 0.33 \\
\hline 
 \multicolumn{19}{c}{\emph{JULE}: SDbw index} \\
 \hline 
Raw score & -0.53 & -0.33 & -0.33 & -0.24 & 0.04 & 0.05 & -0.89 & -0.72 & -1.00 & -0.97 & -0.73 & -0.55 & -0.14 & -0.07 & - & - & -0.51 & -0.40 \\
Paired score & -0.32 & -0.20 & -0.30 & -0.19 & -0.35 & -0.24 & -0.61 & -0.42 & -0.31 & -0.20 & 0.61 & 0.42 & 0.56 & 0.39 & 0.14 & 0.10 & -0.07 & -0.04 \\
Pooled score & -0.58 & -0.31 & -0.39 & -0.26 & 0.51 & 0.39 & -0.70 & -0.54 & -0.75 & -0.64 & 0.11 & 0.08 & 0.07 & 0.10 & 0.67 & 0.46 & -0.13 & -0.09 \\
\textbf{ACE} & -0.51 & -0.22 & -0.39 & -0.26 & 0.69 & 0.48 & -0.25 & -0.17 & -0.50 & -0.50 & 0.54 & 0.38 & 0.57 & 0.40 & 0.80 & 0.59 & 0.12 & 0.09 \\
\hline 
 \multicolumn{19}{c}{\emph{JULE}: CDbw index} \\
 \hline 
Raw score & -0.27 & -0.22 & - & - & - & - & - & - & -0.31 & -0.26 & - & - & - & - & - & - & -0.29 & -0.24 \\
Paired score & -0.41 & -0.28 & -0.43 & -0.30 & -0.48 & -0.31 & -0.73 & -0.54 & -0.12 & -0.07 & 0.10 & 0.08 & 0.35 & 0.22 & 0.41 & 0.28 & -0.16 & -0.12 \\
Pooled score & -0.62 & -0.40 & -0.29 & -0.18 & 0.50 & 0.39 & -0.88 & -0.67 & 0.73 & 0.66 & 0.62 & 0.45 & 0.83 & 0.64 & 0.89 & 0.72 & 0.22 & 0.20 \\
\textbf{ACE} & -0.55 & -0.34 & -0.36 & -0.26 & 0.59 & 0.44 & -0.85 & -0.65 & 0.58 & 0.51 & 0.67 & 0.50 & 0.75 & 0.55 & 0.90 & 0.75 & 0.22 & 0.19 \\
\hline 
 \multicolumn{19}{c}{\emph{DEPICT}: Cubic clustering criterion} \\
 \hline 
Raw score & - & - & - & - & - & - & - & - & - & - &  &  &  &  &  &  & - & - \\
Paired score & 0.53 & 0.35 & 0.59 & 0.43 & 0.94 & 0.81 & 0.88 & 0.69 & 0.87 & 0.68 &  &  &  &  &  &  & 0.76 & 0.59 \\
Pooled score & 0.96 & 0.87 & 0.54 & 0.44 & 0.94 & 0.82 & 0.96 & 0.87 & 0.96 & 0.85 &  &  &  &  &  &  & 0.87 & 0.77 \\
\textbf{ACE} & 0.96 & 0.87 & 0.65 & 0.53 & 0.93 & 0.83 & 0.97 & 0.91 & 0.96 & 0.85 &  &  &  &  &  &  & 0.89 & 0.80 \\
\hline 
 \multicolumn{19}{c}{\emph{DEPICT}: Dunn index} \\
 \hline 
Raw score & 0.56 & 0.39 & 0.42 & 0.27 & 0.68 & 0.50 & 0.80 & 0.64 & 0.10 & 0.03 &  &  &  &  &  &  & 0.51 & 0.37 \\
Paired score & 0.70 & 0.52 & 0.48 & 0.35 & 0.85 & 0.67 & 0.86 & 0.71 & 0.36 & 0.28 &  &  &  &  &  &  & 0.65 & 0.50 \\
Pooled score & 0.70 & 0.53 & 0.66 & 0.50 & 0.84 & 0.67 & 0.88 & 0.70 & 0.80 & 0.63 &  &  &  &  &  &  & 0.78 & 0.61 \\
\textbf{ACE} & 0.79 & 0.63 & 0.62 & 0.49 & 0.67 & 0.53 & 0.77 & 0.61 & 0.93 & 0.79 &  &  &  &  &  &  & 0.76 & 0.61 \\
\hline 
 \multicolumn{19}{c}{\emph{DEPICT}: Cindex} \\
 \hline 
Raw score & -0.31 & -0.20 & -0.23 & -0.18 & 0.45 & 0.36 & 0.18 & 0.10 & -0.22 & -0.25 &  &  &  &  &  &  & -0.02 & -0.03 \\
Paired score & 0.57 & 0.40 & 0.13 & 0.10 & 0.23 & 0.16 & 0.49 & 0.39 & -0.18 & -0.03 &  &  &  &  &  &  & 0.25 & 0.20 \\
Pooled score & 0.91 & 0.74 & 0.61 & 0.46 & 0.92 & 0.77 & 0.95 & 0.84 & 0.55 & 0.41 &  &  &  &  &  &  & 0.79 & 0.64 \\
\textbf{ACE} & 0.92 & 0.82 & 0.68 & 0.52 & 0.88 & 0.71 & -0.35 & -0.20 & 0.76 & 0.63 &  &  &  &  &  &  & 0.58 & 0.49 \\
\hline 
 \multicolumn{19}{c}{\emph{DEPICT}: SDbw index} \\
 \hline 
Raw score & 0.27 & 0.12 & 0.54 & 0.39 & 0.72 & 0.59 & 0.21 & 0.10 & -0.93 & -0.80 &  &  &  &  &  &  & 0.16 & 0.08 \\
Paired score & 0.66 & 0.48 & 0.51 & 0.35 & 0.90 & 0.74 & 0.88 & 0.70 & 0.59 & 0.48 &  &  &  &  &  &  & 0.71 & 0.55 \\
Pooled score & 0.98 & 0.91 & 0.51 & 0.39 & 0.74 & 0.61 & 0.97 & 0.88 & 0.67 & 0.50 &  &  &  &  &  &  & 0.77 & 0.66 \\
\textbf{ACE} & 0.98 & 0.91 & 0.52 & 0.39 & 0.87 & 0.72 & 0.97 & 0.91 & 0.95 & 0.85 &  &  &  &  &  &  & 0.86 & 0.76 \\
\hline 
 \multicolumn{19}{c}{\emph{DEPICT}: CDbw index} \\
 \hline 
Raw score & - & - & - & - & - & - & - & - & 0.40 & 0.31 &  &  &  &  &  &  & 0.40 & 0.31 \\
Paired score & 0.53 & 0.39 & 0.55 & 0.36 & 0.83 & 0.66 & 0.68 & 0.50 & 0.90 & 0.72 &  &  &  &  &  &  & 0.70 & 0.53 \\
Pooled score & 0.86 & 0.74 & 0.54 & 0.41 & 0.82 & 0.66 & 0.44 & 0.31 & 0.92 & 0.76 &  &  &  &  &  &  & 0.71 & 0.58 \\
\textbf{ACE} & 0.79 & 0.62 & 0.54 & 0.39 & 0.86 & 0.69 & 0.44 & 0.31 & 0.94 & 0.80 &  &  &  &  &  &  & 0.71 & 0.56 \\
\bottomrule
\end{tabular}
}
\label{tab:app:acc:hyper1}
\end{table*}

\clearpage
\newpage \subsection{Application 1 - qualitative analysis} \label{app:exp:ar:qa1}
In this section, we present the qualitative analysis results for Application 1 using both JULE and DEPICT. Graphs depicting the rank correlation between the retained spaces after the multimodality test, based on different internal measures, are provided as follows: for JULE, Figures \ref{fig:graph:dav:0} (Davies-Bouldin index), \ref{fig:graph:ch:0} (Calinski-Harabasz index), \ref{fig:graph:cosine:0} (Silhouette score with Cosine distance), and \ref{fig:graph:euclidean:0} (Silhouette score with Euclidean distance); for DEPICT, Figures \ref{fig:graph:dav:1} (Davies-Bouldin index), \ref{fig:graph:ch:1} (Calinski-Harabasz index), \ref{fig:graph:cosine:1} (Silhouette score with Cosine distance), and \ref{fig:graph:euclidean:1} (Silhouette score with Euclidean distance).  In each graph, spaces grouped together by a density-based clustering of rank correlations share the same color, while outlier spaces are uniformly colored in grey.

These figures reveal distinct grouping behaviors within the retained spaces post multimodality test in Phase 1 of stage-wise grouping. In JULE, where approximately $40$ models (or spaces) are generated, multiple groups are often detected. However, a few instances, such as Figure \ref{fig:graph:dav:0} (b), (d), (e), Figure \ref{fig:graph:ch:0} (b), Figure \ref{fig:graph:cosine:0} (b), and Figure \ref{fig:graph:euclidean:0} (b), only a single group is identified. In contrast, DEPICT, which produces around $18$ spaces, tends to yield more instances with a single group. Across these figures, the same retained spaces can exhibit different grouping patterns depending on the internal measure used. As noted in Supplementary Section \ref{app:index}, different measures assess clustering quality differently: the Silhouette score emphasizes individual data points regarding relationships to their own and other clusters, the Davies-Bouldin index considers the overall compactness and separation across clusters, and the Calinski-Harabasz index compares between-cluster variance with within-cluster variance. These differing criteria explain the variations in observed grouping behavior. Notably, when using the Silhouette score, we compare clustering based on cosine and Euclidean distances. The results show that grouping behavior remains more consistent across these two distance metrics than across different internal measures, suggesting that the choice of internal measure has a greater impact than the choice of distance metric.  

We also utilize t-SNE plots \citep{van2008visualizing} to visualize the discriminative capability of embedding spaces between the finally selected embedding spaces by ACE and the spaces excluded by ACE. The t-SNE algorithm, known for its effectiveness in preserving the local structure of data and maintaining relative distances between neighboring points in high-dimensional space, is employed to create a non-linear mapping from the embedding space to a 2-dimensional feature space for visualization. We present this comparison for Application 1 with JULE based on different internal measures in Figures \ref{fig:tsne:dav:0} (Davies-Bouldin index), \ref{fig:tsne:ch:0} (Calinski-Harabasz index), \ref{fig:tsne:cosine:0} (Silhouette score with Cosine distance), and \ref{fig:tsne:euclidean:0} (Silhouette score with Euclidean distance). Similarly, in Figures \ref{fig:tsne:dav:1} (Davies-Bouldin index), \ref{fig:tsne:ch:1} (Calinski-Harabasz index), \ref{fig:tsne:cosine:1} (Silhouette score with Cosine distance), and \ref{fig:tsne:euclidean:1} (Silhouette score with Euclidean distance), we provide the comparison between selected and excluded embedding spaces for DEPICT. In each figure, we plot and compare one selected space with an excluded space for each dataset. In each subfigure, different colors correspond to different true clusters. Due to space constraints and solely for visualization purposes, we selected one representative space from the retained spaces and one from the excluded spaces for a concise comparison.

Across these figures, it is evident that the selected spaces exhibit more compact and well-separated clusters of data points aligned with their true cluster labels. In contrast, many of the excluded spaces demonstrate poor clustering behavior. For instance, in the case of JULE, the comparison between (o) and (p) in Figures \ref{fig:tsne:dav:0}, \ref{fig:tsne:ch:0}, \ref{fig:tsne:cosine:0}, and \ref{fig:tsne:euclidean:0} reveals that the selected spaces showcase clear separation between different clusters, while some excluded spaces exhibit multiple areas with intermixed clusters. Similarly, the comparison between (e) and (f) in Figures \ref{fig:tsne:dav:0}, \ref{fig:tsne:ch:0}, \ref{fig:tsne:cosine:0}, and \ref{fig:tsne:euclidean:0} highlights that the selected spaces present regular cluster shapes, whereas excluded spaces show irregular shapes resembling strings of different clusters. This phenomenon is consistent for DEPICT as well. For instance, the comparison between (g) and (h) in Figures \ref{fig:tsne:dav:1}, \ref{fig:tsne:ch:1}, \ref{fig:tsne:cosine:1}, and \ref{fig:tsne:euclidean:1} reveals that some excluded spaces lack clear clustering behavior, whereas the selected spaces exhibit compact and well-separated clusters. Similarly, between (i) and (j) in Figures \ref{fig:tsne:dav:1}, \ref{fig:tsne:ch:1}, \ref{fig:tsne:cosine:1}, and \ref{fig:tsne:euclidean:1}, the selected spaces demonstrate well-separated clusters, while the excluded spaces group points from different true clusters into the same cluster. \\

\begin{figure}[H]
\centering
\subfigure[USPS]{\includegraphics[width = 0.3\linewidth]{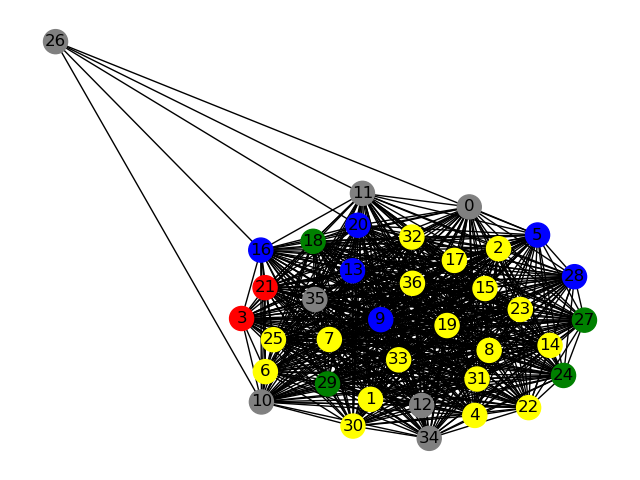}}
\subfigure[UMist]{\includegraphics[width = 0.3\linewidth]{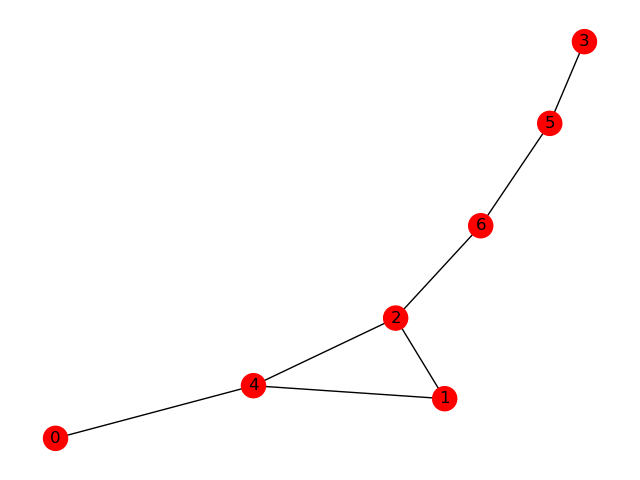}}
\subfigure[COIL-20]{\includegraphics[width = 0.3\linewidth]{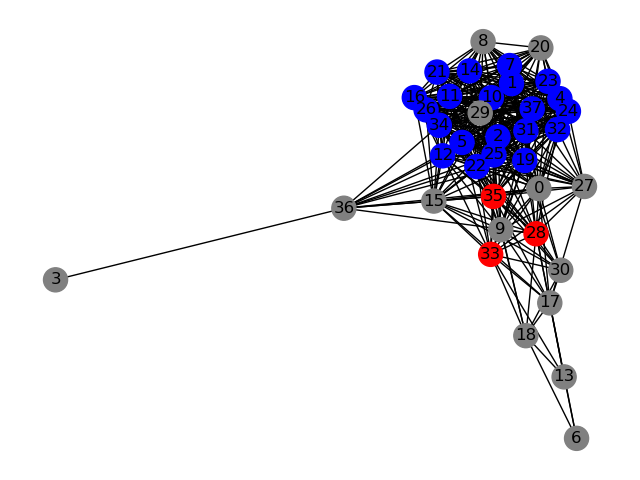}}
\subfigure[COIL-100]{\includegraphics[width = 0.3\linewidth]{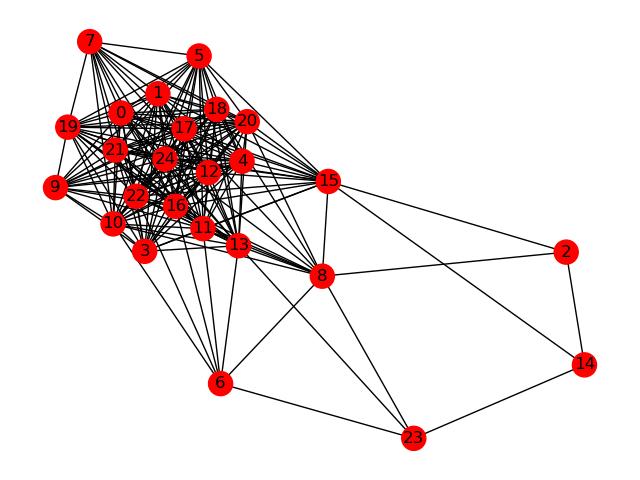}}
\subfigure[YTF]{\includegraphics[width = 0.3\linewidth]{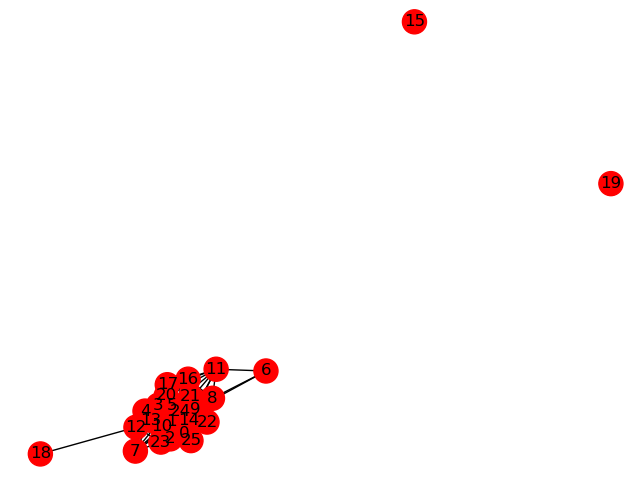}}
\subfigure[FRGC]{\includegraphics[width = 0.3\linewidth]{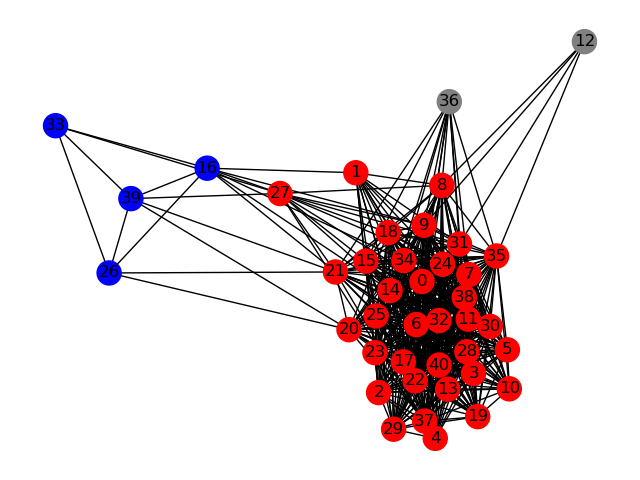}}
\subfigure[MNIST-test]{\includegraphics[width = 0.3\linewidth]{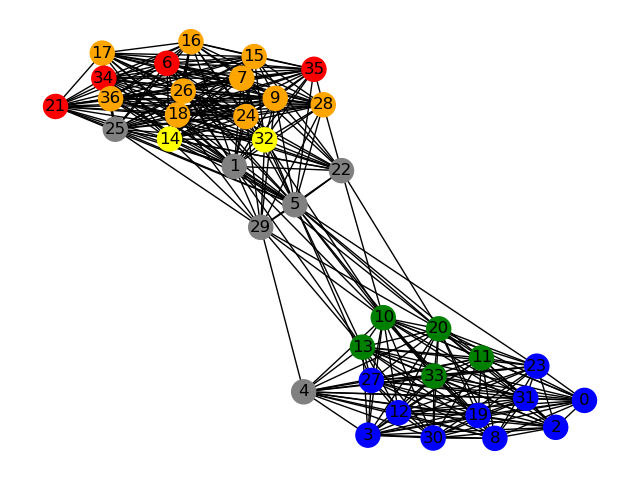}}
\subfigure[CMU-PIE]{\includegraphics[width = 0.3\linewidth]{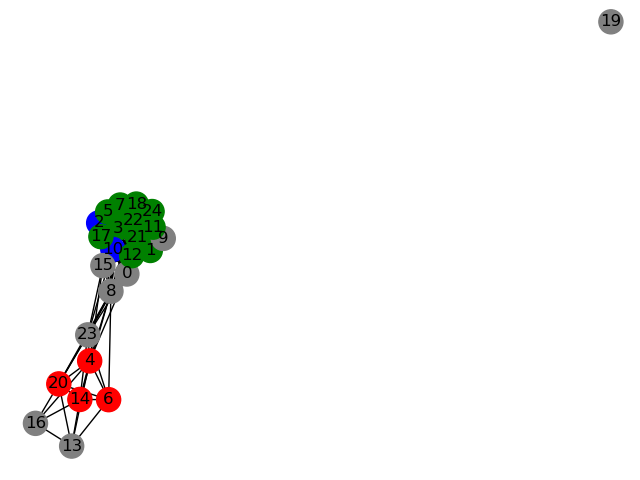}}
\caption{Graph depicting rank correlation based on Davies-Bouldin index among embedding spaces for Application 1 with JULE. Each node represents an embedding space, and each edge signifies a significant rank correlation. Nodes sharing the same color belong to the same Phase 1 group.}
\label{fig:graph:dav:0}
\end{figure}
%%%%%%%%%%%%%%%%%%%%%%%%%
\begin{figure}[htbp!]
\centering
\subfigure[Selected space (USPS)]{\includegraphics[width = 0.24\linewidth]{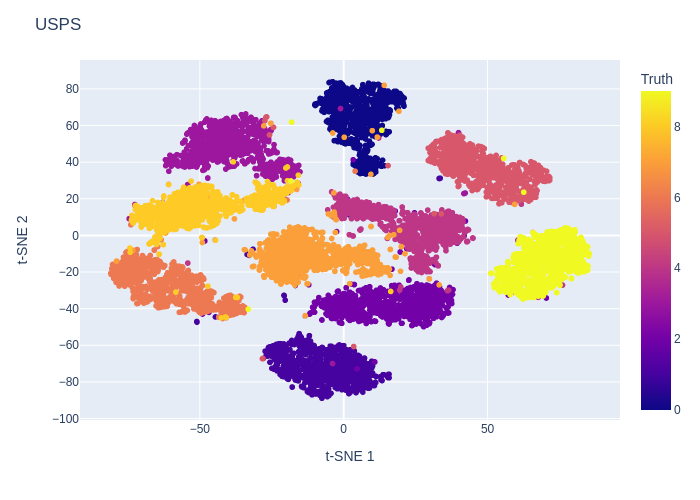}}
\subfigure[Excluded space (USPS)]{\includegraphics[width = 0.24\linewidth]{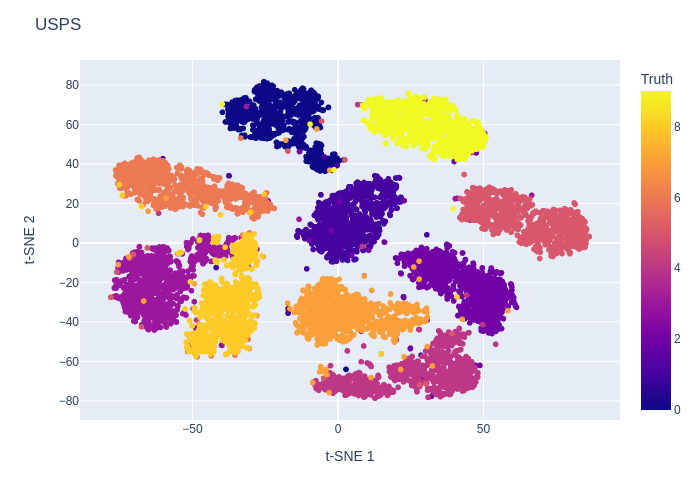}}
\subfigure[Selected space (UMist)]{\includegraphics[width = 0.24\linewidth]{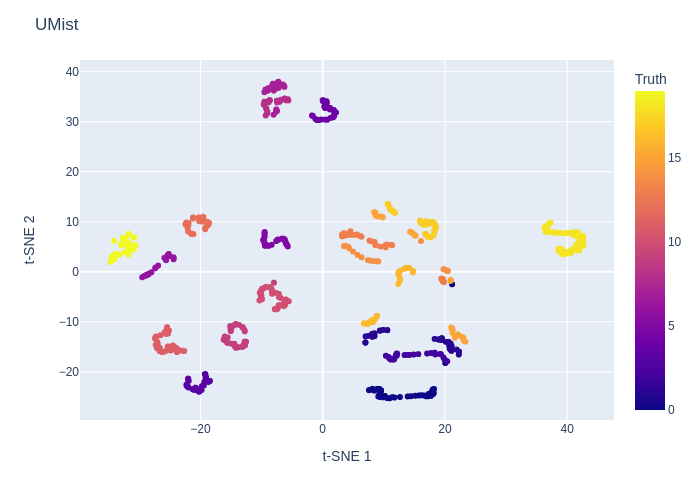}}
\subfigure[Excluded space (UMist)]{\includegraphics[width = 0.24\linewidth]{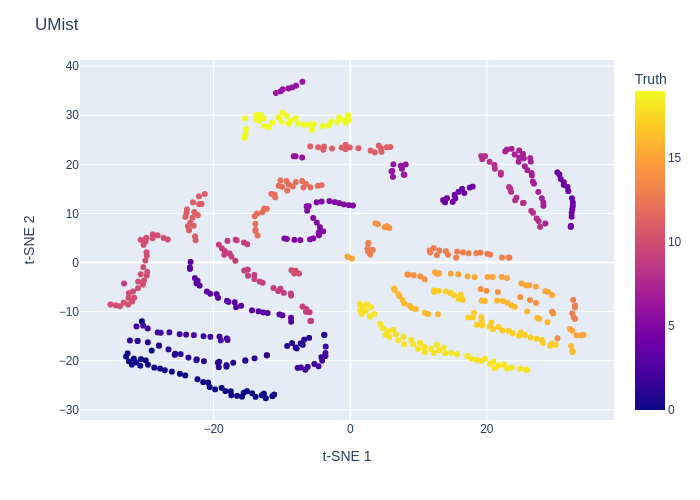}}
\subfigure[Selected space (COIL-20)]{\includegraphics[width = 0.24\linewidth]{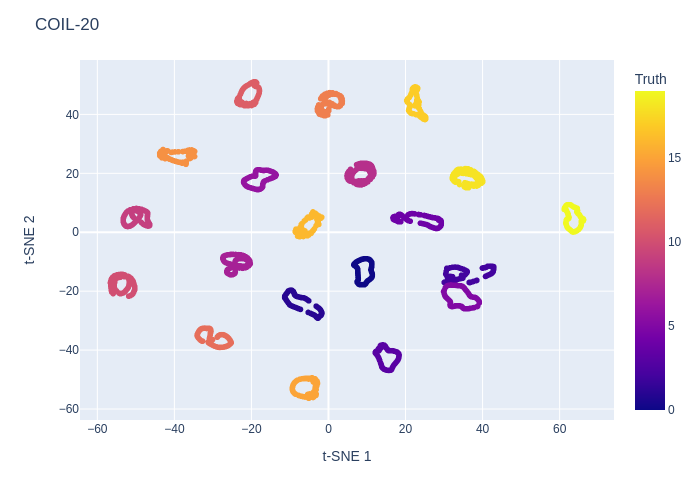}}
\subfigure[Excluded space (COIL-20)]{\includegraphics[width = 0.24\linewidth]{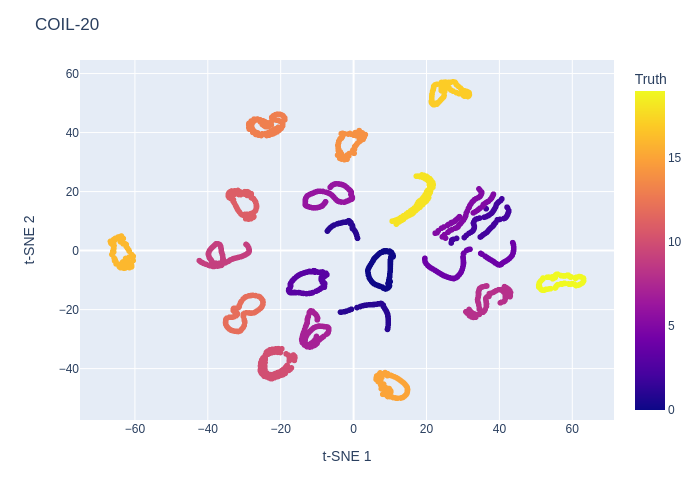}}
\subfigure[Selected space (COIL-100)]{\includegraphics[width = 0.24\linewidth]{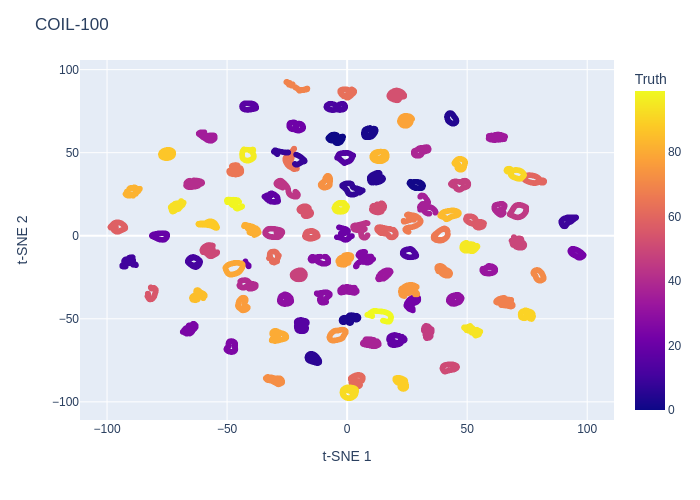}}
\subfigure[Excluded space (COIL-100)]{\includegraphics[width = 0.24\linewidth]{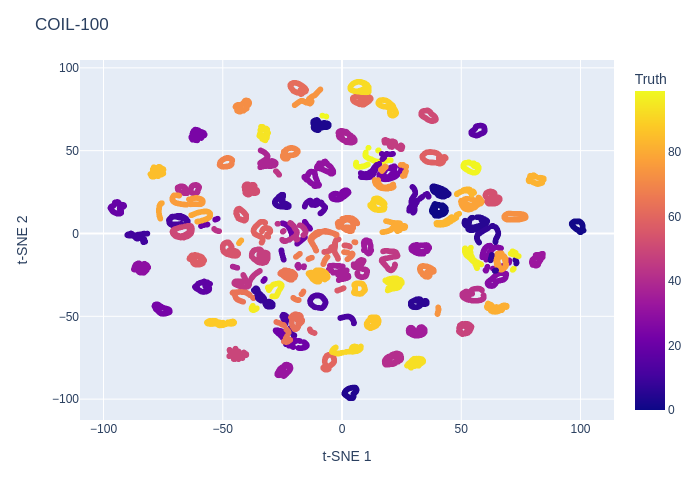}}
\subfigure[Selected space (YTF)]{\includegraphics[width = 0.24\linewidth]{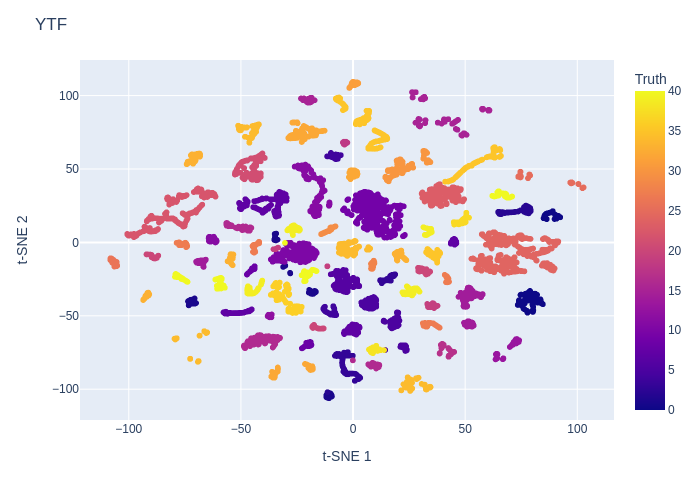}}
\subfigure[Excluded space (YTF)]{\includegraphics[width = 0.24\linewidth]{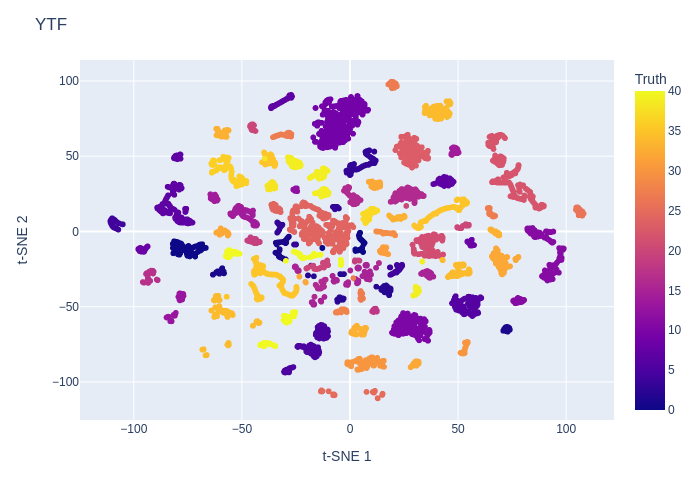}}
\subfigure[Selected space (FRGC)]{\includegraphics[width = 0.24\linewidth]{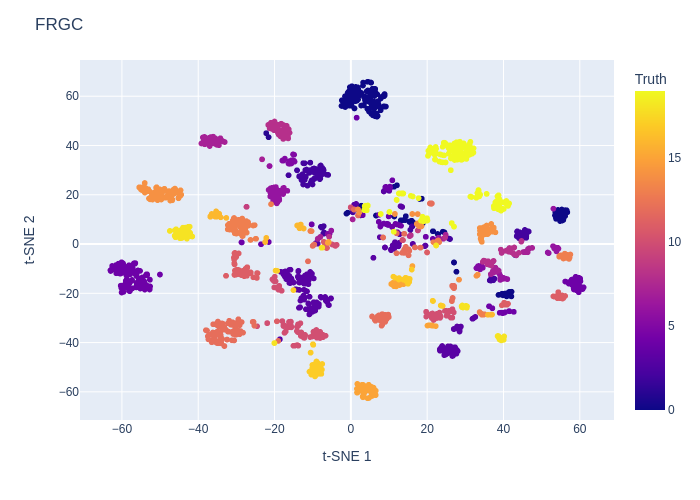}}
\subfigure[Excluded space (FRGC)]{\includegraphics[width = 0.24\linewidth]{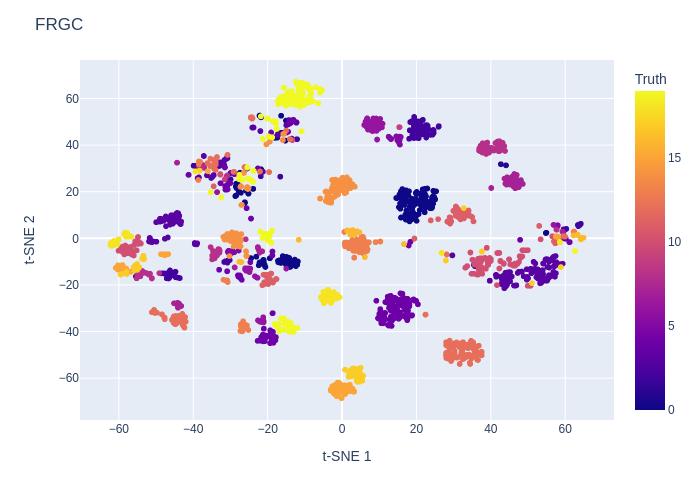}}
\subfigure[Selected space (MNIST-test)]{\includegraphics[width = 0.24\linewidth]{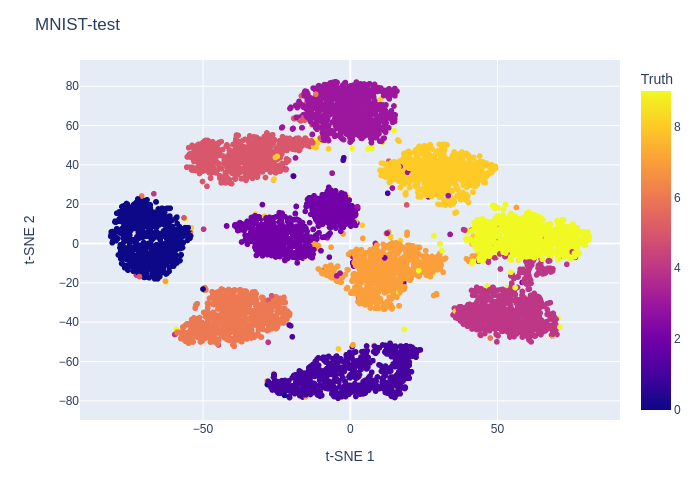}}
\subfigure[Excluded space (MNIST-test)]{\includegraphics[width = 0.24\linewidth]{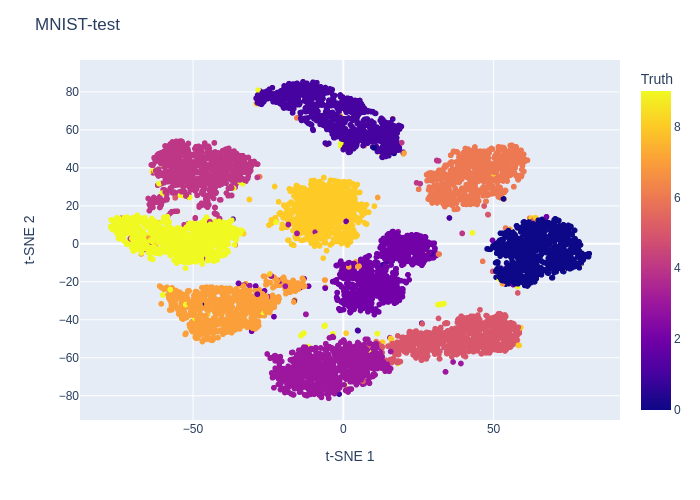}}
\subfigure[Selected space (CMU-PIE)]{\includegraphics[width = 0.24\linewidth]{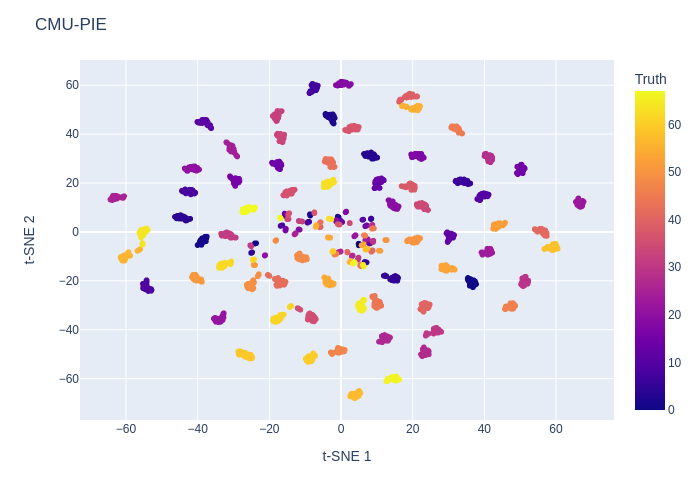}}
\subfigure[Excluded space (CMU-PIE)]{\includegraphics[width = 0.24\linewidth]{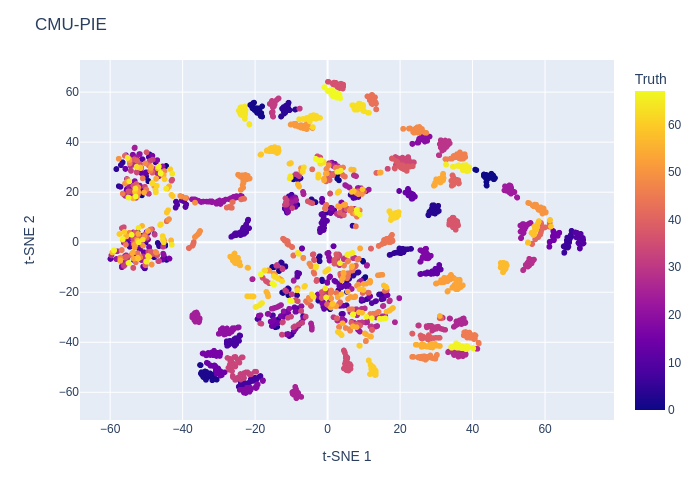}}
\caption{t-SNE visualization illustrating the selected embedding spaces from ACE in comparison to those excluded from ACE, based on Davies-Bouldin index, for Application 1 with JULE. Each data point in the visualizations is assigned a color corresponding to its true cluster label.}
\label{fig:tsne:dav:0}
\end{figure}
%%%%%%%%%%%%%%%%%%%%%%%%%
\begin{figure}[htbp!]
\centering
\subfigure[USPS]{\includegraphics[width = 0.3\linewidth]{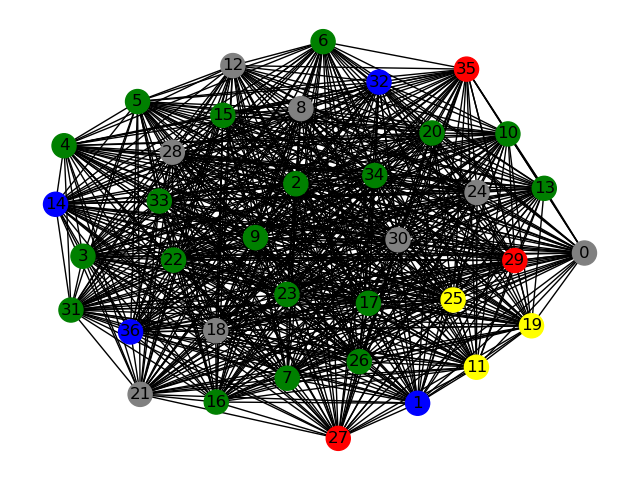}}
\subfigure[UMist]{\includegraphics[width = 0.3\linewidth]{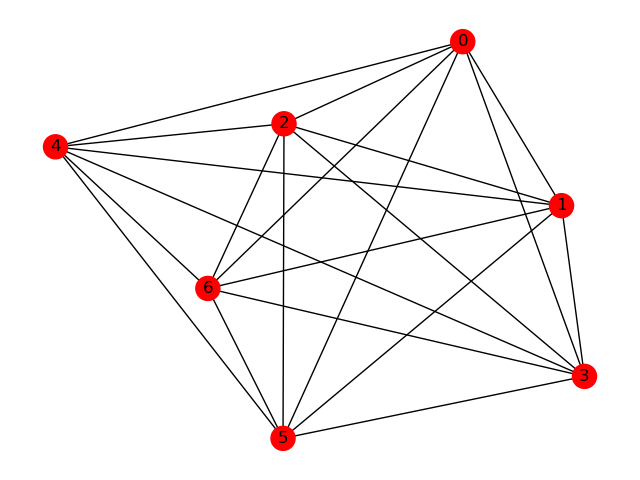}}
\subfigure[COIL-20]{\includegraphics[width = 0.3\linewidth]{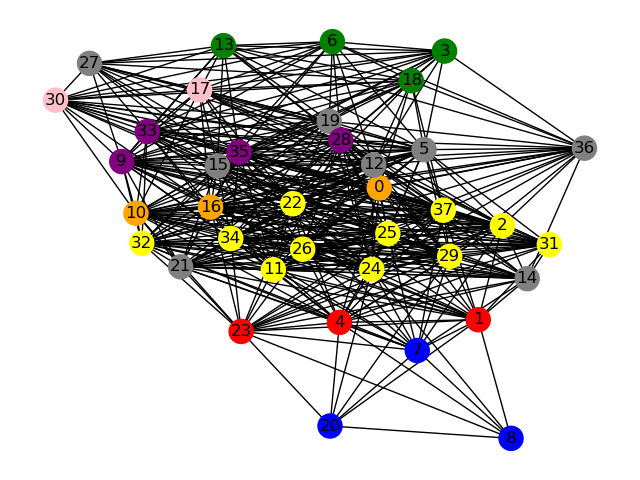}}
\subfigure[COIL-100]{\includegraphics[width = 0.3\linewidth]{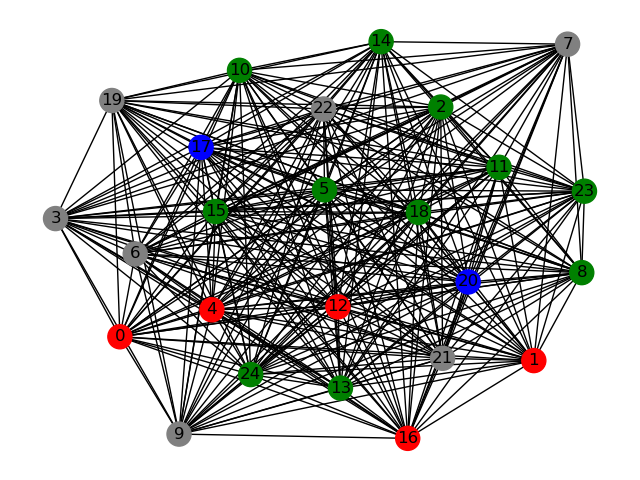}}
\subfigure[YTF]{\includegraphics[width = 0.3\linewidth]{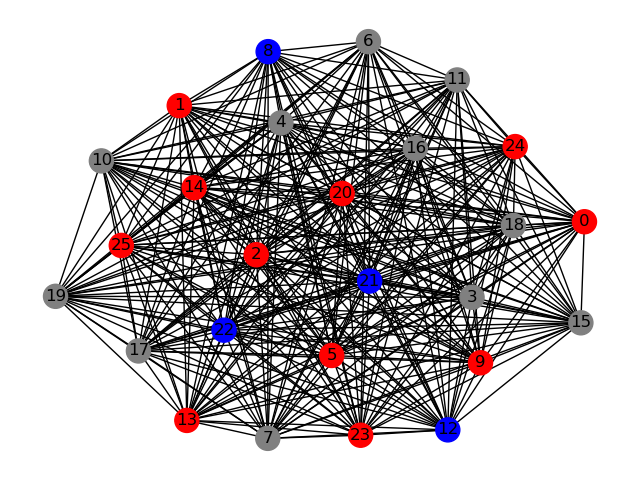}}
\subfigure[FRGC]{\includegraphics[width = 0.3\linewidth]{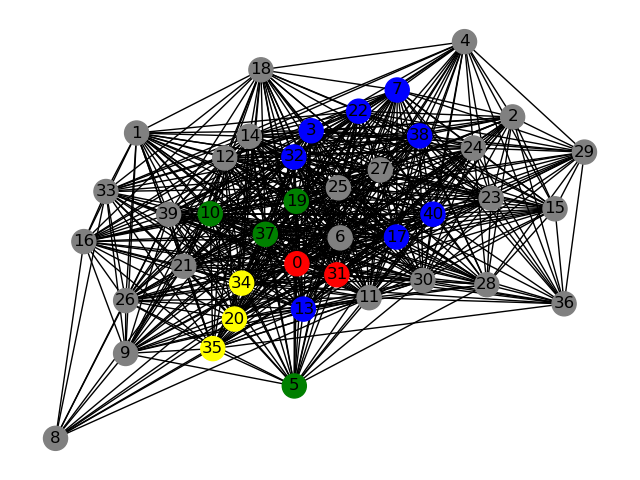}}
\subfigure[MNIST-test]{\includegraphics[width = 0.3\linewidth]{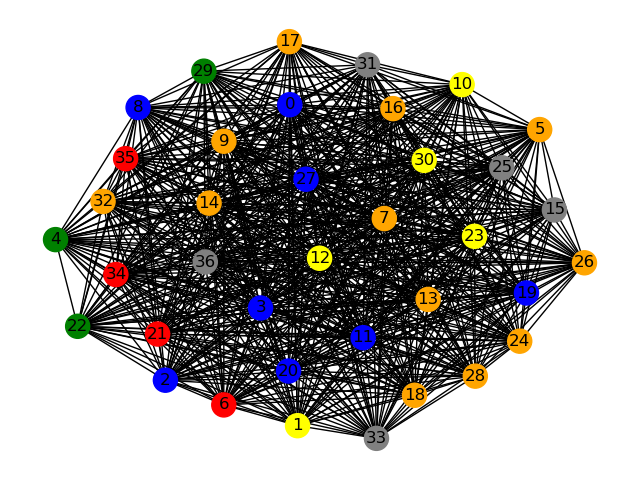}}
\subfigure[CMU-PIE]{\includegraphics[width = 0.3\linewidth]{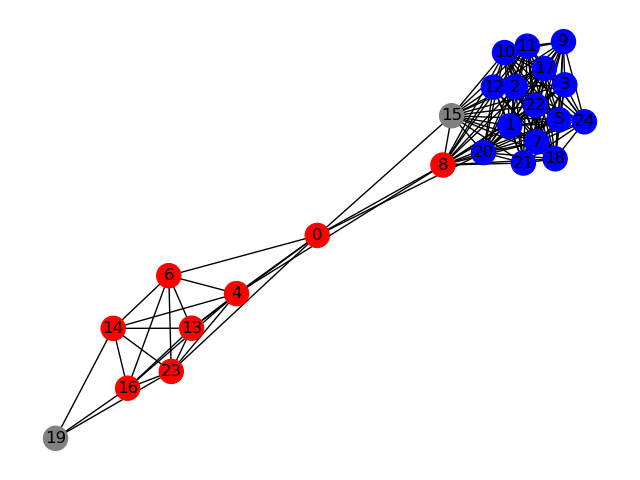}}
\caption{Graph depicting rank correlation based on Calinski-Harabasz index among embedding spaces for Application 1 with JULE. Each node represents an embedding space, and each edge signifies a significant rank correlation.}
\label{fig:graph:ch:0}
\end{figure}
%%%%%%%%%%%%%%%%%%%%%%%%%
\begin{figure}[htbp!]
\centering
\subfigure[Selected space (USPS)]{\includegraphics[width = 0.24\linewidth]{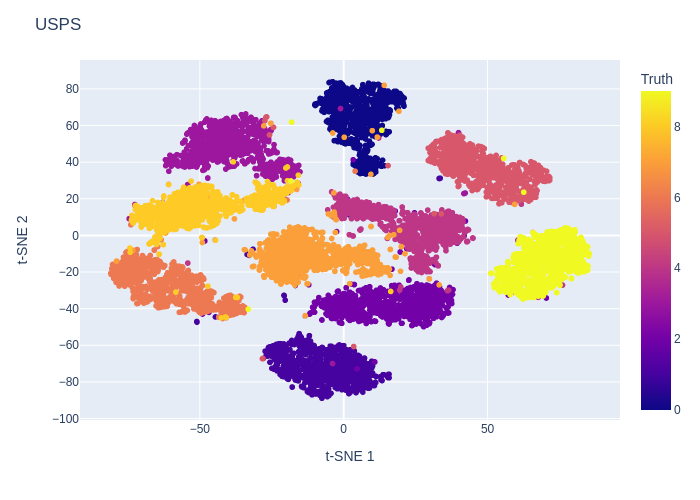}}
\subfigure[Excluded space (USPS)]{\includegraphics[width = 0.24\linewidth]{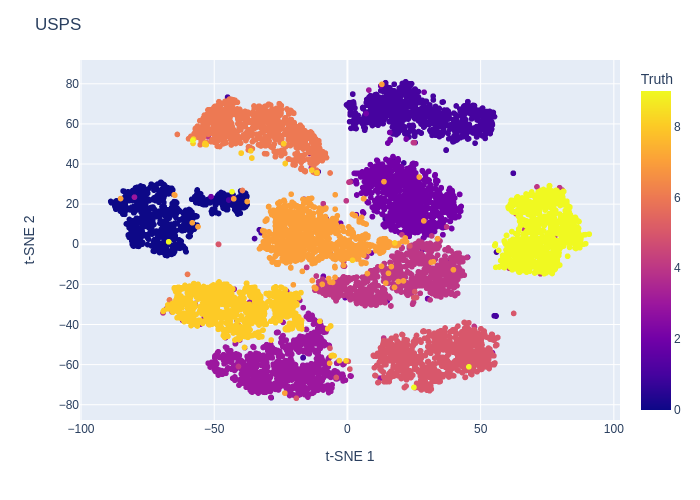}}
\subfigure[Selected space (UMist)]{\includegraphics[width = 0.24\linewidth]{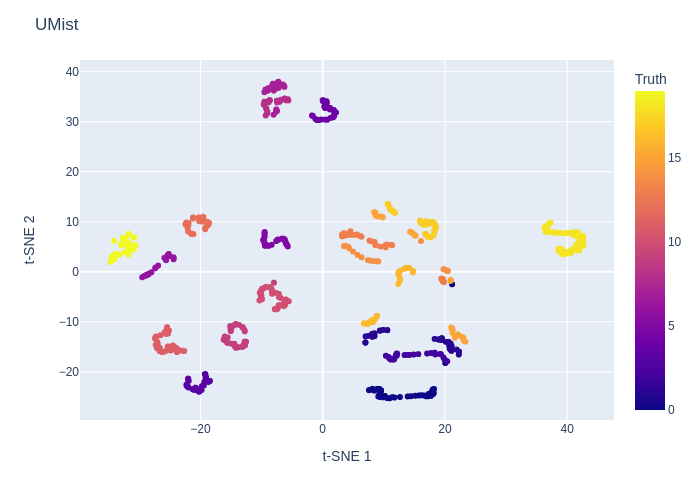}}
\subfigure[Excluded space (UMist)]{\includegraphics[width = 0.24\linewidth]{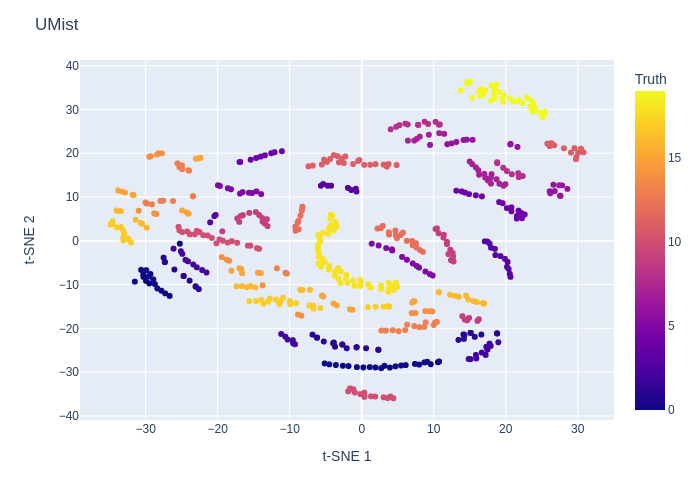}}
\subfigure[Selected space (COIL-20)]{\includegraphics[width = 0.24\linewidth]{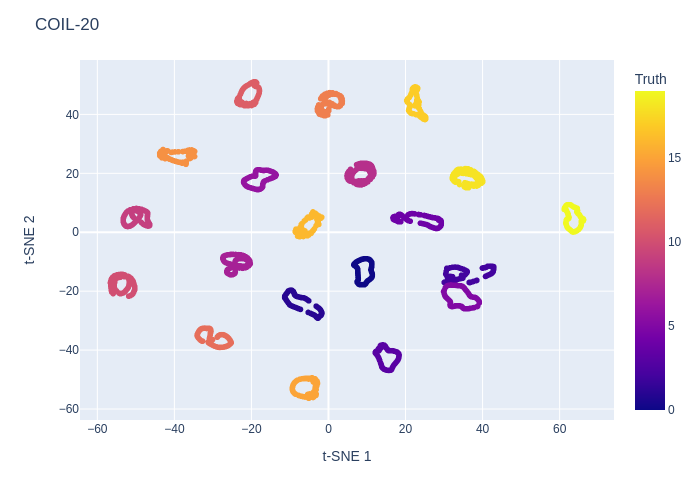}}
\subfigure[Excluded space (COIL-20)]{\includegraphics[width = 0.24\linewidth]{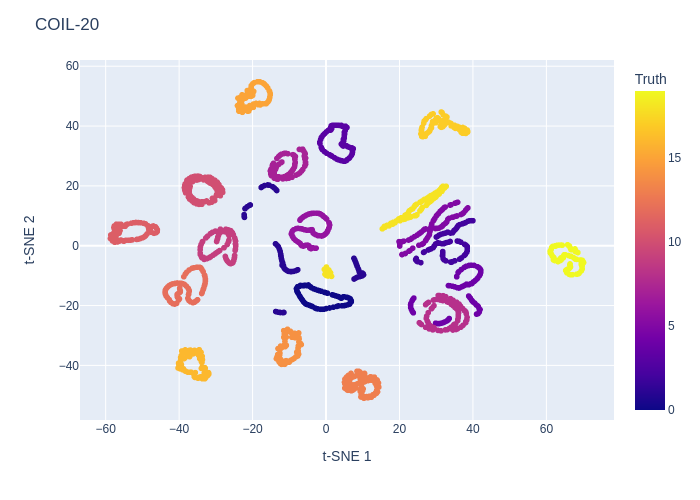}}
\subfigure[Selected space (COIL-100)]{\includegraphics[width = 0.24\linewidth]{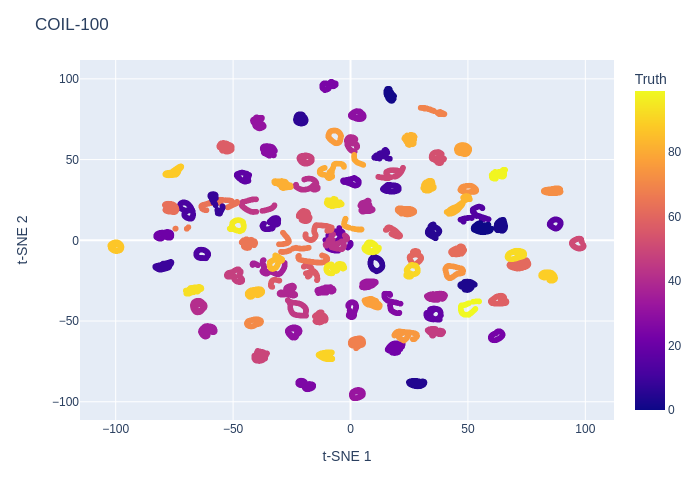}}
\subfigure[Excluded space (COIL-100)]{\includegraphics[width = 0.24\linewidth]{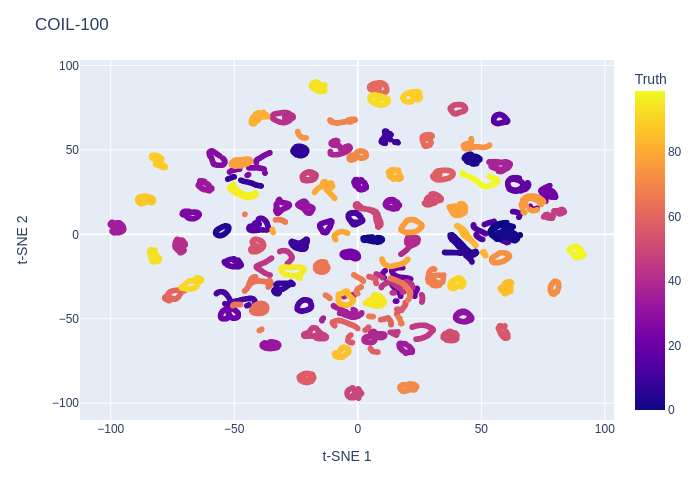}}
\subfigure[Selected space (YTF)]{\includegraphics[width = 0.24\linewidth]{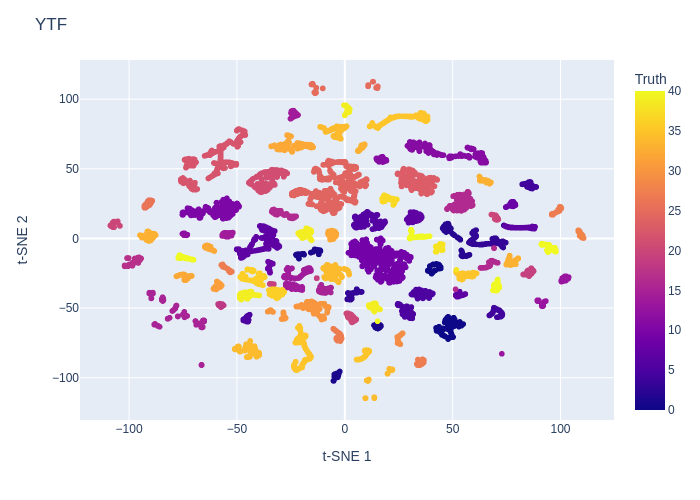}}
\subfigure[Excluded space (YTF)]{\includegraphics[width = 0.24\linewidth]{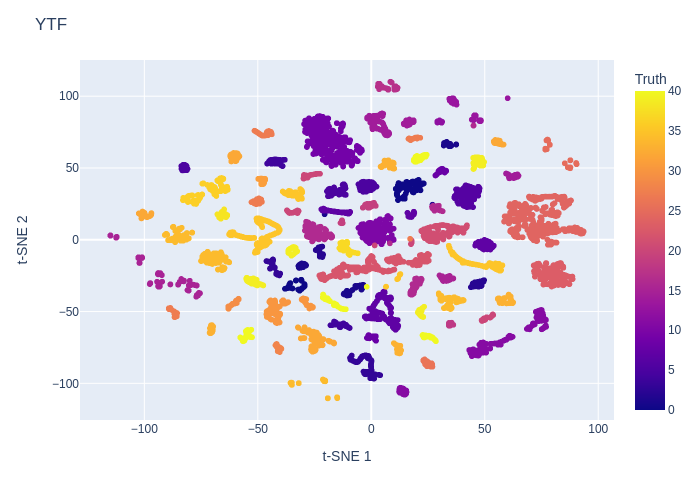}}
\subfigure[Selected space (FRGC)]{\includegraphics[width = 0.24\linewidth]{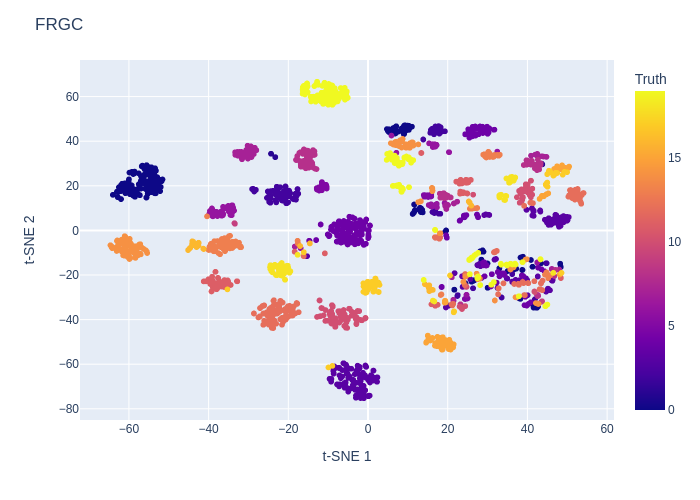}}
\subfigure[Excluded space (FRGC)]{\includegraphics[width = 0.24\linewidth]{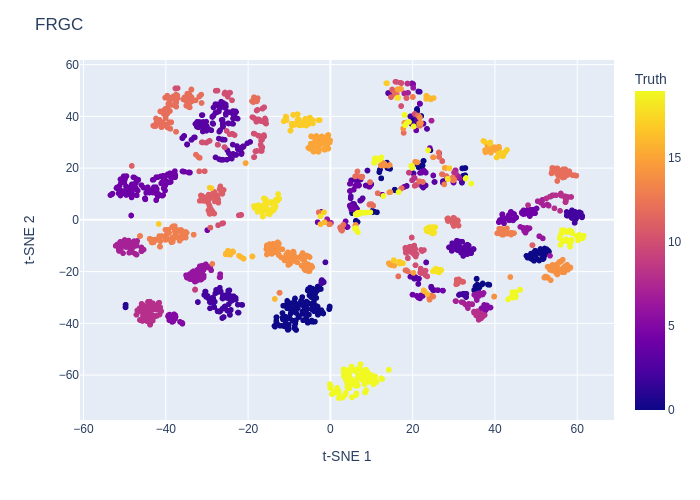}}
\subfigure[Selected space (MNIST-test)]{\includegraphics[width = 0.24\linewidth]{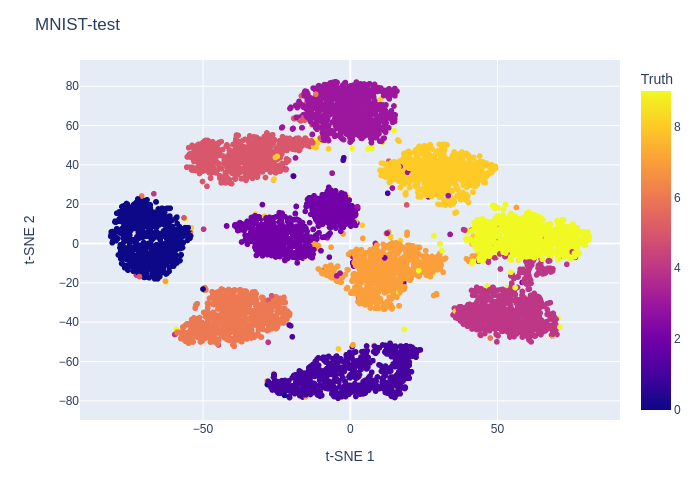}}
\subfigure[Excluded space (MNIST-test)]{\includegraphics[width = 0.24\linewidth]{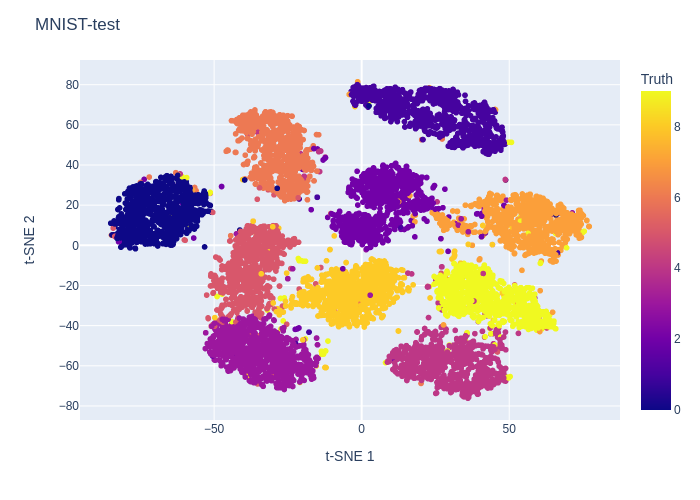}}
\subfigure[Selected space (CMU-PIE)]{\includegraphics[width = 0.24\linewidth]{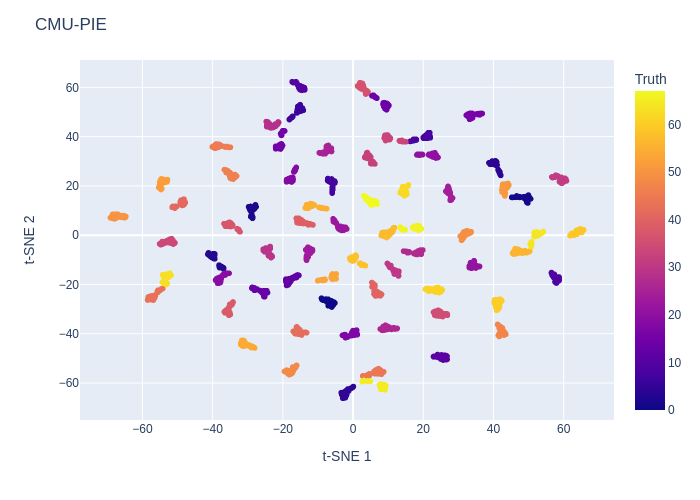}}
\subfigure[Excluded space (CMU-PIE)]{\includegraphics[width = 0.24\linewidth]{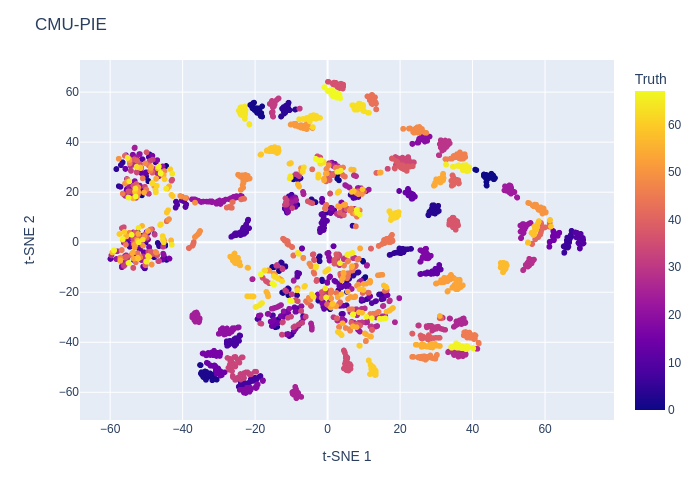}}
\caption{t-SNE visualization illustrating the selected embedding spaces from ACE in comparison to those excluded from ACE, based on Calinski-Harabasz index, for Application 1 with JULE. Each data point in the visualizations is assigned a color corresponding to its true cluster label.}
\label{fig:tsne:ch:0}
\end{figure}
%%%%%%%%%%%%%%%%%%%%%%%%%
\begin{figure}[htbp!]
\centering
\subfigure[USPS]{\includegraphics[width = 0.3\linewidth]{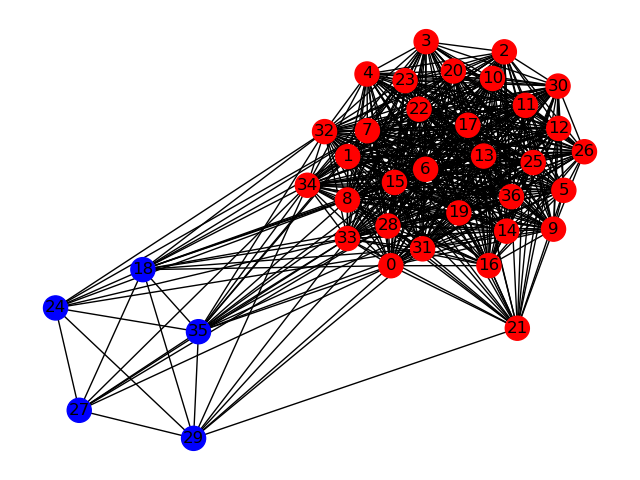}}
\subfigure[UMist]{\includegraphics[width = 0.3\linewidth]{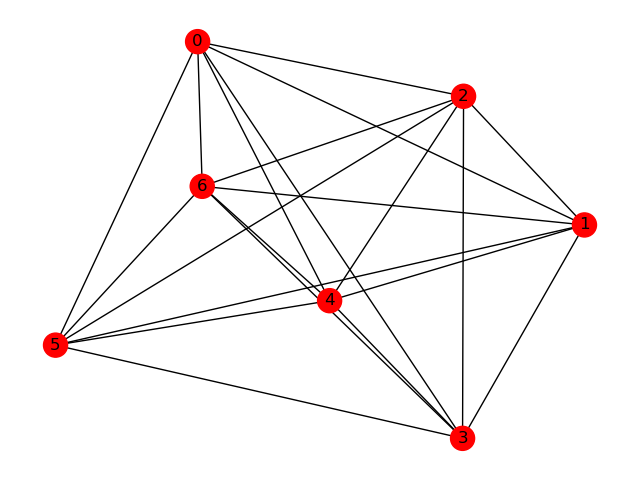}}
\subfigure[COIL-20]{\includegraphics[width = 0.3\linewidth]{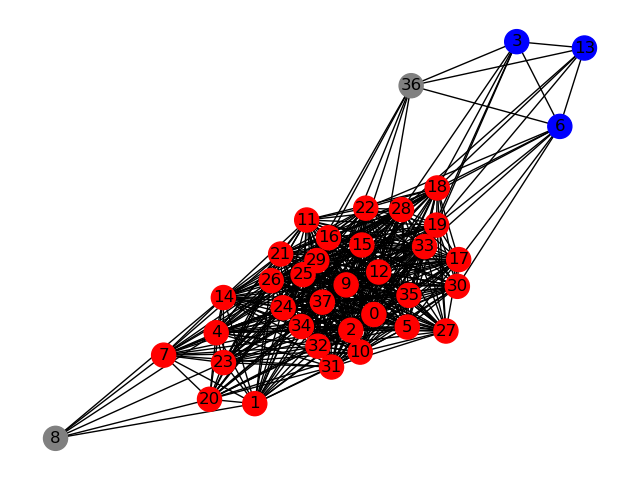}}
\subfigure[COIL-100]{\includegraphics[width = 0.3\linewidth]{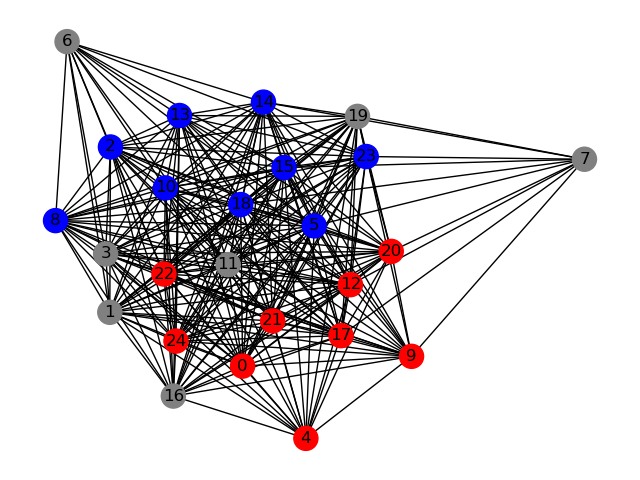}}
\subfigure[YTF]{\includegraphics[width = 0.3\linewidth]{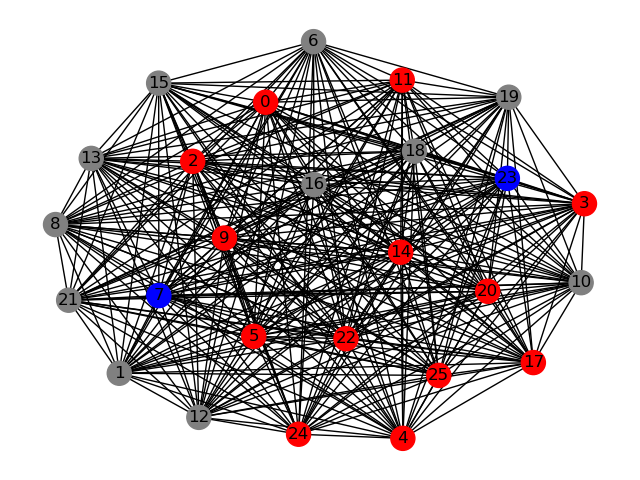}}
\subfigure[FRGC]{\includegraphics[width = 0.3\linewidth]{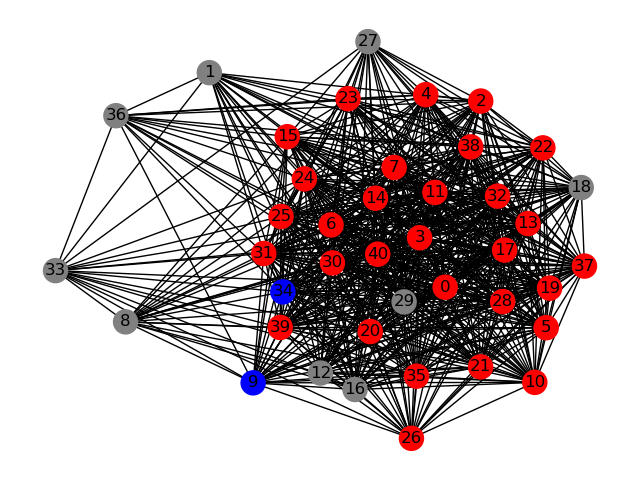}}
\subfigure[MNIST-test]{\includegraphics[width = 0.3\linewidth]{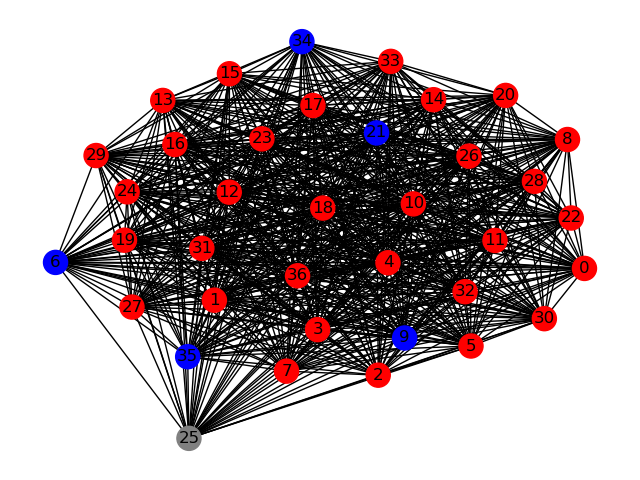}}
\subfigure[CMU-PIE]{\includegraphics[width = 0.3\linewidth]{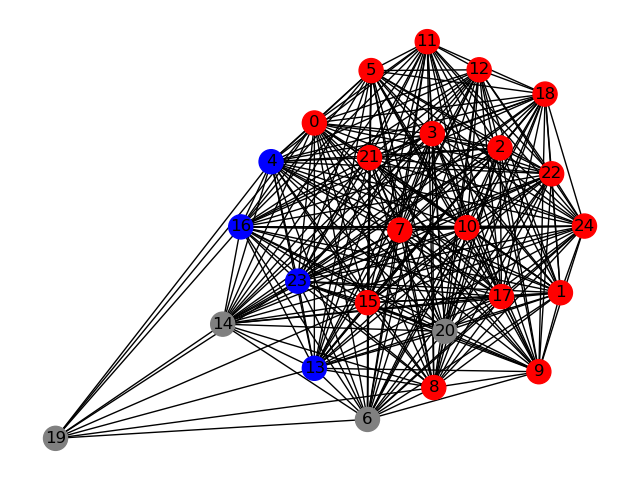}}
\caption{Graph depicting rank correlation based on Silhouette score (Cosine distance) among embedding spaces for Application 1 with JULE. Each node represents an embedding space, and each edge signifies a significant rank correlation.}
\label{fig:graph:cosine:0}
\end{figure}
%%%%%%%%%%%%%%%%%%%%%%%%%
\begin{figure}[htbp!]
\centering
\subfigure[Selected space (USPS)]{\includegraphics[width = 0.24\linewidth]{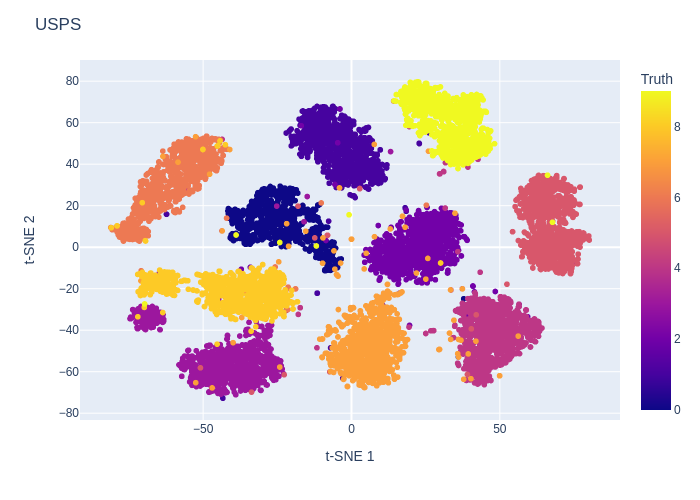}}
\subfigure[Excluded space (USPS)]{\includegraphics[width = 0.24\linewidth]{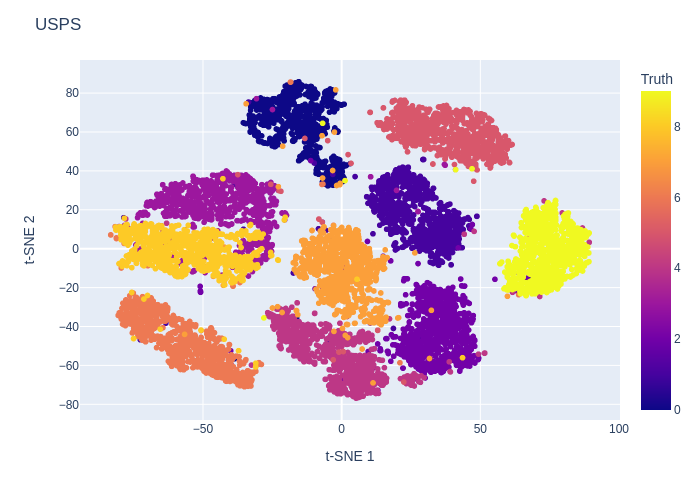}}
\subfigure[Selected space (UMist)]{\includegraphics[width = 0.24\linewidth]{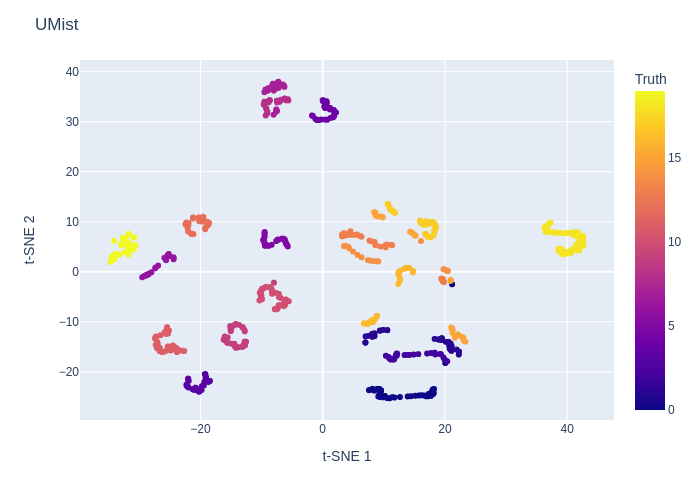}}
\subfigure[Excluded space (UMist)]{\includegraphics[width = 0.24\linewidth]{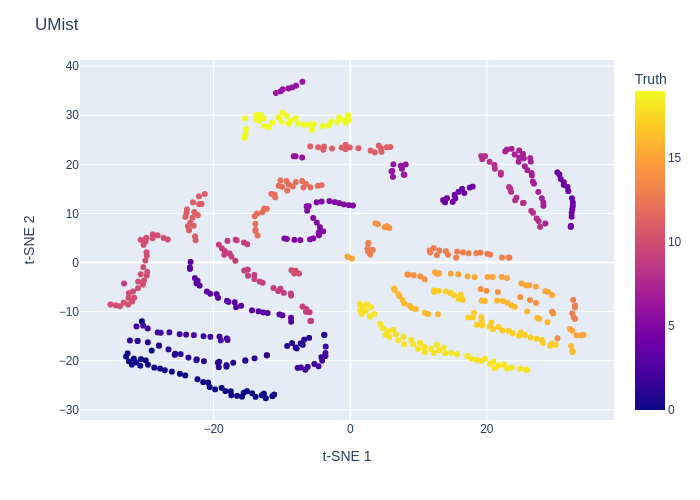}}
\subfigure[Selected space (COIL-20)]{\includegraphics[width = 0.24\linewidth]{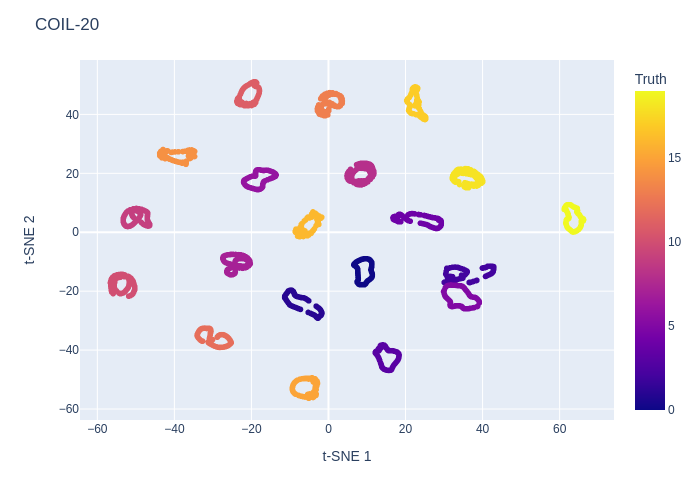}}
\subfigure[Excluded space (COIL-20)]{\includegraphics[width = 0.24\linewidth]{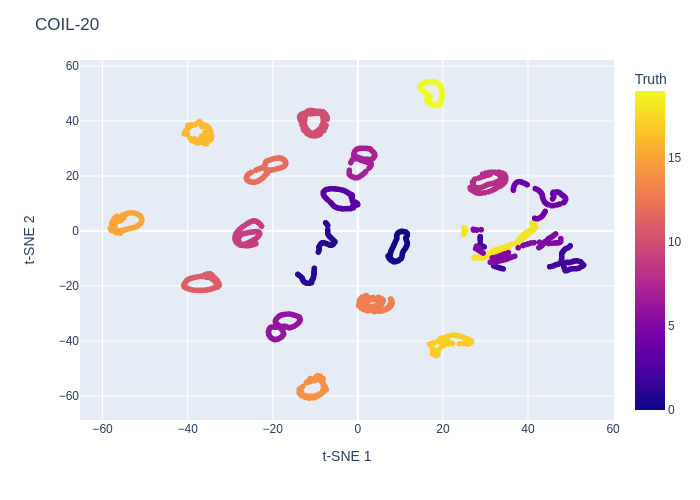}}
\subfigure[Selected space (COIL-100)]{\includegraphics[width = 0.24\linewidth]{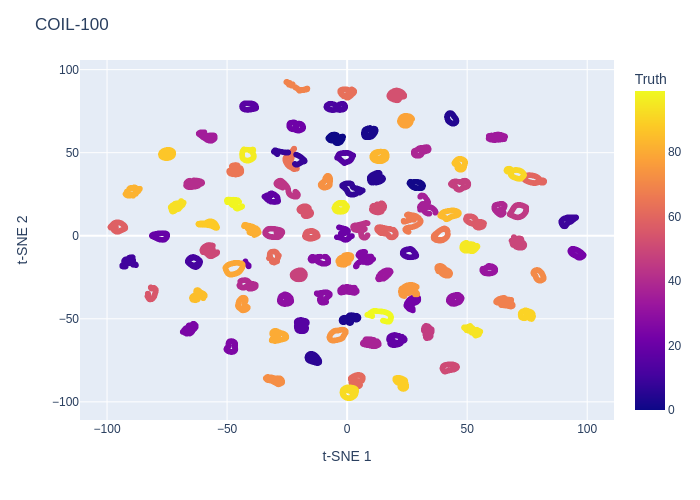}}
\subfigure[Excluded space (COIL-100)]{\includegraphics[width = 0.24\linewidth]{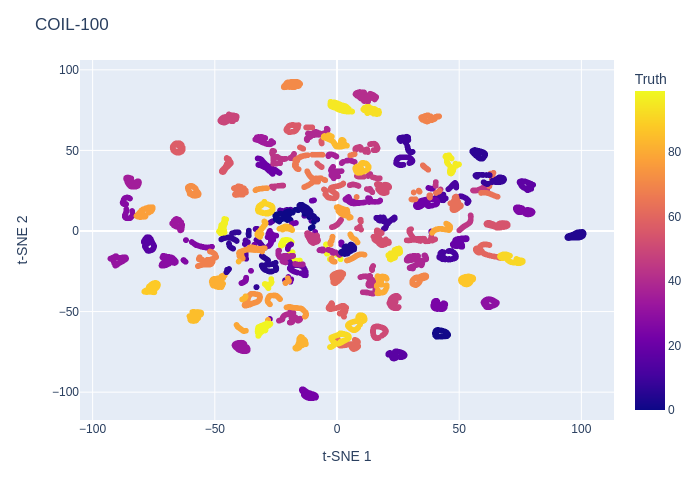}}
\subfigure[Selected space (YTF)]{\includegraphics[width = 0.24\linewidth]{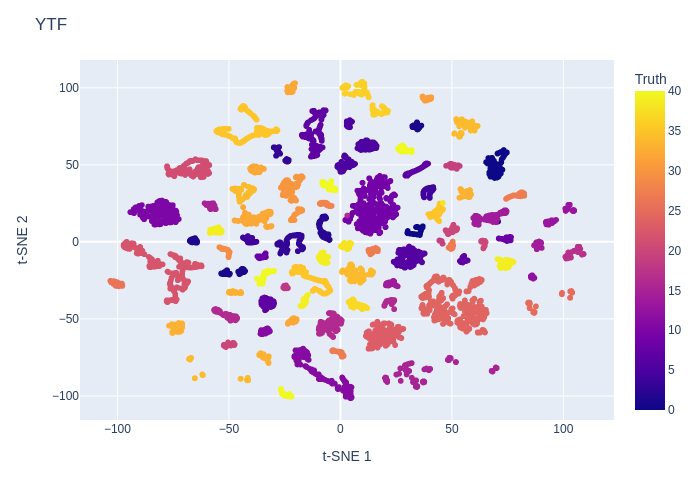}}
\subfigure[Excluded space (YTF)]{\includegraphics[width = 0.24\linewidth]{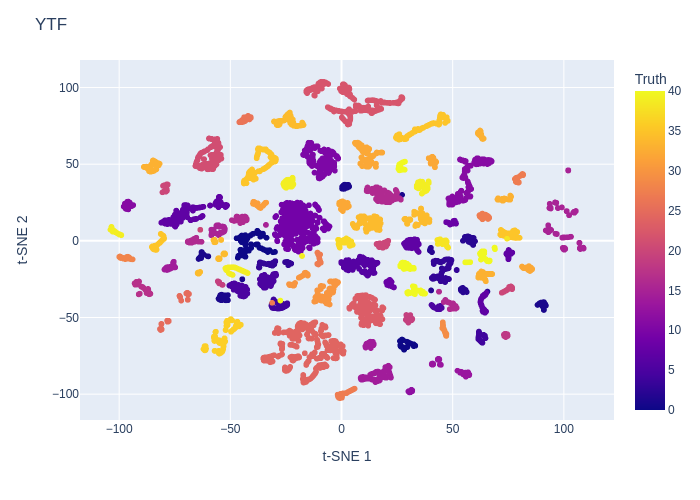}}
\subfigure[Selected space (FRGC)]{\includegraphics[width = 0.24\linewidth]{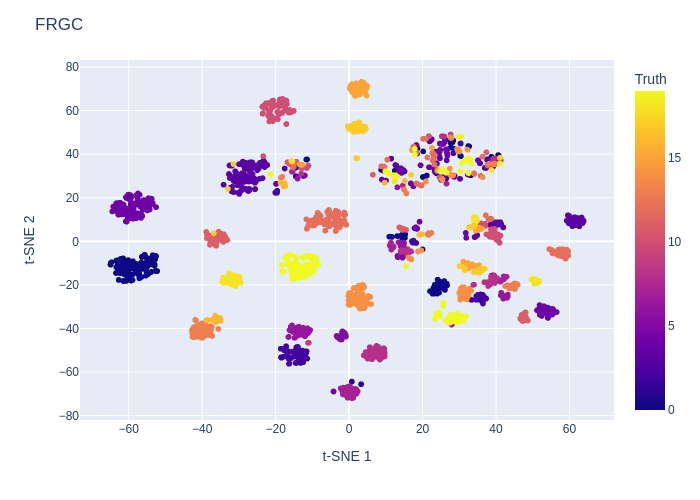}}
\subfigure[Excluded space (FRGC)]{\includegraphics[width = 0.24\linewidth]{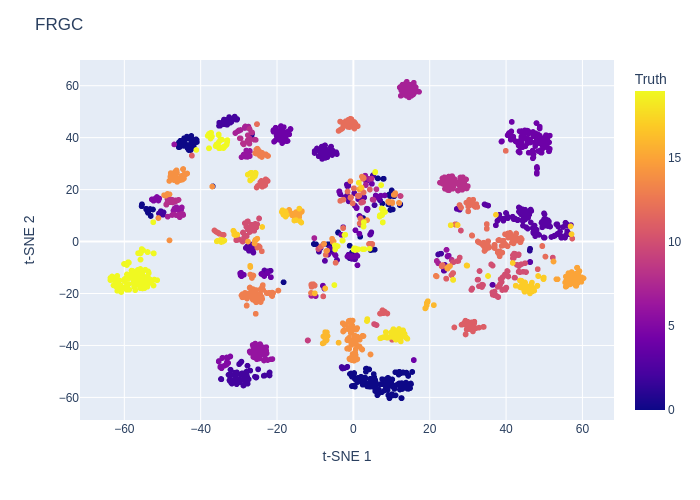}}
\subfigure[Selected space (MNIST-test)]{\includegraphics[width = 0.24\linewidth]{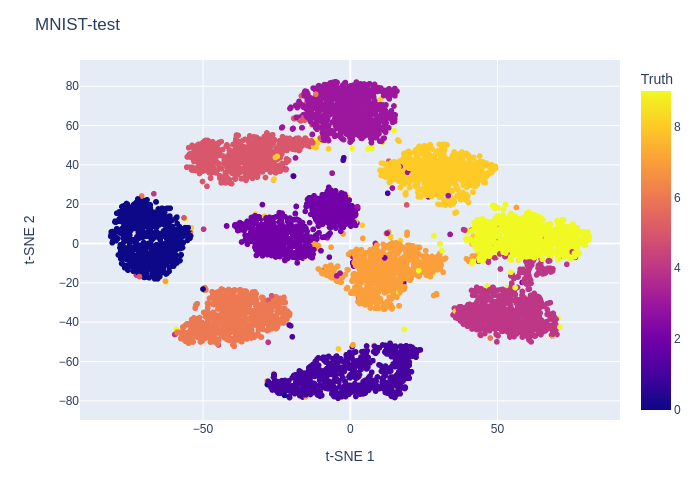}}
\subfigure[Excluded space (MNIST-test)]{\includegraphics[width = 0.24\linewidth]{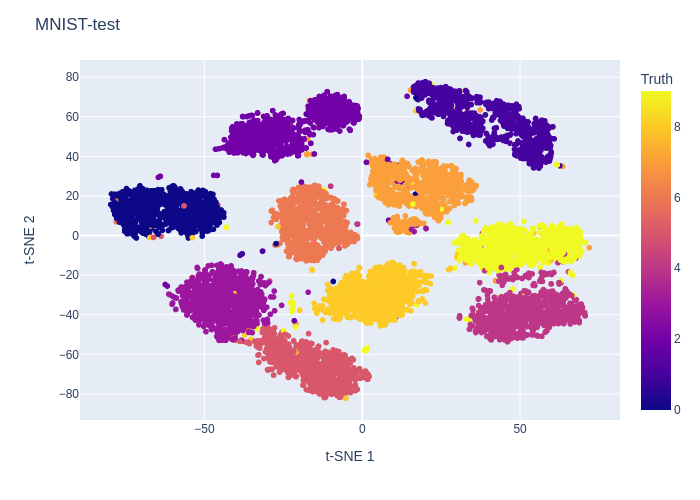}}
\subfigure[Selected space (CMU-PIE)]{\includegraphics[width = 0.24\linewidth]{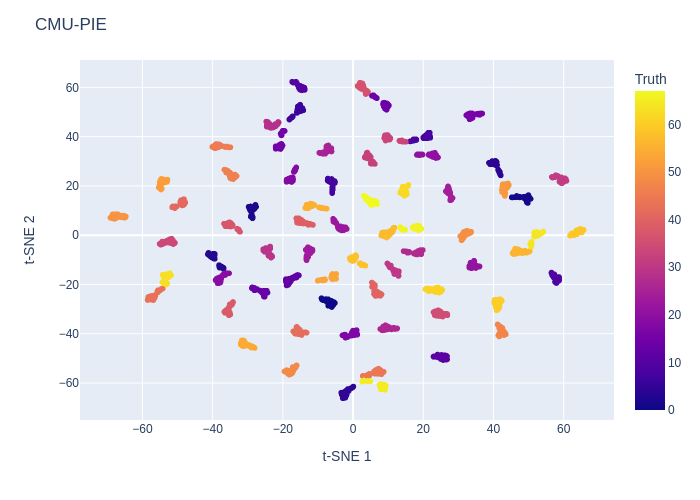}}
\subfigure[Excluded space (CMU-PIE)]{\includegraphics[width = 0.24\linewidth]{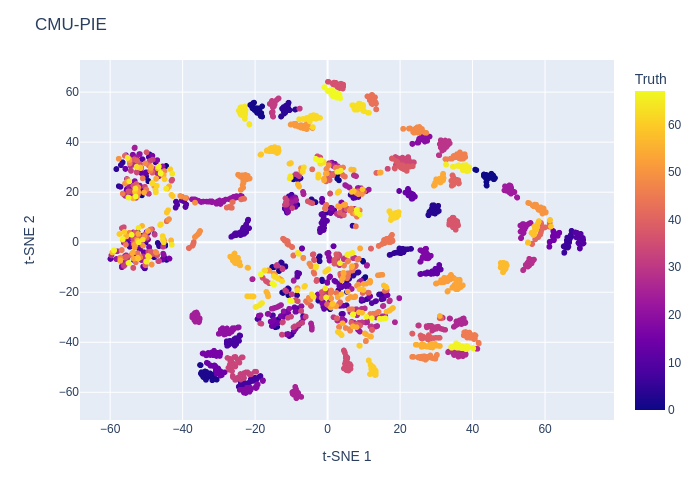}}
\caption{t-SNE visualization illustrating the selected embedding spaces from ACE in comparison to those excluded from ACE, based on Silhouette score (Cosine distance), for Application 1 with JULE. Each data point in the visualizations is assigned a color corresponding to its true cluster label.}
\label{fig:tsne:cosine:0}
\end{figure}
%%%%%%%%%%%%%%%%%%%%%%%%%
\begin{figure}[htbp!]
\centering
\subfigure[USPS]{\includegraphics[width = 0.3\linewidth]{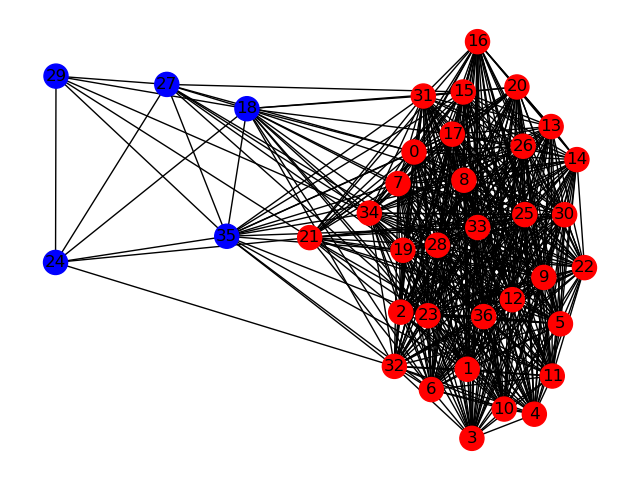}}
\subfigure[UMist]{\includegraphics[width = 0.3\linewidth]{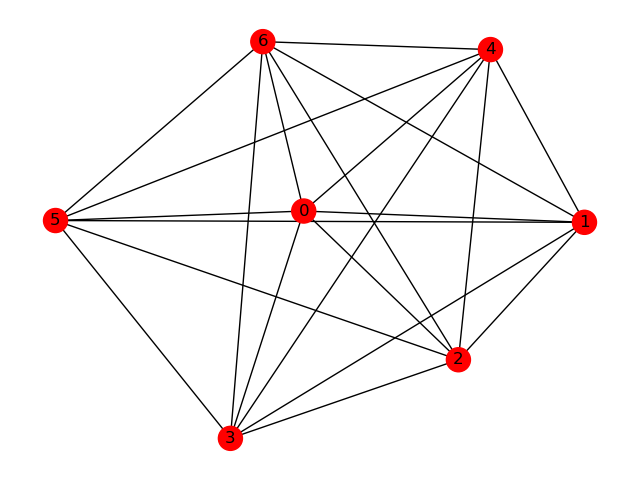}}
\subfigure[COIL-20]{\includegraphics[width = 0.3\linewidth]{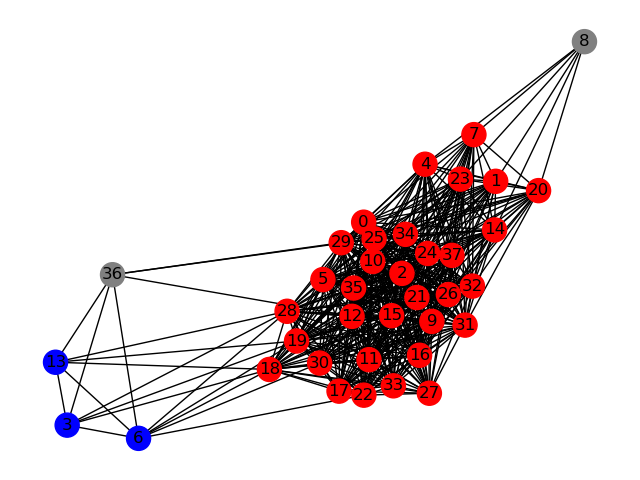}}
\subfigure[COIL-100]{\includegraphics[width = 0.3\linewidth]{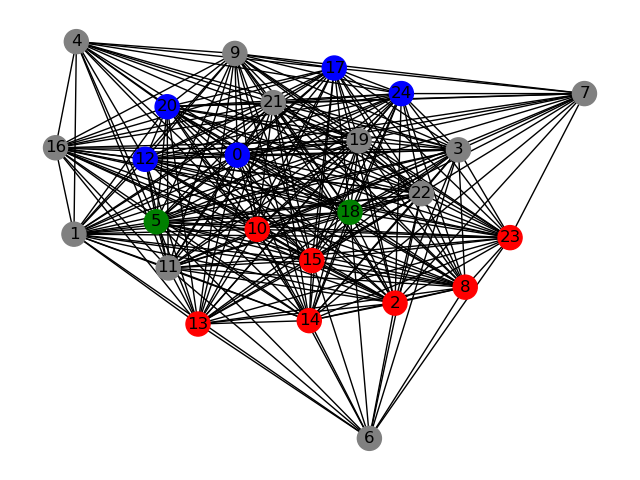}}
\subfigure[YTF]{\includegraphics[width = 0.3\linewidth]{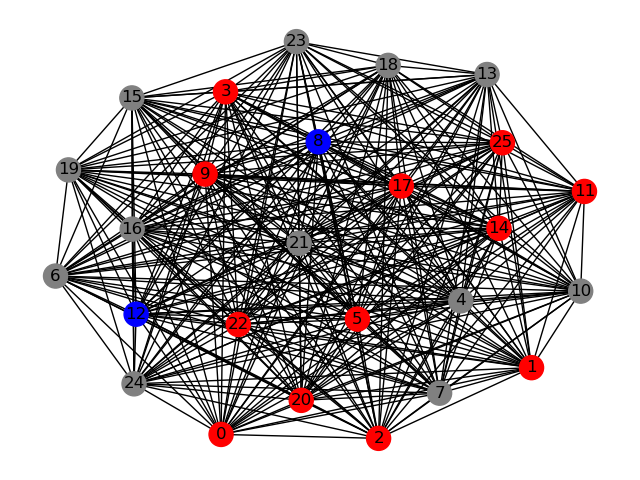}}
\subfigure[FRGC]{\includegraphics[width = 0.3\linewidth]{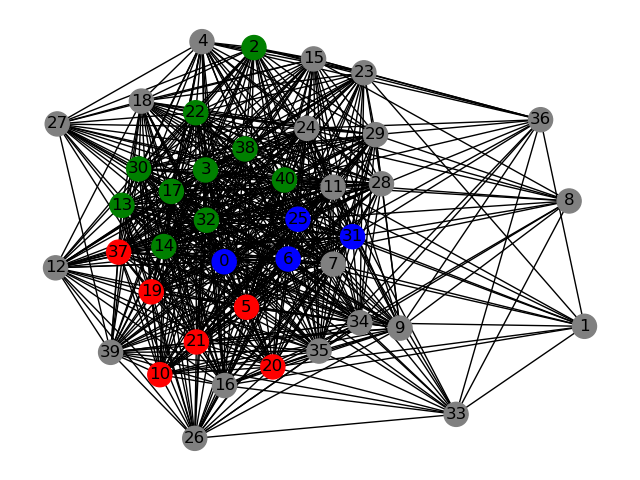}}
\subfigure[MNIST-test]{\includegraphics[width = 0.3\linewidth]{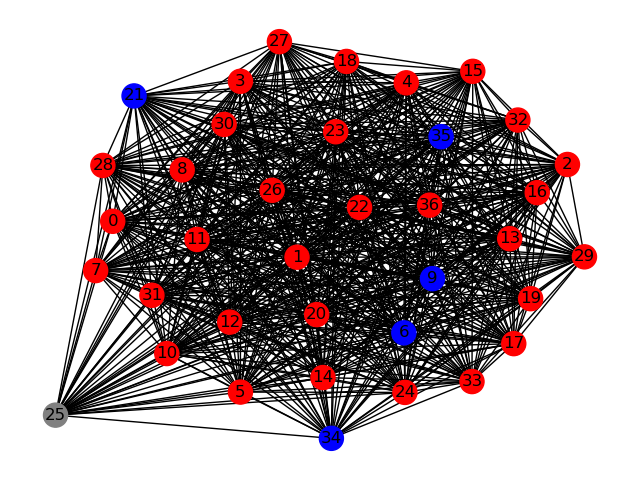}}
\subfigure[CMU-PIE]{\includegraphics[width = 0.3\linewidth]{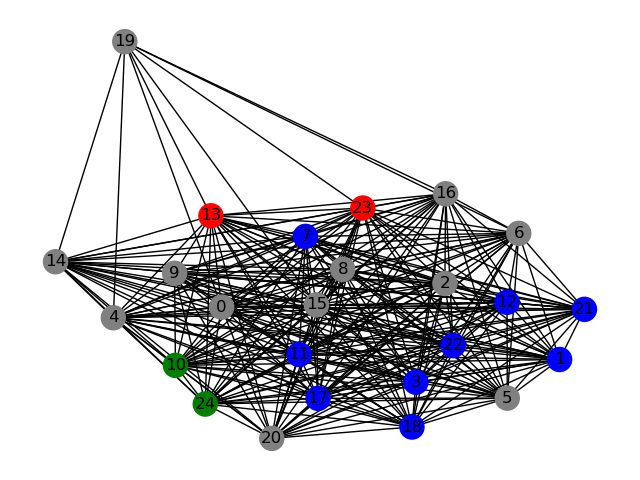}}
\caption{Graph depicting rank correlation based on Silhouette score (Euclidean distance) among embedding spaces for Application 1 with JULE. Each node represents an embedding space, and each edge signifies a significant rank correlation.}
\label{fig:graph:euclidean:0}
\end{figure}
%%%%%%%%%%%%%%%%%%%%%%%%%
\begin{figure}[htbp!]
\centering
\subfigure[Selected space (USPS)]{\includegraphics[width = 0.24\linewidth]{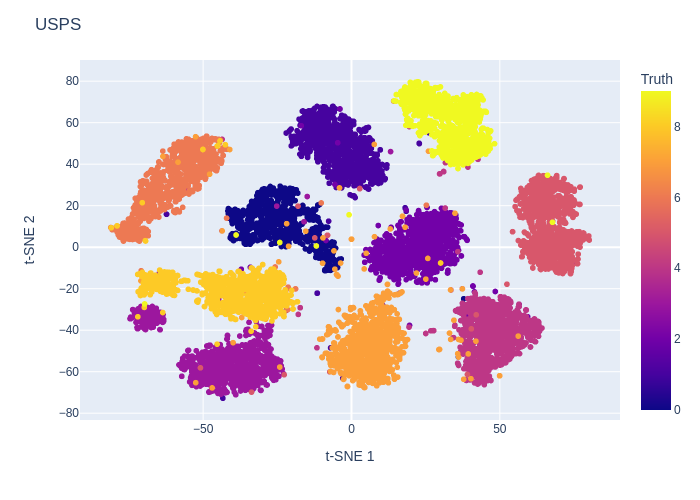}}
\subfigure[Excluded space (USPS)]{\includegraphics[width = 0.24\linewidth]{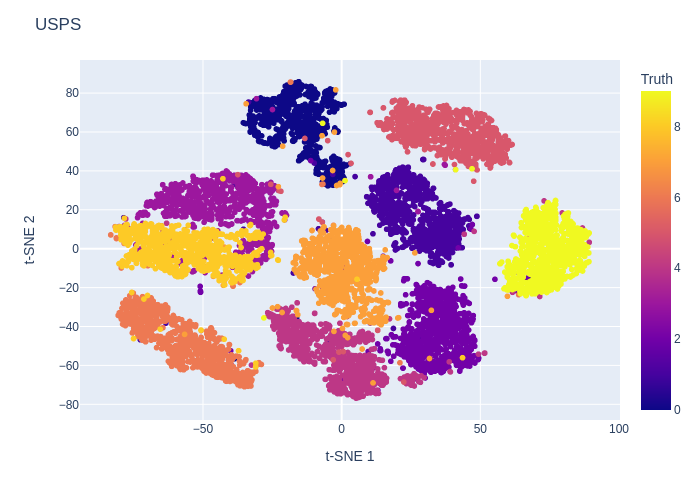}}
\subfigure[Selected space (UMist)]{\includegraphics[width = 0.24\linewidth]{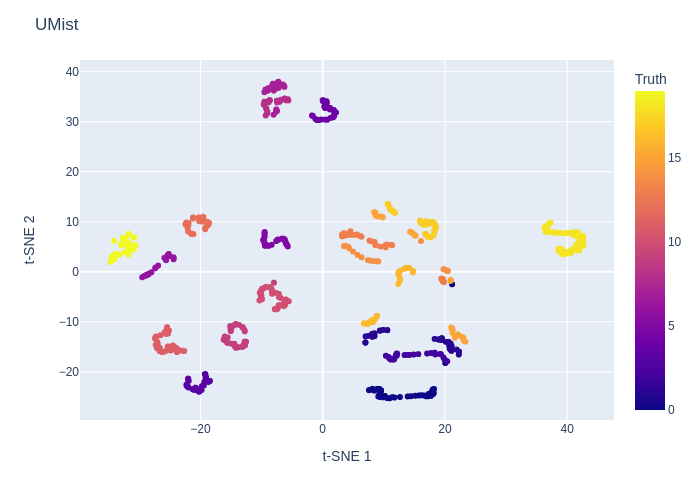}}
\subfigure[Excluded space (UMist)]{\includegraphics[width = 0.24\linewidth]{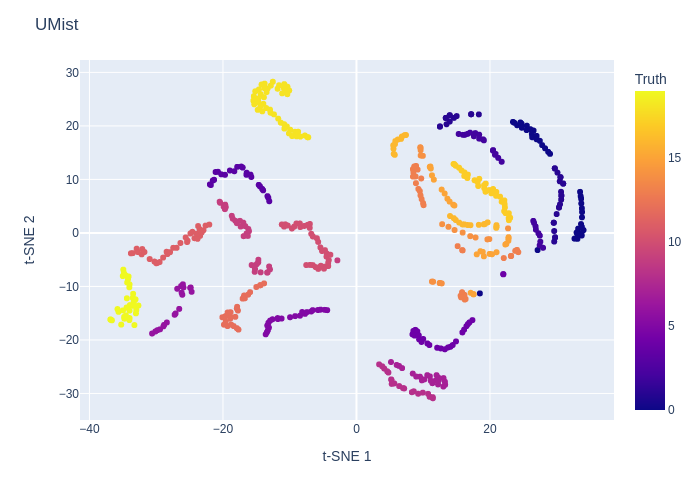}}
\subfigure[Selected space (COIL-20)]{\includegraphics[width = 0.24\linewidth]{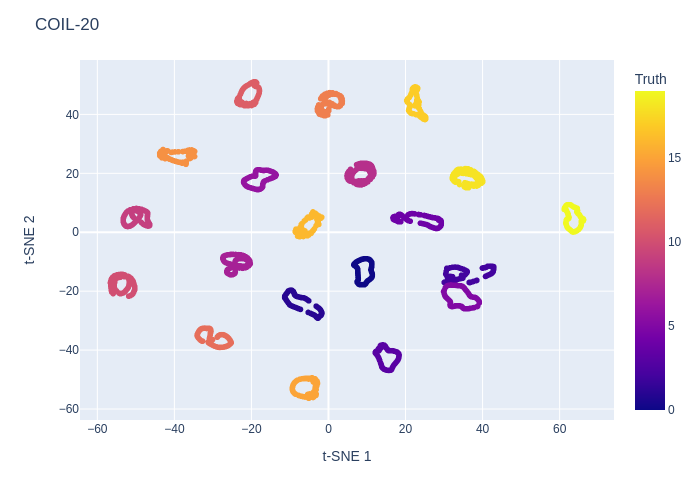}}
\subfigure[Excluded space (COIL-20)]{\includegraphics[width = 0.24\linewidth]{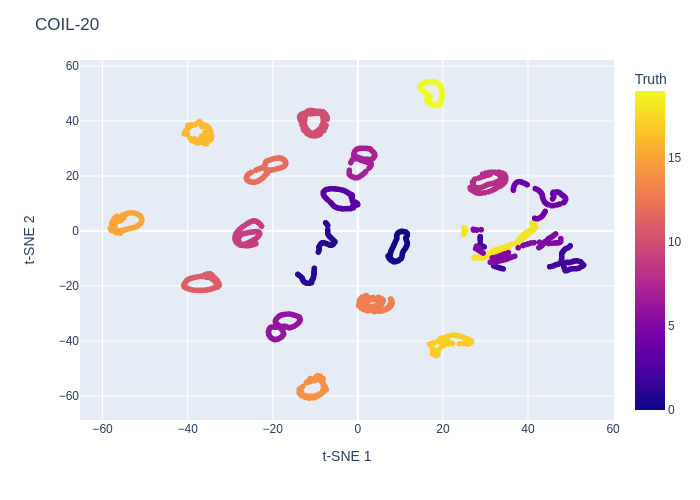}}
\subfigure[Selected space (COIL-100)]{\includegraphics[width = 0.24\linewidth]{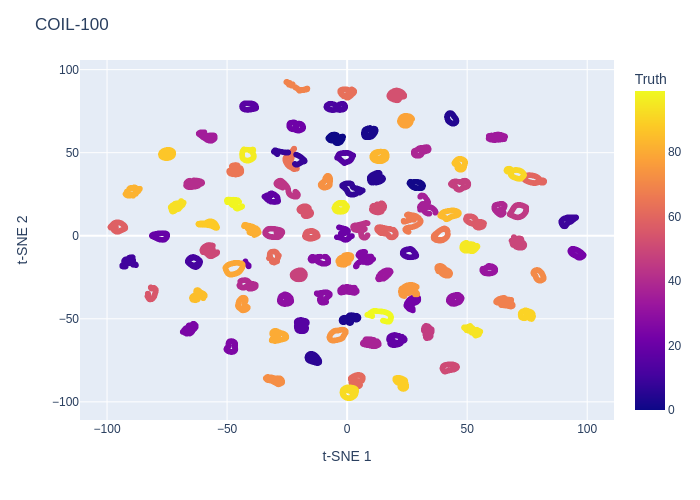}}
\subfigure[Excluded space (COIL-100)]{\includegraphics[width = 0.24\linewidth]{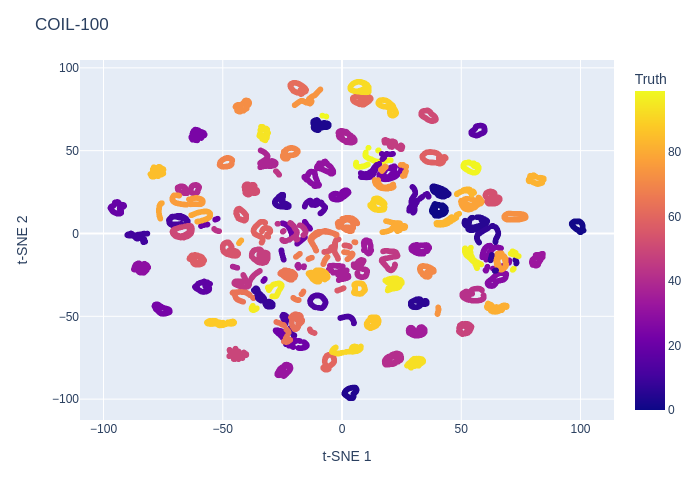}}
\subfigure[Selected space (YTF)]{\includegraphics[width = 0.24\linewidth]{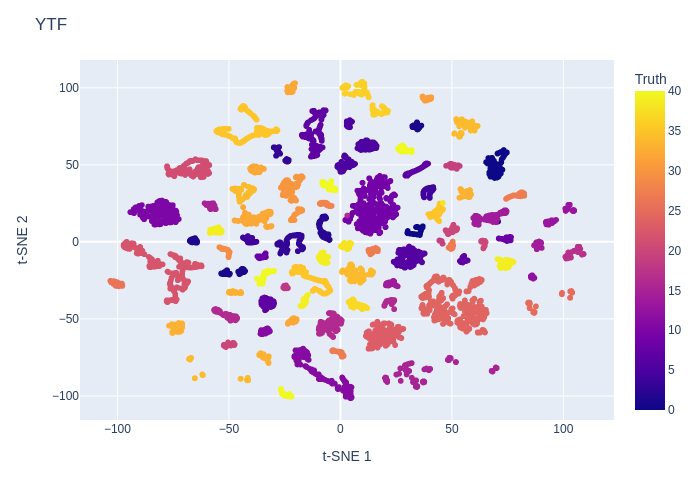}}
\subfigure[Excluded space (YTF)]{\includegraphics[width = 0.24\linewidth]{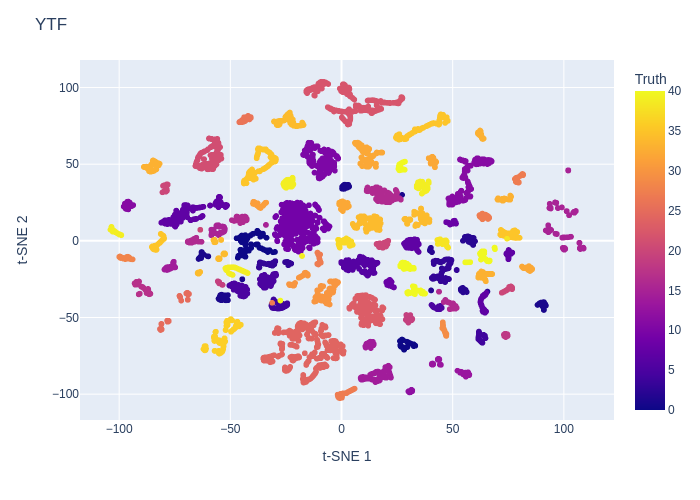}}
\subfigure[Selected space (FRGC)]{\includegraphics[width = 0.24\linewidth]{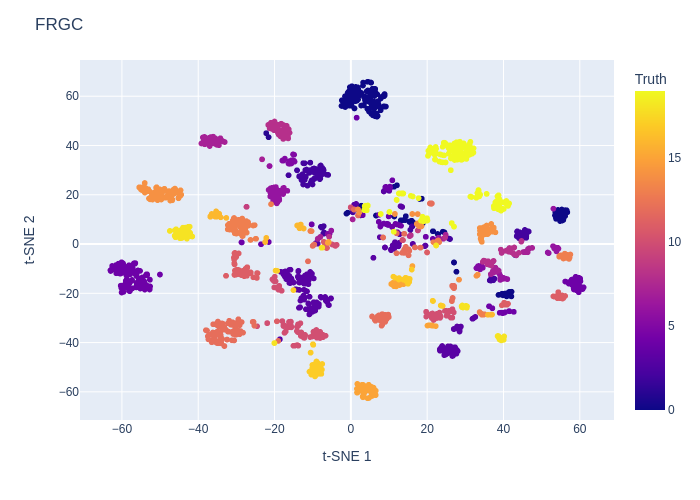}}
\subfigure[Excluded space (FRGC)]{\includegraphics[width = 0.24\linewidth]{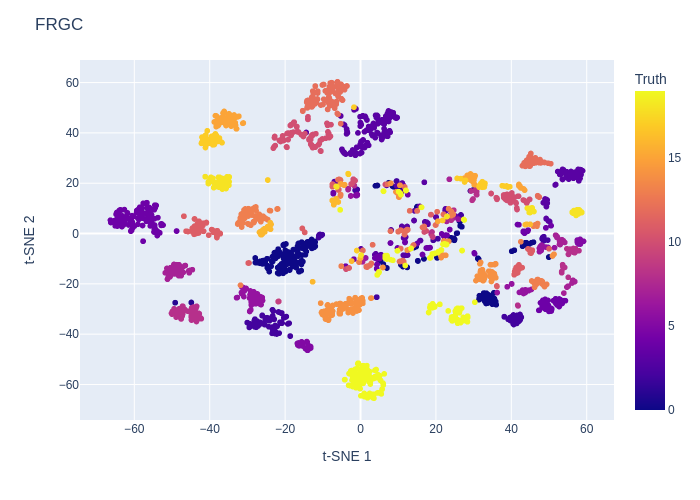}}
\subfigure[Selected space (MNIST-test)]{\includegraphics[width = 0.24\linewidth]{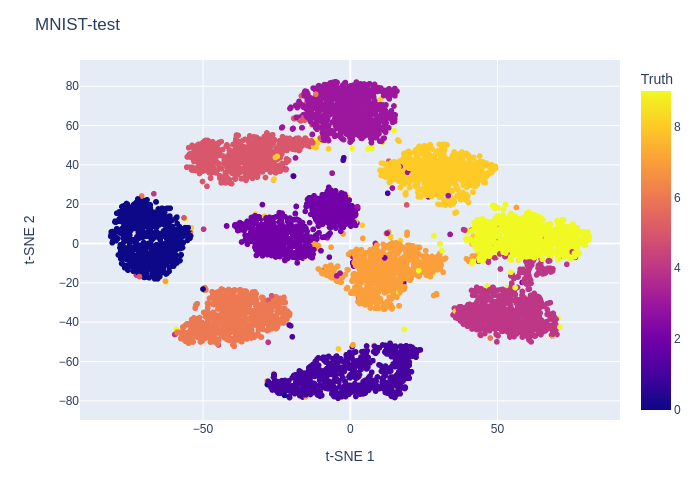}}
\subfigure[Excluded space (MNIST-test)]{\includegraphics[width = 0.24\linewidth]{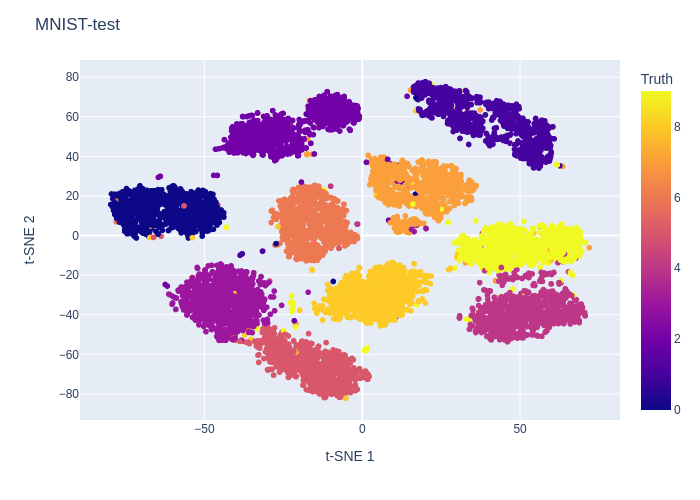}}
\subfigure[Selected space (CMU-PIE)]{\includegraphics[width = 0.24\linewidth]{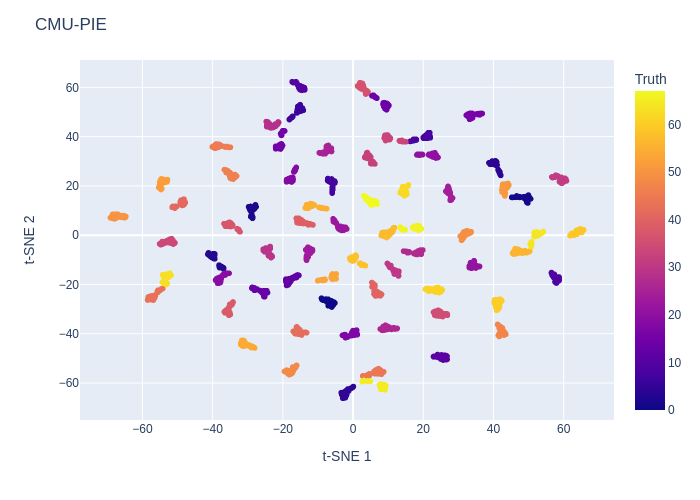}}
\subfigure[Excluded space (CMU-PIE)]{\includegraphics[width = 0.24\linewidth]{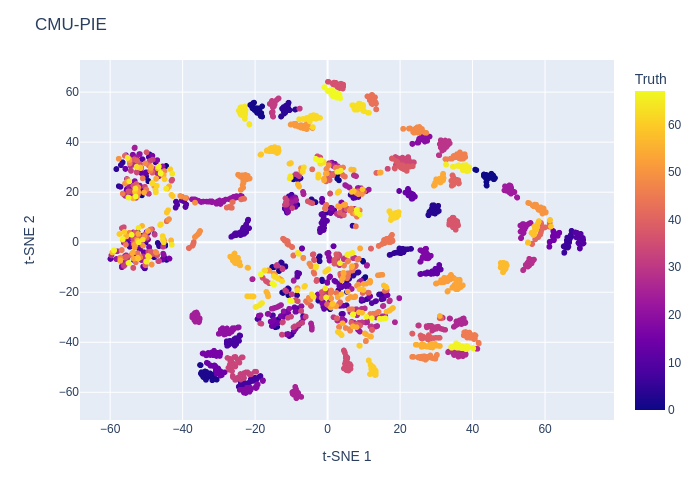}}
\caption{t-SNE visualization illustrating the selected embedding spaces from ACE in comparison to those excluded from ACE, based on Silhouette score (Euclidean distance), for Application 1 with JULE. Each data point in the visualizations is assigned a color corresponding to its true cluster label.}
\label{fig:tsne:euclidean:0}
\end{figure}
%%%%%%%%%%%%%%%%%%%%%%%%%
%%%%%%%%%%%%%%%%%%%%%%%%%
\begin{figure}[htbp!]
\centering
\subfigure[USPS]{\includegraphics[width = 0.3\linewidth]{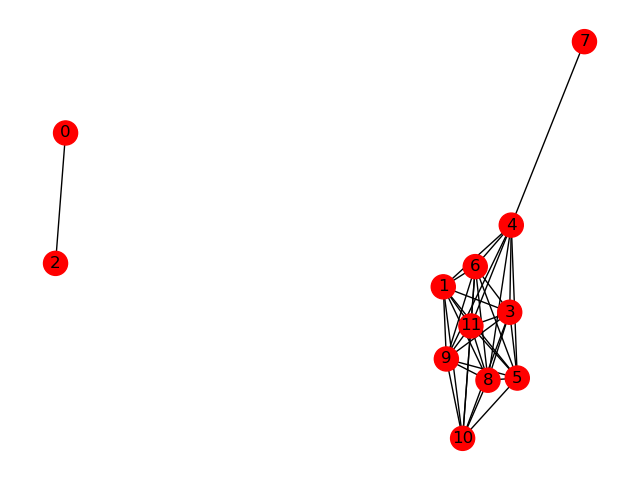}}
\subfigure[YTF]{\includegraphics[width = 0.3\linewidth]{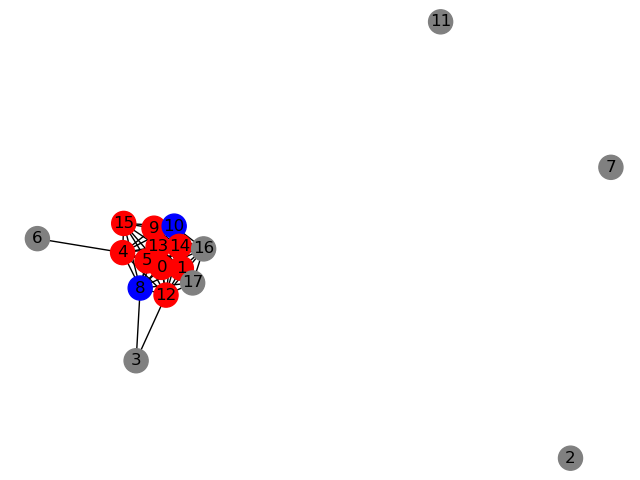}}
\subfigure[FRGC]{\includegraphics[width = 0.3\linewidth]{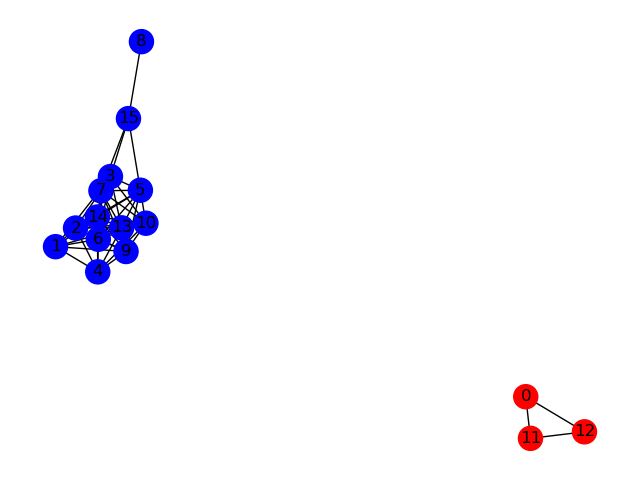}}
\subfigure[MNIST-test]{\includegraphics[width = 0.3\linewidth]{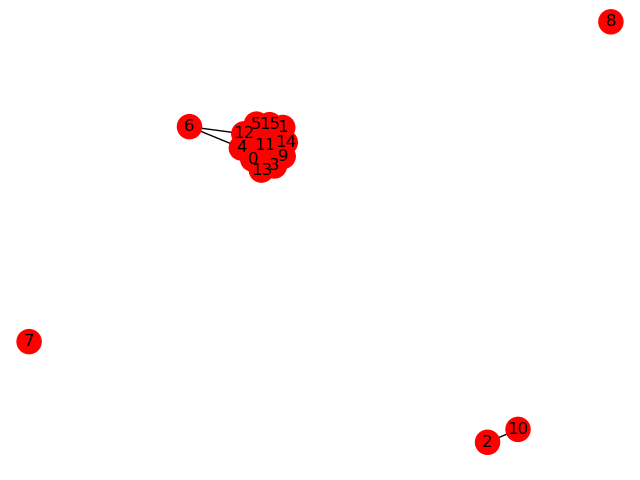}}
\subfigure[CMU-PIE]{\includegraphics[width = 0.3\linewidth]{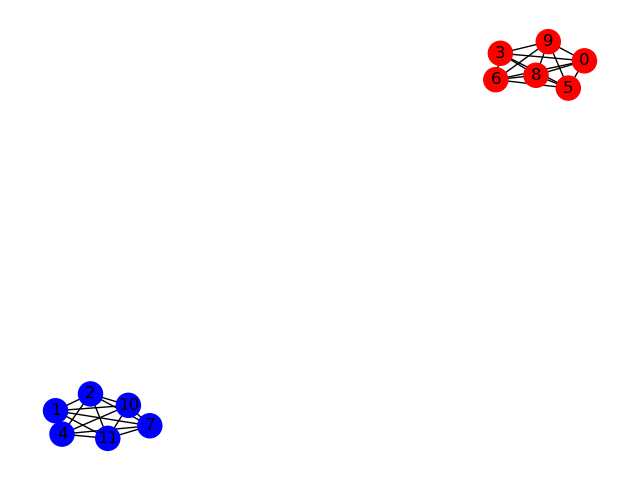}}
\caption{Graph depicting rank correlation based on Davies-Bouldin index among embedding spaces for Application 1 with DEPICT. Each node represents an embedding space, and each edge signifies a significant rank correlation.}
\label{fig:graph:dav:1}
\end{figure}
%%%%%%%%%%%%%%%%%%%%%%%%%
\begin{figure}[htbp!]
\centering
\subfigure[Selected space (USPS)]{\includegraphics[width = 0.24\linewidth]{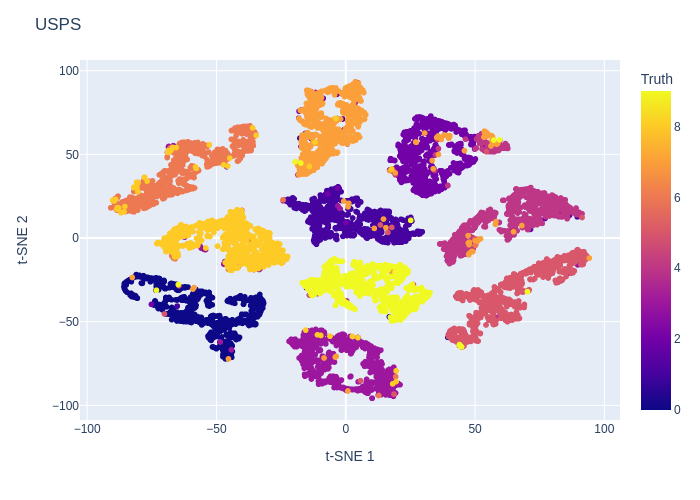}}
\subfigure[Excluded space (USPS)]{\includegraphics[width = 0.24\linewidth]{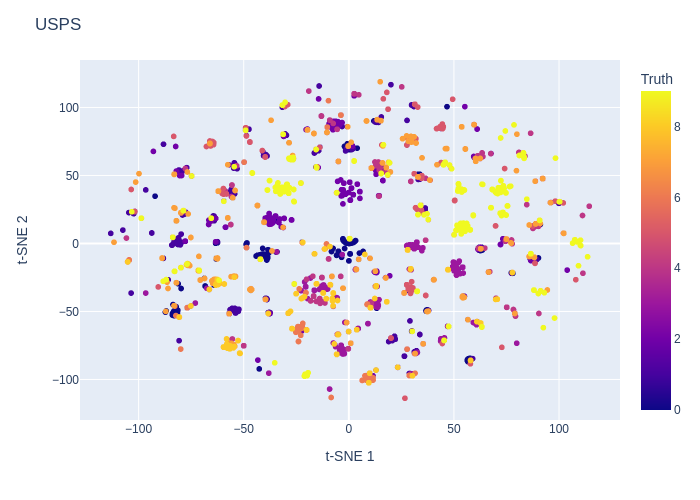}}
\subfigure[Selected space (YTF)]{\includegraphics[width = 0.24\linewidth]{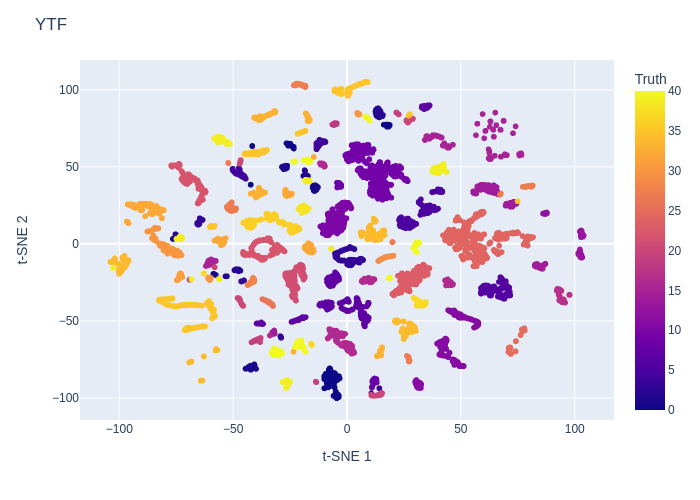}}
\subfigure[Excluded space (YTF)]{\includegraphics[width = 0.24\linewidth]{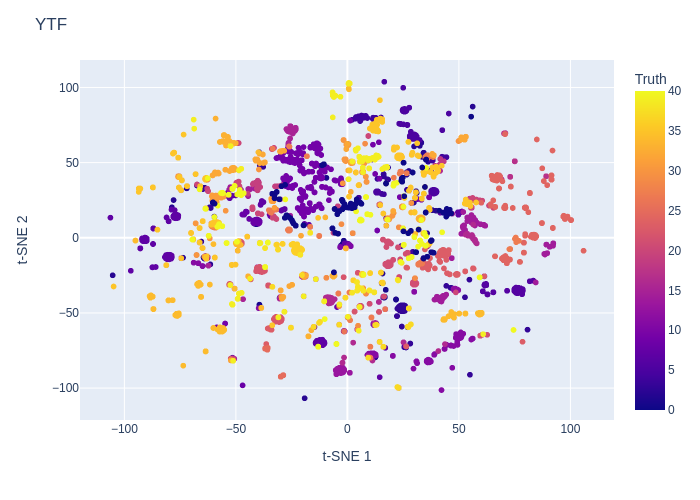}}
\subfigure[Selected space (FRGC)]{\includegraphics[width = 0.24\linewidth]{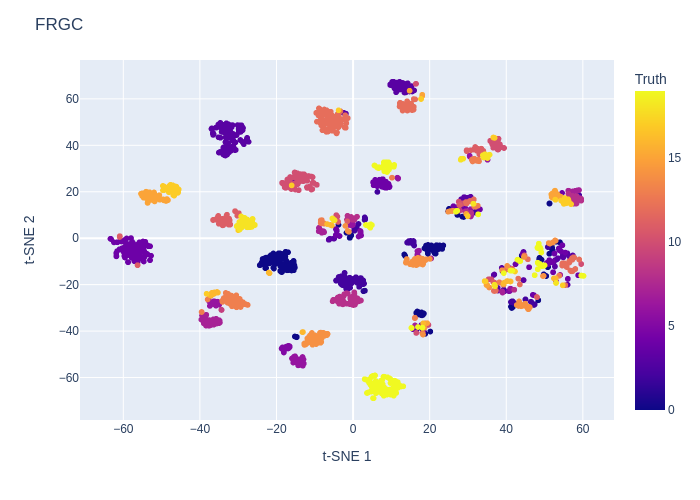}}
\subfigure[Excluded space (FRGC)]{\includegraphics[width = 0.24\linewidth]{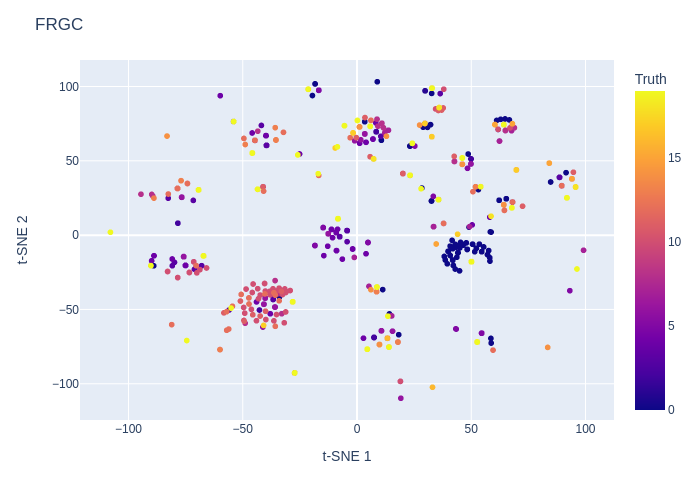}}
\subfigure[Selected space (MNIST-test)]{\includegraphics[width = 0.24\linewidth]{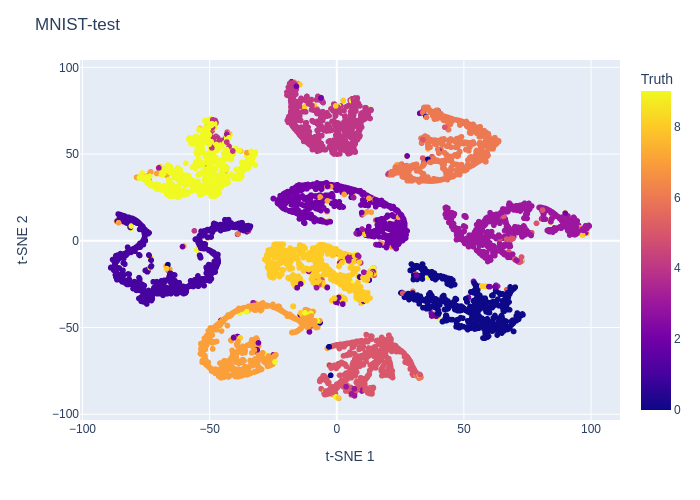}}
\subfigure[Excluded space (MNIST-test)]{\includegraphics[width = 0.24\linewidth]{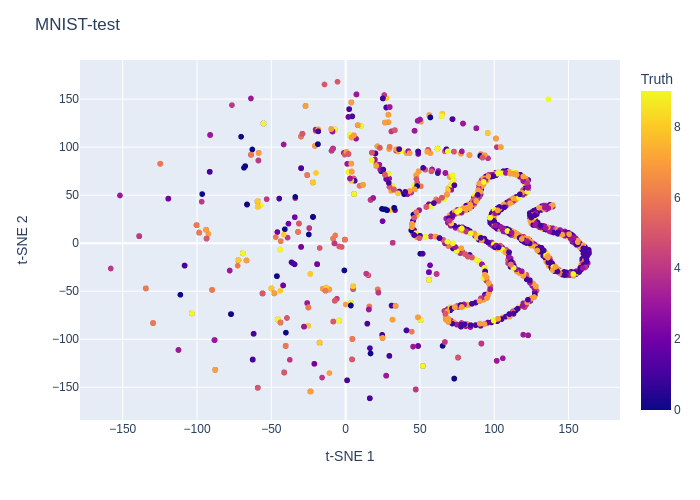}}
\subfigure[Selected space (CMU-PIE)]{\includegraphics[width = 0.24\linewidth]{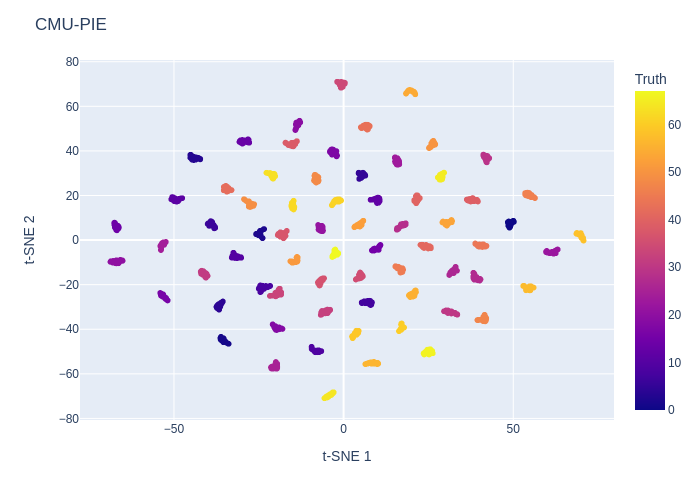}}
\subfigure[Excluded space (CMU-PIE)]{\includegraphics[width = 0.24\linewidth]{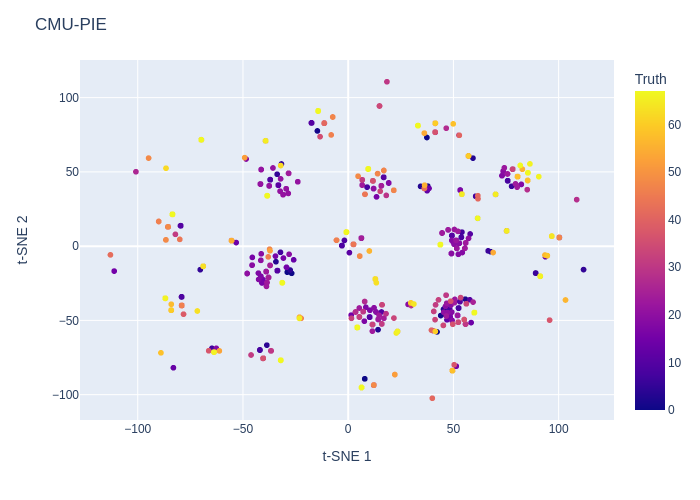}}
\caption{t-SNE visualization illustrating the selected embedding spaces from ACE in comparison to those excluded from ACE, based on Davies-Bouldin index, for Application 1 with DEPICT. Each data point in the visualizations is assigned a color corresponding to its true cluster label.}
\label{fig:tsne:dav:1}
\end{figure}
%%%%%%%%%%%%%%%%%%%%%%%%%
\begin{figure}[htbp!]
\centering
\subfigure[USPS]{\includegraphics[width = 0.3\linewidth]{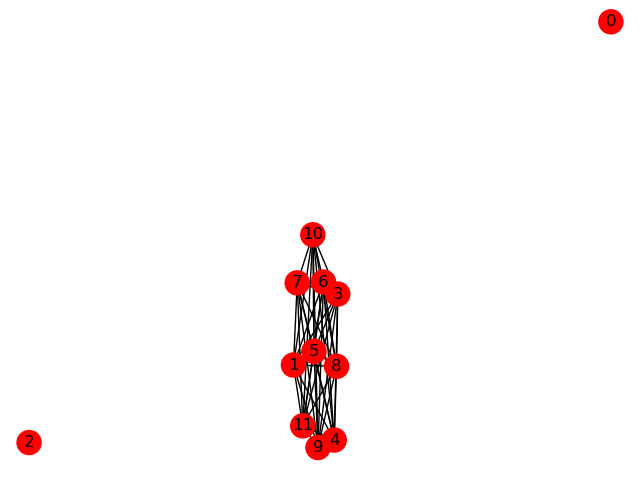}}
\subfigure[YTF]{\includegraphics[width = 0.3\linewidth]{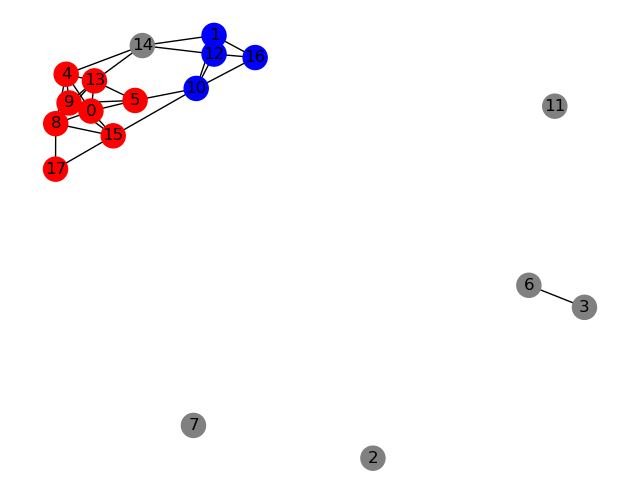}}
\subfigure[FRGC]{\includegraphics[width = 0.3\linewidth]{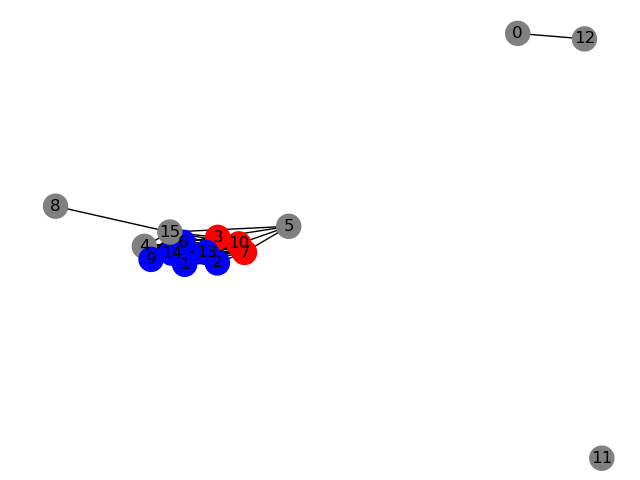}}
\subfigure[MNIST-test]{\includegraphics[width = 0.3\linewidth]{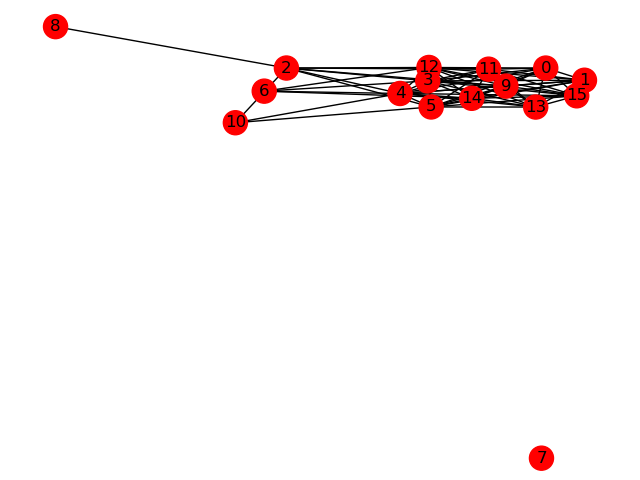}}
\subfigure[CMU-PIE]{\includegraphics[width = 0.3\linewidth]{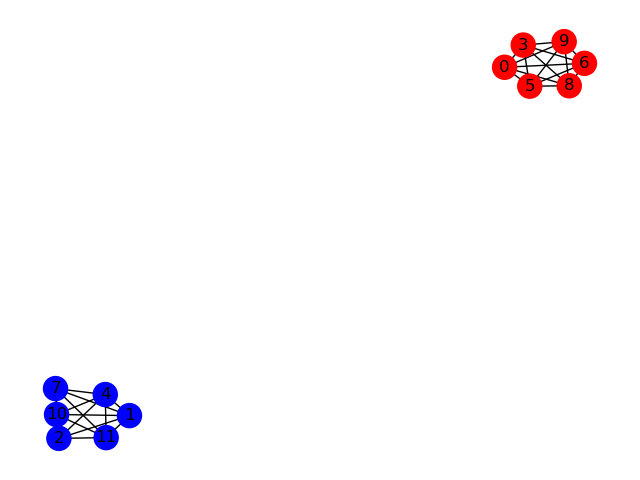}}
\caption{Graph depicting rank correlation based on Calinski-Harabasz index among embedding spaces for Application 1 with DEPICT. Each node represents an embedding space, and each edge signifies a significant rank correlation.}
\label{fig:graph:ch:1}
\end{figure}
%%%%%%%%%%%%%%%%%%%%%%%%%
\begin{figure}[htbp!]
\centering
\subfigure[Selected space (USPS)]{\includegraphics[width = 0.24\linewidth]{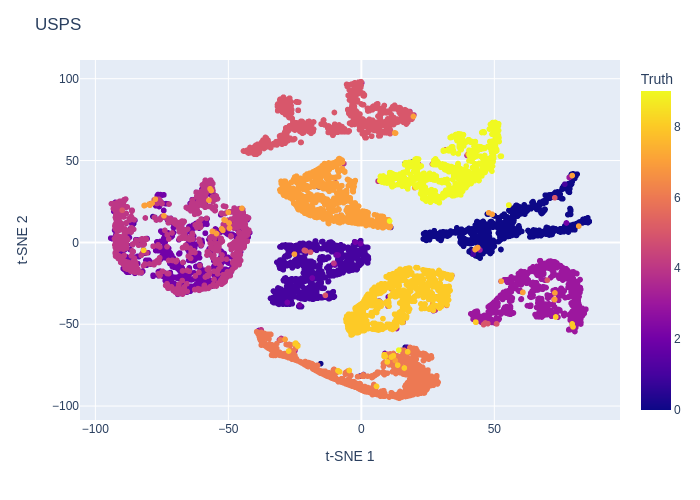}}
\subfigure[Excluded space (USPS)]{\includegraphics[width = 0.24\linewidth]{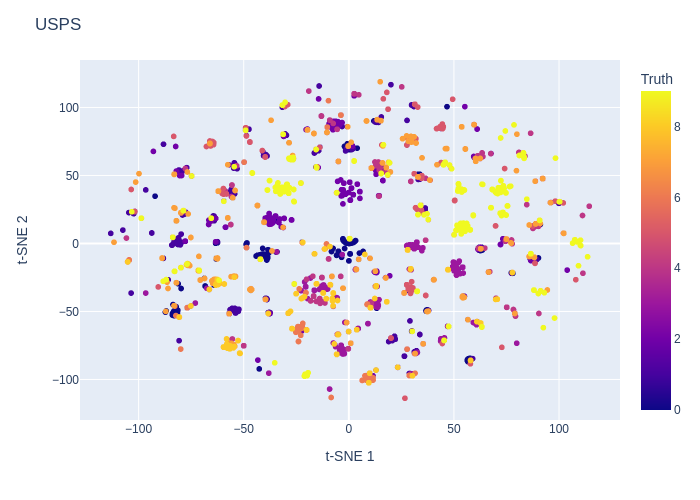}}
\subfigure[Selected space (YTF)]{\includegraphics[width = 0.24\linewidth]{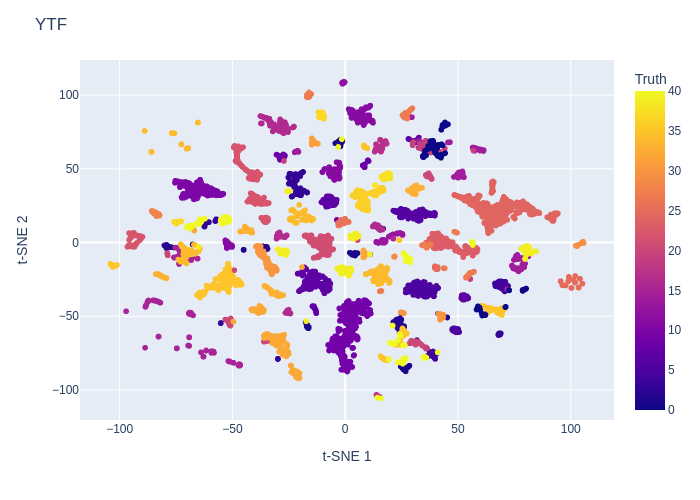}}
\subfigure[Excluded space (YTF)]{\includegraphics[width = 0.24\linewidth]{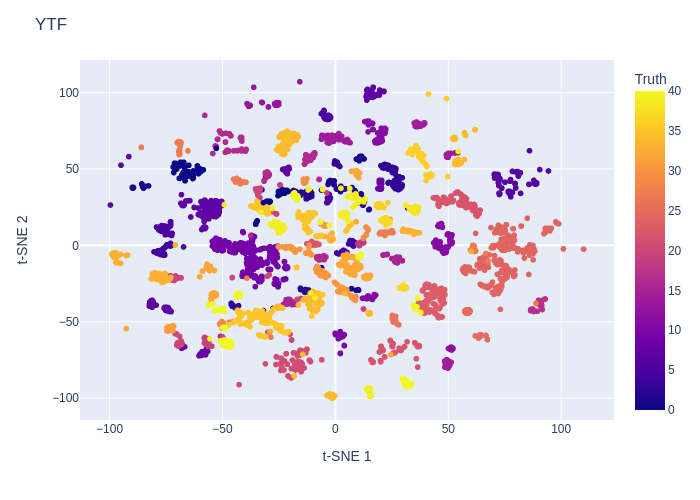}}
\subfigure[Selected space (FRGC)]{\includegraphics[width = 0.24\linewidth]{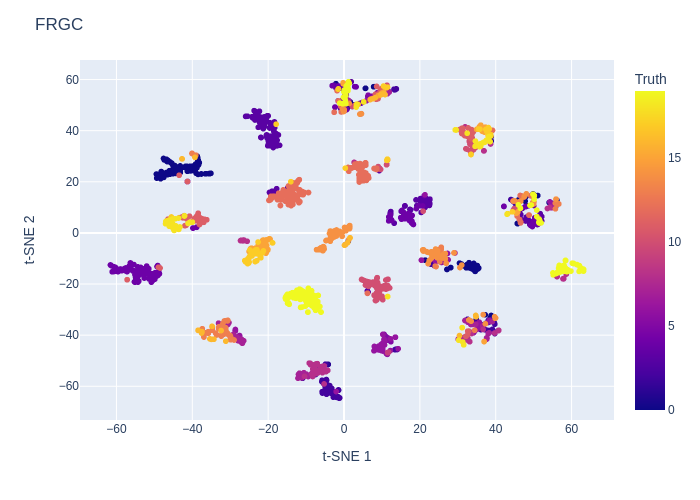}}
\subfigure[Excluded space (FRGC)]{\includegraphics[width = 0.24\linewidth]{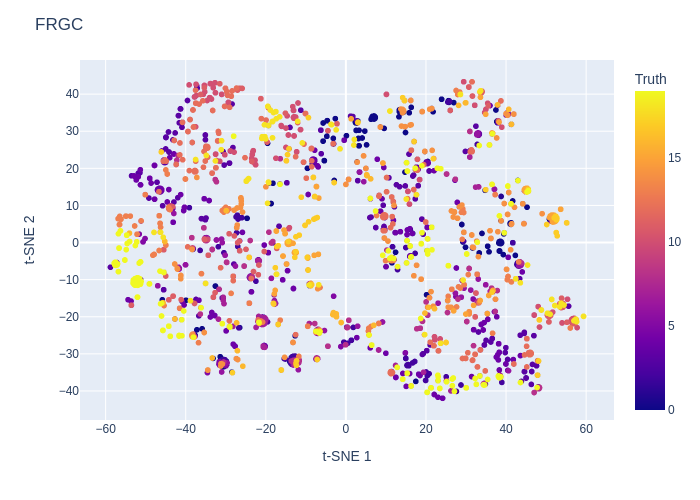}}
\subfigure[Selected space (MNIST-test)]{\includegraphics[width = 0.24\linewidth]{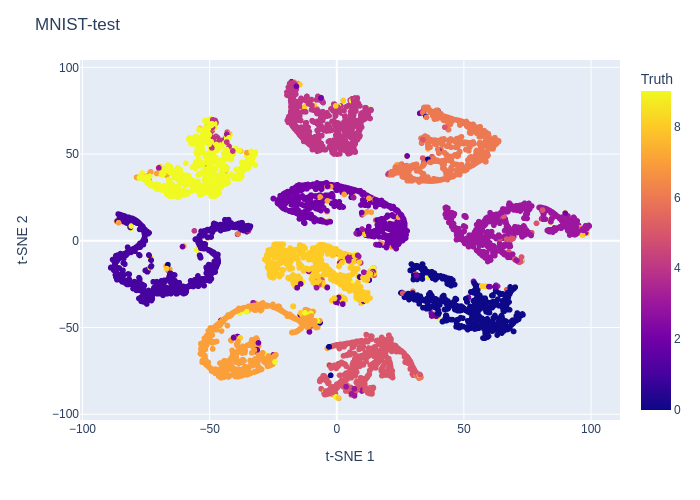}}
\subfigure[Excluded space (MNIST-test)]{\includegraphics[width = 0.24\linewidth]{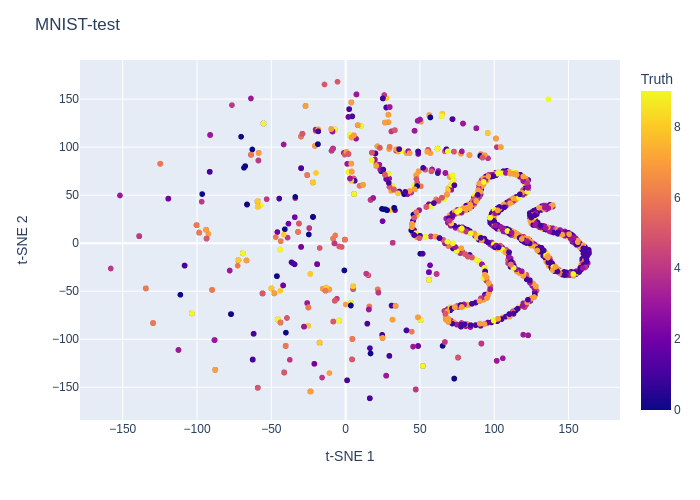}}
\subfigure[Selected space (CMU-PIE)]{\includegraphics[width = 0.24\linewidth]{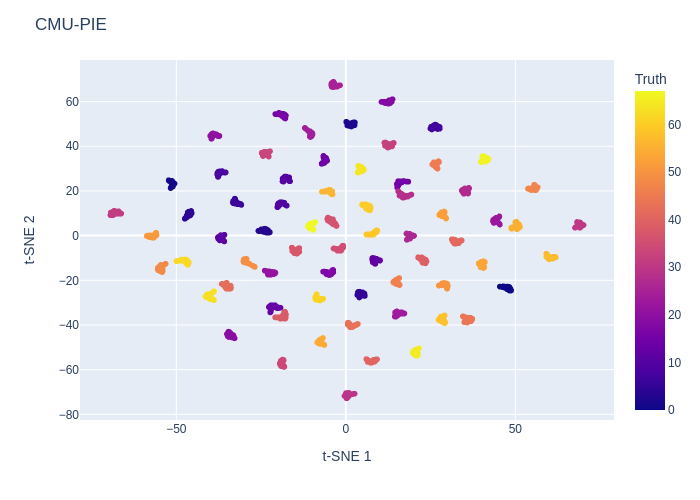}}
\subfigure[Excluded space (CMU-PIE)]{\includegraphics[width = 0.24\linewidth]{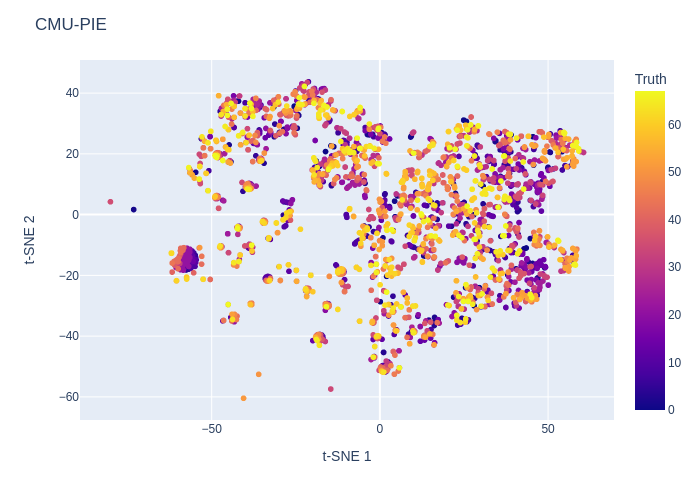}}
\caption{t-SNE visualization illustrating the selected embedding spaces from ACE in comparison to those excluded from ACE, based on Calinski-Harabasz index, for Application 1 with DEPICT. Each data point in the visualizations is assigned a color corresponding to its true cluster label.}
\label{fig:tsne:ch:1}
\end{figure}
%%%%%%%%%%%%%%%%%%%%%%%%%
\begin{figure}[htbp!]
\centering
\subfigure[USPS]{\includegraphics[width = 0.3\linewidth]{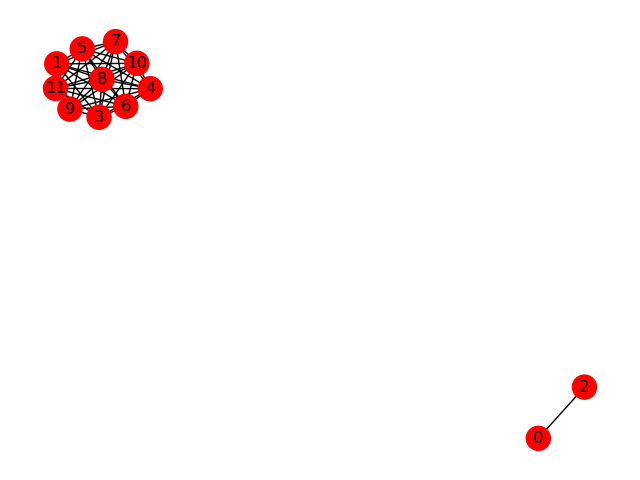}}
\subfigure[YTF]{\includegraphics[width = 0.3\linewidth]{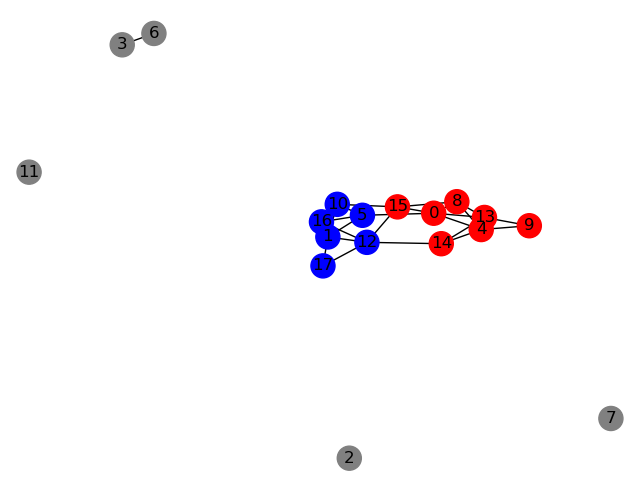}}
\subfigure[FRGC]{\includegraphics[width = 0.3\linewidth]{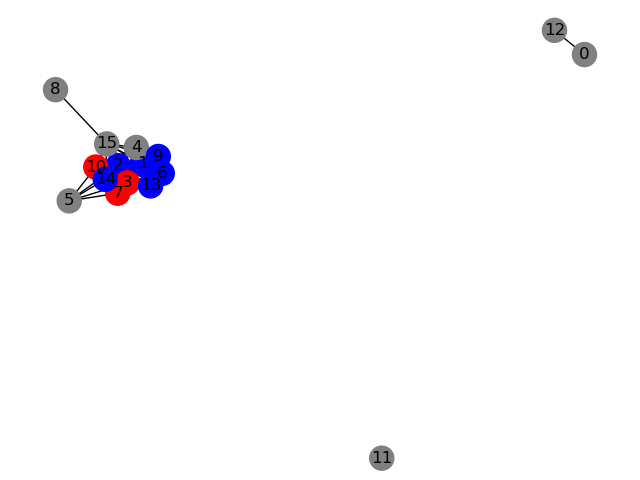}}
\subfigure[MNIST-test]{\includegraphics[width = 0.3\linewidth]{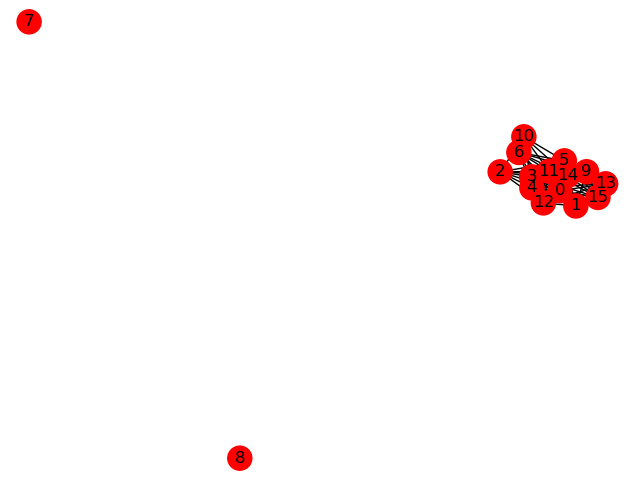}}
\subfigure[CMU-PIE]{\includegraphics[width = 0.3\linewidth]{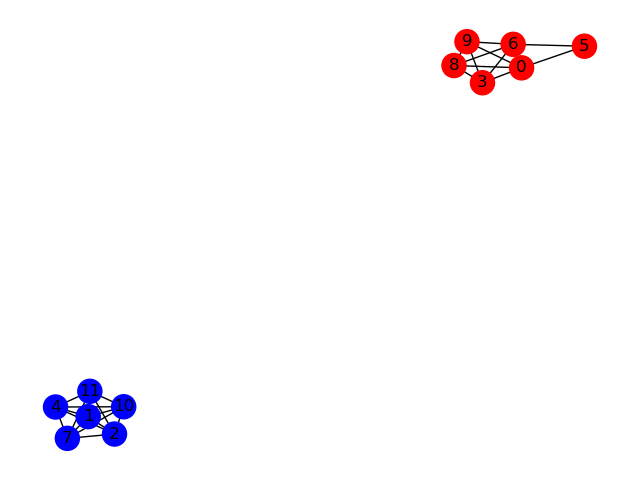}}
\caption{Graph depicting rank correlation based on Silhouette score (Cosine distance) among embedding spaces for Application 1 with DEPICT. Each node represents an embedding space, and each edge signifies a significant rank correlation.}
\label{fig:graph:cosine:1}
\end{figure}
%%%%%%%%%%%%%%%%%%%%%%%%%
\begin{figure}[htbp!]
\centering
\subfigure[Selected space (USPS)]{\includegraphics[width = 0.24\linewidth]{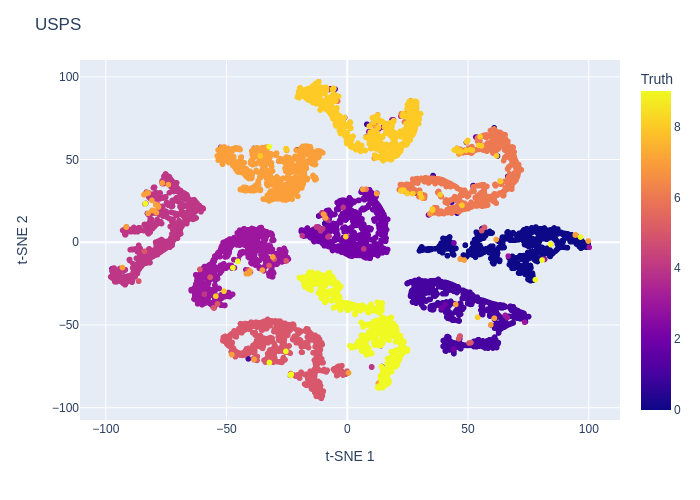}}
\subfigure[Excluded space (USPS)]{\includegraphics[width = 0.24\linewidth]{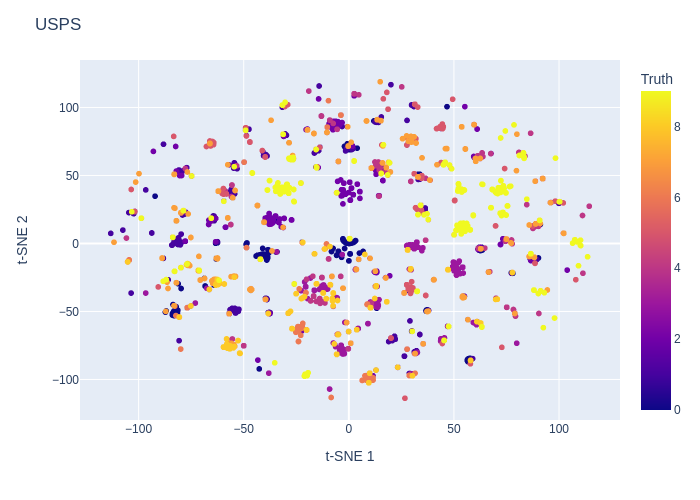}}
\subfigure[Selected space (YTF)]{\includegraphics[width = 0.24\linewidth]{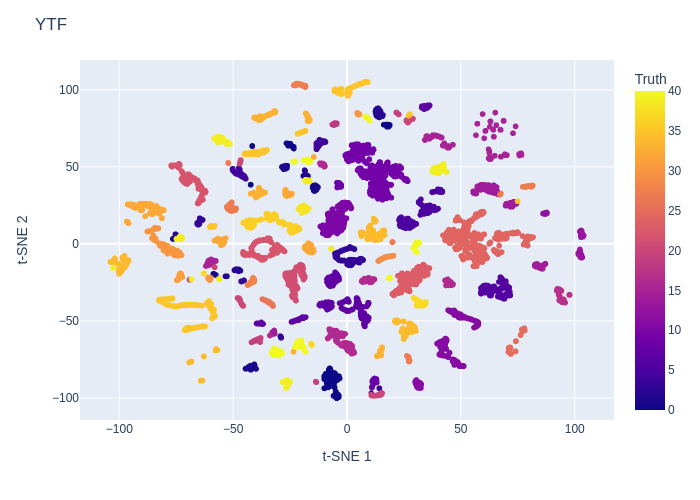}}
\subfigure[Excluded space (YTF)]{\includegraphics[width = 0.24\linewidth]{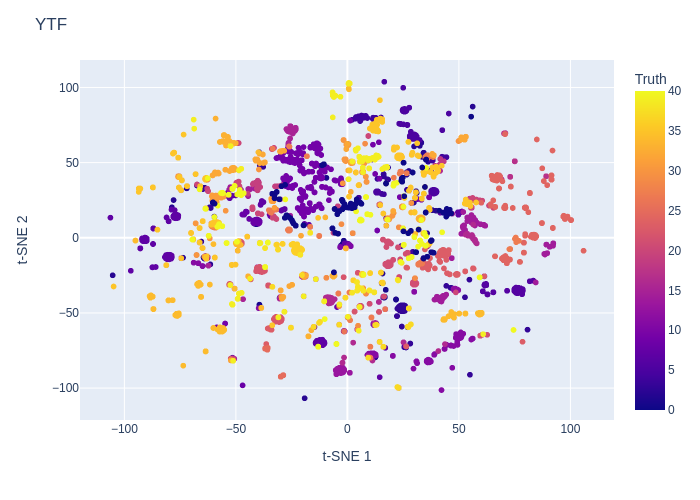}}
\subfigure[Selected space (FRGC)]{\includegraphics[width = 0.24\linewidth]{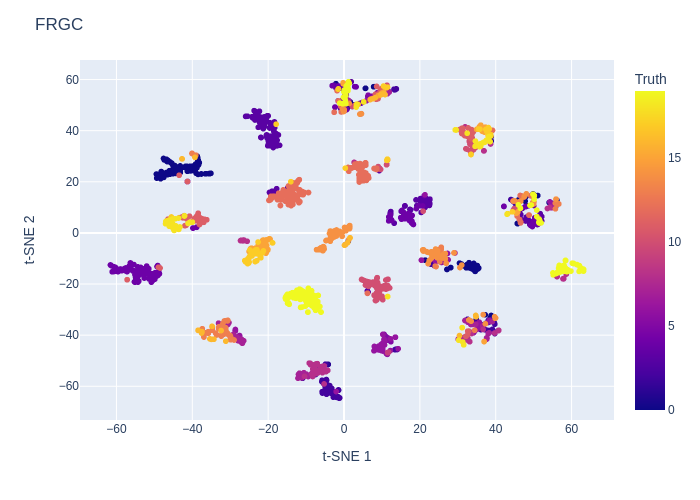}}
\subfigure[Excluded space (FRGC)]{\includegraphics[width = 0.24\linewidth]{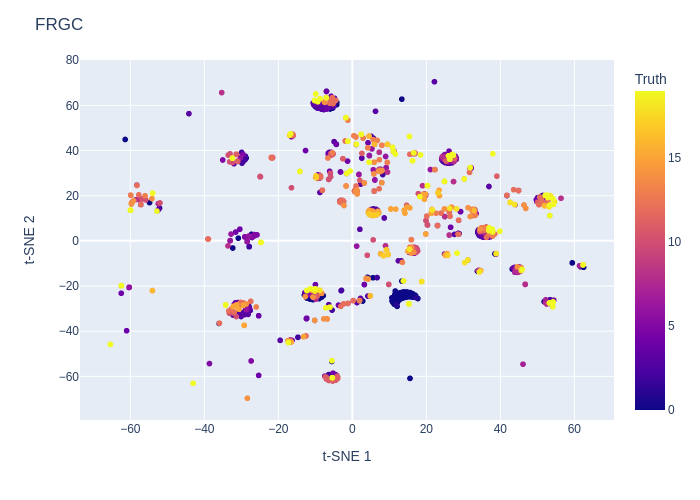}}
\subfigure[Selected space (MNIST-test)]{\includegraphics[width = 0.24\linewidth]{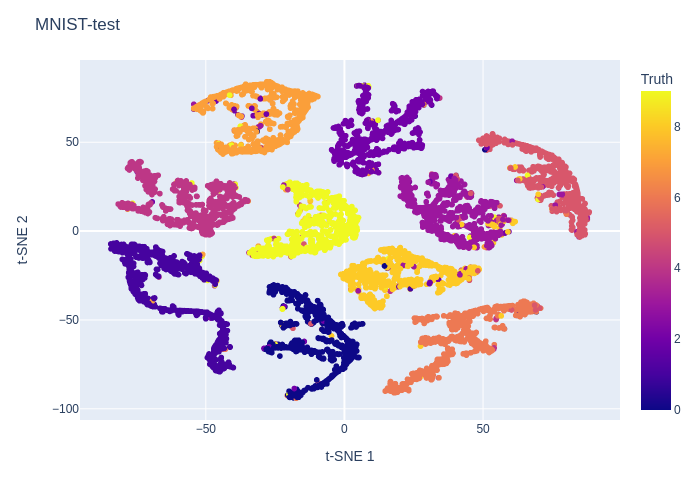}}
\subfigure[Excluded space (MNIST-test)]{\includegraphics[width = 0.24\linewidth]{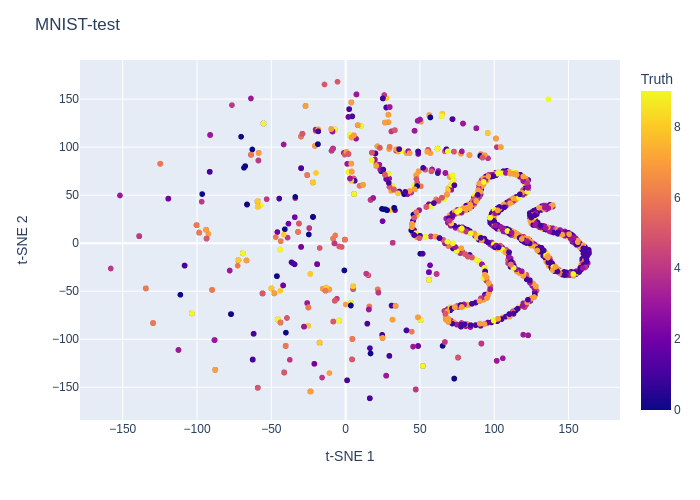}}
\subfigure[Selected space (CMU-PIE)]{\includegraphics[width = 0.24\linewidth]{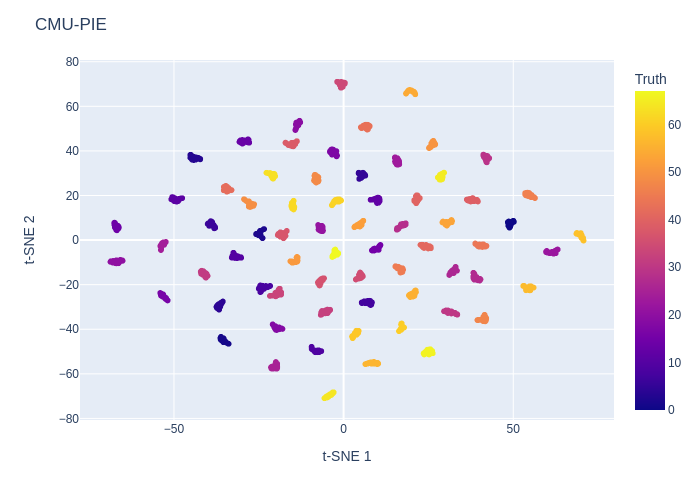}}
\subfigure[Excluded space (CMU-PIE)]{\includegraphics[width = 0.24\linewidth]{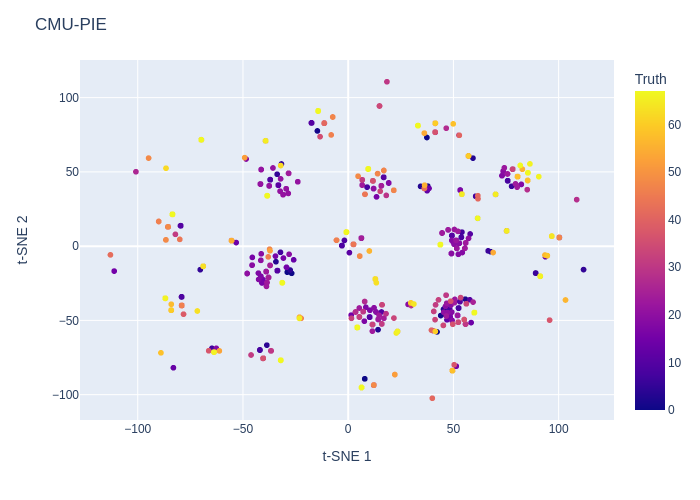}}
\caption{t-SNE visualization illustrating the selected embedding spaces from ACE in comparison to those excluded from ACE, based on Silhouette score (Cosine distance), for Application 1 with DEPICT. Each data point in the visualizations is assigned a color corresponding to its true cluster label.}
\label{fig:tsne:cosine:1}
\end{figure}
%%%%%%%%%%%%%%%%%%%%%%%%%
\begin{figure}[htbp!]
\centering
\subfigure[USPS]{\includegraphics[width = 0.3\linewidth]{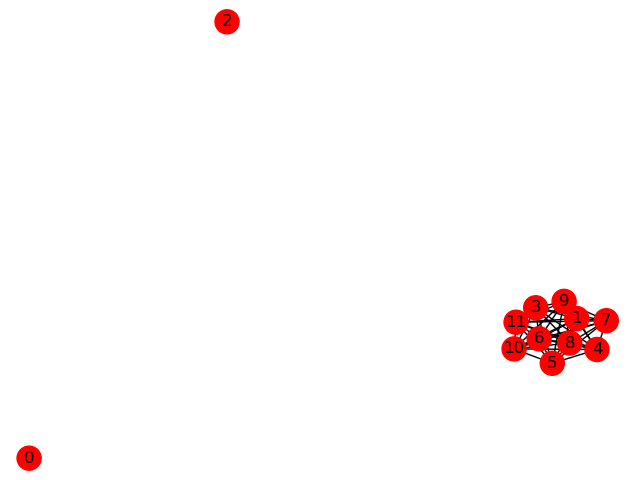}}
\subfigure[YTF]{\includegraphics[width = 0.3\linewidth]{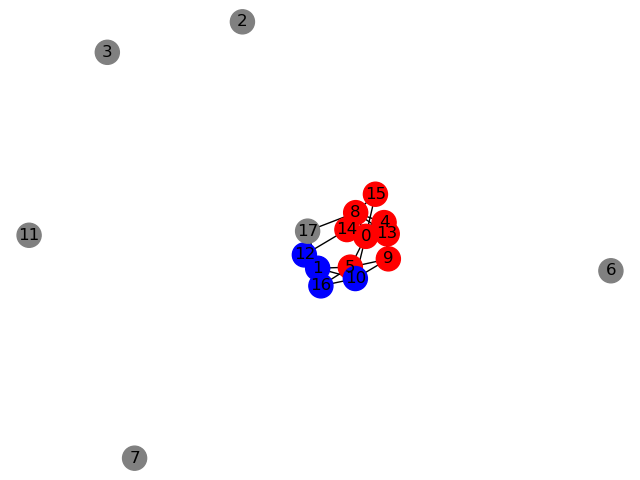}}
\subfigure[FRGC]{\includegraphics[width = 0.3\linewidth]{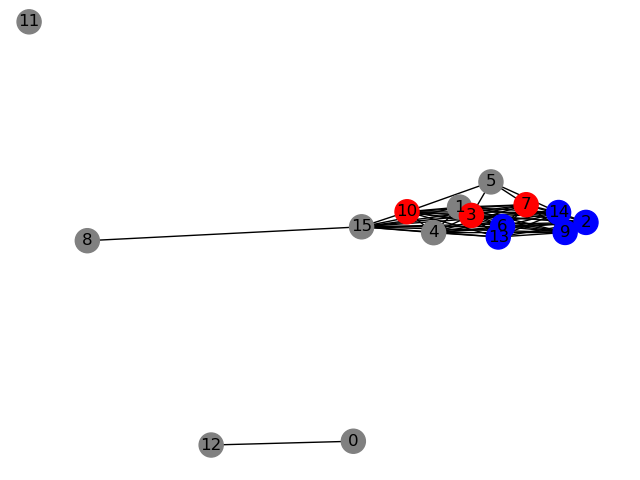}}
\subfigure[MNIST-test]{\includegraphics[width = 0.3\linewidth]{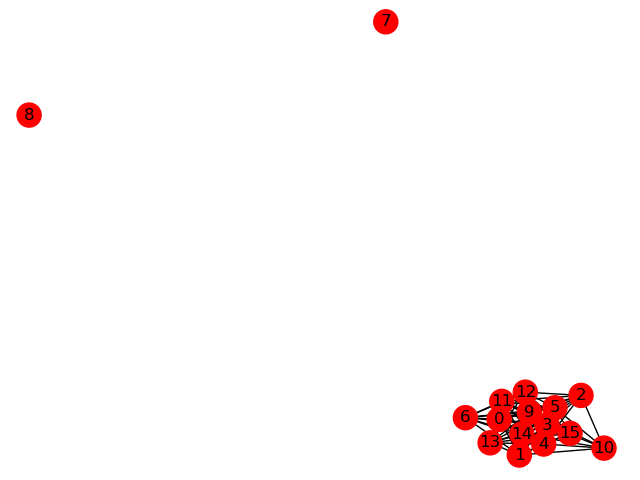}}
\subfigure[CMU-PIE]{\includegraphics[width = 0.3\linewidth]{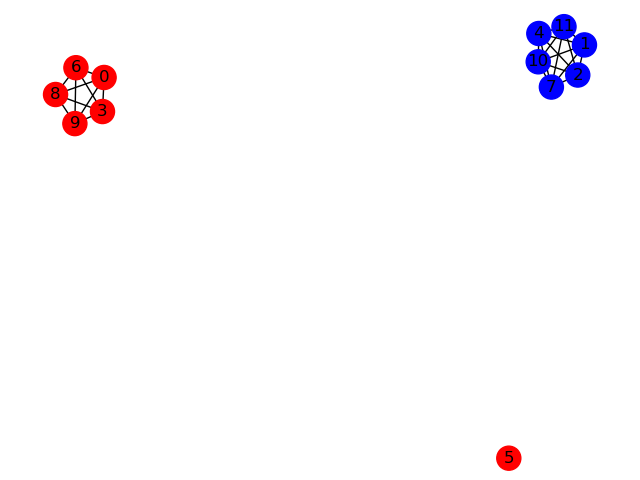}}
\caption{Graph depicting rank correlation based on Silhouette score (Euclidean distance) among embedding spaces for Application 1 with DEPICT. Each node represents an embedding space, and each edge signifies a significant rank correlation.}
\label{fig:graph:euclidean:1}
\end{figure}
%%%%%%%%%%%%%%%%%%%%%%%%%
\begin{figure}[htbp!]
\centering
\subfigure[Selected space (USPS)]{\includegraphics[width = 0.24\linewidth]{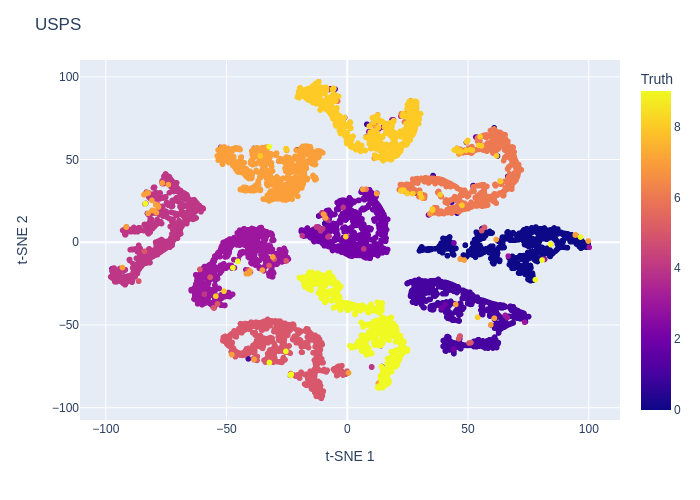}}
\subfigure[Excluded space (USPS)]{\includegraphics[width = 0.24\linewidth]{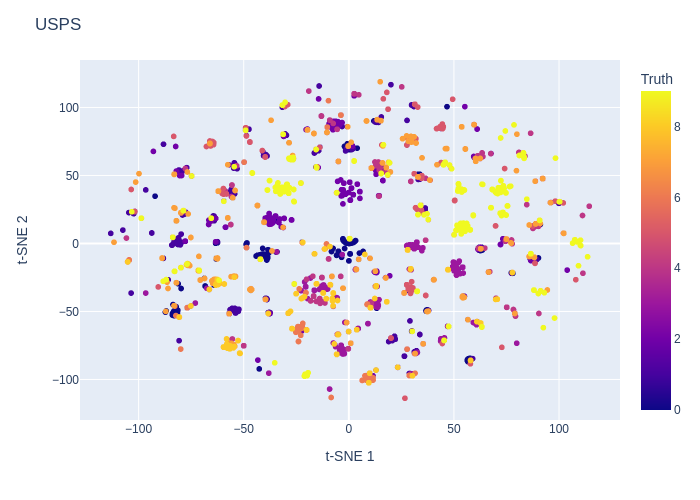}}
\subfigure[Selected space (YTF)]{\includegraphics[width = 0.24\linewidth]{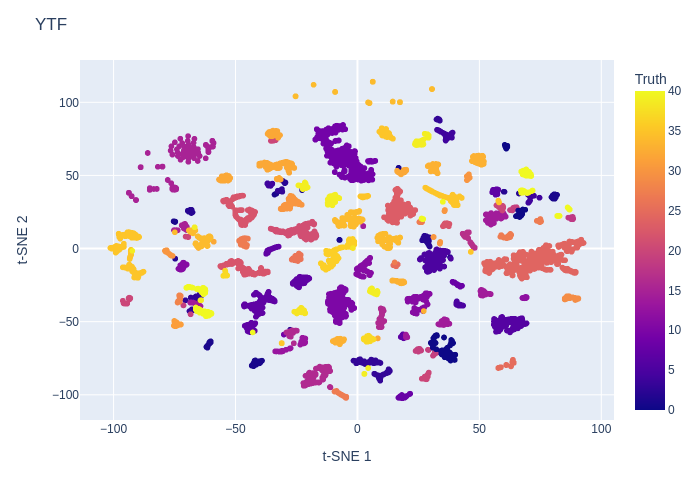}}
\subfigure[Excluded space (YTF)]{\includegraphics[width = 0.24\linewidth]{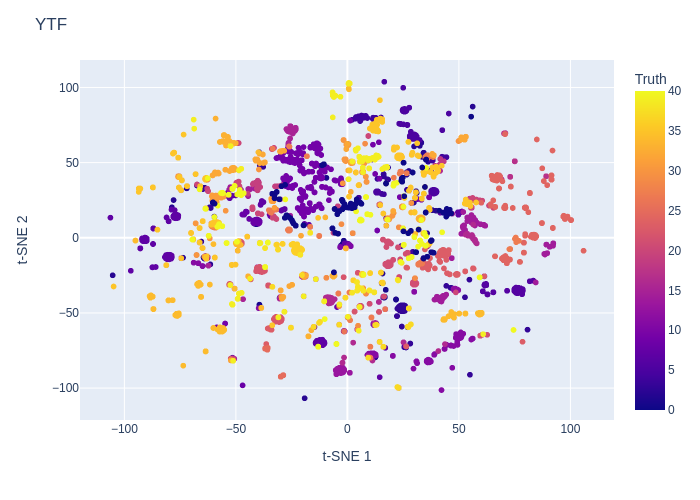}}
\subfigure[Selected space (FRGC)]{\includegraphics[width = 0.24\linewidth]{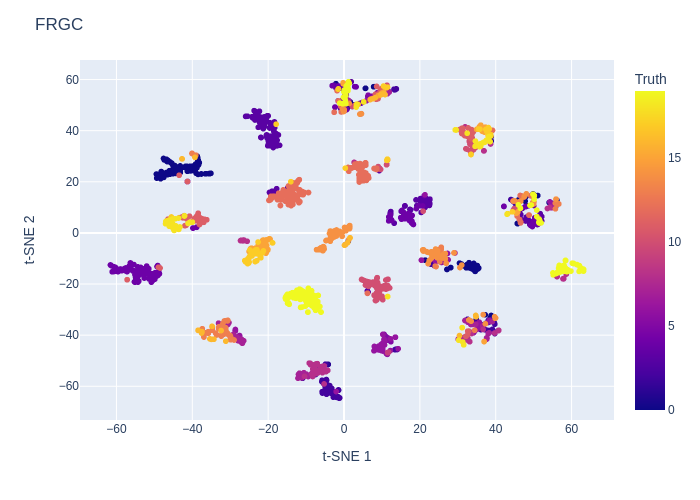}}
\subfigure[Excluded space (FRGC)]{\includegraphics[width = 0.24\linewidth]{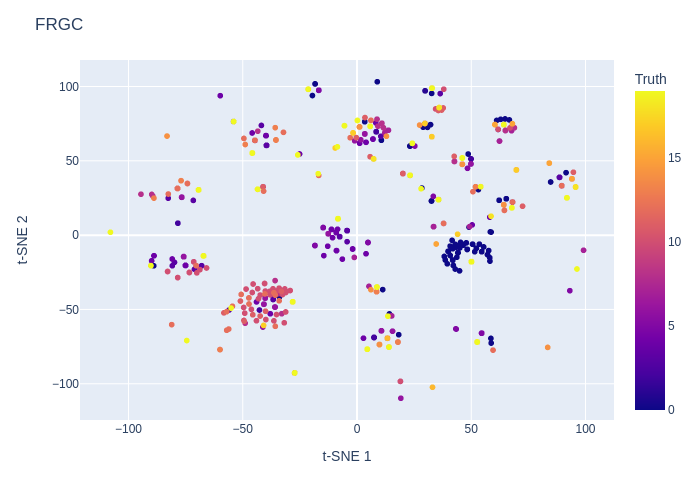}}
\subfigure[Selected space (MNIST-test)]{\includegraphics[width = 0.24\linewidth]{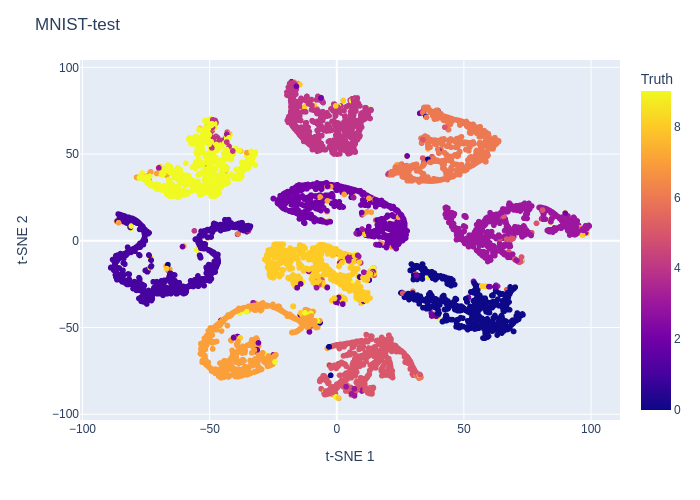}}
\subfigure[Excluded space (MNIST-test)]{\includegraphics[width = 0.24\linewidth]{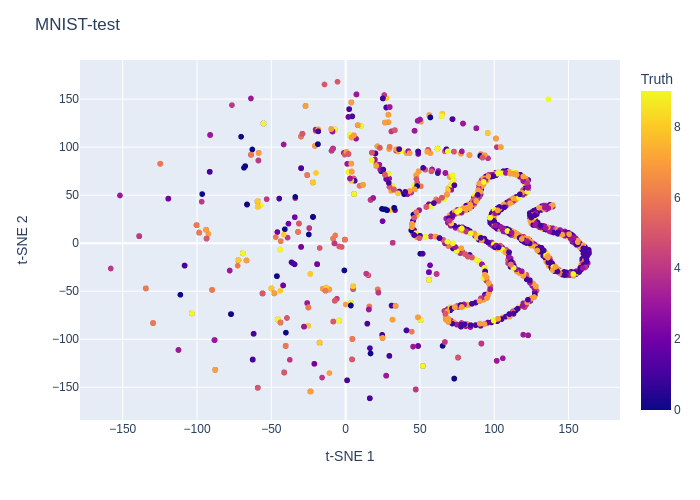}}
\subfigure[Selected space (CMU-PIE)]{\includegraphics[width = 0.24\linewidth]{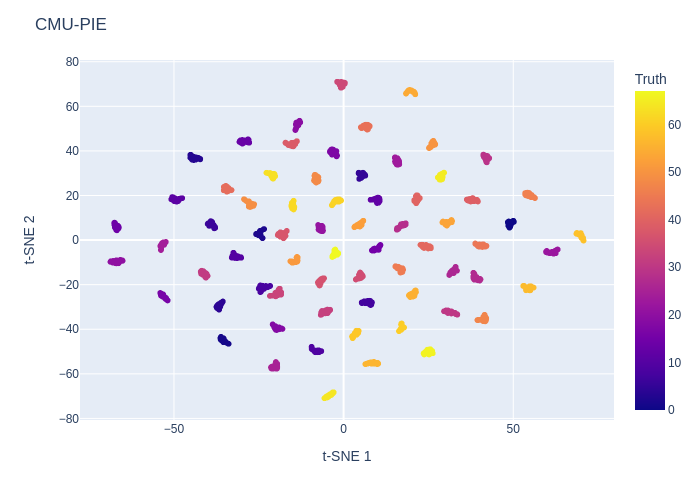}}
\subfigure[Excluded space (CMU-PIE)]{\includegraphics[width = 0.24\linewidth]{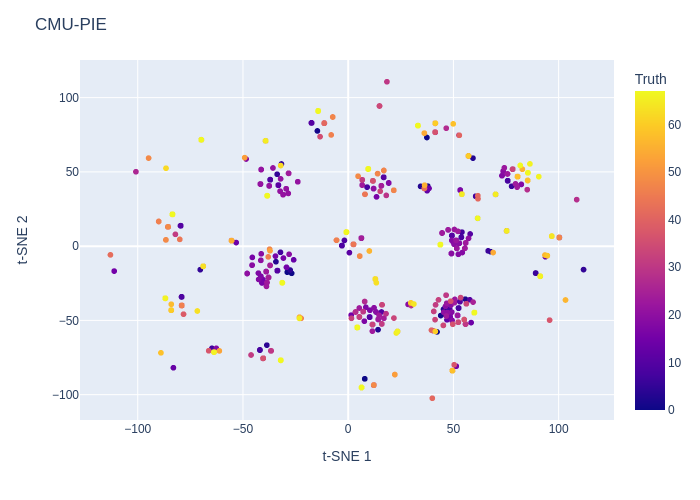}}
\caption{t-SNE visualization illustrating the selected embedding spaces from ACE in comparison to those excluded from ACE, based on Silhouette score (Euclidean distance), for Application 1 with DEPICT. Each data point in the visualizations is assigned a color corresponding to its true cluster label.}
\label{fig:tsne:euclidean:1}
\end{figure}
%%%%%%%%%%%%%%%%%%%%%%%%%
%%%%%%%%%%%%%%%%%%%%%%%%%
%%%%%%%%%%%%%%%%%%%%%%%%%%%%%%%%%%%%%%%%%%%%%%%%%%%%%%%%%%%%%%%%%%%%%%%%%%%%%%%%%%%%%%%%%%%%%%%%%%%%%%%%%%%%%%
\clearpage
\newpage \subsection{Application 2 - supplementary results for additional tables} \label{app:exp:ar:3}

In this section, we present additional tables for Application 2. First, we provide Table \ref{tab:acc:det}, which illustrates rank consistency with ACC based on the three internal measures discussed in the main text across different evaluation approaches. The findings in Table \ref{tab:acc:det} align with those in Table \ref{tab:nmi:det}, demonstrating similar conclusions when using different external measures like NMI and ACC, and further reinforcing our results.

%%%%%%%%%%%%%%%%%%%%%%%%%%%%%%%%%%%%%%%%%%%%%%%%%%%%%
\begin{table*}[htbp!]
  \centering
  \caption{Rank consistency with ACC based on the three internal measures discussed in the main text for different evaluation approaches in Application 2. The top results for each dataset are highlighted in bold.}
  \resizebox{\textwidth}{!}{
\begin{tabular}{lllllllllllllllllll}
\toprule
 & \multicolumn{2}{l}{USPS} & \multicolumn{2}{l}{YTF} & \multicolumn{2}{l}{FRGC} & \multicolumn{2}{l}{MNIST-test} & \multicolumn{2}{l}{CMU-PIE} & \multicolumn{2}{l}{UMist} & \multicolumn{2}{l}{COIL-20} & \multicolumn{2}{l}{COIL-100} & \multicolumn{2}{l}{Average} \\
 & $r_s$ & $\tau_B$ & $r_s$ & $\tau_B$ & $r_s$ & $\tau_B$ & $r_s$ & $\tau_B$ & $r_s$ & $\tau_B$ & $r_s$ & $\tau_B$ & $r_s$ & $\tau_B$ & $r_s$ & $\tau_B$ & $r_s$ & $\tau_B$ \\
\midrule
\hline 
 \multicolumn{19}{c}{JULE: Calinski-Harabasz index} \\
 \hline 
Raw score & 0.71 & 0.64 & \textbf{1.00} & \textbf{1.00} & -0.46 & -0.25 & 0.41 & 0.47 & -0.38 & -0.29 & -0.09 & -0.02 & \textbf{0.76} & \textbf{0.71} & 0.36 & 0.33 & 0.29 & 0.32 \\
Paired score & \textbf{0.84} & \textbf{0.73} & 0.03 & -0.06 & -0.49 & -0.31 & \textbf{0.61} & \textbf{0.56} & -0.09 & -0.07 & \textbf{-0.04} & \textbf{0.07} & 0.74 & 0.64 & 0.60 & 0.51 & 0.27 & 0.26 \\
Pooled score & \textbf{0.84} & \textbf{0.73} & 0.88 & 0.78 & -0.37 & -0.20 & \textbf{0.61} & \textbf{0.56} & \textbf{0.85} & \textbf{0.69} & -0.07 & 0.02 & \textbf{0.76} & \textbf{0.71} & 0.56 & 0.51 & 0.51 & 0.48 \\
\textbf{ACE} & \textbf{0.84} & \textbf{0.73} & 0.92 & 0.83 & \textbf{-0.11} & \textbf{-0.03} & \textbf{0.61} & \textbf{0.56} & 0.83 & \textbf{0.69} & -0.07 & 0.02 & \textbf{0.76} & \textbf{0.71} & \textbf{0.65} & \textbf{0.56} & \textbf{0.55} & \textbf{0.51} \\
\hline 
 \multicolumn{19}{c}{JULE: Davies-Bouldin index} \\
 \hline 
Raw score & -0.49 & -0.38 & \textbf{0.85} & \textbf{0.67} & 0.37 & 0.20 & -0.41 & -0.38 & 0.77 & 0.51 & \textbf{0.02} & \textbf{-0.16} & -0.86 & -0.71 & -0.82 & -0.78 & -0.07 & -0.13 \\
Paired score & 0.39 & 0.29 & 0.10 & 0.06 & 0.37 & 0.25 & 0.49 & 0.33 & 0.83 & 0.60 & -0.28 & -0.29 & \textbf{-0.29} & \textbf{-0.21} & -0.87 & -0.73 & 0.09 & 0.04 \\
Pooled score & \textbf{0.89} & \textbf{0.73} & 0.80 & \textbf{0.67} & \textbf{0.71} & \textbf{0.54} & \textbf{0.83} & \textbf{0.64} & 0.85 & 0.69 & -0.42 & -0.33 & -0.79 & -0.64 & \textbf{-0.79} & \textbf{-0.69} & \textbf{0.26} & \textbf{0.20} \\
\textbf{ACE} & \textbf{0.89} & \textbf{0.73} & 0.80 & \textbf{0.67} & 0.60 & 0.42 & \textbf{0.83} & \textbf{0.64} & \textbf{0.88} & \textbf{0.73} & -0.42 & -0.33 & -0.71 & -0.64 & -0.82 & \textbf{-0.69} & \textbf{0.26} & 0.19 \\
\hline 
 \multicolumn{19}{c}{JULE: Silhouette score (Cosine distance)} \\
 \hline 
Raw score & 0.58 & 0.42 & 0.95 & 0.89 & 0.52 & 0.42 & -0.01 & -0.02 & -0.32 & -0.16 & \textbf{0.08} & \textbf{0.02} & -0.50 & -0.36 & 0.53 & 0.38 & 0.23 & 0.20 \\
Paired score & 0.89 & 0.78 & 0.27 & 0.22 & 0.21 & 0.09 & 0.81 & 0.64 & \textbf{0.99} & \textbf{0.96} & -0.26 & -0.24 & 0.55 & 0.43 & 0.52 & 0.33 & 0.50 & 0.40 \\
Pooled score & \textbf{0.95} & \textbf{0.87} & \textbf{0.98} & \textbf{0.94} & 0.61 & 0.48 & \textbf{0.94} & \textbf{0.82} & \textbf{0.99} & \textbf{0.96} & -0.32 & -0.24 & 0.67 & 0.50 & 0.54 & 0.38 & 0.67 & 0.59 \\
\textbf{ACE} & \textbf{0.95} & \textbf{0.87} & \textbf{0.98} & \textbf{0.94} & \textbf{0.64} & \textbf{0.54} & \textbf{0.94} & \textbf{0.82} & \textbf{0.99} & \textbf{0.96} & -0.32 & -0.24 & \textbf{0.76} & \textbf{0.57} & \textbf{0.60} & \textbf{0.47} & \textbf{0.69} & \textbf{0.61} \\
\hline 
 \multicolumn{19}{c}{JULE: Silhouette score (Euclidean distance)} \\
 \hline 
Raw score & 0.62 & 0.56 & 0.95 & 0.89 & -0.17 & -0.14 & 0.53 & 0.42 & 0.53 & 0.33 & \textbf{0.04} & \textbf{-0.07} & -0.38 & -0.29 & 0.52 & 0.33 & 0.33 & 0.25 \\
Paired score & 0.93 & 0.82 & 0.30 & 0.28 & 0.21 & 0.09 & 0.82 & 0.64 & 0.98 & 0.91 & -0.13 & -0.16 & 0.52 & 0.36 & 0.55 & 0.42 & 0.52 & 0.42 \\
Pooled score & \textbf{0.95} & \textbf{0.87} & 0.97 & 0.89 & \textbf{0.61} & \textbf{0.48} & \textbf{0.92} & \textbf{0.78} & \textbf{0.99} & \textbf{0.96} & -0.03 & -0.11 & \textbf{0.74} & \textbf{0.50} & \textbf{0.59} & \textbf{0.47} & \textbf{0.72} & 0.60 \\
\textbf{ACE} & \textbf{0.95} & \textbf{0.87} & \textbf{0.98} & \textbf{0.94} & 0.57 & \textbf{0.48} & \textbf{0.92} & \textbf{0.78} & \textbf{0.99} & \textbf{0.96} & -0.03 & -0.11 & \textbf{0.74} & \textbf{0.50} & \textbf{0.59} & \textbf{0.47} & 0.71 & \textbf{0.61} \\
\hline 
 \multicolumn{19}{c}{DEPICT: Calinski-Harabasz index} \\
 \hline 
Raw score & \textbf{0.88} & \textbf{0.82} & \textbf{-0.66} & \textbf{-0.51} & -0.40 & -0.28 & \textbf{0.82} & \textbf{0.78} & -0.92 & -0.82 &  &  &  &  &  &  & -0.06 & -0.00 \\
Paired score & \textbf{0.88} & \textbf{0.82} & -0.96 & -0.91 & -0.37 & -0.22 & 0.79 & 0.73 & -0.92 & -0.82 &  &  &  &  &  &  & -0.11 & -0.08 \\
Pooled score & \textbf{0.88} & \textbf{0.82} & -0.94 & -0.87 & -0.37 & -0.22 & \textbf{0.82} & \textbf{0.78} & 0.44 & 0.56 &  &  &  &  &  &  & 0.17 & 0.21 \\
\textbf{ACE} & \textbf{0.88} & \textbf{0.82} & -0.67 & -0.56 & \textbf{0.92} & \textbf{0.78} & \textbf{0.82} & \textbf{0.78} & \textbf{0.92} & \textbf{0.82} &  &  &  &  &  &  & \textbf{0.57} & \textbf{0.53} \\
\hline 
 \multicolumn{19}{c}{DEPICT: Davies-Bouldin index} \\
 \hline 
Raw score & -0.82 & -0.64 & \textbf{1.00} & \textbf{1.00} & 0.03 & -0.11 & -0.50 & -0.33 & 0.92 & 0.82 &  &  &  &  &  &  & 0.13 & 0.15 \\
Paired score & 0.88 & \textbf{0.82} & -0.77 & -0.60 & -0.37 & -0.22 & 0.79 & 0.73 & -0.10 & 0.02 &  &  &  &  &  &  & 0.09 & 0.15 \\
Pooled score & 0.90 & 0.73 & 0.90 & 0.78 & 0.47 & 0.33 & 0.88 & 0.82 & 0.92 & 0.82 &  &  &  &  &  &  & 0.81 & 0.70 \\
\textbf{ACE} & \textbf{0.93} & \textbf{0.82} & 0.96 & 0.91 & \textbf{0.92} & \textbf{0.83} & \textbf{0.93} & \textbf{0.87} & \textbf{0.96} & \textbf{0.91} &  &  &  &  &  &  & \textbf{0.94} & \textbf{0.87} \\
\hline 
 \multicolumn{19}{c}{DEPICT: Silhouette score (Cosine distance)} \\
 \hline 
Raw score & -0.39 & -0.33 & \textbf{0.99} & \textbf{0.96} & 0.52 & 0.39 & 0.76 & 0.56 & -0.43 & -0.33 &  &  &  &  &  &  & 0.29 & 0.25 \\
Paired score & 0.87 & 0.78 & -0.69 & -0.56 & -0.37 & -0.22 & 0.79 & 0.73 & 0.07 & 0.11 &  &  &  &  &  &  & 0.14 & 0.17 \\
Pooled score & 0.90 & 0.73 & 0.67 & 0.51 & 0.68 & 0.56 & 0.90 & 0.82 & 0.98 & 0.91 &  &  &  &  &  &  & 0.83 & 0.71 \\
\textbf{ACE} & \textbf{0.95} & \textbf{0.87} & 0.92 & 0.82 & \textbf{0.80} & \textbf{0.67} & \textbf{0.95} & \textbf{0.87} & \textbf{0.99} & \textbf{0.96} &  &  &  &  &  &  & \textbf{0.92} & \textbf{0.84} \\
\hline 
 \multicolumn{19}{c}{DEPICT: Silhouette score (Euclidean distance)} \\
 \hline 
Raw score & -0.28 & -0.24 & \textbf{0.99} & \textbf{0.96} & -0.20 & -0.17 & 0.66 & 0.51 & -0.43 & -0.33 &  &  &  &  &  &  & 0.15 & 0.14 \\
Paired score & 0.87 & 0.78 & -0.64 & -0.51 & -0.37 & -0.22 & 0.79 & 0.73 & -0.12 & -0.02 &  &  &  &  &  &  & 0.11 & 0.15 \\
Pooled score & \textbf{0.90} & 0.73 & \textbf{0.99} & \textbf{0.96} & 0.68 & \textbf{0.56} & 0.94 & \textbf{0.87} & \textbf{0.99} & \textbf{0.96} &  &  &  &  &  &  & \textbf{0.90} & \textbf{0.81} \\
\textbf{ACE} & 0.88 & \textbf{0.82} & 0.98 & 0.91 & \textbf{0.73} & \textbf{0.56} & \textbf{0.95} & \textbf{0.87} & 0.98 & 0.91 &  &  &  &  &  &  & \textbf{0.90} & \textbf{0.81} \\
\bottomrule
\end{tabular}
}
\label{tab:acc:det}
\end{table*}
%\end{sidewaystable}
%\newpage

Table \ref{tab:app:nmi:det} presents the rank correlation with NMI for five additional internal measures: Cubic Clustering Criterion (CCC), Dunn index, Cindex, SDbw index, and CDbw index. The missing values in raw scores highlight the practical challenges of obtaining them due to computational resource constraints in certain scenarios. In some cases, ACE and pooled scores exhibit superior rank correlation, while in others, they do not outperform paired scores, particularly when all three show negative rank correlation with NMI, which may indicate the absence of admissible spaces for this metric in the dataset. Similar findings are observed in Table \ref{tab:app:acc:det1}, which presents the rank correlation between different scores and ACC across all internal measures.

\begin{table*}[htbp!]
  \centering
  \caption{Rank consistency with NMI for different evaluation approaches based on five additonal internal measures in Application 2. The optimum ($K$) identified by each approach is shown in the cell brackets, and the true ($K$) is indicated in the header brackets.}
  \resizebox{\textwidth}{!}{
\begin{tabular}{lllllllllllllllllll}
\toprule
{} & \multicolumn{2}{l}{USPS (10)} & \multicolumn{2}{l}{YTF (41)} & \multicolumn{2}{l}{FRGC (20)} & \multicolumn{2}{l}{MNIST-test (10)} & \multicolumn{2}{l}{CMU-PIE (68)} & \multicolumn{2}{l}{UMist (20)} & \multicolumn{2}{l}{COIL-20 (20)} & \multicolumn{2}{l}{COIL-100 (100)} & \multicolumn{2}{l}{Average} \\
 & $r_s$ & $\tau_B$ & $r_s$ & $\tau_B$ & $r_s$ & $\tau_B$ & $r_s$ & $\tau_B$ & $r_s$ & $\tau_B$ & $r_s$ & $\tau_B$ & $r_s$ & $\tau_B$ & $r_s$ & $\tau_B$ & $r_s$ & $\tau_B$ \\
\midrule
\hline 
 \multicolumn{19}{c}{\emph{JULE}: Cubic clustering criterion} \\
 \hline 
Raw score & - & - & - & - & - & - & - & - & - & - & - & - & - & - & - & - & - & - \\
Paired score & 0.79 (10) & 0.73 (10) & -0.22 (50) & -0.17 (50) & 0.65 (25) & 0.5 (25) & 0.82 (10) & 0.69 (10) & 0.9 (90) & 0.78 (90) & 0.33 (35) & 0.29 (35) & 0.43 (40) & 0.29 (40) & 0.41 (200) & 0.2 (200) & 0.51 & 0.41 \\
Pooled score & 0.87 (10) & 0.78 (10) & 0.78 (50) & 0.61 (50) & 0.67 (45) & 0.56 (45) & 0.84 (10) & 0.73 (10) & 0.83 (100) & 0.69 (100) & 0.24 (45) & 0.24 (45) & -0.74 (45) & -0.64 (45) & 0.39 (200) & 0.2 (200) & 0.48 & 0.40 \\
\textbf{ACE} & 0.87 (10) & 0.78 (10) & 0.92 (50) & 0.78 (50) & 0.53 (25) & 0.39 (25) & 0.84 (10) & 0.73 (10) & 0.83 (100) & 0.69 (100) & 0.24 (45) & 0.24 (45) & -0.69 (50) & -0.57 (50) & 0.39 (200) & 0.2 (200) & 0.49 & 0.41 \\
\hline 
 \multicolumn{19}{c}{\emph{JULE}: Dunn index} \\
 \hline 
Raw score & 0.42 (10,15,35,5) & 0.4 (10,15,35,5) & -0.9 (15) & -0.78 (15) & 0.58 (25,50) & 0.42 (25,50) & 0.39 (15) & 0.33 (15) & 0.74 (100,70,80,90) & 0.65 (100,70,80,90) & -0.1 (15) & -0.09 (15) & 0.64 (20) & 0.57 (20) & -0.33 (40,60) & -0.23 (40,60) & 0.18 & 0.16 \\
Paired score & 0.28 (5) & 0.29 (5) & -0.48 (25) & -0.33 (25) & 0.6 (50) & 0.5 (50) & 0.44 (15) & 0.42 (15) & 0.55 (50) & 0.38 (50) & -0.2 (5) & -0.11 (5) & 0.64 (15) & 0.5 (15) & 0.12 (60) & 0.11 (60) & 0.24 & 0.22 \\
Pooled score & 0.43 (5) & 0.51 (5) & -0.93 (11) & -0.83 (11) & -0.4 (10) & -0.33 (10) & 0.14 (5) & 0.16 (5) & -0.27 (50) & -0.16 (50) & 0.02 (5) & 0.07 (5) & 0.48 (15) & 0.36 (15) & -0.39 (60) & -0.2 (60) & -0.11 & -0.05 \\
\textbf{ACE} & 0.43 (5) & 0.51 (5) & -0.98 (11) & -0.94 (11) & -0.2 (20) & -0.06 (20) & 0.14 (5) & 0.16 (5) & -0.27 (50) & -0.16 (50) & 0.02 (5) & 0.07 (5) & 0.38 (15) & 0.29 (15) & -0.44 (60) & -0.24 (60) & -0.12 & -0.05 \\
\hline 
 \multicolumn{19}{c}{\emph{JULE}: Cindex} \\
 \hline 
Raw score & -0.41 (40) & -0.47 (40) & -0.02 (40) & 0.11 (40) & 0.75 (35) & 0.56 (35) & -0.19 (30) & -0.11 (30) & 0.77 (100) & 0.56 (100) & 0.12 (35) & 0.11 (35) & -0.69 (50) & -0.57 (50) & 0.41 (180) & 0.24 (180) & 0.09 & 0.05 \\
Paired score & -0.12 (45) & -0.16 (45) & -0.5 (11) & -0.39 (11) & 0.47 (30) & 0.39 (30) & -0.03 (30) & 0.02 (30) & -0.56 (20) & -0.38 (20) & 0.07 (45) & 0.16 (45) & -0.43 (50) & -0.36 (50) & 0.55 (140) & 0.38 (140) & -0.07 & -0.04 \\
Pooled score & -0.44 (45) & -0.56 (45) & 0.88 (50) & 0.72 (50) & 0.98 (50) & 0.94 (50) & -0.34 (40) & -0.38 (40) & 0.77 (100) & 0.6 (100) & 0.12 (50) & 0.02 (50) & -0.59 (50) & -0.5 (50) & 0.44 (200) & 0.29 (200) & 0.23 & 0.14 \\
\textbf{ACE} & -0.42 (45) & -0.51 (45) & 0.63 (25) & 0.5 (25) & 0.98 (50) & 0.94 (50) & -0.34 (40) & -0.38 (40) & 0.76 (100) & 0.56 (100) & 0.13 (50) & 0.07 (50) & -0.59 (50) & -0.5 (50) & 0.39 (200) & 0.2 (200) & 0.19 & 0.11 \\
\hline 
 \multicolumn{19}{c}{\emph{JULE}: SDbw index} \\
 \hline 
Raw score & -0.37 (50) & -0.42 (50) & 0.98 (50) & 0.94 (50) & 0.82 (50) & 0.72 (50) & -0.46 (45) & -0.42 (45) & 0.6 (100) & 0.38 (100) & 0.03 (50) & 0.07 (50) & -0.91 (45) & -0.79 (45) & - & - & 0.10 & 0.07 \\
Paired score & -0.16 (45) & -0.16 (45) & 0.75 (35) & 0.67 (35) & 0.97 (50) & 0.89 (50) & -0.18 (50) & -0.24 (50) & 0.65 (100) & 0.51 (100) & 0.2 (45) & 0.16 (45) & -0.24 (40) & -0.21 (40) & -0.93 (20) & -0.82 (20) & 0.13 & 0.10 \\
Pooled score & -0.43 (45) & -0.51 (45) & 0.98 (50) & 0.94 (50) & 0.98 (50) & 0.94 (50) & -0.39 (45) & -0.42 (45) & 0.77 (100) & 0.56 (100) & 0.19 (50) & 0.16 (50) & -0.74 (50) & -0.64 (50) & -0.99 (20) & -0.96 (20) & 0.05 & 0.01 \\
\textbf{ACE} & -0.43 (45) & -0.51 (45) & 0.98 (50) & 0.94 (50) & 0.98 (50) & 0.94 (50) & -0.39 (45) & -0.42 (45) & 0.77 (100) & 0.56 (100) & 0.19 (50) & 0.16 (50) & -0.74 (50) & -0.64 (50) & -0.99 (20) & -0.96 (20) & 0.05 & 0.01 \\
\hline 
 \multicolumn{19}{c}{\emph{JULE}: CDbw index} \\
 \hline 
Raw score & - & - & - & - & - & - & - & - & - & - & - & - & - & - & - & - & - & - \\
Paired score & 0.01 (15) & -0.02 (15) & -0.25 (30) & -0.11 (30) & 0.82 (45) & 0.61 (45) & -0.19 (45) & -0.16 (45) & -0.52 (20) & -0.33 (20) & 0.09 (45) & 0.11 (45) & 0.26 (15) & 0.14 (15) & -0.73 (20) & -0.6 (20) & -0.06 & -0.04 \\
Pooled score & -0.38 (50) & -0.47 (50) & 0.95 (50) & 0.89 (50) & 0.97 (50) & 0.89 (50) & -0.37 (45) & -0.33 (45) & 0.89 (70) & 0.78 (70) & 0.31 (45) & 0.24 (45) & 0.52 (15) & 0.43 (15) & -0.41 (20) & -0.24 (20) & 0.31 & 0.27 \\
\textbf{ACE} & -0.37 (45) & -0.42 (45) & 0.98 (50) & 0.94 (50) & 0.97 (50) & 0.89 (50) & -0.37 (45) & -0.33 (45) & 0.88 (70) & 0.73 (70) & 0.31 (45) & 0.24 (45) & 0.57 (15) & 0.5 (15) & -0.39 (20) & -0.2 (20) & 0.32 & 0.29 \\
\hline 
 \multicolumn{19}{c}{\emph{DEPICT}: Cubic clustering criterion} \\
 \hline 
Raw score & - & - & - & - & - & - & - & - & - & - &  &  &  &  &  &  & - & - \\
Paired score & -0.08 (25) & -0.07 (25) & 0.98 (50) & 0.91 (50) & 0.53 (25) & 0.39 (25) & 0.13 (35) & 0.11 (35) & 0.98 (80) & 0.91 (80) &  &  &  &  &  &  & 0.51 & 0.45 \\
Pooled score & -0.18 (25) & -0.2 (25) & 1.0 (50) & 1.0 (50) & 0.83 (50) & 0.67 (50) & 0.2 (20) & 0.16 (20) & 0.92 (100) & 0.82 (100) &  &  &  &  &  &  & 0.55 & 0.49 \\
\textbf{ACE} & -0.18 (25) & -0.2 (25) & 1.0 (50) & 1.0 (50) & 0.83 (50) & 0.67 (50) & 0.06 (40) & 0.02 (40) & 0.92 (100) & 0.82 (100) &  &  &  &  &  &  & 0.53 & 0.46 \\
\hline 
 \multicolumn{19}{c}{\emph{DEPICT}: Dunn index} \\
 \hline 
Raw score & -0.16 (5) & -0.11 (5) & 0.82 (50) & 0.69 (50) & 0.83 (35) & 0.67 (35) & 0.07 (10) & 0.02 (10) & 0.2 (100) & 0.24 (100) &  &  &  &  &  &  & 0.35 & 0.30 \\
Paired score & 0.04 (25) & 0.07 (25) & -0.22 (5) & -0.16 (5) & -0.57 (15) & -0.44 (15) & 0.34 (15) & 0.29 (15) & 0.02 (50) & -0.02 (50) &  &  &  &  &  &  & -0.08 & -0.05 \\
Pooled score & -0.12 (5) & -0.07 (5) & -0.32 (5) & -0.24 (5) & 0.0 (30) & 0.0 (30) & 0.24 (10) & 0.2 (10) & 0.22 (50) & 0.16 (50) &  &  &  &  &  &  & 0.00 & 0.01 \\
\textbf{ACE} & -0.38 (5) & -0.29 (5) & 0.04 (15) & 0.02 (15) & 0.0 (30) & 0.0 (30) & 0.24 (5) & 0.24 (5) & 0.22 (50) & 0.16 (50) &  &  &  &  &  &  & 0.02 & 0.03 \\
\hline 
 \multicolumn{19}{c}{\emph{DEPICT}: Cindex} \\
 \hline 
Raw score & -0.22 (40) & -0.24 (40) & 0.65 (35) & 0.42 (35) & 0.72 (40) & 0.56 (40) & -0.42 (45) & -0.38 (45) & 0.85 (100) & 0.73 (100) &  &  &  &  &  &  & 0.32 & 0.22 \\
Paired score & 0.46 (5) & 0.6 (5) & -0.54 (5) & -0.47 (5) & -0.92 (10) & -0.83 (10) & 0.42 (5) & 0.47 (5) & 0.12 (10) & 0.16 (10) &  &  &  &  &  &  & -0.09 & -0.01 \\
Pooled score & -0.44 (50) & -0.56 (50) & 1.0 (50) & 1.0 (50) & 0.85 (50) & 0.72 (50) & -0.37 (50) & -0.42 (50) & 0.92 (100) & 0.82 (100) &  &  &  &  &  &  & 0.39 & 0.31 \\
\textbf{ACE} & -0.44 (50) & -0.56 (50) & 1.0 (50) & 1.0 (50) & 0.78 (50) & 0.61 (50) & -0.37 (50) & -0.42 (50) & 0.92 (100) & 0.82 (100) &  &  &  &  &  &  & 0.38 & 0.29 \\
\hline 
 \multicolumn{19}{c}{\emph{DEPICT}: SDbw index} \\
 \hline 
Raw score & -0.41 (45) & -0.47 (45) & 1.0 (50) & 1.0 (50) & 0.72 (50) & 0.5 (50) & -0.36 (40) & -0.38 (40) & 0.92 (100) & 0.82 (100) &  &  &  &  &  &  & 0.37 & 0.29 \\
Paired score & 0.43 (5) & 0.51 (5) & -0.41 (5) & -0.29 (5) & -0.85 (10) & -0.72 (10) & 0.55 (10) & 0.6 (10) & 0.26 (10) & 0.29 (10) &  &  &  &  &  &  & 0.00 & 0.08 \\
Pooled score & -0.44 (50) & -0.56 (50) & 1.0 (50) & 1.0 (50) & 0.85 (50) & 0.72 (50) & -0.34 (45) & -0.38 (45) & 0.92 (100) & 0.82 (100) &  &  &  &  &  &  & 0.40 & 0.32 \\
\textbf{ACE} & -0.44 (50) & -0.56 (50) & 1.0 (50) & 1.0 (50) & 0.85 (50) & 0.72 (50) & -0.38 (45) & -0.47 (45) & 0.93 (100) & 0.87 (100) &  &  &  &  &  &  & 0.39 & 0.31 \\
\hline 
 \multicolumn{19}{c}{\emph{DEPICT}: CDbw index} \\
 \hline 
Raw score & - & - & - & - & - & - & - & - & -0.04 (100) & 0.0 (100) &  &  &  &  &  &  & -0.04 & 0.00 \\
Paired score & 0.42 (5) & 0.51 (5) & -0.72 (5) & -0.6 (5) & -0.83 (10) & -0.67 (10) & 0.44 (5) & 0.56 (5) & -0.81 (10) & -0.64 (10) &  &  &  &  &  &  & -0.30 & -0.17 \\
Pooled score & -0.55 (50) & -0.47 (50) & -0.07 (5) & 0.07 (5) & -0.9 (10) & -0.78 (10) & -0.76 (5) & -0.56 (5) & 0.85 (100) & 0.73 (100) &  &  &  &  &  &  & -0.29 & -0.20 \\
\textbf{ACE} & -0.01 (50) & -0.11 (50) & 0.39 (5) & 0.42 (5) & -0.25 (10) & -0.17 (10) & -0.73 (5) & -0.51 (5) & 0.9 (100) & 0.78 (100) &  &  &  &  &  &  & 0.06 & 0.08 \\
\bottomrule
\end{tabular}
}
\label{tab:app:nmi:det}
\end{table*}

\begin{table*}[htbp!]
  \centering
  \caption{Rank consistency with ACC for different evaluation approaches based on five additonal internal measures in Application 2.}
  \resizebox{\textwidth}{!}{
\begin{tabular}{lllllllllllllllllll}
\toprule
{} & \multicolumn{2}{l}{USPS (10)} & \multicolumn{2}{l}{YTF (41)} & \multicolumn{2}{l}{FRGC (20)} & \multicolumn{2}{l}{MNIST-test (10)} & \multicolumn{2}{l}{CMU-PIE (68)} & \multicolumn{2}{l}{UMist (20)} & \multicolumn{2}{l}{COIL-20 (20)} & \multicolumn{2}{l}{COIL-100 (100)} & \multicolumn{2}{l}{Average} \\
 & $r_s$ & $\tau_B$ & $r_s$ & $\tau_B$ & $r_s$ & $\tau_B$ & $r_s$ & $\tau_B$ & $r_s$ & $\tau_B$ & $r_s$ & $\tau_B$ & $r_s$ & $\tau_B$ & $r_s$ & $\tau_B$ & $r_s$ & $\tau_B$ \\
\midrule
\hline 
 \multicolumn{19}{c}{\emph{JULE}: Cubic clustering criterion} \\
 \hline 
Raw score & - & - & - & - & - & - & - & - & - & - & - & - & - & - & - & - & - & - \\
Paired score & 0.90 & 0.82 & -0.28 & -0.28 & 0.26 & 0.14 & 0.77 & 0.64 & 0.93 & 0.82 & 0.13 & 0.07 & 0.45 & 0.36 & 0.52 & 0.33 & 0.46 & 0.36 \\
Pooled score & 0.94 & 0.87 & 0.77 & 0.61 & 0.51 & 0.37 & 0.81 & 0.69 & 0.88 & 0.73 & 0.04 & -0.07 & -0.71 & -0.57 & 0.52 & 0.33 & 0.47 & 0.37 \\
\textbf{ACE} & 0.94 & 0.87 & 0.92 & 0.78 & 0.39 & 0.31 & 0.81 & 0.69 & 0.88 & 0.73 & 0.04 & -0.07 & -0.74 & -0.64 & 0.52 & 0.33 & 0.47 & 0.38 \\
\hline 
 \multicolumn{19}{c}{\emph{JULE}: Dunn index} \\
 \hline 
Raw score & 0.71 & 0.60 & -0.83 & -0.67 & 0.31 & 0.23 & 0.42 & 0.38 & 0.78 & 0.69 & -0.10 & -0.04 & 0.62 & 0.50 & -0.52 & -0.36 & 0.17 & 0.17 \\
Paired score & 0.52 & 0.38 & -0.40 & -0.22 & 0.09 & 0.09 & 0.49 & 0.47 & 0.50 & 0.33 & -0.28 & -0.24 & 0.67 & 0.57 & -0.03 & -0.02 & 0.19 & 0.17 \\
Pooled score & 0.69 & 0.60 & -0.85 & -0.72 & -0.35 & -0.20 & 0.20 & 0.20 & -0.37 & -0.20 & -0.07 & -0.07 & 0.52 & 0.43 & -0.54 & -0.33 & -0.10 & -0.04 \\
\textbf{ACE} & 0.69 & 0.60 & -0.93 & -0.83 & -0.31 & -0.25 & 0.20 & 0.20 & -0.37 & -0.20 & -0.07 & -0.07 & 0.45 & 0.36 & -0.59 & -0.38 & -0.12 & -0.07 \\
\hline 
 \multicolumn{19}{c}{\emph{JULE}: Cindex} \\
 \hline 
Raw score & -0.71 & -0.64 & 0.17 & 0.22 & 0.07 & 0.09 & -0.21 & -0.16 & 0.83 & 0.60 & 0.03 & -0.02 & -0.71 & -0.64 & 0.54 & 0.38 & 0.00 & -0.02 \\
Paired score & -0.36 & -0.24 & -0.58 & -0.50 & 0.14 & 0.20 & -0.04 & -0.02 & -0.60 & -0.42 & -0.06 & -0.07 & -0.48 & -0.43 & 0.60 & 0.42 & -0.17 & -0.13 \\
Pooled score & -0.72 & -0.64 & 0.87 & 0.72 & 0.45 & 0.31 & -0.33 & -0.33 & 0.78 & 0.64 & 0.04 & -0.11 & -0.64 & -0.57 & 0.55 & 0.42 & 0.12 & 0.05 \\
\textbf{ACE} & -0.70 & -0.60 & 0.58 & 0.39 & 0.45 & 0.31 & -0.33 & -0.33 & 0.77 & 0.60 & 0.06 & -0.07 & -0.64 & -0.57 & 0.52 & 0.33 & 0.09 & 0.01 \\
\hline 
 \multicolumn{19}{c}{\emph{JULE}: SDbw index} \\
 \hline 
Raw score & -0.61 & -0.51 & 0.93 & 0.83 & 0.18 & 0.09 & -0.41 & -0.38 & 0.66 & 0.42 & -0.15 & -0.24 & -0.88 & -0.71 & - & - & -0.04 & -0.07 \\
Paired score & -0.42 & -0.24 & 0.78 & 0.67 & 0.35 & 0.25 & -0.15 & -0.20 & 0.71 & 0.56 & 0.01 & -0.16 & -0.29 & -0.29 & -0.95 & -0.87 & 0.01 & -0.03 \\
Pooled score & -0.70 & -0.60 & 0.92 & 0.83 & 0.45 & 0.31 & -0.36 & -0.38 & 0.82 & 0.60 & 0.02 & -0.16 & -0.76 & -0.71 & -0.96 & -0.91 & -0.07 & -0.13 \\
\textbf{ACE} & -0.70 & -0.60 & 0.92 & 0.83 & 0.45 & 0.31 & -0.36 & -0.38 & 0.82 & 0.60 & 0.02 & -0.16 & -0.76 & -0.71 & -0.96 & -0.91 & -0.07 & -0.13 \\
\hline 
 \multicolumn{19}{c}{\emph{JULE}: CDbw index} \\
 \hline 
Raw score & - & - & - & - & - & - & - & - & - & - & - & - & - & - & - & - & - & - \\
Paired score & -0.20 & -0.11 & -0.23 & -0.11 & 0.59 & 0.42 & -0.13 & -0.11 & -0.55 & -0.38 & -0.15 & -0.20 & 0.24 & 0.07 & -0.81 & -0.64 & -0.16 & -0.13 \\
Pooled score & -0.65 & -0.56 & 0.87 & 0.78 & 0.54 & 0.37 & -0.31 & -0.29 & 0.90 & 0.82 & 0.06 & -0.07 & 0.43 & 0.36 & -0.53 & -0.38 & 0.16 & 0.13 \\
\textbf{ACE} & -0.64 & -0.51 & 0.92 & 0.83 & 0.54 & 0.37 & -0.31 & -0.29 & 0.89 & 0.78 & 0.06 & -0.07 & 0.48 & 0.43 & -0.52 & -0.33 & 0.18 & 0.15 \\
\hline 
 \multicolumn{19}{c}{\emph{DEPICT}: Cubic clustering criterion} \\
 \hline 
Raw score & - & - & - & - & - & - & - & - & - & - &  &  &  &  &  &  & - & - \\
Paired score & -0.44 & -0.29 & 0.99 & 0.96 & 0.90 & 0.78 & -0.16 & -0.07 & 0.98 & 0.91 &  &  &  &  &  &  & 0.45 & 0.46 \\
Pooled score & -0.54 & -0.42 & 0.99 & 0.96 & 0.38 & 0.28 & -0.07 & -0.02 & 0.92 & 0.82 &  &  &  &  &  &  & 0.34 & 0.32 \\
\textbf{ACE} & -0.54 & -0.42 & 0.99 & 0.96 & 0.38 & 0.28 & -0.25 & -0.16 & 0.92 & 0.82 &  &  &  &  &  &  & 0.30 & 0.30 \\
\hline 
 \multicolumn{19}{c}{\emph{DEPICT}: Dunn index} \\
 \hline 
Raw score & 0.19 & 0.11 & 0.85 & 0.73 & 0.48 & 0.39 & 0.25 & 0.20 & 0.20 & 0.24 &  &  &  &  &  &  & 0.39 & 0.34 \\
Paired score & 0.24 & 0.20 & -0.19 & -0.11 & -0.20 & -0.06 & 0.59 & 0.47 & 0.02 & -0.02 &  &  &  &  &  &  & 0.09 & 0.10 \\
Pooled score & 0.24 & 0.16 & -0.31 & -0.20 & 0.60 & 0.50 & 0.46 & 0.38 & 0.22 & 0.16 &  &  &  &  &  &  & 0.24 & 0.20 \\
\textbf{ACE} & -0.07 & -0.07 & 0.09 & 0.07 & 0.60 & 0.50 & 0.53 & 0.42 & 0.22 & 0.16 &  &  &  &  &  &  & 0.28 & 0.22 \\
\hline 
 \multicolumn{19}{c}{\emph{DEPICT}: Cindex} \\
 \hline 
Raw score & -0.60 & -0.47 & 0.61 & 0.38 & 0.73 & 0.50 & -0.72 & -0.56 & 0.85 & 0.73 &  &  &  &  &  &  & 0.18 & 0.12 \\
Paired score & 0.88 & 0.82 & -0.62 & -0.51 & -0.43 & -0.33 & 0.78 & 0.64 & 0.12 & 0.16 &  &  &  &  &  &  & 0.14 & 0.16 \\
Pooled score & -0.87 & -0.78 & 0.99 & 0.96 & 0.37 & 0.22 & -0.70 & -0.60 & 0.92 & 0.82 &  &  &  &  &  &  & 0.14 & 0.12 \\
\textbf{ACE} & -0.87 & -0.78 & 0.99 & 0.96 & 0.23 & 0.22 & -0.71 & -0.60 & 0.92 & 0.82 &  &  &  &  &  &  & 0.11 & 0.12 \\
\hline 
 \multicolumn{19}{c}{\emph{DEPICT}: SDbw index} \\
 \hline 
Raw score & -0.83 & -0.69 & 0.99 & 0.96 & 0.22 & 0.11 & -0.72 & -0.56 & 0.92 & 0.82 &  &  &  &  &  &  & 0.11 & 0.13 \\
Paired score & 0.85 & 0.73 & -0.38 & -0.24 & -0.37 & -0.22 & 0.85 & 0.78 & 0.26 & 0.29 &  &  &  &  &  &  & 0.24 & 0.27 \\
Pooled score & -0.85 & -0.78 & 0.99 & 0.96 & 0.37 & 0.22 & -0.71 & -0.56 & 0.92 & 0.82 &  &  &  &  &  &  & 0.14 & 0.13 \\
\textbf{ACE} & -0.85 & -0.78 & 0.99 & 0.96 & 0.37 & 0.22 & -0.74 & -0.64 & 0.93 & 0.87 &  &  &  &  &  &  & 0.14 & 0.12 \\
\hline 
 \multicolumn{19}{c}{\emph{DEPICT}: CDbw index} \\
 \hline 
Raw score & - & - & - & - & - & - & - & - & -0.04 & 0.00 &  &  &  &  &  &  & -0.04 & 0.00 \\
Paired score & 0.84 & 0.73 & -0.71 & -0.56 & -0.33 & -0.17 & 0.81 & 0.73 & -0.81 & -0.64 &  &  &  &  &  &  & -0.04 & 0.02 \\
Pooled score & -0.47 & -0.42 & -0.09 & 0.02 & -0.53 & -0.39 & -0.64 & -0.38 & 0.85 & 0.73 &  &  &  &  &  &  & -0.17 & -0.09 \\
\textbf{ACE} & -0.28 & -0.24 & 0.37 & 0.38 & -0.23 & -0.22 & -0.61 & -0.33 & 0.90 & 0.78 &  &  &  &  &  &  & 0.03 & 0.07 \\
\bottomrule
\end{tabular}
}
\label{tab:app:acc:det1}
\end{table*}

\clearpage

\subsection{Application 2 - detected number of clusters}\label{app:exp:num}
We present the optimal number of clusters (\( K \)) identified by each evaluation approach across various datasets used in Application 2. Results for JULE are illustrated in Figure \ref{app:fig:ncl1}, while those for DEPICT are displayed in Figure \ref{app:fig:ncl2}. The ground truth \( K \) is represented by a red, outlined, hollow box, whereas the detected \( K \) values are shown as solid boxes with hatches. Each scoring approach is distinguished by a unique color.

\begin{figure}[htbp!]
\centering
\subfigure[\scriptsize  COIL-20]{\includegraphics[width = 0.35\linewidth]{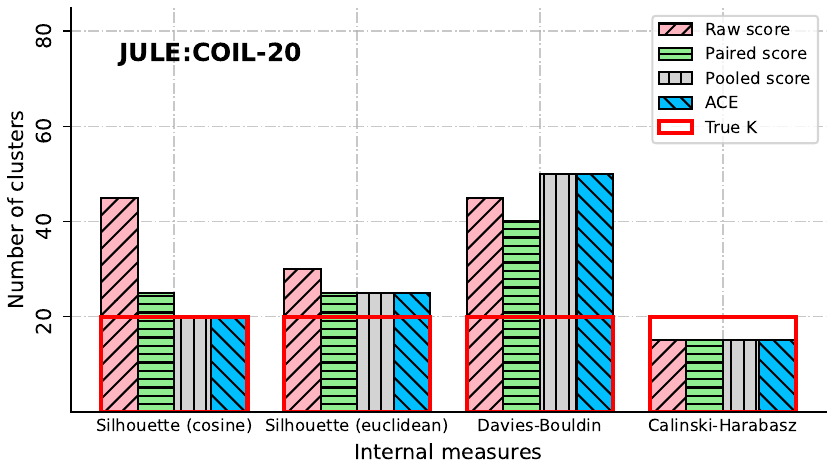}}
\subfigure[\scriptsize  COIL-100]{\includegraphics[width = 0.35\linewidth]{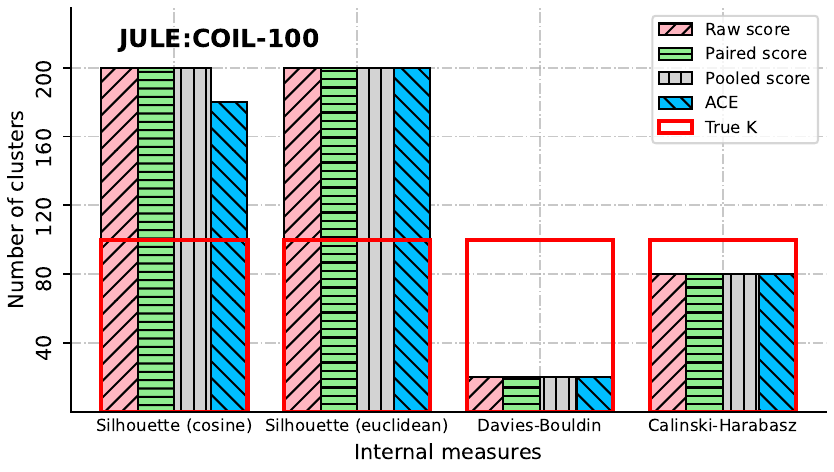}}\\
\subfigure[\scriptsize  CMU-PIE]{\includegraphics[width = 0.35\linewidth]{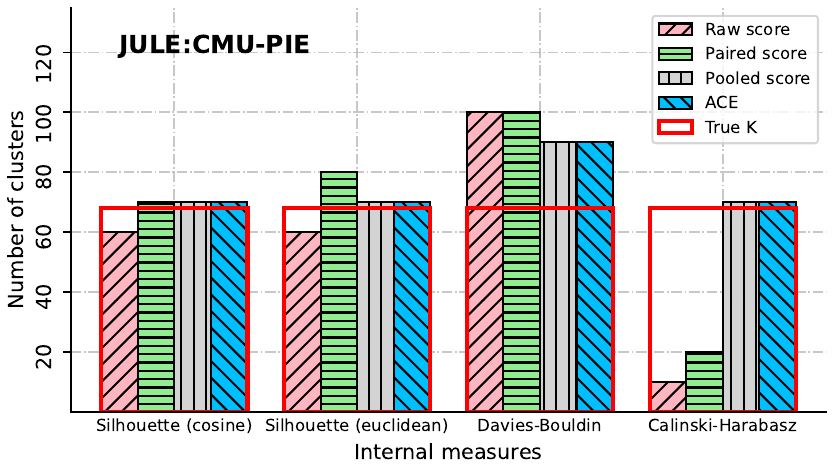}}
\subfigure[\scriptsize  USPS]{\includegraphics[width = 0.35\linewidth]{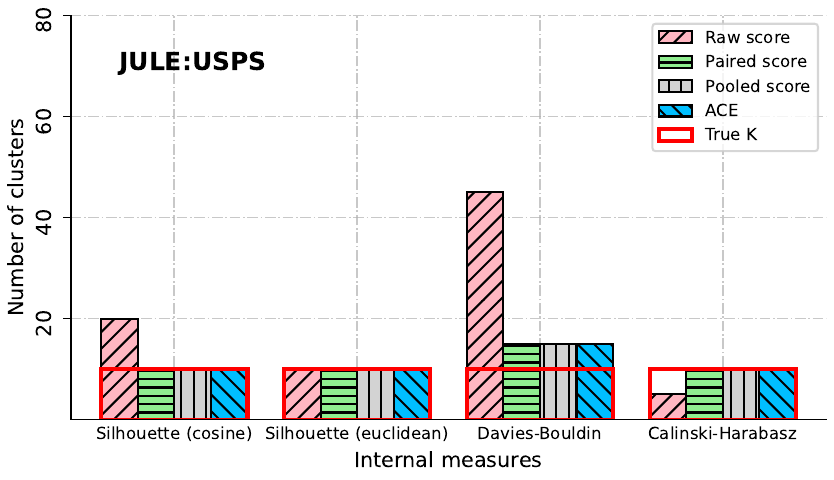}}\\
\subfigure[\scriptsize  MNIST-test]{\includegraphics[width = 0.35\linewidth]{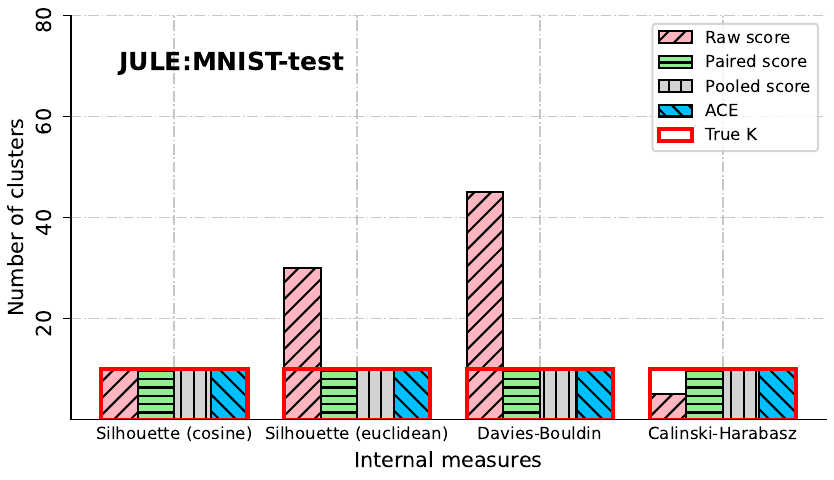}}
\subfigure[\scriptsize  UMist]{\includegraphics[width = 0.35\linewidth]{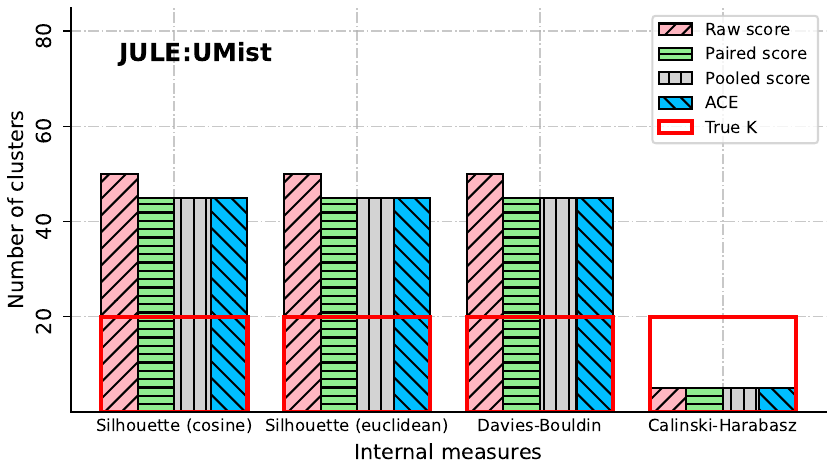}} \\
\subfigure[\scriptsize  YTF]{\includegraphics[width = 0.35\linewidth]{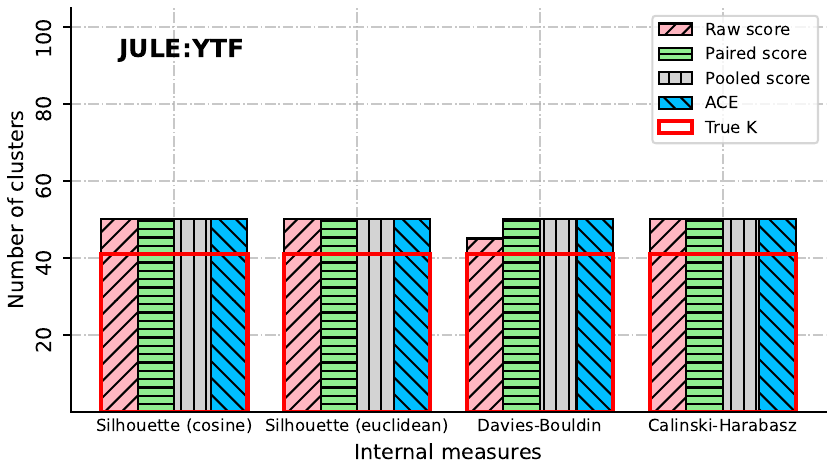}}
\subfigure[\scriptsize  FRGC]{\includegraphics[width = 0.35\linewidth]{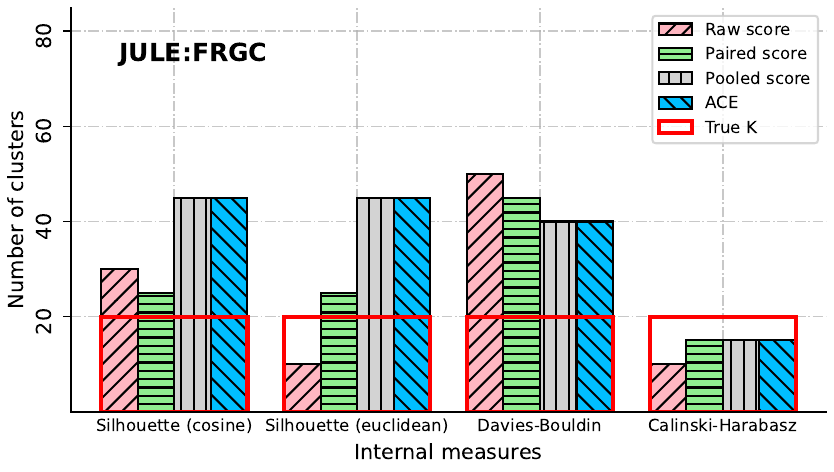}} 
\caption{The optimal \(K\) detected by each scoring approach for JULE is presented using bar plots, with the true \(K\) indicated by a red, outlined, hollow box.}
\label{app:fig:ncl1}
\end{figure}
\begin{figure}[htbp!]
\centering
\subfigure[\scriptsize  YTF]{\includegraphics[width = 0.4\linewidth]{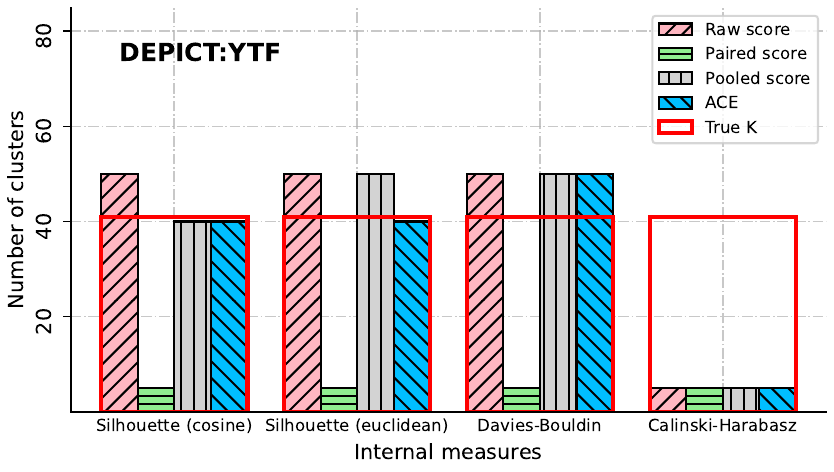}}
\subfigure[\scriptsize  CMU-PIE]{\includegraphics[width = 0.4\linewidth]{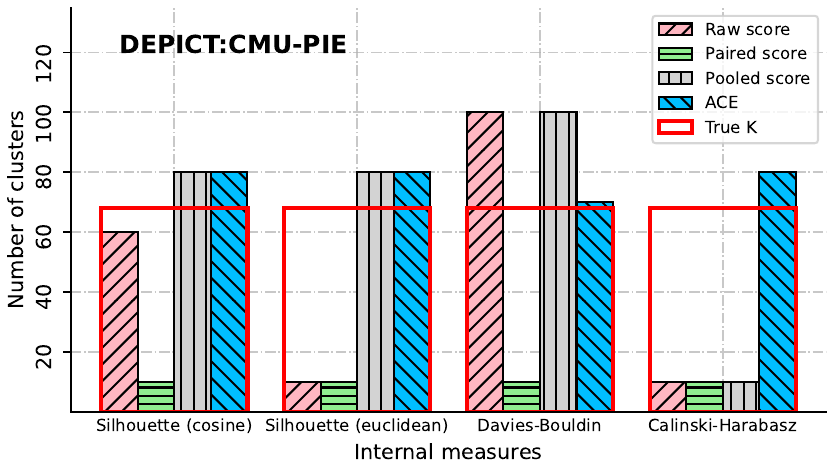}} \\
\subfigure[\scriptsize  USPS]{\includegraphics[width = 0.4\linewidth]{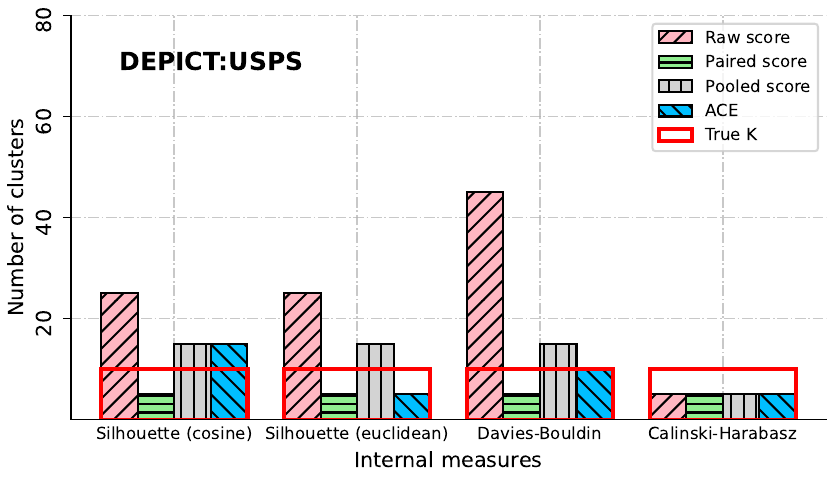}}
\subfigure[\scriptsize  MNIST-Test]{\includegraphics[width = 0.4\linewidth]{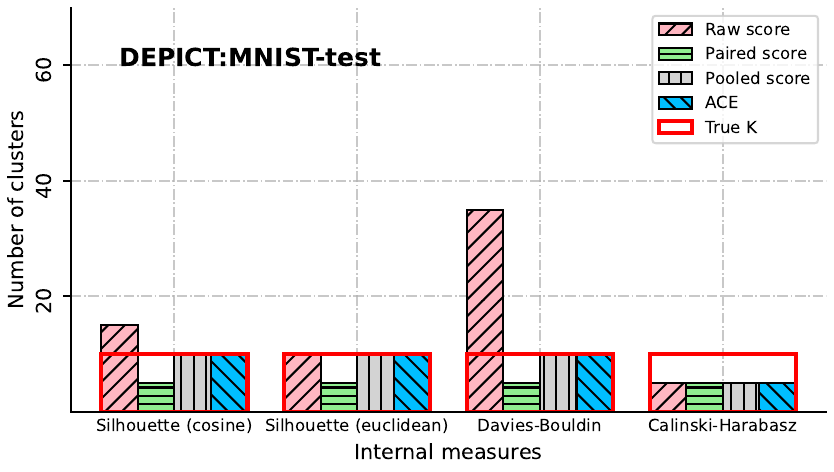}}\\
\subfigure[\scriptsize  FRGC]{\includegraphics[width = 0.4\linewidth]{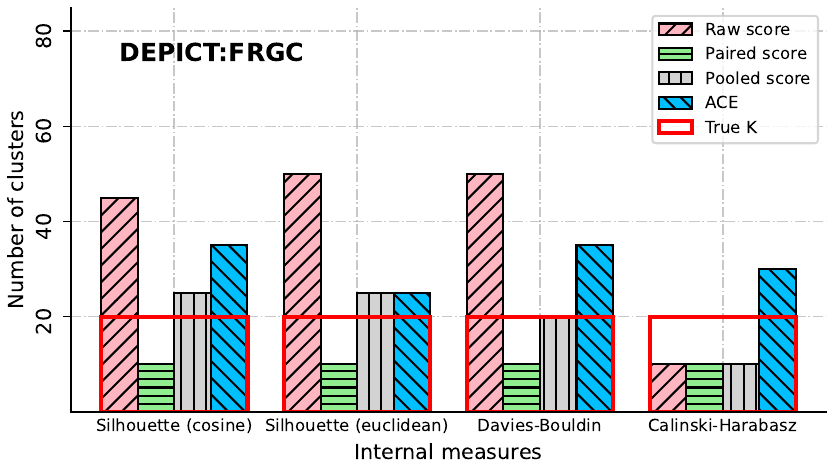}}
\caption{The optimal \(K\) detected by each scoring approach for DEPICT is presented using bar plots, with the true \(K\) indicated by a red, outlined, hollow box.}
\label{app:fig:ncl2}
\end{figure}

\newpage

\newpage \subsection{Application 2 - qualitative analysis} \label{app:exp:ar:qa2}
In this section, we present qualitative analysis results for Application 2 using both JULE and DEPICT. Graphs illustrating the rank correlation between the retained spaces after the multimodality test are provided for JULE in Figure \ref{fig:graph:dav:2} (Davies-Bouldin index), Figure \ref{fig:graph:ch:2} (Calinski-Harabasz index), Figure \ref{fig:graph:cosine:2} (Silhouette score with Cosine distance), and Figure \ref{fig:graph:euclidean:2} (Silhouette score with Euclidean distance). For DEPICT, the corresponding graphs are shown in Figure \ref{fig:graph:dav:3} (Davies-Bouldin index), Figure \ref{fig:graph:ch:3} (Calinski-Harabasz index), Figure \ref{fig:graph:cosine:3} (Silhouette score with Cosine distance), and Figure \ref{fig:graph:euclidean:3} (Silhouette score with Euclidean distance). In each graph, spaces grouped together based on their rank correlations share the same color, while outlier spaces are uniformly colored in grey. Similar to the observations from Application 1, we find that grouping behavior varies depending on the chosen internal measure. However, in this task, where approximately 10 spaces are generated, we observe a greater tendency for a single dominant group to form. This suggests that the grouping behavior of embedding spaces is also influenced by the number of spaces included in the comparison.

Similar to Application 1, we use t-SNE plots \citep{van2008visualizing} to visually evaluate the discriminative capability of embedding spaces selected by ACE compared to those excluded. For JULE, we present comparisons based on different internal measures in Figure \ref{fig:tsne:dav:2} (Davies-Bouldin index), Figure \ref{fig:tsne:ch:2} (Calinski-Harabasz index), Figure \ref{fig:tsne:cosine:2} (Silhouette score with Cosine distance), and Figure \ref{fig:tsne:euclidean:2} (Silhouette score with Euclidean distance). Similarly, for DEPICT, the comparisons are shown in Figure \ref{fig:tsne:dav:3} (Davies-Bouldin index), Figure \ref{fig:tsne:ch:3} (Calinski-Harabasz index), Figure \ref{fig:tsne:cosine:3} (Silhouette score with Cosine distance), and Figure \ref{fig:tsne:euclidean:3} (Silhouette score with Euclidean distance). Due to space limitations, we present a single selected space and a single excluded space for each dataset. In each subfigure, different colors represent distinct true clusters. If a subfigure for an excluded space is missing, it indicates that all spaces were selected by ACE, which can occur when the number of clustering spaces is small. We observe that selected spaces exhibit better clustering behavior than excluded spaces, though the difference is less pronounced than in Application 1. In some cases, all spaces are selected with none excluded, which may occur when the number of available spaces for comparison is limited (e.g., $\sim 10$ in this experiment).\\

%T-SNE, recognized for its ability to preserve local structure and relative distances in high-dimensional space, is utilized to project the embedding space into a 2-dimensional feature space for visualization. 

\begin{figure}[htbp!]
\centering
\subfigure[USPS]{\includegraphics[width = 0.3\linewidth]{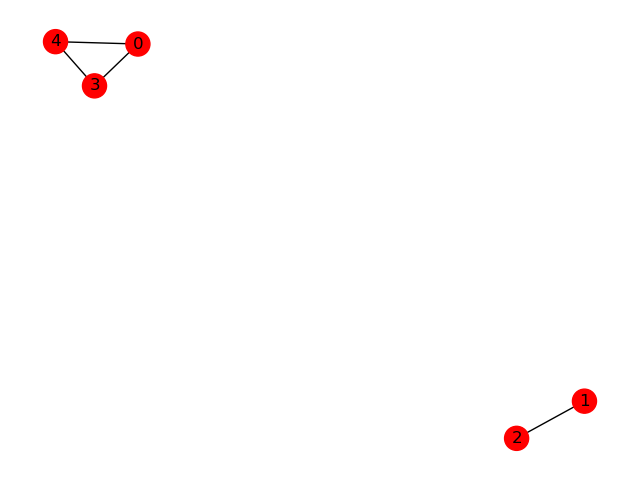}}
\subfigure[UMist]{\includegraphics[width = 0.3\linewidth]{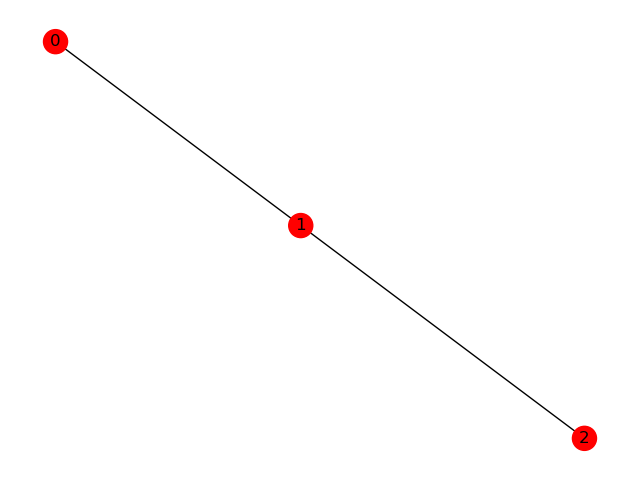}}
\subfigure[COIL-20]{\includegraphics[width = 0.3\linewidth]{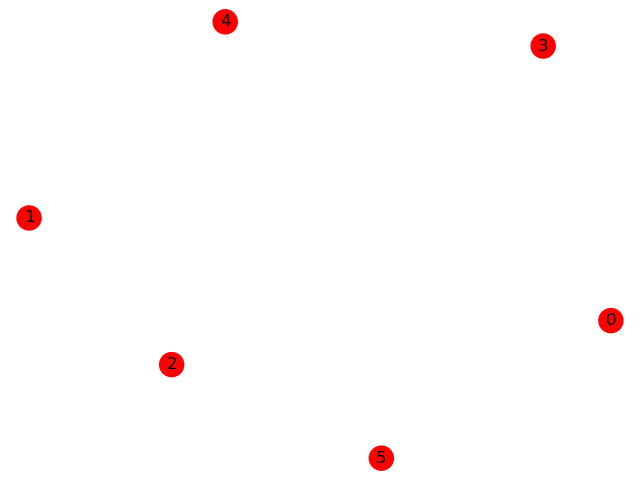}}
\subfigure[COIL-100]{\includegraphics[width = 0.3\linewidth]{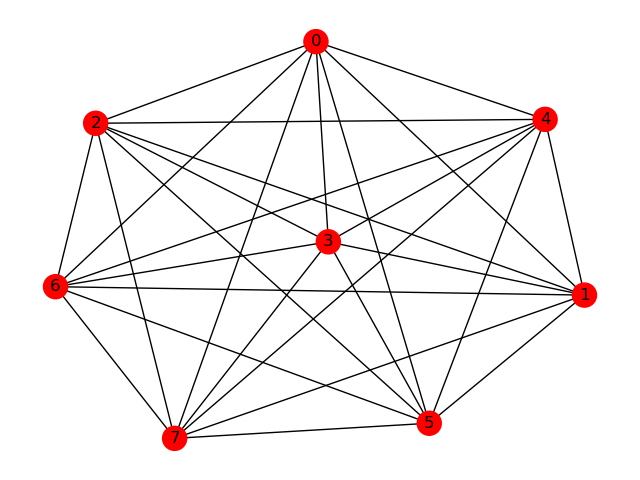}}
\subfigure[YTF]{\includegraphics[width = 0.3\linewidth]{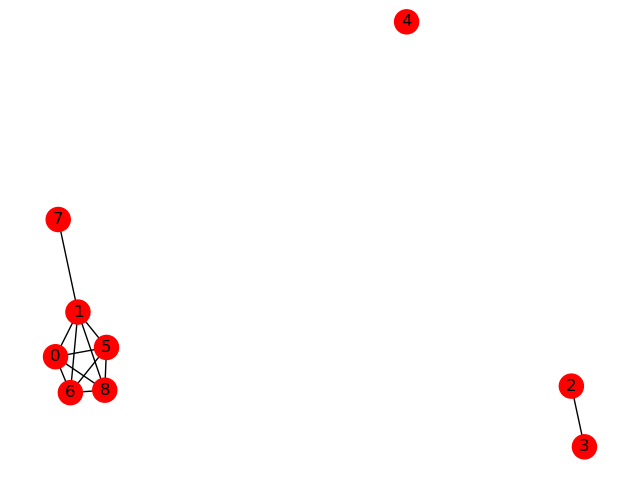}}
\subfigure[FRGC]{\includegraphics[width = 0.3\linewidth]{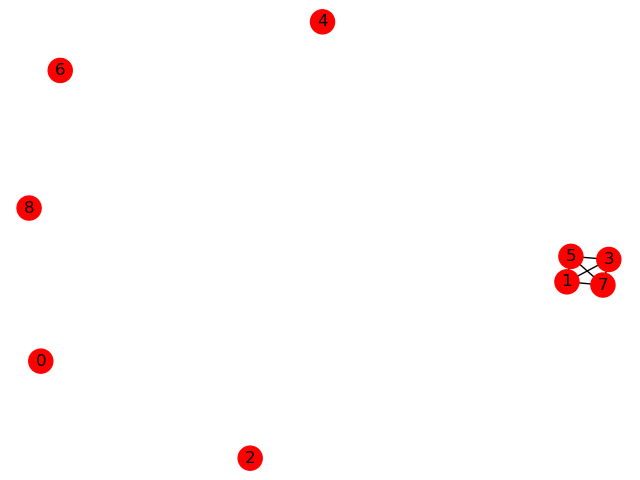}}
\subfigure[MNIST-test]{\includegraphics[width = 0.3\linewidth]{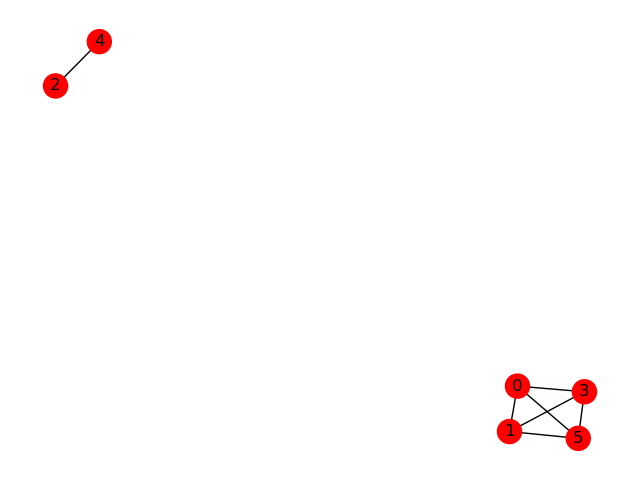}}
\subfigure[CMU-PIE]{\includegraphics[width = 0.3\linewidth]{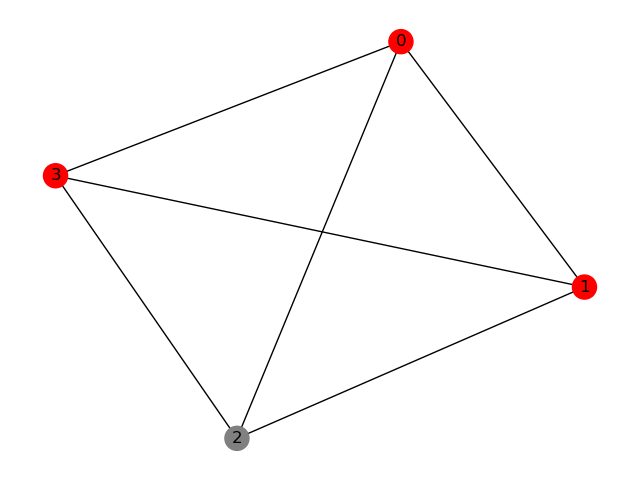}}
\caption{Graph depicting rank correlation based on Davies-Bouldin index among embedding spaces for Application 2 with JULE. Each node represents an embedding space, and each edge signifies a significant rank correlation.}
\label{fig:graph:dav:2}
\end{figure}
%%%%%%%%%%%%%%%%%%%%%%%%%
\begin{figure}[htbp!]
\centering
\subfigure[Selected space (USPS)]{\includegraphics[width = 0.24\linewidth]{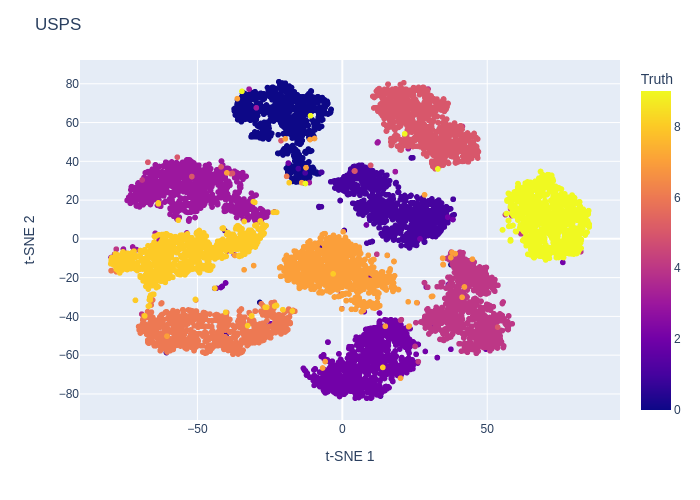}}
\subfigure[Excluded space (USPS)]{\includegraphics[width = 0.24\linewidth]{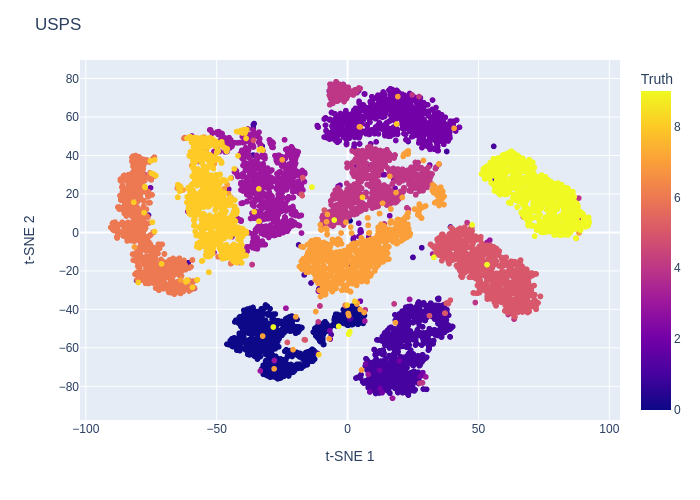}}
\subfigure[Selected space (UMist)]{\includegraphics[width = 0.24\linewidth]{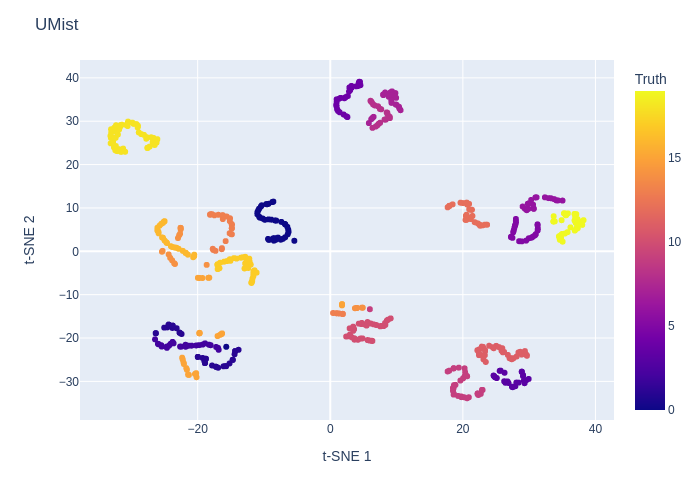}}
\subfigure[Excluded space (UMist)]{\includegraphics[width = 0.24\linewidth]{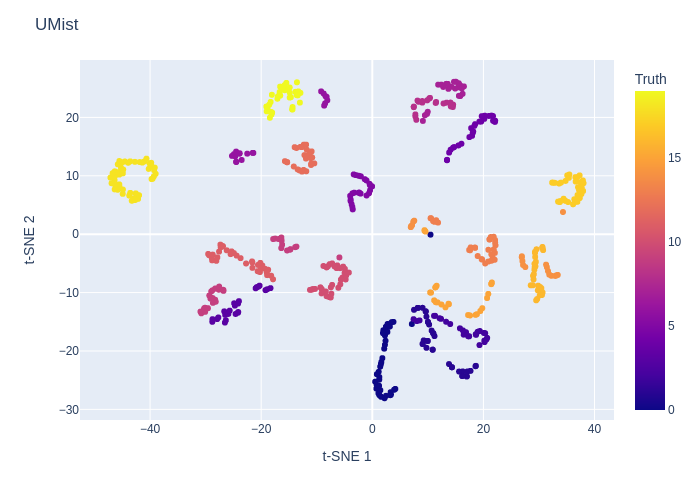}}
\subfigure[Selected space (COIL-20)]{\includegraphics[width = 0.24\linewidth]{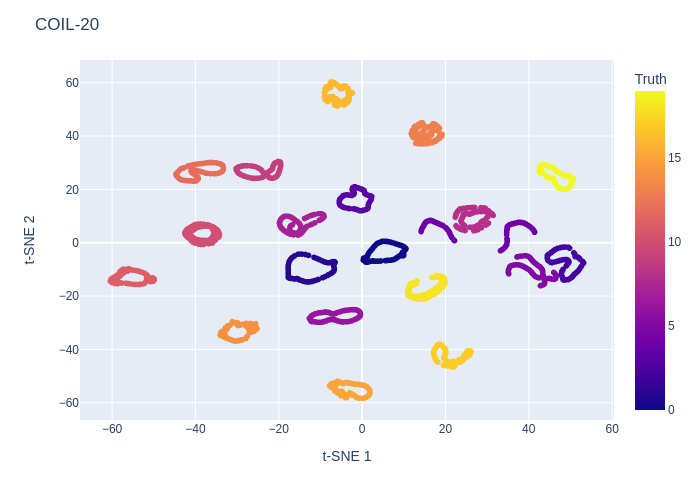}}
\subfigure[Excluded space (COIL-20)]{\includegraphics[width = 0.24\linewidth]{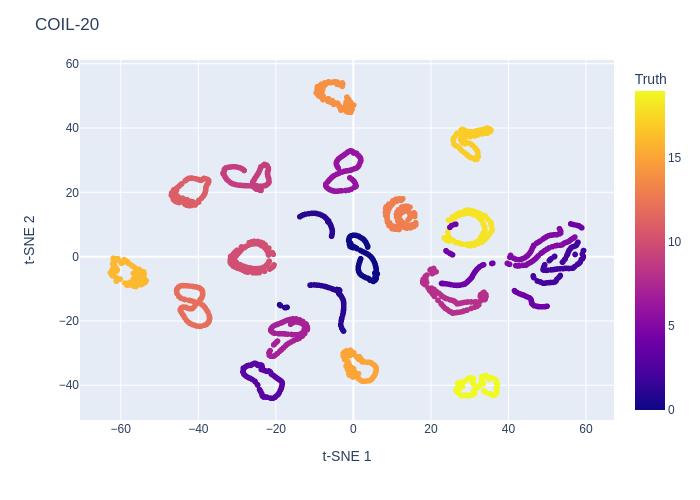}}
\subfigure[Selected space (COIL-100)]{\includegraphics[width = 0.24\linewidth]{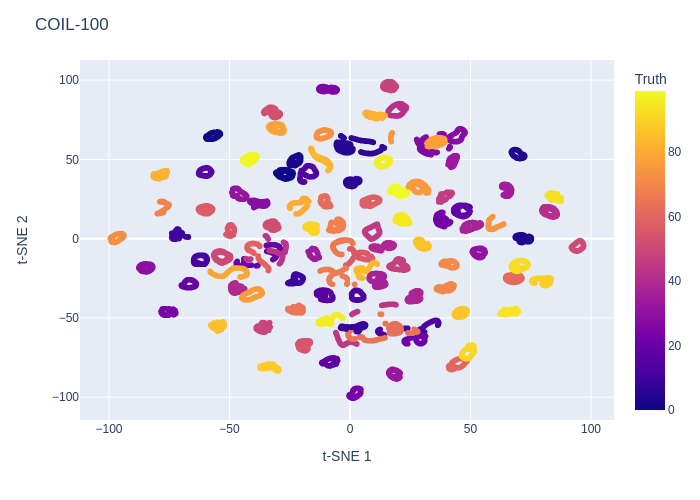}}
\subfigure[Excluded space (COIL-100)]{\includegraphics[width = 0.24\linewidth]{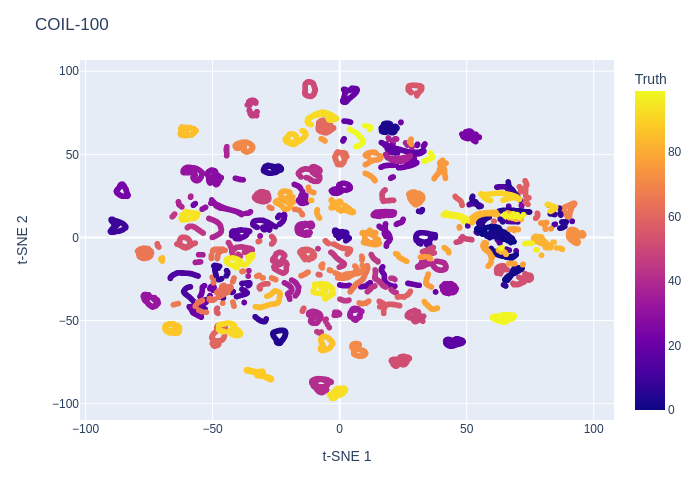}}
\subfigure[Selected space (MNIST-test)]{\includegraphics[width = 0.24\linewidth]{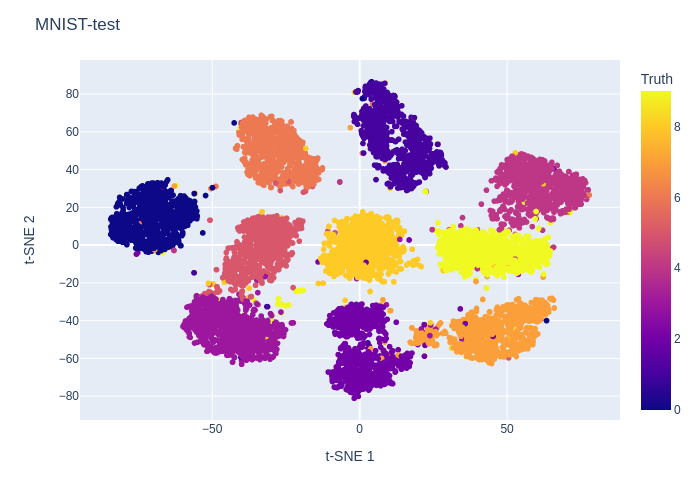}}
\subfigure[Excluded space (MNIST-test)]{\includegraphics[width = 0.24\linewidth]{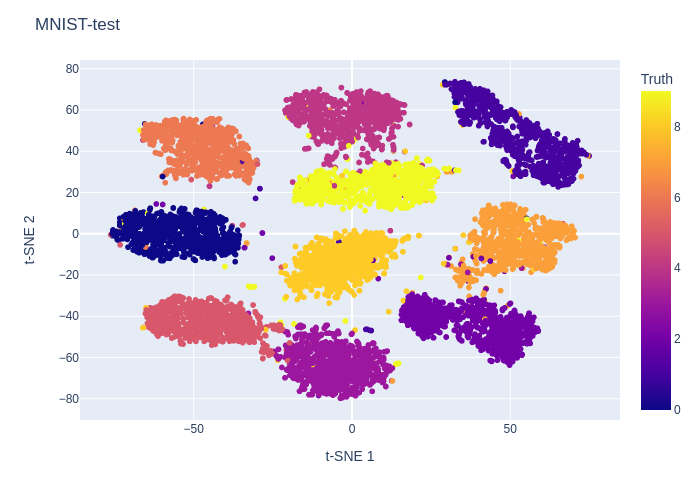}}
\subfigure[Selected space (CMU-PIE)]{\includegraphics[width = 0.24\linewidth]{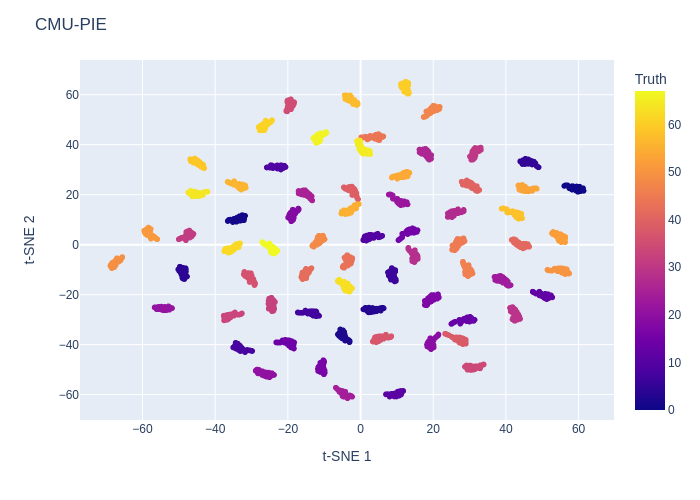}}
\subfigure[Excluded space (CMU-PIE)]{\includegraphics[width = 0.24\linewidth]{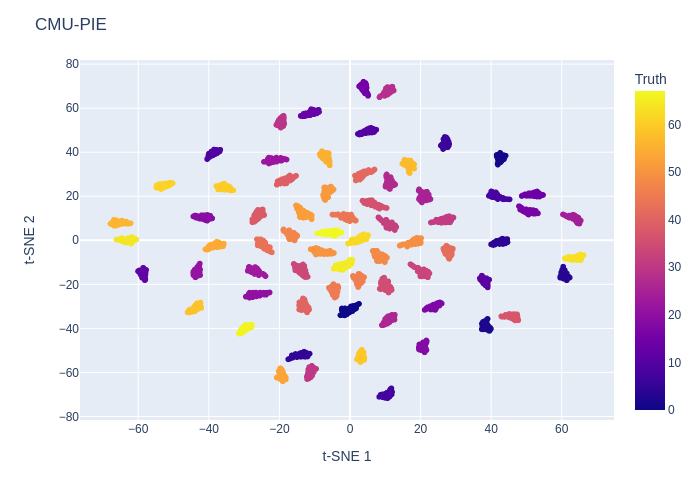}}
\subfigure[Selected space (YTF)]{\includegraphics[width = 0.24\linewidth]{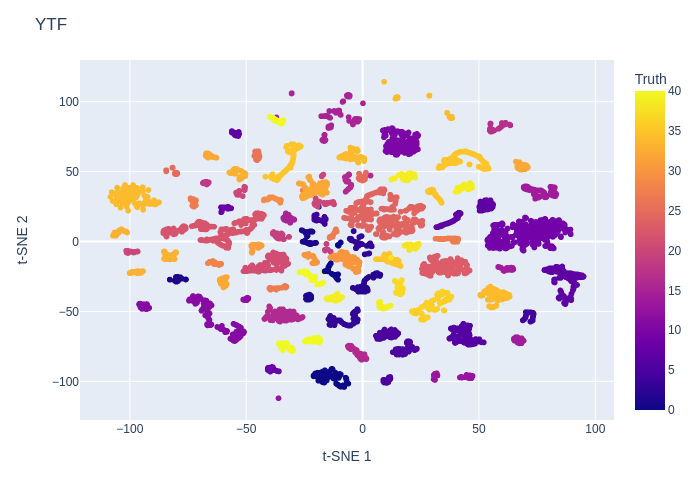}}
\subfigure[Selected space (FRGC)]{\includegraphics[width = 0.24\linewidth]{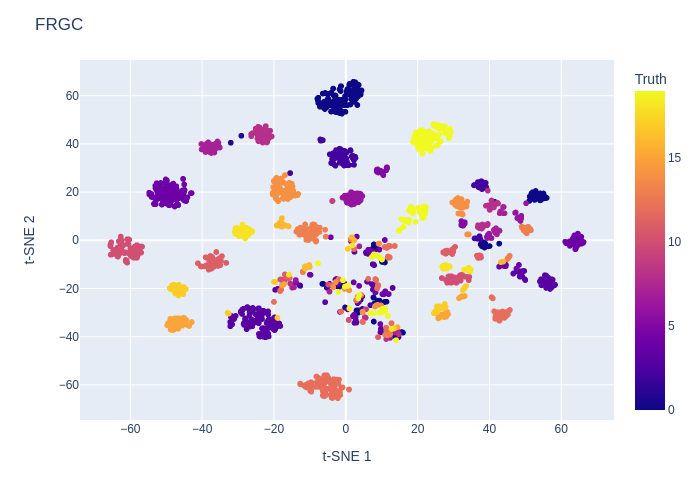}}
\caption{t-SNE visualization illustrating the selected embedding spaces from ACE in comparison to those excluded from ACE, based on Davies-Bouldin index, for Application 2 with JULE. Each data point in the visualizations is assigned a color corresponding to its true cluster label.}
\label{fig:tsne:dav:2}
\end{figure}
%%%%%%%%%%%%%%%%%%%%%%%%%
\begin{figure}[htbp!]
\centering
\subfigure[USPS]{\includegraphics[width = 0.3\linewidth]{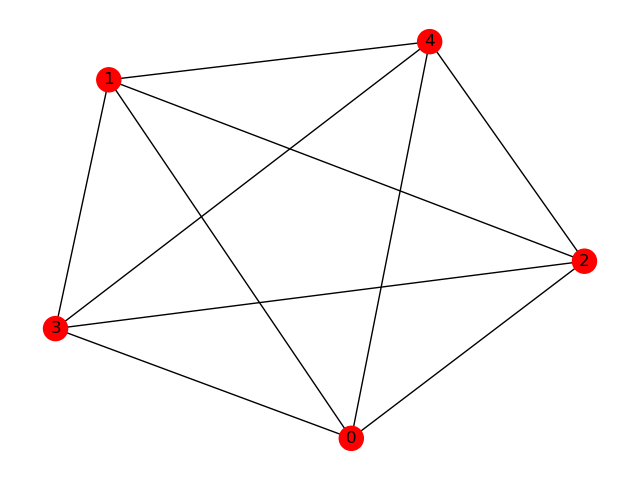}}
\subfigure[UMist]{\includegraphics[width = 0.3\linewidth]{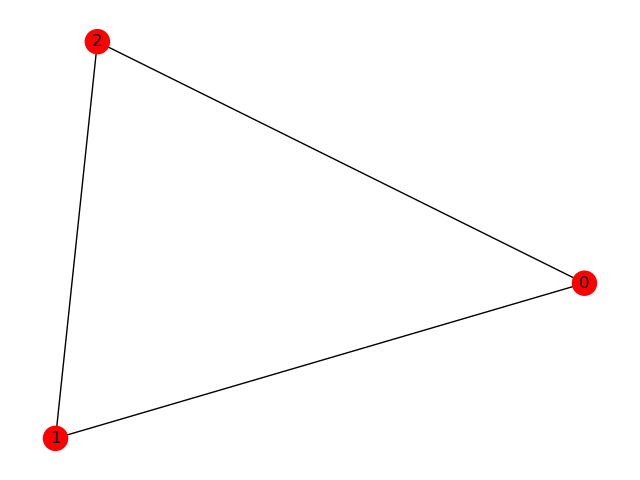}}
\subfigure[COIL-20]{\includegraphics[width = 0.3\linewidth]{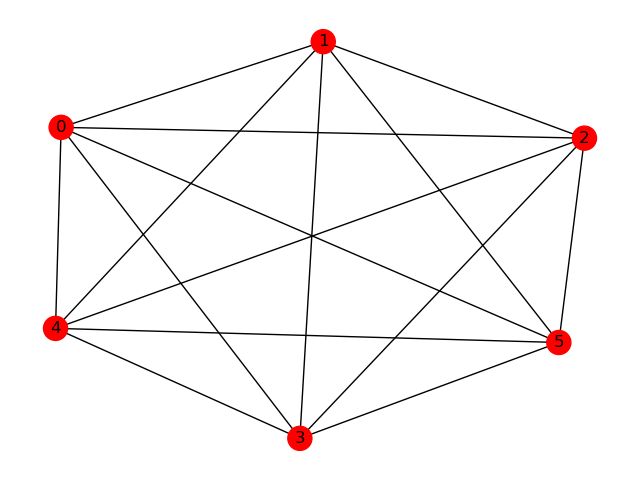}}
\subfigure[COIL-100]{\includegraphics[width = 0.3\linewidth]{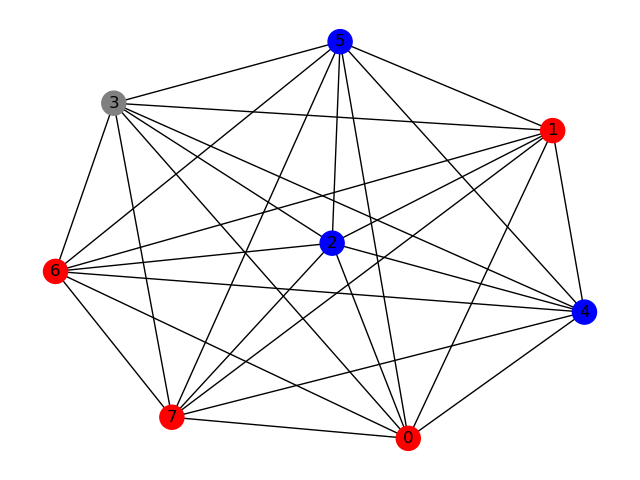}}
\subfigure[YTF]{\includegraphics[width = 0.3\linewidth]{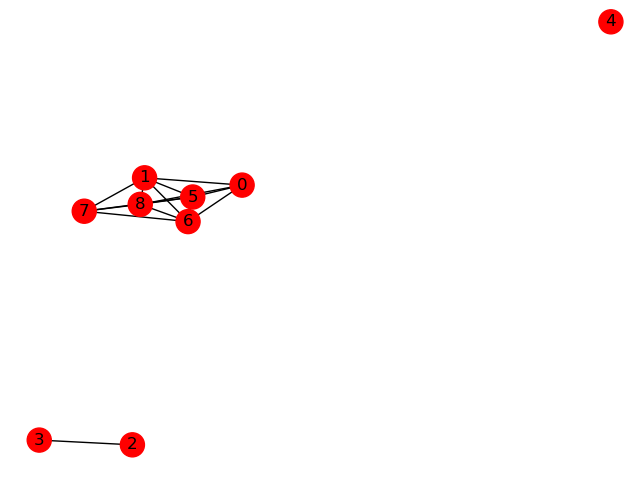}}
\subfigure[FRGC]{\includegraphics[width = 0.3\linewidth]{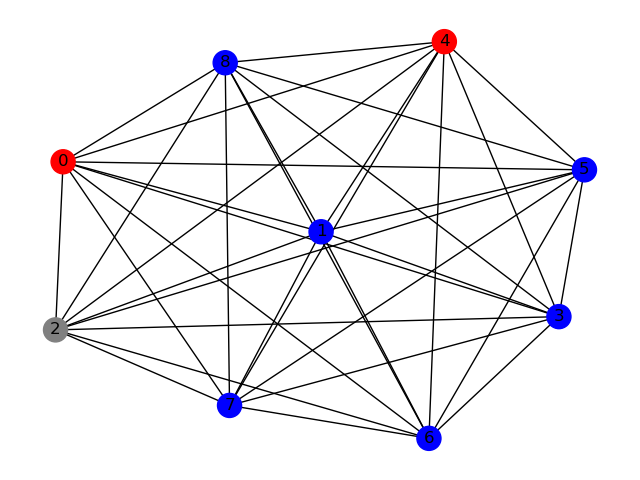}}
\subfigure[MNIST-test]{\includegraphics[width = 0.3\linewidth]{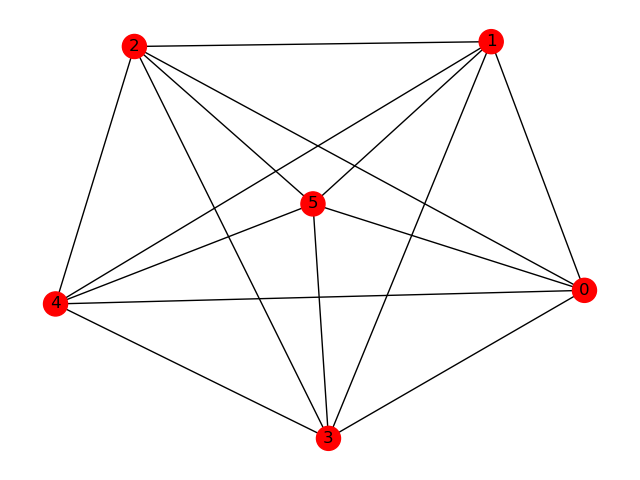}}
\subfigure[CMU-PIE]{\includegraphics[width = 0.3\linewidth]{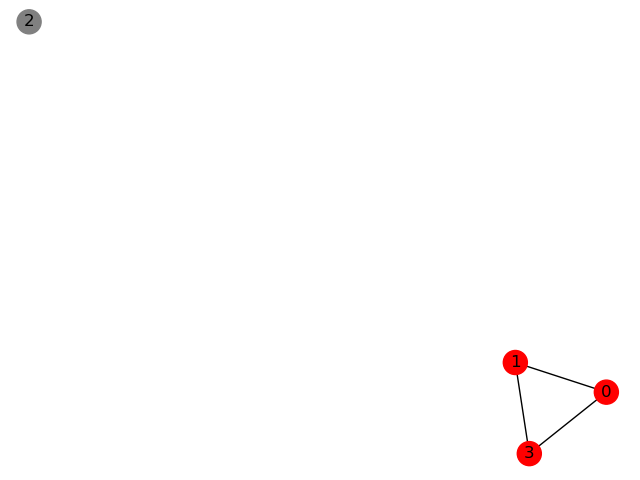}}
\caption{Graph depicting rank correlation based on Calinski-Harabasz index among embedding spaces for Application 2 with JULE. Each node represents an embedding space, and each edge signifies a significant rank correlation.}
\label{fig:graph:ch:2}
\end{figure}
%%%%%%%%%%%%%%%%%%%%%%%%%
\begin{figure}[htbp!]
\centering
\subfigure[Selected space (USPS)]{\includegraphics[width = 0.24\linewidth]{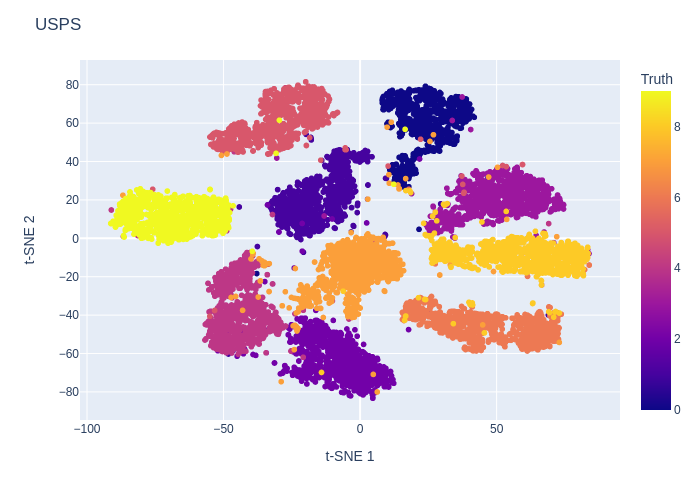}}
\subfigure[Excluded space (USPS)]{\includegraphics[width = 0.24\linewidth]{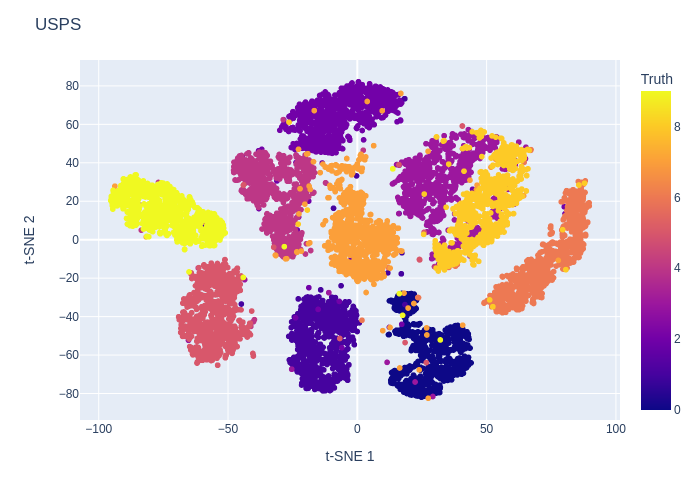}}
\subfigure[Selected space (UMist)]{\includegraphics[width = 0.24\linewidth]{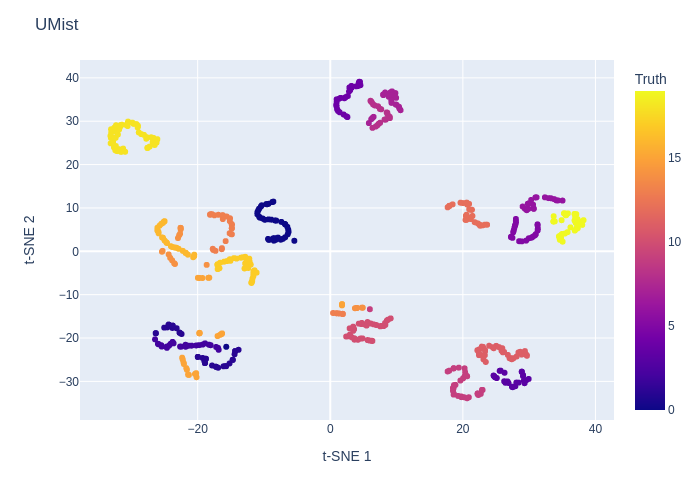}}
\subfigure[Excluded space (UMist)]{\includegraphics[width = 0.24\linewidth]{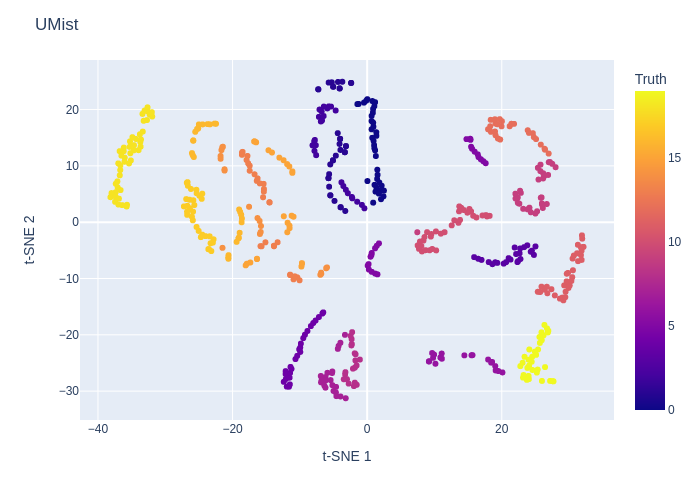}}
\subfigure[Selected space (COIL-20)]{\includegraphics[width = 0.24\linewidth]{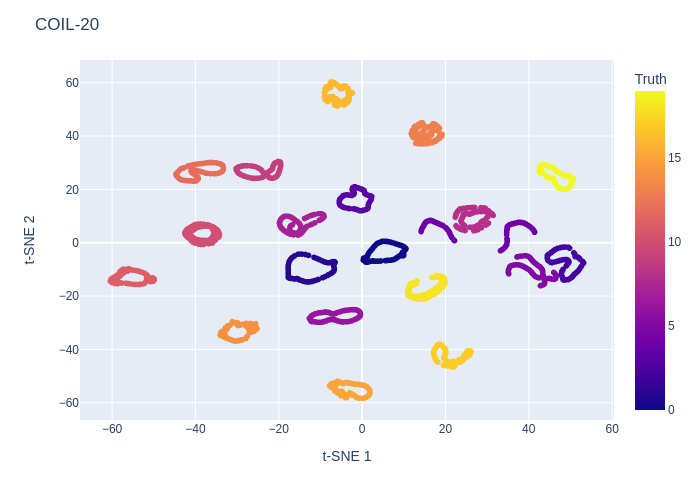}}
\subfigure[Excluded space (COIL-20)]{\includegraphics[width = 0.24\linewidth]{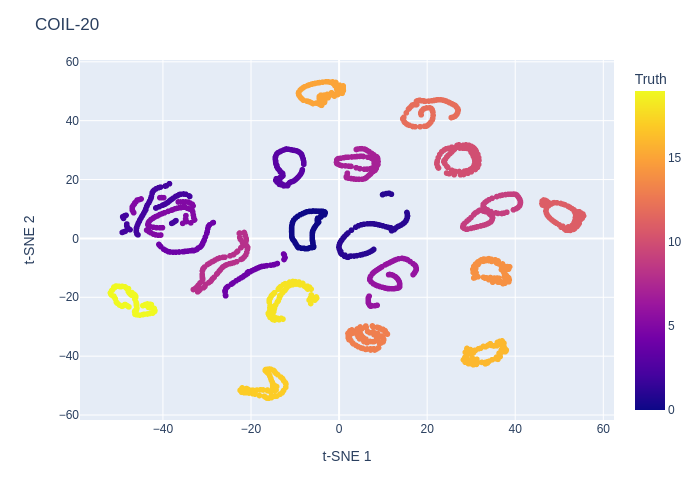}}
\subfigure[Selected space (COIL-100)]{\includegraphics[width = 0.24\linewidth]{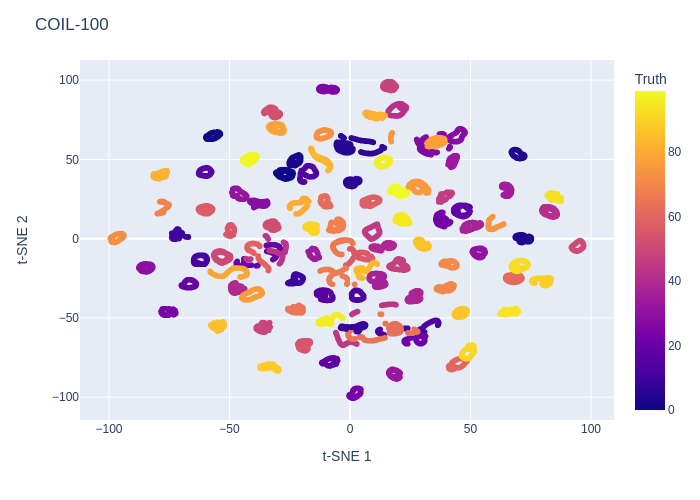}}
\subfigure[Excluded space (COIL-100)]{\includegraphics[width = 0.24\linewidth]{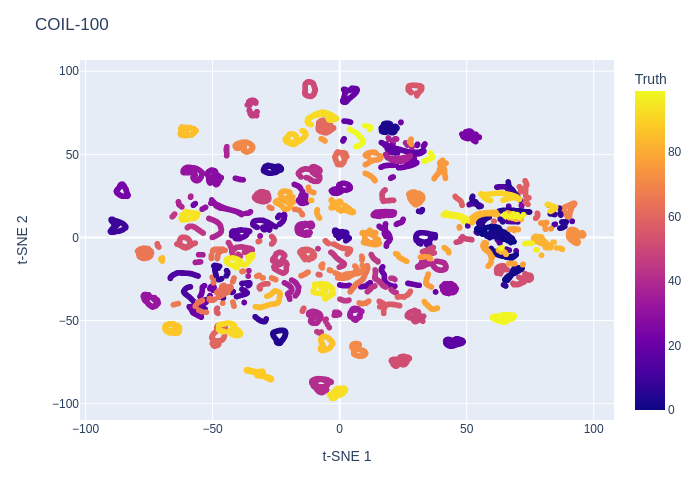}}
\subfigure[Selected space (FRGC)]{\includegraphics[width = 0.24\linewidth]{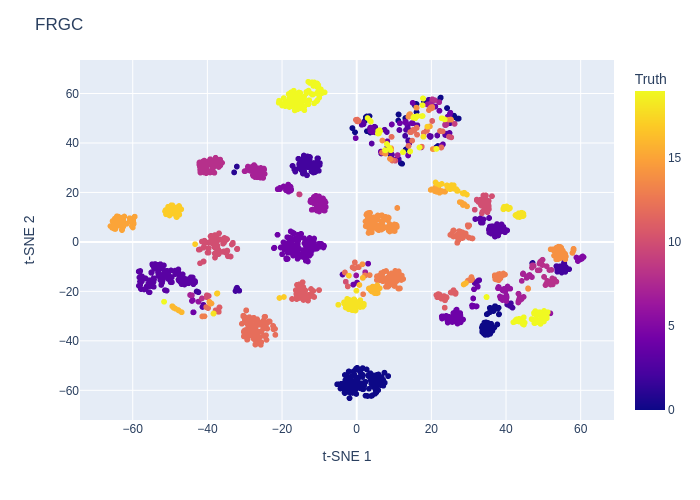}}
\subfigure[Excluded space (FRGC)]{\includegraphics[width = 0.24\linewidth]{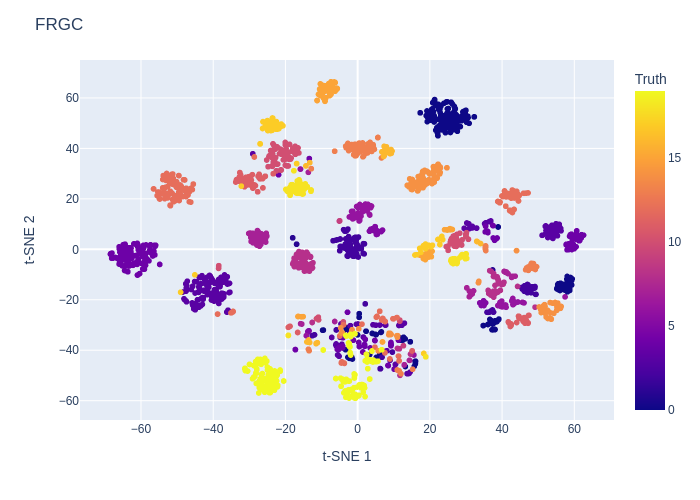}}
\subfigure[Selected space (MNIST-test)]{\includegraphics[width = 0.24\linewidth]{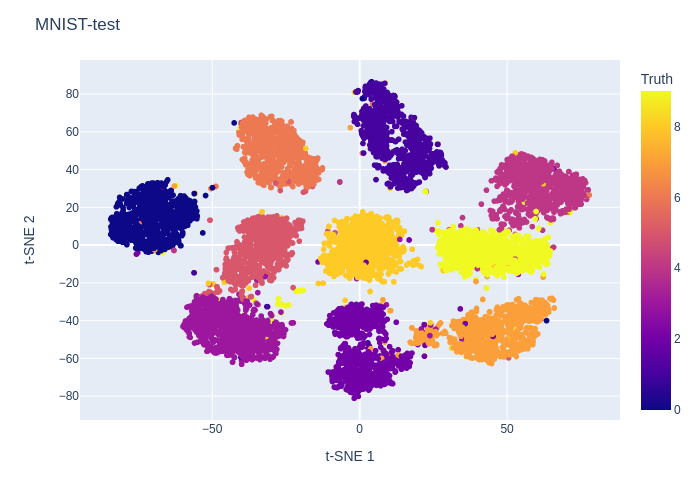}}
\subfigure[Excluded space (MNIST-test)]{\includegraphics[width = 0.24\linewidth]{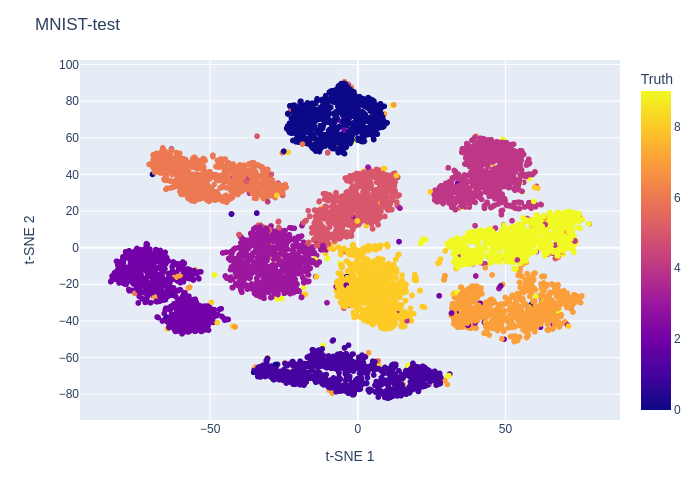}}
\subfigure[Selected space (CMU-PIE)]{\includegraphics[width = 0.24\linewidth]{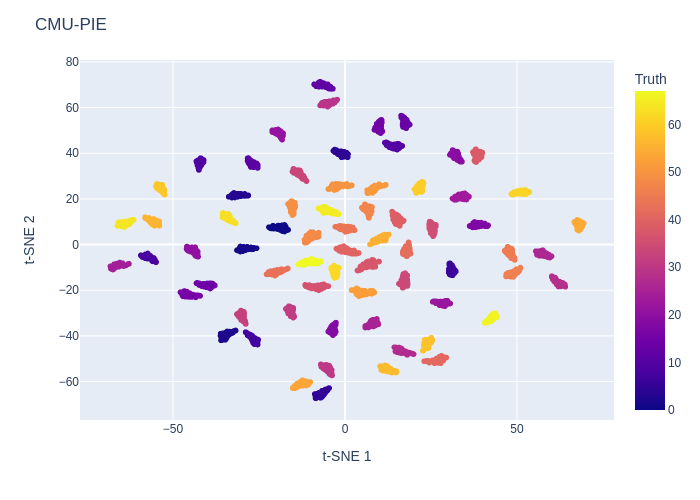}}
\subfigure[Excluded space (CMU-PIE)]{\includegraphics[width = 0.24\linewidth]{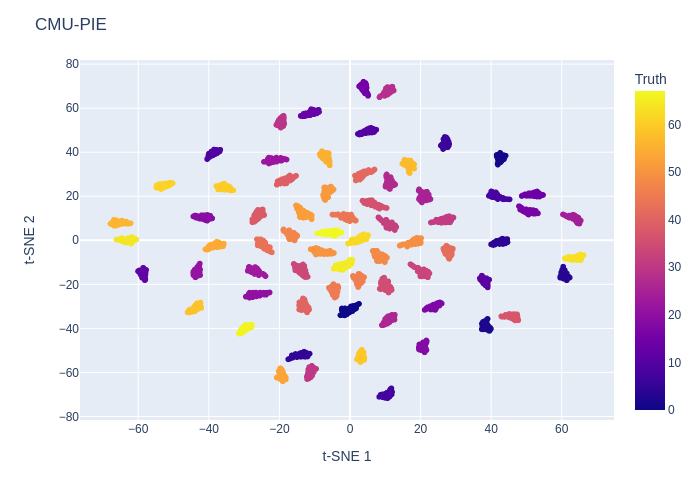}}
\subfigure[Selected space (YTF)]{\includegraphics[width = 0.24\linewidth]{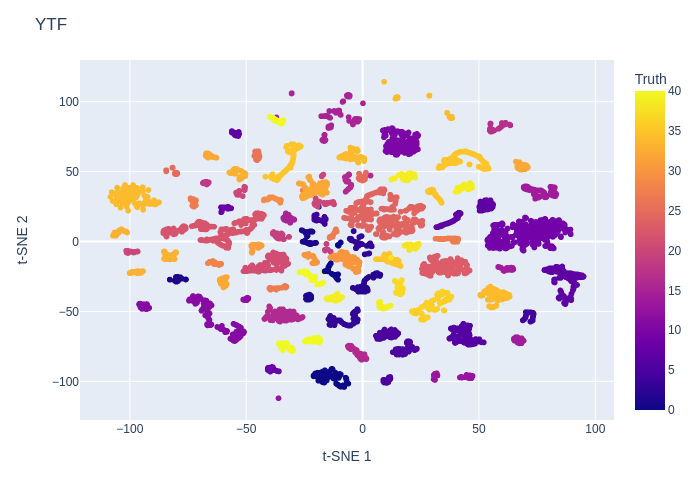}}
\caption{t-SNE visualization illustrating the selected embedding spaces from ACE in comparison to those excluded from ACE, based on Calinski-Harabasz index, for Application 2 with JULE. Each data point in the visualizations is assigned a color corresponding to its true cluster label.}
\label{fig:tsne:ch:2}
\end{figure}
%%%%%%%%%%%%%%%%%%%%%%%%%
\begin{figure}[htbp!]
\centering
\subfigure[USPS]{\includegraphics[width = 0.3\linewidth]{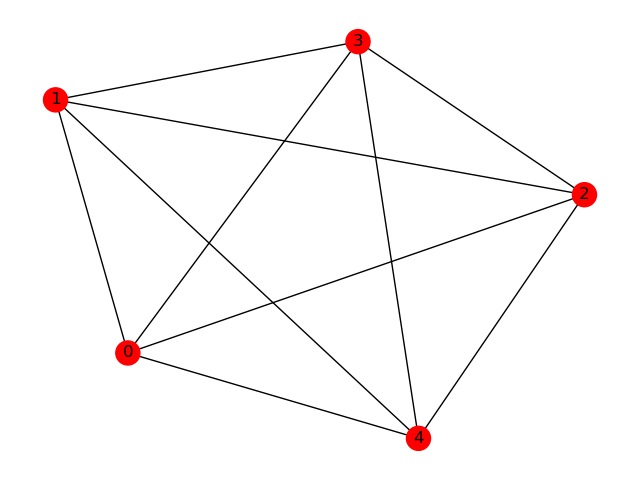}}
\subfigure[UMist]{\includegraphics[width = 0.3\linewidth]{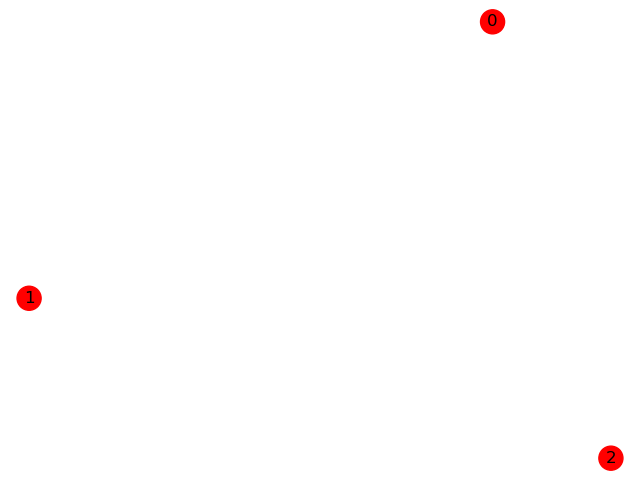}}
\subfigure[COIL-20]{\includegraphics[width = 0.3\linewidth]{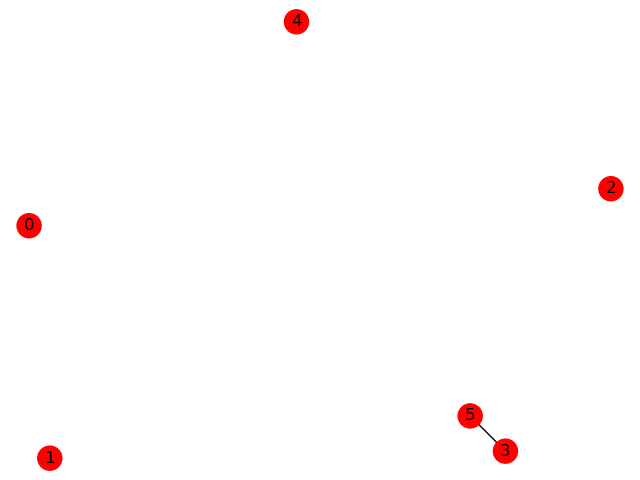}}
\subfigure[COIL-100]{\includegraphics[width = 0.3\linewidth]{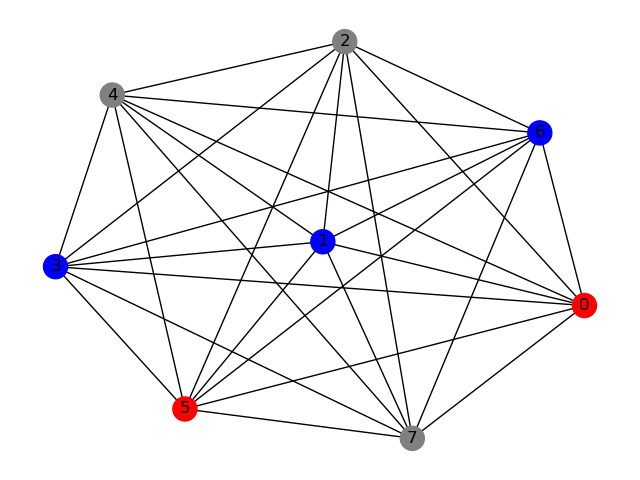}}
\subfigure[YTF]{\includegraphics[width = 0.3\linewidth]{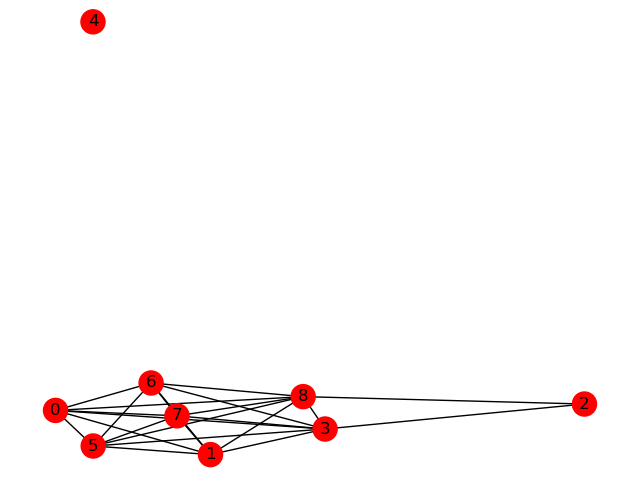}}
\subfigure[FRGC]{\includegraphics[width = 0.3\linewidth]{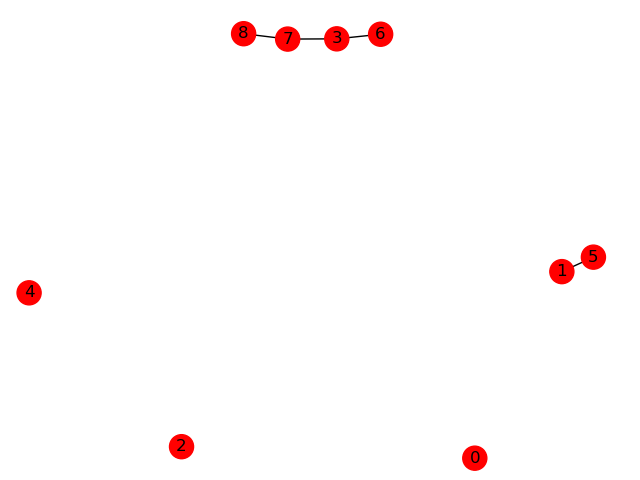}}
\subfigure[MNIST-test]{\includegraphics[width = 0.3\linewidth]{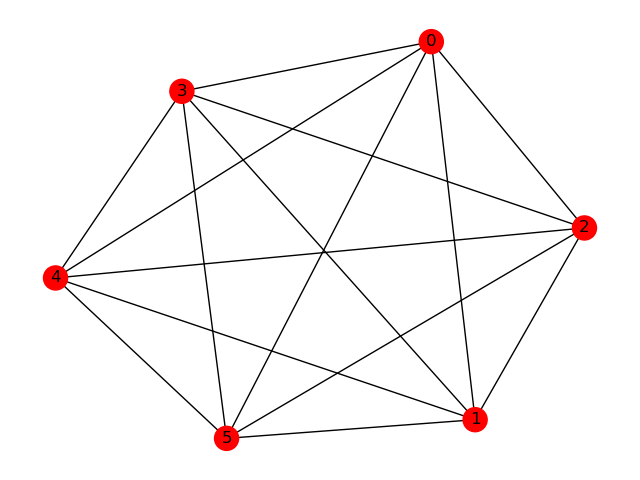}}
\subfigure[CMU-PIE]{\includegraphics[width = 0.3\linewidth]{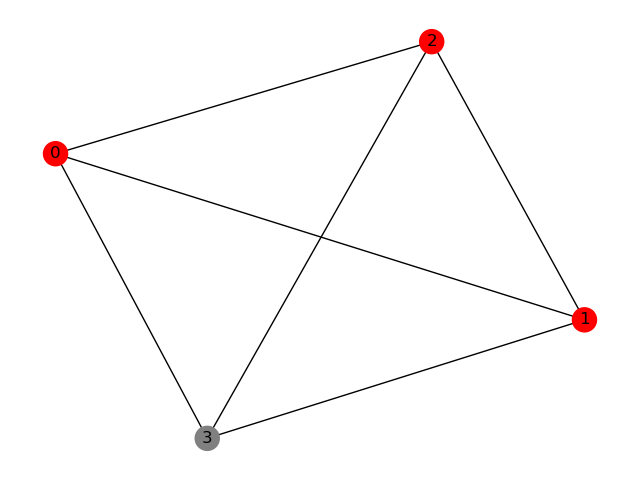}}
\caption{Graph depicting rank correlation based on Silhouette score (Cosine distance) among embedding spaces for Application 2 with JULE. Each node represents an embedding space, and each edge signifies a significant rank correlation.}
\label{fig:graph:cosine:2}
\end{figure}
%%%%%%%%%%%%%%%%%%%%%%%%%
\begin{figure}[htbp!]
\centering
\subfigure[Selected space (USPS)]{\includegraphics[width = 0.24\linewidth]{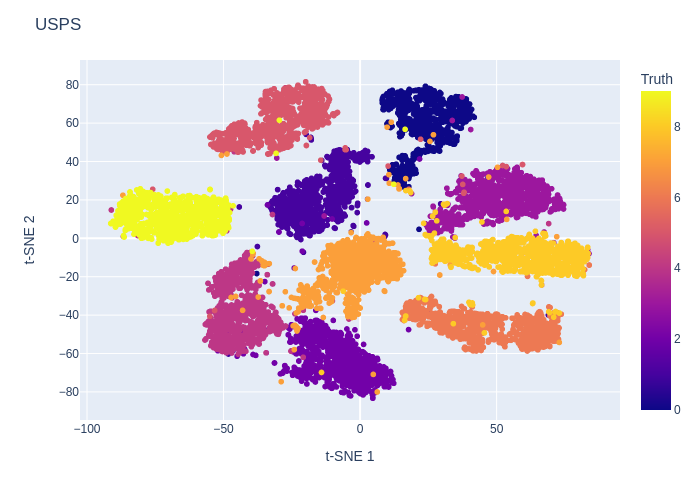}}
\subfigure[Excluded space (USPS)]{\includegraphics[width = 0.24\linewidth]{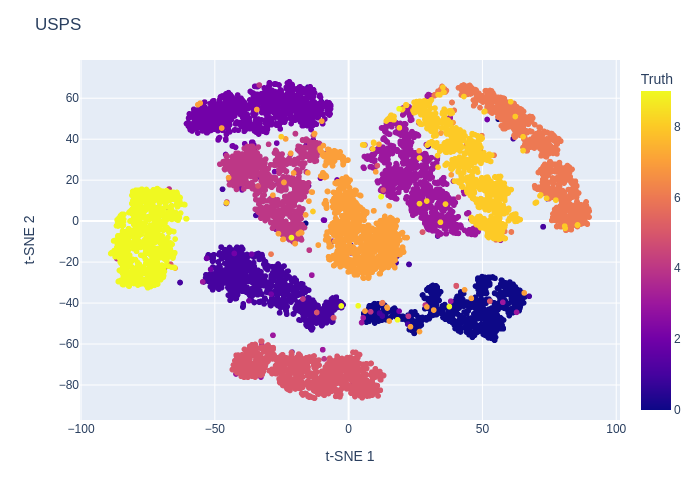}}
\subfigure[Selected space (UMist)]{\includegraphics[width = 0.24\linewidth]{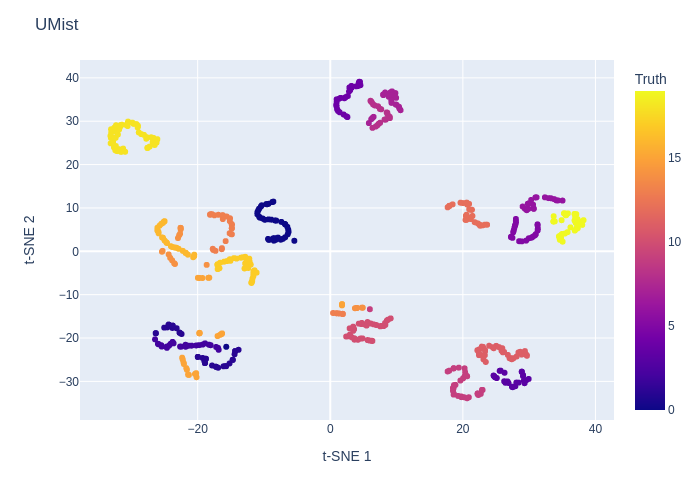}}
\subfigure[Excluded space (UMist)]{\includegraphics[width = 0.24\linewidth]{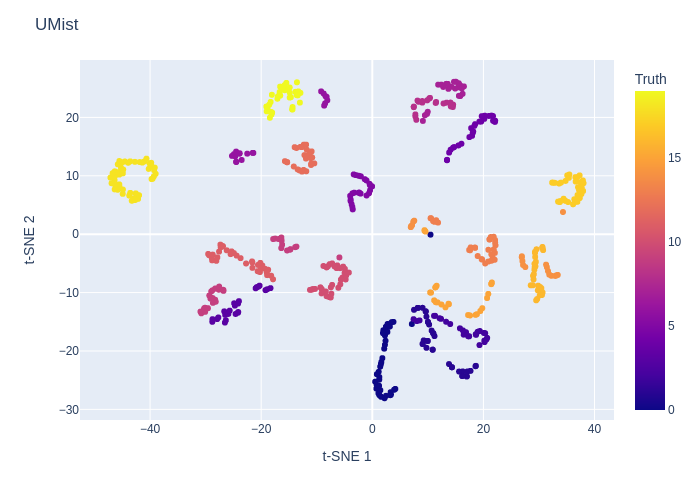}}
\subfigure[Selected space (COIL-20)]{\includegraphics[width = 0.24\linewidth]{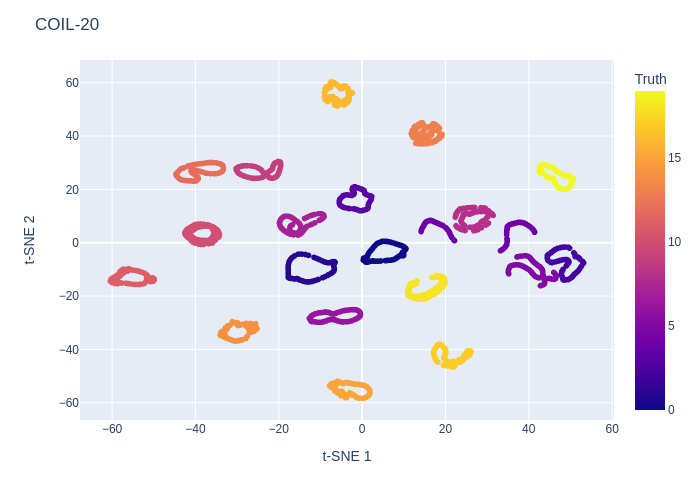}}
\subfigure[Excluded space (COIL-20)]{\includegraphics[width = 0.24\linewidth]{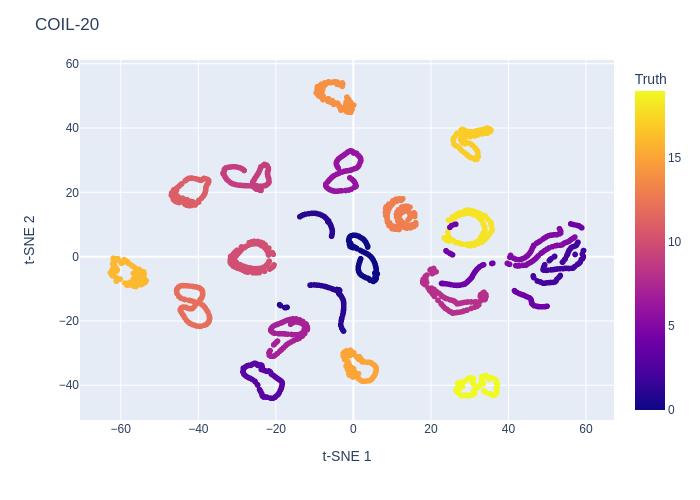}}
\subfigure[Selected space (COIL-100)]{\includegraphics[width = 0.24\linewidth]{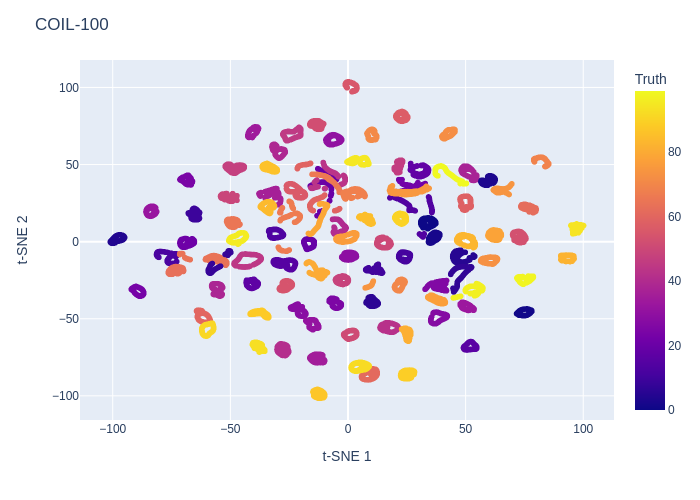}}
\subfigure[Excluded space (COIL-100)]{\includegraphics[width = 0.24\linewidth]{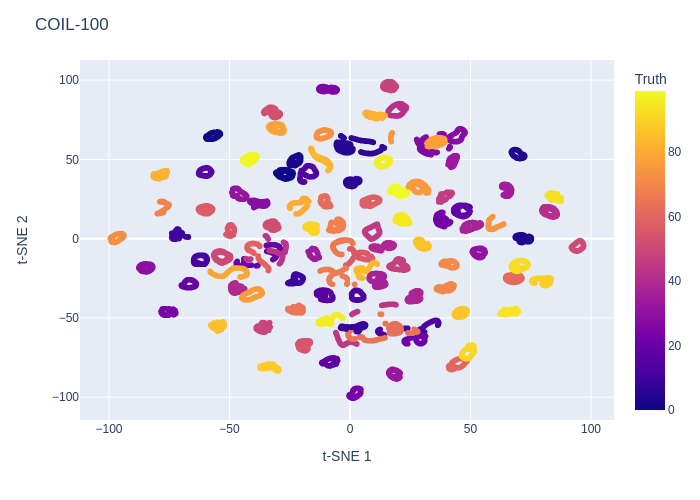}}
\subfigure[Selected space (MNIST-test)]{\includegraphics[width = 0.24\linewidth]{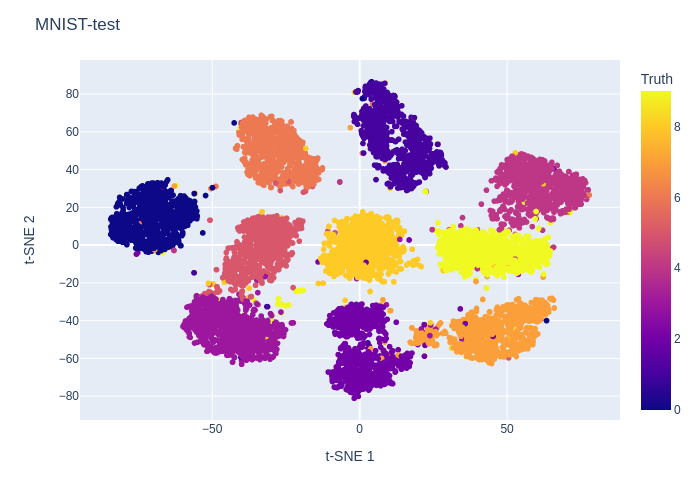}}
\subfigure[Excluded space (MNIST-test)]{\includegraphics[width = 0.24\linewidth]{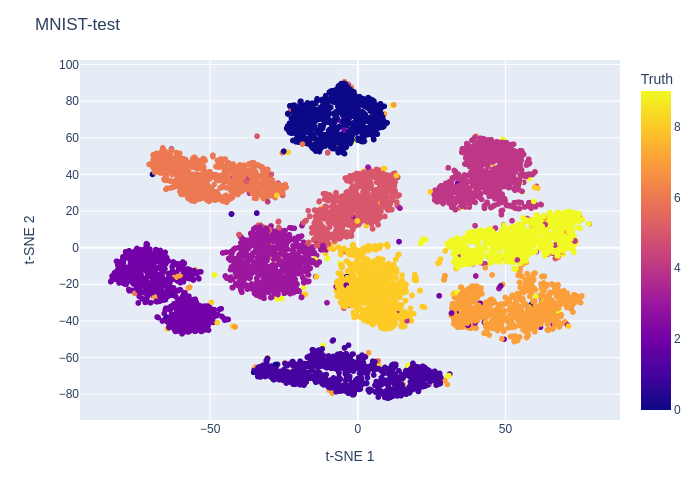}}
\subfigure[Selected space (CMU-PIE)]{\includegraphics[width = 0.24\linewidth]{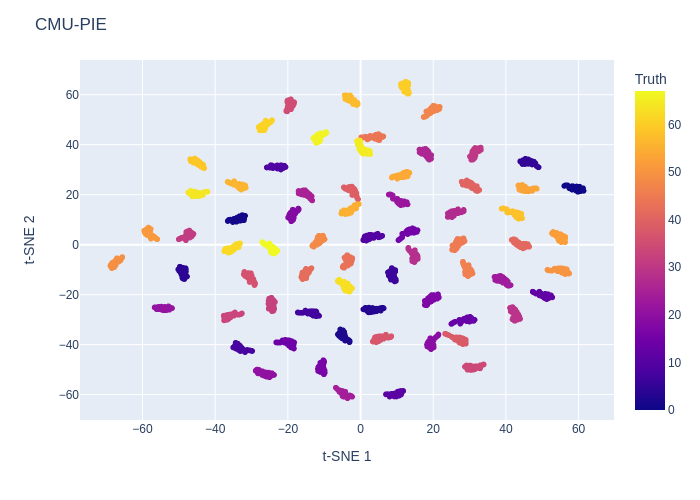}}
\subfigure[Excluded space (CMU-PIE)]{\includegraphics[width = 0.24\linewidth]{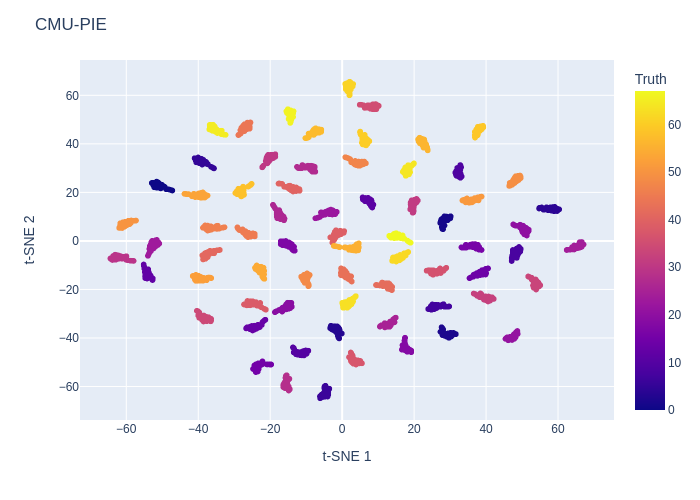}}
\subfigure[Selected space (FRGC)]{\includegraphics[width = 0.24\linewidth]{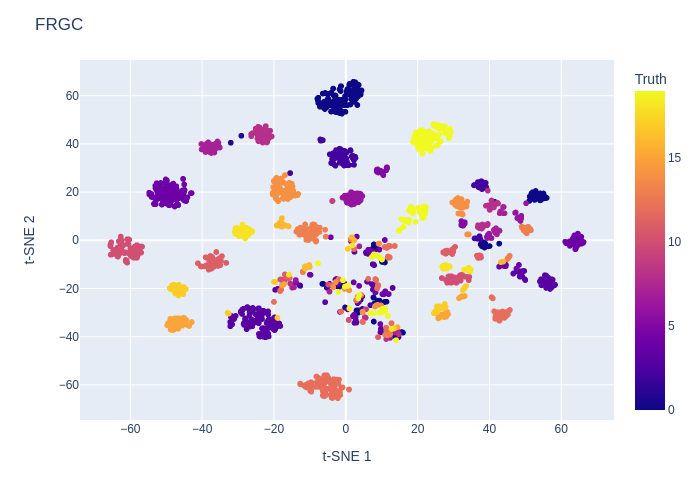}}
\subfigure[Selected space (YTF)]{\includegraphics[width = 0.24\linewidth]{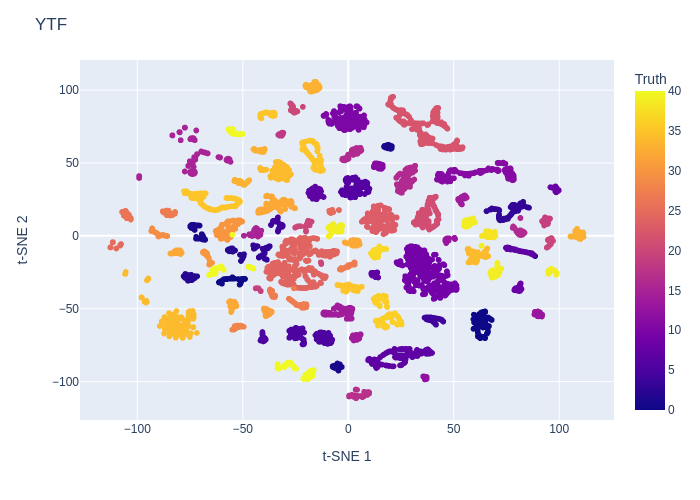}}
\caption{t-SNE visualization illustrating the selected embedding spaces from ACE in comparison to those excluded from ACE, based on Silhouette score (Cosine distance), for Application 2 with JULE. Each data point in the visualizations is assigned a color corresponding to its true cluster label.}
\label{fig:tsne:cosine:2}
\end{figure}
%%%%%%%%%%%%%%%%%%%%%%%%%
\begin{figure}[htbp!]
\centering
\subfigure[USPS]{\includegraphics[width = 0.3\linewidth]{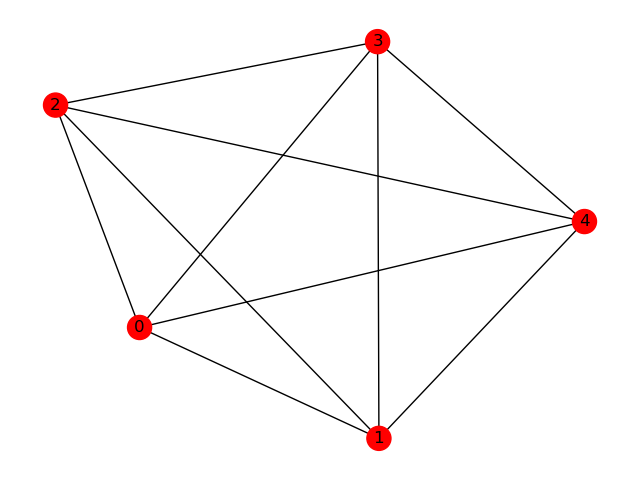}}
\subfigure[UMist]{\includegraphics[width = 0.3\linewidth]{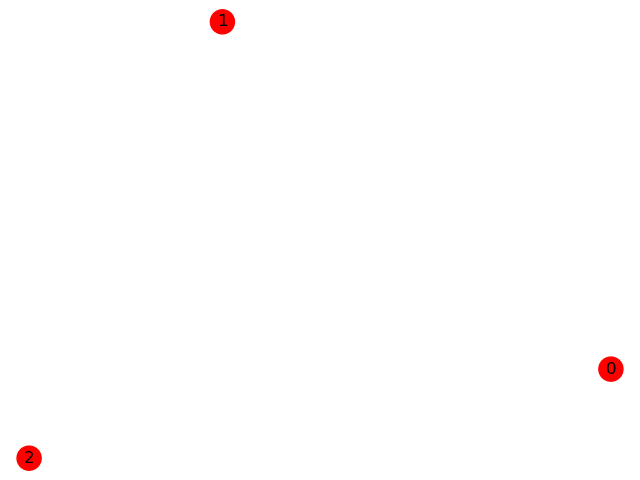}}
\subfigure[COIL-20]{\includegraphics[width = 0.3\linewidth]{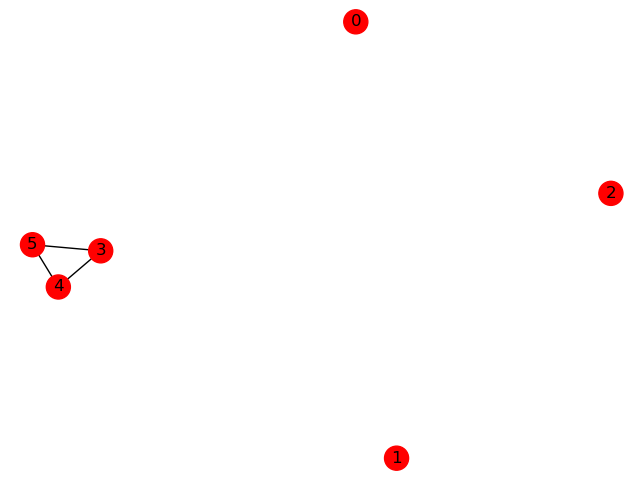}}
\subfigure[COIL-100]{\includegraphics[width = 0.3\linewidth]{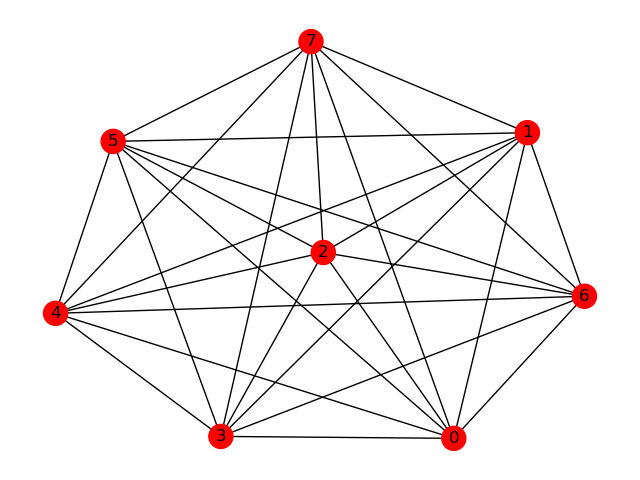}}
\subfigure[YTF]{\includegraphics[width = 0.3\linewidth]{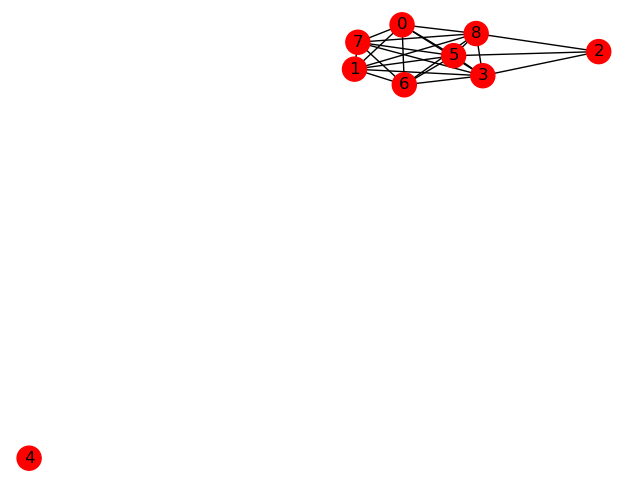}}
\subfigure[FRGC]{\includegraphics[width = 0.3\linewidth]{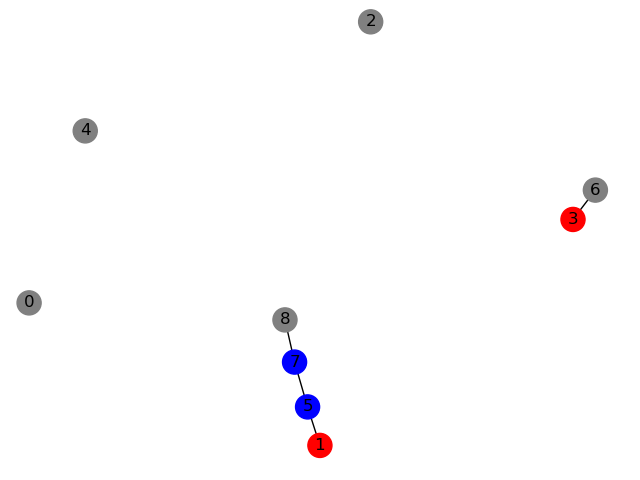}}
\subfigure[MNIST-test]{\includegraphics[width = 0.3\linewidth]{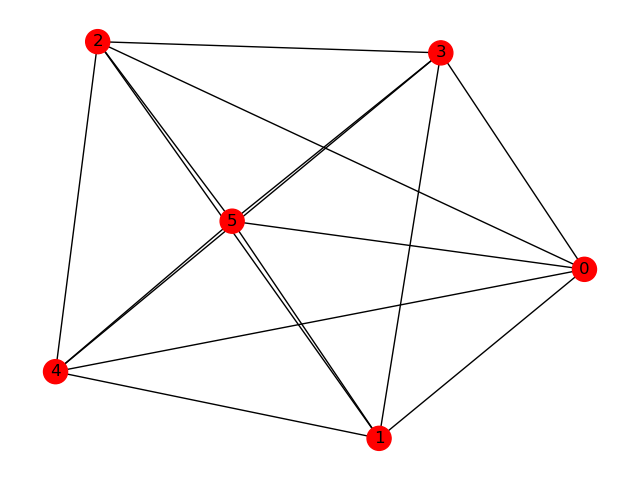}}
\subfigure[CMU-PIE]{\includegraphics[width = 0.3\linewidth]{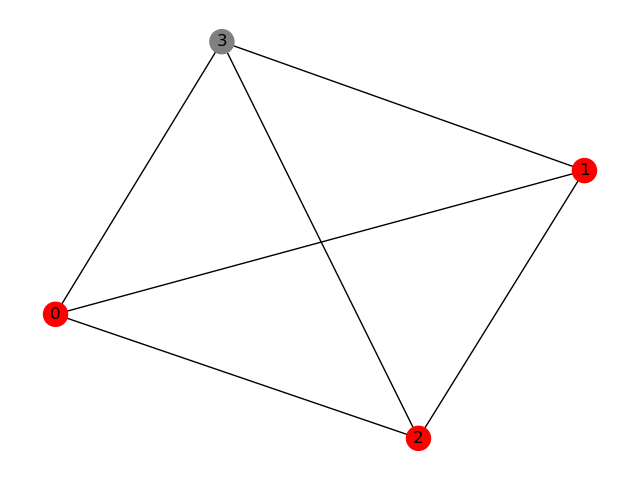}}
\caption{Graph depicting rank correlation based on Silhouette score (Euclidean distance) among embedding spaces for Application 2 with JULE. Each node represents an embedding space, and each edge signifies a significant rank correlation.}
\label{fig:graph:euclidean:2}
\end{figure}
%%%%%%%%%%%%%%%%%%%%%%%%%
\begin{figure}[htbp!]
\centering
\subfigure[Selected space (USPS)]{\includegraphics[width = 0.24\linewidth]{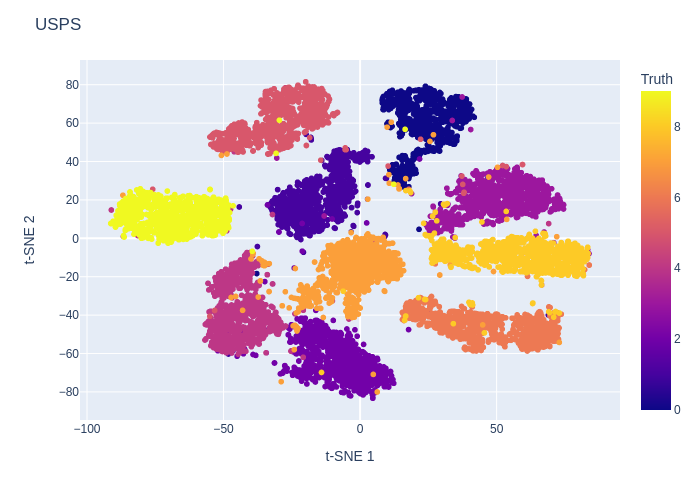}}
\subfigure[Excluded space (USPS)]{\includegraphics[width = 0.24\linewidth]{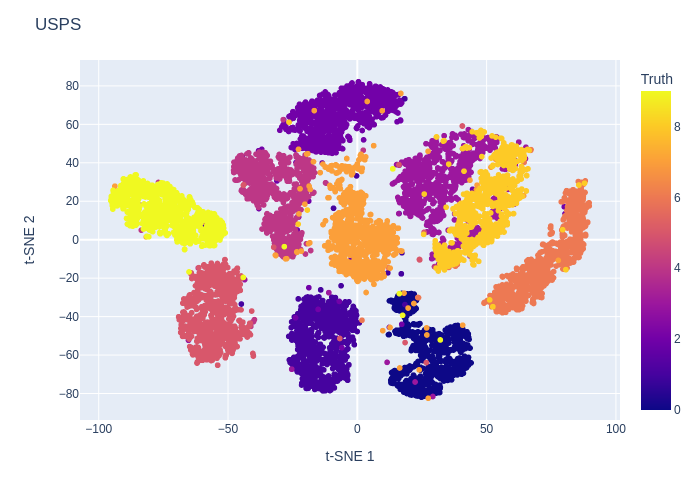}}
\subfigure[Selected space (UMist)]{\includegraphics[width = 0.24\linewidth]{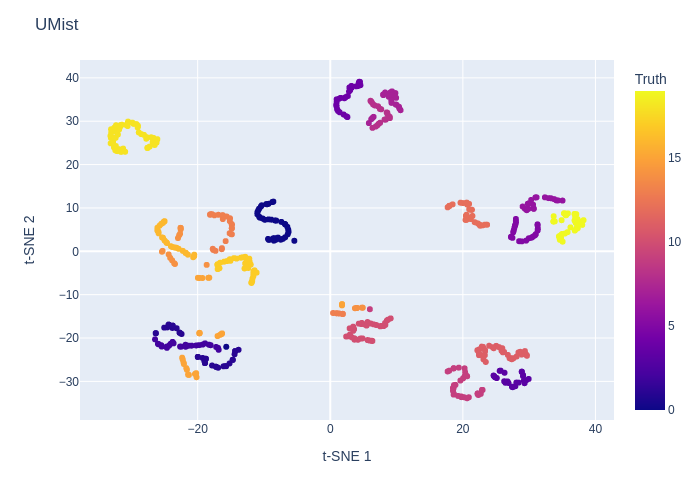}}
\subfigure[Excluded space (UMist)]{\includegraphics[width = 0.24\linewidth]{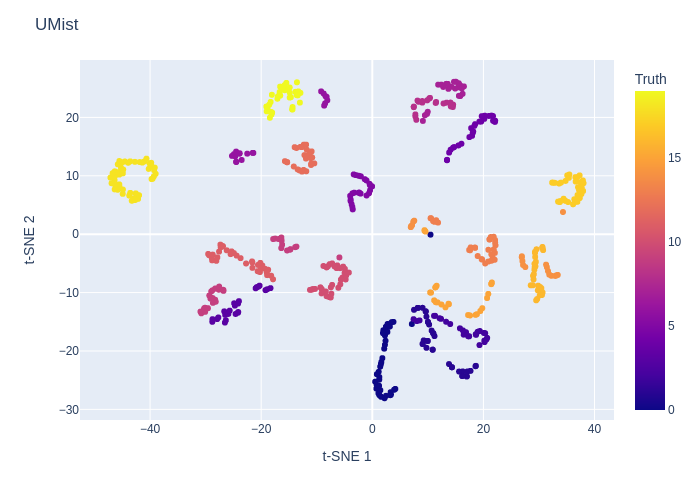}}
\subfigure[Selected space (COIL-20)]{\includegraphics[width = 0.24\linewidth]{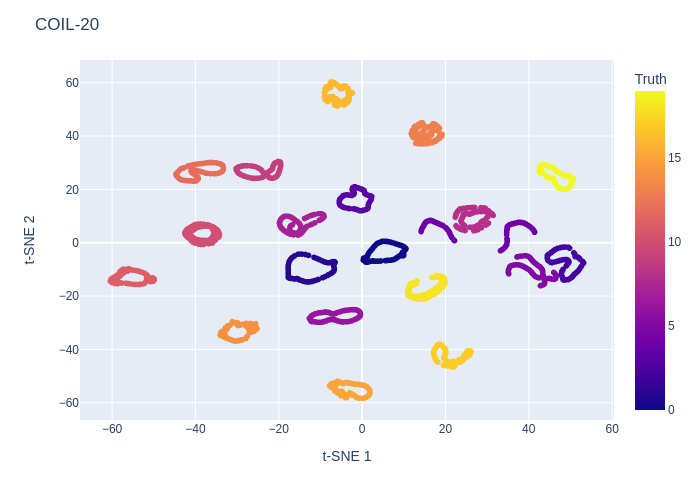}}
\subfigure[Excluded space (COIL-20)]{\includegraphics[width = 0.24\linewidth]{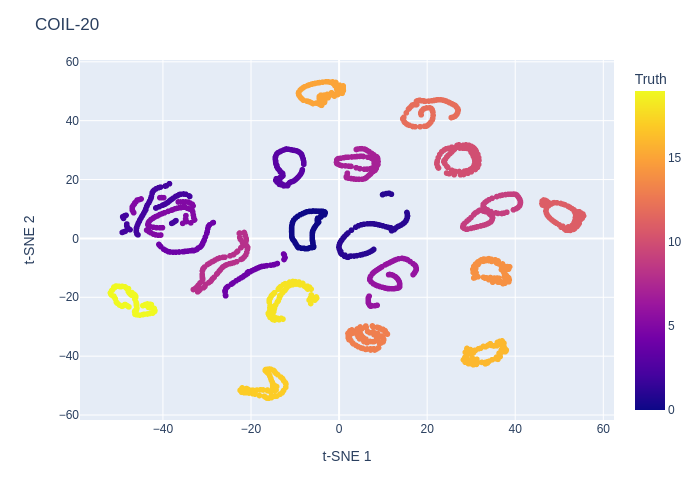}}
\subfigure[Selected space (COIL-100)]{\includegraphics[width = 0.24\linewidth]{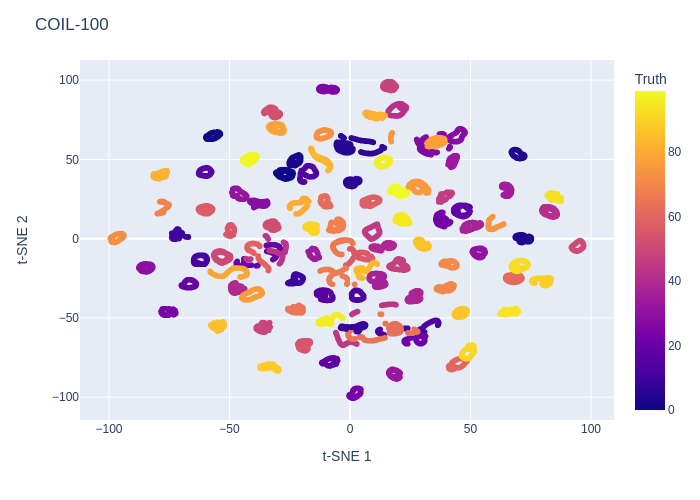}}
\subfigure[Excluded space (COIL-100)]{\includegraphics[width = 0.24\linewidth]{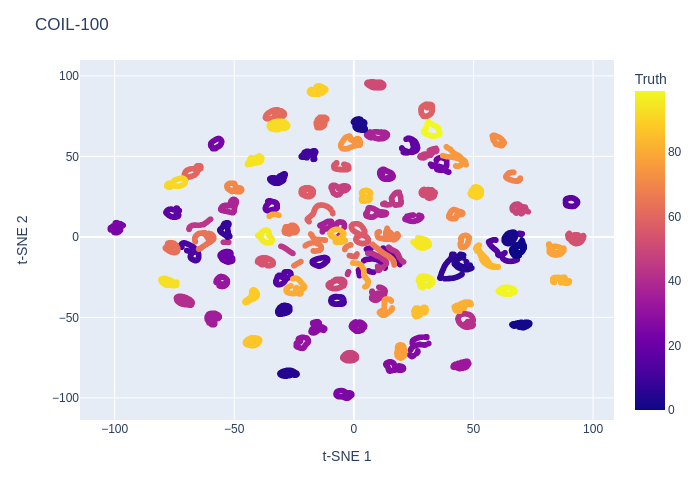}}
\subfigure[Selected space (FRGC)]{\includegraphics[width = 0.24\linewidth]{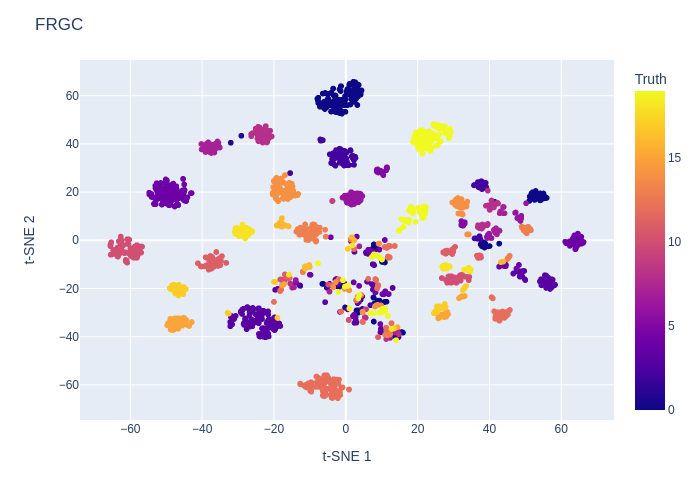}}
\subfigure[Excluded space (FRGC)]{\includegraphics[width = 0.24\linewidth]{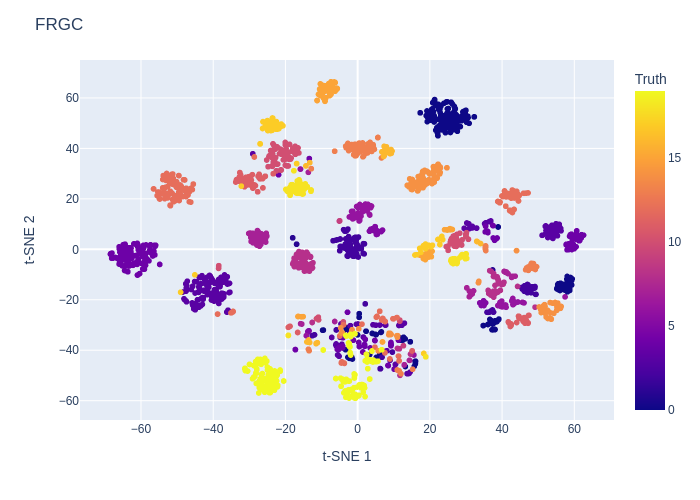}}
\subfigure[Selected space (MNIST-test)]{\includegraphics[width = 0.24\linewidth]{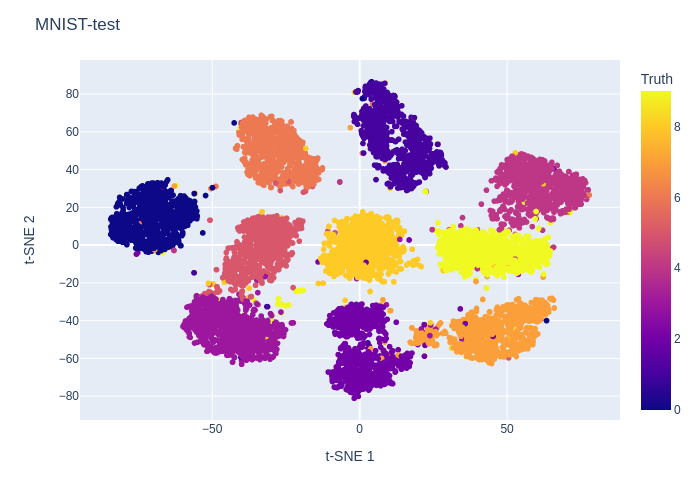}}
\subfigure[Excluded space (MNIST-test)]{\includegraphics[width = 0.24\linewidth]{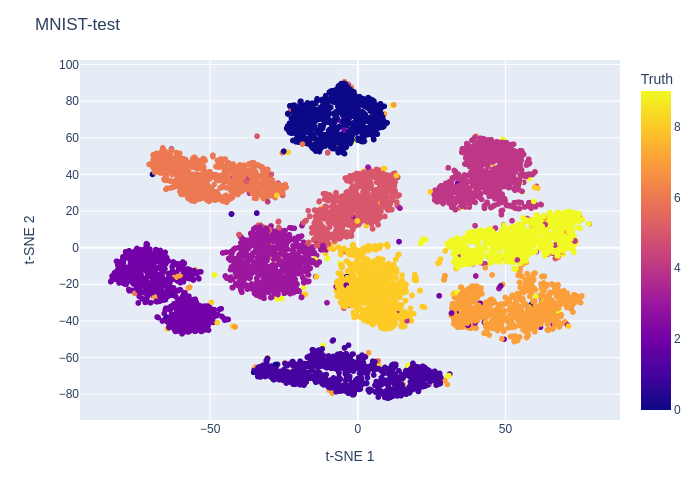}}
\subfigure[Selected space (CMU-PIE)]{\includegraphics[width = 0.24\linewidth]{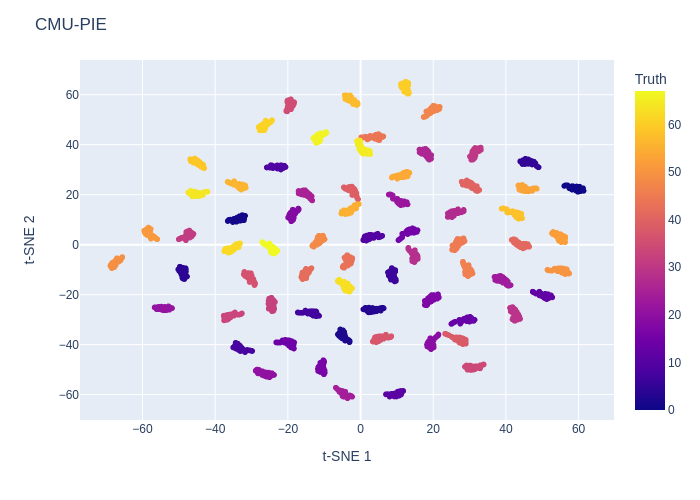}}
\subfigure[Excluded space (CMU-PIE)]{\includegraphics[width = 0.24\linewidth]{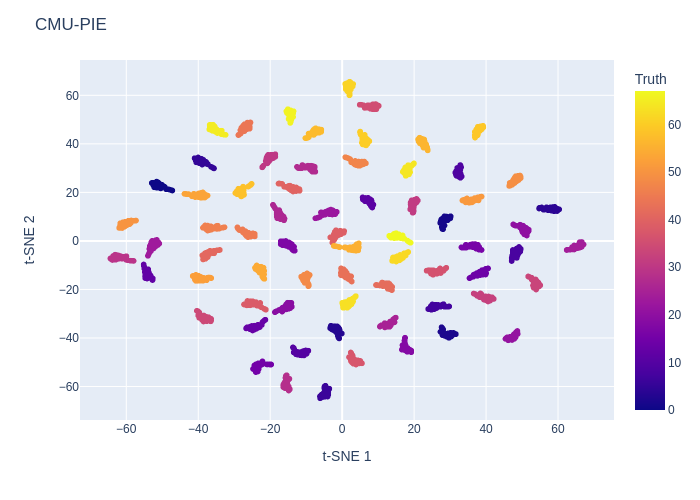}}
\subfigure[Selected space (YTF)]{\includegraphics[width = 0.24\linewidth]{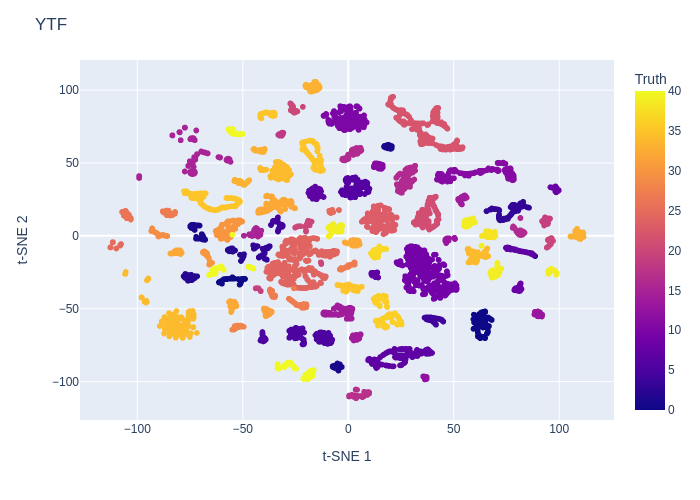}}
\caption{t-SNE visualization illustrating the selected embedding spaces from ACE in comparison to those excluded from ACE, based on Silhouette score (Euclidean distance), for Application 2 with JULE. Each data point in the visualizations is assigned a color corresponding to its true cluster label.}
\label{fig:tsne:euclidean:2}
\end{figure}
%%%%%%%%%%%%%%%%%%%%%%%%%
\begin{figure}[htbp!]
\centering
\subfigure[USPS]{\includegraphics[width = 0.3\linewidth]{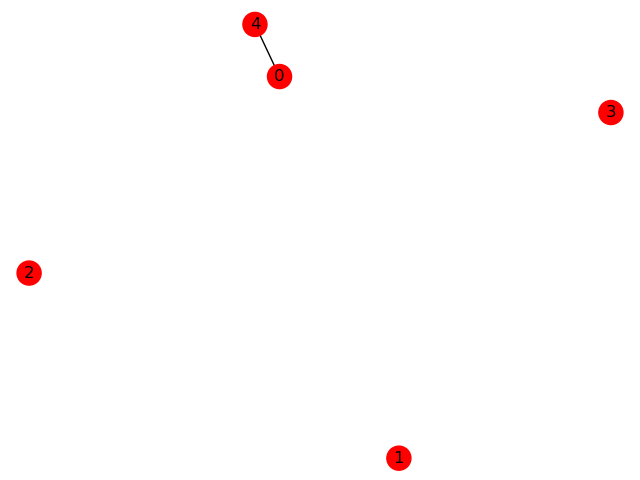}}
\subfigure[YTF]{\includegraphics[width = 0.3\linewidth]{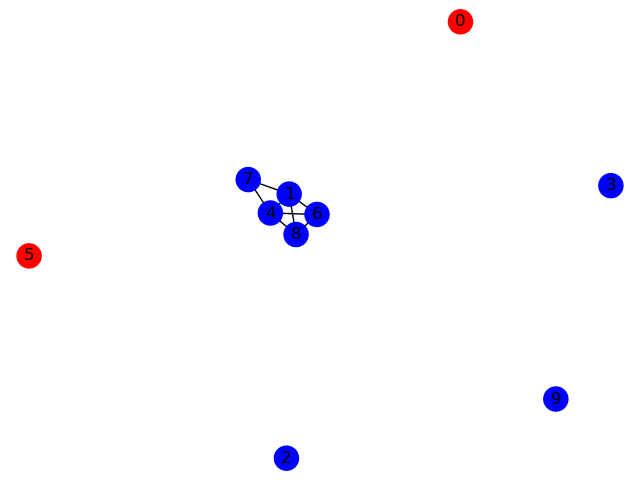}}
\subfigure[FRGC]{\includegraphics[width = 0.3\linewidth]{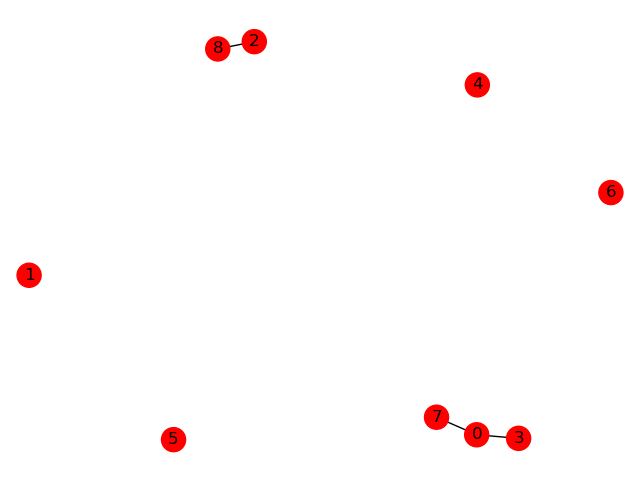}}
\subfigure[MNIST-test]{\includegraphics[width = 0.3\linewidth]{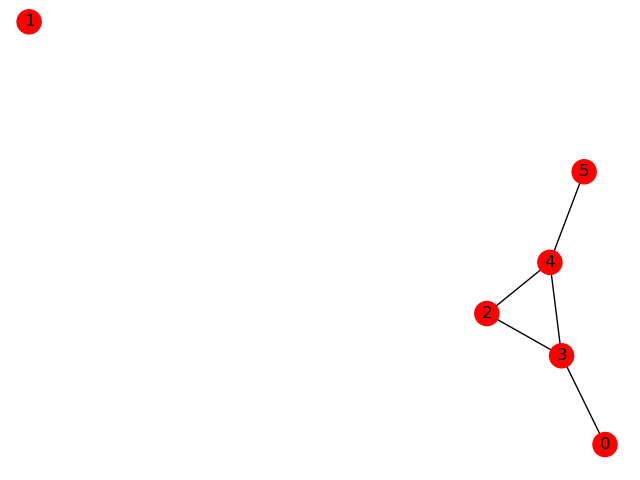}}
\subfigure[CMU-PIE]{\includegraphics[width = 0.3\linewidth]{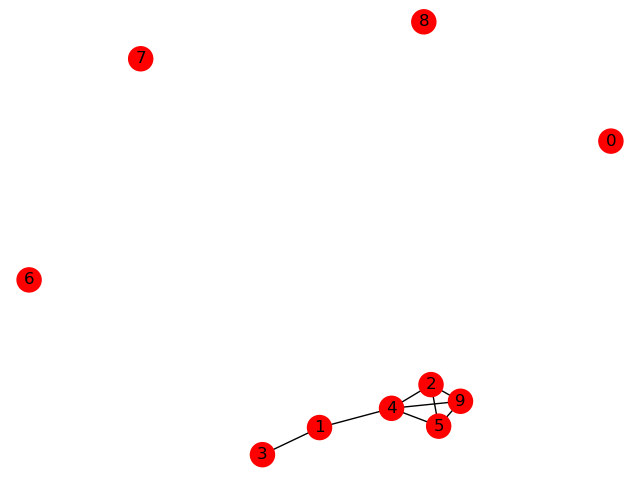}}
\caption{Graph depicting rank correlation based on Davies-Bouldin index among embedding spaces for Application 2 with DEPICT. Each node represents an embedding space, and each edge signifies a significant rank correlation.}
\label{fig:graph:dav:3}
\end{figure}
%%%%%%%%%%%%%%%%%%%%%%%%%
\begin{figure}[htbp!]
\centering
\subfigure[Selected space (USPS)]{\includegraphics[width = 0.24\linewidth]{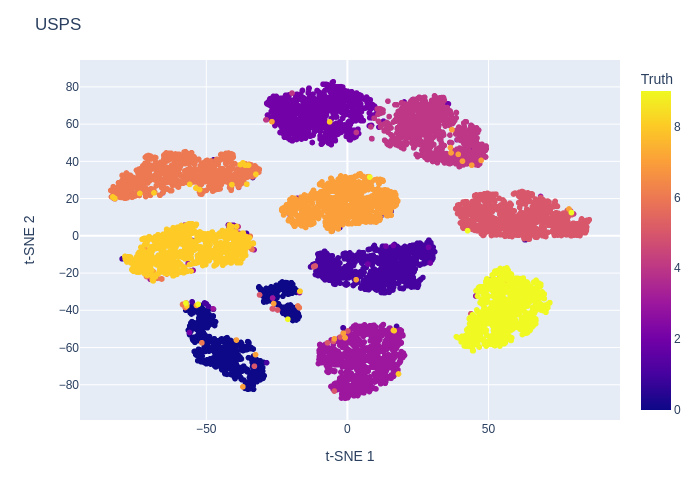}}
\subfigure[Excluded space (USPS)]{\includegraphics[width = 0.24\linewidth]{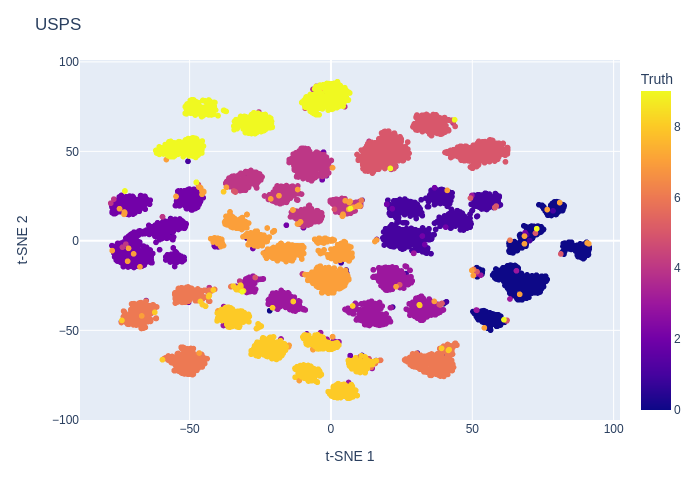}}
\subfigure[Selected space (YTF)]{\includegraphics[width = 0.24\linewidth]{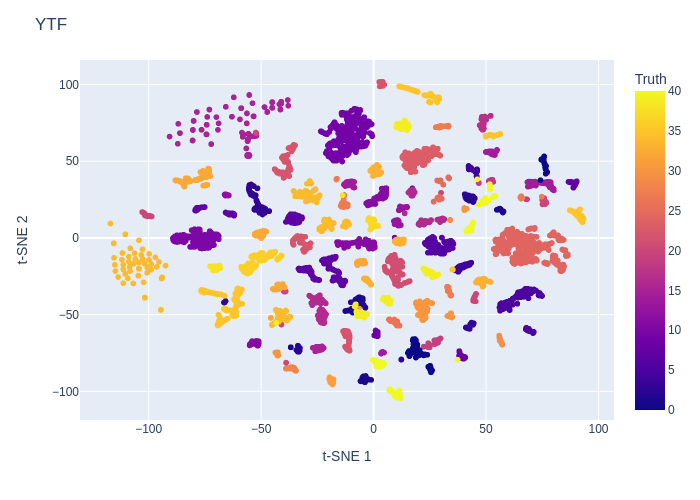}}
\subfigure[Excluded space (YTF)]{\includegraphics[width = 0.24\linewidth]{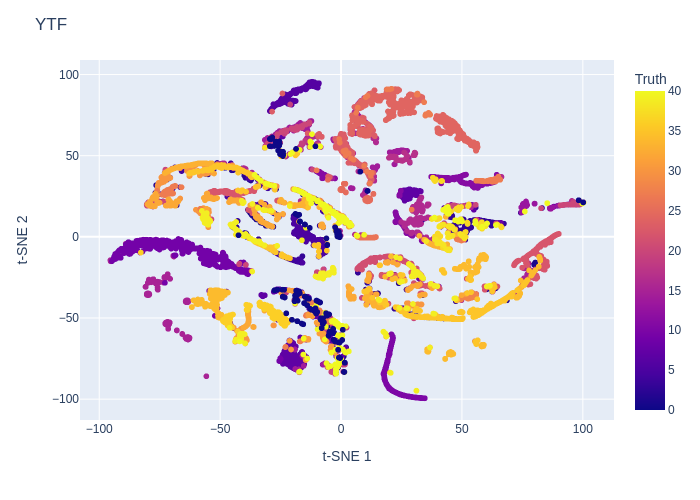}}
\subfigure[Selected space (FRGC)]{\includegraphics[width = 0.24\linewidth]{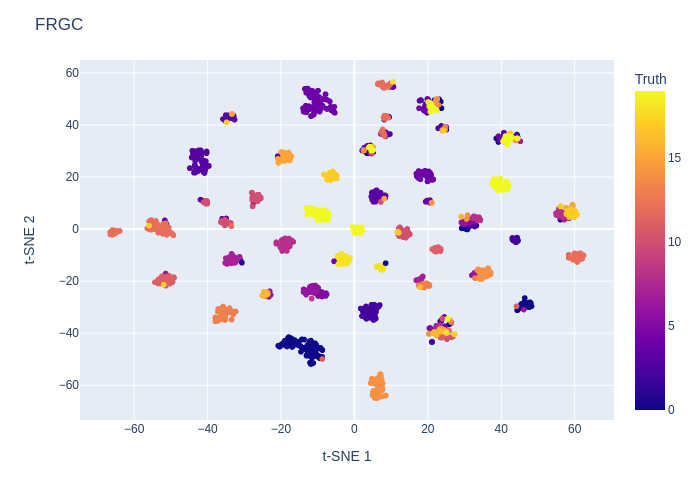}}
\subfigure[Excluded space (FRGC)]{\includegraphics[width = 0.24\linewidth]{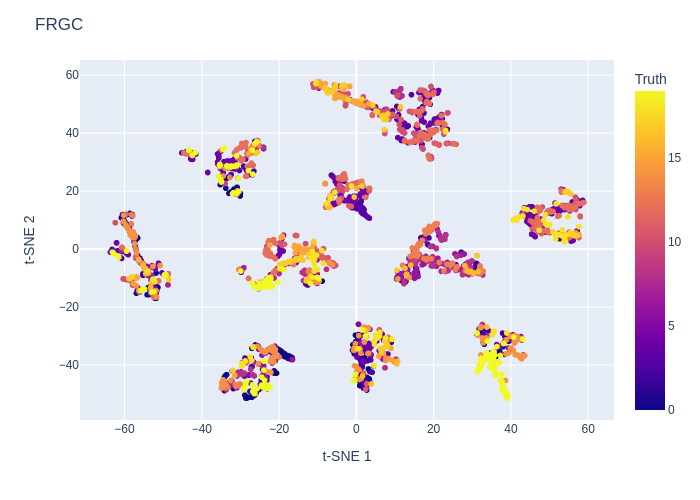}}
\subfigure[Selected space (MNIST-test)]{\includegraphics[width = 0.24\linewidth]{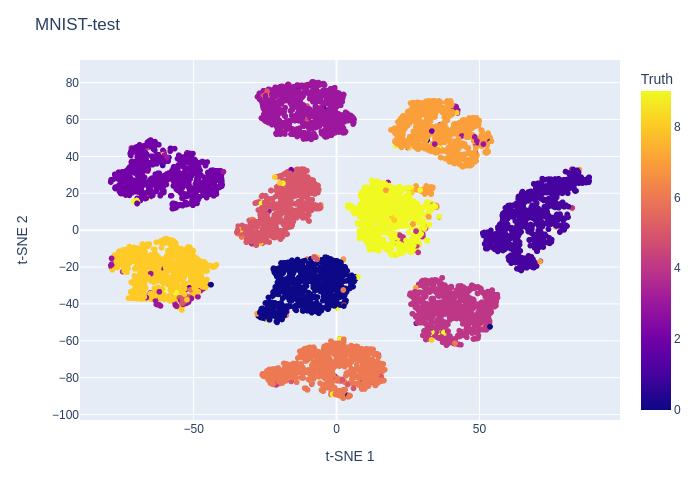}}
\subfigure[Excluded space (MNIST-test)]{\includegraphics[width = 0.24\linewidth]{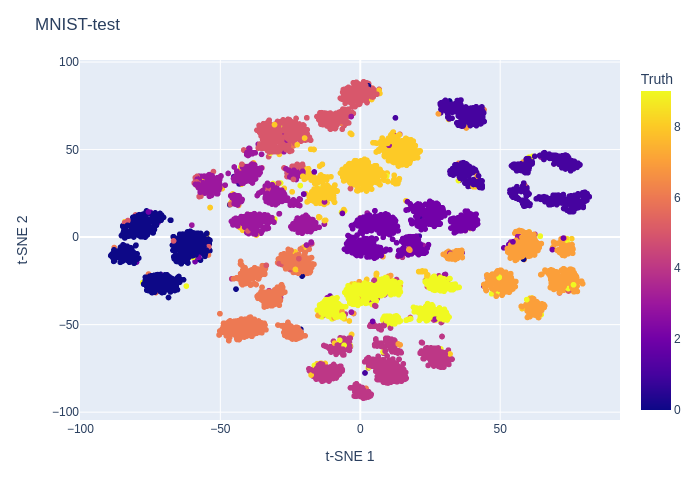}}
\subfigure[Selected space (CMU-PIE)]{\includegraphics[width = 0.24\linewidth]{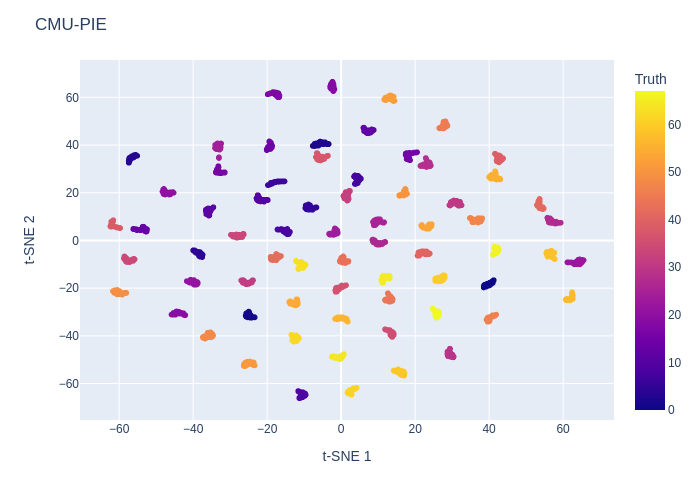}}
\caption{t-SNE visualization illustrating the selected embedding spaces from ACE in comparison to those excluded from ACE, based on Davies-Bouldin index, for Application 2 with DEPICT. Each data point in the visualizations is assigned a color corresponding to its true cluster label.}
\label{fig:tsne:dav:3}
\end{figure}
%%%%%%%%%%%%%%%%%%%%%%%%%
\begin{figure}[htbp!]
\centering
\subfigure[USPS]{\includegraphics[width = 0.3\linewidth]{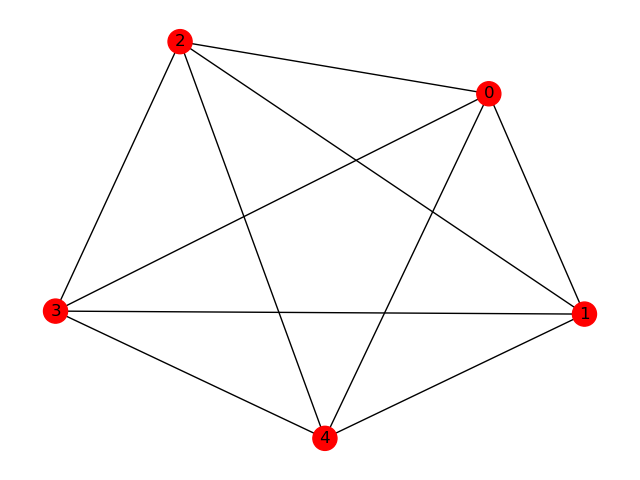}}
\subfigure[YTF]{\includegraphics[width = 0.3\linewidth]{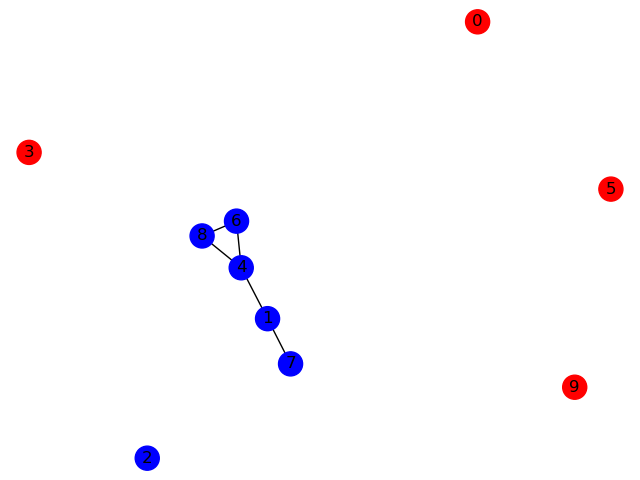}}
\subfigure[FRGC]{\includegraphics[width = 0.3\linewidth]{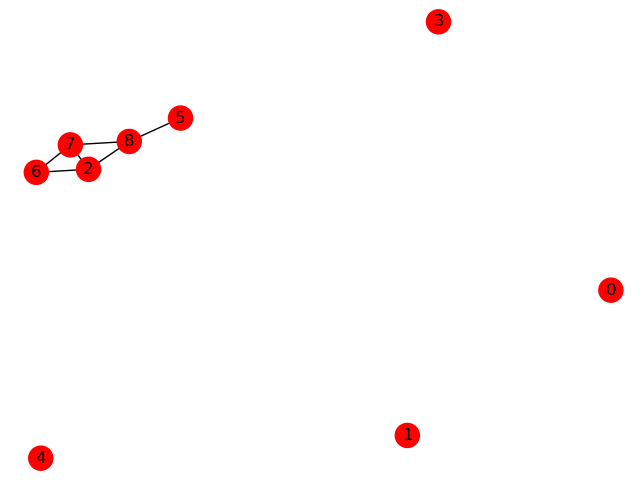}}
\subfigure[MNIST-test]{\includegraphics[width = 0.3\linewidth]{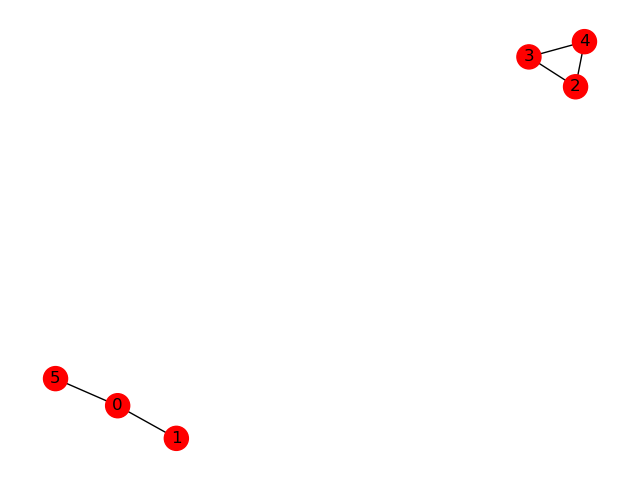}}
\subfigure[CMU-PIE]{\includegraphics[width = 0.3\linewidth]{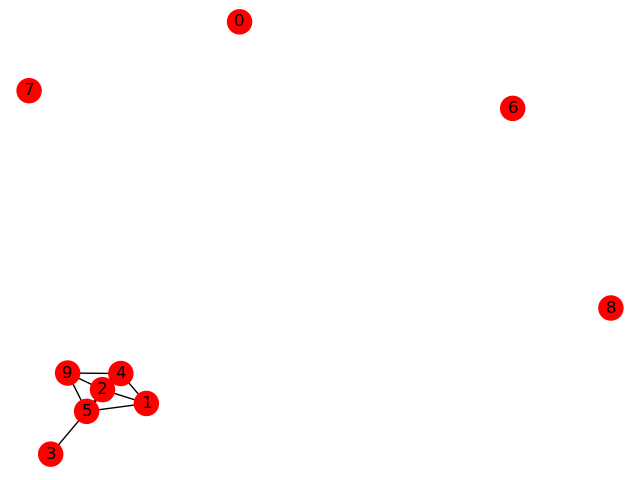}}
\caption{Graph depicting rank correlation based on Calinski-Harabasz index among embedding spaces for Application 2 with DEPICT. Each node represents an embedding space, and each edge signifies a significant rank correlation.}
\label{fig:graph:ch:3}
\end{figure}
%%%%%%%%%%%%%%%%%%%%%%%%%
\begin{figure}[htbp!]
\centering
\subfigure[Selected space (USPS)]{\includegraphics[width = 0.24\linewidth]{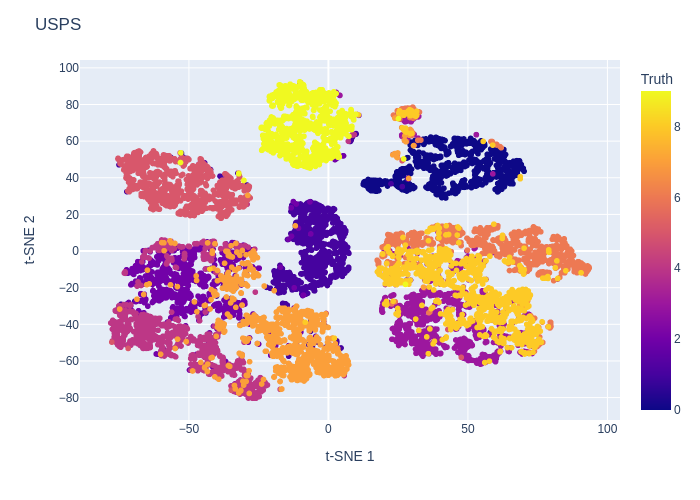}}
\subfigure[Excluded space (USPS)]{\includegraphics[width = 0.24\linewidth]{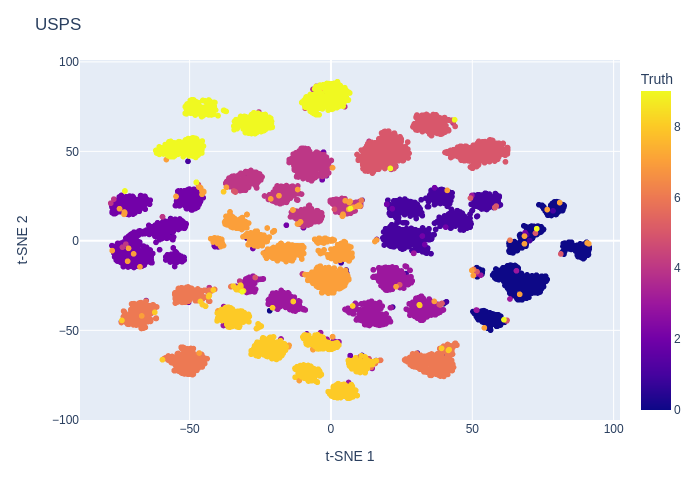}}
\subfigure[Selected space (YTF)]{\includegraphics[width = 0.24\linewidth]{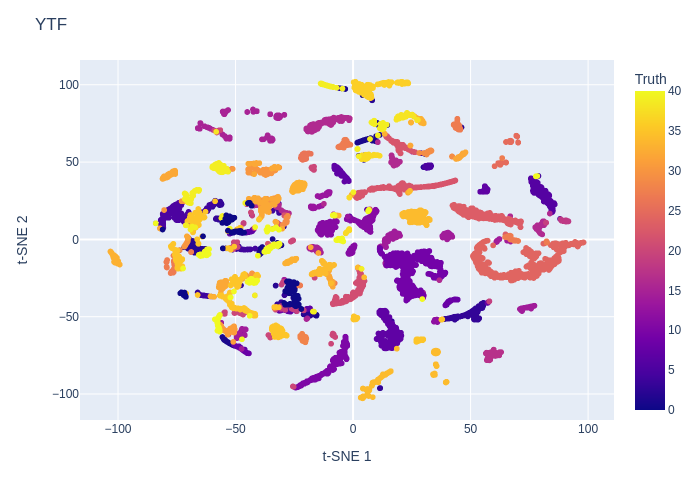}}
\subfigure[Excluded space (YTF)]{\includegraphics[width = 0.24\linewidth]{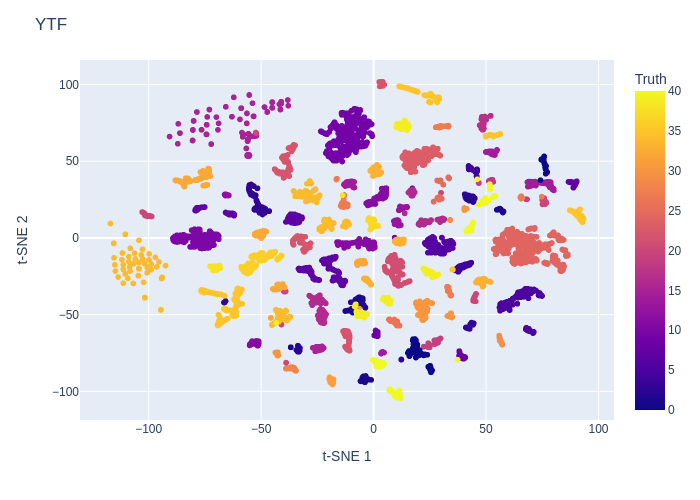}}
\subfigure[Selected space (MNIST-test)]{\includegraphics[width = 0.24\linewidth]{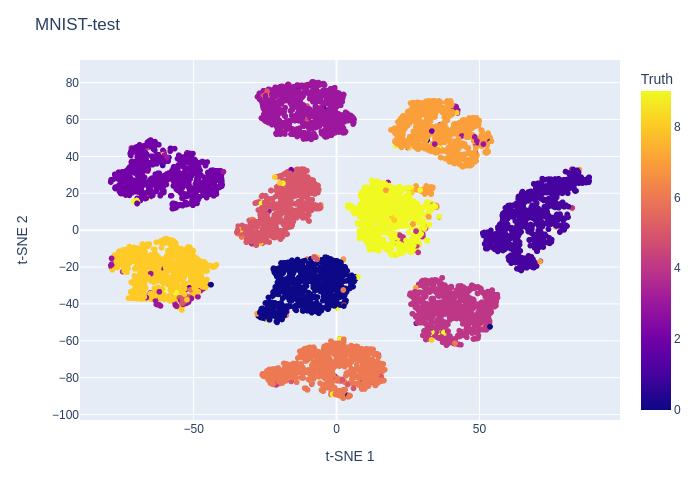}}
\subfigure[Excluded space (MNIST-test)]{\includegraphics[width = 0.24\linewidth]{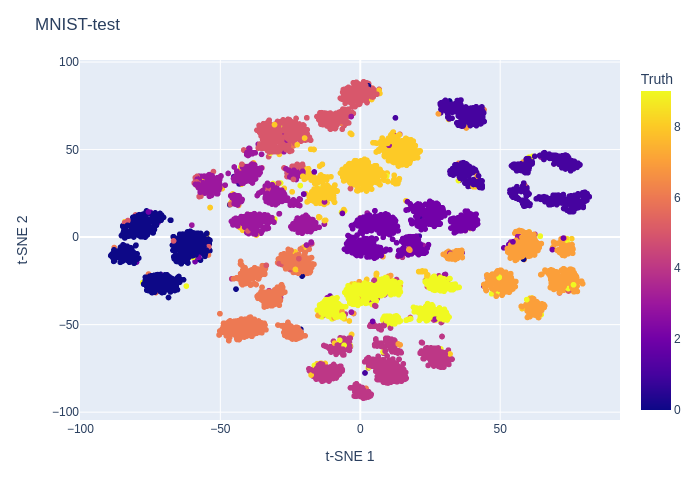}}
\subfigure[Selected space (FRGC)]{\includegraphics[width = 0.24\linewidth]{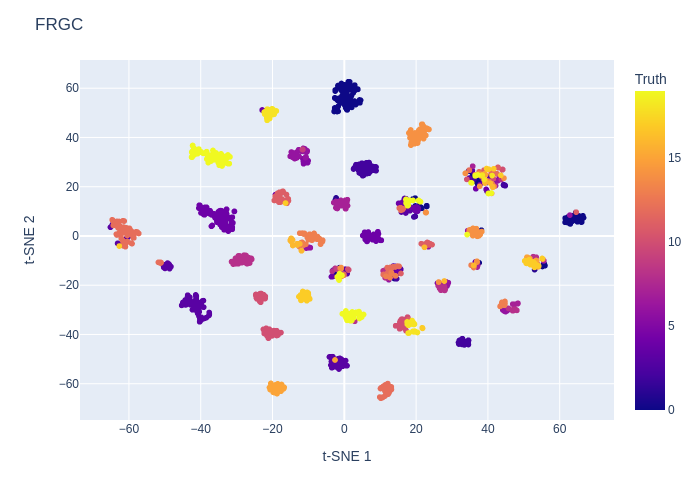}}
\subfigure[Selected space (CMU-PIE)]{\includegraphics[width = 0.24\linewidth]{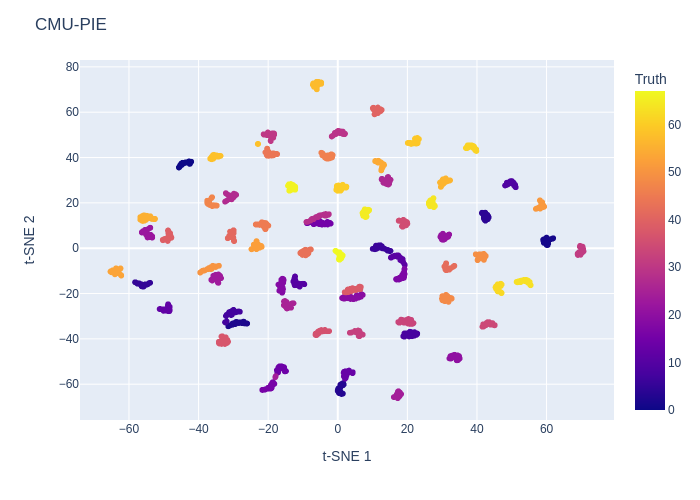}}
\caption{t-SNE visualization illustrating the selected embedding spaces from ACE in comparison to those excluded from ACE, based on Calinski-Harabasz index, for Application 2 with DEPICT. Each data point in the visualizations is assigned a color corresponding to its true cluster label.}
\label{fig:tsne:ch:3}
\end{figure}
%%%%%%%%%%%%%%%%%%%%%%%%%
\begin{figure}[htbp!]
\centering
\subfigure[USPS]{\includegraphics[width = 0.3\linewidth]{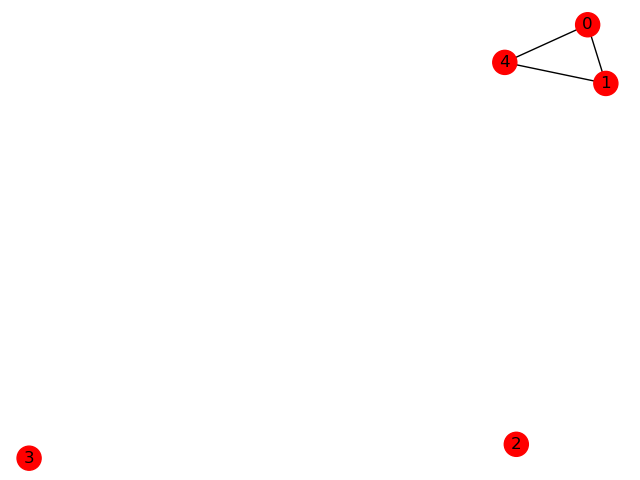}}
\subfigure[YTF]{\includegraphics[width = 0.3\linewidth]{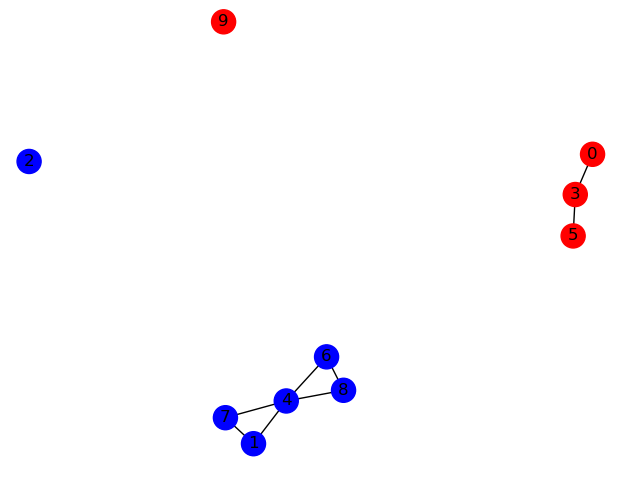}}
\subfigure[FRGC]{\includegraphics[width = 0.3\linewidth]{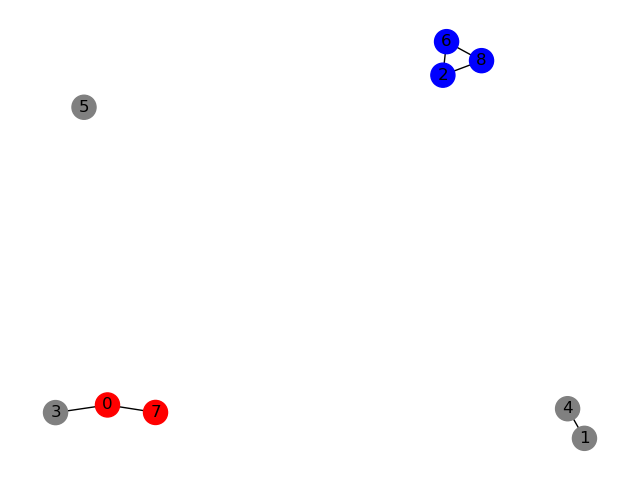}}
\subfigure[MNIST-test]{\includegraphics[width = 0.3\linewidth]{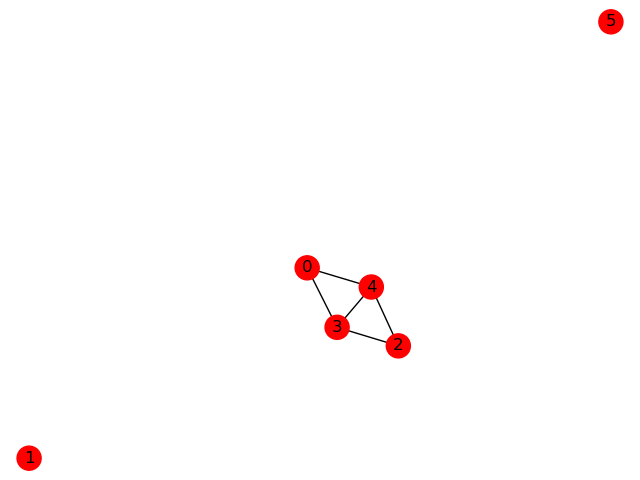}}
\subfigure[CMU-PIE]{\includegraphics[width = 0.3\linewidth]{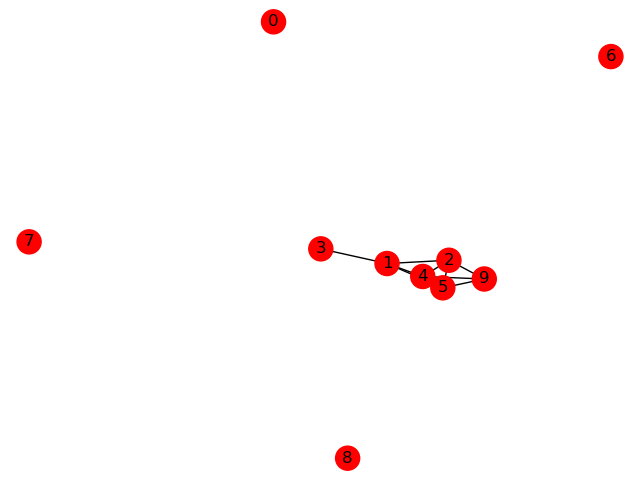}}
\caption{Graph depicting rank correlation based on Silhouette score (Cosine distance) among embedding spaces for Application 2 with DEPICT. Each node represents an embedding space, and each edge signifies a significant rank correlation.}
\label{fig:graph:cosine:3}
\end{figure}
%%%%%%%%%%%%%%%%%%%%%%%%%
\begin{figure}[htbp!]
\centering
\subfigure[Selected space (USPS)]{\includegraphics[width = 0.24\linewidth]{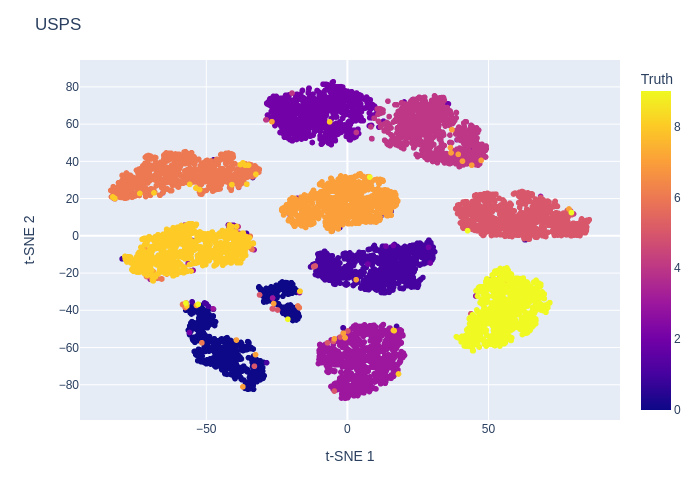}}
\subfigure[Excluded space (USPS)]{\includegraphics[width = 0.24\linewidth]{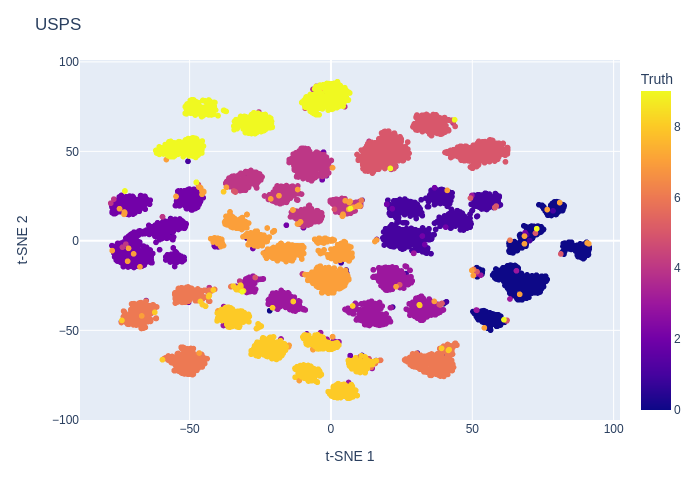}}
\subfigure[Selected space (YTF)]{\includegraphics[width = 0.24\linewidth]{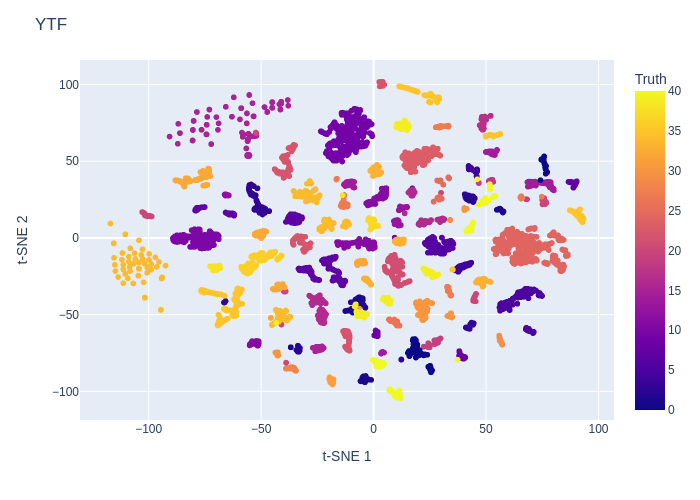}}
\subfigure[Excluded space (YTF)]{\includegraphics[width = 0.24\linewidth]{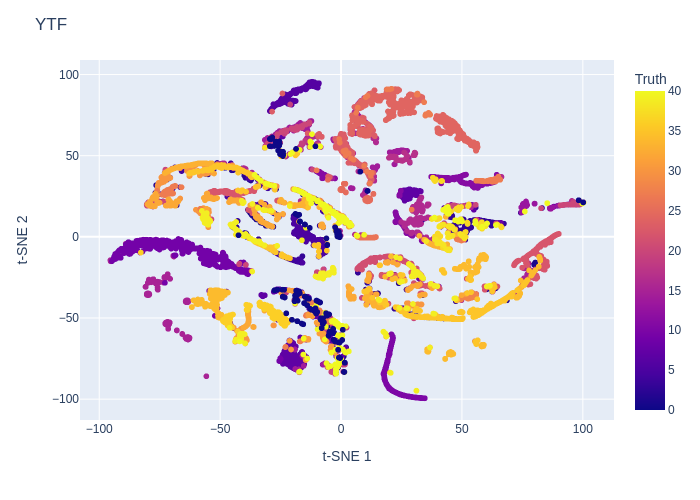}}
\subfigure[Selected space (FRGC)]{\includegraphics[width = 0.24\linewidth]{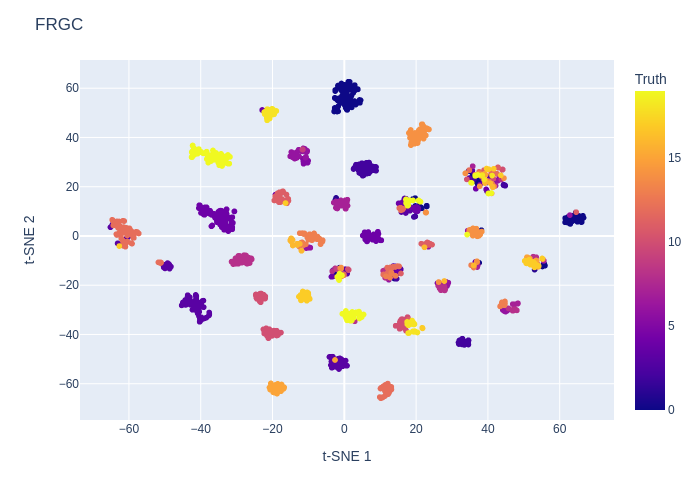}}
\subfigure[Excluded space (FRGC)]{\includegraphics[width = 0.24\linewidth]{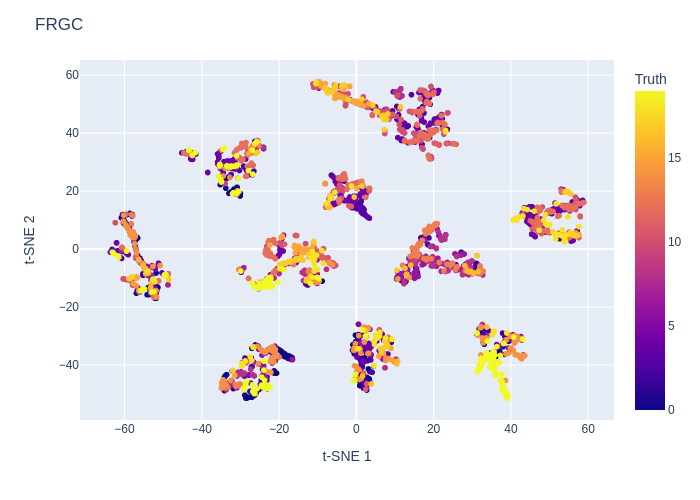}}
\subfigure[Selected space (MNIST-test)]{\includegraphics[width = 0.24\linewidth]{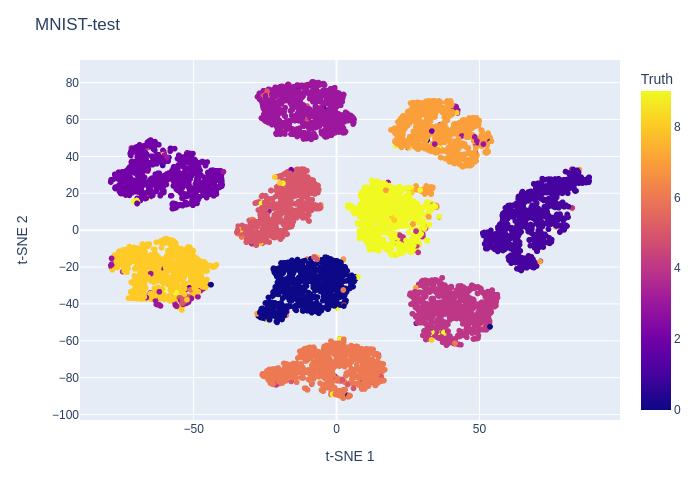}}
\subfigure[Excluded space (MNIST-test)]{\includegraphics[width = 0.24\linewidth]{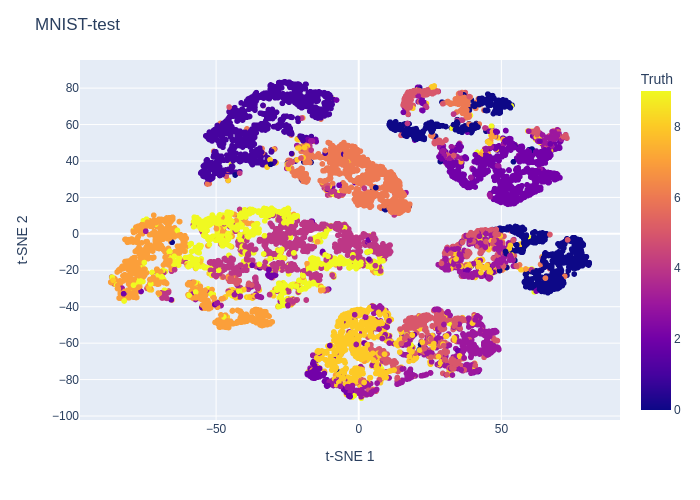}}
\subfigure[Selected space (CMU-PIE)]{\includegraphics[width = 0.24\linewidth]{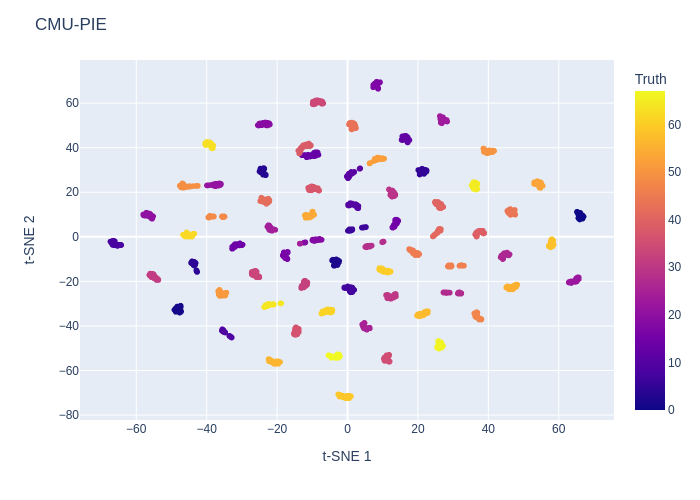}}
\caption{t-SNE visualization illustrating the selected embedding spaces from ACE in comparison to those excluded from ACE, based on Silhouette score (Cosine distance), for Application 2 with DEPICT. Each data point in the visualizations is assigned a color corresponding to its true cluster label.}
\label{fig:tsne:cosine:3}
\end{figure}
%%%%%%%%%%%%%%%%%%%%%%%%%
\begin{figure}[htbp!]
\centering
\subfigure[USPS]{\includegraphics[width = 0.3\linewidth]{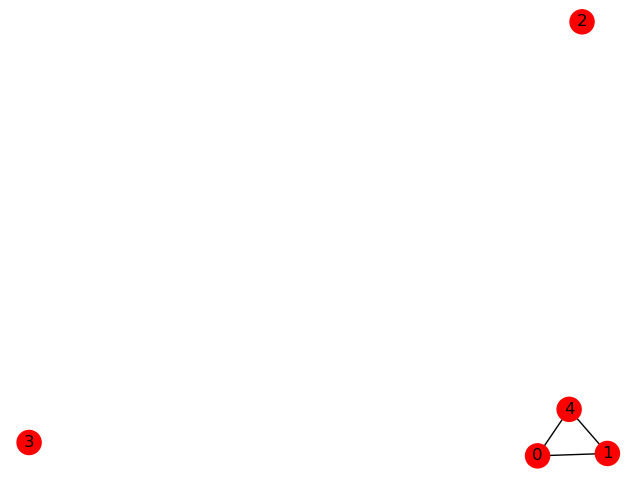}}
\subfigure[YTF]{\includegraphics[width = 0.3\linewidth]{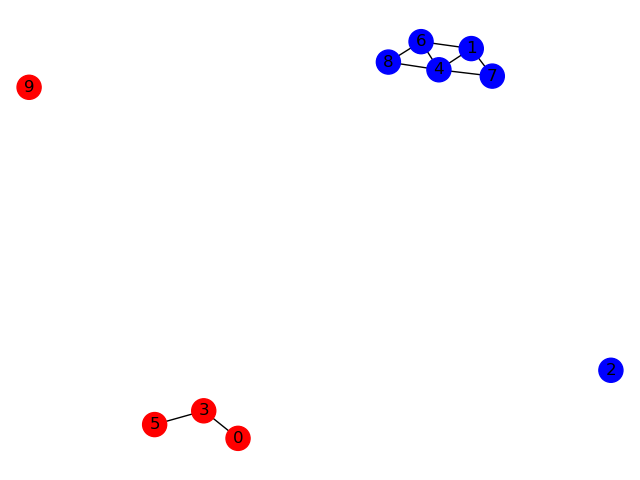}}
\subfigure[FRGC]{\includegraphics[width = 0.3\linewidth]{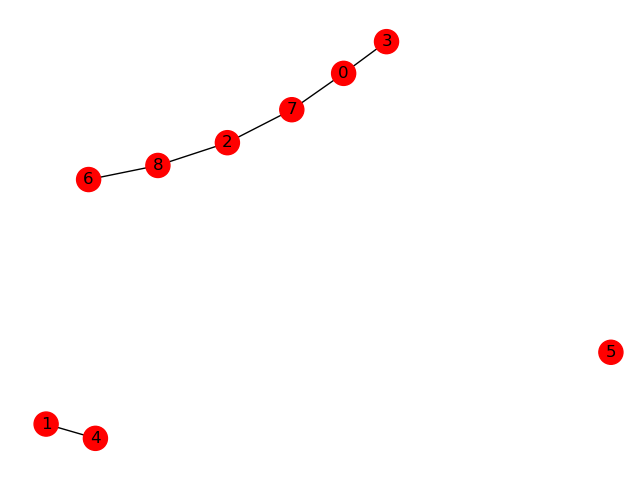}}
\subfigure[MNIST-test]{\includegraphics[width = 0.3\linewidth]{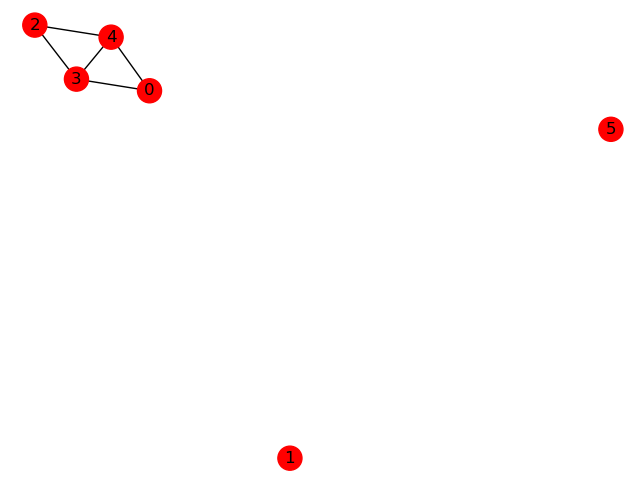}}
\subfigure[CMU-PIE]{\includegraphics[width = 0.3\linewidth]{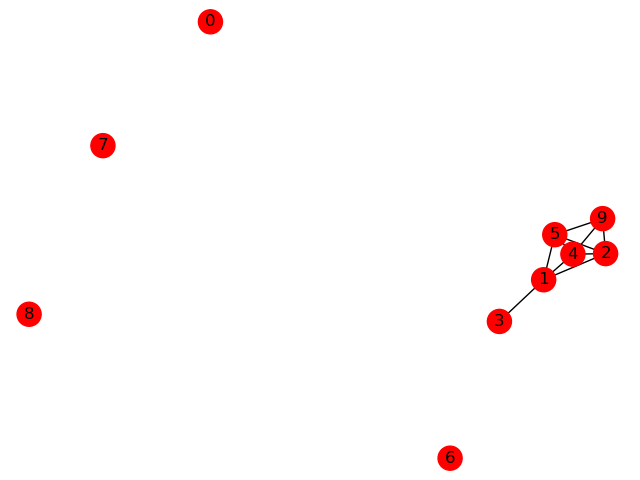}}
\caption{Graph depicting rank correlation based on Silhouette score (Euclidean distance) among embedding spaces for Application 2 with DEPICT. Each node represents an embedding space, and each edge signifies a significant rank correlation.}
\label{fig:graph:euclidean:3}
\end{figure}
%%%%%%%%%%%%%%%%%%%%%%%%%
\begin{figure}[htbp!]
\centering
\subfigure[Selected space (USPS)]{\includegraphics[width = 0.24\linewidth]{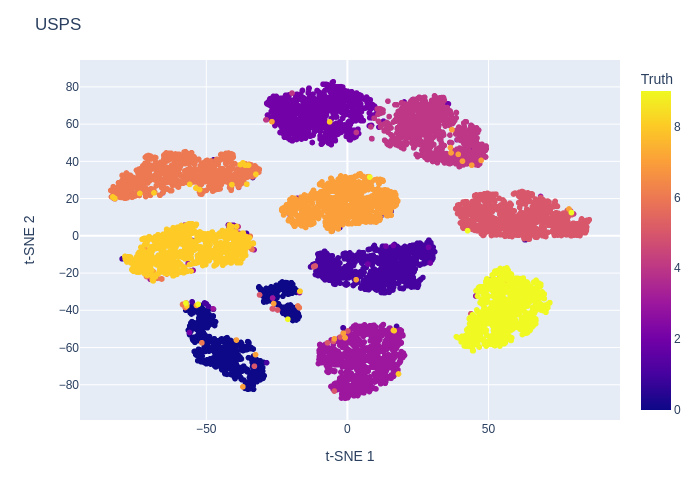}}
\subfigure[Excluded space (USPS)]{\includegraphics[width = 0.24\linewidth]{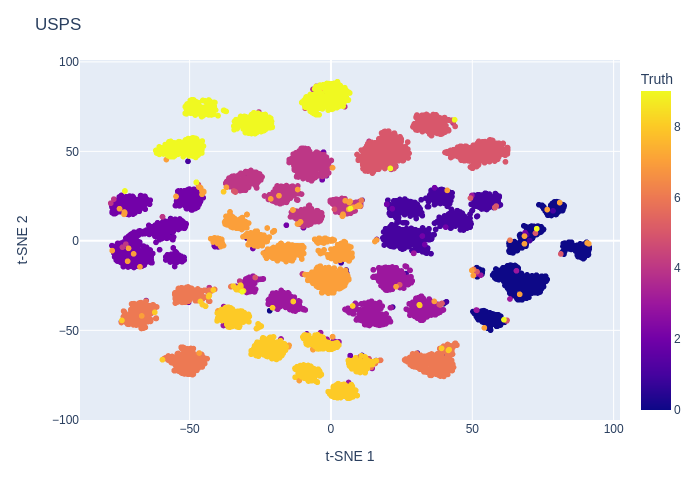}}
\subfigure[Selected space (YTF)]{\includegraphics[width = 0.24\linewidth]{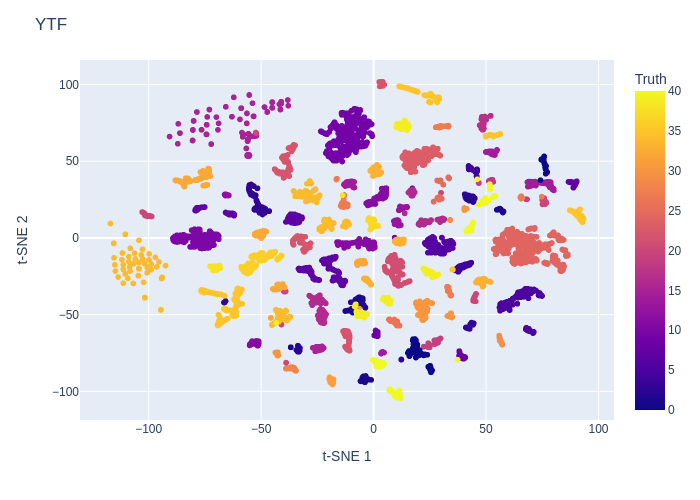}}
\subfigure[Excluded space (YTF)]{\includegraphics[width = 0.24\linewidth]{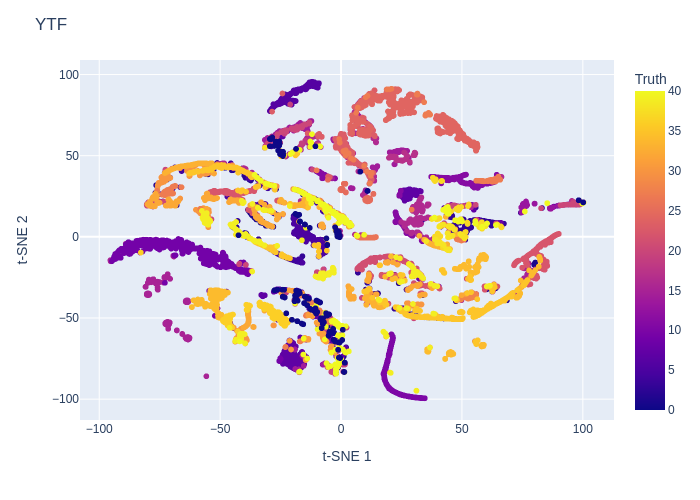}}
\subfigure[Selected space (MNIST-test)]{\includegraphics[width = 0.24\linewidth]{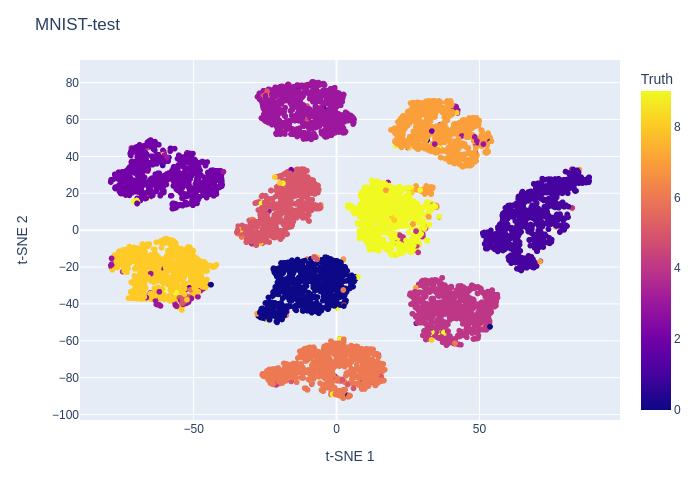}}
\subfigure[Excluded space (MNIST-test)]{\includegraphics[width = 0.24\linewidth]{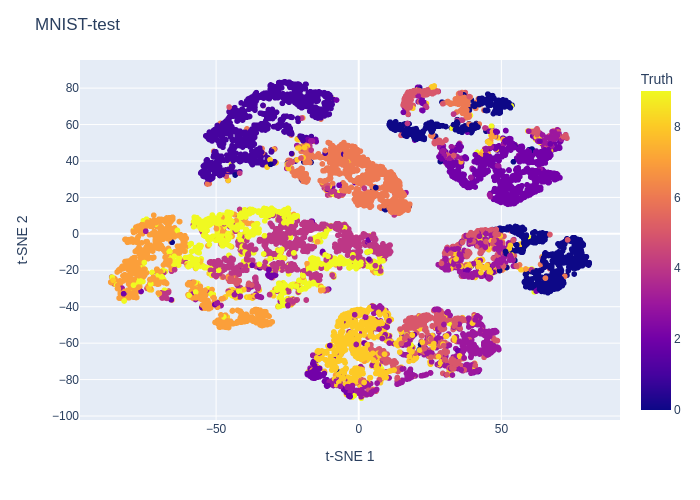}}
\subfigure[Selected space (FRGC)]{\includegraphics[width = 0.24\linewidth]{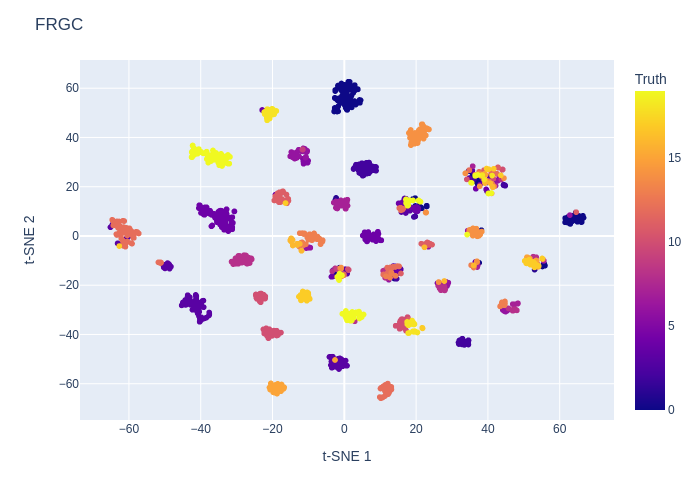}}
\subfigure[Selected space (CMU-PIE)]{\includegraphics[width = 0.24\linewidth]{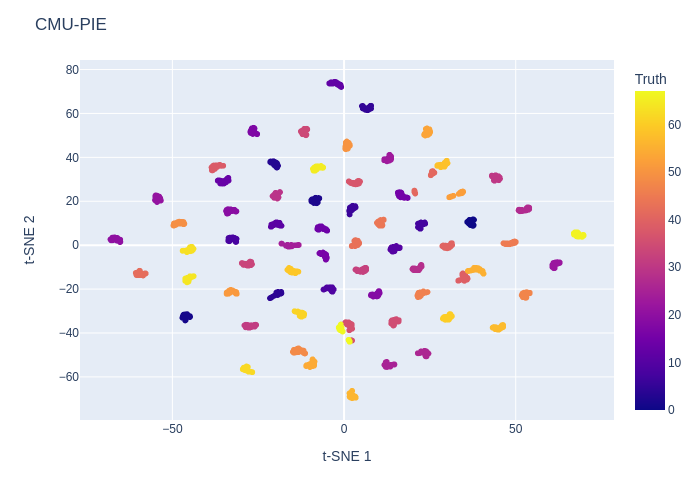}}

\caption{t-SNE visualization illustrating the selected embedding spaces from ACE in comparison to those excluded from ACE, based on Silhouette score (Euclidean distance), for Application 2 with DEPICT. Each data point in the visualizations is assigned a color corresponding to its true cluster label.}
\label{fig:tsne:euclidean:3}
\end{figure}
%%%%%%%%%%%%%%%%%%%%%%%%%

\clearpage
\newpage

\subsection{Selection of checkpoints} \label{app:dcl_res}
In this section, we present the results of the checkpoint selection experiment. Given the significantly time-consuming process of calculating raw scores for this task (approximately $50000$ images with dimensions \(224 \times 224 \times 3\) before being fed into the network), we focus solely on comparing the scores computed in the embedding spaces. We observe that all 20 obtained embedding spaces fail to reject the null hypothesis in the Dip test, as illustrated in the t-SNE visualizations in Figure \ref{fig:app:dcl}, where each subfigure is annotated with its corresponding Dip test p-value. As outlined in Algorithm \ref{Algorithm1}, when no spaces are retained, we proceed with all spaces, meaning ACE is applied to all 20 spaces. For comparison, we also compute the pooled score using all $20$ spaces. Table \ref{tab:app:dcl} reports the rank correlation results with NMI and ACC, using the Silhouette score, Calinski-Harabasz index, and Davies-Bouldin index. The ACE and pooled scores consistently outperform paired scores across all reported indices, highlighting the unreliability of paired scores for evaluation and reinforcing the importance of comparing clustering results within the same space. In this experiment, pooled scores perform similarly to ACE scores. The absence of significant multimodality may suggest a lack of admissible spaces, which could limit ACE's effectiveness, resulting in performance comparable to pooled scores. Since pooled scores can be seen as a naïve version of ACE, this result implies that when no admissible space exists, pooled scores may serve as a practical alternative.
%a reasonable outcome considering the lack of significant multimodality in the spaces and our strategy aimed at selecting and ranking spaces based on differences in quality.

\begin{table}[H]
    \centering
        \caption{Quantitative evaluation of different approaches for selecting checkpoints. The report includes Spearman and Kendall rank correlation coefficients $r_s$ and $\tau_B$ between the generated scores and NMI scores, as well as ACC scores.}
    \label{tab:app:dcl}
\begin{tabular}{lllllllll}
\toprule
 & \multicolumn{2}{c}{Silhouette score} & \multicolumn{2}{c}{Silhouette score} & \multicolumn{2}{c}{Davies-Bouldin} & \multicolumn{2}{c}{Calinski-Harabasz} \\
  & \multicolumn{2}{c}{(cosine)} & \multicolumn{2}{c}{(euclidean)} & \multicolumn{2}{c}{index} & \multicolumn{2}{c}{index} \\
 & $r_s$ & $\tau_B$ & $r_s$ & $\tau_B$ & $r_s$ & $\tau_B$ & $r_s$ & $\tau_B$ \\
\midrule
NMI &&&&&&&&\\
\hline
Paired score & -0.75 & -0.59 & -0.74 & -0.57 & -0.79 & -0.64 & -0.82 & -0.68 \\
Pooled score & 0.43 & 0.34 & 0.43 & 0.34 & 0.29 & 0.21 & 0.52 & 0.42 \\
\textbf{ACE} & 0.40 & 0.29 & 0.40 & 0.29 & 0.39 & 0.25 & 0.52 & 0.37 \\
\hline
ACC &&&&&&&&\\
\hline
Paired score & -0.75 & -0.58 & -0.74 & -0.56 & -0.80 & -0.66 & -0.82 & -0.69 \\
Pooled score & 0.43 & 0.32 & 0.43 & 0.32 & 0.28 & 0.19 & 0.52 & 0.40 \\
\textbf{ACE} & 0.40 & 0.28 & 0.40 & 0.28 & 0.40 & 0.25 & 0.52 & 0.37 \\
\bottomrule
\end{tabular}

\end{table}

%%%%%%%%%%%%%%%%%%%%%%%%%%
\begin{figure}[H]
  \centering
  
\subfigure[Checkpoint 0]{\includegraphics[width=0.23\linewidth]{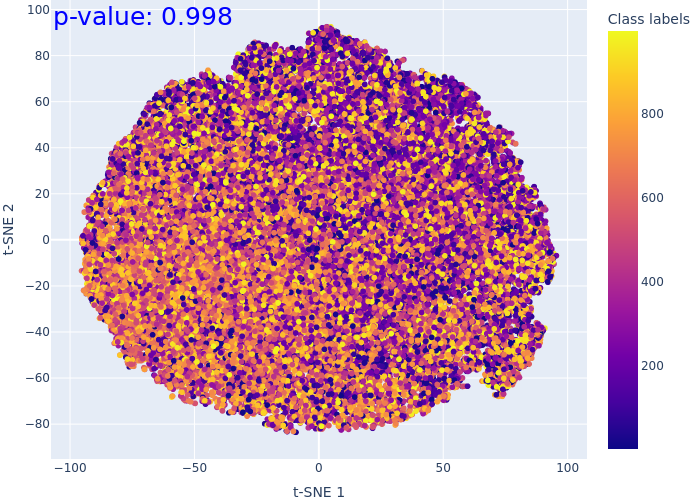}}
\subfigure[Checkpoint 1]{\includegraphics[width=0.23\linewidth]{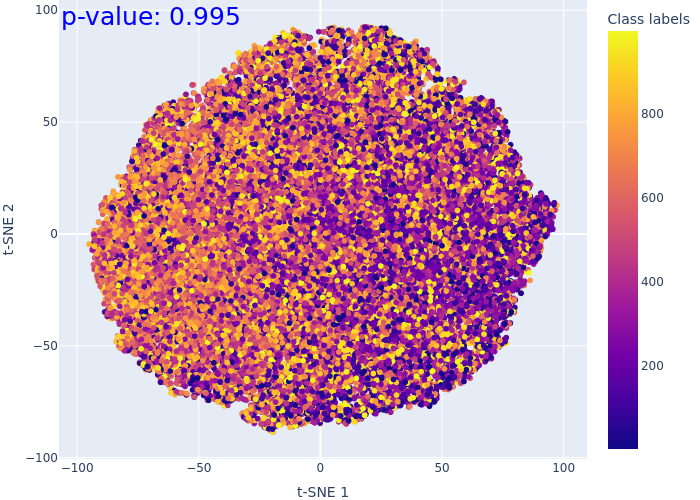}}
\subfigure[Checkpoint 2]{\includegraphics[width=0.23\linewidth]{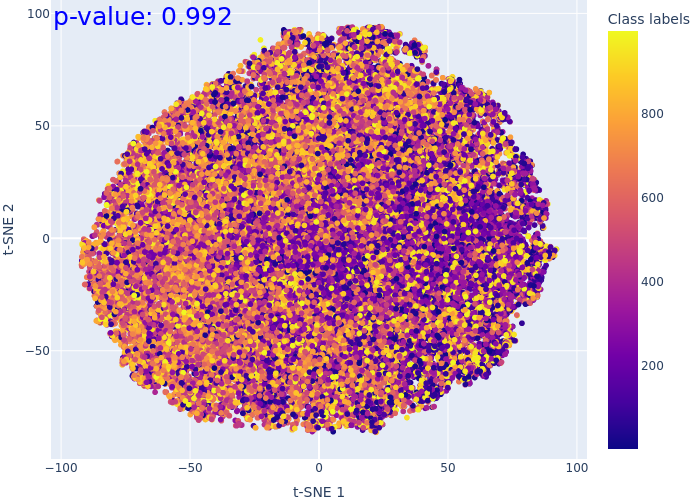}}
\subfigure[Checkpoint 3]{\includegraphics[width=0.23\linewidth]{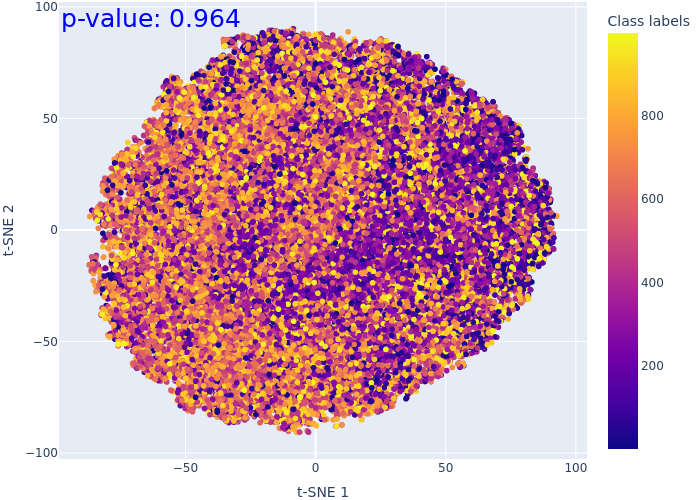}}
\subfigure[Checkpoint 4]{\includegraphics[width=0.23\linewidth]{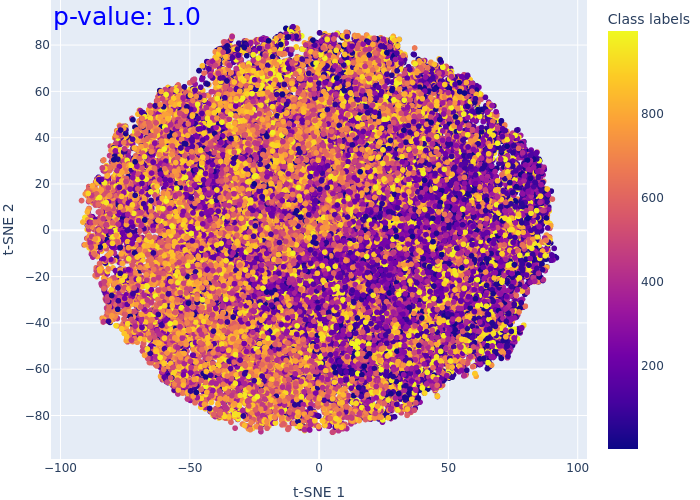}}
\subfigure[Checkpoint 5]{\includegraphics[width=0.23\linewidth]{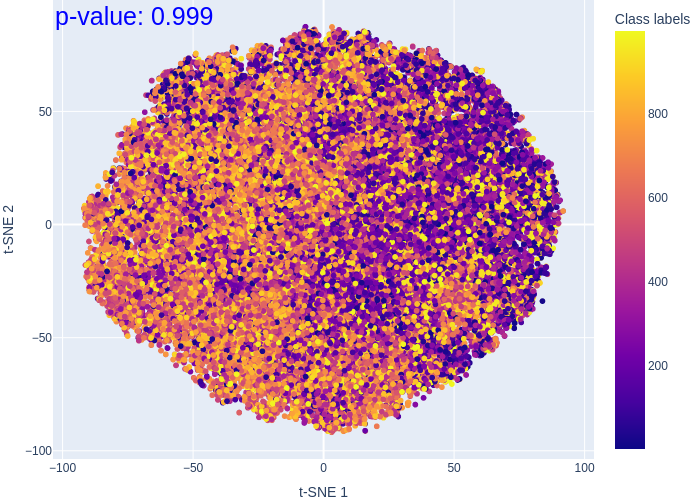}}
\subfigure[Checkpoint 6]{\includegraphics[width=0.23\linewidth]{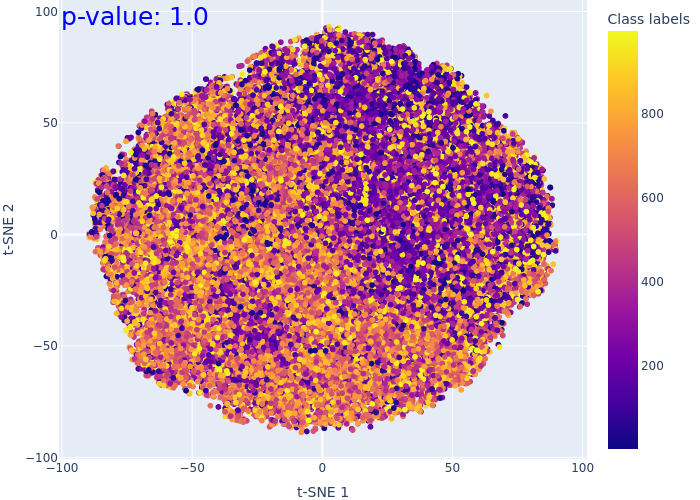}}
\subfigure[Checkpoint 7]{\includegraphics[width=0.23\linewidth]{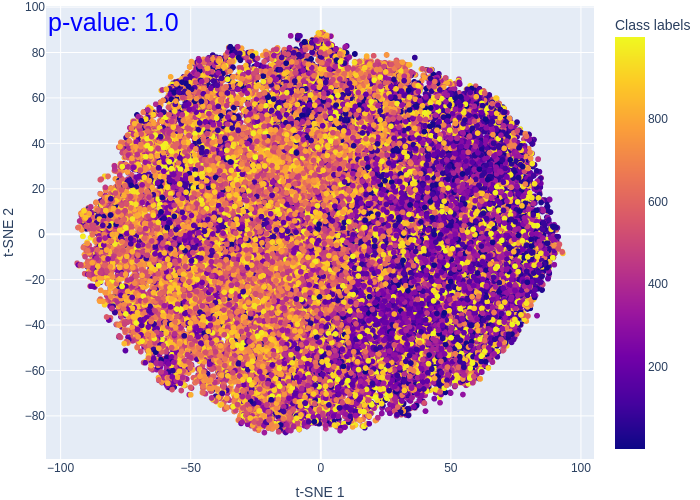}}
\subfigure[Checkpoint 8]{\includegraphics[width=0.23\linewidth]{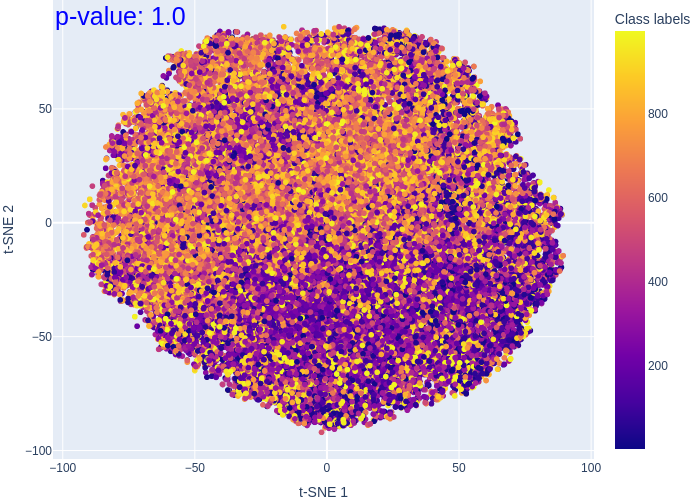}}
\subfigure[Checkpoint 9]{\includegraphics[width=0.23\linewidth]{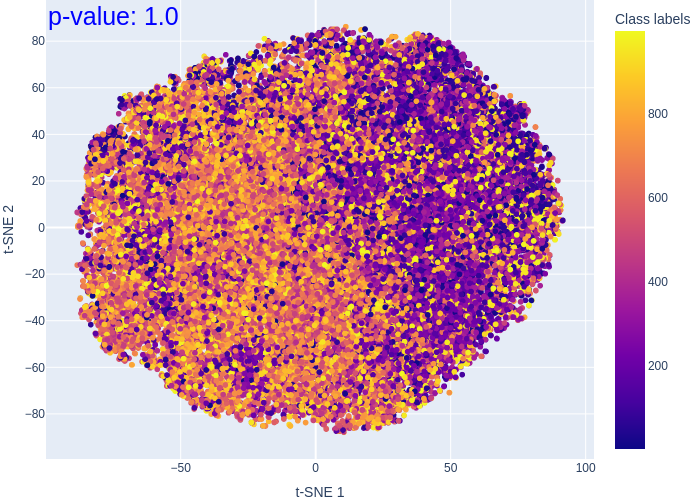}}
\subfigure[Checkpoint 10]{\includegraphics[width=0.23\linewidth]{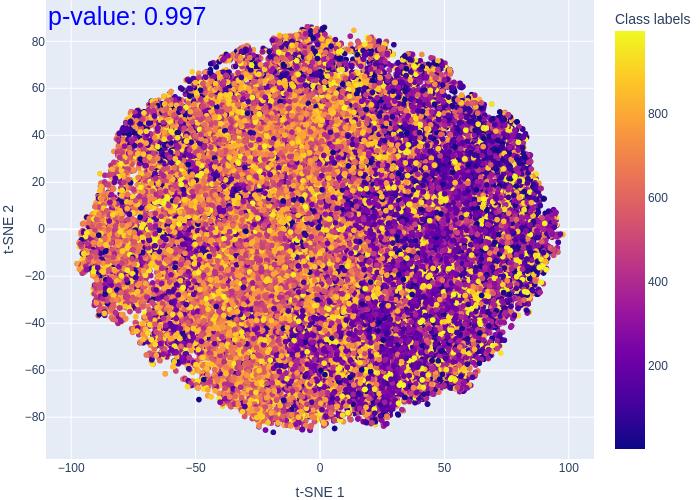}}
\subfigure[Checkpoint 11]{\includegraphics[width=0.23\linewidth]{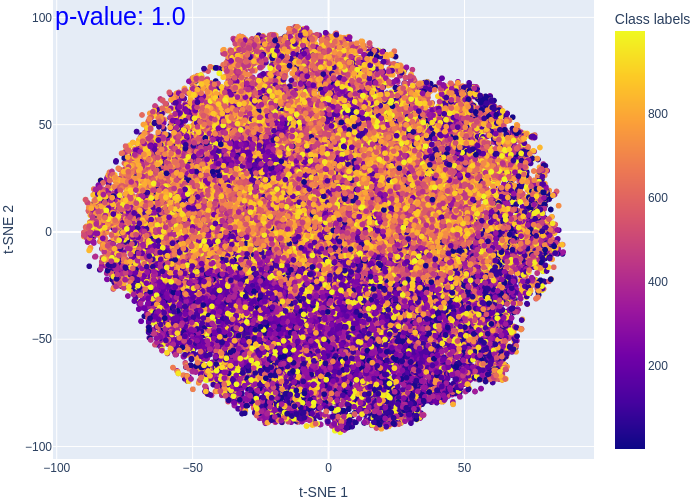}}
\subfigure[Checkpoint 12]{\includegraphics[width=0.23\linewidth]{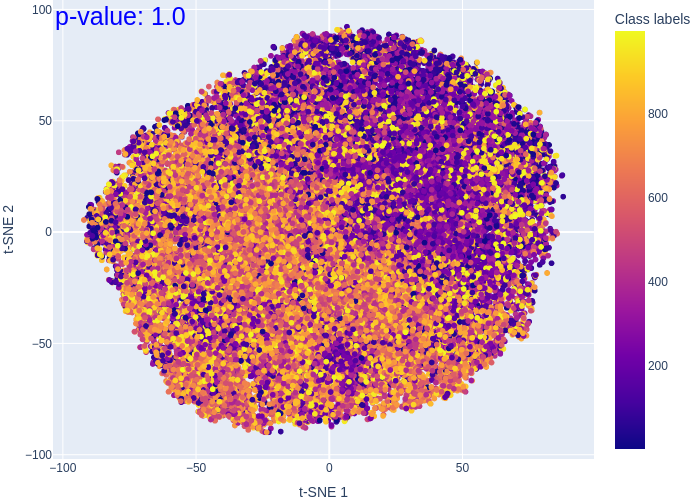}}
\subfigure[Checkpoint 13]{\includegraphics[width=0.23\linewidth]{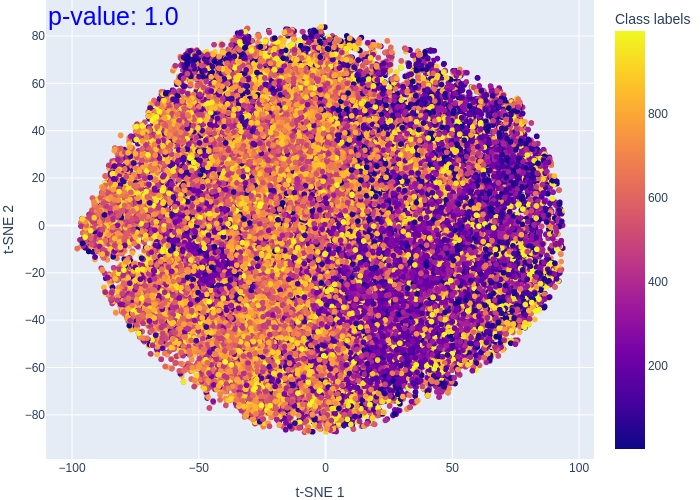}}
\subfigure[Checkpoint 14]{\includegraphics[width=0.23\linewidth]{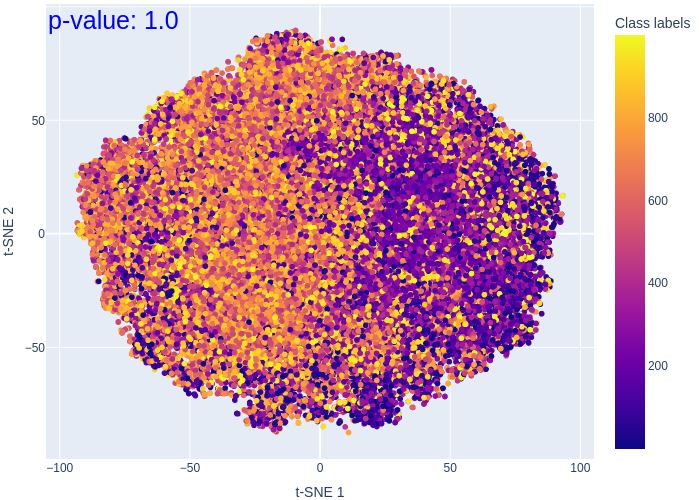}}
\subfigure[Checkpoint 15]{\includegraphics[width=0.23\linewidth]{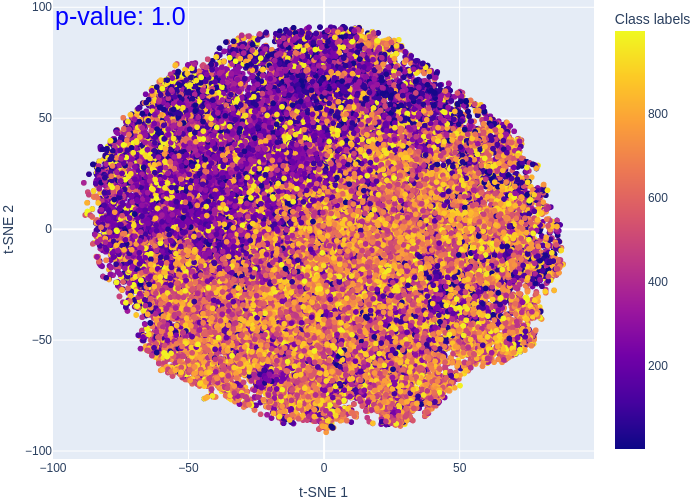}}
\subfigure[Checkpoint 16]{\includegraphics[width=0.23\linewidth]{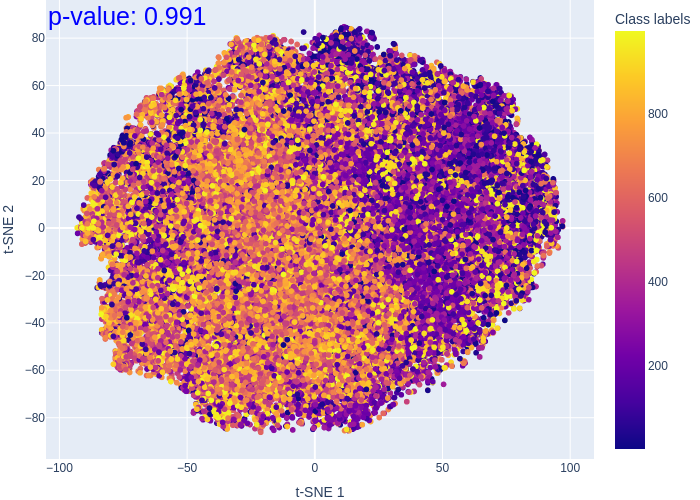}}
\subfigure[Checkpoint 17]{\includegraphics[width=0.23\linewidth]{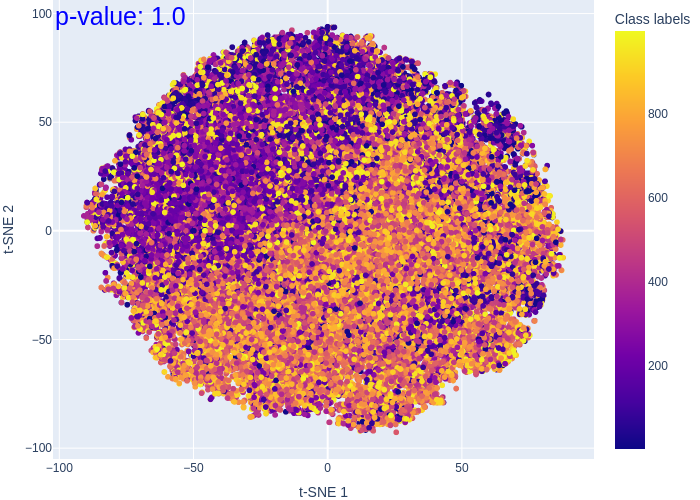}}
\subfigure[Checkpoint 18]{\includegraphics[width=0.23\linewidth]{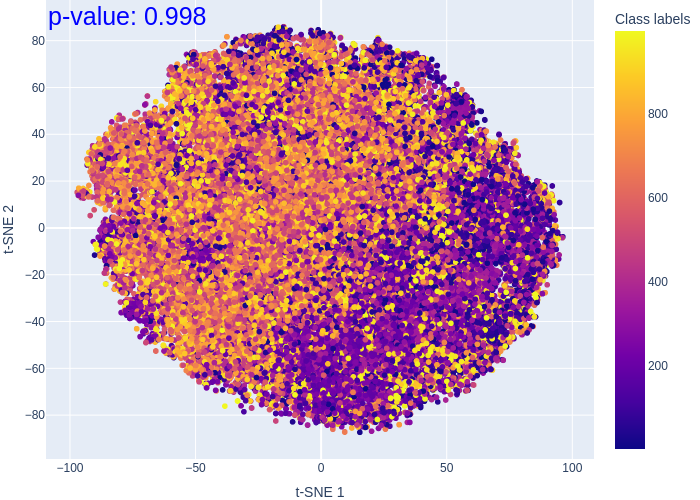}}
\subfigure[Checkpoint 19]{\includegraphics[width=0.23\linewidth]{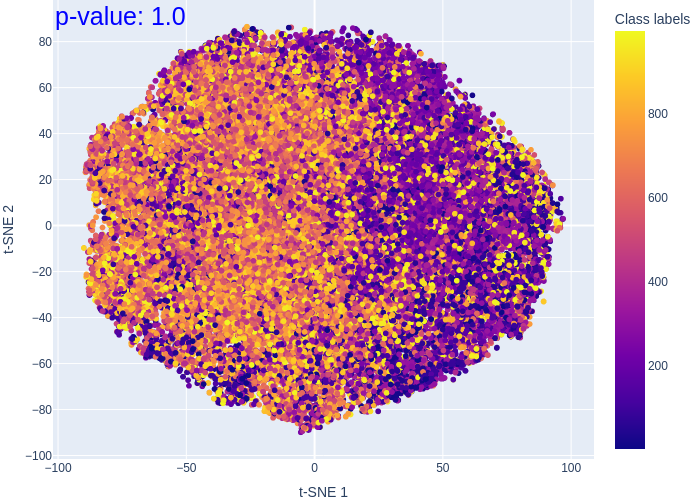}}
\caption{t-SNE visualization illustrating all the embedding spaces obtained in the experiment of checkpoint selection. Each data point in the visualizations is assigned a color corresponding to its true cluster label. The p-value of the Dip test for each space is annotated in the upper-left corner of each subfigure.}
  \label{fig:app:dcl}
\end{figure}

%%%%%%%%%%%%%%%%%%%%%%%%%%%%%%%%%%%%%%%%%%%%%%%%%%%%%%%%%%%%%%%%%%%%%%%%%%%%%%%%%%%%%%%%%%%%%%%%%%%%
%%%%%%%%%%%%%%%%%%%%%%%%%%%%%%%%%%%%%%%%%%%%%%%%%%%%%%%%%%%%%%%%%%%%%%%%%%%%%%%%%%%%%%%%%%%%%%%%%%%%
%%%%%%%%%%%%%%%%%%%%%%%% ABLATION %%%%%%%%%%%%%%%%%%%%%%%%%%%%%%%%%%%%%%
%%%%%%%%%%%%%%%%%%%%%%%%%%%%%%%%%%%%%%%%%%%%%%%%%%%%%%%%%%%%%%%%%%%%%%%%%%%%%%%%%%%%%%%%%%%%%%%%%%%%
%%%%%%%%%%%%%%%%%%%%%%%%%%%%%%%%%%%%%%%%%%%%%%%%%%%%%%%%%%%%%%%%%%%%%%%%%%%%%%%%%%%%%%%%%%%%%%%%%%%%
\clearpage
\newpage
\section{Perturbation Analysis} \label{app:exp:ab}
In this section, we perform a series of perturbation analyses to assess ACE's sensitivity to algorithmic modifications, such as removing Step 1, adjusting parameter values, or using an alternative link analysis algorithm. Specifically, we examine the impact of conducting the Dip test, varying the threshold ($\alpha_{edges}$) in link analysis, applying different link analysis algorithms, and exploring alternative density-based methods. Additionally, we introduce a comparative study that includes the outlier space, differing from our existing approach, which excludes outlier spaces after Phase 1 in stage-wise grouping.

\subsection{Dip test} \label{app:exp:ab:dip}
The Dip test functions as a filtering mechanism in our approach, identifying spaces that exhibit multimodality or clustering behavior.  In the main text, both ACE and pooled scores reported are computed after applying Step 1 of Algorithm \ref{Algorithm1}, namely, the Dip test. To evaluate its impact on model performance, we compare ACE scores and pooled scores with and without the Dip test. Tables \ref{tab:abs:nmi:dip0} and \ref{tab:abs:acc:dip0} present a comparative analysis for Application 1, assessing rank correlation with NMI and ACC, while Tables \ref{tab:abs:nmi:dip1} and \ref{tab:abs:acc:dip1} extend this evaluation to Application 2. Our observations indicate that the application of the Dip test tends to enhance the performance of ACE in certain tasks, while its impact on the pooled score is relatively marginal. Specifically, ACE exhibits improvements (compared to ACE without the Dip test) in hyperparameter tuning for Davies-Bouldin index. Additionally, notable improvements are observed in Application 2 for JULE (with respect to the Davies-Bouldin index and Silhouette score). These improvements can be attributed to ACE’s reliance on the quality of retained spaces. Since ACE determines rankings and scores based on voting within retained spaces, the Dip test helps refine the selection, ultimately enhancing its effectiveness. In contrast, the pooled score, which averages over all retained spaces, is less affected by the Dip test. In summary, our findings suggest that the Dip test contributes to the effectiveness of ACE in specific tasks, while its impact on the pooled score remains limited. \\

\begin{table*}[htbp!]
\centering
\caption{Perturbation analyses of the Dip test for Application 1. $r_s$ and $\tau_B$ between the generated scores and NMI scores are reported. Empty cells and dash marks ($-$) indicate cases where the result is either missing or impractical to obtain.}
\resizebox{\textwidth}{!}{
\begin{tabular}{lllllllllllllllllll}
\toprule
{} & \multicolumn{2}{l}{USPS} & \multicolumn{2}{l}{YTF} & \multicolumn{2}{l}{FRGC} & \multicolumn{2}{l}{MNIST-test} & \multicolumn{2}{l}{CMU-PIE} & \multicolumn{2}{l}{UMist} & \multicolumn{2}{l}{COIL-20} & \multicolumn{2}{l}{COIL-100} & \multicolumn{2}{l}{Average} \\
{} & $r_s$ & $\tau_B$ & $r_s$ & $\tau_B$ & $r_s$ & $\tau_B$ &      $r_s$ & $\tau_B$ &   $r_s$ & $\tau_B$ & $r_s$ & $\tau_B$ &   $r_s$ & $\tau_B$ &    $r_s$ & $\tau_B$ &   $r_s$ & $\tau_B$ \\
\midrule
\hline 
 \multicolumn{19}{c}{JULE: Calinski-Harabasz index} \\
 \hline 
Paired score      &  0.17 &     0.13 &  0.52 &     0.40 & -0.13 &    -0.10 &       0.49 &     0.34 &   -0.13 &    -0.08 &  0.70 &     0.50 &    0.53 &     0.38 &     0.20 &     0.19 &    0.29 &     0.22 \\
Pooled score (w/o. Dip test) &  0.85 &     0.68 &  0.91 &     0.79 &  0.31 &     0.23 &       0.82 &     0.67 &    0.90 &     0.77 &  0.63 &     0.44 &    0.61 &     0.46 &     0.91 &     0.76 &    0.74 &     0.60 \\
Pooled score           &  0.84 &     0.68 &  0.91 &     0.79 &  0.29 &     0.22 &       0.82 &     0.67 &    0.94 &     0.82 &  0.81 &     0.60 &    0.62 &     0.47 &     0.89 &     0.73 &    0.77 &     0.62 \\
\textbf{ACE} (w/o .Dip test)  &  0.80 &     0.63 &  0.90 &     0.73 &  0.42 &     0.30 &       0.86 &     0.70 &    0.98 &     0.93 &  0.71 &     0.51 &    0.92 &     0.76 &     0.92 &     0.79 &    0.81 &     0.67 \\
\textbf{ACE}    &  0.80 &     0.63 &  0.90 &     0.73 &  0.39 &     0.26 &       0.87 &     0.71 &    0.98 &     0.90 &  0.81 &     0.61 &    0.60 &     0.45 &     0.95 &     0.82 &    0.79 &     0.64 \\
\hline 
 \multicolumn{19}{c}{JULE: Davies-Bouldin index} \\
 \hline 
Paired score      & -0.10 &    -0.03 & -0.32 &    -0.21 & -0.08 &    -0.05 &      -0.13 &    -0.06 &    0.26 &     0.20 &  0.62 &     0.44 &    0.61 &     0.42 &     0.43 &     0.35 &    0.16 &     0.13 \\
Pooled score (w/o. Dip test) & -0.26 &    -0.13 & -0.46 &    -0.34 &  0.12 &     0.08 &      -0.15 &    -0.06 &    0.92 &     0.78 & -0.35 &    -0.24 &   -0.24 &    -0.17 &    -0.46 &    -0.35 &   -0.11 &    -0.05 \\
Pooled score           & -0.26 &    -0.12 & -0.46 &    -0.34 &  0.11 &     0.07 &      -0.16 &    -0.07 &    0.92 &     0.78 &  0.30 &     0.20 &   -0.25 &    -0.17 &    -0.46 &    -0.35 &   -0.03 &    0.00 \\
\textbf{ACE} (w/o .Dip test)  & -0.08 &    -0.02 & -0.30 &    -0.21 &  0.22 &     0.16 &       0.73 &     0.55 &    0.03 &    -0.01 &  0.74 &     0.54 &    0.29 &     0.26 &    -0.49 &    -0.39 &    0.14 &     0.11 \\
\textbf{ACE}    & -0.08 &    -0.02 & -0.30 &    -0.21 &  0.22 &     0.16 &       0.73 &     0.55 &    0.10 &     0.06 &  0.38 &     0.27 &    0.23 &     0.22 &     0.48 &     0.33 &    0.22 &     0.17 \\
\hline 
 \multicolumn{19}{c}{JULE: Silhouette score (Cosine distance)} \\
 \hline 
Paired score      &  0.28 &     0.22 &  0.73 &     0.56 &  0.09 &     0.06 &       0.63 &     0.47 &    0.50 &     0.36 &  0.71 &     0.50 &    0.68 &     0.50 &     0.74 &     0.54 &    0.54 &     0.40 \\
Pooled score (w/o. Dip test) &  0.71 &     0.58 &  0.93 &     0.81 &  0.41 &     0.28 &       0.79 &     0.64 &    0.95 &     0.84 &  0.58 &     0.39 &    0.26 &     0.16 &     0.69 &     0.53 &    0.66 &     0.53 \\
Pooled score           &  0.70 &     0.56 &  0.93 &     0.81 &  0.40 &     0.27 &       0.79 &     0.64 &    0.95 &     0.85 &  0.77 &     0.56 &    0.27 &     0.16 &     0.68 &     0.52 &    0.69 &     0.55 \\
\textbf{ACE} (w/o .Dip test)  &  0.89 &     0.73 &  0.93 &     0.83 &  0.52 &     0.35 &       0.81 &     0.66 &    0.99 &     0.94 &  0.83 &     0.65 &    0.44 &     0.38 &     0.91 &     0.77 &    0.79 &     0.66 \\
\textbf{ACE}    &  0.89 &     0.73 &  0.93 &     0.83 &  0.52 &     0.35 &       0.81 &     0.66 &    0.99 &     0.93 &  0.79 &     0.59 &    0.44 &     0.38 &     0.92 &     0.78 &    0.79 &     0.66 \\
\hline 
 \multicolumn{19}{c}{JULE: Silhouette score (Euclidean distance)} \\
 \hline 
Paired score      &  0.27 &     0.20 &  0.72 &     0.55 &  0.04 &     0.03 &       0.56 &     0.41 &    0.42 &     0.30 &  0.70 &     0.50 &    0.64 &     0.46 &     0.55 &     0.41 &    0.49 &     0.36 \\
Pooled score (w/o. Dip test) &  0.70 &     0.57 &  0.90 &     0.77 &  0.41 &     0.28 &       0.78 &     0.63 &    0.95 &     0.84 &  0.64 &     0.43 &    0.25 &     0.16 &     0.71 &     0.54 &    0.67 &     0.53 \\
Pooled score           &  0.71 &     0.58 &  0.90 &     0.77 &  0.41 &     0.28 &       0.78 &     0.63 &    0.96 &     0.85 &  0.79 &     0.57 &    0.26 &     0.16 &     0.70 &     0.54 &    0.69 &     0.55 \\
\textbf{ACE} (w/o .Dip test)  &  0.88 &     0.72 &  0.89 &     0.75 &  0.42 &     0.28 &       0.81 &     0.65 &    0.98 &     0.92 &  0.88 &     0.70 &    0.41 &     0.36 &     0.91 &     0.78 &    0.77 &     0.65 \\
\textbf{ACE}    &  0.88 &     0.72 &  0.89 &     0.75 &  0.42 &     0.28 &       0.81 &     0.65 &    0.98 &     0.90 &  0.88 &     0.70 &    0.41 &     0.36 &     0.92 &     0.78 &    0.77 &     0.64 \\
\hline 
 \multicolumn{19}{c}{DEPICT: Calinski-Harabasz index} \\
 \hline 
Paired score      &  0.76 &     0.57 &  0.44 &     0.26 &  0.76 &     0.57 &       0.89 &     0.72 &    0.49 &     0.44 &    &       &      &       &       &       &    0.67 &     0.51 \\
Pooled score (w/o. Dip test) &  0.96 &     0.84 &  0.53 &     0.41 &  0.90 &     0.77 &       0.96 &     0.87 &    0.73 &     0.59 &    &       &      &       &       &       &    0.82 &     0.70 \\
Pooled score           &  0.96 &     0.83 &  0.53 &     0.41 &  0.90 &     0.77 &       0.96 &     0.87 &    0.61 &     0.56 &    &       &      &       &       &       &    0.79 &     0.69 \\
\textbf{ACE} (w/o .Dip test)  &  0.91 &     0.77 &  0.56 &     0.44 &  0.94 &     0.82 &       0.96 &     0.87 &    0.96 &     0.88 &    &       &      &       &       &       &    0.87 &     0.75 \\
\textbf{ACE}    &  0.91 &     0.77 &  0.56 &     0.44 &  0.94 &     0.82 &       0.96 &     0.87 &    0.96 &     0.87 &    &       &      &       &       &       &    0.87 &     0.75 \\
\hline 
 \multicolumn{19}{c}{DEPICT: Davies-Bouldin index} \\
 \hline 
Paired score      &  0.81 &     0.59 &  0.45 &     0.31 &  0.90 &     0.74 &       0.89 &     0.72 &    0.63 &     0.59 &    &       &      &       &       &       &    0.73 &     0.59 \\
Pooled score (w/o. Dip test) &  0.95 &     0.84 &  0.49 &     0.35 &  0.65 &     0.50 &       0.50 &     0.36 &    0.23 &     0.06 &    &       &      &       &       &       &    0.56 &     0.42 \\
Pooled score           &  0.96 &     0.88 &  0.49 &     0.35 &  0.64 &     0.48 &       0.43 &     0.32 &   -0.77 &    -0.61 &    &       &      &       &       &       &    0.35 &     0.28 \\
\textbf{ACE} (w/o .Dip test)  &  0.90 &     0.79 &  0.76 &     0.58 &  0.91 &     0.79 &       0.95 &     0.83 &    0.63 &     0.49 &    &       &      &       &       &       &    0.83 &     0.70 \\
\textbf{ACE}    &  0.91 &     0.82 &  0.76 &     0.58 &  0.91 &     0.79 &       0.96 &     0.87 &    0.98 &     0.92 &    &       &      &       &       &       &    0.90 &     0.80 \\
\hline 
 \multicolumn{19}{c}{DEPICT: Silhouette score (Cosine distance)} \\
 \hline 
Paired score      &  0.81 &     0.62 &  0.45 &     0.33 &  0.90 &     0.75 &       0.89 &     0.72 &    0.77 &     0.58 &    &       &      &       &       &       &    0.76 &     0.60 \\
Pooled score (w/o. Dip test) &  0.96 &     0.83 &  0.68 &     0.56 &  0.94 &     0.82 &       0.95 &     0.87 &    0.95 &     0.86 &    &       &      &       &       &       &    0.90 &     0.79 \\
Pooled score           &  0.96 &     0.86 &  0.68 &     0.56 &  0.94 &     0.82 &       0.97 &     0.90 &    0.93 &     0.79 &    &       &      &       &       &       &    0.90 &     0.78 \\
\textbf{ACE} (w/o .Dip test)  &  0.97 &     0.90 &  0.71 &     0.56 &  0.94 &     0.82 &       0.98 &     0.91 &    0.95 &     0.84 &    &       &      &       &       &       &    0.91 &     0.80 \\
\textbf{ACE}    &  0.97 &     0.90 &  0.71 &     0.56 &  0.94 &     0.82 &       0.97 &     0.90 &    0.94 &     0.83 &    &       &      &       &       &       &    0.91 &     0.80 \\
\hline 
 \multicolumn{19}{c}{DEPICT: Silhouette score (Euclidean distance)} \\
 \hline 
Paired score      &  0.73 &     0.50 &  0.47 &     0.36 &  0.79 &     0.65 &       0.86 &     0.69 &    0.59 &     0.52 &    &       &      &       &       &       &    0.69 &     0.54 \\
Pooled score (w/o. Dip test) &  0.96 &     0.84 &  0.65 &     0.53 &  0.94 &     0.82 &       0.97 &     0.90 &    0.95 &     0.86 &    &       &      &       &       &       &    0.89 &     0.79 \\
Pooled score           &  0.96 &     0.86 &  0.65 &     0.53 &  0.94 &     0.82 &       0.97 &     0.90 &    0.92 &     0.75 &    &       &      &       &       &       &    0.89 &     0.77 \\
\textbf{ACE} (w/o .Dip test)  &  0.92 &     0.80 &  0.65 &     0.50 &  0.95 &     0.83 &       0.98 &     0.90 &    0.95 &     0.83 &    &       &      &       &       &       &    0.89 &     0.77 \\
\textbf{ACE}    &  0.97 &     0.88 &  0.65 &     0.50 &  0.95 &     0.83 &       0.98 &     0.90 &    0.94 &     0.82 &    &       &      &       &       &       &    0.90 &     0.79 \\
\bottomrule
\end{tabular}
}
 \label{tab:abs:nmi:dip0}
\end{table*}

\begin{table*}[htbp!]
\centering
\caption{Perturbation analyses of the Dip test for Application 1. $r_s$ and $\tau_B$ between the generated scores and ACC scores are reported. Empty cells and dash marks ($-$) indicate cases where the result is either missing or impractical to obtain.}
\resizebox{\textwidth}{!}{
\begin{tabular}{lllllllllllllllllll}
\toprule
{} & \multicolumn{2}{l}{USPS} & \multicolumn{2}{l}{YTF} & \multicolumn{2}{l}{FRGC} & \multicolumn{2}{l}{MNIST-test} & \multicolumn{2}{l}{CMU-PIE} & \multicolumn{2}{l}{UMist} & \multicolumn{2}{l}{COIL-20} & \multicolumn{2}{l}{COIL-100} & \multicolumn{2}{l}{Average} \\
{} & $r_s$ & $\tau_B$ & $r_s$ & $\tau_B$ & $r_s$ & $\tau_B$ &      $r_s$ & $\tau_B$ &   $r_s$ & $\tau_B$ & $r_s$ & $\tau_B$ &   $r_s$ & $\tau_B$ &    $r_s$ & $\tau_B$ &   $r_s$ & $\tau_B$ \\
\midrule
\hline 
 \multicolumn{19}{c}{JULE: Calinski-Harabasz index} \\
 \hline 
Paired score      &  0.04 &     0.05 &  0.39 &     0.27 & -0.26 &    -0.18 &       0.31 &     0.21 &   -0.20 &    -0.12 &  0.64 &     0.45 &    0.57 &     0.40 &     0.09 &     0.08 &    0.20 &     0.14 \\
Pooled score (w/o. Dip test) &  0.92 &     0.79 &  0.78 &     0.61 &  0.30 &     0.21 &       0.91 &     0.77 &    0.91 &     0.78 &  0.65 &     0.47 &    0.57 &     0.42 &     0.91 &     0.78 &    0.74 &     0.60 \\
Pooled score           &  0.91 &     0.78 &  0.78 &     0.61 &  0.30 &     0.21 &       0.91 &     0.77 &    0.95 &     0.83 &  0.81 &     0.60 &    0.58 &     0.43 &     0.90 &     0.75 &    0.77 &     0.62 \\
\textbf{ACE} (w/o .Dip test)  &  0.90 &     0.77 &  0.73 &     0.54 &  0.59 &     0.44 &       0.95 &     0.81 &    0.97 &     0.89 &  0.67 &     0.49 &    0.89 &     0.72 &     0.88 &     0.74 &    0.82 &     0.68 \\
\textbf{ACE}    &  0.90 &     0.77 &  0.73 &     0.54 &  0.49 &     0.36 &       0.95 &     0.82 &    0.97 &     0.87 &  0.81 &     0.61 &    0.57 &     0.40 &     0.93 &     0.81 &    0.79 &     0.65 \\
\hline 
 \multicolumn{19}{c}{JULE: Davies-Bouldin index} \\
 \hline 
Paired score      & -0.27 &    -0.15 & -0.14 &    -0.09 & -0.23 &    -0.14 &      -0.35 &    -0.19 &    0.20 &     0.16 &  0.53 &     0.36 &    0.63 &     0.44 &     0.33 &     0.26 &    0.09 &     0.08 \\
Pooled score (w/o. Dip test) & -0.49 &    -0.21 & -0.35 &    -0.23 &  0.49 &     0.36 &      -0.35 &    -0.20 &    0.89 &     0.76 & -0.47 &    -0.34 &   -0.30 &    -0.22 &    -0.48 &    -0.34 &   -0.13 &    -0.05 \\
Pooled score           & -0.49 &    -0.20 & -0.35 &    -0.23 &  0.48 &     0.36 &      -0.35 &    -0.21 &    0.89 &     0.75 &  0.17 &     0.11 &   -0.29 &    -0.22 &    -0.48 &    -0.34 &   -0.05 &     0.00 \\
\textbf{ACE} (w/o .Dip test)  & -0.30 &    -0.09 & -0.07 &    -0.07 &  0.53 &     0.38 &       0.79 &     0.64 &    0.01 &    -0.04 &  0.66 &     0.45 &    0.27 &     0.23 &    -0.49 &    -0.35 &    0.17 &     0.14 \\
\textbf{ACE}    & -0.30 &    -0.09 & -0.07 &    -0.07 &  0.53 &     0.38 &       0.79 &     0.64 &    0.07 &     0.03 &  0.27 &     0.20 &    0.21 &     0.18 &     0.44 &     0.28 &    0.24 &     0.19 \\
\hline 
 \multicolumn{19}{c}{JULE: Silhouette score (Cosine distance)} \\
 \hline 
Paired score      &  0.17 &     0.14 &  0.59 &     0.41 &  0.07 &     0.06 &       0.47 &     0.33 &    0.45 &     0.33 &  0.64 &     0.46 &    0.70 &     0.51 &     0.64 &     0.45 &    0.47 &     0.34 \\
Pooled score (w/o. Dip test) &  0.75 &     0.70 &  0.73 &     0.55 &  0.71 &     0.53 &       0.90 &     0.73 &    0.96 &     0.87 &  0.57 &     0.38 &    0.19 &     0.10 &     0.60 &     0.44 &    0.68 &     0.54 \\
Pooled score           &  0.74 &     0.68 &  0.73 &     0.55 &  0.71 &     0.53 &       0.90 &     0.73 &    0.96 &     0.88 &  0.75 &     0.55 &    0.20 &     0.11 &     0.61 &     0.44 &    0.70 &     0.56 \\
\textbf{ACE} (w/o .Dip test)  &  0.96 &     0.85 &  0.74 &     0.55 &  0.82 &     0.65 &       0.92 &     0.78 &    0.99 &     0.94 &  0.80 &     0.61 &    0.41 &     0.32 &     0.81 &     0.65 &    0.81 &     0.67 \\
\textbf{ACE}    &  0.96 &     0.85 &  0.74 &     0.55 &  0.82 &     0.65 &       0.92 &     0.78 &    0.98 &     0.92 &  0.78 &     0.58 &    0.41 &     0.32 &     0.84 &     0.68 &    0.81 &     0.67 \\
\hline 
 \multicolumn{19}{c}{JULE: Silhouette score (Euclidean distance)} \\
 \hline 
Paired score      &  0.14 &     0.12 &  0.54 &     0.39 & -0.08 &    -0.02 &       0.41 &     0.27 &    0.36 &     0.27 &  0.64 &     0.46 &    0.67 &     0.48 &     0.44 &     0.31 &    0.39 &     0.28 \\
Pooled score (w/o. Dip test) &  0.74 &     0.68 &  0.66 &     0.49 &  0.71 &     0.53 &       0.89 &     0.72 &    0.96 &     0.87 &  0.64 &     0.43 &    0.19 &     0.10 &     0.62 &     0.45 &    0.68 &     0.53 \\
Pooled score           &  0.73 &     0.67 &  0.66 &     0.49 &  0.70 &     0.53 &       0.89 &     0.72 &    0.97 &     0.88 &  0.77 &     0.57 &    0.20 &     0.11 &     0.62 &     0.45 &    0.69 &     0.55 \\
\textbf{ACE} (w/o .Dip test)  &  0.93 &     0.78 &  0.63 &     0.48 &  0.71 &     0.53 &       0.92 &     0.78 &    0.99 &     0.94 &  0.86 &     0.68 &    0.39 &     0.30 &     0.81 &     0.66 &    0.78 &     0.64 \\
\textbf{ACE}    &  0.93 &     0.78 &  0.63 &     0.48 &  0.71 &     0.53 &       0.92 &     0.78 &    0.98 &     0.91 &  0.86 &     0.68 &    0.39 &     0.30 &     0.84 &     0.68 &    0.78 &     0.64 \\
\hline 
 \multicolumn{19}{c}{DEPICT: Calinski-Harabasz index} \\
 \hline 
Paired score      &  0.56 &     0.40 &  0.54 &     0.35 &  0.76 &     0.57 &       0.88 &     0.69 &    0.48 &     0.43 &    &       &      &       &       &       &    0.64 &     0.49 \\
Pooled score (w/o. Dip test) &  0.94 &     0.83 &  0.54 &     0.45 &  0.92 &     0.79 &       0.95 &     0.86 &    0.74 &     0.62 &    &       &      &       &       &       &    0.82 &     0.71 \\
Pooled score           &  0.94 &     0.82 &  0.54 &     0.45 &  0.92 &     0.79 &       0.95 &     0.86 &    0.62 &     0.55 &    &       &      &       &       &       &    0.79 &     0.69 \\
\textbf{ACE} (w/o .Dip test)  &  0.82 &     0.72 &  0.61 &     0.45 &  0.91 &     0.82 &       0.97 &     0.91 &    0.98 &     0.91 &    &       &      &       &       &       &    0.86 &     0.76 \\
\textbf{ACE}    &  0.82 &     0.72 &  0.61 &     0.45 &  0.91 &     0.82 &       0.97 &     0.91 &    0.96 &     0.87 &    &       &      &       &       &       &    0.86 &     0.75 \\
\hline 
 \multicolumn{19}{c}{DEPICT: Davies-Bouldin index} \\
 \hline 
Paired score      &  0.61 &     0.42 &  0.48 &     0.32 &  0.92 &     0.74 &       0.88 &     0.69 &    0.62 &     0.56 &    &       &      &       &       &       &    0.70 &     0.55 \\
Pooled score (w/o. Dip test) &  0.93 &     0.80 &  0.40 &     0.28 &  0.65 &     0.50 &       0.45 &     0.32 &    0.24 &     0.07 &    &       &      &       &       &       &    0.53 &     0.39 \\
Pooled score           &  0.95 &     0.84 &  0.40 &     0.28 &  0.64 &     0.48 &       0.38 &     0.28 &   -0.76 &    -0.60 &    &       &      &       &       &       &    0.32 &     0.26 \\
\textbf{ACE} (w/o .Dip test)  &  0.99 &     0.96 &  0.65 &     0.46 &  0.90 &     0.74 &       0.99 &     0.92 &    0.60 &     0.46 &    &       &      &       &       &       &    0.82 &     0.71 \\
\textbf{ACE}    &  0.99 &     0.96 &  0.65 &     0.46 &  0.90 &     0.74 &       0.99 &     0.96 &    0.96 &     0.87 &    &       &      &       &       &       &    0.90 &     0.80 \\
\hline 
 \multicolumn{19}{c}{DEPICT: Silhouette score (Cosine distance)} \\
 \hline 
Paired score      &  0.62 &     0.45 &  0.53 &     0.42 &  0.91 &     0.75 &       0.88 &     0.69 &    0.77 &     0.58 &    &       &      &       &       &       &    0.74 &     0.58 \\
Pooled score (w/o. Dip test) &  0.96 &     0.87 &  0.75 &     0.59 &  0.94 &     0.82 &       0.96 &     0.88 &    0.95 &     0.85 &    &       &      &       &       &       &    0.91 &     0.80 \\
Pooled score           &  0.96 &     0.87 &  0.75 &     0.59 &  0.94 &     0.82 &       0.96 &     0.88 &    0.93 &     0.76 &    &       &      &       &       &       &    0.91 &     0.78 \\
\textbf{ACE} (w/o .Dip test)  &  0.95 &     0.88 &  0.70 &     0.54 &  0.91 &     0.77 &       0.96 &     0.90 &    0.96 &     0.87 &    &       &      &       &       &       &    0.90 &     0.79 \\
\textbf{ACE}    &  0.95 &     0.88 &  0.70 &     0.54 &  0.91 &     0.77 &       0.96 &     0.88 &    0.94 &     0.83 &    &       &      &       &       &       &    0.89 &     0.78 \\
\hline 
 \multicolumn{19}{c}{DEPICT: Silhouette score (Euclidean distance)} \\
 \hline 
Paired score      &  0.52 &     0.33 &  0.57 &     0.45 &  0.80 &     0.62 &       0.85 &     0.65 &    0.59 &     0.48 &    &       &      &       &       &       &    0.67 &     0.51 \\
Pooled score (w/o. Dip test) &  0.95 &     0.86 &  0.72 &     0.57 &  0.94 &     0.82 &       0.96 &     0.88 &    0.95 &     0.85 &    &       &      &       &       &       &    0.91 &     0.80 \\
Pooled score           &  0.94 &     0.84 &  0.72 &     0.57 &  0.94 &     0.82 &       0.96 &     0.88 &    0.92 &     0.75 &    &       &      &       &       &       &    0.90 &     0.77 \\
\textbf{ACE} (w/o .Dip test)  &  0.94 &     0.84 &  0.63 &     0.49 &  0.91 &     0.78 &       0.97 &     0.91 &    0.95 &     0.85 &    &       &      &       &       &       &    0.88 &     0.77 \\
\textbf{ACE}    &  0.95 &     0.87 &  0.63 &     0.49 &  0.91 &     0.78 &       0.97 &     0.91 &    0.95 &     0.84 &    &       &      &       &       &       &    0.88 &     0.78 \\
\bottomrule
\end{tabular}
}
 \label{tab:abs:acc:dip0}
\end{table*}
%\newpage \subsection{Determination of the number of clusters  - Dip Test} 
\begin{table*}[htbp!]
\centering
\caption{Perturbation analyses of the Dip test for Application 2. $r_s$ and $\tau_B$ between the generated scores and NMI scores are reported. Empty cells and dash marks ($-$) indicate cases where the result is either missing or impractical to obtain.}
\resizebox{\textwidth}{!}{
\begin{tabular}{lllllllllllllllllll}
\toprule
{} & \multicolumn{2}{l}{USPS (10)} & \multicolumn{2}{l}{YTF (41)} & \multicolumn{2}{l}{FRGC (20)} & \multicolumn{2}{l}{MNIST-test (10)} & \multicolumn{2}{l}{CMU-PIE (68)} & \multicolumn{2}{l}{UMist (20)} & \multicolumn{2}{l}{COIL-20 (20)} & \multicolumn{2}{l}{COIL-100 (100)} & \multicolumn{2}{l}{Average} \\
{} &      $r_s$ &   $\tau_B$ &      $r_s$ &   $\tau_B$ &       $r_s$ &    $\tau_B$ &           $r_s$ &  $\tau_B$ &        $r_s$ &    $\tau_B$ &      $r_s$ &   $\tau_B$ &        $r_s$ &   $\tau_B$ &          $r_s$ &   $\tau_B$ &   $r_s$ & $\tau_B$ \\
\midrule
\hline 
 \multicolumn{19}{c}{JULE: Calinski-Harabasz index} \\
 \hline 
Paired score      &  0.65 (10) &  0.64 (10) &   0.1 (50) &  0.06 (50) &  -0.93 (15) &  -0.83 (15) &       0.64 (10) &  0.6 (10) &   -0.03 (20) &  -0.02 (20) &  -0.13 (5) &  -0.07 (5) &    0.76 (15) &  0.71 (15) &      0.74 (80) &  0.56 (80) &    0.22 &     0.21 \\
Pooled score (w/o. Dip test) &  0.55 (10) &   0.6 (10) &   0.9 (50) &  0.78 (50) &  -0.87 (15) &  -0.72 (15) &       0.64 (10) &  0.6 (10) &    0.88 (70) &   0.73 (70) &  -0.14 (5) &  -0.11 (5) &    0.74 (15) &  0.64 (15) &      0.72 (80) &  0.64 (80) &    0.43 &     0.40 \\
Pooled score           &  0.65 (10) &  0.64 (10) &   0.9 (50) &  0.78 (50) &  -0.87 (15) &  -0.72 (15) &       0.64 (10) &  0.6 (10) &     0.9 (70) &   0.73 (70) &  -0.14 (5) &  -0.11 (5) &    0.74 (15) &  0.64 (15) &      0.72 (80) &  0.64 (80) &    0.44 &     0.40 \\
\textbf{ACE} (w/o .Dip test)  &  0.65 (10) &  0.64 (10) &  0.93 (50) &  0.83 (50) &  -0.72 (15) &  -0.67 (15) &       0.64 (10) &  0.6 (10) &    0.88 (70) &   0.73 (70) &  -0.13 (5) &  -0.07 (5) &    0.74 (15) &  0.64 (15) &      0.79 (80) &  0.69 (80) &    0.47 &     0.42 \\
\textbf{ACE}    &  0.65 (10) &  0.64 (10) &  0.93 (50) &  0.83 (50) &  -0.72 (15) &  -0.67 (15) &       0.64 (10) &  0.6 (10) &    0.88 (70) &   0.73 (70) &  -0.14 (5) &  -0.11 (5) &    0.74 (15) &  0.64 (15) &      0.79 (80) &  0.69 (80) &    0.47 &     0.42 \\
\hline 
 \multicolumn{19}{c}{JULE: Davies-Bouldin index} \\
 \hline 
Paired score      &  0.54 (15) &  0.38 (15) &  0.15 (50) &  0.17 (50) &  0.85 (45) &  0.67 (45) &       0.43 (10) &  0.29 (10) &   0.78 (100) &  0.56 (100) &  -0.08 (45) &   0.02 (45) &   -0.26 (40) &  -0.14 (40) &      -0.9 (20) &  -0.78 (20) &    0.19 &     0.15 \\
Pooled score (w/o. Dip test) &  0.88 (15) &  0.73 (15) &  0.83 (50) &  0.67 (50) &  0.82 (40) &  0.61 (40) &       0.81 (10) &  0.64 (10) &    0.82 (90) &   0.64 (90) &   0.12 (50) &   0.11 (50) &   -0.67 (50) &   -0.5 (50) &     -0.92 (20) &  -0.82 (20) &    0.34 &     0.26 \\
Pooled score           &  0.98 (15) &  0.91 (15) &  0.83 (50) &  0.67 (50) &  0.82 (40) &  0.61 (40) &       0.79 (10) &   0.6 (10) &    0.82 (90) &   0.64 (90) &  -0.21 (45) &  -0.02 (45) &   -0.76 (50) &  -0.57 (50) &     -0.92 (20) &  -0.82 (20) &    0.29 &     0.25 \\
\textbf{ACE} (w/o .Dip test)  &  0.06 (30) &  0.07 (30) &  0.83 (50) &  0.67 (50) &  0.87 (40) &  0.72 (40) &       0.65 (25) &  0.51 (25) &    0.99 (70) &   0.96 (70) &   0.12 (50) &   0.11 (50) &   -0.67 (50) &   -0.5 (50) &     -0.92 (20) &  -0.82 (20) &    0.24 &     0.22 \\
\textbf{ACE}    &  0.98 (15) &  0.91 (15) &  0.83 (50) &  0.67 (50) &  0.87 (40) &  0.72 (40) &       0.79 (10) &   0.6 (10) &    0.85 (90) &   0.69 (90) &  -0.21 (45) &  -0.02 (45) &   -0.69 (50) &  -0.57 (50) &     -0.94 (20) &  -0.82 (20) &    0.31 &     0.27 \\
\hline 
 \multicolumn{19}{c}{JULE: Silhouette score (Cosine distance)} \\
 \hline 
Paired score      &  0.99 (10) &  0.96 (10) &   0.3 (50) &  0.22 (50) &  0.72 (25) &  0.61 (25) &       0.87 (10) &  0.69 (10) &    0.98 (70) &  0.91 (70) &  -0.07 (45) &   0.07 (45) &    0.52 (25) &   0.36 (25) &     0.39 (200) &   0.2 (200) &    0.59 &     0.50 \\
Pooled score (w/o. Dip test) &  0.98 (10) &  0.91 (10) &  0.98 (50) &  0.94 (50) &  0.68 (45) &  0.56 (45) &       0.93 (10) &  0.82 (10) &    0.98 (70) &  0.91 (70) &   0.21 (45) &   0.16 (45) &    0.36 (25) &   0.21 (25) &     0.47 (200) &  0.33 (200) &    0.70 &     0.60 \\
Pooled score           &  0.95 (10) &  0.87 (10) &  0.98 (50) &  0.94 (50) &  0.68 (45) &  0.56 (45) &       0.96 (10) &  0.87 (10) &    0.98 (70) &  0.91 (70) &  -0.07 (45) &  -0.02 (45) &    0.71 (20) &   0.57 (20) &     0.41 (200) &  0.24 (200) &    0.70 &     0.62 \\
\textbf{ACE} (w/o .Dip test)  &  0.92 (10) &  0.82 (10) &  0.98 (50) &  0.94 (50) &   0.7 (45) &  0.61 (45) &       0.99 (10) &  0.96 (10) &    0.98 (70) &  0.91 (70) &   -0.48 (5) &   -0.38 (5) &   -0.24 (45) &  -0.14 (45) &     0.47 (200) &  0.33 (200) &    0.54 &     0.51 \\
\textbf{ACE}    &  0.95 (10) &  0.87 (10) &  0.98 (50) &  0.94 (50) &   0.7 (45) &  0.61 (45) &       0.96 (10) &  0.87 (10) &    0.98 (70) &  0.91 (70) &  -0.07 (45) &  -0.02 (45) &    0.74 (20) &    0.5 (20) &     0.46 (180) &  0.33 (180) &    0.71 &     0.63 \\
\hline 
 \multicolumn{19}{c}{JULE: Silhouette score (Euclidean distance)} \\
 \hline 
Paired score      &  0.85 (10) &  0.73 (10) &  0.33 (50) &  0.28 (50) &  0.72 (25) &  0.61 (25) &       0.88 (10) &  0.69 (10) &    0.96 (80) &  0.87 (80) &  0.07 (45) &  0.16 (45) &    0.55 (25) &  0.43 (25) &     0.44 (200) &  0.29 (200) &    0.60 &     0.51 \\
Pooled score (w/o. Dip test) &  0.98 (10) &  0.91 (10) &  0.97 (50) &  0.89 (50) &  0.68 (45) &  0.56 (45) &       0.93 (10) &  0.82 (10) &    0.98 (70) &  0.91 (70) &  0.21 (45) &  0.16 (45) &    0.36 (25) &  0.21 (25) &     0.47 (200) &  0.33 (200) &    0.70 &     0.60 \\
Pooled score           &  0.95 (10) &  0.87 (10) &  0.97 (50) &  0.89 (50) &  0.68 (45) &  0.56 (45) &       0.95 (10) &  0.82 (10) &    0.98 (70) &  0.91 (70) &  0.14 (45) &  0.11 (45) &    0.76 (25) &  0.57 (25) &     0.47 (200) &  0.33 (200) &    0.74 &     0.63 \\
\textbf{ACE} (w/o .Dip test)  &  0.79 (10) &  0.73 (10) &  0.98 (50) &  0.94 (50) &  0.78 (45) &  0.67 (45) &       0.92 (10) &  0.82 (10) &    0.99 (70) &  0.96 (70) &  -0.69 (5) &  -0.51 (5) &    0.24 (25) &  0.14 (25) &     0.43 (160) &  0.29 (160) &    0.55 &     0.50 \\
\textbf{ACE}    &  0.95 (10) &  0.87 (10) &  0.98 (50) &  0.94 (50) &  0.78 (45) &  0.67 (45) &       0.95 (10) &  0.82 (10) &    0.98 (70) &  0.91 (70) &  0.14 (45) &  0.11 (45) &    0.71 (25) &  0.43 (25) &     0.47 (200) &  0.33 (200) &    0.74 &     0.64 \\
\hline 
 \multicolumn{19}{c}{DEPICT: Calinski-Harabasz index} \\
 \hline 
Paired score      &  0.46 (5) &  0.6 (5) &  -0.99 (5) &  -0.96 (5) &  -0.85 (10) &  -0.72 (10) &        0.44 (5) &  0.56 (5) &   -0.92 (10) &  -0.82 (10) &         &       &           &       &             &       &   -0.37 &    -0.27 \\
Pooled score (w/o. Dip test) &  0.46 (5) &  0.6 (5) &  -0.98 (5) &  -0.91 (5) &  -0.85 (10) &  -0.72 (10) &        0.46 (5) &   0.6 (5) &    0.44 (10) &   0.56 (10) &         &       &           &       &             &       &   -0.09 &     0.03 \\
Pooled score           &  0.46 (5) &  0.6 (5) &  -0.98 (5) &  -0.91 (5) &  -0.85 (10) &  -0.72 (10) &        0.46 (5) &   0.6 (5) &    0.44 (10) &   0.56 (10) &         &       &           &       &             &       &   -0.09 &     0.03 \\
\textbf{ACE} (w/o .Dip test)  &  0.46 (5) &  0.6 (5) &  -0.66 (5) &  -0.51 (5) &   0.77 (30) &   0.61 (30) &        0.44 (5) &  0.56 (5) &    0.92 (80) &   0.82 (80) &         &       &           &       &             &       &    0.39 &     0.42 \\
\textbf{ACE}    &  0.46 (5) &  0.6 (5) &  -0.66 (5) &  -0.51 (5) &   0.77 (30) &   0.61 (30) &        0.46 (5) &   0.6 (5) &    0.92 (80) &   0.82 (80) &         &       &           &       &             &       &    0.39 &     0.42 \\
\hline 
 \multicolumn{19}{c}{DEPICT: Davies-Bouldin index} \\
 \hline 
Paired score      &   0.46 (5) &    0.6 (5) &  -0.78 (5) &  -0.64 (5) &  -0.85 (10) &  -0.72 (10) &        0.44 (5) &   0.56 (5) &    -0.1 (10) &   0.02 (10) &         &       &           &       &             &       &   -0.17 &    -0.04 \\
Pooled score (w/o. Dip test) &   0.7 (15) &  0.64 (15) &  0.88 (50) &  0.73 (50) &  -0.13 (20) &  -0.17 (20) &       0.94 (10) &  0.82 (10) &   0.92 (100) &  0.82 (100) &         &       &           &       &             &       &    0.66 &     0.57 \\
Pooled score           &   0.6 (15) &  0.51 (15) &  0.88 (50) &  0.73 (50) &  -0.13 (20) &  -0.17 (20) &       0.74 (10) &  0.64 (10) &   0.92 (100) &  0.82 (100) &         &       &           &       &             &       &    0.60 &     0.51 \\
\textbf{ACE} (w/o .Dip test)  &  0.94 (15) &  0.82 (15) &  0.95 (50) &  0.87 (50) &   0.77 (35) &   0.67 (35) &       0.93 (15) &  0.78 (15) &    0.96 (70) &   0.91 (70) &         &       &           &       &             &       &    0.91 &     0.81 \\
\textbf{ACE}    &  0.62 (10) &   0.6 (10) &  0.95 (50) &  0.87 (50) &   0.77 (35) &   0.67 (35) &       0.78 (10) &  0.69 (10) &    0.96 (70) &   0.91 (70) &         &       &           &       &             &       &    0.82 &     0.75 \\
\hline 
 \multicolumn{19}{c}{DEPICT: Silhouette score (Cosine distance)} \\
 \hline 
Paired score      &   0.44 (5) &   0.56 (5) &   -0.7 (5) &   -0.6 (5) &  -0.85 (10) &  -0.72 (10) &        0.44 (5) &   0.56 (5) &    0.07 (10) &  0.11 (10) &         &       &           &       &             &       &   -0.12 &    -0.02 \\
Pooled score (w/o. Dip test) &  0.89 (15) &  0.78 (15) &  0.61 (40) &  0.47 (40) &   0.07 (25) &   0.06 (25) &       0.85 (10) &  0.78 (10) &    0.98 (80) &  0.91 (80) &         &       &           &       &             &       &    0.68 &     0.60 \\
Pooled score           &   0.6 (15) &  0.51 (15) &  0.61 (40) &  0.47 (40) &   0.07 (25) &   0.06 (25) &       0.71 (10) &  0.64 (10) &    0.98 (80) &  0.91 (80) &         &       &           &       &             &       &    0.59 &     0.52 \\
\textbf{ACE} (w/o .Dip test)  &  0.83 (25) &  0.69 (25) &  0.87 (40) &  0.78 (40) &   0.93 (35) &   0.83 (35) &       0.92 (10) &  0.82 (10) &    0.99 (80) &  0.96 (80) &         &       &           &       &             &       &    0.91 &     0.82 \\
\textbf{ACE}    &  0.65 (15) &  0.64 (15) &  0.87 (40) &  0.78 (40) &   0.93 (35) &   0.83 (35) &       0.85 (10) &  0.78 (10) &    0.99 (80) &  0.96 (80) &         &       &           &       &             &       &    0.86 &     0.80 \\
\hline 
 \multicolumn{19}{c}{DEPICT: Silhouette score (Euclidean distance)} \\
 \hline 
Paired score      &   0.44 (5) &   0.56 (5) &  -0.61 (5) &  -0.47 (5) &  -0.85 (10) &  -0.72 (10) &        0.44 (5) &   0.56 (5) &   -0.12 (10) &  -0.02 (10) &         &       &           &       &             &       &   -0.14 &    -0.02 \\
Pooled score (w/o. Dip test) &  0.74 (15) &  0.64 (15) &  0.98 (50) &  0.91 (50) &   0.07 (25) &   0.06 (25) &       0.81 (10) &  0.73 (10) &    0.99 (80) &   0.96 (80) &         &       &           &       &             &       &    0.72 &     0.66 \\
Pooled score           &   0.6 (15) &  0.51 (15) &  0.98 (50) &  0.91 (50) &   0.07 (25) &   0.06 (25) &       0.73 (10) &  0.69 (10) &    0.99 (80) &   0.96 (80) &         &       &           &       &             &       &    0.67 &     0.63 \\
\textbf{ACE} (w/o .Dip test)  &  0.65 (15) &  0.64 (15) &  0.94 (40) &  0.87 (40) &   0.02 (25) &   0.06 (25) &        0.9 (15) &  0.78 (15) &    0.98 (80) &   0.91 (80) &         &       &           &       &             &       &    0.70 &     0.65 \\
\textbf{ACE}    &   0.46 (5) &    0.6 (5) &  0.94 (40) &  0.87 (40) &   0.02 (25) &   0.06 (25) &       0.85 (10) &  0.78 (10) &    0.98 (80) &   0.91 (80) &         &       &           &       &             &       &    0.65 &     0.64 \\
\bottomrule
\end{tabular}
}
 \label{tab:abs:nmi:dip1}
\end{table*}
%%%%%%%%%%%%%%%%%%%%%%%%%%%%%%%%%%%%%%%%%%%%%%%%%%%%%%%%%%%%%%%%%%%%%%%%%%%%%%%%%%
\begin{table*}[htbp!]
\centering
\caption{Perturbation analyses of the Dip test for Application 2. $r_s$ and $\tau_B$ between the generated scores and ACC scores are reported. Empty cells and dash marks ($-$) indicate cases where the result is either missing or impractical to obtain.}
\resizebox{\textwidth}{!}{
\begin{tabular}{lllllllllllllllllll}
\toprule
{} & \multicolumn{2}{l}{USPS (10)} & \multicolumn{2}{l}{YTF (41)} & \multicolumn{2}{l}{FRGC (20)} & \multicolumn{2}{l}{MNIST-test (10)} & \multicolumn{2}{l}{CMU-PIE (68)} & \multicolumn{2}{l}{UMist (20)} & \multicolumn{2}{l}{COIL-20 (20)} & \multicolumn{2}{l}{COIL-100 (100)} & \multicolumn{2}{l}{Average} \\
{} &     $r_s$ & $\tau_B$ &    $r_s$ & $\tau_B$ &     $r_s$ & $\tau_B$ &           $r_s$ & $\tau_B$ &        $r_s$ & $\tau_B$ &      $r_s$ & $\tau_B$ &        $r_s$ & $\tau_B$ &          $r_s$ & $\tau_B$ &   $r_s$ & $\tau_B$ \\
\midrule
\hline 
 \multicolumn{19}{c}{JULE: Calinski-Harabasz index} \\
 \hline 
Paired score      &      0.84 &     0.73 &     0.03 &    -0.06 &     -0.49 &    -0.31 &            0.61 &     0.56 &        -0.09 &    -0.07 &      -0.04 &     0.07 &         0.74 &     0.64 &           0.60 &     0.51 &    0.27 &     0.26 \\
Pooled score (w/o. Dip test) &      0.78 &     0.69 &     0.88 &     0.78 &     -0.37 &    -0.20 &            0.61 &     0.56 &         0.83 &     0.69 &      -0.07 &     0.02 &         0.76 &     0.71 &           0.56 &     0.51 &    0.50 &     0.47 \\
Pooled score           &      0.84 &     0.73 &     0.88 &     0.78 &     -0.37 &    -0.20 &            0.61 &     0.56 &         0.85 &     0.69 &      -0.07 &     0.02 &         0.76 &     0.71 &           0.56 &     0.51 &    0.51 &     0.48 \\
\textbf{ACE} (w/o .Dip test)  &      0.84 &     0.73 &     0.92 &     0.83 &     -0.11 &    -0.03 &            0.61 &     0.56 &         0.83 &     0.69 &      -0.04 &     0.07 &         0.76 &     0.71 &           0.65 &     0.56 &    0.56 &     0.52 \\
\textbf{ACE}    &      0.84 &     0.73 &     0.92 &     0.83 &     -0.11 &    -0.03 &            0.61 &     0.56 &         0.83 &     0.69 &      -0.07 &     0.02 &         0.76 &     0.71 &           0.65 &     0.56 &    0.55 &     0.51 \\
\hline 
 \multicolumn{19}{c}{JULE: Davies-Bouldin index} \\
 \hline 
Paired score      &      0.39 &     0.29 &     0.10 &     0.06 &      0.37 &     0.25 &            0.49 &     0.33 &         0.83 &     0.60 &      -0.28 &    -0.29 &        -0.29 &    -0.21 &          -0.87 &    -0.73 &    0.09 &     0.04 \\
Pooled score (w/o. Dip test) &      0.77 &     0.56 &     0.80 &     0.67 &      0.71 &     0.54 &            0.84 &     0.69 &         0.85 &     0.69 &      -0.06 &    -0.20 &        -0.69 &    -0.57 &          -0.79 &    -0.69 &    0.30 &     0.21 \\
Pooled score           &      0.89 &     0.73 &     0.80 &     0.67 &      0.71 &     0.54 &            0.83 &     0.64 &         0.85 &     0.69 &      -0.42 &    -0.33 &        -0.79 &    -0.64 &          -0.79 &    -0.69 &    0.26 &     0.20 \\
\textbf{ACE} (w/o .Dip test)  &     -0.15 &    -0.11 &     0.80 &     0.67 &      0.60 &     0.42 &            0.67 &     0.56 &         1.00 &     1.00 &      -0.06 &    -0.20 &        -0.69 &    -0.57 &          -0.79 &    -0.69 &    0.17 &     0.13 \\
\textbf{ACE}    &      0.89 &     0.73 &     0.80 &     0.67 &      0.60 &     0.42 &            0.83 &     0.64 &         0.88 &     0.73 &      -0.42 &    -0.33 &        -0.71 &    -0.64 &          -0.82 &    -0.69 &    0.26 &     0.19 \\
\hline 
 \multicolumn{19}{c}{JULE: Silhouette score (Cosine distance)} \\
 \hline 
Paired score      &      0.89 &     0.78 &     0.27 &     0.22 &      0.21 &     0.09 &            0.81 &     0.64 &         0.99 &     0.96 &      -0.26 &    -0.24 &         0.55 &     0.43 &           0.52 &     0.33 &    0.50 &     0.40 \\
Pooled score (w/o. Dip test) &      0.88 &     0.73 &     0.98 &     0.94 &      0.61 &     0.48 &            0.90 &     0.78 &         0.99 &     0.96 &       0.04 &    -0.07 &         0.38 &     0.29 &           0.59 &     0.47 &    0.67 &     0.57 \\
Pooled score           &      0.95 &     0.87 &     0.98 &     0.94 &      0.61 &     0.48 &            0.94 &     0.82 &         0.99 &     0.96 &      -0.32 &    -0.24 &         0.67 &     0.50 &           0.54 &     0.38 &    0.67 &     0.59 \\
\textbf{ACE} (w/o .Dip test)  &      0.96 &     0.91 &     0.98 &     0.94 &      0.64 &     0.54 &            0.98 &     0.91 &         0.99 &     0.96 &      -0.76 &    -0.60 &        -0.21 &    -0.07 &           0.59 &     0.47 &    0.52 &     0.51 \\
\textbf{ACE}    &      0.95 &     0.87 &     0.98 &     0.94 &      0.64 &     0.54 &            0.94 &     0.82 &         0.99 &     0.96 &      -0.32 &    -0.24 &         0.76 &     0.57 &           0.60 &     0.47 &    0.69 &     0.61 \\
\hline 
 \multicolumn{19}{c}{JULE: Silhouette score (Euclidean distance)} \\
 \hline 
Paired score      &      0.93 &     0.82 &     0.30 &     0.28 &      0.21 &     0.09 &            0.82 &     0.64 &         0.98 &     0.91 &      -0.13 &    -0.16 &         0.52 &     0.36 &           0.55 &     0.42 &    0.52 &     0.42 \\
Pooled score (w/o. Dip test) &      0.88 &     0.73 &     0.97 &     0.89 &      0.61 &     0.48 &            0.90 &     0.78 &         0.99 &     0.96 &       0.04 &    -0.07 &         0.33 &     0.14 &           0.59 &     0.47 &    0.66 &     0.55 \\
Pooled score           &      0.95 &     0.87 &     0.97 &     0.89 &      0.61 &     0.48 &            0.92 &     0.78 &         0.99 &     0.96 &      -0.03 &    -0.11 &         0.74 &     0.50 &           0.59 &     0.47 &    0.72 &     0.60 \\
\textbf{ACE} (w/o .Dip test)  &      0.90 &     0.82 &     0.98 &     0.94 &      0.57 &     0.48 &            0.89 &     0.78 &         1.00 &     1.00 &      -0.89 &    -0.73 &         0.31 &     0.21 &           0.56 &     0.42 &    0.54 &     0.49 \\
\textbf{ACE}    &      0.95 &     0.87 &     0.98 &     0.94 &      0.57 &     0.48 &            0.92 &     0.78 &         0.99 &     0.96 &      -0.03 &    -0.11 &         0.74 &     0.50 &           0.59 &     0.47 &    0.71 &     0.61 \\
\hline 
 \multicolumn{19}{c}{DEPICT: Calinski-Harabasz index} \\
 \hline 
Paired score      &      0.88 &     0.82 &    -0.96 &    -0.91 &     -0.37 &    -0.22 &            0.79 &     0.73 &        -0.92 &    -0.82 &         &       &           &       &             &       &   -0.11 &    -0.08 \\
Pooled score (w/o. Dip test) &      0.88 &     0.82 &    -0.94 &    -0.87 &     -0.37 &    -0.22 &            0.82 &     0.78 &         0.44 &     0.56 &         &       &           &       &             &       &    0.17 &     0.21 \\
Pooled score           &      0.88 &     0.82 &    -0.94 &    -0.87 &     -0.37 &    -0.22 &            0.82 &     0.78 &         0.44 &     0.56 &         &       &           &       &             &       &    0.17 &     0.21 \\
\textbf{ACE} (w/o .Dip test)  &      0.88 &     0.82 &    -0.67 &    -0.56 &      0.92 &     0.78 &            0.81 &     0.73 &         0.92 &     0.82 &         &       &           &       &             &       &    0.57 &     0.52 \\
\textbf{ACE}    &      0.88 &     0.82 &    -0.67 &    -0.56 &      0.92 &     0.78 &            0.82 &     0.78 &         0.92 &     0.82 &         &       &           &       &             &       &    0.57 &     0.53 \\
\hline 
 \multicolumn{19}{c}{DEPICT: Davies-Bouldin index} \\
 \hline 
Paired score      &      0.88 &     0.82 &    -0.77 &    -0.60 &     -0.37 &    -0.22 &            0.79 &     0.73 &        -0.10 &     0.02 &         &       &           &       &             &       &    0.09 &     0.15 \\
Pooled score (w/o. Dip test) &      0.48 &     0.42 &     0.90 &     0.78 &      0.47 &     0.33 &            0.85 &     0.73 &         0.92 &     0.82 &         &       &           &       &             &       &    0.72 &     0.62 \\
Pooled score           &      0.90 &     0.73 &     0.90 &     0.78 &      0.47 &     0.33 &            0.88 &     0.82 &         0.92 &     0.82 &         &       &           &       &             &       &    0.81 &     0.70 \\
\textbf{ACE} (w/o .Dip test)  &      0.83 &     0.69 &     0.96 &     0.91 &      0.92 &     0.83 &            0.84 &     0.69 &         0.96 &     0.91 &         &       &           &       &             &       &    0.90 &     0.81 \\
\textbf{ACE}    &      0.93 &     0.82 &     0.96 &     0.91 &      0.92 &     0.83 &            0.93 &     0.87 &         0.96 &     0.91 &         &       &           &       &             &       &    0.94 &     0.87 \\
\hline 
 \multicolumn{19}{c}{DEPICT: Silhouette score (Cosine distance)} \\
 \hline 
Paired score      &      0.87 &     0.78 &    -0.69 &    -0.56 &     -0.37 &    -0.22 &            0.79 &     0.73 &         0.07 &     0.11 &         &       &           &       &             &       &    0.14 &     0.17 \\
Pooled score (w/o. Dip test) &      0.85 &     0.73 &     0.67 &     0.51 &      0.68 &     0.56 &            0.95 &     0.87 &         0.98 &     0.91 &         &       &           &       &             &       &    0.83 &     0.72 \\
Pooled score           &      0.90 &     0.73 &     0.67 &     0.51 &      0.68 &     0.56 &            0.90 &     0.82 &         0.98 &     0.91 &         &       &           &       &             &       &    0.83 &     0.71 \\
\textbf{ACE} (w/o .Dip test)  &      0.64 &     0.47 &     0.92 &     0.82 &      0.80 &     0.67 &            0.96 &     0.91 &         0.99 &     0.96 &         &       &           &       &             &       &    0.86 &     0.76 \\
\textbf{ACE}    &      0.95 &     0.87 &     0.92 &     0.82 &      0.80 &     0.67 &            0.95 &     0.87 &         0.99 &     0.96 &         &       &           &       &             &       &    0.92 &     0.84 \\
\hline 
 \multicolumn{19}{c}{DEPICT: Silhouette score (Euclidean distance)} \\
 \hline 
Paired score      &      0.87 &     0.78 &    -0.64 &    -0.51 &     -0.37 &    -0.22 &            0.79 &     0.73 &        -0.12 &    -0.02 &         &       &           &       &             &       &    0.11 &     0.15 \\
Pooled score (w/o. Dip test) &      0.93 &     0.87 &     0.99 &     0.96 &      0.68 &     0.56 &            0.96 &     0.91 &         0.99 &     0.96 &         &       &           &       &             &       &    0.91 &     0.85 \\
Pooled score           &      0.90 &     0.73 &     0.99 &     0.96 &      0.68 &     0.56 &            0.94 &     0.87 &         0.99 &     0.96 &         &       &           &       &             &       &    0.90 &     0.81 \\
\textbf{ACE} (w/o .Dip test)  &      0.95 &     0.87 &     0.98 &     0.91 &      0.73 &     0.56 &            0.95 &     0.87 &         0.98 &     0.91 &         &       &           &       &             &       &    0.92 &     0.82 \\
\textbf{ACE}    &      0.88 &     0.82 &     0.98 &     0.91 &      0.73 &     0.56 &            0.95 &     0.87 &         0.98 &     0.91 &         &       &           &       &             &       &    0.90 &     0.81 \\
\bottomrule
\end{tabular}
}

 \label{tab:abs:acc:dip1}
\end{table*}

%%%%%%%%%%%%%%%%%%%%%%%%%%%%%%%%%%%%%%%%%%%%%%%%%%%%%%%%%%%%%%%%%%%%%%%%%%%%%%%%%%%%%%%%%
%%%%%%%%%%%%%%%%%%%%%%%%%%%%%%%%%%%%%%%%%%%%%%%%%%%%%%%%%%%%%%%%%%%%%%%%%%%%%%%%%%%%%%%%%
%%%%%%%%%%%%%%%%%%%%%%%%%%%%%%%%%%%%%%%%%%%%%%%%%%%%%%%%%%%%%%%%%%%%%%%%%%%%%%%%%%%%%%%%%
\clearpage
\newpage \subsection{Different $\alpha_{edges}$} \label{app:exp:ab:alpha}
In this section, we examine the impact of different significance thresholds controlling family-wise error rate ($\alpha_{edges}$) on edge inclusion in link analysis. In Algorithm \ref{Algorithm1}, a multiple testing procedure (the Holm–Bonferroni method in this paper) that controls FWER at $\alpha_{edges}$, followed by the exclusion of all negative edges, is applied to ensure that only edges with significant positive rank correlation are included in link analysis. Beyond the main experiments using $\alpha_{edges} = 0.1$, we conduct additional analyses with $\alpha_{edges} = 0.05$, which imposes a stricter criterion for edge inclusion, and with all positive edges included (referred to as ``include all'' in the tables for simplicity). The comparative results for Application 1 are presented in Tables \ref{tab:abs:nmi:alpha0} and \ref{tab:abs:acc:alpha0}, while those for Application 2 are in Tables \ref{tab:abs:nmi:alpha1} and \ref{tab:abs:acc:alpha1}. Overall, ACE demonstrates robustness to the choice of $\alpha_{edges}$, as performance remains consistent between $\alpha_{edges} = 0.1$ and $\alpha_{edges} = 0.05$ in most cases. Including all positive edges (by setting $\alpha$ to a very large value while excluding negative edges) also yields similar results in the majority of instances. However, in certain cases, such as DEPICT (Calinski-Harabasz index) in Tables \ref{tab:abs:nmi:alpha1} and \ref{tab:abs:acc:alpha1}, including all positive edges leads to a notably lower correlation. This suggests the probable importance of using a multiple testing procedure to selectively include only significantly rank-correlated edges in link analysis.

\begin{table*}[htbp!]
\centering
\caption{Perturbation analyses on different $\alpha_{edges}$ for Application 1. $r_s$ and $\tau_B$ between the generated scores and NMI scores are reported. Empty cells and dash marks ($-$) indicate cases where the result is either missing or impractical to obtain.}
 \label{tab:abs:nmi:alpha0}
\resizebox{\textwidth}{!}{
\begin{tabular}{lllllllllllllllllll}
\toprule
{} & \multicolumn{2}{l}{USPS} & \multicolumn{2}{l}{YTF} & \multicolumn{2}{l}{FRGC} & \multicolumn{2}{l}{MNIST-test} & \multicolumn{2}{l}{CMU-PIE} & \multicolumn{2}{l}{UMist} & \multicolumn{2}{l}{COIL-20} & \multicolumn{2}{l}{COIL-100} & \multicolumn{2}{l}{Average} \\
{} & $r_s$ & $\tau_B$ & $r_s$ & $\tau_B$ & $r_s$ & $\tau_B$ &      $r_s$ & $\tau_B$ &   $r_s$ & $\tau_B$ & $r_s$ & $\tau_B$ &   $r_s$ & $\tau_B$ &    $r_s$ & $\tau_B$ &   $r_s$ & $\tau_B$ \\
\midrule
\hline 
 \multicolumn{19}{c}{JULE: Calinski-Harabasz index} \\
 \hline 
Paired score     &  0.17 &     0.13 &  0.52 &     0.40 & -0.13 &    -0.10 &       0.49 &     0.34 &   -0.13 &    -0.08 &  0.70 &     0.50 &    0.53 &     0.38 &     0.20 &     0.19 &    0.29 &     0.22 \\
\textbf{ACE} (include all) &  0.80 &     0.63 &  0.90 &     0.73 &  0.39 &     0.26 &       0.87 &     0.71 &    0.98 &     0.90 &  0.81 &     0.61 &    0.60 &     0.45 &     0.95 &     0.82 &    0.79 &     0.64 \\
\textbf{ACE} ($\alpha_{edges}=0.1$)   &  0.80 &     0.63 &  0.90 &     0.73 &  0.39 &     0.26 &       0.87 &     0.71 &    0.98 &     0.90 &  0.81 &     0.61 &    0.60 &     0.45 &     0.95 &     0.82 &    0.79 &     0.64 \\
\textbf{ACE} ($\alpha_{edges}=0.05$) &  0.80 &     0.63 &  0.90 &     0.73 &  0.39 &     0.26 &       0.87 &     0.71 &    0.98 &     0.90 &  0.81 &     0.61 &    0.60 &     0.45 &     0.95 &     0.82 &    0.79 &     0.64 \\
\hline 
 \multicolumn{19}{c}{JULE: Davies-Bouldin index} \\
 \hline 
Paired score     & -0.10 &    -0.03 & -0.32 &    -0.21 & -0.08 &    -0.05 &      -0.13 &    -0.06 &    0.26 &     0.20 &  0.62 &     0.44 &    0.61 &     0.42 &     0.43 &     0.35 &    0.16 &     0.13 \\
\textbf{ACE} (include all) & -0.08 &    -0.02 & -0.30 &    -0.21 &  0.22 &     0.16 &       0.73 &     0.55 &    0.10 &     0.06 &  0.36 &     0.25 &    0.23 &     0.22 &     0.54 &     0.38 &    0.23 &     0.17 \\
\textbf{ACE} ($\alpha_{edges}=0.1$)   & -0.08 &    -0.02 & -0.30 &    -0.21 &  0.22 &     0.16 &       0.73 &     0.55 &    0.10 &     0.06 &  0.38 &     0.27 &    0.23 &     0.22 &     0.48 &     0.33 &    0.22 &     0.17 \\
\textbf{ACE} ($\alpha_{edges}=0.05$) & -0.08 &    -0.02 & -0.30 &    -0.21 &  0.22 &     0.16 &       0.73 &     0.55 &    0.10 &     0.06 &  0.30 &     0.20 &    0.23 &     0.22 &     0.48 &     0.33 &    0.21 &     0.16 \\
\hline 
 \multicolumn{19}{c}{JULE: Silhouette score (Cosine distance)} \\
 \hline 
Paired score     &  0.28 &     0.22 &  0.73 &     0.56 &  0.09 &     0.06 &       0.63 &     0.47 &    0.50 &     0.36 &  0.71 &     0.50 &    0.68 &     0.50 &     0.74 &     0.54 &    0.54 &     0.40 \\
\textbf{ACE} (include all) &  0.89 &     0.73 &  0.93 &     0.83 &  0.52 &     0.35 &       0.81 &     0.66 &    0.99 &     0.93 &  0.79 &     0.59 &    0.44 &     0.38 &     0.92 &     0.78 &    0.79 &     0.66 \\
\textbf{ACE} ($\alpha_{edges}=0.1$)   &  0.89 &     0.73 &  0.93 &     0.83 &  0.52 &     0.35 &       0.81 &     0.66 &    0.99 &     0.93 &  0.79 &     0.59 &    0.44 &     0.38 &     0.92 &     0.78 &    0.79 &     0.66 \\
\textbf{ACE} ($\alpha_{edges}=0.05$) &  0.89 &     0.73 &  0.93 &     0.83 &  0.52 &     0.35 &       0.81 &     0.66 &    0.99 &     0.93 &  0.80 &     0.59 &    0.44 &     0.38 &     0.92 &     0.78 &    0.79 &     0.66 \\
\hline 
 \multicolumn{19}{c}{JULE: Silhouette score (Euclidean distance)} \\
 \hline 
Paired score     &  0.27 &     0.20 &  0.72 &     0.55 &  0.04 &     0.03 &       0.56 &     0.41 &    0.42 &     0.30 &  0.70 &     0.50 &    0.64 &     0.46 &     0.55 &     0.41 &    0.49 &     0.36 \\
\textbf{ACE} (include all) &  0.88 &     0.72 &  0.89 &     0.75 &  0.42 &     0.28 &       0.81 &     0.65 &    0.98 &     0.90 &  0.88 &     0.70 &    0.41 &     0.36 &     0.92 &     0.78 &    0.77 &     0.64 \\
\textbf{ACE} ($\alpha_{edges}=0.1$)   &  0.88 &     0.72 &  0.89 &     0.75 &  0.42 &     0.28 &       0.81 &     0.65 &    0.98 &     0.90 &  0.88 &     0.70 &    0.41 &     0.36 &     0.92 &     0.78 &    0.77 &     0.64 \\
\textbf{ACE} ($\alpha_{edges}=0.05$) &  0.88 &     0.72 &  0.89 &     0.75 &  0.42 &     0.28 &       0.81 &     0.65 &    0.98 &     0.90 &  0.88 &     0.70 &    0.41 &     0.36 &     0.92 &     0.78 &    0.77 &     0.64 \\
\hline 
 \multicolumn{19}{c}{DEPICT: Calinski-Harabasz index} \\
 \hline 
Paired score     &  0.76 &     0.57 &  0.44 &     0.26 &  0.76 &     0.57 &       0.89 &     0.72 &    0.49 &     0.44 &    &       &      &       &       &       &    0.67 &     0.51 \\
\textbf{ACE} (include all) &  0.91 &     0.77 &  0.56 &     0.44 &  0.94 &     0.82 &       0.96 &     0.87 &    0.96 &     0.87 &    &       &      &       &       &       &    0.87 &     0.75 \\
\textbf{ACE} ($\alpha_{edges}=0.1$)   &  0.91 &     0.77 &  0.56 &     0.44 &  0.94 &     0.82 &       0.96 &     0.87 &    0.96 &     0.87 &    &       &      &       &       &       &    0.87 &     0.75 \\
\textbf{ACE} ($\alpha_{edges}=0.05$) &  0.91 &     0.77 &  0.56 &     0.44 &  0.94 &     0.82 &       0.96 &     0.87 &    0.95 &     0.84 &    &       &      &       &       &       &    0.87 &     0.75 \\
\hline 
 \multicolumn{19}{c}{DEPICT: Davies-Bouldin index} \\
 \hline 
Paired score     &  0.81 &     0.59 &  0.45 &     0.31 &  0.90 &     0.74 &       0.89 &     0.72 &    0.63 &     0.59 &    &       &      &       &       &       &    0.73 &     0.59 \\
\textbf{ACE} (include all) &  0.91 &     0.82 &  0.76 &     0.58 &  0.89 &     0.75 &       0.96 &     0.87 &    0.98 &     0.92 &    &       &      &       &       &       &    0.90 &     0.79 \\
\textbf{ACE} ($\alpha_{edges}=0.1$)   &  0.91 &     0.82 &  0.76 &     0.58 &  0.91 &     0.79 &       0.96 &     0.87 &    0.98 &     0.92 &    &       &      &       &       &       &    0.90 &     0.80 \\
\textbf{ACE} ($\alpha_{edges}=0.05$) &  0.91 &     0.82 &  0.76 &     0.58 &  0.91 &     0.79 &       0.96 &     0.87 &    0.98 &     0.92 &    &       &      &       &       &       &    0.90 &     0.80 \\
\hline 
 \multicolumn{19}{c}{DEPICT: Silhouette score (Cosine distance)} \\
 \hline 
Paired score     &  0.81 &     0.62 &  0.45 &     0.33 &  0.90 &     0.75 &       0.89 &     0.72 &    0.77 &     0.58 &    &       &      &       &       &       &    0.76 &     0.60 \\
\textbf{ACE} (include all) &  0.97 &     0.90 &  0.56 &     0.45 &  0.94 &     0.82 &       0.97 &     0.90 &    0.94 &     0.83 &    &       &      &       &       &       &    0.88 &     0.78 \\
\textbf{ACE} ($\alpha_{edges}=0.1$)   &  0.97 &     0.90 &  0.71 &     0.56 &  0.94 &     0.82 &       0.97 &     0.90 &    0.94 &     0.83 &    &       &      &       &       &       &    0.91 &     0.80 \\
\textbf{ACE} ($\alpha_{edges}=0.05$) &  0.97 &     0.90 &  0.71 &     0.56 &  0.94 &     0.82 &       0.97 &     0.90 &    0.94 &     0.83 &    &       &      &       &       &       &    0.91 &     0.80 \\
\hline 
 \multicolumn{19}{c}{DEPICT: Silhouette score (Euclidean distance)} \\
 \hline 
Paired score     &  0.73 &     0.50 &  0.47 &     0.36 &  0.79 &     0.65 &       0.86 &     0.69 &    0.59 &     0.52 &    &       &      &       &       &       &    0.69 &     0.54 \\
\textbf{ACE} (include all) &  0.97 &     0.88 &  0.65 &     0.50 &  0.95 &     0.83 &       0.98 &     0.90 &    0.94 &     0.82 &    &       &      &       &       &       &    0.90 &     0.79 \\
\textbf{ACE} ($\alpha_{edges}=0.1$)   &  0.97 &     0.88 &  0.65 &     0.50 &  0.95 &     0.83 &       0.98 &     0.90 &    0.94 &     0.82 &    &       &      &       &       &       &    0.90 &     0.79 \\
\textbf{ACE} ($\alpha_{edges}=0.05$) &  0.97 &     0.88 &  0.67 &     0.52 &  0.95 &     0.83 &       0.98 &     0.90 &    0.94 &     0.82 &    &       &      &       &       &       &    0.90 &     0.79 \\
\bottomrule
\end{tabular}
}
\end{table*}
 %%%%%%%%%%%%%%%%%%%%%%%%%%%%%%
\begin{table*}[htbp!]
\centering
\caption{Perturbation analyses on different $\alpha_{edges}$ for Application 1. $r_s$ and $\tau_B$ between the generated scores and ACC scores are reported. Empty cells and dash marks ($-$) indicate cases where the result is either missing or impractical to obtain.}
\resizebox{\textwidth}{!}{
\begin{tabular}{lllllllllllllllllll}
\toprule
{} & \multicolumn{2}{l}{USPS} & \multicolumn{2}{l}{YTF} & \multicolumn{2}{l}{FRGC} & \multicolumn{2}{l}{MNIST-test} & \multicolumn{2}{l}{CMU-PIE} & \multicolumn{2}{l}{UMist} & \multicolumn{2}{l}{COIL-20} & \multicolumn{2}{l}{COIL-100} & \multicolumn{2}{l}{Average} \\
{} & $r_s$ & $\tau_B$ & $r_s$ & $\tau_B$ & $r_s$ & $\tau_B$ &      $r_s$ & $\tau_B$ &   $r_s$ & $\tau_B$ & $r_s$ & $\tau_B$ &   $r_s$ & $\tau_B$ &    $r_s$ & $\tau_B$ &   $r_s$ & $\tau_B$ \\
\midrule
\hline 
 \multicolumn{19}{c}{JULE: Calinski-Harabasz index} \\
 \hline 
Paired score     &  0.04 &     0.05 &  0.39 &     0.27 & -0.26 &    -0.18 &       0.31 &     0.21 &   -0.20 &    -0.12 &  0.64 &     0.45 &    0.57 &     0.40 &     0.09 &     0.08 &    0.20 &     0.14 \\
\textbf{ACE} (include all) &  0.90 &     0.77 &  0.73 &     0.54 &  0.49 &     0.36 &       0.95 &     0.82 &    0.97 &     0.87 &  0.81 &     0.61 &    0.57 &     0.40 &     0.93 &     0.81 &    0.79 &     0.65 \\
\textbf{ACE} ($\alpha_{edges}=0.1$)   &  0.90 &     0.77 &  0.73 &     0.54 &  0.49 &     0.36 &       0.95 &     0.82 &    0.97 &     0.87 &  0.81 &     0.61 &    0.57 &     0.40 &     0.93 &     0.81 &    0.79 &     0.65 \\
\textbf{ACE} ($\alpha_{edges}=0.05$) &  0.90 &     0.77 &  0.73 &     0.54 &  0.49 &     0.36 &       0.95 &     0.82 &    0.97 &     0.87 &  0.81 &     0.61 &    0.57 &     0.40 &     0.93 &     0.81 &    0.79 &     0.65 \\
\hline 
 \multicolumn{19}{c}{JULE: Davies-Bouldin index} \\
 \hline 
Paired score     & -0.27 &    -0.15 & -0.14 &    -0.09 & -0.23 &    -0.14 &      -0.35 &    -0.19 &    0.20 &     0.16 &  0.53 &     0.36 &    0.63 &     0.44 &     0.33 &     0.26 &    0.09 &     0.08 \\
\textbf{ACE} (include all) & -0.30 &    -0.09 & -0.07 &    -0.07 &  0.53 &     0.38 &       0.79 &     0.64 &    0.07 &     0.03 &  0.24 &     0.17 &    0.21 &     0.18 &     0.49 &     0.33 &    0.24 &     0.20 \\
\textbf{ACE} ($\alpha_{edges}=0.1$)   & -0.30 &    -0.09 & -0.07 &    -0.07 &  0.53 &     0.38 &       0.79 &     0.64 &    0.07 &     0.03 &  0.27 &     0.20 &    0.21 &     0.18 &     0.44 &     0.28 &    0.24 &     0.19 \\
\textbf{ACE} ($\alpha_{edges}=0.05$) & -0.30 &    -0.09 & -0.07 &    -0.07 &  0.53 &     0.38 &       0.79 &     0.64 &    0.07 &     0.03 &  0.17 &     0.11 &    0.21 &     0.18 &     0.44 &     0.28 &    0.23 &     0.18 \\
\hline 
 \multicolumn{19}{c}{JULE: Silhouette score (Cosine distance)} \\
 \hline 
Paired score     &  0.17 &     0.14 &  0.59 &     0.41 &  0.07 &     0.06 &       0.47 &     0.33 &    0.45 &     0.33 &  0.64 &     0.46 &    0.70 &     0.51 &     0.64 &     0.45 &    0.47 &     0.34 \\
\textbf{ACE} (include all) &  0.96 &     0.85 &  0.74 &     0.55 &  0.82 &     0.65 &       0.92 &     0.78 &    0.98 &     0.92 &  0.78 &     0.58 &    0.41 &     0.32 &     0.84 &     0.68 &    0.81 &     0.67 \\
\textbf{ACE} ($\alpha_{edges}=0.1$)   &  0.96 &     0.85 &  0.74 &     0.55 &  0.82 &     0.65 &       0.92 &     0.78 &    0.98 &     0.92 &  0.78 &     0.58 &    0.41 &     0.32 &     0.84 &     0.68 &    0.81 &     0.67 \\
\textbf{ACE} ($\alpha_{edges}=0.05$) &  0.96 &     0.85 &  0.74 &     0.55 &  0.82 &     0.65 &       0.92 &     0.78 &    0.98 &     0.92 &  0.78 &     0.57 &    0.41 &     0.32 &     0.84 &     0.68 &    0.81 &     0.66 \\
\hline 
 \multicolumn{19}{c}{JULE: Silhouette score (Euclidean distance)} \\
 \hline 
Paired score     &  0.14 &     0.12 &  0.54 &     0.39 & -0.08 &    -0.02 &       0.41 &     0.27 &    0.36 &     0.27 &  0.64 &     0.46 &    0.67 &     0.48 &     0.44 &     0.31 &    0.39 &     0.28 \\
\textbf{ACE} (include all) &  0.93 &     0.78 &  0.63 &     0.48 &  0.71 &     0.53 &       0.92 &     0.78 &    0.98 &     0.91 &  0.86 &     0.68 &    0.39 &     0.30 &     0.84 &     0.68 &    0.78 &     0.64 \\
\textbf{ACE} ($\alpha_{edges}=0.1$)   &  0.93 &     0.78 &  0.63 &     0.48 &  0.71 &     0.53 &       0.92 &     0.78 &    0.98 &     0.91 &  0.86 &     0.68 &    0.39 &     0.30 &     0.84 &     0.68 &    0.78 &     0.64 \\
\textbf{ACE} ($\alpha_{edges}=0.05$) &  0.93 &     0.78 &  0.63 &     0.48 &  0.71 &     0.53 &       0.92 &     0.78 &    0.98 &     0.91 &  0.86 &     0.68 &    0.39 &     0.30 &     0.84 &     0.68 &    0.78 &     0.64 \\
\hline 
 \multicolumn{19}{c}{DEPICT: Calinski-Harabasz index} \\
 \hline 
Paired score     &  0.56 &     0.40 &  0.54 &     0.35 &  0.76 &     0.57 &       0.88 &     0.69 &    0.48 &     0.43 &    &       &      &       &       &       &    0.64 &     0.49 \\
\textbf{ACE} (include all) &  0.82 &     0.72 &  0.61 &     0.45 &  0.91 &     0.82 &       0.97 &     0.91 &    0.96 &     0.87 &    &       &      &       &       &       &    0.86 &     0.75 \\
\textbf{ACE} ($\alpha_{edges}=0.1$)   &  0.82 &     0.72 &  0.61 &     0.45 &  0.91 &     0.82 &       0.97 &     0.91 &    0.96 &     0.87 &    &       &      &       &       &       &    0.86 &     0.75 \\
\textbf{ACE} ($\alpha_{edges}=0.05$) &  0.82 &     0.72 &  0.61 &     0.45 &  0.91 &     0.82 &       0.97 &     0.91 &    0.96 &     0.87 &    &       &      &       &       &       &    0.86 &     0.75 \\
\hline 
 \multicolumn{19}{c}{DEPICT: Davies-Bouldin index} \\
 \hline 
Paired score     &  0.61 &     0.42 &  0.48 &     0.32 &  0.92 &     0.74 &       0.88 &     0.69 &    0.62 &     0.56 &    &       &      &       &       &       &    0.70 &     0.55 \\
\textbf{ACE} (include all) &  0.99 &     0.96 &  0.65 &     0.46 &  0.88 &     0.72 &       0.99 &     0.96 &    0.96 &     0.87 &    &       &      &       &       &       &    0.89 &     0.80 \\
\textbf{ACE} ($\alpha_{edges}=0.1$)   &  0.99 &     0.96 &  0.65 &     0.46 &  0.90 &     0.74 &       0.99 &     0.96 &    0.96 &     0.87 &    &       &      &       &       &       &    0.90 &     0.80 \\
\textbf{ACE} ($\alpha_{edges}=0.05$) &  0.99 &     0.96 &  0.65 &     0.46 &  0.90 &     0.74 &       0.99 &     0.96 &    0.96 &     0.87 &    &       &      &       &       &       &    0.90 &     0.80 \\
\hline 
 \multicolumn{19}{c}{DEPICT: Silhouette score (Cosine distance)} \\
 \hline 
Paired score     &  0.62 &     0.45 &  0.53 &     0.42 &  0.91 &     0.75 &       0.88 &     0.69 &    0.77 &     0.58 &    &       &      &       &       &       &    0.74 &     0.58 \\
\textbf{ACE} (include all) &  0.95 &     0.88 &  0.60 &     0.44 &  0.91 &     0.77 &       0.96 &     0.88 &    0.94 &     0.83 &    &       &      &       &       &       &    0.87 &     0.76 \\
\textbf{ACE} ($\alpha_{edges}=0.1$)   &  0.95 &     0.88 &  0.70 &     0.54 &  0.91 &     0.77 &       0.96 &     0.88 &    0.94 &     0.83 &    &       &      &       &       &       &    0.89 &     0.78 \\
\textbf{ACE} ($\alpha_{edges}=0.05$) &  0.95 &     0.88 &  0.70 &     0.54 &  0.91 &     0.77 &       0.96 &     0.88 &    0.94 &     0.83 &    &       &      &       &       &       &    0.89 &     0.78 \\
\hline 
 \multicolumn{19}{c}{DEPICT: Silhouette score (Euclidean distance)} \\
 \hline 
Paired score     &  0.52 &     0.33 &  0.57 &     0.45 &  0.80 &     0.62 &       0.85 &     0.65 &    0.59 &     0.48 &    &       &      &       &       &       &    0.67 &     0.51 \\
\textbf{ACE} (include all) &  0.95 &     0.87 &  0.63 &     0.49 &  0.91 &     0.78 &       0.97 &     0.91 &    0.95 &     0.84 &    &       &      &       &       &       &    0.88 &     0.78 \\
\textbf{ACE} ($\alpha_{edges}=0.1$)   &  0.95 &     0.87 &  0.63 &     0.49 &  0.91 &     0.78 &       0.97 &     0.91 &    0.95 &     0.84 &    &       &      &       &       &       &    0.88 &     0.78 \\
\textbf{ACE} ($\alpha_{edges}=0.05$) &  0.95 &     0.87 &  0.64 &     0.50 &  0.91 &     0.78 &       0.97 &     0.91 &    0.95 &     0.84 &    &       &      &       &       &       &    0.88 &     0.78 \\
\bottomrule
\end{tabular}
}
 \label{tab:abs:acc:alpha0}
\end{table*}
%%%%%%%%%%%%%%%%%%%%%%%%%\newpage \subsection{Determination of the number of clusters  - Different $\alpha_{edges}$} 
\begin{table*}[htbp!]
\centering
\caption{Perturbation analyses on different $\alpha_{edges}$ for Application 2. $r_s$ and $\tau_B$ between the generated scores and NMI scores are reported. Empty cells and dash marks ($-$) indicate cases where the result is either missing or impractical to obtain.}
\resizebox{\textwidth}{!}{
\begin{tabular}{lllllllllllllllllll}
\toprule
{} & \multicolumn{2}{l}{USPS (10)} & \multicolumn{2}{l}{YTF (41)} & \multicolumn{2}{l}{FRGC (20)} & \multicolumn{2}{l}{MNIST-test (10)} & \multicolumn{2}{l}{CMU-PIE (68)} & \multicolumn{2}{l}{UMist (20)} & \multicolumn{2}{l}{COIL-20 (20)} & \multicolumn{2}{l}{COIL-100 (100)} & \multicolumn{2}{l}{Average} \\
{} &      $r_s$ &   $\tau_B$ &      $r_s$ &   $\tau_B$ &       $r_s$ &    $\tau_B$ &           $r_s$ &  $\tau_B$ &        $r_s$ &    $\tau_B$ &      $r_s$ &   $\tau_B$ &        $r_s$ &   $\tau_B$ &          $r_s$ &   $\tau_B$ &   $r_s$ & $\tau_B$ \\
\midrule
\hline 
 \multicolumn{19}{c}{JULE: Calinski-Harabasz index} \\
 \hline 
Paired score     &  0.65 (10) &  0.64 (10) &   0.1 (50) &  0.06 (50) &  -0.93 (15) &  -0.83 (15) &       0.64 (10) &  0.6 (10) &   -0.03 (20) &  -0.02 (20) &  -0.13 (5) &  -0.07 (5) &    0.76 (15) &  0.71 (15) &      0.74 (80) &  0.56 (80) &    0.22 &     0.21 \\
\textbf{ACE} (include all) &  0.65 (10) &  0.64 (10) &  0.93 (50) &  0.83 (50) &  -0.72 (15) &  -0.67 (15) &       0.64 (10) &  0.6 (10) &    0.88 (70) &   0.73 (70) &  -0.14 (5) &  -0.11 (5) &    0.74 (15) &  0.64 (15) &      0.79 (80) &  0.69 (80) &    0.47 &     0.42 \\
\textbf{ACE} ($\alpha_{edges}=0.1$)   &  0.65 (10) &  0.64 (10) &  0.93 (50) &  0.83 (50) &  -0.72 (15) &  -0.67 (15) &       0.64 (10) &  0.6 (10) &    0.88 (70) &   0.73 (70) &  -0.14 (5) &  -0.11 (5) &    0.74 (15) &  0.64 (15) &      0.79 (80) &  0.69 (80) &    0.47 &     0.42 \\
\textbf{ACE} ($\alpha_{edges}=0.05$) &  0.65 (10) &  0.64 (10) &  0.93 (50) &  0.83 (50) &  -0.72 (15) &  -0.67 (15) &       0.64 (10) &  0.6 (10) &    0.88 (70) &   0.73 (70) &  -0.14 (5) &  -0.11 (5) &    0.74 (15) &  0.64 (15) &      0.79 (80) &  0.69 (80) &    0.47 &     0.42 \\
\hline 
 \multicolumn{19}{c}{JULE: Davies-Bouldin index} \\
 \hline 
Paired score     &  0.54 (15) &  0.38 (15) &  0.15 (50) &  0.17 (50) &  0.85 (45) &  0.67 (45) &       0.43 (10) &  0.29 (10) &   0.78 (100) &  0.56 (100) &  -0.08 (45) &   0.02 (45) &   -0.26 (40) &  -0.14 (40) &      -0.9 (20) &  -0.78 (20) &    0.19 &     0.15 \\
\textbf{ACE} (include all) &  0.98 (15) &  0.91 (15) &  0.83 (50) &  0.67 (50) &  0.83 (40) &  0.67 (40) &       0.81 (10) &  0.64 (10) &    0.85 (90) &   0.69 (90) &  -0.33 (45) &  -0.11 (45) &   -0.83 (50) &  -0.71 (50) &     -0.94 (20) &  -0.82 (20) &    0.28 &     0.24 \\
\textbf{ACE} ($\alpha_{edges}=0.1$)   &  0.98 (15) &  0.91 (15) &  0.83 (50) &  0.67 (50) &  0.87 (40) &  0.72 (40) &       0.79 (10) &   0.6 (10) &    0.85 (90) &   0.69 (90) &  -0.21 (45) &  -0.02 (45) &   -0.69 (50) &  -0.57 (50) &     -0.94 (20) &  -0.82 (20) &    0.31 &     0.27 \\
\textbf{ACE} ($\alpha_{edges}=0.05$) &  0.72 (15) &  0.64 (15) &  0.92 (50) &  0.78 (50) &  0.87 (40) &  0.72 (40) &       0.79 (10) &   0.6 (10) &    0.85 (90) &   0.69 (90) &  -0.49 (50) &  -0.38 (50) &   -0.69 (50) &  -0.57 (50) &     -0.94 (20) &  -0.82 (20) &    0.25 &     0.21 \\
\hline 
 \multicolumn{19}{c}{JULE: Silhouette score (Cosine distance)} \\
 \hline 
Paired score     &  0.99 (10) &  0.96 (10) &   0.3 (50) &  0.22 (50) &  0.72 (25) &  0.61 (25) &       0.87 (10) &  0.69 (10) &    0.98 (70) &  0.91 (70) &  -0.07 (45) &   0.07 (45) &    0.52 (25) &  0.36 (25) &     0.39 (200) &   0.2 (200) &    0.59 &     0.50 \\
\textbf{ACE} (include all) &  0.95 (10) &  0.87 (10) &  0.98 (50) &  0.94 (50) &  0.68 (45) &  0.56 (45) &       0.96 (10) &  0.87 (10) &    0.98 (70) &  0.91 (70) &  -0.22 (45) &  -0.16 (45) &    0.76 (20) &  0.57 (20) &     0.46 (180) &  0.33 (180) &    0.69 &     0.61 \\
\textbf{ACE} ($\alpha_{edges}=0.1$)   &  0.95 (10) &  0.87 (10) &  0.98 (50) &  0.94 (50) &   0.7 (45) &  0.61 (45) &       0.96 (10) &  0.87 (10) &    0.98 (70) &  0.91 (70) &  -0.07 (45) &  -0.02 (45) &    0.74 (20) &   0.5 (20) &     0.46 (180) &  0.33 (180) &    0.71 &     0.63 \\
\textbf{ACE} ($\alpha_{edges}=0.05$) &  0.95 (10) &  0.87 (10) &  0.98 (50) &  0.94 (50) &  0.83 (45) &  0.72 (45) &       0.96 (10) &  0.87 (10) &    0.98 (70) &  0.91 (70) &  -0.07 (45) &  -0.02 (45) &    0.74 (20) &   0.5 (20) &     0.46 (180) &  0.33 (180) &    0.73 &     0.64 \\
\hline 
 \multicolumn{19}{c}{JULE: Silhouette score (Euclidean distance)} \\
 \hline 
Paired score     &  0.85 (10) &  0.73 (10) &  0.33 (50) &  0.28 (50) &  0.72 (25) &  0.61 (25) &       0.88 (10) &  0.69 (10) &    0.96 (80) &  0.87 (80) &  0.07 (45) &  0.16 (45) &    0.55 (25) &  0.43 (25) &     0.44 (200) &  0.29 (200) &    0.60 &     0.51 \\
\textbf{ACE} (include all) &  0.95 (10) &  0.87 (10) &  0.98 (50) &  0.94 (50) &  0.78 (45) &  0.67 (45) &       0.95 (10) &  0.82 (10) &    0.98 (70) &  0.91 (70) &  0.14 (45) &  0.11 (45) &    0.76 (25) &  0.57 (25) &     0.47 (200) &  0.33 (200) &    0.75 &     0.65 \\
\textbf{ACE} ($\alpha_{edges}=0.1$)   &  0.95 (10) &  0.87 (10) &  0.98 (50) &  0.94 (50) &  0.78 (45) &  0.67 (45) &       0.95 (10) &  0.82 (10) &    0.98 (70) &  0.91 (70) &  0.14 (45) &  0.11 (45) &    0.71 (25) &  0.43 (25) &     0.47 (200) &  0.33 (200) &    0.74 &     0.64 \\
\textbf{ACE} ($\alpha_{edges}=0.05$) &  0.95 (10) &  0.87 (10) &  0.98 (50) &  0.94 (50) &  0.78 (45) &  0.67 (45) &       0.95 (10) &  0.82 (10) &    0.98 (70) &  0.91 (70) &  0.14 (45) &  0.11 (45) &    0.71 (25) &  0.43 (25) &     0.47 (200) &  0.33 (200) &    0.74 &     0.64 \\
\hline 
 \multicolumn{19}{c}{DEPICT: Calinski-Harabasz index} \\
 \hline 
Paired score     &  0.46 (5) &  0.6 (5) &  -0.99 (5) &  -0.96 (5) &  -0.85 (10) &  -0.72 (10) &        0.44 (5) &  0.56 (5) &   -0.92 (10) &  -0.82 (10) &         &       &           &       &             &       &   -0.37 &    -0.27 \\
\textbf{ACE} (include all) &  0.46 (5) &  0.6 (5) &  -0.65 (5) &  -0.47 (5) &  -0.75 (10) &  -0.56 (10) &        0.46 (5) &   0.6 (5) &    0.72 (80) &   0.69 (80) &         &       &           &       &             &       &    0.05 &     0.17 \\
\textbf{ACE} ($\alpha_{edges}=0.1$)   &  0.46 (5) &  0.6 (5) &  -0.66 (5) &  -0.51 (5) &   0.77 (30) &   0.61 (30) &        0.46 (5) &   0.6 (5) &    0.92 (80) &   0.82 (80) &         &       &           &       &             &       &    0.39 &     0.42 \\
\textbf{ACE} ($\alpha_{edges}=0.05$) &  0.46 (5) &  0.6 (5) &  -0.66 (5) &  -0.51 (5) &   0.87 (35) &   0.72 (35) &        0.46 (5) &   0.6 (5) &    0.92 (80) &   0.82 (80) &         &       &           &       &             &       &    0.41 &     0.45 \\
\hline 
 \multicolumn{19}{c}{DEPICT: Davies-Bouldin index} \\
 \hline 
Paired score     &   0.46 (5) &    0.6 (5) &  -0.78 (5) &  -0.64 (5) &  -0.85 (10) &  -0.72 (10) &        0.44 (5) &   0.56 (5) &    -0.1 (10) &  0.02 (10) &         &       &           &       &             &       &   -0.17 &    -0.04 \\
\textbf{ACE} (include all) &  0.61 (15) &  0.56 (15) &  0.96 (50) &  0.91 (50) &   0.88 (35) &   0.78 (35) &       0.87 (10) &  0.78 (10) &    0.95 (80) &  0.87 (80) &         &       &           &       &             &       &    0.85 &     0.78 \\
\textbf{ACE} ($\alpha_{edges}=0.1$)   &  0.62 (10) &   0.6 (10) &  0.95 (50) &  0.87 (50) &   0.77 (35) &   0.67 (35) &       0.78 (10) &  0.69 (10) &    0.96 (70) &  0.91 (70) &         &       &           &       &             &       &    0.82 &     0.75 \\
\textbf{ACE} ($\alpha_{edges}=0.05$) &  0.62 (10) &   0.6 (10) &  0.96 (50) &  0.91 (50) &   0.77 (35) &   0.67 (35) &       0.87 (10) &  0.78 (10) &     1.0 (80) &   1.0 (80) &         &       &           &       &             &       &    0.84 &     0.79 \\
\hline 
 \multicolumn{19}{c}{DEPICT: Silhouette score (Cosine distance)} \\
 \hline 
Paired score     &   0.44 (5) &   0.56 (5) &   -0.7 (5) &   -0.6 (5) &  -0.85 (10) &  -0.72 (10) &        0.44 (5) &   0.56 (5) &    0.07 (10) &  0.11 (10) &         &       &           &       &             &       &   -0.12 &    -0.02 \\
\textbf{ACE} (include all) &  0.64 (15) &   0.6 (15) &  0.82 (40) &  0.73 (40) &   0.93 (35) &   0.83 (35) &       0.93 (10) &  0.82 (10) &    0.98 (80) &  0.91 (80) &         &       &           &       &             &       &    0.86 &     0.78 \\
\textbf{ACE} ($\alpha_{edges}=0.1$)   &  0.65 (15) &  0.64 (15) &  0.87 (40) &  0.78 (40) &   0.93 (35) &   0.83 (35) &       0.85 (10) &  0.78 (10) &    0.99 (80) &  0.96 (80) &         &       &           &       &             &       &    0.86 &     0.80 \\
\textbf{ACE} ($\alpha_{edges}=0.05$) &  0.65 (15) &  0.64 (15) &  0.87 (40) &  0.78 (40) &   0.93 (35) &   0.83 (35) &       0.85 (10) &  0.78 (10) &    0.99 (80) &  0.96 (80) &         &       &           &       &             &       &    0.86 &     0.80 \\
\hline 
 \multicolumn{19}{c}{DEPICT: Silhouette score (Euclidean distance)} \\
 \hline 
Paired score     &   0.44 (5) &  0.56 (5) &  -0.61 (5) &  -0.47 (5) &  -0.85 (10) &  -0.72 (10) &        0.44 (5) &   0.56 (5) &   -0.12 (10) &  -0.02 (10) &         &       &           &       &             &       &   -0.14 &    -0.02 \\
\textbf{ACE} (include all) &  0.64 (15) &  0.6 (15) &  0.94 (40) &  0.87 (40) &    0.3 (25) &   0.22 (25) &       0.87 (10) &  0.78 (10) &    0.99 (80) &   0.96 (80) &         &       &           &       &             &       &    0.75 &     0.69 \\
\textbf{ACE} ($\alpha_{edges}=0.1$)   &   0.46 (5) &   0.6 (5) &  0.94 (40) &  0.87 (40) &   0.02 (25) &   0.06 (25) &       0.85 (10) &  0.78 (10) &    0.98 (80) &   0.91 (80) &         &       &           &       &             &       &    0.65 &     0.64 \\
\textbf{ACE} ($\alpha_{edges}=0.05$) &   0.46 (5) &   0.6 (5) &  0.94 (40) &  0.87 (40) &   0.18 (25) &   0.17 (25) &       0.85 (10) &  0.78 (10) &    0.99 (80) &   0.96 (80) &         &       &           &       &             &       &    0.68 &     0.68 \\
\bottomrule
\end{tabular}
}

 \label{tab:abs:nmi:alpha1}
\end{table*}
 %%%%%%%%%%%%%%%%%%%%%%%%%%%%%%
\begin{table*}[htbp!]
\centering
\caption{Perturbation analyses on different $\alpha_{edges}$ for Application 2. $r_s$ and $\tau_B$ between the generated scores and ACC scores are reported. Empty cells and dash marks ($-$) indicate cases where the result is either missing or impractical to obtain.}
\resizebox{\textwidth}{!}{
\begin{tabular}{lllllllllllllllllll}
\toprule
{} & \multicolumn{2}{l}{USPS (10)} & \multicolumn{2}{l}{YTF (41)} & \multicolumn{2}{l}{FRGC (20)} & \multicolumn{2}{l}{MNIST-test (10)} & \multicolumn{2}{l}{CMU-PIE (68)} & \multicolumn{2}{l}{UMist (20)} & \multicolumn{2}{l}{COIL-20 (20)} & \multicolumn{2}{l}{COIL-100 (100)} & \multicolumn{2}{l}{Average} \\
{} &     $r_s$ & $\tau_B$ &    $r_s$ & $\tau_B$ &     $r_s$ & $\tau_B$ &           $r_s$ & $\tau_B$ &        $r_s$ & $\tau_B$ &      $r_s$ & $\tau_B$ &        $r_s$ & $\tau_B$ &          $r_s$ & $\tau_B$ &   $r_s$ & $\tau_B$ \\
\midrule
\hline 
 \multicolumn{19}{c}{JULE: Calinski-Harabasz index} \\
 \hline 
Paired score     &      0.84 &     0.73 &     0.03 &    -0.06 &     -0.49 &    -0.31 &            0.61 &     0.56 &        -0.09 &    -0.07 &      -0.04 &     0.07 &         0.74 &     0.64 &           0.60 &     0.51 &    0.27 &     0.26 \\
\textbf{ACE} (include all) &      0.84 &     0.73 &     0.92 &     0.83 &     -0.11 &    -0.03 &            0.61 &     0.56 &         0.83 &     0.69 &      -0.07 &     0.02 &         0.76 &     0.71 &           0.65 &     0.56 &    0.55 &     0.51 \\
\textbf{ACE} ($\alpha_{edges}=0.1$)   &      0.84 &     0.73 &     0.92 &     0.83 &     -0.11 &    -0.03 &            0.61 &     0.56 &         0.83 &     0.69 &      -0.07 &     0.02 &         0.76 &     0.71 &           0.65 &     0.56 &    0.55 &     0.51 \\
\textbf{ACE} ($\alpha_{edges}=0.05$) &      0.84 &     0.73 &     0.92 &     0.83 &     -0.11 &    -0.03 &            0.61 &     0.56 &         0.83 &     0.69 &      -0.07 &     0.02 &         0.76 &     0.71 &           0.65 &     0.56 &    0.55 &     0.51 \\
\hline 
 \multicolumn{19}{c}{JULE: Davies-Bouldin index} \\
 \hline 
Paired score     &      0.39 &     0.29 &     0.10 &     0.06 &      0.37 &     0.25 &            0.49 &     0.33 &         0.83 &     0.60 &      -0.28 &    -0.29 &        -0.29 &    -0.21 &          -0.87 &    -0.73 &    0.09 &     0.04 \\
\textbf{ACE} (include all) &      0.89 &     0.73 &     0.80 &     0.67 &      0.63 &     0.48 &            0.84 &     0.69 &         0.88 &     0.73 &      -0.58 &    -0.42 &        -0.86 &    -0.79 &          -0.82 &    -0.69 &    0.22 &     0.18 \\
\textbf{ACE} ($\alpha_{edges}=0.1$)   &      0.89 &     0.73 &     0.80 &     0.67 &      0.60 &     0.42 &            0.83 &     0.64 &         0.88 &     0.73 &      -0.42 &    -0.33 &        -0.71 &    -0.64 &          -0.82 &    -0.69 &    0.26 &     0.19 \\
\textbf{ACE} ($\alpha_{edges}=0.05$) &      0.89 &     0.73 &     0.90 &     0.78 &      0.60 &     0.42 &            0.83 &     0.64 &         0.88 &     0.73 &      -0.83 &    -0.69 &        -0.71 &    -0.64 &          -0.82 &    -0.69 &    0.22 &     0.16 \\
\hline 
 \multicolumn{19}{c}{JULE: Silhouette score (Cosine distance)} \\
 \hline 
Paired score     &      0.89 &     0.78 &     0.27 &     0.22 &      0.21 &     0.09 &            0.81 &     0.64 &         0.99 &     0.96 &      -0.26 &    -0.24 &         0.55 &     0.43 &           0.52 &     0.33 &    0.50 &     0.40 \\
\textbf{ACE} (include all) &      0.95 &     0.87 &     0.98 &     0.94 &      0.61 &     0.48 &            0.94 &     0.82 &         0.99 &     0.96 &      -0.60 &    -0.38 &         0.79 &     0.64 &           0.60 &     0.47 &    0.66 &     0.60 \\
\textbf{ACE} ($\alpha_{edges}=0.1$)   &      0.95 &     0.87 &     0.98 &     0.94 &      0.64 &     0.54 &            0.94 &     0.82 &         0.99 &     0.96 &      -0.32 &    -0.24 &         0.76 &     0.57 &           0.60 &     0.47 &    0.69 &     0.61 \\
\textbf{ACE} ($\alpha_{edges}=0.05$) &      0.95 &     0.87 &     0.98 &     0.94 &      0.54 &     0.42 &            0.94 &     0.82 &         0.99 &     0.96 &      -0.32 &    -0.24 &         0.76 &     0.57 &           0.60 &     0.47 &    0.68 &     0.60 \\
\hline 
 \multicolumn{19}{c}{JULE: Silhouette score (Euclidean distance)} \\
 \hline 
Paired score     &      0.93 &     0.82 &     0.30 &     0.28 &      0.21 &     0.09 &            0.82 &     0.64 &         0.98 &     0.91 &      -0.13 &    -0.16 &         0.52 &     0.36 &           0.55 &     0.42 &    0.52 &     0.42 \\
\textbf{ACE} (include all) &      0.95 &     0.87 &     0.98 &     0.94 &      0.57 &     0.48 &            0.92 &     0.78 &         0.99 &     0.96 &      -0.03 &    -0.11 &         0.74 &     0.50 &           0.59 &     0.47 &    0.71 &     0.61 \\
\textbf{ACE} ($\alpha_{edges}=0.1$)   &      0.95 &     0.87 &     0.98 &     0.94 &      0.57 &     0.48 &            0.92 &     0.78 &         0.99 &     0.96 &      -0.03 &    -0.11 &         0.74 &     0.50 &           0.59 &     0.47 &    0.71 &     0.61 \\
\textbf{ACE} ($\alpha_{edges}=0.05$) &      0.95 &     0.87 &     0.98 &     0.94 &      0.57 &     0.48 &            0.92 &     0.78 &         0.99 &     0.96 &      -0.03 &    -0.11 &         0.74 &     0.50 &           0.59 &     0.47 &    0.71 &     0.61 \\
\hline 
 \multicolumn{19}{c}{DEPICT: Calinski-Harabasz index} \\
 \hline 
Paired score     &      0.88 &     0.82 &    -0.96 &    -0.91 &     -0.37 &    -0.22 &            0.79 &     0.73 &        -0.92 &    -0.82 &         &       &           &       &             &       &   -0.11 &    -0.08 \\
\textbf{ACE} (include all) &      0.88 &     0.82 &    -0.66 &    -0.51 &     -0.13 &    -0.06 &            0.82 &     0.78 &         0.72 &     0.69 &         &       &           &       &             &       &    0.32 &     0.34 \\
\textbf{ACE} ($\alpha_{edges}=0.1$)   &      0.88 &     0.82 &    -0.67 &    -0.56 &      0.92 &     0.78 &            0.82 &     0.78 &         0.92 &     0.82 &         &       &           &       &             &       &    0.57 &     0.53 \\
\textbf{ACE} ($\alpha_{edges}=0.05$) &      0.88 &     0.82 &    -0.67 &    -0.56 &      0.83 &     0.67 &            0.82 &     0.78 &         0.92 &     0.82 &         &       &           &       &             &       &    0.55 &     0.51 \\
\hline 
 \multicolumn{19}{c}{DEPICT: Davies-Bouldin index} \\
 \hline 
Paired score     &      0.88 &     0.82 &    -0.77 &    -0.60 &     -0.37 &    -0.22 &            0.79 &     0.73 &        -0.10 &     0.02 &         &       &           &       &             &       &    0.09 &     0.15 \\
\textbf{ACE} (include all) &      0.92 &     0.78 &     0.99 &     0.96 &      0.87 &     0.72 &            0.89 &     0.78 &         0.95 &     0.87 &         &       &           &       &             &       &    0.92 &     0.82 \\
\textbf{ACE} ($\alpha_{edges}=0.1$)   &      0.93 &     0.82 &     0.96 &     0.91 &      0.92 &     0.83 &            0.93 &     0.87 &         0.96 &     0.91 &         &       &           &       &             &       &    0.94 &     0.87 \\
\textbf{ACE} ($\alpha_{edges}=0.05$) &      0.93 &     0.82 &     0.99 &     0.96 &      0.92 &     0.83 &            0.89 &     0.78 &         1.00 &     1.00 &         &       &           &       &             &       &    0.94 &     0.88 \\
\hline 
 \multicolumn{19}{c}{DEPICT: Silhouette score (Cosine distance)} \\
 \hline 
Paired score     &      0.87 &     0.78 &    -0.69 &    -0.56 &     -0.37 &    -0.22 &            0.79 &     0.73 &         0.07 &     0.11 &         &       &           &       &             &       &    0.14 &     0.17 \\
\textbf{ACE} (include all) &      0.94 &     0.82 &     0.87 &     0.78 &      0.80 &     0.67 &            0.90 &     0.82 &         0.98 &     0.91 &         &       &           &       &             &       &    0.90 &     0.80 \\
\textbf{ACE} ($\alpha_{edges}=0.1$)   &      0.95 &     0.87 &     0.92 &     0.82 &      0.80 &     0.67 &            0.95 &     0.87 &         0.99 &     0.96 &         &       &           &       &             &       &    0.92 &     0.84 \\
\textbf{ACE} ($\alpha_{edges}=0.05$) &      0.95 &     0.87 &     0.92 &     0.82 &      0.80 &     0.67 &            0.95 &     0.87 &         0.99 &     0.96 &         &       &           &       &             &       &    0.92 &     0.84 \\
\hline 
 \multicolumn{19}{c}{DEPICT: Silhouette score (Euclidean distance)} \\
 \hline 
Paired score     &      0.87 &     0.78 &    -0.64 &    -0.51 &     -0.37 &    -0.22 &            0.79 &     0.73 &        -0.12 &    -0.02 &         &       &           &       &             &       &    0.11 &     0.15 \\
\textbf{ACE} (include all) &      0.94 &     0.82 &     0.98 &     0.91 &      0.88 &     0.72 &            0.89 &     0.78 &         0.99 &     0.96 &         &       &           &       &             &       &    0.94 &     0.84 \\
\textbf{ACE} ($\alpha_{edges}=0.1$)   &      0.88 &     0.82 &     0.98 &     0.91 &      0.73 &     0.56 &            0.95 &     0.87 &         0.98 &     0.91 &         &       &           &       &             &       &    0.90 &     0.81 \\
\textbf{ACE} ($\alpha_{edges}=0.05$) &      0.88 &     0.82 &     0.98 &     0.91 &      0.83 &     0.67 &            0.95 &     0.87 &         0.99 &     0.96 &         &       &           &       &             &       &    0.93 &     0.84 \\
\bottomrule
\end{tabular}
}
 \label{tab:abs:acc:alpha1}
\end{table*}
%%%%%%%%%%%%%%%%%%%%%%%%%%%%%%%%%%%%%%%%%%%%%%%%%%%%%%%%%%%%%%%%%%%%%%%%%%%%%%%%%%%%%%%%%
%%%%%%%%%%%%%%%%%%%%%%%%%%%%%%%%%%%%%%%%%%%%%%%%%%%%%%%%%%%%%%%%%%%%%%%%%%%%%%%%%%%%%%%%%
%%%%%%%%%%%%%%%%%%%%%%%%%%%%%%%%%%%%%%%%%%%%%%%%%%%%%%%%%%%%%%%%%%%%%%%%%%%%%%%%%%%%%%%%%

\clearpage
\newpage \subsection{HDBSCAN vs. DBSCAN} \label{app:exp:ab:db}
In Algorithm \ref{Algorithm1}, we employ a density-based clustering approach to group embedding spaces based on their rank correlation. Density-based methods are advantageous as they do not necessitate prior knowledge of the number of groups and can identify outlier spaces with low rank correlation. In the main text, we present the results using HDBSCAN, a density-based clustering algorithm that requires minimal parameter tuning compared to alternatives like DBSCAN. In this section, we extend our exploration of clustering based on rank correlation in Phase 1 by conducting additional experiments with the DBSCAN function from scikit-learn. Specifically, we vary the critical parameter $eps$, which plays a pivotal role in DBSCAN, setting it to $0.1$ and $0.2$ respectively. Since the DBSCAN function does not provide outlier scores, these experiments omit the post hoc refinement step (i.e., the score-based filtering technique used with HBSCAN). These results are compared with HDBSCAN. Tables \ref{tab:abs:nmi:db0} and \ref{tab:abs:acc:db0} showcase the evaluation performance for Application 1, while Tables \ref{tab:abs:nmi:db1} and \ref{tab:abs:acc:db1} present the results for Application 2. Our observations reveal that, in certain cases (e.g., JULE for Application 1), DBSCAN can even yield higher correlations with NMI and ACC. Conversely, in other scenarios, HDBSCAN outperforms (e.g., JULE for Application 2). Considering the advantage of not needing to tune the parameter $eps$, we opt for HDBSCAN as the grouping method and report its performance in the main text.

\begin{table*}[htbp!]
\centering
\caption{Perturbation analyses comparing HDBSCAN and DBSCAN for Application 1. $r_s$ and $\tau_B$ between the generated scores and NMI scores are reported. Empty cells and dash marks ($-$) indicate cases where the result is either missing or impractical to obtain.}
\resizebox{\textwidth}{!}{
\begin{tabular}{lllllllllllllllllll}
\toprule
{} & \multicolumn{2}{l}{USPS} & \multicolumn{2}{l}{YTF} & \multicolumn{2}{l}{FRGC} & \multicolumn{2}{l}{MNIST-test} & \multicolumn{2}{l}{CMU-PIE} & \multicolumn{2}{l}{UMist} & \multicolumn{2}{l}{COIL-20} & \multicolumn{2}{l}{COIL-100} & \multicolumn{2}{l}{Average} \\
{} & $r_s$ & $\tau_B$ & $r_s$ & $\tau_B$ & $r_s$ & $\tau_B$ &      $r_s$ & $\tau_B$ &   $r_s$ & $\tau_B$ & $r_s$ & $\tau_B$ &   $r_s$ & $\tau_B$ &    $r_s$ & $\tau_B$ &   $r_s$ & $\tau_B$ \\
\midrule
\hline 
 \multicolumn{19}{c}{JULE: Calinski-Harabasz index} \\
 \hline 
Paired score     &  0.17 &     0.13 &  0.52 &     0.40 & -0.13 &    -0.10 &       0.49 &     0.34 &   -0.13 &    -0.08 &  0.70 &     0.50 &    0.53 &     0.38 &     0.20 &     0.19 &    0.29 &     0.22 \\
\textbf{ACE} ($DBSCAN_{eps=0.1}$) &  0.74 &     0.59 &  0.88 &     0.70 &  0.37 &     0.25 &       0.87 &     0.71 &    0.96 &     0.85 &  0.88 &     0.68 &    0.93 &     0.78 &     0.95 &     0.82 &    0.82 &     0.67 \\
\textbf{ACE} ($DBSCAN_{eps=0.2}$) &  0.74 &     0.59 &  0.71 &     0.54 &  0.08 &     0.04 &       0.87 &     0.71 &    0.96 &     0.85 &  0.87 &     0.68 &    0.92 &     0.76 &     0.94 &     0.80 &    0.76 &     0.62 \\
\textbf{ACE} ($HDBSCAN$)   &  0.80 &     0.63 &  0.90 &     0.73 &  0.39 &     0.26 &       0.87 &     0.71 &    0.98 &     0.90 &  0.81 &     0.61 &    0.60 &     0.45 &     0.95 &     0.82 &    0.79 &     0.64 \\
\hline 
 \multicolumn{19}{c}{JULE: Davies-Bouldin index} \\
 \hline 
Paired score     & -0.10 &    -0.03 & -0.32 &    -0.21 & -0.08 &    -0.05 &      -0.13 &    -0.06 &    0.26 &     0.20 &  0.62 &     0.44 &    0.61 &     0.42 &     0.43 &     0.35 &    0.16 &     0.13 \\
\textbf{ACE} ($DBSCAN_{eps=0.1}$) & -0.14 &    -0.07 & -0.57 &    -0.40 &  0.48 &     0.32 &       0.73 &     0.55 &    0.96 &     0.87 &  0.59 &     0.41 &    0.29 &     0.26 &    -0.48 &    -0.34 &    0.23 &     0.20 \\
\textbf{ACE} ($DBSCAN_{eps=0.2}$) &  0.01 &     0.05 & -0.54 &    -0.39 &  0.22 &     0.16 &       0.73 &     0.55 &    0.96 &     0.87 &  0.59 &     0.41 &    0.26 &     0.25 &    -0.41 &    -0.29 &    0.23 &     0.20 \\
\textbf{ACE} ($HDBSCAN$)   & -0.08 &    -0.02 & -0.30 &    -0.21 &  0.22 &     0.16 &       0.73 &     0.55 &    0.10 &     0.06 &  0.38 &     0.27 &    0.23 &     0.22 &     0.48 &     0.33 &    0.22 &     0.17 \\
\hline 
 \multicolumn{19}{c}{JULE: Silhouette score (Cosine distance)} \\
 \hline 
Paired score     &  0.28 &     0.22 &  0.73 &     0.56 &  0.09 &     0.06 &       0.63 &     0.47 &    0.50 &     0.36 &  0.71 &     0.50 &    0.68 &     0.50 &     0.74 &     0.54 &    0.54 &     0.40 \\
\textbf{ACE} ($DBSCAN_{eps=0.1}$) &  0.89 &     0.73 &  0.92 &     0.80 &  0.58 &     0.40 &       0.81 &     0.66 &    0.57 &     0.49 &  0.87 &     0.70 &    0.92 &     0.78 &     0.92 &     0.78 &    0.81 &     0.67 \\
\textbf{ACE} ($DBSCAN_{eps=0.2}$) &  0.89 &     0.73 &  0.92 &     0.80 &  0.52 &     0.35 &       0.81 &     0.66 &    0.97 &     0.90 &  0.88 &     0.70 &    0.44 &     0.38 &     0.92 &     0.78 &    0.79 &     0.66 \\
\textbf{ACE} ($HDBSCAN$)   &  0.89 &     0.73 &  0.93 &     0.83 &  0.52 &     0.35 &       0.81 &     0.66 &    0.99 &     0.93 &  0.79 &     0.59 &    0.44 &     0.38 &     0.92 &     0.78 &    0.79 &     0.66 \\
\hline 
 \multicolumn{19}{c}{JULE: Silhouette score (Euclidean distance)} \\
 \hline 
Paired score     &  0.27 &     0.20 &  0.72 &     0.55 &  0.04 &     0.03 &       0.56 &     0.41 &    0.42 &     0.30 &  0.70 &     0.50 &    0.64 &     0.46 &     0.55 &     0.41 &    0.49 &     0.36 \\
\textbf{ACE} ($DBSCAN_{eps=0.1}$) &  0.88 &     0.72 &  0.90 &     0.77 &  0.58 &     0.41 &       0.81 &     0.65 &    0.99 &     0.93 &  0.88 &     0.70 &    0.92 &     0.77 &     0.91 &     0.78 &    0.86 &     0.71 \\
\textbf{ACE} ($DBSCAN_{eps=0.2}$) &  0.88 &     0.72 &  0.90 &     0.77 &  0.53 &     0.36 &       0.81 &     0.65 &    0.99 &     0.93 &  0.89 &     0.70 &    0.41 &     0.36 &     0.92 &     0.78 &    0.79 &     0.66 \\
\textbf{ACE} ($HDBSCAN$)   &  0.88 &     0.72 &  0.89 &     0.75 &  0.42 &     0.28 &       0.81 &     0.65 &    0.98 &     0.90 &  0.88 &     0.70 &    0.41 &     0.36 &     0.92 &     0.78 &    0.77 &     0.64 \\
\hline 
 \multicolumn{19}{c}{DEPICT: Calinski-Harabasz index} \\
 \hline 
Paired score     &  0.76 &     0.57 &  0.44 &     0.26 &  0.76 &     0.57 &       0.89 &     0.72 &    0.49 &     0.44 &    &       &      &       &       &       &    0.67 &     0.51 \\
\textbf{ACE} ($DBSCAN_{eps=0.1}$) &  0.91 &     0.77 &  0.58 &     0.44 &  0.94 &     0.82 &       0.96 &     0.87 &    0.97 &     0.90 &    &       &      &       &       &       &    0.87 &     0.76 \\
\textbf{ACE} ($DBSCAN_{eps=0.2}$) &  0.91 &     0.77 &  0.67 &     0.54 &  0.91 &     0.79 &       0.96 &     0.87 &    0.96 &     0.87 &    &       &      &       &       &       &    0.88 &     0.77 \\
\textbf{ACE} ($HDBSCAN$)   &  0.91 &     0.77 &  0.56 &     0.44 &  0.94 &     0.82 &       0.96 &     0.87 &    0.96 &     0.87 &    &       &      &       &       &       &    0.87 &     0.75 \\
\hline 
 \multicolumn{19}{c}{DEPICT: Davies-Bouldin index} \\
 \hline 
Paired score     &  0.81 &     0.59 &  0.45 &     0.31 &  0.90 &     0.74 &       0.89 &     0.72 &    0.63 &     0.59 &    &       &      &       &       &       &    0.73 &     0.59 \\
\textbf{ACE} ($DBSCAN_{eps=0.1}$) &  0.90 &     0.79 &  0.57 &     0.42 &  0.92 &     0.80 &       0.95 &     0.83 &    0.99 &     0.95 &    &       &      &       &       &       &    0.87 &     0.76 \\
\textbf{ACE} ($DBSCAN_{eps=0.2}$) &  0.95 &     0.86 &  0.54 &     0.39 &  0.91 &     0.79 &       0.95 &     0.83 &    0.98 &     0.92 &    &       &      &       &       &       &    0.87 &     0.76 \\
\textbf{ACE} ($HDBSCAN$)   &  0.91 &     0.82 &  0.76 &     0.58 &  0.91 &     0.79 &       0.96 &     0.87 &    0.98 &     0.92 &    &       &      &       &       &       &    0.90 &     0.80 \\
\hline 
 \multicolumn{19}{c}{DEPICT: Silhouette score (Cosine distance)} \\
 \hline 
Paired score     &  0.81 &     0.62 &  0.45 &     0.33 &  0.90 &     0.75 &       0.89 &     0.72 &    0.77 &     0.58 &    &       &      &       &       &       &    0.76 &     0.60 \\
\textbf{ACE} ($DBSCAN_{eps=0.1}$) &  0.97 &     0.90 &  0.59 &     0.48 &  0.95 &     0.83 &       0.98 &     0.91 &    0.94 &     0.84 &    &       &      &       &       &       &    0.89 &     0.79 \\
\textbf{ACE} ($DBSCAN_{eps=0.2}$) &  0.96 &     0.87 &  0.62 &     0.49 &  0.95 &     0.83 &       0.98 &     0.91 &    0.94 &     0.83 &    &       &      &       &       &       &    0.89 &     0.79 \\
\textbf{ACE} ($HDBSCAN$)   &  0.97 &     0.90 &  0.71 &     0.56 &  0.94 &     0.82 &       0.97 &     0.90 &    0.94 &     0.83 &    &       &      &       &       &       &    0.91 &     0.80 \\
\hline 
 \multicolumn{19}{c}{DEPICT: Silhouette score (Euclidean distance)} \\
 \hline 
Paired score     &  0.73 &     0.50 &  0.47 &     0.36 &  0.79 &     0.65 &       0.86 &     0.69 &    0.59 &     0.52 &    &       &      &       &       &       &    0.69 &     0.54 \\
\textbf{ACE} ($DBSCAN_{eps=0.1}$) &  0.97 &     0.88 &  0.58 &     0.45 &  0.95 &     0.83 &       0.98 &     0.90 &    0.94 &     0.82 &    &       &      &       &       &       &    0.88 &     0.78 \\
\textbf{ACE} ($DBSCAN_{eps=0.2}$) &  0.97 &     0.88 &  0.62 &     0.48 &  0.95 &     0.84 &       0.98 &     0.90 &    0.94 &     0.82 &    &       &      &       &       &       &    0.89 &     0.78 \\
\textbf{ACE} ($HDBSCAN$)   &  0.97 &     0.88 &  0.65 &     0.50 &  0.95 &     0.83 &       0.98 &     0.90 &    0.94 &     0.82 &    &       &      &       &       &       &    0.90 &     0.79 \\
\bottomrule
\end{tabular}
}
 \label{tab:abs:nmi:db0}
\end{table*}

%%%%%%%%%%%%%%%%%%%%%%%%%%%%%%%%%%%%%%%%%%%%%%%%%%%%%%%%%%%%%%%%%%%%%%%%%%%%%%%%%%%%%%%%%
\begin{table*}[htbp!]
\centering
\caption{Perturbation analyses comparing HDBSCAN and DBSCAN for Application 1. $r_s$ and $\tau_B$ between the generated scores and ACC scores are reported. Empty cells and dash marks ($-$) indicate cases where the result is either missing or impractical to obtain.}
\resizebox{\textwidth}{!}{
\begin{tabular}{lllllllllllllllllll}
\toprule
{} & \multicolumn{2}{l}{USPS} & \multicolumn{2}{l}{YTF} & \multicolumn{2}{l}{FRGC} & \multicolumn{2}{l}{MNIST-test} & \multicolumn{2}{l}{CMU-PIE} & \multicolumn{2}{l}{UMist} & \multicolumn{2}{l}{COIL-20} & \multicolumn{2}{l}{COIL-100} & \multicolumn{2}{l}{Average} \\
{} & $r_s$ & $\tau_B$ & $r_s$ & $\tau_B$ & $r_s$ & $\tau_B$ &      $r_s$ & $\tau_B$ &   $r_s$ & $\tau_B$ & $r_s$ & $\tau_B$ &   $r_s$ & $\tau_B$ &    $r_s$ & $\tau_B$ &   $r_s$ & $\tau_B$ \\
\midrule
\hline 
 \multicolumn{19}{c}{JULE: Calinski-Harabasz index} \\
 \hline 
Paired score     &  0.04 &     0.05 &  0.39 &     0.27 & -0.26 &    -0.18 &       0.31 &     0.21 &   -0.20 &    -0.12 &  0.64 &     0.45 &    0.57 &     0.40 &     0.09 &     0.08 &    0.20 &     0.14 \\
\textbf{ACE} ($DBSCAN_{eps=0.1}$) &  0.71 &     0.58 &  0.74 &     0.55 &  0.49 &     0.38 &       0.95 &     0.82 &    0.94 &     0.82 &  0.88 &     0.69 &    0.90 &     0.74 &     0.93 &     0.81 &    0.82 &     0.67 \\
\textbf{ACE} ($DBSCAN_{eps=0.2}$) &  0.71 &     0.58 &  0.61 &     0.46 &  0.13 &     0.09 &       0.95 &     0.82 &    0.94 &     0.82 &  0.87 &     0.69 &    0.89 &     0.72 &     0.92 &     0.79 &    0.75 &     0.62 \\
\textbf{ACE} ($HDBSCAN$)   &  0.90 &     0.77 &  0.73 &     0.54 &  0.49 &     0.36 &       0.95 &     0.82 &    0.97 &     0.87 &  0.81 &     0.61 &    0.57 &     0.40 &     0.93 &     0.81 &    0.79 &     0.65 \\
\hline 
 \multicolumn{19}{c}{JULE: Davies-Bouldin index} \\
 \hline 
Paired score     & -0.27 &    -0.15 & -0.14 &    -0.09 & -0.23 &    -0.14 &      -0.35 &    -0.19 &    0.20 &     0.16 &  0.53 &     0.36 &    0.63 &     0.44 &     0.33 &     0.26 &    0.09 &     0.08 \\
\textbf{ACE} ($DBSCAN_{eps=0.1}$) & -0.36 &    -0.14 & -0.43 &    -0.30 &  0.83 &     0.64 &       0.79 &     0.64 &    0.95 &     0.85 &  0.50 &     0.36 &    0.27 &     0.23 &    -0.46 &    -0.32 &    0.26 &     0.24 \\
\textbf{ACE} ($DBSCAN_{eps=0.2}$) & -0.28 &    -0.12 & -0.42 &    -0.29 &  0.53 &     0.38 &       0.79 &     0.64 &    0.95 &     0.85 &  0.50 &     0.36 &    0.23 &     0.21 &    -0.42 &    -0.29 &    0.23 &     0.22 \\
\textbf{ACE} ($HDBSCAN$)   & -0.30 &    -0.09 & -0.07 &    -0.07 &  0.53 &     0.38 &       0.79 &     0.64 &    0.07 &     0.03 &  0.27 &     0.20 &    0.21 &     0.18 &     0.44 &     0.28 &    0.24 &     0.19 \\
\hline 
 \multicolumn{19}{c}{JULE: Silhouette score (Cosine distance)} \\
 \hline 
Paired score     &  0.17 &     0.14 &  0.59 &     0.41 &  0.07 &     0.06 &       0.47 &     0.33 &    0.45 &     0.33 &  0.64 &     0.46 &    0.70 &     0.51 &     0.64 &     0.45 &    0.47 &     0.34 \\
\textbf{ACE} ($DBSCAN_{eps=0.1}$) &  0.96 &     0.85 &  0.73 &     0.55 &  0.88 &     0.69 &       0.92 &     0.78 &    0.58 &     0.52 &  0.85 &     0.67 &    0.90 &     0.72 &     0.85 &     0.68 &    0.83 &     0.68 \\
\textbf{ACE} ($DBSCAN_{eps=0.2}$) &  0.96 &     0.85 &  0.73 &     0.55 &  0.82 &     0.65 &       0.92 &     0.78 &    0.98 &     0.90 &  0.87 &     0.68 &    0.41 &     0.32 &     0.84 &     0.68 &    0.82 &     0.67 \\
\textbf{ACE} ($HDBSCAN$)   &  0.96 &     0.85 &  0.74 &     0.55 &  0.82 &     0.65 &       0.92 &     0.78 &    0.98 &     0.92 &  0.78 &     0.58 &    0.41 &     0.32 &     0.84 &     0.68 &    0.81 &     0.67 \\
\hline 
 \multicolumn{19}{c}{JULE: Silhouette score (Euclidean distance)} \\
 \hline 
Paired score     &  0.14 &     0.12 &  0.54 &     0.39 & -0.08 &    -0.02 &       0.41 &     0.27 &    0.36 &     0.27 &  0.64 &     0.46 &    0.67 &     0.48 &     0.44 &     0.31 &    0.39 &     0.28 \\
\textbf{ACE} ($DBSCAN_{eps=0.1}$) &  0.93 &     0.78 &  0.66 &     0.49 &  0.88 &     0.68 &       0.92 &     0.78 &    0.99 &     0.93 &  0.86 &     0.68 &    0.89 &     0.71 &     0.82 &     0.67 &    0.87 &     0.71 \\
\textbf{ACE} ($DBSCAN_{eps=0.2}$) &  0.93 &     0.78 &  0.66 &     0.49 &  0.82 &     0.63 &       0.92 &     0.78 &    0.99 &     0.93 &  0.88 &     0.69 &    0.39 &     0.30 &     0.84 &     0.68 &    0.80 &     0.66 \\
\textbf{ACE} ($HDBSCAN$)   &  0.93 &     0.78 &  0.63 &     0.48 &  0.71 &     0.53 &       0.92 &     0.78 &    0.98 &     0.91 &  0.86 &     0.68 &    0.39 &     0.30 &     0.84 &     0.68 &    0.78 &     0.64 \\
\hline 
 \multicolumn{19}{c}{DEPICT: Calinski-Harabasz index} \\
 \hline 
Paired score     &  0.56 &     0.40 &  0.54 &     0.35 &  0.76 &     0.57 &       0.88 &     0.69 &    0.48 &     0.43 &    &       &      &       &       &       &    0.64 &     0.49 \\
\textbf{ACE} ($DBSCAN_{eps=0.1}$) &  0.82 &     0.72 &  0.53 &     0.40 &  0.91 &     0.82 &       0.95 &     0.86 &    0.98 &     0.92 &    &       &      &       &       &       &    0.84 &     0.74 \\
\textbf{ACE} ($DBSCAN_{eps=0.2}$) &  0.82 &     0.72 &  0.59 &     0.45 &  0.93 &     0.82 &       0.95 &     0.86 &    0.96 &     0.87 &    &       &      &       &       &       &    0.85 &     0.74 \\
\textbf{ACE} ($HDBSCAN$)   &  0.82 &     0.72 &  0.61 &     0.45 &  0.91 &     0.82 &       0.97 &     0.91 &    0.96 &     0.87 &    &       &      &       &       &       &    0.86 &     0.75 \\
\hline 
 \multicolumn{19}{c}{DEPICT: Davies-Bouldin index} \\
 \hline 
Paired score     &  0.61 &     0.42 &  0.48 &     0.32 &  0.92 &     0.74 &       0.88 &     0.69 &    0.62 &     0.56 &    &       &      &       &       &       &    0.70 &     0.55 \\
\textbf{ACE} ($DBSCAN_{eps=0.1}$) &  0.99 &     0.96 &  0.50 &     0.39 &  0.90 &     0.75 &       0.99 &     0.92 &    0.97 &     0.89 &    &       &      &       &       &       &    0.87 &     0.78 \\
\textbf{ACE} ($DBSCAN_{eps=0.2}$) &  0.96 &     0.87 &  0.44 &     0.32 &  0.91 &     0.77 &       0.99 &     0.92 &    0.96 &     0.87 &    &       &      &       &       &       &    0.85 &     0.75 \\
\textbf{ACE} ($HDBSCAN$)   &  0.99 &     0.96 &  0.65 &     0.46 &  0.90 &     0.74 &       0.99 &     0.96 &    0.96 &     0.87 &    &       &      &       &       &       &    0.90 &     0.80 \\
\hline 
 \multicolumn{19}{c}{DEPICT: Silhouette score (Cosine distance)} \\
 \hline 
Paired score     &  0.62 &     0.45 &  0.53 &     0.42 &  0.91 &     0.75 &       0.88 &     0.69 &    0.77 &     0.58 &    &       &      &       &       &       &    0.74 &     0.58 \\
\textbf{ACE} ($DBSCAN_{eps=0.1}$) &  0.95 &     0.88 &  0.55 &     0.44 &  0.94 &     0.80 &       0.96 &     0.90 &    0.95 &     0.84 &    &       &      &       &       &       &    0.87 &     0.77 \\
\textbf{ACE} ($DBSCAN_{eps=0.2}$) &  0.96 &     0.88 &  0.59 &     0.45 &  0.94 &     0.80 &       0.96 &     0.90 &    0.94 &     0.83 &    &       &      &       &       &       &    0.88 &     0.77 \\
\textbf{ACE} ($HDBSCAN$)   &  0.95 &     0.88 &  0.70 &     0.54 &  0.91 &     0.77 &       0.96 &     0.88 &    0.94 &     0.83 &    &       &      &       &       &       &    0.89 &     0.78 \\
\hline 
 \multicolumn{19}{c}{DEPICT: Silhouette score (Euclidean distance)} \\
 \hline 
Paired score     &  0.52 &     0.33 &  0.57 &     0.45 &  0.80 &     0.62 &       0.85 &     0.65 &    0.59 &     0.48 &    &       &      &       &       &       &    0.67 &     0.51 \\
\textbf{ACE} ($DBSCAN_{eps=0.1}$) &  0.95 &     0.87 &  0.56 &     0.44 &  0.94 &     0.80 &       0.97 &     0.91 &    0.95 &     0.84 &    &       &      &       &       &       &    0.87 &     0.77 \\
\textbf{ACE} ($DBSCAN_{eps=0.2}$) &  0.95 &     0.87 &  0.60 &     0.46 &  0.94 &     0.82 &       0.97 &     0.91 &    0.95 &     0.84 &    &       &      &       &       &       &    0.88 &     0.78 \\
\textbf{ACE} ($HDBSCAN$)   &  0.95 &     0.87 &  0.63 &     0.49 &  0.91 &     0.78 &       0.97 &     0.91 &    0.95 &     0.84 &    &       &      &       &       &       &    0.88 &     0.78 \\
\bottomrule
\end{tabular}
}
 \label{tab:abs:acc:db0}
\end{table*}
%\newpage \subsection{Determination of the number of clusters  - HDBSCAN vs. DBSCAN}
\begin{table*}[htbp!]
\centering
\caption{Perturbation analyses comparing HDBSCAN and DBSCAN for Application 2. $r_s$ and $\tau_B$ between the generated scores and NMI scores are reported. Empty cells and dash marks ($-$) indicate cases where the result is either missing or impractical to obtain.}
\resizebox{\textwidth}{!}{
\begin{tabular}{lllllllllllllllllll}
\toprule
{} & \multicolumn{2}{l}{USPS (10)} & \multicolumn{2}{l}{YTF (41)} & \multicolumn{2}{l}{FRGC (20)} & \multicolumn{2}{l}{MNIST-test (10)} & \multicolumn{2}{l}{CMU-PIE (68)} & \multicolumn{2}{l}{UMist (20)} & \multicolumn{2}{l}{COIL-20 (20)} & \multicolumn{2}{l}{COIL-100 (100)} & \multicolumn{2}{l}{Average} \\
{} &      $r_s$ &   $\tau_B$ &      $r_s$ &   $\tau_B$ &       $r_s$ &    $\tau_B$ &           $r_s$ &  $\tau_B$ &        $r_s$ &    $\tau_B$ &      $r_s$ &   $\tau_B$ &        $r_s$ &   $\tau_B$ &          $r_s$ &   $\tau_B$ &   $r_s$ & $\tau_B$ \\
\midrule
\hline 
 \multicolumn{19}{c}{JULE: Calinski-Harabasz index} \\
 \hline 
Paired score     &  0.65 (10) &  0.64 (10) &   0.1 (50) &  0.06 (50) &  -0.93 (15) &  -0.83 (15) &       0.64 (10) &  0.6 (10) &   -0.03 (20) &  -0.02 (20) &  -0.13 (5) &  -0.07 (5) &    0.76 (15) &  0.71 (15) &      0.74 (80) &  0.56 (80) &    0.22 &     0.21 \\
\textbf{ACE} ($DBSCAN_{eps=0.1}$) &  0.65 (10) &  0.64 (10) &  0.93 (50) &  0.83 (50) &  -0.87 (15) &  -0.72 (15) &       0.64 (10) &  0.6 (10) &    0.88 (70) &   0.73 (70) &  -0.13 (5) &  -0.07 (5) &    0.74 (15) &  0.64 (15) &      0.72 (80) &  0.64 (80) &    0.45 &     0.41 \\
\textbf{ACE} ($DBSCAN_{eps=0.2}$) &  0.65 (10) &  0.64 (10) &   0.3 (20) &  0.17 (20) &  -0.87 (15) &  -0.72 (15) &       0.64 (10) &  0.6 (10) &    0.88 (70) &   0.73 (70) &  -0.14 (5) &  -0.11 (5) &    0.74 (15) &  0.64 (15) &      0.72 (80) &  0.64 (80) &    0.36 &     0.32 \\
\textbf{ACE} ($HDBSCAN$)   &  0.65 (10) &  0.64 (10) &  0.93 (50) &  0.83 (50) &  -0.72 (15) &  -0.67 (15) &       0.64 (10) &  0.6 (10) &    0.88 (70) &   0.73 (70) &  -0.14 (5) &  -0.11 (5) &    0.74 (15) &  0.64 (15) &      0.79 (80) &  0.69 (80) &    0.47 &     0.42 \\
\hline 
 \multicolumn{19}{c}{JULE: Davies-Bouldin index} \\
 \hline 
Paired score     &  0.54 (15) &  0.38 (15) &  0.15 (50) &  0.17 (50) &  0.85 (45) &  0.67 (45) &       0.43 (10) &  0.29 (10) &   0.78 (100) &  0.56 (100) &  -0.08 (45) &   0.02 (45) &   -0.26 (40) &  -0.14 (40) &      -0.9 (20) &  -0.78 (20) &    0.19 &     0.15 \\
\textbf{ACE} ($DBSCAN_{eps=0.1}$) &  0.73 (10) &  0.69 (10) &  0.92 (50) &  0.78 (50) &  0.87 (40) &  0.72 (40) &       0.65 (25) &  0.51 (25) &    0.85 (90) &   0.69 (90) &    -0.6 (5) &   -0.47 (5) &   -0.67 (50) &   -0.5 (50) &     -0.95 (20) &  -0.87 (20) &    0.22 &     0.19 \\
\textbf{ACE} ($DBSCAN_{eps=0.2}$) &  0.73 (10) &  0.69 (10) &  0.32 (20) &  0.17 (20) &  0.87 (40) &  0.72 (40) &       0.65 (25) &  0.51 (25) &    0.82 (90) &   0.64 (90) &  -0.49 (50) &  -0.38 (50) &   -0.67 (50) &   -0.5 (50) &     -0.94 (20) &  -0.82 (20) &    0.16 &     0.13 \\
\textbf{ACE} ($HDBSCAN$)   &  0.98 (15) &  0.91 (15) &  0.83 (50) &  0.67 (50) &  0.87 (40) &  0.72 (40) &       0.79 (10) &   0.6 (10) &    0.85 (90) &   0.69 (90) &  -0.21 (45) &  -0.02 (45) &   -0.69 (50) &  -0.57 (50) &     -0.94 (20) &  -0.82 (20) &    0.31 &     0.27 \\
\hline 
 \multicolumn{19}{c}{JULE: Silhouette score (Cosine distance)} \\
 \hline 
Paired score     &  0.99 (10) &  0.96 (10) &   0.3 (50) &  0.22 (50) &  0.72 (25) &  0.61 (25) &       0.87 (10) &  0.69 (10) &    0.98 (70) &  0.91 (70) &  -0.07 (45) &   0.07 (45) &    0.52 (25) &  0.36 (25) &     0.39 (200) &   0.2 (200) &    0.59 &     0.50 \\
\textbf{ACE} ($DBSCAN_{eps=0.1}$) &  0.92 (10) &  0.82 (10) &  0.98 (50) &  0.94 (50) &  0.88 (45) &  0.78 (45) &       0.98 (10) &  0.91 (10) &    0.98 (70) &  0.91 (70) &   -0.48 (5) &   -0.38 (5) &    0.69 (20) &  0.43 (20) &     0.46 (180) &  0.33 (180) &    0.68 &     0.59 \\
\textbf{ACE} ($DBSCAN_{eps=0.2}$) &  0.92 (10) &  0.82 (10) &  0.78 (50) &  0.67 (50) &   0.7 (45) &  0.61 (45) &       0.96 (10) &  0.87 (10) &    0.98 (70) &  0.91 (70) &   -0.48 (5) &   -0.38 (5) &    0.69 (20) &  0.43 (20) &     0.46 (180) &  0.33 (180) &    0.63 &     0.53 \\
\textbf{ACE} ($HDBSCAN$)   &  0.95 (10) &  0.87 (10) &  0.98 (50) &  0.94 (50) &   0.7 (45) &  0.61 (45) &       0.96 (10) &  0.87 (10) &    0.98 (70) &  0.91 (70) &  -0.07 (45) &  -0.02 (45) &    0.74 (20) &   0.5 (20) &     0.46 (180) &  0.33 (180) &    0.71 &     0.63 \\
\hline 
 \multicolumn{19}{c}{JULE: Silhouette score (Euclidean distance)} \\
 \hline 
Paired score     &  0.85 (10) &  0.73 (10) &  0.33 (50) &  0.28 (50) &  0.72 (25) &  0.61 (25) &       0.88 (10) &  0.69 (10) &    0.96 (80) &  0.87 (80) &  0.07 (45) &  0.16 (45) &    0.55 (25) &  0.43 (25) &     0.44 (200) &  0.29 (200) &    0.60 &     0.51 \\
\textbf{ACE} ($DBSCAN_{eps=0.1}$) &  0.79 (10) &  0.73 (10) &  0.98 (50) &  0.94 (50) &  0.83 (45) &  0.72 (45) &       0.92 (10) &  0.82 (10) &    0.98 (70) &  0.91 (70) &  -0.69 (5) &  -0.51 (5) &    0.71 (25) &  0.43 (25) &     0.47 (200) &  0.33 (200) &    0.62 &     0.55 \\
\textbf{ACE} ($DBSCAN_{eps=0.2}$) &  0.79 (10) &  0.73 (10) &  0.98 (50) &  0.94 (50) &  0.65 (45) &  0.56 (45) &       0.92 (10) &  0.82 (10) &    0.98 (70) &  0.91 (70) &  -0.69 (5) &  -0.51 (5) &    0.71 (25) &  0.43 (25) &     0.47 (200) &  0.33 (200) &    0.60 &     0.53 \\
\textbf{ACE} ($HDBSCAN$)   &  0.95 (10) &  0.87 (10) &  0.98 (50) &  0.94 (50) &  0.78 (45) &  0.67 (45) &       0.95 (10) &  0.82 (10) &    0.98 (70) &  0.91 (70) &  0.14 (45) &  0.11 (45) &    0.71 (25) &  0.43 (25) &     0.47 (200) &  0.33 (200) &    0.74 &     0.64 \\
\hline 
 \multicolumn{19}{c}{DEPICT: Calinski-Harabasz index} \\
 \hline 
Paired score     &  0.46 (5) &  0.6 (5) &  -0.99 (5) &  -0.96 (5) &  -0.85 (10) &  -0.72 (10) &        0.44 (5) &   0.56 (5) &   -0.92 (10) &  -0.82 (10) &         &       &           &       &             &       &   -0.37 &    -0.27 \\
\textbf{ACE} ($DBSCAN_{eps=0.1}$) &  0.46 (5) &  0.6 (5) &  0.88 (35) &  0.73 (35) &   0.97 (35) &   0.89 (35) &       0.95 (10) &  0.87 (10) &    0.95 (80) &   0.87 (80) &         &       &           &       &             &       &    0.84 &     0.79 \\
\textbf{ACE} ($DBSCAN_{eps=0.2}$) &  0.46 (5) &  0.6 (5) &  0.84 (40) &  0.69 (40) &   0.22 (20) &   0.11 (20) &       0.95 (10) &  0.87 (10) &    0.95 (80) &   0.87 (80) &         &       &           &       &             &       &    0.68 &     0.63 \\
\textbf{ACE} ($HDBSCAN$)   &  0.46 (5) &  0.6 (5) &  -0.66 (5) &  -0.51 (5) &   0.77 (30) &   0.61 (30) &        0.46 (5) &    0.6 (5) &    0.92 (80) &   0.82 (80) &         &       &           &       &             &       &    0.39 &     0.42 \\
\hline 
 \multicolumn{19}{c}{DEPICT: Davies-Bouldin index} \\
 \hline 
Paired score     &   0.46 (5) &   0.6 (5) &  -0.78 (5) &  -0.64 (5) &  -0.85 (10) &  -0.72 (10) &        0.44 (5) &   0.56 (5) &    -0.1 (10) &  0.02 (10) &         &       &           &       &             &       &   -0.17 &    -0.04 \\
\textbf{ACE} ($DBSCAN_{eps=0.1}$) &  0.62 (10) &  0.6 (10) &  0.99 (50) &  0.96 (50) &   0.68 (35) &   0.61 (35) &        0.9 (15) &  0.73 (15) &    0.87 (70) &  0.78 (70) &         &       &           &       &             &       &    0.81 &     0.74 \\
\textbf{ACE} ($DBSCAN_{eps=0.2}$) &  0.62 (10) &  0.6 (10) &  0.95 (50) &  0.87 (50) &   0.68 (35) &   0.61 (35) &       0.93 (10) &  0.82 (10) &    0.64 (50) &  0.47 (50) &         &       &           &       &             &       &    0.76 &     0.67 \\
\textbf{ACE} ($HDBSCAN$)   &  0.62 (10) &  0.6 (10) &  0.95 (50) &  0.87 (50) &   0.77 (35) &   0.67 (35) &       0.78 (10) &  0.69 (10) &    0.96 (70) &  0.91 (70) &         &       &           &       &             &       &    0.82 &     0.75 \\
\hline 
 \multicolumn{19}{c}{DEPICT: Silhouette score (Cosine distance)} \\
 \hline 
Paired score     &   0.44 (5) &   0.56 (5) &   -0.7 (5) &   -0.6 (5) &  -0.85 (10) &  -0.72 (10) &        0.44 (5) &   0.56 (5) &    0.07 (10) &  0.11 (10) &         &       &           &       &             &       &   -0.12 &    -0.02 \\
\textbf{ACE} ($DBSCAN_{eps=0.1}$) &   0.46 (5) &    0.6 (5) &  0.99 (50) &  0.96 (50) &   0.83 (35) &   0.72 (35) &       0.85 (10) &  0.78 (10) &     1.0 (80) &   1.0 (80) &         &       &           &       &             &       &    0.83 &     0.81 \\
\textbf{ACE} ($DBSCAN_{eps=0.2}$) &   0.46 (5) &    0.6 (5) &  0.87 (40) &  0.78 (40) &   0.93 (35) &   0.83 (35) &       0.73 (10) &  0.69 (10) &    0.72 (50) &  0.56 (50) &         &       &           &       &             &       &    0.74 &     0.69 \\
\textbf{ACE} ($HDBSCAN$)   &  0.65 (15) &  0.64 (15) &  0.87 (40) &  0.78 (40) &   0.93 (35) &   0.83 (35) &       0.85 (10) &  0.78 (10) &    0.99 (80) &  0.96 (80) &         &       &           &       &             &       &    0.86 &     0.80 \\
\hline 
 \multicolumn{19}{c}{DEPICT: Silhouette score (Euclidean distance)} \\
 \hline 
Paired score     &  0.44 (5) &  0.56 (5) &  -0.61 (5) &  -0.47 (5) &  -0.85 (10) &  -0.72 (10) &        0.44 (5) &   0.56 (5) &   -0.12 (10) &  -0.02 (10) &         &       &           &       &             &       &   -0.14 &    -0.02 \\
\textbf{ACE} ($DBSCAN_{eps=0.1}$) &  0.46 (5) &   0.6 (5) &  0.94 (40) &  0.87 (40) &   0.77 (35) &   0.67 (35) &       0.73 (10) &  0.69 (10) &    0.98 (80) &   0.91 (80) &         &       &           &       &             &       &    0.78 &     0.75 \\
\textbf{ACE} ($DBSCAN_{eps=0.2}$) &  0.46 (5) &   0.6 (5) &  0.94 (40) &  0.87 (40) &   0.45 (30) &   0.39 (30) &       0.73 (10) &  0.69 (10) &    0.98 (80) &   0.91 (80) &         &       &           &       &             &       &    0.71 &     0.69 \\
\textbf{ACE} ($HDBSCAN$)   &  0.46 (5) &   0.6 (5) &  0.94 (40) &  0.87 (40) &   0.02 (25) &   0.06 (25) &       0.85 (10) &  0.78 (10) &    0.98 (80) &   0.91 (80) &         &       &           &       &             &       &    0.65 &     0.64 \\
\bottomrule
\end{tabular}
}
 \label{tab:abs:nmi:db1}
\end{table*}
%%%%%%%%%%%%%%%%%%%%%%%%%%%%%%%%%%%%%%%%%%%%%%%%%%%%%%%%%%%%%%%%%%%%%%%%%%%%%%%%%%%%%%%%
\begin{table*}[htbp!]
\centering
\caption{Perturbation analyses comparing HDBSCAN and DBSCAN for Application 2. $r_s$ and $\tau_B$ between the generated scores and ACC scores are reported. Empty cells and dash marks ($-$) indicate cases where the result is either missing or impractical to obtain.}
\resizebox{\textwidth}{!}{
\begin{tabular}{lllllllllllllllllll}
\toprule
{} & \multicolumn{2}{l}{USPS (10)} & \multicolumn{2}{l}{YTF (41)} & \multicolumn{2}{l}{FRGC (20)} & \multicolumn{2}{l}{MNIST-test (10)} & \multicolumn{2}{l}{CMU-PIE (68)} & \multicolumn{2}{l}{UMist (20)} & \multicolumn{2}{l}{COIL-20 (20)} & \multicolumn{2}{l}{COIL-100 (100)} & \multicolumn{2}{l}{Average} \\
{} &     $r_s$ & $\tau_B$ &    $r_s$ & $\tau_B$ &     $r_s$ & $\tau_B$ &           $r_s$ & $\tau_B$ &        $r_s$ & $\tau_B$ &      $r_s$ & $\tau_B$ &        $r_s$ & $\tau_B$ &          $r_s$ & $\tau_B$ &   $r_s$ & $\tau_B$ \\
\midrule
\hline 
 \multicolumn{19}{c}{JULE: Calinski-Harabasz index} \\
 \hline 
Paired score     &      0.84 &     0.73 &     0.03 &    -0.06 &     -0.49 &    -0.31 &            0.61 &     0.56 &        -0.09 &    -0.07 &      -0.04 &     0.07 &         0.74 &     0.64 &           0.60 &     0.51 &    0.27 &     0.26 \\
\textbf{ACE} ($DBSCAN_{eps=0.1}$) &      0.84 &     0.73 &     0.92 &     0.83 &     -0.37 &    -0.20 &            0.61 &     0.56 &         0.83 &     0.69 &      -0.04 &     0.07 &         0.76 &     0.71 &           0.56 &     0.51 &    0.51 &     0.49 \\
\textbf{ACE} ($DBSCAN_{eps=0.2}$) &      0.84 &     0.73 &     0.17 &     0.06 &     -0.37 &    -0.20 &            0.61 &     0.56 &         0.83 &     0.69 &      -0.07 &     0.02 &         0.76 &     0.71 &           0.56 &     0.51 &    0.42 &     0.39 \\
\textbf{ACE} ($HDBSCAN$)   &      0.84 &     0.73 &     0.92 &     0.83 &     -0.11 &    -0.03 &            0.61 &     0.56 &         0.83 &     0.69 &      -0.07 &     0.02 &         0.76 &     0.71 &           0.65 &     0.56 &    0.55 &     0.51 \\
\hline 
 \multicolumn{19}{c}{JULE: Davies-Bouldin index} \\
 \hline 
Paired score     &      0.39 &     0.29 &     0.10 &     0.06 &      0.37 &     0.25 &            0.49 &     0.33 &         0.83 &     0.60 &      -0.28 &    -0.29 &        -0.29 &    -0.21 &          -0.87 &    -0.73 &    0.09 &     0.04 \\
\textbf{ACE} ($DBSCAN_{eps=0.1}$) &      0.90 &     0.78 &     0.90 &     0.78 &      0.60 &     0.42 &            0.67 &     0.56 &         0.88 &     0.73 &      -0.89 &    -0.78 &        -0.71 &    -0.57 &          -0.83 &    -0.73 &    0.19 &     0.15 \\
\textbf{ACE} ($DBSCAN_{eps=0.2}$) &      0.90 &     0.78 &     0.30 &     0.17 &      0.60 &     0.42 &            0.67 &     0.56 &         0.85 &     0.69 &      -0.83 &    -0.69 &        -0.71 &    -0.57 &          -0.82 &    -0.69 &    0.12 &     0.08 \\
\textbf{ACE} ($HDBSCAN$)   &      0.89 &     0.73 &     0.80 &     0.67 &      0.60 &     0.42 &            0.83 &     0.64 &         0.88 &     0.73 &      -0.42 &    -0.33 &        -0.71 &    -0.64 &          -0.82 &    -0.69 &    0.26 &     0.19 \\
\hline 
 \multicolumn{19}{c}{JULE: Silhouette score (Cosine distance)} \\
 \hline 
Paired score     &      0.89 &     0.78 &     0.27 &     0.22 &      0.21 &     0.09 &            0.81 &     0.64 &         0.99 &     0.96 &      -0.26 &    -0.24 &         0.55 &     0.43 &           0.52 &     0.33 &    0.50 &     0.40 \\
\textbf{ACE} ($DBSCAN_{eps=0.1}$) &      0.96 &     0.91 &     0.98 &     0.94 &      0.46 &     0.37 &            0.96 &     0.87 &         0.99 &     0.96 &      -0.76 &    -0.60 &         0.71 &     0.50 &           0.60 &     0.47 &    0.61 &     0.55 \\
\textbf{ACE} ($DBSCAN_{eps=0.2}$) &      0.96 &     0.91 &     0.70 &     0.56 &      0.64 &     0.54 &            0.94 &     0.82 &         0.99 &     0.96 &      -0.76 &    -0.60 &         0.71 &     0.50 &           0.60 &     0.47 &    0.60 &     0.52 \\
\textbf{ACE} ($HDBSCAN$)   &      0.95 &     0.87 &     0.98 &     0.94 &      0.64 &     0.54 &            0.94 &     0.82 &         0.99 &     0.96 &      -0.32 &    -0.24 &         0.76 &     0.57 &           0.60 &     0.47 &    0.69 &     0.61 \\
\hline 
 \multicolumn{19}{c}{JULE: Silhouette score (Euclidean distance)} \\
 \hline 
Paired score     &      0.93 &     0.82 &     0.30 &     0.28 &      0.21 &     0.09 &            0.82 &     0.64 &         0.98 &     0.91 &      -0.13 &    -0.16 &         0.52 &     0.36 &           0.55 &     0.42 &    0.52 &     0.42 \\
\textbf{ACE} ($DBSCAN_{eps=0.1}$) &      0.90 &     0.82 &     0.98 &     0.94 &      0.54 &     0.42 &            0.89 &     0.78 &         0.99 &     0.96 &      -0.89 &    -0.73 &         0.74 &     0.50 &           0.59 &     0.47 &    0.59 &     0.52 \\
\textbf{ACE} ($DBSCAN_{eps=0.2}$) &      0.90 &     0.82 &     0.98 &     0.94 &      0.60 &     0.48 &            0.89 &     0.78 &         0.99 &     0.96 &      -0.89 &    -0.73 &         0.74 &     0.50 &           0.59 &     0.47 &    0.60 &     0.53 \\
\textbf{ACE} ($HDBSCAN$)   &      0.95 &     0.87 &     0.98 &     0.94 &      0.57 &     0.48 &            0.92 &     0.78 &         0.99 &     0.96 &      -0.03 &    -0.11 &         0.74 &     0.50 &           0.59 &     0.47 &    0.71 &     0.61 \\
\hline 
 \multicolumn{19}{c}{DEPICT: Calinski-Harabasz index} \\
 \hline 
Paired score     &      0.88 &     0.82 &    -0.96 &    -0.91 &     -0.37 &    -0.22 &            0.79 &     0.73 &        -0.92 &    -0.82 &         &       &           &       &             &       &   -0.11 &    -0.08 \\
\textbf{ACE} ($DBSCAN_{eps=0.1}$) &      0.88 &     0.82 &     0.90 &     0.78 &      0.73 &     0.61 &            0.94 &     0.87 &         0.95 &     0.87 &         &       &           &       &             &       &    0.88 &     0.79 \\
\textbf{ACE} ($DBSCAN_{eps=0.2}$) &      0.88 &     0.82 &     0.88 &     0.73 &      0.70 &     0.61 &            0.94 &     0.87 &         0.95 &     0.87 &         &       &           &       &             &       &    0.87 &     0.78 \\
\textbf{ACE} ($HDBSCAN$)   &      0.88 &     0.82 &    -0.67 &    -0.56 &      0.92 &     0.78 &            0.82 &     0.78 &         0.92 &     0.82 &         &       &           &       &             &       &    0.57 &     0.53 \\
\hline 
 \multicolumn{19}{c}{DEPICT: Davies-Bouldin index} \\
 \hline 
Paired score     &      0.88 &     0.82 &    -0.77 &    -0.60 &     -0.37 &    -0.22 &            0.79 &     0.73 &        -0.10 &     0.02 &         &       &           &       &             &       &    0.09 &     0.15 \\
\textbf{ACE} ($DBSCAN_{eps=0.1}$) &      0.93 &     0.82 &     1.00 &     1.00 &      0.90 &     0.78 &            0.88 &     0.73 &         0.87 &     0.78 &         &       &           &       &             &       &    0.91 &     0.82 \\
\textbf{ACE} ($DBSCAN_{eps=0.2}$) &      0.93 &     0.82 &     0.96 &     0.91 &      0.90 &     0.78 &            0.90 &     0.82 &         0.64 &     0.47 &         &       &           &       &             &       &    0.87 &     0.76 \\
\textbf{ACE} ($HDBSCAN$)   &      0.93 &     0.82 &     0.96 &     0.91 &      0.92 &     0.83 &            0.93 &     0.87 &         0.96 &     0.91 &         &       &           &       &             &       &    0.94 &     0.87 \\
\hline 
 \multicolumn{19}{c}{DEPICT: Silhouette score (Cosine distance)} \\
 \hline 
Paired score     &      0.87 &     0.78 &    -0.69 &    -0.56 &     -0.37 &    -0.22 &            0.79 &     0.73 &         0.07 &     0.11 &         &       &           &       &             &       &    0.14 &     0.17 \\
\textbf{ACE} ($DBSCAN_{eps=0.1}$) &      0.88 &     0.82 &     1.00 &     1.00 &      0.88 &     0.78 &            0.95 &     0.87 &         1.00 &     1.00 &         &       &           &       &             &       &    0.94 &     0.89 \\
\textbf{ACE} ($DBSCAN_{eps=0.2}$) &      0.88 &     0.82 &     0.92 &     0.82 &      0.80 &     0.67 &            0.94 &     0.87 &         0.72 &     0.56 &         &       &           &       &             &       &    0.85 &     0.75 \\
\textbf{ACE} ($HDBSCAN$)   &      0.95 &     0.87 &     0.92 &     0.82 &      0.80 &     0.67 &            0.95 &     0.87 &         0.99 &     0.96 &         &       &           &       &             &       &    0.92 &     0.84 \\
\hline 
 \multicolumn{19}{c}{DEPICT: Silhouette score (Euclidean distance)} \\
 \hline 
Paired score     &      0.87 &     0.78 &    -0.64 &    -0.51 &     -0.37 &    -0.22 &            0.79 &     0.73 &        -0.12 &    -0.02 &         &       &           &       &             &       &    0.11 &     0.15 \\
\textbf{ACE} ($DBSCAN_{eps=0.1}$) &      0.88 &     0.82 &     0.98 &     0.91 &      0.92 &     0.83 &            0.94 &     0.87 &         0.98 &     0.91 &         &       &           &       &             &       &    0.94 &     0.87 \\
\textbf{ACE} ($DBSCAN_{eps=0.2}$) &      0.88 &     0.82 &     0.98 &     0.91 &      0.97 &     0.89 &            0.94 &     0.87 &         0.98 &     0.91 &         &       &           &       &             &       &    0.95 &     0.88 \\
\textbf{ACE} ($HDBSCAN$)   &      0.88 &     0.82 &     0.98 &     0.91 &      0.73 &     0.56 &            0.95 &     0.87 &         0.98 &     0.91 &         &       &           &       &             &       &    0.90 &     0.81 \\
\bottomrule
\end{tabular}
}
 \label{tab:abs:acc:db1}
\end{table*}
%%%%%%%%%%%%%%%%%%%%%%%%%%%%%%%%%%%%%%%%%%%%%%%%%%%%%%%%%%%%%%%%%%%%%%%%%%%%%%%%%%%%%%%%%
%%%%%%%%%%%%%%%%%%%%%%%%%%%%%%%%%%%%%%%%%%%%%%%%%%%%%%%%%%%%%%%%%%%%%%%%%%%%%%%%%%%%%%%%%
%%%%%%%%%%%%%%%%%%%%%%%%%%%%%%%%%%%%%%%%%%%%%%%%%%%%%%%%%%%%%%%%%%%%%%%%%%%%%%%%%%%%%%%%%
\clearpage
\newpage \subsection{PageRank vs. HITS} \label{app:exp:ab:link}
ACE incorporates link analysis to score and rank each space within the selected group of embedding spaces based on its connectivity within the group. Two widely used link analysis algorithms, HITS and PageRank, are introduced in Appendix \ref{app:link}. In our main text, we opted for PageRank, as it simultaneously considers both incoming and outgoing links. To compare their performance, we conducted experiments using both algorithms. For HITS, we report results based on the authority value. If any negative scores are generated by these algorithms, we assigned equal weights to all spaces in the ensemble analysis. Comparative performance results for Application 1 are presented in Tables \ref{tab:abs:nmi:link0} and \ref{tab:abs:acc:link0}, while those for Application 2 are shown in Tables \ref{tab:abs:nmi:link1} and \ref{tab:abs:acc:link1}. Across the experiments, both algorithms produced similar results, though PageRank sometimes yielded higher correlation, particularly in Application 2. Overall, their performance was comparable with PageRank performing only slightly better than HITS.\\

\begin{table*}[htbp!]
\centering
\caption{Perturbation analyses comparing PageRank and HITS for Application 1. $r_s$ and $\tau_B$ between the generated scores and NMI scores are reported. Empty cells and dash marks ($-$) indicate cases where the result is either missing or impractical to obtain.}
\resizebox{\textwidth}{!}{
\begin{tabular}{lllllllllllllllllll}
\toprule
 {} & \multicolumn{2}{l}{USPS} & \multicolumn{2}{l}{YTF} & \multicolumn{2}{l}{FRGC} & \multicolumn{2}{l}{MNIST-test} & \multicolumn{2}{l}{CMU-PIE} & \multicolumn{2}{l}{UMist} & \multicolumn{2}{l}{COIL-20} & \multicolumn{2}{l}{COIL-100} & \multicolumn{2}{l}{Average} \\
{} & $r_s$ & $\tau_B$ & $r_s$ & $\tau_B$ & $r_s$ & $\tau_B$ &      $r_s$ & $\tau_B$ &   $r_s$ & $\tau_B$ & $r_s$ & $\tau_B$ &   $r_s$ & $\tau_B$ &    $r_s$ & $\tau_B$ &   $r_s$ & $\tau_B$ \\
\midrule
\hline 
 \multicolumn{19}{c}{\emph{JULE}: Calinski-Harabasz index} \\
 \hline 
Paired score & 0.17 & 0.13 & 0.52 & 0.40 & -0.13 & -0.10 & 0.49 & 0.34 & -0.14 & -0.08 & 0.70 & 0.50 & 0.53 & 0.38 & 0.20 & 0.19 & 0.29 & 0.22 \\
\textbf{ACE} (HITS) & 0.80 & 0.63 & 0.90 & 0.73 & 0.39 & 0.26 & 0.87 & 0.71 & 0.98 & 0.91 & 0.82 & 0.62 & 0.61 & 0.47 & 0.95 & 0.82 & 0.79 & 0.64 \\
\textbf{ACE} (PR) & 0.80 & 0.63 & 0.90 & 0.73 & 0.39 & 0.26 & 0.87 & 0.71 & 0.98 & 0.91 & 0.81 & 0.61 & 0.61 & 0.47 & 0.95 & 0.82 & 0.79 & 0.64 \\
\hline 
 \multicolumn{19}{c}{\emph{JULE}: Davies-Bouldin index} \\
 \hline 
Paired score & -0.10 & -0.03 & -0.32 & -0.21 & -0.08 & -0.05 & -0.13 & -0.06 & 0.26 & 0.19 & 0.62 & 0.44 & 0.61 & 0.43 & 0.43 & 0.35 & 0.16 & 0.13 \\
\textbf{ACE} (HITS) & -0.08 & -0.02 & -0.30 & -0.21 & 0.21 & 0.15 & 0.73 & 0.55 & 0.10 & 0.06 & 0.46 & 0.34 & 0.23 & 0.22 & 0.61 & 0.44 & 0.25 & 0.19 \\
\textbf{ACE} (PR) & -0.08 & -0.02 & -0.30 & -0.21 & 0.22 & 0.16 & 0.73 & 0.55 & 0.10 & 0.06 & 0.38 & 0.27 & 0.24 & 0.22 & 0.48 & 0.33 & 0.22 & 0.17 \\
\hline 
 \multicolumn{19}{c}{\emph{JULE}: Silhouette score (cosine distance)} \\
 \hline 
Paired score & 0.28 & 0.22 & 0.73 & 0.56 & 0.09 & 0.06 & 0.63 & 0.47 & 0.50 & 0.36 & 0.71 & 0.50 & 0.68 & 0.51 & 0.74 & 0.54 & 0.54 & 0.40 \\
\textbf{ACE} (HITS) & 0.89 & 0.73 & 0.93 & 0.83 & 0.52 & 0.35 & 0.81 & 0.66 & 0.99 & 0.93 & 0.80 & 0.60 & 0.44 & 0.39 & 0.92 & 0.78 & 0.79 & 0.66 \\
\textbf{ACE} (PR) & 0.89 & 0.73 & 0.93 & 0.83 & 0.52 & 0.35 & 0.81 & 0.66 & 0.99 & 0.93 & 0.79 & 0.59 & 0.44 & 0.39 & 0.92 & 0.78 & 0.79 & 0.66 \\
\hline 
 \multicolumn{19}{c}{\emph{JULE}: Silhouette score (euclidean distance)} \\
 \hline 
Paired score & 0.27 & 0.20 & 0.72 & 0.55 & 0.04 & 0.03 & 0.56 & 0.41 & 0.41 & 0.30 & 0.70 & 0.50 & 0.64 & 0.47 & 0.55 & 0.41 & 0.49 & 0.36 \\
\textbf{ACE} (HITS) & 0.88 & 0.72 & 0.89 & 0.75 & 0.42 & 0.28 & 0.81 & 0.65 & 0.98 & 0.91 & 0.88 & 0.70 & 0.42 & 0.37 & 0.92 & 0.78 & 0.77 & 0.65 \\
\textbf{ACE} (PR) & 0.88 & 0.72 & 0.89 & 0.75 & 0.42 & 0.28 & 0.81 & 0.65 & 0.98 & 0.91 & 0.88 & 0.70 & 0.42 & 0.37 & 0.92 & 0.78 & 0.77 & 0.65 \\
\hline 
 \multicolumn{19}{c}{\emph{DEPICT}: Calinski-Harabasz index} \\
 \hline 
Paired score & 0.76 & 0.57 & 0.44 & 0.26 & 0.76 & 0.57 & 0.89 & 0.72 & 0.49 & 0.44 &  &  &  &  &  &  & 0.67 & 0.51 \\
\textbf{ACE} (HITS) & 0.91 & 0.77 & 0.56 & 0.44 & 0.94 & 0.82 & 0.96 & 0.87 & 0.96 & 0.87 &  &  &  &  &  &  & 0.87 & 0.75 \\
\textbf{ACE} (PR) & 0.91 & 0.77 & 0.56 & 0.44 & 0.94 & 0.82 & 0.96 & 0.87 & 0.96 & 0.87 &  &  &  &  &  &  & 0.87 & 0.75 \\
\hline 
 \multicolumn{19}{c}{\emph{DEPICT}: Davies-Bouldin index} \\
 \hline 
Paired score & 0.81 & 0.59 & 0.45 & 0.31 & 0.90 & 0.74 & 0.89 & 0.72 & 0.63 & 0.59 &  &  &  &  &  &  & 0.73 & 0.59 \\
\textbf{ACE} (HITS) & 0.91 & 0.82 & 0.75 & 0.58 & 0.92 & 0.80 & 0.96 & 0.87 & 0.98 & 0.92 &  &  &  &  &  &  & 0.90 & 0.80 \\
\textbf{ACE} (PR) & 0.91 & 0.82 & 0.76 & 0.58 & 0.91 & 0.79 & 0.96 & 0.87 & 0.98 & 0.92 &  &  &  &  &  &  & 0.90 & 0.80 \\
\hline 
 \multicolumn{19}{c}{\emph{DEPICT}: Silhouette score (cosine distance)} \\
 \hline 
Paired score & 0.81 & 0.62 & 0.45 & 0.33 & 0.90 & 0.75 & 0.89 & 0.72 & 0.77 & 0.58 &  &  &  &  &  &  & 0.76 & 0.60 \\
\textbf{ACE} (HITS) & 0.97 & 0.90 & 0.71 & 0.56 & 0.94 & 0.82 & 0.97 & 0.90 & 0.94 & 0.83 &  &  &  &  &  &  & 0.91 & 0.80 \\
\textbf{ACE} (PR) & 0.97 & 0.90 & 0.71 & 0.56 & 0.94 & 0.82 & 0.97 & 0.90 & 0.94 & 0.83 &  &  &  &  &  &  & 0.91 & 0.80 \\
\hline 
 \multicolumn{19}{c}{\emph{DEPICT}: Silhouette score (euclidean distance)} \\
 \hline 
Paired score & 0.73 & 0.50 & 0.47 & 0.36 & 0.79 & 0.65 & 0.86 & 0.69 & 0.59 & 0.52 &  &  &  &  &  &  & 0.69 & 0.54 \\
\textbf{ACE} (HITS) & 0.97 & 0.88 & 0.62 & 0.49 & 0.95 & 0.83 & 0.98 & 0.90 & 0.94 & 0.82 &  &  &  &  &  &  & 0.89 & 0.78 \\
\textbf{ACE} (PR) & 0.97 & 0.88 & 0.65 & 0.50 & 0.95 & 0.83 & 0.98 & 0.90 & 0.94 & 0.82 &  &  &  &  &  &  & 0.90 & 0.79 \\
\bottomrule
\end{tabular}
}
 \label{tab:abs:nmi:link0}
\end{table*}
%%%%%%%%%%%%%%%%%%%%%%%%%%%%%%%%%%%%%%%%%%%%%%%%%%%%%%%%%%%%%%%%%%%%%%%%%%%%%%%%%%%%%%%%%
\begin{table*}[htbp!]
\centering
\caption{Perturbation analyses comparing PageRank and HITS for Application 1. $r_s$ and $\tau_B$ between the generated scores and ACC scores are reported. Empty cells and dash marks ($-$) indicate cases where the result is either missing or impractical to obtain.}
\resizebox{\textwidth}{!}{
\begin{tabular}{lllllllllllllllllll}
\toprule
 {} & \multicolumn{2}{l}{USPS} & \multicolumn{2}{l}{YTF} & \multicolumn{2}{l}{FRGC} & \multicolumn{2}{l}{MNIST-test} & \multicolumn{2}{l}{CMU-PIE} & \multicolumn{2}{l}{UMist} & \multicolumn{2}{l}{COIL-20} & \multicolumn{2}{l}{COIL-100} & \multicolumn{2}{l}{Average} \\
{} & $r_s$ & $\tau_B$ & $r_s$ & $\tau_B$ & $r_s$ & $\tau_B$ &      $r_s$ & $\tau_B$ &   $r_s$ & $\tau_B$ & $r_s$ & $\tau_B$ &   $r_s$ & $\tau_B$ &    $r_s$ & $\tau_B$ &   $r_s$ & $\tau_B$ \\
\midrule
\hline 
 \multicolumn{19}{c}{\emph{JULE}: Calinski-Harabasz index} \\
 \hline 
Paired score & 0.04 & 0.05 & 0.39 & 0.27 & -0.26 & -0.18 & 0.31 & 0.21 & -0.20 & -0.12 & 0.64 & 0.45 & 0.57 & 0.40 & 0.09 & 0.08 & 0.20 & 0.14 \\
\textbf{ACE} (HITS) & 0.90 & 0.77 & 0.73 & 0.54 & 0.49 & 0.36 & 0.95 & 0.82 & 0.97 & 0.87 & 0.82 & 0.62 & 0.58 & 0.40 & 0.93 & 0.81 & 0.80 & 0.65 \\
\textbf{ACE} (PR) & 0.90 & 0.77 & 0.73 & 0.54 & 0.49 & 0.36 & 0.95 & 0.82 & 0.97 & 0.87 & 0.81 & 0.61 & 0.57 & 0.40 & 0.93 & 0.81 & 0.79 & 0.65 \\
\hline 
 \multicolumn{19}{c}{\emph{JULE}: Davies-Bouldin index} \\
 \hline 
Paired score & -0.27 & -0.15 & -0.14 & -0.09 & -0.23 & -0.14 & -0.35 & -0.19 & 0.20 & 0.16 & 0.53 & 0.36 & 0.63 & 0.44 & 0.33 & 0.26 & 0.09 & 0.08 \\
\textbf{ACE} (HITS) & -0.31 & -0.10 & -0.07 & -0.07 & 0.52 & 0.38 & 0.79 & 0.64 & 0.07 & 0.03 & 0.36 & 0.25 & 0.20 & 0.18 & 0.57 & 0.39 & 0.27 & 0.21 \\
\textbf{ACE} (PR) & -0.30 & -0.09 & -0.07 & -0.07 & 0.53 & 0.38 & 0.79 & 0.64 & 0.07 & 0.03 & 0.27 & 0.20 & 0.21 & 0.18 & 0.44 & 0.28 & 0.24 & 0.19 \\
\hline 
 \multicolumn{19}{c}{\emph{JULE}: Silhouette score (cosine distance)} \\
 \hline 
Paired score & 0.17 & 0.14 & 0.59 & 0.41 & 0.07 & 0.06 & 0.47 & 0.33 & 0.45 & 0.33 & 0.64 & 0.46 & 0.70 & 0.51 & 0.64 & 0.45 & 0.47 & 0.34 \\
\textbf{ACE} (HITS) & 0.96 & 0.85 & 0.74 & 0.55 & 0.82 & 0.65 & 0.92 & 0.78 & 0.98 & 0.92 & 0.79 & 0.58 & 0.41 & 0.32 & 0.84 & 0.68 & 0.81 & 0.67 \\
\textbf{ACE} (PR) & 0.96 & 0.85 & 0.74 & 0.55 & 0.82 & 0.65 & 0.92 & 0.78 & 0.98 & 0.92 & 0.78 & 0.58 & 0.41 & 0.32 & 0.84 & 0.68 & 0.81 & 0.67 \\
\hline 
 \multicolumn{19}{c}{\emph{JULE}: Silhouette score (euclidean distance)} \\
 \hline 
Paired score & 0.14 & 0.12 & 0.54 & 0.39 & -0.08 & -0.02 & 0.41 & 0.27 & 0.36 & 0.27 & 0.64 & 0.46 & 0.67 & 0.48 & 0.44 & 0.31 & 0.39 & 0.28 \\
\textbf{ACE} (HITS) & 0.93 & 0.78 & 0.63 & 0.48 & 0.71 & 0.53 & 0.92 & 0.78 & 0.98 & 0.91 & 0.86 & 0.68 & 0.39 & 0.30 & 0.84 & 0.68 & 0.78 & 0.64 \\
\textbf{ACE} (PR) & 0.93 & 0.78 & 0.63 & 0.48 & 0.71 & 0.53 & 0.92 & 0.78 & 0.98 & 0.91 & 0.86 & 0.68 & 0.39 & 0.30 & 0.84 & 0.68 & 0.78 & 0.64 \\
\hline 
 \multicolumn{19}{c}{\emph{DEPICT}: Calinski-Harabasz index} \\
 \hline 
Paired score & 0.56 & 0.40 & 0.54 & 0.35 & 0.76 & 0.57 & 0.88 & 0.69 & 0.48 & 0.43 &  &  &  &  &  &  & 0.64 & 0.49 \\
\textbf{ACE} (HITS) & 0.82 & 0.72 & 0.61 & 0.45 & 0.91 & 0.82 & 0.97 & 0.91 & 0.96 & 0.87 &  &  &  &  &  &  & 0.86 & 0.75 \\
\textbf{ACE} (PR) & 0.82 & 0.72 & 0.61 & 0.45 & 0.91 & 0.82 & 0.97 & 0.91 & 0.96 & 0.87 &  &  &  &  &  &  & 0.86 & 0.75 \\
\hline 
 \multicolumn{19}{c}{\emph{DEPICT}: Davies-Bouldin index} \\
 \hline 
Paired score & 0.61 & 0.42 & 0.48 & 0.32 & 0.92 & 0.74 & 0.88 & 0.69 & 0.62 & 0.56 &  &  &  &  &  &  & 0.70 & 0.55 \\
\textbf{ACE} (HITS) & 0.99 & 0.96 & 0.65 & 0.46 & 0.90 & 0.75 & 0.99 & 0.96 & 0.96 & 0.87 &  &  &  &  &  &  & 0.90 & 0.80 \\
\textbf{ACE} (PR) & 0.99 & 0.96 & 0.65 & 0.46 & 0.90 & 0.74 & 0.99 & 0.96 & 0.96 & 0.87 &  &  &  &  &  &  & 0.90 & 0.80 \\
\hline 
 \multicolumn{19}{c}{\emph{DEPICT}: Silhouette score (cosine distance)} \\
 \hline 
Paired score & 0.62 & 0.45 & 0.53 & 0.42 & 0.91 & 0.75 & 0.88 & 0.69 & 0.77 & 0.58 &  &  &  &  &  &  & 0.74 & 0.58 \\
\textbf{ACE} (HITS) & 0.95 & 0.88 & 0.70 & 0.54 & 0.91 & 0.77 & 0.96 & 0.88 & 0.94 & 0.83 &  &  &  &  &  &  & 0.89 & 0.78 \\
\textbf{ACE} (PR) & 0.95 & 0.88 & 0.70 & 0.54 & 0.91 & 0.77 & 0.96 & 0.88 & 0.94 & 0.83 &  &  &  &  &  &  & 0.89 & 0.78 \\
\hline 
 \multicolumn{19}{c}{\emph{DEPICT}: Silhouette score (euclidean distance)} \\
 \hline 
Paired score & 0.52 & 0.33 & 0.57 & 0.45 & 0.80 & 0.62 & 0.85 & 0.65 & 0.59 & 0.48 &  &  &  &  &  &  & 0.67 & 0.51 \\
\textbf{ACE} (HITS) & 0.95 & 0.87 & 0.61 & 0.48 & 0.91 & 0.78 & 0.97 & 0.91 & 0.95 & 0.84 &  &  &  &  &  &  & 0.88 & 0.77 \\
\textbf{ACE} (PR) & 0.95 & 0.87 & 0.63 & 0.49 & 0.91 & 0.78 & 0.97 & 0.91 & 0.95 & 0.84 &  &  &  &  &  &  & 0.88 & 0.78 \\
\bottomrule
\end{tabular}
}
 \label{tab:abs:acc:link0}
\end{table*}

%%%%%%%%%%%%%%%%%%%%%%%%%%%%%%%%%%%%%%%%%%%%%%%%%%%%%%%%%%%%%%%%%%%%%%%%%%%%%%%%%%%%%%%%%
\begin{table*}[htbp!]
\centering
\caption{Perturbation analyses comparing PageRank and HITS for Application 2. $r_s$ and $\tau_B$ between the generated scores and NMI scores are reported. Empty cells and dash marks ($-$) indicate cases where the result is either missing or impractical to obtain.}
\resizebox{\textwidth}{!}{
\begin{tabular}{lllllllllllllllllll}
\toprule
{} & \multicolumn{2}{l}{USPS (10)} & \multicolumn{2}{l}{YTF (41)} & \multicolumn{2}{l}{FRGC (20)} & \multicolumn{2}{l}{MNIST-test (10)} & \multicolumn{2}{l}{CMU-PIE (68)} & \multicolumn{2}{l}{UMist (20)} & \multicolumn{2}{l}{COIL-20 (20)} & \multicolumn{2}{l}{COIL-100 (100)} & \multicolumn{2}{l}{Average} \\
{} &      $r_s$ &   $\tau_B$ &      $r_s$ &   $\tau_B$ &       $r_s$ &    $\tau_B$ &           $r_s$ &  $\tau_B$ &        $r_s$ &    $\tau_B$ &      $r_s$ &   $\tau_B$ &        $r_s$ &   $\tau_B$ &          $r_s$ &   $\tau_B$ &   $r_s$ & $\tau_B$ \\
\midrule
\hline 
 \multicolumn{19}{c}{\emph{JULE}: Calinski-Harabasz index} \\
 \hline 
Paired score & 0.65 (10) & 0.64 (10) & 0.1 (50) & 0.06 (50) & -0.93 (15) & -0.83 (15) & 0.64 (10) & 0.6 (10) & -0.03 (20) & -0.02 (20) & -0.13 (5) & -0.07 (5) & 0.76 (15) & 0.71 (15) & 0.74 (80) & 0.56 (80) & 0.22 & 0.21 \\
\textbf{ACE} (HITS) & 0.65 (10) & 0.64 (10) & 0.93 (50) & 0.83 (50) & -0.7 (15) & -0.61 (15) & 0.64 (10) & 0.6 (10) & 0.88 (70) & 0.73 (70) & -0.14 (5) & -0.11 (5) & 0.74 (15) & 0.64 (15) & 0.79 (80) & 0.69 (80) & 0.47 & 0.43 \\
\textbf{ACE} (PR) & 0.65 (10) & 0.64 (10) & 0.93 (50) & 0.83 (50) & -0.72 (15) & -0.67 (15) & 0.64 (10) & 0.6 (10) & 0.88 (70) & 0.73 (70) & -0.14 (5) & -0.11 (5) & 0.74 (15) & 0.64 (15) & 0.79 (80) & 0.69 (80) & 0.47 & 0.42 \\
\hline 
 \multicolumn{19}{c}{\emph{JULE}: Davies-Bouldin index} \\
 \hline 
Paired score & 0.54 (15) & 0.38 (15) & 0.15 (50) & 0.17 (50) & 0.85 (45) & 0.67 (45) & 0.43 (10) & 0.29 (10) & 0.78 (100) & 0.56 (100) & -0.08 (45) & 0.02 (45) & -0.26 (40) & -0.14 (40) & -0.9 (20) & -0.78 (20) & 0.19 & 0.15 \\
\textbf{ACE} (HITS) & 0.73 (10) & 0.69 (10) & 0.83 (50) & 0.67 (50) & 0.82 (40) & 0.61 (40) & 0.79 (10) & 0.6 (10) & 0.85 (90) & 0.69 (90) & -0.36 (45) & -0.16 (45) & -0.69 (50) & -0.57 (50) & -0.94 (20) & -0.82 (20) & 0.25 & 0.21 \\
\textbf{ACE} (PR) & 0.98 (15) & 0.91 (15) & 0.83 (50) & 0.67 (50) & 0.87 (40) & 0.72 (40) & 0.79 (10) & 0.6 (10) & 0.85 (90) & 0.69 (90) & -0.21 (45) & -0.02 (45) & -0.69 (50) & -0.57 (50) & -0.94 (20) & -0.82 (20) & 0.31 & 0.27 \\
\hline 
 \multicolumn{19}{c}{\emph{JULE}: Silhouette score (cosine distance)} \\
 \hline 
Paired score & 0.99 (10) & 0.96 (10) & 0.3 (50) & 0.22 (50) & 0.72 (25) & 0.61 (25) & 0.87 (10) & 0.69 (10) & 0.98 (70) & 0.91 (70) & -0.07 (45) & 0.07 (45) & 0.52 (25) & 0.36 (25) & 0.39 (200) & 0.2 (200) & 0.59 & 0.50 \\
\textbf{ACE} (HITS) & 0.95 (10) & 0.87 (10) & 0.98 (50) & 0.94 (50) & 0.68 (45) & 0.56 (45) & 0.96 (10) & 0.87 (10) & 0.98 (70) & 0.91 (70) & -0.07 (45) & -0.02 (45) & 0.79 (20) & 0.64 (20) & 0.55 (180) & 0.47 (180) & 0.73 & 0.66 \\
\textbf{ACE} (PR) & 0.95 (10) & 0.87 (10) & 0.98 (50) & 0.94 (50) & 0.7 (45) & 0.61 (45) & 0.96 (10) & 0.87 (10) & 0.98 (70) & 0.91 (70) & -0.07 (45) & -0.02 (45) & 0.74 (20) & 0.5 (20) & 0.46 (180) & 0.33 (180) & 0.71 & 0.63 \\
\hline 
 \multicolumn{19}{c}{\emph{JULE}: Silhouette score (euclidean distance)} \\
 \hline 
Paired score & 0.85 (10) & 0.73 (10) & 0.33 (50) & 0.28 (50) & 0.72 (25) & 0.61 (25) & 0.88 (10) & 0.69 (10) & 0.96 (80) & 0.87 (80) & 0.07 (45) & 0.16 (45) & 0.55 (25) & 0.43 (25) & 0.44 (200) & 0.29 (200) & 0.60 & 0.51 \\
\textbf{ACE} (HITS) & 0.95 (10) & 0.87 (10) & 0.97 (50) & 0.89 (50) & 0.78 (45) & 0.67 (45) & 0.95 (10) & 0.82 (10) & 0.98 (70) & 0.91 (70) & 0.14 (45) & 0.11 (45) & 0.76 (25) & 0.57 (25) & 0.47 (200) & 0.33 (200) & 0.75 & 0.65 \\
\textbf{ACE} (PR) & 0.95 (10) & 0.87 (10) & 0.98 (50) & 0.94 (50) & 0.78 (45) & 0.67 (45) & 0.95 (10) & 0.82 (10) & 0.98 (70) & 0.91 (70) & 0.14 (45) & 0.11 (45) & 0.71 (25) & 0.43 (25) & 0.47 (200) & 0.33 (200) & 0.74 & 0.64 \\
\hline 
 \multicolumn{19}{c}{\emph{DEPICT}: Calinski-Harabasz index} \\
 \hline 
Paired score & 0.46 (5) & 0.6 (5) & -0.99 (5) & -0.96 (5) & -0.85 (10) & -0.72 (10) & 0.44 (5) & 0.56 (5) & -0.92 (10) & -0.82 (10) &  &  &  &  &  &  & -0.37 & -0.27 \\
\textbf{ACE} (HITS) & 0.46 (5) & 0.6 (5) & -0.66 (5) & -0.51 (5) & -0.85 (10) & -0.72 (10) & 0.95 (10) & 0.87 (10) & 0.44 (10) & 0.56 (10) &  &  &  &  &  &  & 0.07 & 0.16 \\
\textbf{ACE} (PR) & 0.46 (5) & 0.6 (5) & -0.66 (5) & -0.51 (5) & 0.77 (30) & 0.61 (30) & 0.46 (5) & 0.6 (5) & 0.92 (80) & 0.82 (80) &  &  &  &  &  &  & 0.39 & 0.42 \\
\hline 
 \multicolumn{19}{c}{\emph{DEPICT}: Davies-Bouldin index} \\
 \hline 
Paired score & 0.46 (5) & 0.6 (5) & -0.78 (5) & -0.64 (5) & -0.85 (10) & -0.72 (10) & 0.44 (5) & 0.56 (5) & -0.1 (10) & 0.02 (10) &  &  &  &  &  &  & -0.17 & -0.04 \\
\textbf{ACE} (HITS) & 0.6 (15) & 0.51 (15) & 0.95 (50) & 0.87 (50) & 0.65 (35) & 0.56 (35) & 0.78 (10) & 0.69 (10) & 0.92 (100) & 0.82 (100) &  &  &  &  &  &  & 0.78 & 0.69 \\
\textbf{ACE} (PR) & 0.62 (10) & 0.6 (10) & 0.95 (50) & 0.87 (50) & 0.77 (35) & 0.67 (35) & 0.78 (10) & 0.69 (10) & 0.96 (70) & 0.91 (70) &  &  &  &  &  &  & 0.82 & 0.75 \\
\hline 
 \multicolumn{19}{c}{\emph{DEPICT}: Silhouette score (cosine distance)} \\
 \hline 
Paired score & 0.44 (5) & 0.56 (5) & -0.7 (5) & -0.6 (5) & -0.85 (10) & -0.72 (10) & 0.44 (5) & 0.56 (5) & 0.07 (10) & 0.11 (10) &  &  &  &  &  &  & -0.12 & -0.02 \\
\textbf{ACE} (HITS) & 0.6 (15) & 0.51 (15) & 0.82 (40) & 0.73 (40) & 0.93 (35) & 0.83 (35) & 0.85 (10) & 0.78 (10) & 0.98 (80) & 0.91 (80) &  &  &  &  &  &  & 0.84 & 0.75 \\
\textbf{ACE} (PR) & 0.65 (15) & 0.64 (15) & 0.87 (40) & 0.78 (40) & 0.93 (35) & 0.83 (35) & 0.85 (10) & 0.78 (10) & 0.99 (80) & 0.96 (80) &  &  &  &  &  &  & 0.86 & 0.80 \\
\hline 
 \multicolumn{19}{c}{\emph{DEPICT}: Silhouette score (euclidean distance)} \\
 \hline 
Paired score & 0.44 (5) & 0.56 (5) & -0.61 (5) & -0.47 (5) & -0.85 (10) & -0.72 (10) & 0.44 (5) & 0.56 (5) & -0.12 (10) & -0.02 (10) &  &  &  &  &  &  & -0.14 & -0.02 \\
\textbf{ACE} (HITS) & 0.6 (15) & 0.51 (15) & 0.94 (40) & 0.87 (40) & 0.45 (30) & 0.39 (30) & 0.85 (10) & 0.78 (10) & 0.98 (80) & 0.91 (80) &  &  &  &  &  &  & 0.76 & 0.69 \\
\textbf{ACE} (PR) & 0.46 (5) & 0.6 (5) & 0.94 (40) & 0.87 (40) & 0.02 (25) & 0.06 (25) & 0.85 (10) & 0.78 (10) & 0.98 (80) & 0.91 (80) &  &  &  &  &  &  & 0.65 & 0.64 \\
\bottomrule
\end{tabular}
}
 \label{tab:abs:nmi:link1}
\end{table*}
%%%%%%%%%%%%%%%%%%%%%%%%%%%%%%%%%%%%%%%%%%%%%%%%%%%%%%%%%%%%%%%%%%%%%%%%%%%%%%%%%%%%%%%%%
\begin{table*}[htbp!]
\centering
\caption{Perturbation analyses comparing PageRank and HITS for Application 2. $r_s$ and $\tau_B$ between the generated scores and ACC scores are reported. Empty cells and dash marks ($-$) indicate cases where the result is either missing or impractical to obtain.}
\resizebox{\textwidth}{!}{
\begin{tabular}{lllllllllllllllllll}
\toprule
 {} & \multicolumn{2}{l}{USPS (10)} & \multicolumn{2}{l}{YTF (41)} & \multicolumn{2}{l}{FRGC (20)} & \multicolumn{2}{l}{MNIST-test (10)} & \multicolumn{2}{l}{CMU-PIE (68)} & \multicolumn{2}{l}{UMist (20)} & \multicolumn{2}{l}{COIL-20 (20)} & \multicolumn{2}{l}{COIL-100 (100)} & \multicolumn{2}{l}{Average} \\
{} &     $r_s$ & $\tau_B$ &    $r_s$ & $\tau_B$ &     $r_s$ & $\tau_B$ &           $r_s$ & $\tau_B$ &        $r_s$ & $\tau_B$ &      $r_s$ & $\tau_B$ &        $r_s$ & $\tau_B$ &          $r_s$ & $\tau_B$ &   $r_s$ & $\tau_B$ \\
\midrule
\hline 
 \multicolumn{19}{c}{\emph{JULE}: Calinski-Harabasz index} \\
 \hline 
Paired score & 0.84 & 0.73 & 0.03 & -0.06 & -0.49 & -0.31 & 0.61 & 0.56 & -0.09 & -0.07 & -0.04 & 0.07 & 0.74 & 0.64 & 0.60 & 0.51 & 0.27 & 0.26 \\
\textbf{ACE} (HITS) & 0.84 & 0.73 & 0.92 & 0.83 & -0.19 & -0.09 & 0.61 & 0.56 & 0.83 & 0.69 & -0.07 & 0.02 & 0.76 & 0.71 & 0.65 & 0.56 & 0.54 & 0.50 \\
\textbf{ACE} (PR) & 0.84 & 0.73 & 0.92 & 0.83 & -0.11 & -0.03 & 0.61 & 0.56 & 0.83 & 0.69 & -0.07 & 0.02 & 0.76 & 0.71 & 0.65 & 0.56 & 0.55 & 0.51 \\
\hline 
 \multicolumn{19}{c}{\emph{JULE}: Davies-Bouldin index} \\
 \hline 
Paired score & 0.39 & 0.29 & 0.10 & 0.06 & 0.37 & 0.25 & 0.49 & 0.33 & 0.83 & 0.60 & -0.28 & -0.29 & -0.29 & -0.21 & -0.87 & -0.73 & 0.09 & 0.04 \\
\textbf{ACE} (HITS) & 0.90 & 0.78 & 0.80 & 0.67 & 0.71 & 0.54 & 0.83 & 0.64 & 0.88 & 0.73 & -0.65 & -0.47 & -0.71 & -0.64 & -0.82 & -0.69 & 0.24 & 0.19 \\
\textbf{ACE} (PR) & 0.89 & 0.73 & 0.80 & 0.67 & 0.60 & 0.42 & 0.83 & 0.64 & 0.88 & 0.73 & -0.42 & -0.33 & -0.71 & -0.64 & -0.82 & -0.69 & 0.26 & 0.19 \\
\hline 
 \multicolumn{19}{c}{\emph{JULE}: Silhouette score (cosine distance)} \\
 \hline 
Paired score & 0.89 & 0.78 & 0.27 & 0.22 & 0.21 & 0.09 & 0.81 & 0.64 & 0.99 & 0.96 & -0.26 & -0.24 & 0.55 & 0.43 & 0.52 & 0.33 & 0.50 & 0.40 \\
\textbf{ACE} (HITS) & 0.95 & 0.87 & 0.98 & 0.94 & 0.61 & 0.48 & 0.94 & 0.82 & 0.99 & 0.96 & -0.32 & -0.24 & 0.76 & 0.57 & 0.65 & 0.60 & 0.70 & 0.62 \\
\textbf{ACE} (PR) & 0.95 & 0.87 & 0.98 & 0.94 & 0.64 & 0.54 & 0.94 & 0.82 & 0.99 & 0.96 & -0.32 & -0.24 & 0.76 & 0.57 & 0.60 & 0.47 & 0.69 & 0.61 \\
\hline 
 \multicolumn{19}{c}{\emph{JULE}: Silhouette score (euclidean distance)} \\
 \hline 
Paired score & 0.93 & 0.82 & 0.30 & 0.28 & 0.21 & 0.09 & 0.82 & 0.64 & 0.98 & 0.91 & -0.13 & -0.16 & 0.52 & 0.36 & 0.55 & 0.42 & 0.52 & 0.42 \\
\textbf{ACE} (HITS) & 0.95 & 0.87 & 0.97 & 0.89 & 0.57 & 0.48 & 0.92 & 0.78 & 0.99 & 0.96 & -0.03 & -0.11 & 0.74 & 0.50 & 0.59 & 0.47 & 0.71 & 0.60 \\
\textbf{ACE} (PR) & 0.95 & 0.87 & 0.98 & 0.94 & 0.57 & 0.48 & 0.92 & 0.78 & 0.99 & 0.96 & -0.03 & -0.11 & 0.74 & 0.50 & 0.59 & 0.47 & 0.71 & 0.61 \\
\hline 
 \multicolumn{19}{c}{\emph{DEPICT}: Calinski-Harabasz index} \\
 \hline 
Paired score & 0.88 & 0.82 & -0.96 & -0.91 & -0.37 & -0.22 & 0.79 & 0.73 & -0.92 & -0.82 &  &  &  &  &  &  & -0.11 & -0.08 \\
\textbf{ACE} (HITS) & 0.88 & 0.82 & -0.67 & -0.56 & -0.37 & -0.22 & 0.94 & 0.87 & 0.44 & 0.56 &  &  &  &  &  &  & 0.24 & 0.29 \\
\textbf{ACE} (PR) & 0.88 & 0.82 & -0.67 & -0.56 & 0.92 & 0.78 & 0.82 & 0.78 & 0.92 & 0.82 &  &  &  &  &  &  & 0.57 & 0.53 \\
\hline 
 \multicolumn{19}{c}{\emph{DEPICT}: Davies-Bouldin index} \\
 \hline 
Paired score & 0.88 & 0.82 & -0.77 & -0.60 & -0.37 & -0.22 & 0.79 & 0.73 & -0.10 & 0.02 &  &  &  &  &  &  & 0.09 & 0.15 \\
\textbf{ACE} (HITS) & 0.90 & 0.73 & 0.98 & 0.91 & 0.95 & 0.83 & 0.93 & 0.87 & 0.92 & 0.82 &  &  &  &  &  &  & 0.93 & 0.83 \\
\textbf{ACE} (PR) & 0.93 & 0.82 & 0.96 & 0.91 & 0.92 & 0.83 & 0.93 & 0.87 & 0.96 & 0.91 &  &  &  &  &  &  & 0.94 & 0.87 \\
\hline 
 \multicolumn{19}{c}{\emph{DEPICT}: Silhouette score (cosine distance)} \\
 \hline 
Paired score & 0.87 & 0.78 & -0.69 & -0.56 & -0.37 & -0.22 & 0.79 & 0.73 & 0.07 & 0.11 &  &  &  &  &  &  & 0.14 & 0.17 \\
\textbf{ACE} (HITS) & 0.90 & 0.73 & 0.87 & 0.78 & 0.80 & 0.67 & 0.95 & 0.87 & 0.98 & 0.91 &  &  &  &  &  &  & 0.90 & 0.79 \\
\textbf{ACE} (PR) & 0.95 & 0.87 & 0.92 & 0.82 & 0.80 & 0.67 & 0.95 & 0.87 & 0.99 & 0.96 &  &  &  &  &  &  & 0.92 & 0.84 \\
\hline 
 \multicolumn{19}{c}{\emph{DEPICT}: Silhouette score (euclidean distance)} \\
 \hline 
Paired score & 0.87 & 0.78 & -0.64 & -0.51 & -0.37 & -0.22 & 0.79 & 0.73 & -0.12 & -0.02 &  &  &  &  &  &  & 0.11 & 0.15 \\
\textbf{ACE} (HITS) & 0.90 & 0.73 & 0.98 & 0.91 & 0.97 & 0.89 & 0.95 & 0.87 & 0.98 & 0.91 &  &  &  &  &  &  & 0.95 & 0.86 \\
\textbf{ACE} (PR) & 0.88 & 0.82 & 0.98 & 0.91 & 0.73 & 0.56 & 0.95 & 0.87 & 0.98 & 0.91 &  &  &  &  &  &  & 0.90 & 0.81 \\
\bottomrule
\end{tabular}
}
 \label{tab:abs:acc:link1}
\end{table*}

%%%%%%%%%%%%%%%%%%%%%%%%%%%%%%%%%%%%%%%%%%%%%%%%%%%%%%%%%%%%%%%%%%%%%%%%%%%%%%%%%%%%%%%%%
%%%%%%%%%%%%%%%%%%%%%%%%%%%%%%%%%%%%%%%%%%%%%%%%%%%%%%%%%%%%%%%%%%%%%%%%%%%%%%%%%%%%%%%%%
%%%%%%%%%%%%%%%%%%%%%%%%%%%%%%%%%%%%%%%%%%%%%%%%%%%%%%%%%%%%%%%%%%%%%%%%%%%%%%%%%%%%%%%%% 
\clearpage
\newpage \subsection{Outlier space (approximately rank uncorrelated space)} \label{app:outlier}

In Algorithms \ref{Algorithm1} and \ref{Algorithm2}, outlier spaces are excluded after Phase 1 of the stage-wise grouping algorithm, and treated as rank uncorrelated spaces for ensemble analysis. However, a challenge arises when the number of admissible spaces among the candidate embedding spaces is small. In such cases, the few admissible spaces that do exist may be mistakenly classified as outliers during Phase 1 of our stage-wise grouping strategy. As a result, the current version of ACE risks excluding the only admissible spaces, limiting its effectiveness in such scenarios.

To address this issue, instead of discarding the outlier space after Phase 1, we identify the outlier space \(\mathcal{Z}^{outlier^*}\) with the highest average score and compare its scores \(\{\pi(\rho_{m} | G_{outlier^*})\}_{m=1}^M\) with those of \(\{\pi(\rho_{m} | G_{s^*})\}_{m=1}^M\), obtained at the end of Algorithm \ref{Algorithm1}. To enforce a stringent selection criterion, we first verify whether \(G_{outlier^*}\) exceeds \(G_{s^*}\) in terms of the average score. Additionally, we perform a paired t-test to determine if the scores generated from \(G_{outlier^*}\) are significantly greater than those from \(G_{s^*}\) at a significance level of 0.05. If both conditions hold, we select the scores from this ``best'' outlier space as the final output. While this approach helps mitigate the issue described above, it also increases variance by raising the likelihood of relying on a single space instead of a group of spaces for score calculation, potentially introducing fluctuations and reducing performance in other cases. In other words, finding a universal solution remains challenging. To evaluate the impact of this modification, we implement an alternative version of ACE (with $\mathcal{Z}_{outlier}$) and compare its performance against the original ACE framework presented in the main text. The comparative results for Application 1 are reported in Tables \ref{tab:abs:nmi:ot0} and \ref{tab:abs:acc:ot0}, while those for Application 2 are presented in Tables \ref{tab:abs:nmi:ot1} and \ref{tab:abs:acc:ot1}.

From the comparison, we observe that in most cases, ACE and ACE with \(\mathcal{Z}_{outlier}\) exhibit similar performance, suggesting that the edge case we discussed may not occur frequently in our experiments and that the modified version of ACE does not noticeably degrade performance when the edge case does not occur. Both strategies outperform the use of paired scores, indicating that our proposed approaches are more effective than directly computing internal measure values from paired embedding spaces. In specific cases, such as COIL-20 for Application 1 with JULE (with respect to Davies-Bouldin index), where ACE underperforms compared to paired scores, incorporating the outlier space in ACE leads to a notable performance improvement. Further investigation reveals that, in this case, only a few spaces are included in the selected Phase 1 group. This observation suggests that the alternative strategy can serve as a remedy in certain edge cases where the standard approach struggles due to a possibly limited number of admissible spaces.

\begin{table*}[htbp!]
\centering
\caption{Perturbation analyses comparing inclusion and exclusion of outlier spaces for Application 1. $r_s$ and $\tau_B$ between the generated scores and NMI scores are reported. Empty cells and dash marks ($-$) indicate cases where the result is either missing or impractical to obtain.}
 \label{tab:abs:nmi:ot0}
\resizebox{\textwidth}{!}{
\begin{tabular}{lllllllllllllllllll}
\toprule
{} & \multicolumn{2}{l}{USPS} & \multicolumn{2}{l}{YTF} & \multicolumn{2}{l}{FRGC} & \multicolumn{2}{l}{MNIST-test} & \multicolumn{2}{l}{CMU-PIE} & \multicolumn{2}{l}{UMist} & \multicolumn{2}{l}{COIL-20} & \multicolumn{2}{l}{COIL-100} & \multicolumn{2}{l}{Average} \\
{} & $r_s$ & $\tau_B$ & $r_s$ & $\tau_B$ & $r_s$ & $\tau_B$ &      $r_s$ & $\tau_B$ &   $r_s$ & $\tau_B$ & $r_s$ & $\tau_B$ &   $r_s$ & $\tau_B$ &    $r_s$ & $\tau_B$ &   $r_s$ & $\tau_B$ \\
\midrule
\hline 
 \multicolumn{19}{c}{JULE: Calinski-Harabasz index} \\
 \hline 
Paired score     &  0.17 &     0.13 &  0.52 &     0.40 & -0.13 &    -0.10 &       0.49 &     0.34 &   -0.13 &    -0.08 &  0.70 &     0.50 &    0.53 &     0.38 &     0.20 &     0.19 &    0.29 &     0.22 \\
\textbf{ACE} (with $\mathcal{Z}_{outlier}$) &  0.81 &     0.64 &  0.71 &     0.54 &  0.08 &     0.04 &       0.87 &     0.71 &    0.98 &     0.90 &  0.81 &     0.61 &    0.71 &     0.54 &     0.61 &     0.47 &    0.70 &     0.56 \\
\textbf{ACE}   &  0.80 &     0.63 &  0.90 &     0.73 &  0.39 &     0.26 &       0.87 &     0.71 &    0.98 &     0.90 &  0.81 &     0.61 &    0.60 &     0.45 &     0.95 &     0.82 &    0.79 &     0.64 \\
\hline 
 \multicolumn{19}{c}{JULE: Davies-Bouldin index} \\
 \hline 
Paired score     & -0.10 &    -0.03 & -0.32 &    -0.21 & -0.08 &    -0.05 &      -0.13 &    -0.06 &    0.26 &     0.20 &  0.62 &     0.44 &    0.61 &     0.42 &     0.43 &     0.35 &    0.16 &     0.13 \\
\textbf{ACE} (with $\mathcal{Z}_{outlier}$) &  0.01 &     0.05 & -0.30 &    -0.21 &  0.22 &     0.16 &       0.73 &     0.55 &    0.83 &     0.67 &  0.38 &     0.27 &    0.86 &     0.66 &     0.48 &     0.33 &    0.40 &     0.31 \\
\textbf{ACE}   & -0.08 &    -0.02 & -0.30 &    -0.21 &  0.22 &     0.16 &       0.73 &     0.55 &    0.10 &     0.06 &  0.38 &     0.27 &    0.23 &     0.22 &     0.48 &     0.33 &    0.22 &     0.17 \\
\hline 
 \multicolumn{19}{c}{JULE: Silhouette score (Cosine distance)} \\
 \hline 
Paired score     &  0.28 &     0.22 &  0.73 &     0.56 &  0.09 &     0.06 &       0.63 &     0.47 &    0.50 &     0.36 &  0.71 &     0.50 &    0.68 &     0.50 &     0.74 &     0.54 &    0.54 &     0.40 \\
\textbf{ACE} (with $\mathcal{Z}_{outlier}$) &  0.89 &     0.73 &  0.93 &     0.83 &  0.52 &     0.35 &       0.81 &     0.66 &    0.99 &     0.93 &  0.79 &     0.59 &    0.44 &     0.38 &     0.92 &     0.78 &    0.79 &     0.66 \\
\textbf{ACE}   &  0.89 &     0.73 &  0.93 &     0.83 &  0.52 &     0.35 &       0.81 &     0.66 &    0.99 &     0.93 &  0.79 &     0.59 &    0.44 &     0.38 &     0.92 &     0.78 &    0.79 &     0.66 \\
\hline 
 \multicolumn{19}{c}{JULE: Silhouette score (Euclidean distance)} \\
 \hline 
Paired score     &  0.27 &     0.20 &  0.72 &     0.55 &  0.04 &     0.03 &       0.56 &     0.41 &    0.42 &     0.30 &  0.70 &     0.50 &    0.64 &     0.46 &     0.55 &     0.41 &    0.49 &     0.36 \\
\textbf{ACE} (with $\mathcal{Z}_{outlier}$) &  0.88 &     0.72 &  0.89 &     0.75 &  0.53 &     0.36 &       0.81 &     0.65 &    0.52 &     0.44 &  0.88 &     0.70 &    0.41 &     0.36 &     0.92 &     0.78 &    0.73 &     0.60 \\
\textbf{ACE}   &  0.88 &     0.72 &  0.89 &     0.75 &  0.42 &     0.28 &       0.81 &     0.65 &    0.98 &     0.90 &  0.88 &     0.70 &    0.41 &     0.36 &     0.92 &     0.78 &    0.77 &     0.64 \\
\hline 
 \multicolumn{19}{c}{DEPICT: Calinski-Harabasz index} \\
 \hline 
Paired score     &  0.76 &     0.57 &  0.44 &     0.26 &  0.76 &     0.57 &       0.89 &     0.72 &    0.49 &     0.44 &    &       &      &       &       &       &    0.67 &     0.51 \\
\textbf{ACE} (with $\mathcal{Z}_{outlier}$) &  0.91 &     0.77 &  0.56 &     0.44 &  0.94 &     0.82 &       0.96 &     0.87 &    0.96 &     0.87 &    &       &      &       &       &       &    0.87 &     0.75 \\
\textbf{ACE}   &  0.91 &     0.77 &  0.56 &     0.44 &  0.94 &     0.82 &       0.96 &     0.87 &    0.96 &     0.87 &    &       &      &       &       &       &    0.87 &     0.75 \\
\hline 
 \multicolumn{19}{c}{DEPICT: Davies-Bouldin index} \\
 \hline 
Paired score     &  0.81 &     0.59 &  0.45 &     0.31 &  0.90 &     0.74 &       0.89 &     0.72 &    0.63 &     0.59 &    &       &      &       &       &       &    0.73 &     0.59 \\
\textbf{ACE} (with $\mathcal{Z}_{outlier}$) &  0.91 &     0.82 &  0.76 &     0.58 &  0.91 &     0.79 &       0.96 &     0.87 &    0.98 &     0.92 &    &       &      &       &       &       &    0.90 &     0.80 \\
\textbf{ACE}   &  0.91 &     0.82 &  0.76 &     0.58 &  0.91 &     0.79 &       0.96 &     0.87 &    0.98 &     0.92 &    &       &      &       &       &       &    0.90 &     0.80 \\
\hline 
 \multicolumn{19}{c}{DEPICT: Silhouette score (Cosine distance)} \\
 \hline 
Paired score     &  0.81 &     0.62 &  0.45 &     0.33 &  0.90 &     0.75 &       0.89 &     0.72 &    0.77 &     0.58 &    &       &      &       &       &       &    0.76 &     0.60 \\
\textbf{ACE} (with $\mathcal{Z}_{outlier}$) &  0.97 &     0.90 &  0.71 &     0.56 &  0.94 &     0.82 &       0.97 &     0.90 &    0.94 &     0.83 &    &       &      &       &       &       &    0.91 &     0.80 \\
\textbf{ACE}   &  0.97 &     0.90 &  0.71 &     0.56 &  0.94 &     0.82 &       0.97 &     0.90 &    0.94 &     0.83 &    &       &      &       &       &       &    0.91 &     0.80 \\
\hline 
 \multicolumn{19}{c}{DEPICT: Silhouette score (Euclidean distance)} \\
 \hline 
Paired score     &  0.73 &     0.50 &  0.47 &     0.36 &  0.79 &     0.65 &       0.86 &     0.69 &    0.59 &     0.52 &    &       &      &       &       &       &    0.69 &     0.54 \\
\textbf{ACE} (with $\mathcal{Z}_{outlier}$) &  0.97 &     0.88 &  0.65 &     0.50 &  0.95 &     0.83 &       0.98 &     0.90 &    0.94 &     0.82 &    &       &      &       &       &       &    0.90 &     0.79 \\
\textbf{ACE}   &  0.97 &     0.88 &  0.65 &     0.50 &  0.95 &     0.83 &       0.98 &     0.90 &    0.94 &     0.82 &    &       &      &       &       &       &    0.90 &     0.79 \\
\bottomrule
\end{tabular}
}

\end{table*}
\begin{table*}[htbp!]
\centering
\caption{Perturbation analyses comparing inclusion and exclusion of outlier spaces for Application 1. $r_s$ and $\tau_B$ between the generated scores and ACC scores are reported. Empty cells and dash marks ($-$) indicate cases where the result is either missing or impractical to obtain.}
\resizebox{\textwidth}{!}{
\begin{tabular}{lllllllllllllllllll}
\toprule
{} & \multicolumn{2}{l}{USPS} & \multicolumn{2}{l}{YTF} & \multicolumn{2}{l}{FRGC} & \multicolumn{2}{l}{MNIST-test} & \multicolumn{2}{l}{CMU-PIE} & \multicolumn{2}{l}{UMist} & \multicolumn{2}{l}{COIL-20} & \multicolumn{2}{l}{COIL-100} & \multicolumn{2}{l}{Average} \\
{} & $r_s$ & $\tau_B$ & $r_s$ & $\tau_B$ & $r_s$ & $\tau_B$ &      $r_s$ & $\tau_B$ &   $r_s$ & $\tau_B$ & $r_s$ & $\tau_B$ &   $r_s$ & $\tau_B$ &    $r_s$ & $\tau_B$ &   $r_s$ & $\tau_B$ \\
\midrule
\hline 
 \multicolumn{19}{c}{JULE: Calinski-Harabasz index} \\
 \hline 
Paired score     &  0.04 &     0.05 &  0.39 &     0.27 & -0.26 &    -0.18 &       0.31 &     0.21 &   -0.20 &    -0.12 &  0.64 &     0.45 &    0.57 &     0.40 &     0.09 &     0.08 &    0.20 &     0.14 \\
\textbf{ACE} (with $\mathcal{Z}_{outlier}$) &  0.85 &     0.70 &  0.61 &     0.46 &  0.13 &     0.09 &       0.95 &     0.82 &    0.97 &     0.87 &  0.81 &     0.61 &    0.68 &     0.51 &     0.59 &     0.46 &    0.70 &     0.57 \\
\textbf{ACE}   &  0.90 &     0.77 &  0.73 &     0.54 &  0.49 &     0.36 &       0.95 &     0.82 &    0.97 &     0.87 &  0.81 &     0.61 &    0.57 &     0.40 &     0.93 &     0.81 &    0.79 &     0.65 \\
\hline 
 \multicolumn{19}{c}{JULE: Davies-Bouldin index} \\
 \hline 
Paired score     & -0.27 &    -0.15 & -0.14 &    -0.09 & -0.23 &    -0.14 &      -0.35 &    -0.19 &    0.20 &     0.16 &  0.53 &     0.36 &    0.63 &     0.44 &     0.33 &     0.26 &    0.09 &     0.08 \\
\textbf{ACE} (with $\mathcal{Z}_{outlier}$) & -0.28 &    -0.12 & -0.07 &    -0.07 &  0.53 &     0.38 &       0.79 &     0.64 &    0.78 &     0.62 &  0.27 &     0.20 &    0.84 &     0.64 &     0.44 &     0.28 &    0.41 &     0.32 \\
\textbf{ACE}   & -0.30 &    -0.09 & -0.07 &    -0.07 &  0.53 &     0.38 &       0.79 &     0.64 &    0.07 &     0.03 &  0.27 &     0.20 &    0.21 &     0.18 &     0.44 &     0.28 &    0.24 &     0.19 \\
\hline 
 \multicolumn{19}{c}{JULE: Silhouette score (Cosine distance)} \\
 \hline 
Paired score     &  0.17 &     0.14 &  0.59 &     0.41 &  0.07 &     0.06 &       0.47 &     0.33 &    0.45 &     0.33 &  0.64 &     0.46 &    0.70 &     0.51 &     0.64 &     0.45 &    0.47 &     0.34 \\
\textbf{ACE} (with $\mathcal{Z}_{outlier}$) &  0.96 &     0.85 &  0.74 &     0.55 &  0.82 &     0.65 &       0.92 &     0.78 &    0.98 &     0.92 &  0.78 &     0.58 &    0.41 &     0.32 &     0.84 &     0.68 &    0.81 &     0.67 \\
\textbf{ACE}   &  0.96 &     0.85 &  0.74 &     0.55 &  0.82 &     0.65 &       0.92 &     0.78 &    0.98 &     0.92 &  0.78 &     0.58 &    0.41 &     0.32 &     0.84 &     0.68 &    0.81 &     0.67 \\
\hline 
 \multicolumn{19}{c}{JULE: Silhouette score (Euclidean distance)} \\
 \hline 
Paired score     &  0.14 &     0.12 &  0.54 &     0.39 & -0.08 &    -0.02 &       0.41 &     0.27 &    0.36 &     0.27 &  0.64 &     0.46 &    0.67 &     0.48 &     0.44 &     0.31 &    0.39 &     0.28 \\
\textbf{ACE} (with $\mathcal{Z}_{outlier}$) &  0.93 &     0.78 &  0.63 &     0.48 &  0.82 &     0.63 &       0.92 &     0.78 &    0.54 &     0.47 &  0.86 &     0.68 &    0.39 &     0.30 &     0.84 &     0.68 &    0.74 &     0.60 \\
\textbf{ACE}   &  0.93 &     0.78 &  0.63 &     0.48 &  0.71 &     0.53 &       0.92 &     0.78 &    0.98 &     0.91 &  0.86 &     0.68 &    0.39 &     0.30 &     0.84 &     0.68 &    0.78 &     0.64 \\
\hline 
 \multicolumn{19}{c}{DEPICT: Calinski-Harabasz index} \\
 \hline 
Paired score     &  0.56 &     0.40 &  0.54 &     0.35 &  0.76 &     0.57 &       0.88 &     0.69 &    0.48 &     0.43 &    &       &      &       &       &       &    0.64 &     0.49 \\
\textbf{ACE} (with $\mathcal{Z}_{outlier}$) &  0.82 &     0.72 &  0.61 &     0.45 &  0.91 &     0.82 &       0.97 &     0.91 &    0.96 &     0.87 &    &       &      &       &       &       &    0.86 &     0.75 \\
\textbf{ACE}   &  0.82 &     0.72 &  0.61 &     0.45 &  0.91 &     0.82 &       0.97 &     0.91 &    0.96 &     0.87 &    &       &      &       &       &       &    0.86 &     0.75 \\
\hline 
 \multicolumn{19}{c}{DEPICT: Davies-Bouldin index} \\
 \hline 
Paired score     &  0.61 &     0.42 &  0.48 &     0.32 &  0.92 &     0.74 &       0.88 &     0.69 &    0.62 &     0.56 &    &       &      &       &       &       &    0.70 &     0.55 \\
\textbf{ACE} (with $\mathcal{Z}_{outlier}$) &  0.99 &     0.96 &  0.65 &     0.46 &  0.90 &     0.74 &       0.99 &     0.96 &    0.96 &     0.87 &    &       &      &       &       &       &    0.90 &     0.80 \\
\textbf{ACE}   &  0.99 &     0.96 &  0.65 &     0.46 &  0.90 &     0.74 &       0.99 &     0.96 &    0.96 &     0.87 &    &       &      &       &       &       &    0.90 &     0.80 \\
\hline 
 \multicolumn{19}{c}{DEPICT: Silhouette score (Cosine distance)} \\
 \hline 
Paired score     &  0.62 &     0.45 &  0.53 &     0.42 &  0.91 &     0.75 &       0.88 &     0.69 &    0.77 &     0.58 &    &       &      &       &       &       &    0.74 &     0.58 \\
\textbf{ACE} (with $\mathcal{Z}_{outlier}$) &  0.95 &     0.88 &  0.70 &     0.54 &  0.91 &     0.77 &       0.96 &     0.88 &    0.94 &     0.83 &    &       &      &       &       &       &    0.89 &     0.78 \\
\textbf{ACE}   &  0.95 &     0.88 &  0.70 &     0.54 &  0.91 &     0.77 &       0.96 &     0.88 &    0.94 &     0.83 &    &       &      &       &       &       &    0.89 &     0.78 \\
\hline 
 \multicolumn{19}{c}{DEPICT: Silhouette score (Euclidean distance)} \\
 \hline 
Paired score     &  0.52 &     0.33 &  0.57 &     0.45 &  0.80 &     0.62 &       0.85 &     0.65 &    0.59 &     0.48 &    &       &      &       &       &       &    0.67 &     0.51 \\
\textbf{ACE} (with $\mathcal{Z}_{outlier}$) &  0.95 &     0.87 &  0.63 &     0.49 &  0.91 &     0.78 &       0.97 &     0.91 &    0.95 &     0.84 &    &       &      &       &       &       &    0.88 &     0.78 \\
\textbf{ACE}   &  0.95 &     0.87 &  0.63 &     0.49 &  0.91 &     0.78 &       0.97 &     0.91 &    0.95 &     0.84 &    &       &      &       &       &       &    0.88 &     0.78 \\
\bottomrule
\end{tabular}
}

 \label{tab:abs:acc:ot0}
\end{table*}
%%%%%%%%%%%%%%%%%%%%%%%%%%%%%%%%%%%%%%%%%%%%%%%%%%%%%%%%%%%%%%%%%%%%%%%%%%%%%%%%%%%%%%%%%
\begin{table*}[htbp!]
\centering
\caption{Perturbation analyses comparing inclusion and exclusion of outlier spaces for Application 2. $r_s$ and $\tau_B$ between the generated scores and NMI scores are reported. Empty cells and dash marks ($-$) indicate cases where the result is either missing or impractical to obtain.}
\resizebox{\textwidth}{!}{
\begin{tabular}{lllllllllllllllllll}
\toprule
{} & \multicolumn{2}{l}{USPS (10)} & \multicolumn{2}{l}{YTF (41)} & \multicolumn{2}{l}{FRGC (20)} & \multicolumn{2}{l}{MNIST-test (10)} & \multicolumn{2}{l}{CMU-PIE (68)} & \multicolumn{2}{l}{UMist (20)} & \multicolumn{2}{l}{COIL-20 (20)} & \multicolumn{2}{l}{COIL-100 (100)} & \multicolumn{2}{l}{Average} \\
{} &      $r_s$ &   $\tau_B$ &      $r_s$ &   $\tau_B$ &       $r_s$ &    $\tau_B$ &           $r_s$ &  $\tau_B$ &        $r_s$ &    $\tau_B$ &      $r_s$ &   $\tau_B$ &        $r_s$ &   $\tau_B$ &          $r_s$ &   $\tau_B$ &   $r_s$ & $\tau_B$ \\
\midrule
\hline 
 \multicolumn{19}{c}{JULE: Calinski-Harabasz index} \\
 \hline 
Paired score     &  0.65 (10) &  0.64 (10) &   0.1 (50) &  0.06 (50) &  -0.93 (15) &  -0.83 (15) &       0.64 (10) &  0.6 (10) &   -0.03 (20) &  -0.02 (20) &  -0.13 (5) &  -0.07 (5) &    0.76 (15) &  0.71 (15) &      0.74 (80) &  0.56 (80) &    0.22 &     0.21 \\
\textbf{ACE} (with $\mathcal{Z}_{outlier}$) &  0.65 (10) &  0.64 (10) &  0.93 (50) &  0.83 (50) &  -0.93 (10) &  -0.83 (10) &       0.64 (10) &  0.6 (10) &    0.14 (20) &   0.16 (20) &  -0.14 (5) &  -0.11 (5) &    0.74 (15) &  0.64 (15) &      0.79 (80) &  0.69 (80) &    0.35 &     0.33 \\
\textbf{ACE}   &  0.65 (10) &  0.64 (10) &  0.93 (50) &  0.83 (50) &  -0.72 (15) &  -0.67 (15) &       0.64 (10) &  0.6 (10) &    0.88 (70) &   0.73 (70) &  -0.14 (5) &  -0.11 (5) &    0.74 (15) &  0.64 (15) &      0.79 (80) &  0.69 (80) &    0.47 &     0.42 \\
\hline 
 \multicolumn{19}{c}{JULE: Davies-Bouldin index} \\
 \hline 
Paired score     &  0.54 (15) &  0.38 (15) &  0.15 (50) &  0.17 (50) &  0.85 (45) &  0.67 (45) &       0.43 (10) &  0.29 (10) &   0.78 (100) &  0.56 (100) &  -0.08 (45) &   0.02 (45) &   -0.26 (40) &  -0.14 (40) &      -0.9 (20) &  -0.78 (20) &    0.19 &     0.15 \\
\textbf{ACE} (with $\mathcal{Z}_{outlier}$) &  0.98 (15) &  0.91 (15) &  0.83 (50) &  0.67 (50) &  0.87 (40) &  0.72 (40) &       0.79 (10) &   0.6 (10) &    0.85 (90) &   0.69 (90) &  -0.21 (45) &  -0.02 (45) &   -0.69 (50) &  -0.57 (50) &     -0.94 (20) &  -0.82 (20) &    0.31 &     0.27 \\
\textbf{ACE}   &  0.98 (15) &  0.91 (15) &  0.83 (50) &  0.67 (50) &  0.87 (40) &  0.72 (40) &       0.79 (10) &   0.6 (10) &    0.85 (90) &   0.69 (90) &  -0.21 (45) &  -0.02 (45) &   -0.69 (50) &  -0.57 (50) &     -0.94 (20) &  -0.82 (20) &    0.31 &     0.27 \\
\hline 
 \multicolumn{19}{c}{JULE: Silhouette score (Cosine distance)} \\
 \hline 
Paired score     &  0.99 (10) &  0.96 (10) &   0.3 (50) &  0.22 (50) &  0.72 (25) &  0.61 (25) &       0.87 (10) &  0.69 (10) &    0.98 (70) &  0.91 (70) &  -0.07 (45) &   0.07 (45) &    0.52 (25) &  0.36 (25) &     0.39 (200) &   0.2 (200) &    0.59 &     0.50 \\
\textbf{ACE} (with $\mathcal{Z}_{outlier}$) &  0.95 (10) &  0.87 (10) &  0.98 (50) &  0.94 (50) &   0.7 (45) &  0.61 (45) &       0.96 (10) &  0.87 (10) &    0.95 (90) &  0.87 (90) &  -0.07 (45) &  -0.02 (45) &    0.74 (20) &   0.5 (20) &     0.43 (160) &  0.29 (160) &    0.70 &     0.62 \\
\textbf{ACE}   &  0.95 (10) &  0.87 (10) &  0.98 (50) &  0.94 (50) &   0.7 (45) &  0.61 (45) &       0.96 (10) &  0.87 (10) &    0.98 (70) &  0.91 (70) &  -0.07 (45) &  -0.02 (45) &    0.74 (20) &   0.5 (20) &     0.46 (180) &  0.33 (180) &    0.71 &     0.63 \\
\hline 
 \multicolumn{19}{c}{JULE: Silhouette score (Euclidean distance)} \\
 \hline 
Paired score     &  0.85 (10) &  0.73 (10) &  0.33 (50) &  0.28 (50) &  0.72 (25) &  0.61 (25) &       0.88 (10) &  0.69 (10) &    0.96 (80) &  0.87 (80) &  0.07 (45) &  0.16 (45) &    0.55 (25) &  0.43 (25) &     0.44 (200) &  0.29 (200) &    0.60 &     0.51 \\
\textbf{ACE} (with $\mathcal{Z}_{outlier}$) &  0.95 (10) &  0.87 (10) &  0.98 (50) &  0.94 (50) &  0.78 (45) &  0.67 (45) &       0.95 (10) &  0.82 (10) &    0.95 (90) &  0.87 (90) &  0.14 (45) &  0.11 (45) &    0.71 (25) &  0.43 (25) &     0.47 (200) &  0.33 (200) &    0.74 &     0.63 \\
\textbf{ACE}   &  0.95 (10) &  0.87 (10) &  0.98 (50) &  0.94 (50) &  0.78 (45) &  0.67 (45) &       0.95 (10) &  0.82 (10) &    0.98 (70) &  0.91 (70) &  0.14 (45) &  0.11 (45) &    0.71 (25) &  0.43 (25) &     0.47 (200) &  0.33 (200) &    0.74 &     0.64 \\
\hline 
 \multicolumn{19}{c}{DEPICT: Calinski-Harabasz index} \\
 \hline 
Paired score     &  0.46 (5) &  0.6 (5) &  -0.99 (5) &  -0.96 (5) &  -0.85 (10) &  -0.72 (10) &        0.44 (5) &  0.56 (5) &   -0.92 (10) &  -0.82 (10) &         &       &           &       &             &       &   -0.37 &    -0.27 \\
\textbf{ACE} (with $\mathcal{Z}_{outlier}$) &  0.46 (5) &  0.6 (5) &  -0.66 (5) &  -0.51 (5) &   0.77 (30) &   0.61 (30) &        0.46 (5) &   0.6 (5) &    0.92 (80) &   0.82 (80) &         &       &           &       &             &       &    0.39 &     0.42 \\
\textbf{ACE}   &  0.46 (5) &  0.6 (5) &  -0.66 (5) &  -0.51 (5) &   0.77 (30) &   0.61 (30) &        0.46 (5) &   0.6 (5) &    0.92 (80) &   0.82 (80) &         &       &           &       &             &       &    0.39 &     0.42 \\
\hline 
 \multicolumn{19}{c}{DEPICT: Davies-Bouldin index} \\
 \hline 
Paired score     &   0.46 (5) &   0.6 (5) &  -0.78 (5) &  -0.64 (5) &  -0.85 (10) &  -0.72 (10) &        0.44 (5) &   0.56 (5) &    -0.1 (10) &  0.02 (10) &         &       &           &       &             &       &   -0.17 &    -0.04 \\
\textbf{ACE} (with $\mathcal{Z}_{outlier}$) &  0.62 (10) &  0.6 (10) &  0.95 (50) &  0.87 (50) &   0.77 (35) &   0.67 (35) &       0.78 (10) &  0.69 (10) &    0.96 (70) &  0.91 (70) &         &       &           &       &             &       &    0.82 &     0.75 \\
\textbf{ACE}   &  0.62 (10) &  0.6 (10) &  0.95 (50) &  0.87 (50) &   0.77 (35) &   0.67 (35) &       0.78 (10) &  0.69 (10) &    0.96 (70) &  0.91 (70) &         &       &           &       &             &       &    0.82 &     0.75 \\
\hline 
 \multicolumn{19}{c}{DEPICT: Silhouette score (Cosine distance)} \\
 \hline 
Paired score     &   0.44 (5) &   0.56 (5) &   -0.7 (5) &   -0.6 (5) &  -0.85 (10) &  -0.72 (10) &        0.44 (5) &   0.56 (5) &    0.07 (10) &  0.11 (10) &         &       &           &       &             &       &   -0.12 &    -0.02 \\
\textbf{ACE} (with $\mathcal{Z}_{outlier}$) &  0.65 (15) &  0.64 (15) &  0.87 (40) &  0.78 (40) &   0.93 (35) &   0.83 (35) &       0.85 (10) &  0.78 (10) &    0.99 (80) &  0.96 (80) &         &       &           &       &             &       &    0.86 &     0.80 \\
\textbf{ACE}   &  0.65 (15) &  0.64 (15) &  0.87 (40) &  0.78 (40) &   0.93 (35) &   0.83 (35) &       0.85 (10) &  0.78 (10) &    0.99 (80) &  0.96 (80) &         &       &           &       &             &       &    0.86 &     0.80 \\
\hline 
 \multicolumn{19}{c}{DEPICT: Silhouette score (Euclidean distance)} \\
 \hline 
Paired score     &  0.44 (5) &  0.56 (5) &  -0.61 (5) &  -0.47 (5) &  -0.85 (10) &  -0.72 (10) &        0.44 (5) &   0.56 (5) &   -0.12 (10) &  -0.02 (10) &         &       &           &       &             &       &   -0.14 &    -0.02 \\
\textbf{ACE} (with $\mathcal{Z}_{outlier}$) &  0.46 (5) &   0.6 (5) &  0.94 (40) &  0.87 (40) &   0.02 (25) &   0.06 (25) &       0.85 (10) &  0.78 (10) &    0.98 (80) &   0.91 (80) &         &       &           &       &             &       &    0.65 &     0.64 \\
\textbf{ACE}   &  0.46 (5) &   0.6 (5) &  0.94 (40) &  0.87 (40) &   0.02 (25) &   0.06 (25) &       0.85 (10) &  0.78 (10) &    0.98 (80) &   0.91 (80) &         &       &           &       &             &       &    0.65 &     0.64 \\
\bottomrule
\end{tabular}
}
 \label{tab:abs:nmi:ot1}
\end{table*}
\begin{table*}[htbp!]
\centering
\caption{Perturbation analyses comparing inclusion and exclusion of outlier spaces for Application 2. $r_s$ and $\tau_B$ between the generated scores and ACC scores are reported. Empty cells and dash marks ($-$) indicate cases where the result is either missing or impractical to obtain.}
\resizebox{\textwidth}{!}{
\begin{tabular}{lllllllllllllllllll}
\toprule
{} & \multicolumn{2}{l}{USPS (10)} & \multicolumn{2}{l}{YTF (41)} & \multicolumn{2}{l}{FRGC (20)} & \multicolumn{2}{l}{MNIST-test (10)} & \multicolumn{2}{l}{CMU-PIE (68)} & \multicolumn{2}{l}{UMist (20)} & \multicolumn{2}{l}{COIL-20 (20)} & \multicolumn{2}{l}{COIL-100 (100)} & \multicolumn{2}{l}{Average} \\
{} &     $r_s$ & $\tau_B$ &    $r_s$ & $\tau_B$ &     $r_s$ & $\tau_B$ &           $r_s$ & $\tau_B$ &        $r_s$ & $\tau_B$ &      $r_s$ & $\tau_B$ &        $r_s$ & $\tau_B$ &          $r_s$ & $\tau_B$ &   $r_s$ & $\tau_B$ \\
\midrule
\hline 
 \multicolumn{19}{c}{JULE: Calinski-Harabasz index} \\
 \hline 
Paired score     &      0.84 &     0.73 &     0.03 &    -0.06 &     -0.49 &    -0.31 &            0.61 &     0.56 &        -0.09 &    -0.07 &      -0.04 &     0.07 &         0.74 &     0.64 &           0.60 &     0.51 &    0.27 &     0.26 \\
\textbf{ACE} (with $\mathcal{Z}_{outlier}$) &      0.84 &     0.73 &     0.92 &     0.83 &     -0.59 &    -0.42 &            0.61 &     0.56 &         0.08 &     0.11 &      -0.07 &     0.02 &         0.76 &     0.71 &           0.65 &     0.56 &    0.40 &     0.39 \\
\textbf{ACE}   &      0.84 &     0.73 &     0.92 &     0.83 &     -0.11 &    -0.03 &            0.61 &     0.56 &         0.83 &     0.69 &      -0.07 &     0.02 &         0.76 &     0.71 &           0.65 &     0.56 &    0.55 &     0.51 \\
\hline 
 \multicolumn{19}{c}{JULE: Davies-Bouldin index} \\
 \hline 
Paired score     &      0.39 &     0.29 &     0.10 &     0.06 &      0.37 &     0.25 &            0.49 &     0.33 &         0.83 &     0.60 &      -0.28 &    -0.29 &        -0.29 &    -0.21 &          -0.87 &    -0.73 &    0.09 &     0.04 \\
\textbf{ACE} (with $\mathcal{Z}_{outlier}$) &      0.89 &     0.73 &     0.80 &     0.67 &      0.60 &     0.42 &            0.83 &     0.64 &         0.88 &     0.73 &      -0.42 &    -0.33 &        -0.71 &    -0.64 &          -0.82 &    -0.69 &    0.26 &     0.19 \\
\textbf{ACE}   &      0.89 &     0.73 &     0.80 &     0.67 &      0.60 &     0.42 &            0.83 &     0.64 &         0.88 &     0.73 &      -0.42 &    -0.33 &        -0.71 &    -0.64 &          -0.82 &    -0.69 &    0.26 &     0.19 \\
\hline 
 \multicolumn{19}{c}{JULE: Silhouette score (Cosine distance)} \\
 \hline 
Paired score     &      0.89 &     0.78 &     0.27 &     0.22 &      0.21 &     0.09 &            0.81 &     0.64 &         0.99 &     0.96 &      -0.26 &    -0.24 &         0.55 &     0.43 &           0.52 &     0.33 &    0.50 &     0.40 \\
\textbf{ACE} (with $\mathcal{Z}_{outlier}$) &      0.95 &     0.87 &     0.98 &     0.94 &      0.64 &     0.54 &            0.94 &     0.82 &         0.96 &     0.91 &      -0.32 &    -0.24 &         0.76 &     0.57 &           0.56 &     0.42 &    0.69 &     0.60 \\
\textbf{ACE}   &      0.95 &     0.87 &     0.98 &     0.94 &      0.64 &     0.54 &            0.94 &     0.82 &         0.99 &     0.96 &      -0.32 &    -0.24 &         0.76 &     0.57 &           0.60 &     0.47 &    0.69 &     0.61 \\
\hline 
 \multicolumn{19}{c}{JULE: Silhouette score (Euclidean distance)} \\
 \hline 
Paired score     &      0.93 &     0.82 &     0.30 &     0.28 &      0.21 &     0.09 &            0.82 &     0.64 &         0.98 &     0.91 &      -0.13 &    -0.16 &         0.52 &     0.36 &           0.55 &     0.42 &    0.52 &     0.42 \\
\textbf{ACE} (with $\mathcal{Z}_{outlier}$) &      0.95 &     0.87 &     0.98 &     0.94 &      0.57 &     0.48 &            0.92 &     0.78 &         0.96 &     0.91 &      -0.03 &    -0.11 &         0.74 &     0.50 &           0.59 &     0.47 &    0.71 &     0.60 \\
\textbf{ACE}   &      0.95 &     0.87 &     0.98 &     0.94 &      0.57 &     0.48 &            0.92 &     0.78 &         0.99 &     0.96 &      -0.03 &    -0.11 &         0.74 &     0.50 &           0.59 &     0.47 &    0.71 &     0.61 \\
\hline 
 \multicolumn{19}{c}{DEPICT: Calinski-Harabasz index} \\
 \hline 
Paired score     &      0.88 &     0.82 &    -0.96 &    -0.91 &     -0.37 &    -0.22 &            0.79 &     0.73 &        -0.92 &    -0.82 &         &       &           &       &             &       &   -0.11 &    -0.08 \\
\textbf{ACE} (with $\mathcal{Z}_{outlier}$) &      0.88 &     0.82 &    -0.67 &    -0.56 &      0.92 &     0.78 &            0.82 &     0.78 &         0.92 &     0.82 &         &       &           &       &             &       &    0.57 &     0.53 \\
\textbf{ACE}   &      0.88 &     0.82 &    -0.67 &    -0.56 &      0.92 &     0.78 &            0.82 &     0.78 &         0.92 &     0.82 &         &       &           &       &             &       &    0.57 &     0.53 \\
\hline 
 \multicolumn{19}{c}{DEPICT: Davies-Bouldin index} \\
 \hline 
Paired score     &      0.88 &     0.82 &    -0.77 &    -0.60 &     -0.37 &    -0.22 &            0.79 &     0.73 &        -0.10 &     0.02 &         &       &           &       &             &       &    0.09 &     0.15 \\
\textbf{ACE} (with $\mathcal{Z}_{outlier}$) &      0.93 &     0.82 &     0.96 &     0.91 &      0.92 &     0.83 &            0.93 &     0.87 &         0.96 &     0.91 &         &       &           &       &             &       &    0.94 &     0.87 \\
\textbf{ACE}   &      0.93 &     0.82 &     0.96 &     0.91 &      0.92 &     0.83 &            0.93 &     0.87 &         0.96 &     0.91 &         &       &           &       &             &       &    0.94 &     0.87 \\
\hline 
 \multicolumn{19}{c}{DEPICT: Silhouette score (Cosine distance)} \\
 \hline 
Paired score     &      0.87 &     0.78 &    -0.69 &    -0.56 &     -0.37 &    -0.22 &            0.79 &     0.73 &         0.07 &     0.11 &         &       &           &       &             &       &    0.14 &     0.17 \\
\textbf{ACE} (with $\mathcal{Z}_{outlier}$) &      0.95 &     0.87 &     0.92 &     0.82 &      0.80 &     0.67 &            0.95 &     0.87 &         0.99 &     0.96 &         &       &           &       &             &       &    0.92 &     0.84 \\
\textbf{ACE}   &      0.95 &     0.87 &     0.92 &     0.82 &      0.80 &     0.67 &            0.95 &     0.87 &         0.99 &     0.96 &         &       &           &       &             &       &    0.92 &     0.84 \\
\hline 
 \multicolumn{19}{c}{DEPICT: Silhouette score (Euclidean distance)} \\
 \hline 
Paired score     &      0.87 &     0.78 &    -0.64 &    -0.51 &     -0.37 &    -0.22 &            0.79 &     0.73 &        -0.12 &    -0.02 &         &       &           &       &             &       &    0.11 &     0.15 \\
\textbf{ACE} (with $\mathcal{Z}_{outlier}$) &      0.88 &     0.82 &     0.98 &     0.91 &      0.73 &     0.56 &            0.95 &     0.87 &         0.98 &     0.91 &         &       &           &       &             &       &    0.90 &     0.81 \\
\textbf{ACE}   &      0.88 &     0.82 &     0.98 &     0.91 &      0.73 &     0.56 &            0.95 &     0.87 &         0.98 &     0.91 &         &       &           &       &             &       &    0.90 &     0.81 \\
\bottomrule
\end{tabular}
}
 \label{tab:abs:acc:ot1}
\end{table*}
%%%%%%%%%%%%%%%%%%%%%
\newpage
\section{Additional Discussion} \label{app:discussion}

\subsection{Application of ACE on pretrained embeddings}
In our discussion, we noted that a potential extension of ACE is its use for identifying the pretrained model that yields the most reliable clustering results. This direction is motivated by the common practice of leveraging pretrained encoders to obtain low-dimensional embeddings prior to applying clustering techniques, especially with the recent advances in pretrained models. To explore this potential, we conducted an additional experiment using ACE on the pretrained embeddings of the test set from the widely used benchmark dataset, \textbf{CIFAR-10} \citep{krizhevsky2009learning}. Specifically, we collected embeddings from a set of widely adopted pretrained models, including \texttt{ResNet18}, \texttt{ResNet34}, \texttt{ResNet50}, \texttt{ResNet101}, \texttt{ResNet152} \citep{he2016deep}, \texttt{VGG11}, \texttt{VGG13}, \texttt{VGG16}, \texttt{VGG19} \citep{simonyan2014very}, \texttt{DenseNet121}, \texttt{DenseNet169}, \texttt{DenseNet201} \citep{huang2017densely}, \texttt{MobileNetV2} \citep{sandler2018mobilenetv2}, \texttt{MobileNetV3-Large} \citep{howard2019searching}, \texttt{EfficientNet-B0}, \texttt{EfficientNet-B1}, \texttt{EfficientNet-B2} \citep{tan2019efficientnet}, \texttt{GoogleNet} \citep{szegedy2015going}, \texttt{ShuffleNetV2-x1.0} \citep{ma2018shufflenet}, \texttt{ConvNeXt-Tiny} \citep{liu2022convnet}, \texttt{ViT} (ViT-B/16) \citep{dosovitskiy2020image}, and \texttt{CLIP} \citep{radford2021learning}. For CLIP embeddings, we use ViT-B-32 as the backbone and load the original CLIP weights released by OpenAI, pretrained on CLIP’s large-scale image–text dataset. All other embeddings are obtained from models pretrained on the ImageNet dataset \citep{deng2009imagenet}. For each embedding type, we apply the corresponding preprocessing pipeline to the input images to obtain the final embeddings. Following standard practice, we apply L2 normalization to each set of embeddings and then run \texttt{KMeans} clustering to obtain the corresponding clustering assignments.

Following the same evaluation protocol used in our simulation and real-data experiments, we assessed clustering results using three internal measures: the Silhouette score (with both Euclidean and Cosine distances), the Calinski-Harabasz index, and the Davies-Bouldin index. We then compared ACE against three baseline evaluation strategies: Raw score, Paired score, and Pooled score. To quantify the consistency between internal and external measures, specifically, Normalized Mutual Information (NMI) and clustering accuracy (ACC), we computed Spearman’s rank correlation coefficient ($r_s$) and Kendall’s rank coefficient ($\tau_B$). The results are presented in Figure \ref{fig:pretrained}.

As shown in Figure \ref{fig:pretrained}, ACE consistently outperforms the baseline methods across all internal metrics and correlation measures. Raw score performs the worst, exhibiting negative rank correlations in some cases.  Overall, the Paired score tends to underperform compared to both the Pooled and ACE scores, although it can slightly outperform the Pooled score when the Silhouette score is used as the internal measure. These results indicate that ACE is not limited to deep clustering scenarios. It can also be effectively applied to evaluate clustering results when using pretrained embeddings.

\begin{figure}[h!]
\centering
\subfigure[$r_s$: NMI]{
    \includegraphics[width = 0.45\linewidth]{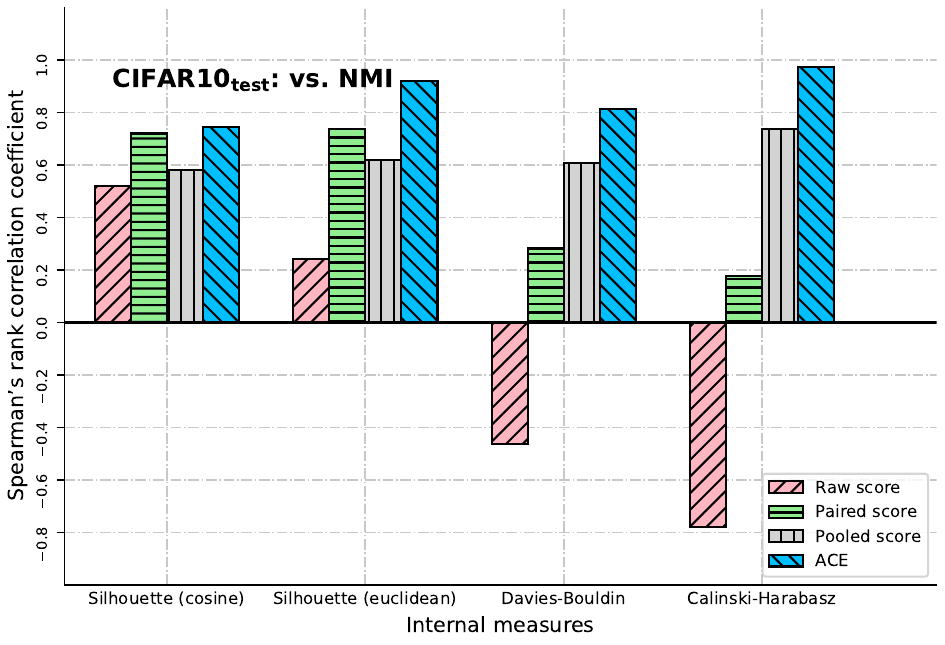}
}
\subfigure[$r_s$: ACC]{
    \includegraphics[width = 0.45\linewidth]{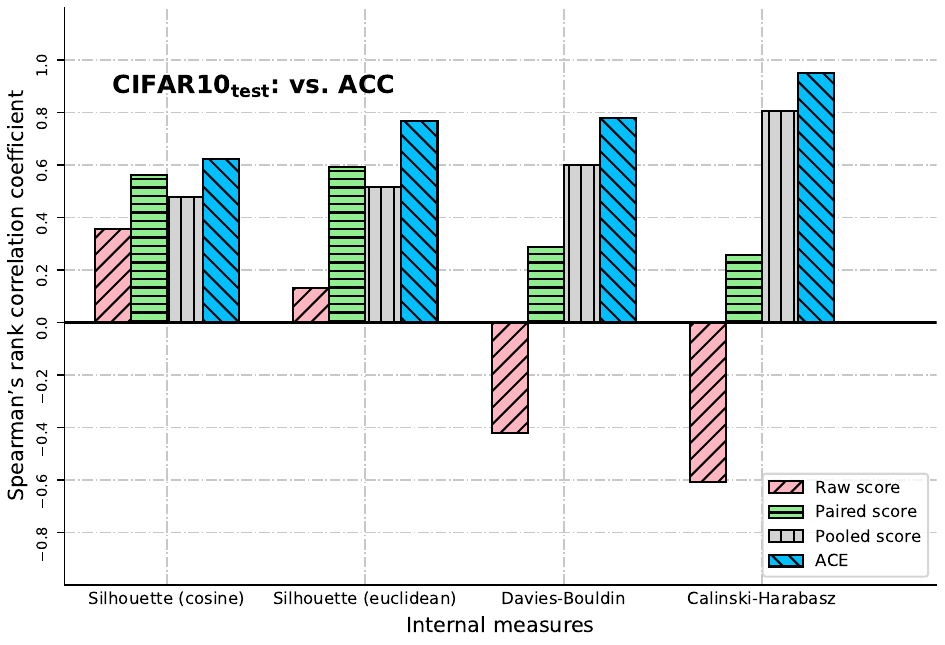}
}\\
\subfigure[$\tau_B$: NMI]{
    \includegraphics[width = 0.45\linewidth]{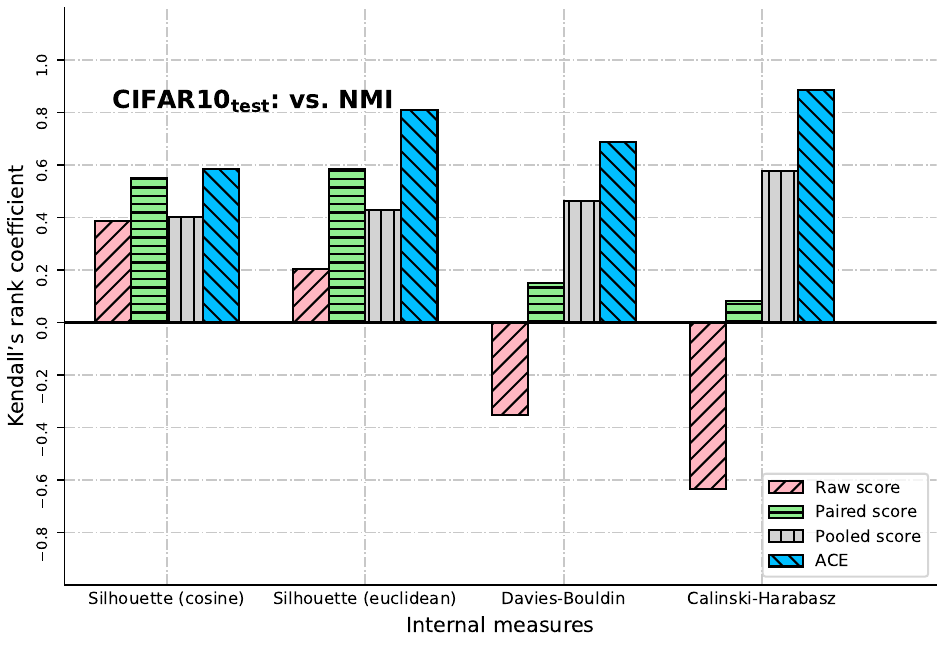}
}
\subfigure[$\tau_B$: ACC]{
    \includegraphics[width = 0.45\linewidth]{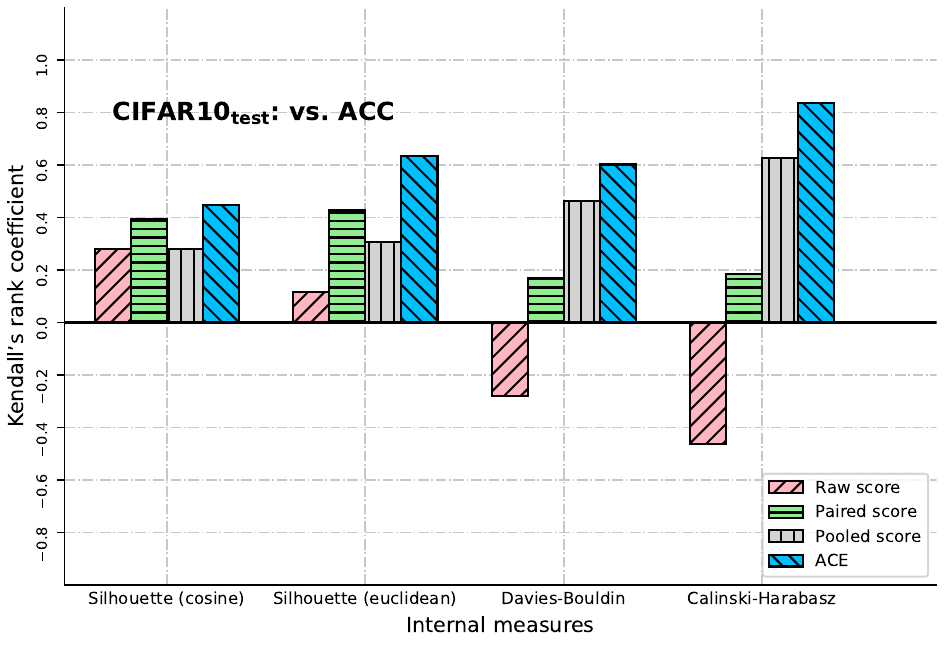}
}
\caption{Barplots illustrate rank consistency based on rank correlation coefficients with external measures for the application of ACE on pretrained embeddings.}
\label{fig:pretrained}
\end{figure}

\subsection{Connection to Uncertainty Quantification}
Although ACE does not directly produce uncertainty quantification for clustering results, it can be adapted for this purpose. In general, uncertainty for a clustering method can be quantified by generating $M$ clustering results through variations in factors that induce uncertainty in the base method (e.g., hyperparameters, initializations, data resampling or perturbation, or the number of clusters). Applying ACE to these $M$ results allows us to compute: 1) the proportion of spaces $\mathcal{Z}_m$ retained by ACE, 2) the weight assigned to each retained space and 3) the uncertainty metric of interest (e.g., the proportion of co-assignment of two points among the retained spaces). More retained spaces and more evenly distributed weights (e.g., smaller KL divergence from the discrete uniform distribution) indicate higher stability. We refer to this type of uncertainty as \emph{stability under perturbation} \citep{https://doi.org/10.1002/wics.1575,10.3150/13-BEJSP14}.

ACE can also be applied to quantify uncertainty for a single clustering result $(\mathcal{Z}, \rho)$. Candidate clusterings $\{(\mathcal{Z}_m, \rho_m)\}_{m=1}^M$ are first generated either using different clustering methods or by varying uncertainty-related factors of the same method. ACE then produces an ensemble-based evaluation criterion $\pi(\cdot \mid G_s^*)$ associated with a group of reliable embedding spaces $G_s^*$. Then the quantity $\sum_{m \in G_s^*} w_m\cdot(\pi(\rho\mid \mathcal{Z}_m)-\pi(\rho \mid G_s^*))^2$ can serve as an uncertainty measure for any clustering result $\rho$. 
We call this type of uncertainty as \emph{internal-measure-related uncertainty} as it leverages $\pi$, a measure of clustering quality (e.g., cluster compactness and separation), to assess the reliability of the clustering.

\newpage
\bibliographystyle{agsm}
\bibliography{reference}

\end{document}